\documentclass[11pt,a4paper]{article}
\usepackage[hyperref]{emnlp2018}
\usepackage{times}
\usepackage{latexsym}
\usepackage{amsmath}
\usepackage{amssymb}
\usepackage{graphicx}
\usepackage{url}
\usepackage{enumitem}
\usepackage{comment}
\usepackage{tabularx}
\usepackage{xcolor}
\usepackage{subcaption}

\aclfinalcopy 

\title{Learning Disentangled Representations of Texts \\ with Application to Biomedical Abstracts}

\author{Sarthak Jain \\
  Northeastern University \\
  {\tt \small jain.sar@husky.neu.edu} \\\And
  Edward Banner \\
  Northeastern University \\
  {\tt \small ebanner@ccs.neu.edu} \\\And
  Jan-Willem van de Meent \\
  Northeastern University \\
  {\tt \small j.vandemeent@northeastern.edu} \\\AND
  Iain J. Marshall \\
  King's College London \\
  {\tt \small iain.marshall@kcl.ac.uk} \\\And
  Byron C. Wallace \\
  Northeastern University \\
  {\tt \small b.wallace@northeastern.edu}
  }

\date{}

\begin{document}
\maketitle
\begin{abstract}

We propose a method for learning \emph{disentangled} representations of texts that code for distinct and complementary aspects, with the aim of affording efficient model transfer and interpretability. To induce disentangled embeddings, we propose an adversarial objective based on the (dis)similarity between triplets of documents with respect to specific aspects. Our motivating application is embedding biomedical abstracts describing clinical trials in a manner that disentangles the \emph{populations}, \emph{interventions}, and \emph{outcomes} in a given trial. We show that our method learns representations that encode these clinically salient aspects, and that these can be effectively used to perform aspect-specific retrieval. We demonstrate that the approach generalizes beyond our motivating application in experiments on two multi-aspect review corpora. 
\end{abstract}

\section{Introduction}
\vspace{-.25em}

A classic problem that arises in (distributed) representation learning is that it is difficult to determine what information individual dimensions in an embedding encode. When training a classifier to distinguish between images of people and landscapes, we do not know \emph{a priori} whether the model is sensitive to differences in color, contrast, shapes or textures. Analogously, in the case of natural language, when we calculate similarities between document embeddings of user reviews, we cannot know if this similarity primarily reflects user sentiment, the product discussed, or syntactic patterns. This lack of interpretability 
makes it difficult to assess whether a learned representations is likely to generalize to a new task or domain, hindering model transferability. Disentangled representations with known semantics could allow more efficient training in settings in which supervision is expensive to obtain (e.g., biomedical NLP).

Thus far in NLP, learned distributed representations have, with few exceptions \cite{ruder2016hierarchical,he-2017,zhang2017aspect}, been \emph{entangled}: they indiscriminately encode all aspects of texts. 
Rather than representing text via a monolithic vector, we propose to estimate multiple embeddings that capture complementary aspects of texts, drawing inspiration from the ML in vision community \cite{whitney2016disentangled,veit2017conditional}.

As a motivating example we consider documents that describe clinical trials. Such publications constitute the evidence drawn upon to support \emph{evidence-based medicine} (EBM), in which one formulates precise clinical questions with respect to the Populations, Interventions, Comparators and Outcomes (PICO elements) of interest \cite{sackett1996evidence}.\footnote{We collapse I and C because the distinction is arbitrary.} Ideally, learned representations of such articles would factorize into embeddings for the respective PICO elements. This would enable aspect-specific similarity measures, in turn facilitating retrieval of evidence concerning a given condition of interest (i.e., in a specific patient population), regardless of the interventions and outcomes considered. Better representations may reduce the amount of supervision needed, which is expensive in this domain.

Our work is one of the first efforts to induce disentangled representations of texts,\footnote{We review the few recent related works that do exist in Section \ref{section:related-work}.} which we believe may be broadly useful in NLP. Concretely, our contributions in this paper are as follows:

\vspace{-.4em}
\begin{itemize}[leftmargin=1.0em]
\item We formalize the problem of learning disentangled representations of texts, and develop a relatively general approach for learning these from aspect-specific similarity judgments expressed as triplets $(s, d, o)_a$, which indicate that document $d$ is more similar to document $s$ than to document $o$, with respect to aspect $a$.
\vspace{-.5em}
\item We perform extensive experiments that provide evidence that our approach yields disentangled representations of texts, both for our motivating task of learning PICO-specific embeddings of biomedical abstracts, and, more generally, for multi-aspect sentiment corpora.
\end{itemize}

\vspace{-.65em}
\section{Framework and Models}
\vspace{-.5em}
Recent approaches in computer vision have emphasized unsupervised learning of disentangled representations by incorporating information-theoretic regularizers into the objective \cite{chen2016infogan,higgins2017beta-vae}. These approaches do not require explicit manual annotations, but consequently they require post-hoc manual assignment of meaningful interpretations to learned representations. We believe it is more natural to use weak supervision to induce meaningful aspect embeddings.

\vspace{-.25em}
\subsection{Learning from Aspect Triplets}

As a general strategy for learning disentangled representations, we propose exploiting aspect-specific document triplets $(s,d,o)_a$: this signals that $s$ and $d$ are \emph{more} similar than are $d$ and $o$, with respect to aspect $a$ \cite{karaletsos2015bayesian,veit2016conditional}, i.e., $\text{sim}_a (d, s) > \text{sim}_a (d, o)$, where sim$_a$ quantifies similarity w.r.t.~aspect $a$. 

We associate with each aspect an encoder \emph{enc}$_a$ (encoders share low-level layer parameters; see Section \ref{section:encoder} for architecture details). This is used to obtain text embeddings $(\mathbf{e}^a_s, \mathbf{e}^a_d, \mathbf{e}^a_o)$. To estimate the parameters of these encoders we adopt a simple objective that seeks to maximize the similarity between $(\mathbf{e}^a_d, \mathbf{e}^a_s)$ and minimize similarity between $(\mathbf{e}^a_d, \mathbf{e}^a_o)$, via the following maximum margin loss
\vspace{-.2em}
\begin{equation}
\begin{aligned}
\mathcal{L}(\mathbf{e}^a_s, \mathbf{e}^a_d, \mathbf{e}^a_o) = \text{\it max}\{0,  1 &- \text{\it sim}(\mathbf{e}^a_d, \mathbf{e}^a_s) \\&+ \text{\it sim}(\mathbf{e}^a_d, \mathbf{e}^a_o)\}
\end{aligned}
\end{equation}
\noindent Where similarity between documents $i$ and $j$ with respect to a particular aspect $a$,  $\text{\it sim}_a(i, j)$, is simply the cosine similarity between the aspect embeddings $\mathbf{e}^a_i$ and $\mathbf{e}^a_j$. This allows for the same documents to be similar with respect to some aspects while dissimilar in terms of others.

The above setup depends on the correlation between aspects in the training data. At one extreme, when triplets enforce identical similarities for all aspects, the model cannot distinguish between aspects at all. At the other extreme, triplets are present for only one aspect $a$, and absent for all other aspects $a'$: 
In this case the model will use only the embeddings for aspect $a$ to represent similarities.
In general, we expect a compromise between these extremes, and propose using negative sampling to enable the model to learn targeted aspect-specific encodings.


\begin{figure}
	\centering
    \includegraphics[width=\columnwidth]{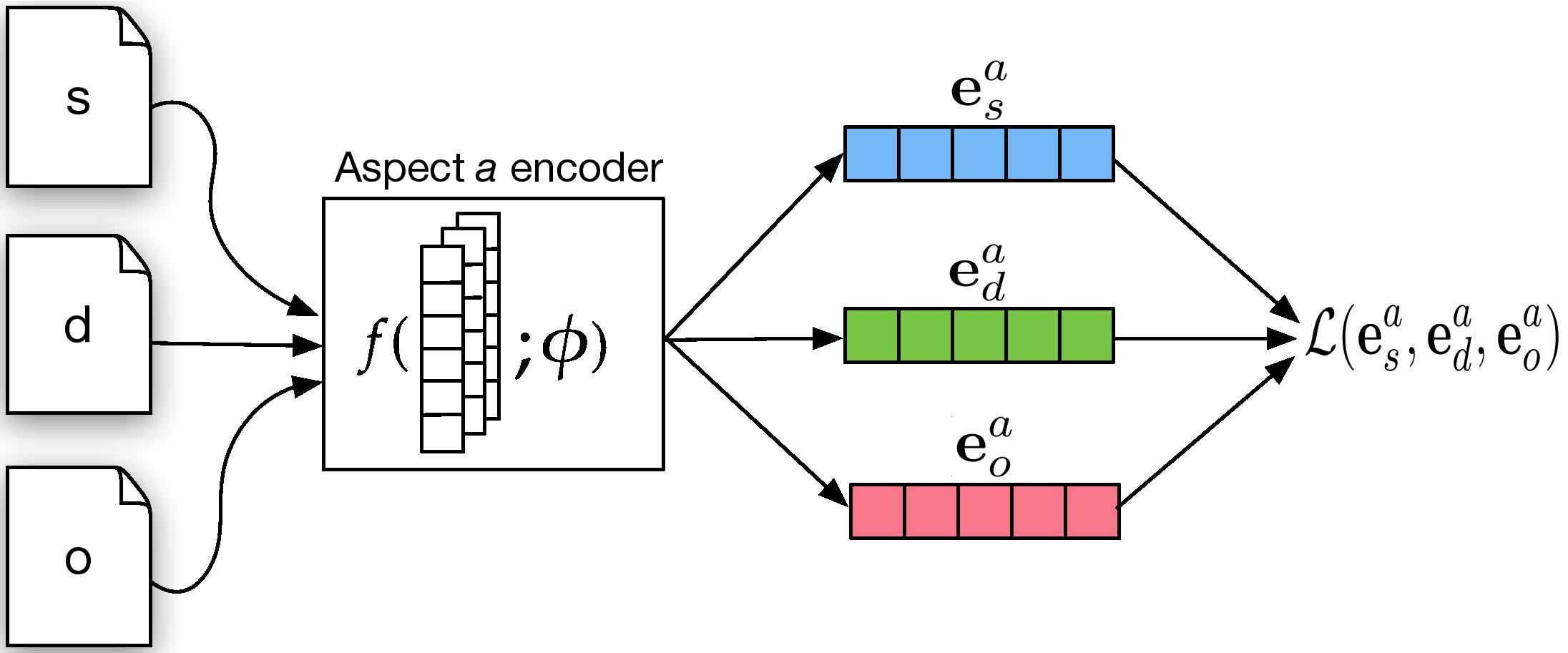}
    \vspace{-1.5em}
    \caption{We propose associating aspects with encoders (low-level parameters are shared across aspects; this is not shown) and training these with triplets codifying aspect-wise relative similarities.}
    \label{fig:the-idea}
    \vspace{-.75em}
\end{figure}

\vspace{-.25em}
\subsection{Encoder Architecture}
\label{section:encoder}
\vspace{-.25em}

Designing an aspect-based model requires specifying an encoder architecture. One consideration here is interpretability: a desirable property for aspect encoders is the ability to identify salient words for a given aspect. With this in mind, we propose using gated CNNs, which afford introspection via the token-wise gate activations. 

\begin{figure}
	\centering
    \includegraphics[width=\columnwidth]{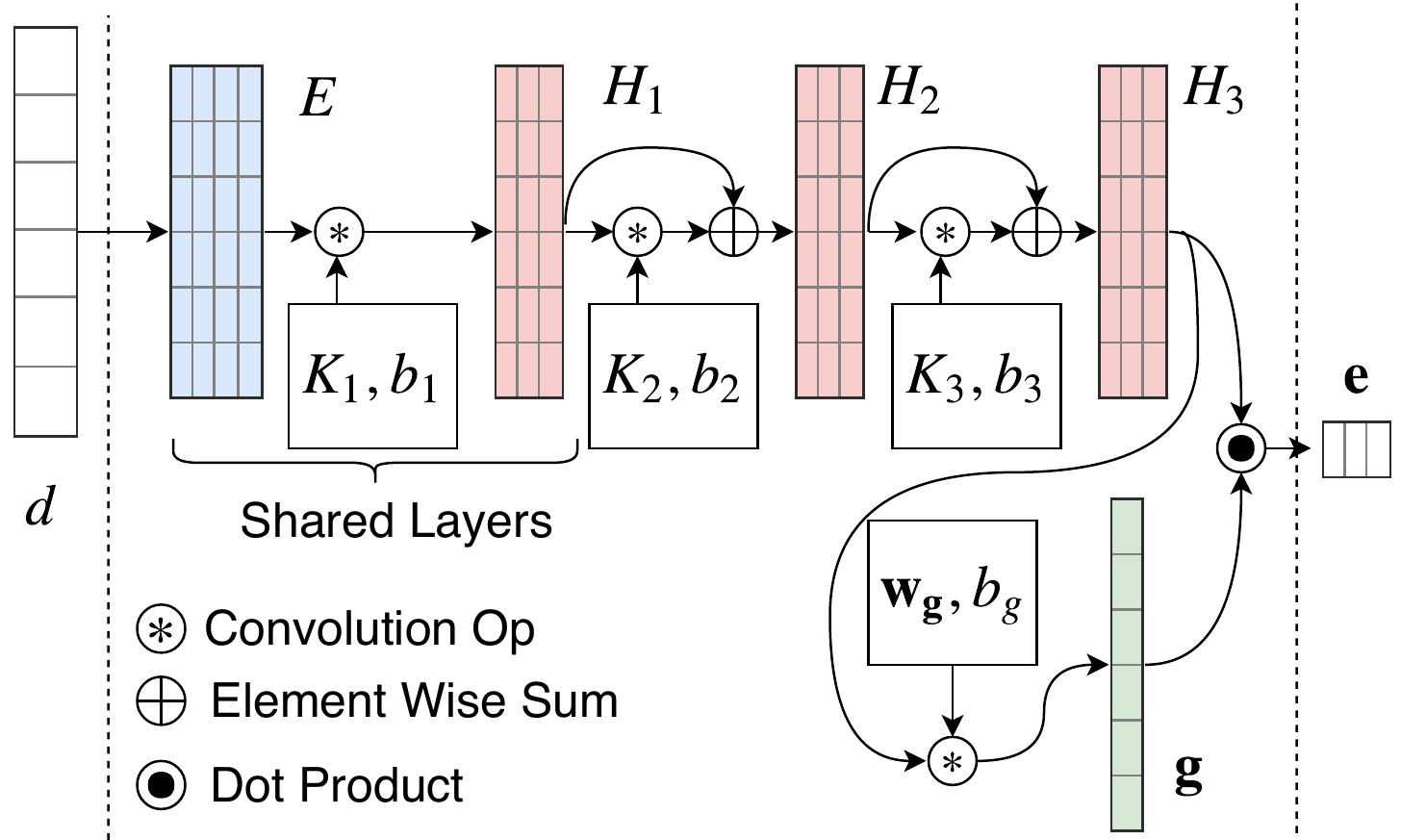}
     \vspace{-.75em}
    \caption{Schematic of our encoder architecture.}
    \label{fig:encoder}
    \vspace{-.65em}
\end{figure}

Figure \ref{fig:encoder} schematizes our encoder architecture. The input is a sequence of word indices $d = (w_{1}, ..., w_{N})$ which are mapped to $m$-dimensional word embeddings and stacked into a matrix $E = [{\mathbf e}_1, ..., {\mathbf e}_N]$. These are passed through sequential convolutional layers $C_1, ..., C_L$, which induce representations $H_l \in \mathbb{R}^{N \times k}$:

\vspace{-.5em}
\begin{equation}
\vspace{-.2em}
H_l = f_e(X * K_l + {\mathbf b_l})
\end{equation}

\noindent where $X \in \mathbb{R}^{N \times k}$ is the input to layer $C_l$ (either a set of $n$-gram embeddings or $H_{l-1}$) and $k$ is the number of feature maps. Kernel $K_l \in \mathbb{R}^{F \times k \times k}$ and ${\mathbf b_l} \in \mathbb{R}^{k}$ are parameters to be estimated, where $F$ is the size of kernel window.\footnote{The input to $C_1$ is $E \in \mathbb{R}^{N \times m}$, thus $K_1 \in \mathbb{R}^{F \times m \times k}$.} An activation function $f_e$ is applied element-wise to the output of the convolution operations. We fix the size of $H_{l-1} \in \mathbb{R}^{N \times k}$ by zero-padding where necessary. Keeping the size of feature maps constant across layers allows us to introduce residual connections; the output of layer $l$ is summed with the outputs of preceding layers before being passed forward. 

We multiply the output of the last convolutional layer $H_L \in \mathbb{R}^{N \times k}$ with gates ${\mathbf g} \in \mathbb{R}^{N \times 1}$ to yield our final embedding ${\mathbf e_d} \in \mathbb{R}^{1 \times k}$: 

\begin{equation}
\begin{aligned}
{\mathbf g} &= \sigma(H_L \cdot {\mathbf w}_g + b_g) \\
{\mathbf e_d} &= {\mathbf g}^T H_L
\end{aligned}
\label{eq:gate}
\end{equation}

\noindent where ${\mathbf w}_g \in \mathbb{R}^{k \times 1}$ and $b_g \in \mathbb{R}$ are learned parameters and $\sigma$ is the sigmoid activation function. We impose a sparsity-inducing constraint on $\mathbf{g}$ via the $\ell_1$ norm; this allows the gates to effectively serve as an attention mechanism over the input. Additionally, to capture potential cross-aspect correlation, weights in the embedding and first convolutional layers are shared between aspect encoders. 



\vspace{.2em}
\noindent {\bf Alternative encoders}. To assess the relative importance of the specific encoder model architecture used, we conduct experiments in which we fine-tune standard document representation models via triplet-based training. Specifically, we consider a single-layer MLP with BoW inputs, and a Neural Variational Document Model (NVDM) \cite{miao2016neural}. For the NVDM we take a weighted sum of the original loss function and the triplet-loss over the learned embeddings, where the weight is a model hyperparameter.



\section{Varieties of Supervision}
Our approach entails learning from triplets that codify relative similarity judgments with respect to specific aspects. We consider two approaches to acquiring such triplets: the first exploits aspect-specific summaries written for texts, and the second assumes a more general scenario in which we solicit aspect-wise triplet judgments directly.

\subsection{Deriving Triplets from Aspect Summaries}

In the case of our motivating example 
-- 
disentangled representations for articles describing clinical trials 
-- 
we have obtained aspect-specific summaries from the \emph{Cochrane Database of Systematic Reviews (CDSR)}. Cochrane is an international organization that creates and curates biomedical \emph{systematic reviews}. Briefly, such reviews seek to formally synthesize all relevant articles to answer precise clinical questions, i.e., questions that specify a particular PICO frame. The CDSR consists of a set of reviews $\{R_i\}$. Reviews include multiple articles (studies) $\{S_{ij}\}$. Each study $S$ consists of an abstract $A$ and a set of free text summaries $(s_P, s_I, s_O)$ written by reviewers describing the respective P, I and O elements in $S$.





Reviews implicitly specify PICO frames, and thus two studies in any given review may be viewed as equivalent with respect to their PICO aspects. We use this observation to derive document triplets. Recall that triplets for a given aspect include two comparatively similar texts ($s$, $d$) and one relatively dissimilar ($o$). Suppose the aspect of interest is the trial population. Here we match a given abstract ($d$) with its matched population summary from the CDSR ($s$); this encourages the encoder to yield similar embeddings for the abstract and the population description. The dissimilar $o$ is constructed to distinguish the given abstract from (1) other aspect encodings (of interventions, outcomes), and, (2) abstracts for trials with different populations.

Concretely, to construct a triplet  $(s, d, o)$ for the PICO data, we draw two reviews $R_1$ and $R_2$ from the CDSR at random, and sample two studies from the first ($s_1, s_1'$) and one from the second ($s_2$). Intuitively, $s_2$ will (very likely) comprise entirely different PICO elements than ($s_1, s_1'$), by virtue of belonging to a different review. To formalize the preceding description, our triplet is then: $(s = [s_1'^P], d = [s_1^{\text{abstract}}], o = [s_2^P|s_1'^I|s_1'^O])$, where $s_1^{\text{abstract}}$ is the abstract for study $s_1$, and aspect summaries for studies are denoted by superscripts. We include a concrete example of triplet construction in the Appendix, Section D.

\subsection{Learning Directly from Aspect-Wise Similarity Judgments}
The preceding setup assumes a somewhat unique case in which we have access to aspect-specific summaries written for texts. As a more general setting, we also consider learning directly from triplet-wise supervision concerning relative similarity with respect to particular aspects \cite{amid2015multiview,veit2017conditional,wilber2014cost}. The assumption is that such judgments can be solicited directly from annotators, and thus the approach may be applied to arbitrary domains, so long as meaningful aspects can be defined implicitly via pairwise similarities regarding them. 

We do not currently have corpora with such judgments in NLP, so we constructed two datasets using aspect-specific sentiment ratings. Note that this highlights the flexibility of exploiting aspect-wise triplet supervision as a means of learning disentangled representations: existing annotations can often be repurposed into such triplets. 



\section{Datasets and Experiments}
\label{section:experiments}

\begin{table*}
\small
\setlength{\tabcolsep}{3pt}
\centering
\begin{tabular}{l r r r r r|r r || r r r r}
Study &  TF-IDF &  Doc2Vec&  LDA &  NVDM &  ABAE&  RR-TF &  DSSM & $\mathbf{P}$ & $\mathbf{I}$ & $\mathbf{O}$ & $\mathbf{[P|I|O]}$ \\ 
\hline
ACEInhib.  & 0.81 & 0.74 & 0.72 & 0.85 & 0.81 & 0.67 & 0.85 & 0.83 & 0.88 & 0.84 &  \textbf{0.92} \\
ADHD  & 0.90 & 0.82 & 0.83 &  \textbf{0.93}  & 0.77 & 0.85 & 0.83 & 0.86 & 0.75 & 0.91 &  0.89 \\ 
Antihist. & 0.81 & 0.73 & 0.67 & 0.79 & 0.84 & 0.72 & \textbf{0.91} & 0.88 & 0.84 & 0.89 &  \textbf{0.91} \\  
Antipsych.  & 0.75 & 0.85 & 0.88 & 0.89 & 0.81 & 0.63 & 0.96 & 0.91 & 0.93 &  \textbf{0.97} &  \textbf{0.97} \\ 
BetaBlockers  & 0.67 & 0.65 & 0.61 & 0.76 & 0.70 & 0.56 & 0.68 & 0.71 & 0.75 & 0.77 &  \textbf{0.81} \\ 
CCBlockers  & 0.67 & 0.60 & 0.67 & 0.70 & 0.69 & 0.58 & 0.76 & 0.73 & 0.69 & 0.74 &  \textbf{0.77} \\ 
Estrogens  & 0.87 & 0.85 & 0.60 & 0.94 & 0.85 & 0.82 & 0.96 &  \textbf{1.00} & 0.98 & 0.83 &  \textbf{1.00} \\ 
NSAIDS  & 0.85 & 0.77 & 0.73 & 0.9 & 0.77 & 0.74 & 0.89 & 0.94 &  \textbf{0.95} & 0.8 &  \textbf{0.95} \\ 
Opioids  & 0.81 & 0.75 & 0.80 & 0.83 & 0.77 & 0.76 & 0.86 & 0.80 & 0.83 &  \textbf{0.92} &  \textbf{0.92} \\ 
OHG  & 0.79 & 0.80 & 0.70 & 0.89 & 0.90 & 0.72 & 0.90 & 0.90 & 0.95 & 0.95 &  \textbf{0.96} \\
PPI  & 0.81 & 0.79 & 0.74 & 0.85 & 0.82 & 0.68 & 0.94 & 0.94 & 0.87 & 0.87 &  \textbf{0.95} \\  
MuscleRelax.  & 0.60 & 0.67 & 0.74 & 0.75 & 0.61 & 0.57 & 0.77 & 0.68 & 0.62 &  \textbf{0.78}&  0.75 \\ 
Statins  & 0.79 & 0.76 & 0.66 & 0.87 & 0.77 & 0.68 & 0.87 & 0.82 &  \textbf{0.94} & 0.87 &  \textbf{0.94} \\  
Triptans  & 0.92 & 0.82 & 0.83 & 0.92 & 0.75 & 0.81 & \textbf{0.97}& 0.93 & 0.79 &  \textbf{0.97}  &  \textbf{0.97} \\  \hline
Mean & 0.79	& 0.76& 	0.73& 	0.85& 	0.78& 	0.70& 	0.87& 	0.85& 	0.84& 	0.87& 	\textbf{0.91}
\end{tabular}
\vspace{-.65em}
\caption{AUCs achieved using different representations on the Cohen et al. corpus. Models to the right of the $\vert$ are supervised; those to the right of $\vert \vert$ constitute the proposed disentangled embeddings. } 
\label{table:cohenauc}
\end{table*}

We present a series of experiments on three corpora to assess the degree to which the learned representations are disentangled, and to evaluate the utility of these embeddings in simple downstream retrieval tasks. We are particularly interested in the ability to identify documents similar w.r.t.~a target aspect. All parameter settings for baselines are reported in the Appendix (along with additional experimental results). The code is available at \\ {\small \url{https://github.com/successar/neural-nlp}}.

\subsection{PICO (EBM) Domain}
\label{section:PICO}
\vspace{-.25em}

We first evaluate embeddings quantitatively with respect to retrieval performance. In particular, we assess whether the induced representations afford improved retrieval of abstracts relevant to a particular systematic review \cite{cohen2006reducing,wallace2010semi}. We then perform two evaluations that explicitly assess the degree of disentanglement realized by the learned embeddings.

The PICO dataset comprises 41K abstracts of articles describing clinical trials extracted from the CDSR. Each abstract is associated with a review and three summaries, one per aspect (P/I/O). We keep all words that occur in $\geq$ 5 documents, converting all others to $\texttt{unk}$. We truncate documents to a fixed length (set to the 95th percentile).



\vspace{-.1em}
\subsubsection{Quantitative Evaluation}
\vspace{-.25em}

\noindent \textbf{Baselines}. We compare the proposed {\bf P}, {\bf I} and {\bf O} embeddings and their concatenation $\mathbf{[P|I|O]}$ to the following. {\bf TF-IDF}: standard TF-IDF representation of abstracts. {\bf RR-TF}: concatenated TF-IDF vectors of sentences predicted to describe the respective PICO elements, i.e., sentence predictions made using the pre-trained model from \cite{wallace2016extracting} --- this model was trained using distant supervision derived from the CDSR. {\bf doc2vec}: standard (entangled) distributed representations of abstracts \cite{le2014distributed}. {\bf LDA}: Latent Dirichlet Allocation. {\bf NVDM}: A generative model of text where the representation is a vector of log-frequencies that encode a topic \cite{miao2016neural}. {\bf ABAE}: An autoencoder model that discovers latent aspects in sentences \cite{he-2017}. We obtain document embeddings by summing over constituent sentence embeddings. {\bf DSSM}: A CNN based encoder trained with triplet loss over abstracts \cite{shen2014latent}.

\vspace{.15em}
\noindent \textbf{Hyperparameters and Settings}. We use three layers for our CNN-based encoder (with 200 filters in each layer; window size of 5) and the PReLU activation function \cite{he2015delving} as $f_e$. We use 200d word embeddings, initialized via pretraining over a corpus of PubMed abstracts \cite{moen2013distributional}. We used the Adam optimization function with default parameters \cite{kingma2014adam}. We imposed $\ell_2$ regularization over all parameters, the value of which was selected from the range ($1e$-$2$, $1e$-$6$) as $1e$-$5$. The $\ell_1$ regularization parameter for gates was chosen from the range ($1e$-$2$, $1e$-$8$) as $1e$-$6$. All model hyperparameters for our models and baselines were chosen via line search over a 10\% validation set.

\vspace{.15em}
\noindent \textbf{Metric}. For this evaluation, we used a held out set of 15 systematic reviews (comprising 2,223 studies) compiled by \newcite{cohen2006reducing}. The idea is that good representations should map abstracts in the same review (which describe studies with the same PICO frame) relatively near to one another. To compute AUCs over reviews, we first calculate all pairwise study similarities (i.e., over all studies in the Cohen corpus). We can then construct an ROC for a given abstract $a$ from a particular review to calculate its AUC: this measures the probability that a study drawn from the same review will be nearer to $a$ than a study from a different review. A summary AUC for a review is taken as the mean of the study AUCs in that review.



\vspace{.15em}
\noindent \textbf{Results}. Table \ref{table:cohenauc} reports the mean AUCs over individual reviews in the \newcite{cohen2006reducing} corpus, and grand means over these (bottom row). 
In brief: The proposed PICO embeddings (concatenated) obtain an equivalent or higher AUC than baseline strategies on 12/14 reviews, and strictly higher AUCs in 11/14. It is unsurprising that we outperform unsupervised approaches, but we also best RR-TF, which was trained with the same CDSR corpus \cite{wallace2016extracting}, and DSSM \cite{shen2014latent}, which exploits the same triplet loss as our model. We outperform the latter by an average performance gain of 4 points AUC (significant at 95\% level using independent 2-sample t-test).

\begin{table}
\small
    \centering
    \begin{tabularx}{\columnwidth}{X}
         {\setlength{\fboxsep}{0pt}\colorbox{red!16!white}{\strut in}}
{\setlength{\fboxsep}{0pt}\colorbox{red!22!white}{\strut a}}
{\setlength{\fboxsep}{0pt}\colorbox{red!24!white}{\strut clinical}}
{\setlength{\fboxsep}{0pt}\colorbox{red!13!white}{\strut trial}}
{\setlength{\fboxsep}{0pt}\colorbox{red!12!white}{\strut mainly}}
{\setlength{\fboxsep}{0pt}\colorbox{red!18!white}{\strut involving}}
{\setlength{\fboxsep}{0pt}\colorbox{red!23!white}{\strut patients}}
{\setlength{\fboxsep}{0pt}\colorbox{red!49!white}{\strut over}}
{\setlength{\fboxsep}{0pt}\colorbox{red!80!white}{\strut qqq}}
{\setlength{\fboxsep}{0pt}\colorbox{red!100!white}{\strut with}}
{\setlength{\fboxsep}{0pt}\colorbox{red!77!white}{\strut coronary}}
{\setlength{\fboxsep}{0pt}\colorbox{red!74!white}{\strut heart}}
{\setlength{\fboxsep}{0pt}\colorbox{red!82!white}{\strut disease}}
{\setlength{\fboxsep}{0pt}\colorbox{red!82!white}{\strut ,}}
{\setlength{\fboxsep}{0pt}\colorbox{red!47!white}{\strut ramipril}}
{\setlength{\fboxsep}{0pt}\colorbox{red!9!white}{\strut reduced}}
{\setlength{\fboxsep}{0pt}\colorbox{red!0!white}{\strut mortality}}
{\setlength{\fboxsep}{0pt}\colorbox{red!0!white}{\strut while}}
{\setlength{\fboxsep}{0pt}\colorbox{red!0!white}{\strut vitamin}}
{\setlength{\fboxsep}{0pt}\colorbox{red!0!white}{\strut e}}
{\setlength{\fboxsep}{0pt}\colorbox{red!0!white}{\strut had}}
{\setlength{\fboxsep}{0pt}\colorbox{red!10!white}{\strut no}}
{\setlength{\fboxsep}{0pt}\colorbox{red!10!white}{\strut preventive}}
{\setlength{\fboxsep}{0pt}\colorbox{red!9!white}{\strut effect}}
{\setlength{\fboxsep}{0pt}\colorbox{red!0!white}{\strut .}} \\[2pt] \hline
{\setlength{\fboxsep}{0pt}\colorbox{green!11!white}{\strut in}}
{\setlength{\fboxsep}{0pt}\colorbox{green!15!white}{\strut a}}
{\setlength{\fboxsep}{0pt}\colorbox{green!16!white}{\strut clinical}}
{\setlength{\fboxsep}{0pt}\colorbox{green!4!white}{\strut trial}}
{\setlength{\fboxsep}{0pt}\colorbox{green!1!white}{\strut mainly}}
{\setlength{\fboxsep}{0pt}\colorbox{green!1!white}{\strut involving}}
{\setlength{\fboxsep}{0pt}\colorbox{green!3!white}{\strut patients}}
{\setlength{\fboxsep}{0pt}\colorbox{green!5!white}{\strut over}}
{\setlength{\fboxsep}{0pt}\colorbox{green!8!white}{\strut qqq}}
{\setlength{\fboxsep}{0pt}\colorbox{green!6!white}{\strut with}}
{\setlength{\fboxsep}{0pt}\colorbox{green!3!white}{\strut coronary}}
{\setlength{\fboxsep}{0pt}\colorbox{green!1!white}{\strut heart}}
{\setlength{\fboxsep}{0pt}\colorbox{green!46!white}{\strut disease}}
{\setlength{\fboxsep}{0pt}\colorbox{green!74!white}{\strut ,}}
{\setlength{\fboxsep}{0pt}\colorbox{green!100!white}{\strut ramipril}}
{\setlength{\fboxsep}{0pt}\colorbox{green!54!white}{\strut reduced}}
{\setlength{\fboxsep}{0pt}\colorbox{green!26!white}{\strut mortality}}
{\setlength{\fboxsep}{0pt}\colorbox{green!0!white}{\strut while}}
{\setlength{\fboxsep}{0pt}\colorbox{green!36!white}{\strut vitamin}}
{\setlength{\fboxsep}{0pt}\colorbox{green!68!white}{\strut e}}
{\setlength{\fboxsep}{0pt}\colorbox{green!85!white}{\strut had}}
{\setlength{\fboxsep}{0pt}\colorbox{green!65!white}{\strut no}}
{\setlength{\fboxsep}{0pt}\colorbox{green!52!white}{\strut preventive}}
{\setlength{\fboxsep}{0pt}\colorbox{green!44!white}{\strut effect}}
{\setlength{\fboxsep}{0pt}\colorbox{green!27!white}{\strut .}} \\[2pt] \hline
{\setlength{\fboxsep}{0pt}\colorbox{cyan!6!white}{\strut in}}
{\setlength{\fboxsep}{0pt}\colorbox{cyan!10!white}{\strut a}}
{\setlength{\fboxsep}{0pt}\colorbox{cyan!11!white}{\strut clinical}}
{\setlength{\fboxsep}{0pt}\colorbox{cyan!4!white}{\strut trial}}
{\setlength{\fboxsep}{0pt}\colorbox{cyan!0!white}{\strut mainly}}
{\setlength{\fboxsep}{0pt}\colorbox{cyan!0!white}{\strut involving}}
{\setlength{\fboxsep}{0pt}\colorbox{cyan!0!white}{\strut patients}}
{\setlength{\fboxsep}{0pt}\colorbox{cyan!5!white}{\strut over}}
{\setlength{\fboxsep}{0pt}\colorbox{cyan!7!white}{\strut qqq}}
{\setlength{\fboxsep}{0pt}\colorbox{cyan!11!white}{\strut with}}
{\setlength{\fboxsep}{0pt}\colorbox{cyan!22!white}{\strut coronary}}
{\setlength{\fboxsep}{0pt}\colorbox{cyan!41!white}{\strut heart}}
{\setlength{\fboxsep}{0pt}\colorbox{cyan!40!white}{\strut disease}}
{\setlength{\fboxsep}{0pt}\colorbox{cyan!49!white}{\strut ,}}
{\setlength{\fboxsep}{0pt}\colorbox{cyan!57!white}{\strut ramipril}}
{\setlength{\fboxsep}{0pt}\colorbox{cyan!99!white}{\strut reduced}}
{\setlength{\fboxsep}{0pt}\colorbox{cyan!100!white}{\strut mortality}}
{\setlength{\fboxsep}{0pt}\colorbox{cyan!92!white}{\strut while}}
{\setlength{\fboxsep}{0pt}\colorbox{cyan!54!white}{\strut vitamin}}
{\setlength{\fboxsep}{0pt}\colorbox{cyan!29!white}{\strut e}}
{\setlength{\fboxsep}{0pt}\colorbox{cyan!7!white}{\strut had}}
{\setlength{\fboxsep}{0pt}\colorbox{cyan!1!white}{\strut no}}
{\setlength{\fboxsep}{0pt}\colorbox{cyan!13!white}{\strut preventive}}
{\setlength{\fboxsep}{0pt}\colorbox{cyan!22!white}{\strut effect}}
{\setlength{\fboxsep}{0pt}\colorbox{cyan!21!white}{\strut .}}
    \end{tabularx}
    \caption{Gate activations for each aspect in a PICO abstract. Note that because gates are calculated at the final convolution layer, activations are not in exact 1-1 correspondence with words.}
    \label{tab:gate-pico}
\end{table}

We now turn to the more important questions: are the learned representations actually disentangled, and do they encode the target aspects? Table \ref{tab:gate-pico} shows aspect-wise gate activations for PICO elements over a single abstract; this qualitatively suggests disentanglement, but we next investigate this in greater detail.


\vspace{-.1em}
\subsubsection{Qualitative Evaluation} 
\vspace{-.1em}
To assess the degree to which our PICO embeddings are disentangled -- i.e., capture complementary information relevant to the targeted aspects -- we performed two qualitative studies.

First, we assembled 87 articles (not seen in training) describing clinical trials from a review on the effectiveness of decision aids \cite{o2009decision} for: women with, at risk for, and genetically at risk for, breast cancer (BCt, BCs and BCg, respectively); type II diabetes (D); menopausal women (MW); pregnant women generally (PW) and those who have undergone a C-section previously (PWc); people at risk for colon cancer (CC); men with and at risk of prostate cancer (PCt and PCs, respectively) and individuals with atrial fibrillation (AF). This review is unusual in that it \emph{studies a single intervention (decision aids) across different populations}. Thus, if the model is successful in learning disentangled representations, the corresponding P vectors should roughly cluster, while the I/C should not. 

\begin{figure*}
\includegraphics[width=\linewidth]{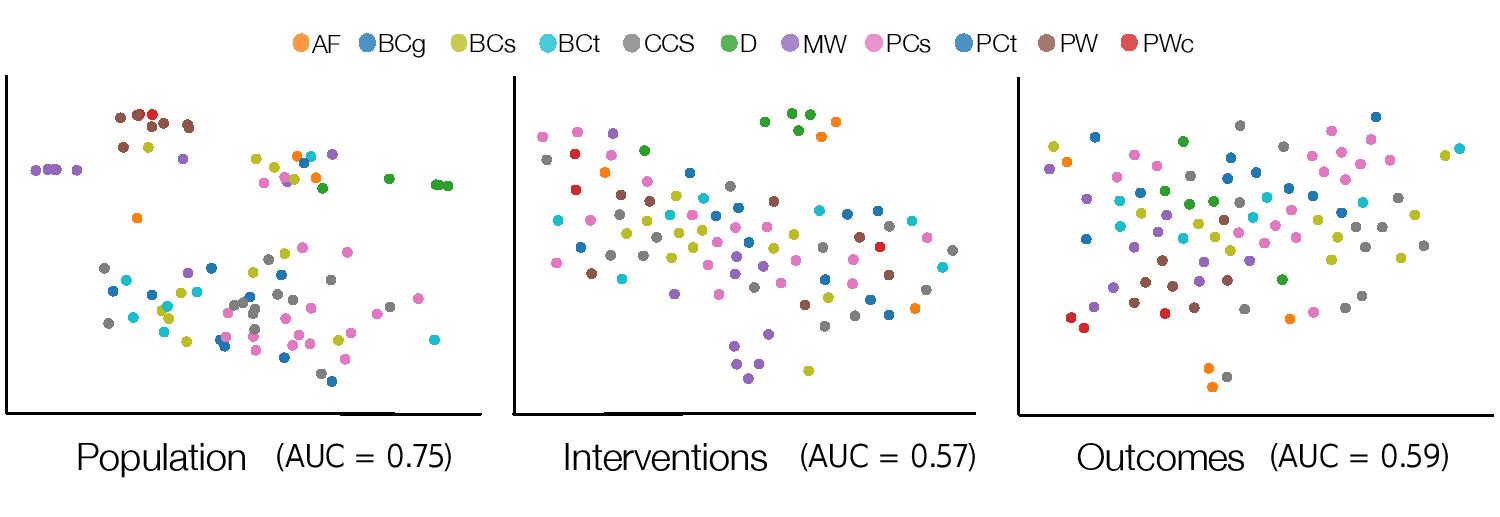}
\vspace{-2em}
\caption{TSNE-reduced scatter of disentangled PICO embeddings of abstracts involving ``decision aid" interventions. Abstracts are colored by known population group (see legend). Population embeddings for studies in the same group co-localize, more so than in the intervention and outcome space.}
\label{figure:DATSNE}
\vspace{-.75em}
\end{figure*}

Figure \ref{figure:DATSNE} shows a TSNE-reduced plot of the P, I/C and O embeddings induced by our model for these studies. Abstracts are color-coded to indicate the populations enumerated above. As hypothesized, P embeddings realize the clearest separation with respect to the populations, while the I and O embeddings of studies do not co-localize to the same degree. This is reflected quantitatively in the AUC values achieved using each aspect embedding (listed on the Figure). This result implies disentanglement along the desired axes.

Next we assembled 50 abstracts describing trials involving \emph{hip replacement arthroplasty} (HipRepl). We selected this topic because HipRepl will either describe the trial population (i.e., patients who have received hip replacements) or it will be the intervention, but not both. Thus, we would expect that abstracts describing trials in which HipRepl describes the population cluster in the corresponding embedding space, but not in the intervention space (and vice-versa). To test this, we first manually annotated the 50 abstracts, associating HipRepl with either P or I. We used these labels to calculate pairwise AUCs, reported in Table \ref{table:pubmed-AUCS}. The results imply that the population embeddings discriminate between studies that enrolled patients with HipRepl and other studies. Likewise, studies in which HipRepl was the intervention are grouped in the interventions embedding space, but not in the populations space. 
\begin{table}
\small
\begin{tabularx}{\columnwidth}{c|X X X}
& HipRepl I & HipRepl P & Mean \\ \hline
Population & 0.62 & \textbf{0.68} & \textbf{0.66} \\ 
Intervention & \textbf{0.91} & 0.46 & 0.57 \\ 
Outcome & 0.89 & 0.42 & 0.54 \\ 
\end{tabularx}
\caption{AUCs realized over HipRepl studies using different embeddings. Column: Study label (HipRepl as P or I). Row: Aspect embedding used.}
\label{table:pubmed-AUCS}
\end{table}

\begin{table}
\small
\centering
\begin{tabularx}{\columnwidth}{X}
{\bf Population} \emph{american}, \emph{area}, \emph{breast}, \emph{colorectal}, \emph{diagnosis}, \emph{inpatients}, \emph{outpatients}, \emph{stage}, \emph{their}, \emph{uk} \\ \hline
{\bf Intervention} \emph{adjunct}, \emph{alone}, \emph{an}, \emph{discussion}, \emph{intervention}, \emph{methods}, \emph{reduced}, \emph{started}, \emph{took}, \emph{written} \\ \hline
{\bf Outcome} \emph{adults}, \emph{area}, \emph{either}, \emph{eligible}, \emph{importance}, \emph{improve}, \emph{mortality}, \emph{pre}, \emph{reduces}, \emph{survival} \\ 
\end{tabularx}
\caption{Top ten most activated words, as determined by the gating mechanism.}
\label{table:picomost}
\vspace{-1em}
\end{table}

\vspace{.15em}
\noindent \textbf{Aspect words}. In Table \ref{table:picomost}, we report the most activated unigrams for each aspect embedding on the decision aids corpus. To derive these we use the outputs of the gating mechanism (Eq. \ref{eq:gate}), which is applied to all words in the input text. For each word, we average the activations across all abstracts and find the top ten words for each aspect. The words align nicely with the PICO aspects, providing further evidence that our model learns to focus on aspect-specific information. 

\subsection{Multi-Aspect Reviews}

We now turn from the specialized domain of biomedical abstracts to more general applications. In particular, we consider learning disentangled representations of beer, hotel and restaurant reviews. Learned embeddings should capture different aspects, e.g., taste or look in the case of beer.

\subsubsection{Beer Reviews (BeerAdvocate)}
\begin{table}
\setlength{\tabcolsep}{2pt}
\footnotesize
    \centering
    \begin{tabularx}{\columnwidth}{ l X X X X }
     Baseline & Look & Aroma & Palate & Taste  \\
    \hline
    TF-IDF & 0.63 & 0.62 & 0.62 & 0.61  \\
    LDA & 0.73 & 0.73 & 0.73 & 0.73 \\
    Doc2Vec & 0.61 & 0.61 & 0.61 & 0.61  \\
    NVDM & 0.68 & 0.69 & 0.69 & 0.70 \\
    ABAE & 0.50 & 0.50 & 0.50 & 0.50 \\
    \hline
    BoW + Triplet & 0.85 & 0.90 & 0.90 & 0.92 \\
    NVDM + Triplet & 0.90 & 0.91 & 0.92 & 0.95 \\
    DSSM + Triplet &  0.87 & 0.90 & 0.90 & 0.92 \\
    CNN + Triplet & \textbf{0.92} & \textbf{0.93} & \textbf{0.94} & \textbf{0.96} \\
    \end{tabularx}
    \caption{AUC results for different representations on the BeerAdvocate data. Models beneath the second line are supervised.}
    \label{table:beerauc}
\end{table}

\begin{table}
\vspace{-.25em}
\footnotesize
    \centering
    \begin{tabularx}{\columnwidth}{ l X X X X }
     & Look & Aroma & Palate & Taste \\ \hline
    Look & 0.92 & 0.89 & 0.88 & 0.87 \\ 
    Aroma & 0.90 & 0.93 & 0.91 & 0.92 \\ 
    Palate & 0.89 & 0.92 & 0.94 & 0.95 \\ 
    Taste & 0.90 & 0.94 & 0.95 & 0.96 \\ 
    \end{tabularx}
    \caption{Cross AUC results for different representations on the BeerAdvocate data. Row: Embedding used. Column: Aspect evaluated against. }
    \label{table:beercrossauc}
\end{table}

\begin{table}
\setlength{\tabcolsep}{2pt}
\footnotesize
\centering
\begin{tabularx}{\columnwidth}{l|X X|X X|X X|X X}
 & \multicolumn{2}{c|}{Look}  & \multicolumn{2}{c|}{Aroma}  & \multicolumn{2}{c|}{Palate} & \multicolumn{2}{c}{Taste} \\ \hline
Look & - & - & 0.42 &	0.60 & 0.40	& 0.63	& 0.38	& 0.65 \\ 
Aroma & 0.33 & 0.69 & - & - & 0.41 & 0.59 & 0.41 & 0.60 \\ 
Palate & 0.32 & 0.70 & 0.46 & 0.54 & - & - & 0.49 & 0.52 \\ 
Taste & 0.23 & 0.80 & 0.35 & 0.66 & 0.33 & 0.67 & - & - \\
\end{tabularx}
\caption{`Decorrelated' cross-AUC results on the BeerAdvocate data, which attempt to mitigate confounding due to overall sentiment being correlated. Each cell reports metrics over subsets of reviews in which the sentiment differs between the row and column aspects. The numbers in each cell are the AUCs w.r.t. sentiment regarding the column aspect achieved using the row and column aspect representations, respectively.}
\label{table:decorrelated}
\end{table}

We conducted experiments on the BeerAdvocate dataset \cite{mcauley2012learning}, which contains 1.5M reviews of beers that address four aspects: \emph{appearance}, \emph{aroma}, \emph{palate}, and \emph{taste}. Free-text reviews are associated with aspect-specific numerical ratings for each of these, ranging from 1 to 5. We consider ratings $<3$ as negative, and $>3$ as positive, and use these to generate triplets of reviews. For each aspect $a$, we construct triplets $(s, d, o)_a$ by first randomly sampling a review $d$. We then select $s$ to be a review with the same sentiment with respect to $a$ as $d$, and $o$ to be a review with the opposite sentiment regarding $a$. We selected 90K reviews for experiments, such that we had an equal number of positive and negative reviews for each aspect. We only keep words appearing in at least 5 documents, converting all others to $\texttt{unk}$. We truncated reviews to 95 percentile length. We split our data into 80/10/10 ratio for training, validation and testing, respectively. 

\vspace{.25em}
\noindent \textbf{Baselines}. We used the same baselines as for the PICO domain, save for \emph{RR-TF}, which was domain-specific. Here we also evaluate the result of replacing the CNN-based encoder with NVDM, BoW and DSSM based encoders, respectively, each trained using triplet loss.

\vspace{.25em}
\noindent  \textbf{Hyperparameters and Settings}. For the CNN-based encoder, we used settings and hyperparameters as described for the PICO domain. For the BoW encoder, we used 800d output embeddings and a PReLU activation function with $\ell_2$ regularization set to $1e$-$5$. For the NVDM based encoder, we used 200d embeddings. 

\vspace{.25em}
\noindent \textbf{Metrics}. We again performed an IR-type evaluation to assess the utility of representations. For each aspect $k$, we constructed an affinity matrix $A^k$ such that $A^k_{ij} = {\textit sim}_k(r_i, r_j)$ for beer reviews $r_i$ and $r_j$. We consider two reviews similar under a given aspect $k$ if they have the same (dichotomized) sentiment value for said aspect. We compute AUCs for each review and aspect using the affinity matrix $A_k$. The AUC values are averaged over reviews in the test set to obtain a final AUC metric for each aspect. We also report cross AUC measures in which we use embeddings for aspect $k$ to distinguish reviews under aspect $k'$. 

\vspace{.25em}
\noindent \textbf{Results} We report the AUC measures for each aspect on our test set using different representations in Table \ref{table:beerauc}. Our model consistently outperforms baseline strategies over all aspects. Unsurprisingly, the model outperforms unsupervised approaches.\footnote{We are not sure why ABAE \cite{he-2017} performs so poorly on the review corpora. It may simply fail to prominently encode sentiment, which is important for these tasks. We note that this model performs reasonably well on the PICO data above, and qualitatively seems to recover reasonable aspects (though not specifically sentiment).} We realize consistent though modest improvement over triplet-supervised approaches that use alternative encoders. 

In Table \ref{table:beercrossauc} we present cross AUC evaluations. Rows correspond to the embedding used and columns to the aspect evaluated against. As expected, aspect-embeddings perform better w.r.t.~the aspects for which they code, suggesting some disentanglement. However, the reduction in performance when using one aspect representation to discriminate w.r.t.~others is not as pronounced as above. This is because aspect ratings are highly correlated: if taste is positive, aroma is very likely to be as well. Effectively, here sentiment entangles all of these aspects.\footnote{Another view is that we are in fact inducing representations of $<$aspect, sentiment$>$ pairs, and only the aspect varies across these; thus representations remain discriminative (w.r.t.~sentiment) across aspects.} 

In Table \ref{table:decorrelated}, we evaluate cross AUC performance for beer by first `decorrelating' the aspects. Specifically, for each cell $(k, k')$ in the table, we first retrieve the subset of reviews in which the sentiment w.r.t. $k$ differs from the sentiment w.r.t. $k'$. Then we evaluate the AUC similarity of these reviews on the basis of sentiment concerning $k'$ using both $k$ and $k'$ embeddings, yielding a pair of AUCs (listed respectively). We observe that the using $k'$ embeddings to evaluate aspect $k'$ similarity yields better results than using $k$ embeddings.

We present the most activated words for each aspect (as per the gating mechanism) in Table \ref{table:Beermost}. And we present an illustrative review color-coded with aspect-wise gate activations in Table \ref{table:beer-example}. For completeness, we reproduce the top words for aspects discovered using \newcite{he-2017} in the Appendix; these do not obviously align with the target aspects, which is unsurprising given that this is an unsupervised method.

\begin{table}
\vspace{-.5em}
\footnotesize
\centering
\begin{tabularx}{\columnwidth}{X} 
{\bf Look} \emph{attractive}, \emph{beautiful}, \emph{fingers}, \emph{pumpkin}, \emph{quarter}, \emph{received}, \emph{retention}, \emph{sheets}, \emph{sipper}, \emph{well-balanced} \\\hline 
{\bf Aroma} \emph{beer}, \emph{cardboard}, \emph{cheap}, \emph{down}, \emph{follows}, \emph{medium-light}, \emph{rice}, \emph{settled}, \emph{skunked}, \emph{skunky} \\\hline 
{\bf Palate} \emph{bother}, \emph{crafted}, \emph{luscious}, \emph{mellow}, \emph{mint}, \emph{range}, \emph{recommended}, \emph{roasted}, \emph{tasting}, \emph{weight} \\\hline 
{\bf Taste} \emph{amazingly}, \emph{down}, \emph{highly}, \emph{product}, \emph{recommended}, \emph{tasted}, \emph{thoroughly}, \emph{to}, \emph{truly}, \emph{wow} \\
\end{tabularx}
\caption{Most activated words for aspects on the beer corpus, as per the gating mechanism.}
\label{table:Beermost}
\vspace{-1em}
\end{table}

\begin{table*}
\footnotesize
\centering
\begin{tabularx}{\textwidth}{X|X|X|X}
\textbf{Look} : {\setlength{\fboxsep}{0pt}\colorbox{red!99!white}{\strut deep}} {\setlength{\fboxsep}{0pt}\colorbox{red!99!white}{\strut amber}} {\setlength{\fboxsep}{0pt}\colorbox{red!80!white}{\strut hue}} {\setlength{\fboxsep}{0pt}\colorbox{red!58!white}{\strut ,}} {\setlength{\fboxsep}{0pt}\colorbox{red!31!white}{\strut this}} {\setlength{\fboxsep}{0pt}\colorbox{red!34!white}{\strut brew}} {\setlength{\fboxsep}{0pt}\colorbox{red!48!white}{\strut is}} {\setlength{\fboxsep}{0pt}\colorbox{red!75!white}{\strut topped}} {\setlength{\fboxsep}{0pt}\colorbox{red!92!white}{\strut with}} {\setlength{\fboxsep}{0pt}\colorbox{red!99!white}{\strut a}} {\setlength{\fboxsep}{0pt}\colorbox{red!99!white}{\strut finger}} {\setlength{\fboxsep}{0pt}\colorbox{red!99!white}{\strut of}} {\setlength{\fboxsep}{0pt}\colorbox{red!87!white}{\strut off}} {\setlength{\fboxsep}{0pt}\colorbox{red!53!white}{\strut white}} {\setlength{\fboxsep}{0pt}\colorbox{red!20!white}{\strut head}} {\setlength{\fboxsep}{0pt}\colorbox{red!0!white}{\strut .}} {\setlength{\fboxsep}{0pt}\colorbox{red!0!white}{\strut smell}} {\setlength{\fboxsep}{0pt}\colorbox{red!0!white}{\strut of}} {\setlength{\fboxsep}{0pt}\colorbox{red!0!white}{\strut dog}} {\setlength{\fboxsep}{0pt}\colorbox{red!0!white}{\strut unk}} {\setlength{\fboxsep}{0pt}\colorbox{red!0!white}{\strut ,}} {\setlength{\fboxsep}{0pt}\colorbox{red!0!white}{\strut green}} {\setlength{\fboxsep}{0pt}\colorbox{red!0!white}{\strut unk}} {\setlength{\fboxsep}{0pt}\colorbox{red!0!white}{\strut ,}} {\setlength{\fboxsep}{0pt}\colorbox{red!0!white}{\strut and}} {\setlength{\fboxsep}{0pt}\colorbox{red!0!white}{\strut slightly}} {\setlength{\fboxsep}{0pt}\colorbox{red!0!white}{\strut fruity}} {\setlength{\fboxsep}{0pt}\colorbox{red!0!white}{\strut .}} {\setlength{\fboxsep}{0pt}\colorbox{red!27!white}{\strut taste}} {\setlength{\fboxsep}{0pt}\colorbox{red!27!white}{\strut of}} {\setlength{\fboxsep}{0pt}\colorbox{red!27!white}{\strut belgian}} {\setlength{\fboxsep}{0pt}\colorbox{red!0!white}{\strut yeast}} {\setlength{\fboxsep}{0pt}\colorbox{red!0!white}{\strut ,}} {\setlength{\fboxsep}{0pt}\colorbox{red!0!white}{\strut coriander}} {\setlength{\fboxsep}{0pt}\colorbox{red!0!white}{\strut ,}} {\setlength{\fboxsep}{0pt}\colorbox{red!2!white}{\strut hard}} {\setlength{\fboxsep}{0pt}\colorbox{red!5!white}{\strut water}} {\setlength{\fboxsep}{0pt}\colorbox{red!5!white}{\strut and}} {\setlength{\fboxsep}{0pt}\colorbox{red!2!white}{\strut bready}} {\setlength{\fboxsep}{0pt}\colorbox{red!0!white}{\strut malt}} {\setlength{\fboxsep}{0pt}\colorbox{red!0!white}{\strut .}} {\setlength{\fboxsep}{0pt}\colorbox{red!0!white}{\strut light}} {\setlength{\fboxsep}{0pt}\colorbox{red!0!white}{\strut body}} {\setlength{\fboxsep}{0pt}\colorbox{red!0!white}{\strut ,}} {\setlength{\fboxsep}{0pt}\colorbox{red!0!white}{\strut with}} {\setlength{\fboxsep}{0pt}\colorbox{red!31!white}{\strut little}} {\setlength{\fboxsep}{0pt}\colorbox{red!31!white}{\strut carbonation}} {\setlength{\fboxsep}{0pt}\colorbox{red!31!white}{\strut .}} & \textbf{Aroma} : {\setlength{\fboxsep}{0pt}\colorbox{green!1!white}{\strut deep}} {\setlength{\fboxsep}{0pt}\colorbox{green!1!white}{\strut amber}} {\setlength{\fboxsep}{0pt}\colorbox{green!0!white}{\strut hue}} {\setlength{\fboxsep}{0pt}\colorbox{green!0!white}{\strut ,}} {\setlength{\fboxsep}{0pt}\colorbox{green!15!white}{\strut this}} {\setlength{\fboxsep}{0pt}\colorbox{green!16!white}{\strut brew}} {\setlength{\fboxsep}{0pt}\colorbox{green!16!white}{\strut is}} {\setlength{\fboxsep}{0pt}\colorbox{green!0!white}{\strut topped}} {\setlength{\fboxsep}{0pt}\colorbox{green!0!white}{\strut with}} {\setlength{\fboxsep}{0pt}\colorbox{green!3!white}{\strut a}} {\setlength{\fboxsep}{0pt}\colorbox{green!3!white}{\strut finger}} {\setlength{\fboxsep}{0pt}\colorbox{green!3!white}{\strut of}} {\setlength{\fboxsep}{0pt}\colorbox{green!0!white}{\strut off}} {\setlength{\fboxsep}{0pt}\colorbox{green!47!white}{\strut white}} {\setlength{\fboxsep}{0pt}\colorbox{green!47!white}{\strut head}} {\setlength{\fboxsep}{0pt}\colorbox{green!100!white}{\strut .}} {\setlength{\fboxsep}{0pt}\colorbox{green!52!white}{\strut smell}} {\setlength{\fboxsep}{0pt}\colorbox{green!97!white}{\strut of}} {\setlength{\fboxsep}{0pt}\colorbox{green!97!white}{\strut dog}} {\setlength{\fboxsep}{0pt}\colorbox{green!97!white}{\strut unk}} {\setlength{\fboxsep}{0pt}\colorbox{green!51!white}{\strut ,}} {\setlength{\fboxsep}{0pt}\colorbox{green!50!white}{\strut green}} {\setlength{\fboxsep}{0pt}\colorbox{green!53!white}{\strut unk}} {\setlength{\fboxsep}{0pt}\colorbox{green!53!white}{\strut ,}} {\setlength{\fboxsep}{0pt}\colorbox{green!3!white}{\strut and}} {\setlength{\fboxsep}{0pt}\colorbox{green!27!white}{\strut slightly}} {\setlength{\fboxsep}{0pt}\colorbox{green!27!white}{\strut fruity}} {\setlength{\fboxsep}{0pt}\colorbox{green!71!white}{\strut .}} {\setlength{\fboxsep}{0pt}\colorbox{green!58!white}{\strut taste}} {\setlength{\fboxsep}{0pt}\colorbox{green!58!white}{\strut of}} {\setlength{\fboxsep}{0pt}\colorbox{green!13!white}{\strut belgian}} {\setlength{\fboxsep}{0pt}\colorbox{green!0!white}{\strut yeast}} {\setlength{\fboxsep}{0pt}\colorbox{green!0!white}{\strut ,}} {\setlength{\fboxsep}{0pt}\colorbox{green!0!white}{\strut coriander}} {\setlength{\fboxsep}{0pt}\colorbox{green!0!white}{\strut ,}} {\setlength{\fboxsep}{0pt}\colorbox{green!18!white}{\strut hard}} {\setlength{\fboxsep}{0pt}\colorbox{green!18!white}{\strut water}} {\setlength{\fboxsep}{0pt}\colorbox{green!18!white}{\strut and}} {\setlength{\fboxsep}{0pt}\colorbox{green!0!white}{\strut bready}} {\setlength{\fboxsep}{0pt}\colorbox{green!0!white}{\strut malt}} {\setlength{\fboxsep}{0pt}\colorbox{green!0!white}{\strut .}} {\setlength{\fboxsep}{0pt}\colorbox{green!0!white}{\strut light}} {\setlength{\fboxsep}{0pt}\colorbox{green!0!white}{\strut body}} {\setlength{\fboxsep}{0pt}\colorbox{green!0!white}{\strut ,}} {\setlength{\fboxsep}{0pt}\colorbox{green!0!white}{\strut with}} {\setlength{\fboxsep}{0pt}\colorbox{green!52!white}{\strut little}} {\setlength{\fboxsep}{0pt}\colorbox{green!52!white}{\strut carbonation}} {\setlength{\fboxsep}{0pt}\colorbox{green!52!white}{\strut .}} & \textbf{Palate} : {\setlength{\fboxsep}{0pt}\colorbox{cyan!20!white}{\strut deep}} {\setlength{\fboxsep}{0pt}\colorbox{cyan!10!white}{\strut amber}} {\setlength{\fboxsep}{0pt}\colorbox{cyan!33!white}{\strut hue}} {\setlength{\fboxsep}{0pt}\colorbox{cyan!0!white}{\strut ,}} {\setlength{\fboxsep}{0pt}\colorbox{cyan!4!white}{\strut this}} {\setlength{\fboxsep}{0pt}\colorbox{cyan!36!white}{\strut brew}} {\setlength{\fboxsep}{0pt}\colorbox{cyan!39!white}{\strut is}} {\setlength{\fboxsep}{0pt}\colorbox{cyan!34!white}{\strut topped}} {\setlength{\fboxsep}{0pt}\colorbox{cyan!3!white}{\strut with}} {\setlength{\fboxsep}{0pt}\colorbox{cyan!0!white}{\strut a}} {\setlength{\fboxsep}{0pt}\colorbox{cyan!0!white}{\strut finger}} {\setlength{\fboxsep}{0pt}\colorbox{cyan!0!white}{\strut of}} {\setlength{\fboxsep}{0pt}\colorbox{cyan!0!white}{\strut off}} {\setlength{\fboxsep}{0pt}\colorbox{cyan!0!white}{\strut white}} {\setlength{\fboxsep}{0pt}\colorbox{cyan!0!white}{\strut head}} {\setlength{\fboxsep}{0pt}\colorbox{cyan!0!white}{\strut .}} {\setlength{\fboxsep}{0pt}\colorbox{cyan!0!white}{\strut smell}} {\setlength{\fboxsep}{0pt}\colorbox{cyan!0!white}{\strut of}} {\setlength{\fboxsep}{0pt}\colorbox{cyan!0!white}{\strut dog}} {\setlength{\fboxsep}{0pt}\colorbox{cyan!0!white}{\strut unk}} {\setlength{\fboxsep}{0pt}\colorbox{cyan!0!white}{\strut ,}} {\setlength{\fboxsep}{0pt}\colorbox{cyan!0!white}{\strut green}} {\setlength{\fboxsep}{0pt}\colorbox{cyan!0!white}{\strut unk}} {\setlength{\fboxsep}{0pt}\colorbox{cyan!0!white}{\strut ,}} {\setlength{\fboxsep}{0pt}\colorbox{cyan!0!white}{\strut and}} {\setlength{\fboxsep}{0pt}\colorbox{cyan!0!white}{\strut slightly}} {\setlength{\fboxsep}{0pt}\colorbox{cyan!3!white}{\strut fruity}} {\setlength{\fboxsep}{0pt}\colorbox{cyan!3!white}{\strut .}} {\setlength{\fboxsep}{0pt}\colorbox{cyan!5!white}{\strut taste}} {\setlength{\fboxsep}{0pt}\colorbox{cyan!1!white}{\strut of}} {\setlength{\fboxsep}{0pt}\colorbox{cyan!1!white}{\strut belgian}} {\setlength{\fboxsep}{0pt}\colorbox{cyan!0!white}{\strut yeast}} {\setlength{\fboxsep}{0pt}\colorbox{cyan!0!white}{\strut ,}} {\setlength{\fboxsep}{0pt}\colorbox{cyan!2!white}{\strut coriander}} {\setlength{\fboxsep}{0pt}\colorbox{cyan!35!white}{\strut ,}} {\setlength{\fboxsep}{0pt}\colorbox{cyan!69!white}{\strut hard}} {\setlength{\fboxsep}{0pt}\colorbox{cyan!98!white}{\strut water}} {\setlength{\fboxsep}{0pt}\colorbox{cyan!64!white}{\strut and}} {\setlength{\fboxsep}{0pt}\colorbox{cyan!31!white}{\strut bready}} {\setlength{\fboxsep}{0pt}\colorbox{cyan!33!white}{\strut malt}} {\setlength{\fboxsep}{0pt}\colorbox{cyan!66!white}{\strut .}} {\setlength{\fboxsep}{0pt}\colorbox{cyan!100!white}{\strut light}} {\setlength{\fboxsep}{0pt}\colorbox{cyan!99!white}{\strut body}} {\setlength{\fboxsep}{0pt}\colorbox{cyan!99!white}{\strut ,}} {\setlength{\fboxsep}{0pt}\colorbox{cyan!99!white}{\strut with}} {\setlength{\fboxsep}{0pt}\colorbox{cyan!99!white}{\strut little}} {\setlength{\fboxsep}{0pt}\colorbox{cyan!98!white}{\strut carbonation}} {\setlength{\fboxsep}{0pt}\colorbox{cyan!65!white}{\strut .}} & \textbf{Taste} :{\setlength{\fboxsep}{0pt}\colorbox{yellow!40!white}{\strut deep}} {\setlength{\fboxsep}{0pt}\colorbox{yellow!42!white}{\strut amber}} {\setlength{\fboxsep}{0pt}\colorbox{yellow!12!white}{\strut hue}} {\setlength{\fboxsep}{0pt}\colorbox{yellow!6!white}{\strut ,}} {\setlength{\fboxsep}{0pt}\colorbox{yellow!49!white}{\strut this}} {\setlength{\fboxsep}{0pt}\colorbox{yellow!94!white}{\strut brew}} {\setlength{\fboxsep}{0pt}\colorbox{yellow!100!white}{\strut is}} {\setlength{\fboxsep}{0pt}\colorbox{yellow!55!white}{\strut topped}} {\setlength{\fboxsep}{0pt}\colorbox{yellow!10!white}{\strut with}} {\setlength{\fboxsep}{0pt}\colorbox{yellow!4!white}{\strut a}} {\setlength{\fboxsep}{0pt}\colorbox{yellow!4!white}{\strut finger}} {\setlength{\fboxsep}{0pt}\colorbox{yellow!4!white}{\strut of}} {\setlength{\fboxsep}{0pt}\colorbox{yellow!0!white}{\strut off}} {\setlength{\fboxsep}{0pt}\colorbox{yellow!0!white}{\strut white}} {\setlength{\fboxsep}{0pt}\colorbox{yellow!0!white}{\strut head}} {\setlength{\fboxsep}{0pt}\colorbox{yellow!11!white}{\strut .}} {\setlength{\fboxsep}{0pt}\colorbox{yellow!11!white}{\strut smell}} {\setlength{\fboxsep}{0pt}\colorbox{yellow!53!white}{\strut of}} {\setlength{\fboxsep}{0pt}\colorbox{yellow!43!white}{\strut dog}} {\setlength{\fboxsep}{0pt}\colorbox{yellow!43!white}{\strut unk}} {\setlength{\fboxsep}{0pt}\colorbox{yellow!34!white}{\strut ,}} {\setlength{\fboxsep}{0pt}\colorbox{yellow!35!white}{\strut green}} {\setlength{\fboxsep}{0pt}\colorbox{yellow!37!white}{\strut unk}} {\setlength{\fboxsep}{0pt}\colorbox{yellow!3!white}{\strut ,}} {\setlength{\fboxsep}{0pt}\colorbox{yellow!41!white}{\strut and}} {\setlength{\fboxsep}{0pt}\colorbox{yellow!39!white}{\strut slightly}} {\setlength{\fboxsep}{0pt}\colorbox{yellow!39!white}{\strut fruity}} {\setlength{\fboxsep}{0pt}\colorbox{yellow!0!white}{\strut .}} {\setlength{\fboxsep}{0pt}\colorbox{yellow!0!white}{\strut taste}} {\setlength{\fboxsep}{0pt}\colorbox{yellow!44!white}{\strut of}} {\setlength{\fboxsep}{0pt}\colorbox{yellow!48!white}{\strut belgian}} {\setlength{\fboxsep}{0pt}\colorbox{yellow!48!white}{\strut yeast}} {\setlength{\fboxsep}{0pt}\colorbox{yellow!6!white}{\strut ,}} {\setlength{\fboxsep}{0pt}\colorbox{yellow!35!white}{\strut coriander}} {\setlength{\fboxsep}{0pt}\colorbox{yellow!66!white}{\strut ,}} {\setlength{\fboxsep}{0pt}\colorbox{yellow!94!white}{\strut hard}} {\setlength{\fboxsep}{0pt}\colorbox{yellow!61!white}{\strut water}} {\setlength{\fboxsep}{0pt}\colorbox{yellow!29!white}{\strut and}} {\setlength{\fboxsep}{0pt}\colorbox{yellow!4!white}{\strut bready}} {\setlength{\fboxsep}{0pt}\colorbox{yellow!15!white}{\strut malt}} {\setlength{\fboxsep}{0pt}\colorbox{yellow!35!white}{\strut .}} {\setlength{\fboxsep}{0pt}\colorbox{yellow!70!white}{\strut light}} {\setlength{\fboxsep}{0pt}\colorbox{yellow!58!white}{\strut body}} {\setlength{\fboxsep}{0pt}\colorbox{yellow!38!white}{\strut ,}} {\setlength{\fboxsep}{0pt}\colorbox{yellow!2!white}{\strut with}} {\setlength{\fboxsep}{0pt}\colorbox{yellow!19!white}{\strut little}} {\setlength{\fboxsep}{0pt}\colorbox{yellow!20!white}{\strut carbonation}} {\setlength{\fboxsep}{0pt}\colorbox{yellow!17!white}{\strut .}}
\end{tabularx}
\caption{Gate activations for each aspect in an example beer review. }
\label{table:beer-example}
\end{table*} 

\subsubsection{Hotel \& Restaurant Reviews}

Finally, we attempt to learn embeddings that disentangle domain from sentiment in reviews. For this we use a combination of TripAdvisor and Yelp! ratings data. The former comprises reviews of hotels, the latter of restaurants; both use a scale of 1 to 5. We convert ratings into positive/negative labels as above. Here we consider aspects to be the domain (hotel or restaurant) and the sentiment (positive or negative). We aim to generate embeddings that capture information about only one of these aspects. We use 50K reviews from each dataset for training and 5K for testing. 

\vspace{.2em}
\noindent \textbf{Baselines}. We use the same baselines as for the BeerAdvocate data, and similarly use different encoder models trained under triplet loss.

\begin{table}
\footnotesize
    \centering
    \begin{tabularx}{\columnwidth}{l X X}
     Baseline & Domain & Sentiment  \\
    \hline
    TF-IDF & 0.59 & 0.52   \\
    Doc2Vec & 0.83 & 0.56 \\
    LDA & 0.90 & 0.62 \\
    NVDM & 0.79 & 0.63  \\
    ABAE & 0.50 & 0.50 \\
    \hline
    BoW + Triplet & \textbf{0.99} & 0.91 \\
    NVDM + Triplet & \textbf{0.99} & 0.91 \\
    DSSM + Triplet &  \textbf{0.99} & 0.90 \\
    CNN + Triplet & \textbf{0.99} & \textbf{0.92}  \\
    \end{tabularx}
    \caption{AUC results for different representations on the Yelp!/TripAdvisor Data. Models beneath the second line are supervised.}
    \label{table:foodauc}
\end{table}

\vspace{.2em}
\noindent \textbf{Evaluation Metrics}. We perform AUC and cross-AUC evaluation as in the preceding section. For the domain aspect, we consider two reviews similar if they are from the same domain, irrespective of sentiment. Similarly, reviews are considered similar with respect to the sentiment aspect if they share a sentiment value, regardless of domain.




\begin{table}
\footnotesize
    \centering
    \begin{tabularx}{\columnwidth}{l X X}
     Baseline & Domain & Sentiment  \\
    \hline
    Domain & 0.988 & 0.512   \\
    Sentiment & 0.510 & 0.917  \\
    \end{tabularx}
    \caption{Cross AUC results for different representations for Yelp!/TripAdvisor Dataset.}
    \label{table:foodcrossauc}
\end{table}

\vspace{.25em}
\noindent  \textbf{Results}. In Table \ref{table:foodauc} we report the AUCs for each aspect on our test set using different representations. Baselines perform reasonably well on the domain aspect because reviews from different domains are quite dissimilar. Capturing sentiment information irrespective of domain is more difficult, and most unsupervised models fail in this respect. In Table \ref{table:foodcrossauc}, we observe that cross AUC results are much more pronounced than for the BeerAdvocate data, as the domain and sentiment are uncorrelated (i.e., sentiment is independent of domain).

\section{Related Work}
\label{section:related-work}

Work in representation learning for NLP has largely focused on improving word embeddings \cite{levy2014dependency,faruqui2014retrofitting,huang2012improving}. But efforts have also been made to embed other textual units, e.g. characters \cite{kim2016character}, and lengthier texts including sentences, paragraphs, and documents \cite{le2014distributed,kiros2015skip}. 

Triplet-based judgments have been used in multiple domains, including vision and NLP, to estimate similarity information implicitly. For example, triplet-based similarity embeddings may be learned using `crowdkernels' with applications to multi-view clustering \cite{amid2015multiview}. Models combining similarity with neural networks mainly revolve around Siamese networks \cite{chopra2005learning} which use pairwise distances to learn embeddings \cite{schroff2015facenet}, a tactic we have followed here. Similarity judgments have also been used to generate document embeddings for IR tasks \cite{shen2014latent,das2016together}.

Recently, \newcite{he-2017} introduced a neural model for aspect extraction that relies on an attention mechanism to identify aspect words. They proposed an autoencoder variant designed to tease apart aspects. In contrast to the method we propose, their approach is unsupervised; discovered aspects may thus not have a clear interpretation. Experiments reported here support this hypothesis, and we provide additional results using their model in the Appendix.



Other recent work has focused on text \emph{generation} from factorized representations \cite{larsson2017disentangled}. And \newcite{zhang2017aspect} proposed a lightly supervised method for \emph{domain adaptation} using aspect-augmented neural networks. They exploited \emph{source} document labels to train a classifier for a \emph{target} aspect. They leveraged sentence-level scores codifying sentence relevance w.r.t.~individual aspects, which were derived from terms \emph{a priori} associated with aspects. This supervision is used to construct a composite loss that captures both classification performance on the source task and a term that enforces \emph{invariance} between source and target representations.



There is also a large body of work that uses probabilistic generative models to recover latent structure in texts. Many of these models derive from Latent Dirichlet Allocation (LDA) \cite{blei2003latent}, and some variants have explicitly represented topics and aspects jointly for sentiment tasks \cite{brody2010unsupervised,sauper2010incorporating,sauper2011content,mukherjee2012aspect,sauper2013automatic,kim2013hierarchical}. 

A bit more generally, \emph{aspects} have also been interpreted as properties spanning entire texts, e.g., a perspective or theme which may then color the discussion of topics \cite{paul2010two}. This intuition led to the development of the \emph{factorial LDA} family of topic models \cite{paul2012factorial,wallace2014large}; these model individual word probability as a product of multiple latent factors characterizing a text. This is similar to the Sparse Additive Generative (SAGE) model of text proposed by \newcite{SAGE}.


\section{Conclusions}

We have proposed an approach for inducing disentangled representations of text. To learn such representations we have relied on supervision codified in aspect-wise similarity judgments expressed as document triplets. This provides a general supervision framework and objective. We evaluated this approach on three datasets, each with different aspects. Our experimental results demonstrate that this approach indeed induces aspect-specific embeddings that are qualitatively interpretable and achieve superior performance on information retrieval tasks. 

Going forward, disentangled representations may afford additional advantages in NLP, e.g., by facilitating transfer \cite{zhang2017aspect}, or supporting aspect-focused summarization models.

\section{Acknowledgements}
This work was supported in part by National Library of Medicine (NLM) of the National Institutes of Health (NIH) award R01LM012086, by the Army Research Office (ARO) award W911NF1810328, and research funds from Northeastern University. The content is solely the responsibility of the authors.

\bibliography{emnlp2018}

\begin{thebibliography}{43}
\expandafter\ifx\csname natexlab\endcsname\relax\def\natexlab#1{#1}\fi

\bibitem[{Amid and Ukkonen(2015)}]{amid2015multiview}
Ehsan Amid and Antti Ukkonen. 2015.
\newblock {Multiview triplet embedding: Learning attributes in multiple maps}.
\newblock In \emph{International Conference on Machine Learning}.

\bibitem[{Blei et~al.(2003)Blei, Ng, and Jordan}]{blei2003latent}
David~M Blei, Andrew~Y Ng, and Michael~I Jordan. 2003.
\newblock {Latent dirichlet allocation}.
\newblock \emph{Journal of machine Learning research}, 3(Jan).

\bibitem[{Brody and Elhadad(2010)}]{brody2010unsupervised}
Samuel Brody and Noemie Elhadad. 2010.
\newblock {An unsupervised aspect-sentiment model for online reviews}.
\newblock In \emph{Proceedings of the Conference of the North American Chapter
  of the Association for Computational Linguistics (NAACL)}. Association for
  Computational Linguistics.

\bibitem[{Chen et~al.(2016)Chen, Duan, Houthooft, Schulman, Sutskever, and
  Abbeel}]{chen2016infogan}
Xi~Chen, Yan Duan, Rein Houthooft, John Schulman, Ilya Sutskever, and Pieter
  Abbeel. 2016.
\newblock {Infogan: Interpretable Representation Learning by Information
  Maximizing Generative Adversarial Nets}.
\newblock In \emph{Advances in Neural Information Processing Systems}.

\bibitem[{Chopra et~al.(2005)Chopra, Hadsell, and LeCun}]{chopra2005learning}
Sumit Chopra, Raia Hadsell, and Yann LeCun. 2005.
\newblock {Learning a similarity metric discriminatively, with application to
  face verification}.
\newblock In \emph{CVPR 2005. IEEE Computer Society Conference on Computer
  Vision and Pattern Recognition, 2005.}

\bibitem[{Cohen et~al.(2006)Cohen, Hersh, Peterson, and
  Yen}]{cohen2006reducing}
Aaron~M Cohen, William~R Hersh, K~Peterson, and Po-Yin Yen. 2006.
\newblock {Reducing workload in systematic review preparation using automated
  citation classification}.
\newblock \emph{Journal of the American Medical Informatics Association},
  13(2).

\bibitem[{Das et~al.(2016)Das, Yenala, Chinnakotla, and
  Shrivastava}]{das2016together}
Arpita Das, Harish Yenala, Manoj Chinnakotla, and Manish Shrivastava. 2016.
\newblock {Together we stand: Siamese networks for similar question retrieval}.
\newblock In \emph{Proceedings of the 54th Annual Meeting of the Association
  for Computational Linguistics}.

\bibitem[{Eisenstein et~al.(2011)Eisenstein, Ahmed, and Xing}]{SAGE}
Jacob Eisenstein, Amr Ahmed, and Eric~P Xing. 2011.
\newblock {Sparse Additive Generative Models of Text.}
\newblock In \emph{Proceedings of the International Conference on Machine
  Learning (ICML)}.

\bibitem[{Faruqui et~al.(2015)Faruqui, Dodge, Jauhar, Dyer, Hovy, and
  Smith}]{faruqui2014retrofitting}
Manaal Faruqui, Jesse Dodge, Sujay~K Jauhar, Chris Dyer, Eduard Hovy, and
  Noah~A Smith. 2015.
\newblock {Retrofitting word vectors to semantic lexicons}.
\newblock \emph{Proceedings of the Conference of the North American Chapter of
  the Association for Computational Linguistics (NAACL)}.

\bibitem[{He et~al.(2015)He, Zhang, Ren, and Sun}]{he2015delving}
Kaiming He, Xiangyu Zhang, Shaoqing Ren, and Jian Sun. 2015.
\newblock {Delving deep into rectifiers: Surpassing human-level performance on
  imagenet classification}.
\newblock In \emph{Proceedings of the IEEE international conference on computer
  vision}.

\bibitem[{He et~al.(2017)He, Lee, Ng, and Dahlmeier}]{he-2017}
Ruidan He, Wee~Sun Lee, Hwee~Tou Ng, and Daniel Dahlmeier. 2017.
\newblock {An Unsupervised Neural Attention Model for Aspect Extraction}.
\newblock In \emph{Proceedings of the Association for Computational Linguistics
  (ACL)}. Association for Computational Linguistics.

\bibitem[{Higgins et~al.(2017)Higgins, Matthey, Pal, Burgess, Glorot,
  Botvinick, Mohamed, and Lerchner}]{higgins2017beta-vae}
Irina Higgins, Loic Matthey, Arka Pal, Christopher Burgess, Xavier Glorot,
  Matthew Botvinick, Shakir Mohamed, and Alexander Lerchner. 2017.
\newblock {Beta-Vae: Learning Basic Visual Concepts with a Constrained
  Variational Framework}.
\newblock \emph{International Conference on Learning Representations}.

\bibitem[{Huang et~al.(2012)Huang, Socher, Manning, and
  Ng}]{huang2012improving}
Eric~H Huang, Richard Socher, Christopher~D Manning, and Andrew~Y Ng. 2012.
\newblock {Improving word representations via global context and multiple word
  prototypes}.
\newblock In \emph{Proceedings of the 50th Annual Meeting of the Association
  for Computational Linguistics}. Association for Computational Linguistics.

\bibitem[{Karaletsos et~al.(2015)Karaletsos, Belongie, and
  R{\"{a}}tsch}]{karaletsos2015bayesian}
Theofanis Karaletsos, Serge Belongie, and Gunnar R{\"{a}}tsch. 2015.
\newblock {Bayesian representation learning with oracle constraints}.
\newblock \emph{arXiv preprint arXiv:1506.05011}.

\bibitem[{Kim et~al.(2013)Kim, Zhang, Chen, Oh, and Liu}]{kim2013hierarchical}
Suin Kim, Jianwen Zhang, Zheng Chen, Alice~H Oh, and Shixia Liu. 2013.
\newblock {A Hierarchical Aspect-Sentiment Model for Online Reviews.}
\newblock In \emph{Proceedings of the AAAI Conference on Artificial
  Intelligence}.

\bibitem[{Kim et~al.(2016)Kim, Jernite, Sontag, and Rush}]{kim2016character}
Yoon Kim, Yacine Jernite, David Sontag, and Alexander~M Rush. 2016.
\newblock {Character-aware neural language models}.
\newblock In \emph{Proceedings of the AAAI Conference on Artificial
  Intelligence}.

\bibitem[{Kingma and Ba(2014)}]{kingma2014adam}
Diederik Kingma and Jimmy Ba. 2014.
\newblock {Adam: A method for stochastic optimization}.
\newblock \emph{arXiv preprint arXiv:1412.6980}.

\bibitem[{Kiros et~al.(2015)Kiros, Zhu, Salakhutdinov, Zemel, Urtasun,
  Torralba, and Fidler}]{kiros2015skip}
Ryan Kiros, Yukun Zhu, Ruslan~R Salakhutdinov, Richard Zemel, Raquel Urtasun,
  Antonio Torralba, and Sanja Fidler. 2015.
\newblock {Skip-thought vectors}.
\newblock In \emph{Advances in neural information processing systems}.

\bibitem[{Larsson et~al.(2017)Larsson, Nilsson, and
  K{\aa}geb{\"{a}}ck}]{larsson2017disentangled}
Maria Larsson, Amanda Nilsson, and Mikael K{\aa}geb{\"{a}}ck. 2017.
\newblock {Disentangled Representations for Manipulation of Sentiment in Text}.
\newblock \emph{arXiv preprint arXiv:1712.10066}.

\bibitem[{Le and Mikolov(2014)}]{le2014distributed}
Quoc~V Le and Tomas Mikolov. 2014.
\newblock {Distributed Representations of Sentences and Documents.}
\newblock In \emph{Proceedings of the International Conference on Machine
  Learning (ICML)}.

\bibitem[{Levy and Goldberg(2014)}]{levy2014dependency}
Omer Levy and Yoav Goldberg. 2014.
\newblock {Dependency-Based Word Embeddings.}
\newblock In \emph{Proceedings of the Association of Computational Linguistics
  (ACL)}.

\bibitem[{McAuley et~al.(2012)McAuley, Leskovec, and
  Jurafsky}]{mcauley2012learning}
Julian McAuley, Jure Leskovec, and Dan Jurafsky. 2012.
\newblock {Learning attitudes and attributes from multi-aspect reviews}.
\newblock In \emph{2012 IEEE 12th International Conference on Data Mining
  (ICDM)}.

\bibitem[{Miao et~al.(2016)Miao, Yu, and Blunsom}]{miao2016neural}
Yishu Miao, Lei Yu, and Phil Blunsom. 2016.
\newblock {Neural variational inference for text processing}.
\newblock In \emph{International Conference on Machine Learning}.

\bibitem[{Mukherjee and Liu(2012)}]{mukherjee2012aspect}
Arjun Mukherjee and Bing Liu. 2012.
\newblock {Aspect extraction through semi-supervised modeling}.
\newblock In \emph{Proceedings of the 50th Annual Meeting of the Association
  for Computational Linguistics: Long Papers-Volume 1}. Association for
  Computational Linguistics.

\bibitem[{Paul and Dredze(2012)}]{paul2012factorial}
Michael Paul and Mark Dredze. 2012.
\newblock {Factorial LDA: Sparse multi-dimensional text models}.
\newblock In \emph{Advances in Neural Information Processing Systems}.

\bibitem[{Paul and Girju(2010)}]{paul2010two}
Michael Paul and Roxana Girju. 2010.
\newblock {A two-dimensional topic-aspect model for discovering multi-faceted
  topics}.
\newblock In \emph{Proceedings of the AAAI Conference on Artificial
  Intelligence}.

\bibitem[{Pyysalo et~al.(2013)Pyysalo, Ginter, Moen, Salakoski, and
  Ananiadou}]{moen2013distributional}
Sampo Pyysalo, Filip Ginter, Hans Moen, Tapio Salakoski, and Sophia Ananiadou.
  2013.
\newblock {Distributional semantics resources for biomedical text processing}.
\newblock In \emph{Proceedings of the 5th International Symposium on Languages
  in Biology and Medicine}.

\bibitem[{Ruder et~al.(2016)Ruder, Ghaffari, and
  Breslin}]{ruder2016hierarchical}
Sebastian Ruder, Parsa Ghaffari, and John~G Breslin. 2016.
\newblock {A Hierarchical Model of Reviews for Aspect-based Sentiment
  Analysis}.
\newblock \emph{Proceedings of the 2016 Conference on Empirical Methods in
  Natural Language Processing}.

\bibitem[{Sackett et~al.(1996)Sackett, Rosenberg, Gray, Haynes, and
  Richardson}]{sackett1996evidence}
David~L Sackett, William M~C Rosenberg, J~A~Muir Gray, R~Brian Haynes, and
  W~Scott Richardson. 1996.
\newblock {Evidence based medicine: what it is and what it isn't}.
\newblock \emph{Bmj}, 312(7023).

\bibitem[{Sauper and Barzilay(2013)}]{sauper2013automatic}
Christina Sauper and Regina Barzilay. 2013.
\newblock {Automatic aggregation by joint modeling of aspects and values}.
\newblock \emph{Journal of Artificial Intelligence Research}.

\bibitem[{Sauper et~al.(2010)Sauper, Haghighi, and
  Barzilay}]{sauper2010incorporating}
Christina Sauper, Aria Haghighi, and Regina Barzilay. 2010.
\newblock {Incorporating content structure into text analysis applications}.
\newblock In \emph{Proceedings of the 2010 Conference on Empirical Methods in
  Natural Language Processing}.

\bibitem[{Sauper et~al.(2011)Sauper, Haghighi, and
  Barzilay}]{sauper2011content}
Christina Sauper, Aria Haghighi, and Regina Barzilay. 2011.
\newblock {Content models with attitude}.
\newblock In \emph{Proceedings of the 49th Annual Meeting of the Association
  for Computational Linguistics: Human Language Technologies-Volume 1}.

\bibitem[{Schroff et~al.(2015)Schroff, Kalenichenko, and
  Philbin}]{schroff2015facenet}
Florian Schroff, Dmitry Kalenichenko, and James Philbin. 2015.
\newblock {Facenet: A unified embedding for face recognition and clustering}.
\newblock In \emph{Proceedings of the IEEE Conference on Computer Vision and
  Pattern Recognition}.

\bibitem[{Shen et~al.(2014)Shen, He, Gao, Deng, and Mesnil}]{shen2014latent}
Yelong Shen, Xiaodong He, Jianfeng Gao, Li~Deng, and Gr{\'{e}}goire Mesnil.
  2014.
\newblock {A latent semantic model with convolutional-pooling structure for
  information retrieval}.
\newblock In \emph{Proceedings of the 23rd ACM International Conference on
  Conference on Information and Knowledge Management}.

\bibitem[{Stacey et~al.(2014)Stacey, L{\'{e}}gar{\'{e}}, Col, Bennett, Barry,
  Eden, Thomas, Lyddiatt, Thomson, Trevena, and Wu}]{o2009decision}
D~Stacey, F~L{\'{e}}gar{\'{e}}, N~F Col, C~L Bennett, M~J Barry, K~B Eden,
  H~Thomas, A~Lyddiatt, R~Thomson, L~Trevena, and J~H~C Wu. 2014.
\newblock {Decision aids for people facing health treatment or screening
  decisions}.
\newblock \emph{Cochrane Database of Systematic Reviews}, 1.

\bibitem[{Veit et~al.(2017{\natexlab{a}})Veit, Belongie, and
  Karaletsos}]{veit2017conditional}
Andreas Veit, Serge Belongie, and Theofanis Karaletsos. 2017{\natexlab{a}}.
\newblock {Conditional Similarity Networks}.
\newblock In \emph{Proceedings of the IEEE Conference on Computer Vision and
  Pattern Recognition}.

\bibitem[{Veit et~al.(2017{\natexlab{b}})Veit, Belongie, and
  Karaletsos}]{veit2016conditional}
Andreas Veit, Serge Belongie, and Theofanis Karaletsos. 2017{\natexlab{b}}.
\newblock {Conditional Similarity Networks}.
\newblock In \emph{Computer Vision and Pattern Recognition (CVPR)}.

\bibitem[{Wallace et~al.(2016)Wallace, Kuiper, Sharma, Zhu, and
  Marshall}]{wallace2016extracting}
Byron~C Wallace, Jo{\"{e}}l Kuiper, Aakash Sharma, Mingxi~Brian Zhu, and Iain~J
  Marshall. 2016.
\newblock {Extracting PICO sentences from clinical trial reports using
  supervised distant supervision}.
\newblock \emph{Journal of Machine Learning Research}, 17(132).

\bibitem[{Wallace et~al.(2014)Wallace, Paul, Sarkar, Trikalinos, and
  Dredze}]{wallace2014large}
Byron~C Wallace, Michael~J Paul, Urmimala Sarkar, Thomas~A Trikalinos, and Mark
  Dredze. 2014.
\newblock {A large-scale quantitative analysis of latent factors and sentiment
  in online doctor reviews}.
\newblock \emph{Journal of the American Medical Informatics Association},
  21(6).

\bibitem[{Wallace et~al.(2010)Wallace, Trikalinos, Lau, Brodley, and
  Schmid}]{wallace2010semi}
Byron~C Wallace, Thomas~A Trikalinos, Joseph Lau, Carla Brodley, and
  Christopher~H Schmid. 2010.
\newblock {Semi-automated screening of biomedical citations for systematic
  reviews}.
\newblock \emph{BMC bioinformatics}, 11(1).

\bibitem[{Whitney(2016)}]{whitney2016disentangled}
William Whitney. 2016.
\newblock {Disentangled Representations in Neural Models}.
\newblock \emph{arXiv preprint arXiv:1602.02383}.

\bibitem[{Wilber et~al.(2014)Wilber, Kwak, and Belongie}]{wilber2014cost}
Michael~J Wilber, Iljung~S Kwak, and Serge~J Belongie. 2014.
\newblock {Cost-effective hits for relative similarity comparisons}.
\newblock In \emph{Second AAAI Conference on Human Computation and
  Crowdsourcing}.

\bibitem[{Zhang et~al.(2017)Zhang, Barzilay, and Jaakkola}]{zhang2017aspect}
Yuan Zhang, Regina Barzilay, and Tommi Jaakkola. 2017.
\newblock {Aspect-augmented Adversarial Networks for Domain Adaptation}.
\newblock In \emph{Transactions of the Association for Computational
  Linguistics}.

\end{thebibliography}
\bibliographystyle{acl_natbib_nourl}

\appendix
\section{Example of PICO Triplet generation}
As an illustrative example, we walk through construction of a single triplet for the PICO domain in detail. Recall that we first randomly draw two reviews, $R_1$ and $R_2$. In this case, $R_1$ consists of studies involving nocturnal enuresis and review $R2$ concerns asthma. From $R_1$ we randomly select two studies $S, S'$. 

(1) Here we sample abstract $A$ for study $S$, shown below:

{\footnotesize \begin{quote}
In recent years the treatment of primary nocturnal enuresis (PNE) with desmopressin (DDAVP) has been promising. The route of administration until now had been intranasal, but because the tablets were introduced for the treatment of diabetes insipidus they have also become available for the treatment of PNE. To find the optimal dosage of desmopressin tablets and to compare desmopressin's efficacy with placebo in a group of adolescents with severe monosymptomatic enuresis. The long-term safety of desmopressin was also studied in the same group of patients. The effect of oral desmopressin (1-deamino-8-D-arginine-vasopressin) (DDAVP tablets, Minirin) was investigated in 25 adolescents (ages 11 to 21 years) with severe monosymptomatic nocturnal enuresis. The first part of the dose-ranging study comprised a single-blind dose titration period, followed by a double-blind, crossover efficacy period comparing desmopressin with placebo. The final part was an open long-term study consisting of two 12-week treatment periods. The efficacy of the drug was measured in reductions of the number of wet nights per week. During the first dose-titration period, the majority of the patients were given desmopressin 400 micrograms, and the number of wet nights decreased from a mean of 4.9 to 2.8. During the double-blind period, a significant reduction of wet nights was observed (1.8 vs 4.1 for placebo). During the two long-term periods, 48\% and 53\% of the patients could be classified as responders (0 to 1 wet night per week) and 22\% and 23.5\% as intermediate responders (2 to 3 wet nights per week). No weight gain was observed due to water retention. After cessation of the drug, 44\% of the patients had a significant decrease in the number of wet nights. Oral desmopressin has a clinically significant effect on patients with PNE, and therapy is safe when administered as long-term treatment. \end{quote}}

For study $S$, the summaries in the CDSR are as follow. First the P summary $(s_P)$ ,

{\footnotesize \begin{quote} Number of children: 10 Inclusion criteria: adolescents (puberty stage at least 2, at least 12 years)  Exclusion criteria: treatment in previous 2 weeks, daytime wetting, UTI, urinary tract abnormalities  Previous treatment: failed using alarms, desmopressin, other drugs Age range 11-21, median 13 years Baseline wetting 4.7 (SD 1.1) wet nights/week Department of Paediatric Surgery, Sweden\end{quote}}

\noindent I Summary $(s_I)$: 

{\footnotesize \begin{quote} A : desmopressin orally (dosage based on titration period) B : placebo Duration 4 weeks each
\end{quote}}

\noindent The O summary $(s_O)$

{\footnotesize \begin{quote} Wet nights during trial (number, mean, SD): A: 10, 1.8 (SD 1.4); B: 10, 4.1 (1.5) Side effects: headache (5); abdominal pain (6); nausea and vertigo (1) All resolved while treatment continued \end{quote}}

\vspace{1em}
(2) From study $S'$, the summaries in the CDSR are reproduced as follows. First, the P Summary $(s'_P)$:

{\footnotesize \begin{quote} Number of children: 135 Dropouts: 23 excluded for non-compliance, and 39 lost to follow up including 12 failed with alarms Inclusion criteria: monosymptomatic nocturnal enuresis, age $>5$ years Exclusion criteria: previous treatment with DDAVP or alarm, urological pathology, diurnal enuresis, UTI Age, mean years: 11.2  Baseline wetting: A 21\% dry nights, B 14\% dry nights\end{quote}}

\noindent I Summary $(s'_I)$: 

{\footnotesize \begin{quote} A (62): desmopressin 20 g intranasally increasing to 40 g if response partial B (73): alarm (pad-and-bell) Duration of treatment 3 months. If failed at that time, changed to alternative arm\end{quote}}

\noindent The O summary $(s'_O)$

{\footnotesize \begin{quote} DRY nights at 3 months: A 85\%; B: 90\% Number not achieving 14 dry nights: A 12/39; B: 6/37 Side effects: not mentioned \end{quote}}

\vspace{1em}
(3) From review $R_2$, we sample one study $S''$. Matched summaries in the CDSR are as follows. 

\noindent P Summary $(s''_P)$: 

{\footnotesize \begin{quote} n = 8 Mean age = 52 Inclusion: intrinsic asthma, constant reversibility $>$ 20\% None had an acute exacerbation at time of study Exclusion: none listed \end{quote}}

\noindent I summary $(s''_I)$:

{\footnotesize \begin{quote} \#1: Atenolol 100 mg Metoprolol 100 mg Placebo \#2: Terbutaline (IV then inhaled) after Tx or placebo \end{quote}}

\noindent O summary $(s''_O)$: 

{\footnotesize \begin{quote} FEV1 Symptoms \end{quote}}

\vspace{1em}

(4) We note how the summaries for $S$ and $S'$ are similar to each other (but not identical) since they belong in the same review whereas they are quite different from summaries for $S''$ which belongs in a different review. Now we construct the triplet $(s, d, o)_P$ as follows:

\vspace{1em}
\noindent $s = s'_P$:

{\footnotesize \begin{quote} Number of children: 135 Dropouts: 23 excluded for non-compliance, and 39 lost to follow up including 12 failed with alarms Inclusion criteria: monosymptomatic nocturnal enuresis, age $>5$ years Exclusion criteria: previous treatment with DDAVP or alarm, urological pathology, diurnal enuresis, UTI Age, mean years: 11.2  Baseline wetting: A 21\% dry nights, B 14\% dry nights\end{quote}}

\noindent So we have $d$ : A and $o = [s''_P|s'_I|s'_O]$: 

{\footnotesize \begin{quote} n = 8 Mean age = 52 Inclusion: intrinsic asthma, constant reversibility $>$ 20\% None had an acute exacerbation at time of study Exclusion: none listed | A (62): desmopressin 20 g intranasally increasing to 40 g if response partial B (73): alarm (pad-and-bell) Duration of treatment 3 months. If failed at that time, changed to alternative arm | DRY nights at 3 months: A 85\%; B: 90\% Number not achieving 14 dry nights: A 12/39; B: 6/37 Side effects: not mentioned \end{quote}}

\section{Implementation Details \& Baseline Hyperparameters}
All tokenisation has been done using default spaCy\footnote{\url{https://spacy.io/}} tokenizer. 

\subsection{PICO Domain}

\subsubsection{Baselines} For the TF-IDF baseline, we use the implementation in scikit-learn\footnote{\url{http://scikit-learn.org/}}. The TF-IDF was fit on all CDSR data. The resulting transformation was applied to Cohen corpus. We use cosine similarity to evaluate TF-IDF model.

For Doc2Vec, we used the gensim implementation\footnote{\url{https://radimrehurek.com/gensim/}} with an 800d embedding size, and a window size of 10. These parameters were selected via random search over validation data. We otherwise used the default settings in gensim. The vocabulary used was the same as above. 

For LDA, we again used the scikit-learn implementation, setting the number of topics to 7; this was selected via line search over the validation set across the range 1 to 50. 

For the NVDM baseline, we used hidden layer comprising 500 dimensions and an output embedding dimension of 200. We used Tanh activation functions. The model was trained using the Adam optimizer, using a learning rate of 5e-5. We performed early stopping based on validation perplexity. 

For the ABAE model, we use the same hyperparameters as reported in the original He \emph{et al}. (2017) paper. We experimented with 20, 50 and 100 aspects. Of these, 100 performed best, which is what we used to generate the reported results. 

For the DSSM model, we used 300d filters with filter windows of sizes (1, 6) and output embedding dimension of 256. We used tanh activation for both the convolution and output layers, and imposed l2 regularization on weights ($1e$-$5$). The model was trained using the Adam optimizer with early stopping on validation loss. 

\subsection{Beer Advocate Domain}
\subsubsection{Baselines} For the TF-IDF baseline, we used the scikit-learn implementation. TF-IDF parameters were fit on the available training set. We used cosine similarity for evaluation.

For Doc2Vec, we used the same method as above, here setting (after line search) the embedding size to 800d and window size to 7. 

For the LDA baseline, we followed the same strategy as above, arriving at 4 topics. 

For NVDM, we set the number of dimensions in the hidden layer to 500, and set the embedding dimension to 145. As an activation for the hidden layer we used Tanh. Again the model was trained using the Adam optimizer, with a learning rate of 5e-5 and early stopping based on validation perplexity. 

For the ABAE baseline, we use the same hyperparameters as in He \emph{et al}. (2017) paper, since BeerAdvocate was one of the evaluation datasets used in said work.

For NVDM + Triplet Loss, we used same settings as reported above for NVDM, with loss weighting of 0.001. 

For the DSSM baseline, we used 200d filters with filter windows of sizes (2, 4) and output embedding dimension of 256. We used tanh activations for both convolution and output layer and l2 regularization on weights of $1e$-$6$. The model was trained using Adam optimizer with early stopping on validation loss.

\subsection{Yelp! and TripAdvisor Domain}

\subsubsection{Baselines}
To induce the baseline TF-IDF vectors,  we used the scikit-learn implementation. TF-IDF parameters were fit on the available training set. Our vocabulary comprised those words that appeared in at least 5 unique reviews, which corresponded to 10K words. We used cosine similarity for evaluation. 

For Doc2Vec, we set the embedding size to 800d and the window size to 7. For LDA, we again used the scikit-learn implementation, setting the number of topics to 10. 

For NVDM, we set the number of dimensions in the hidden layer to 500, and set the embedding dimension to 200. As an activation for the hidden layer we used Tanh. Again the model was trained using the Adam optimizer, with a learning rate of $5e$-$5$ and early stopping based on validation perplexity. 

For ABAE, we use the same hyperparameters in He \emph{et al}. (2017), since Yelp was another one of the evaluation datasets used in their work.

For NVDM + Triplet Loss, we used same settings as above for NVDM, and again used loss weighting of 0.001. 

For the DSSM baseline, we used 200d filters with filter windows of sizes (1, 4) and output embedding dimension of 256. We used tanh activations for both convolution and output layer and l2 regularization on weights of $1e$-$6$. The model was trained using the Adam optimizer with early stopping on validation loss.

\section{Yelp!/TripAdvisor Dataset Word Activations}
\begin{table}[h]
\centering
\begin{tabularx}{\columnwidth}{X}
\textbf{Domain} \emph{celebrity}, \emph{cypress}, \emph{free}, \emph{inadequate}, \emph{kitchens}, \emph{maroma}, \emph{maya}, \emph{scratchy}, \emph{suburban}, \emph{supply} \\\hline 
\textbf{Sentiment} \emph{awful}, \emph{disgusting}, \emph{horrible}, \emph{mediocre}, \emph{poor}, \emph{rude}, \emph{sells}, \emph{shame}, \emph{unacceptable}, \emph{worst} \\ 
\end{tabularx}
\caption{Most activated words for hotels/restaurant data.}
\label{table:foodmost}
\end{table}

We did not have sufficient space to include the most activated words per aspect (inferred via the gating mechanism) for the Yelp!/TripAdvisor corpus in the manuscript (as we did for the other domains) and so we include them here. 

\section{Highlighted Text}

To visualize the output of gating mechanism $g \in R^{N \times 1}$, we perform the following transformations:
\begin{enumerate}
\item Since the gating is applied on the top convolution layer, there is no one-to-one correspondence between words and values in $g$. Hence, we convolved the output a 5 length mean filter to smooth the gating output.
\item We normalize the range of values in $g$ to $[0, 1]$ using the following equation:
\begin{equation}
g = \frac{g - min(g)}{max(g) - min(g)}
\end{equation}
\end{enumerate}

In the following subsections, we provide 3 examples from each of our datasets highlighted with corresponding gating output.
\subsection{PICO Domain}
Color Legend : \setlength{\fboxsep}{0pt}\def\cbRGB{\colorbox[RGB]}\expandafter\cbRGB\expandafter{\detokenize{255,0,0}}{Population\strut} 
\setlength{\fboxsep}{0pt}\def\cbRGB{\colorbox[RGB]}\expandafter\cbRGB\expandafter{\detokenize{0,255,0}}{Intervention\strut} 
\setlength{\fboxsep}{0pt}\def\cbRGB{\colorbox[RGB]}\expandafter\cbRGB\expandafter{\detokenize{0,0,255}}{Outcome\strut} 

\small
\par
\textbf{Example 1}
\par
\setlength{\fboxsep}{0pt}\def\cbRGB{\colorbox[RGB]}\expandafter\cbRGB\expandafter{\detokenize{255,254,254}}{determined\strut} \setlength{\fboxsep}{0pt}\def\cbRGB{\colorbox[RGB]}\expandafter\cbRGB\expandafter{\detokenize{255,246,246}}{the\strut} \setlength{\fboxsep}{0pt}\def\cbRGB{\colorbox[RGB]}\expandafter\cbRGB\expandafter{\detokenize{255,245,245}}{efficacy\strut} \setlength{\fboxsep}{0pt}\def\cbRGB{\colorbox[RGB]}\expandafter\cbRGB\expandafter{\detokenize{255,246,246}}{of\strut} \setlength{\fboxsep}{0pt}\def\cbRGB{\colorbox[RGB]}\expandafter\cbRGB\expandafter{\detokenize{255,247,247}}{unk\strut} \setlength{\fboxsep}{0pt}\def\cbRGB{\colorbox[RGB]}\expandafter\cbRGB\expandafter{\detokenize{255,248,248}}{(\strut} \setlength{\fboxsep}{0pt}\def\cbRGB{\colorbox[RGB]}\expandafter\cbRGB\expandafter{\detokenize{255,248,248}}{unk\strut} \setlength{\fboxsep}{0pt}\def\cbRGB{\colorbox[RGB]}\expandafter\cbRGB\expandafter{\detokenize{255,248,248}}{)\strut} \setlength{\fboxsep}{0pt}\def\cbRGB{\colorbox[RGB]}\expandafter\cbRGB\expandafter{\detokenize{255,247,247}}{in\strut} \setlength{\fboxsep}{0pt}\def\cbRGB{\colorbox[RGB]}\expandafter\cbRGB\expandafter{\detokenize{255,246,246}}{a\strut} \setlength{\fboxsep}{0pt}\def\cbRGB{\colorbox[RGB]}\expandafter\cbRGB\expandafter{\detokenize{255,244,244}}{clinical\strut} \setlength{\fboxsep}{0pt}\def\cbRGB{\colorbox[RGB]}\expandafter\cbRGB\expandafter{\detokenize{255,244,244}}{population\strut} \setlength{\fboxsep}{0pt}\def\cbRGB{\colorbox[RGB]}\expandafter\cbRGB\expandafter{\detokenize{255,242,242}}{of\strut} \setlength{\fboxsep}{0pt}\def\cbRGB{\colorbox[RGB]}\expandafter\cbRGB\expandafter{\detokenize{255,225,225}}{aggressive\strut} \setlength{\fboxsep}{0pt}\def\cbRGB{\colorbox[RGB]}\expandafter\cbRGB\expandafter{\detokenize{255,190,190}}{,\strut} \setlength{\fboxsep}{0pt}\def\cbRGB{\colorbox[RGB]}\expandafter\cbRGB\expandafter{\detokenize{255,145,145}}{urban\strut} \setlength{\fboxsep}{0pt}\def\cbRGB{\colorbox[RGB]}\expandafter\cbRGB\expandafter{\detokenize{255,63,63}}{children\strut} \setlength{\fboxsep}{0pt}\def\cbRGB{\colorbox[RGB]}\expandafter\cbRGB\expandafter{\detokenize{255,8,8}}{diagnosed\strut} \setlength{\fboxsep}{0pt}\def\cbRGB{\colorbox[RGB]}\expandafter\cbRGB\expandafter{\detokenize{255,16,16}}{with\strut} \setlength{\fboxsep}{0pt}\def\cbRGB{\colorbox[RGB]}\expandafter\cbRGB\expandafter{\detokenize{255,99,99}}{attention\strut} \setlength{\fboxsep}{0pt}\def\cbRGB{\colorbox[RGB]}\expandafter\cbRGB\expandafter{\detokenize{255,181,181}}{deficit\strut} \setlength{\fboxsep}{0pt}\def\cbRGB{\colorbox[RGB]}\expandafter\cbRGB\expandafter{\detokenize{255,209,209}}{unk\strut} \setlength{\fboxsep}{0pt}\def\cbRGB{\colorbox[RGB]}\expandafter\cbRGB\expandafter{\detokenize{255,209,209}}{disorder\strut} \setlength{\fboxsep}{0pt}\def\cbRGB{\colorbox[RGB]}\expandafter\cbRGB\expandafter{\detokenize{255,222,222}}{(\strut} \setlength{\fboxsep}{0pt}\def\cbRGB{\colorbox[RGB]}\expandafter\cbRGB\expandafter{\detokenize{255,229,229}}{adhd\strut} \setlength{\fboxsep}{0pt}\def\cbRGB{\colorbox[RGB]}\expandafter\cbRGB\expandafter{\detokenize{255,244,244}}{)\strut} \setlength{\fboxsep}{0pt}\def\cbRGB{\colorbox[RGB]}\expandafter\cbRGB\expandafter{\detokenize{255,240,240}}{.\strut} \setlength{\fboxsep}{0pt}\def\cbRGB{\colorbox[RGB]}\expandafter\cbRGB\expandafter{\detokenize{255,246,246}}{in\strut} \setlength{\fboxsep}{0pt}\def\cbRGB{\colorbox[RGB]}\expandafter\cbRGB\expandafter{\detokenize{255,246,246}}{previous\strut} \setlength{\fboxsep}{0pt}\def\cbRGB{\colorbox[RGB]}\expandafter\cbRGB\expandafter{\detokenize{255,239,239}}{studies\strut} \setlength{\fboxsep}{0pt}\def\cbRGB{\colorbox[RGB]}\expandafter\cbRGB\expandafter{\detokenize{255,220,220}}{of\strut} \setlength{\fboxsep}{0pt}\def\cbRGB{\colorbox[RGB]}\expandafter\cbRGB\expandafter{\detokenize{255,160,160}}{unk\strut} \setlength{\fboxsep}{0pt}\def\cbRGB{\colorbox[RGB]}\expandafter\cbRGB\expandafter{\detokenize{255,89,89}}{children\strut} \setlength{\fboxsep}{0pt}\def\cbRGB{\colorbox[RGB]}\expandafter\cbRGB\expandafter{\detokenize{255,97,97}}{with\strut} \setlength{\fboxsep}{0pt}\def\cbRGB{\colorbox[RGB]}\expandafter\cbRGB\expandafter{\detokenize{255,119,119}}{adhd\strut} \setlength{\fboxsep}{0pt}\def\cbRGB{\colorbox[RGB]}\expandafter\cbRGB\expandafter{\detokenize{255,191,191}}{,\strut} \setlength{\fboxsep}{0pt}\def\cbRGB{\colorbox[RGB]}\expandafter\cbRGB\expandafter{\detokenize{255,204,204}}{unk\strut} \setlength{\fboxsep}{0pt}\def\cbRGB{\colorbox[RGB]}\expandafter\cbRGB\expandafter{\detokenize{255,242,242}}{has\strut} \setlength{\fboxsep}{0pt}\def\cbRGB{\colorbox[RGB]}\expandafter\cbRGB\expandafter{\detokenize{255,247,247}}{been\strut} \setlength{\fboxsep}{0pt}\def\cbRGB{\colorbox[RGB]}\expandafter\cbRGB\expandafter{\detokenize{255,244,244}}{shown\strut} \setlength{\fboxsep}{0pt}\def\cbRGB{\colorbox[RGB]}\expandafter\cbRGB\expandafter{\detokenize{255,245,245}}{to\strut} \setlength{\fboxsep}{0pt}\def\cbRGB{\colorbox[RGB]}\expandafter\cbRGB\expandafter{\detokenize{255,246,246}}{be\strut} \setlength{\fboxsep}{0pt}\def\cbRGB{\colorbox[RGB]}\expandafter\cbRGB\expandafter{\detokenize{255,243,243}}{effective\strut} \setlength{\fboxsep}{0pt}\def\cbRGB{\colorbox[RGB]}\expandafter\cbRGB\expandafter{\detokenize{255,239,239}}{when\strut} \setlength{\fboxsep}{0pt}\def\cbRGB{\colorbox[RGB]}\expandafter\cbRGB\expandafter{\detokenize{255,236,236}}{compared\strut} \setlength{\fboxsep}{0pt}\def\cbRGB{\colorbox[RGB]}\expandafter\cbRGB\expandafter{\detokenize{255,233,233}}{with\strut} \setlength{\fboxsep}{0pt}\def\cbRGB{\colorbox[RGB]}\expandafter\cbRGB\expandafter{\detokenize{255,232,232}}{placebo\strut} \setlength{\fboxsep}{0pt}\def\cbRGB{\colorbox[RGB]}\expandafter\cbRGB\expandafter{\detokenize{255,230,230}}{.\strut} \setlength{\fboxsep}{0pt}\def\cbRGB{\colorbox[RGB]}\expandafter\cbRGB\expandafter{\detokenize{255,228,228}}{eighteen\strut} \setlength{\fboxsep}{0pt}\def\cbRGB{\colorbox[RGB]}\expandafter\cbRGB\expandafter{\detokenize{255,229,229}}{unk\strut} \setlength{\fboxsep}{0pt}\def\cbRGB{\colorbox[RGB]}\expandafter\cbRGB\expandafter{\detokenize{255,220,220}}{children\strut} \setlength{\fboxsep}{0pt}\def\cbRGB{\colorbox[RGB]}\expandafter\cbRGB\expandafter{\detokenize{255,223,223}}{(\strut} \setlength{\fboxsep}{0pt}\def\cbRGB{\colorbox[RGB]}\expandafter\cbRGB\expandafter{\detokenize{255,223,223}}{ages\strut} \setlength{\fboxsep}{0pt}\def\cbRGB{\colorbox[RGB]}\expandafter\cbRGB\expandafter{\detokenize{255,234,234}}{qqq\strut} \setlength{\fboxsep}{0pt}\def\cbRGB{\colorbox[RGB]}\expandafter\cbRGB\expandafter{\detokenize{255,238,238}}{to\strut} \setlength{\fboxsep}{0pt}\def\cbRGB{\colorbox[RGB]}\expandafter\cbRGB\expandafter{\detokenize{255,244,244}}{qqq\strut} \setlength{\fboxsep}{0pt}\def\cbRGB{\colorbox[RGB]}\expandafter\cbRGB\expandafter{\detokenize{255,244,244}}{years\strut} \setlength{\fboxsep}{0pt}\def\cbRGB{\colorbox[RGB]}\expandafter\cbRGB\expandafter{\detokenize{255,224,224}}{)\strut} \setlength{\fboxsep}{0pt}\def\cbRGB{\colorbox[RGB]}\expandafter\cbRGB\expandafter{\detokenize{255,128,128}}{,\strut} \setlength{\fboxsep}{0pt}\def\cbRGB{\colorbox[RGB]}\expandafter\cbRGB\expandafter{\detokenize{255,27,27}}{diagnosed\strut} \setlength{\fboxsep}{0pt}\def\cbRGB{\colorbox[RGB]}\expandafter\cbRGB\expandafter{\detokenize{255,0,0}}{with\strut} \setlength{\fboxsep}{0pt}\def\cbRGB{\colorbox[RGB]}\expandafter\cbRGB\expandafter{\detokenize{255,15,15}}{adhd\strut} \setlength{\fboxsep}{0pt}\def\cbRGB{\colorbox[RGB]}\expandafter\cbRGB\expandafter{\detokenize{255,110,110}}{and\strut} \setlength{\fboxsep}{0pt}\def\cbRGB{\colorbox[RGB]}\expandafter\cbRGB\expandafter{\detokenize{255,158,158}}{attending\strut} \setlength{\fboxsep}{0pt}\def\cbRGB{\colorbox[RGB]}\expandafter\cbRGB\expandafter{\detokenize{255,234,234}}{a\strut} \setlength{\fboxsep}{0pt}\def\cbRGB{\colorbox[RGB]}\expandafter\cbRGB\expandafter{\detokenize{255,242,242}}{unk\strut} \setlength{\fboxsep}{0pt}\def\cbRGB{\colorbox[RGB]}\expandafter\cbRGB\expandafter{\detokenize{255,244,244}}{treatment\strut} \setlength{\fboxsep}{0pt}\def\cbRGB{\colorbox[RGB]}\expandafter\cbRGB\expandafter{\detokenize{255,240,240}}{program\strut} \setlength{\fboxsep}{0pt}\def\cbRGB{\colorbox[RGB]}\expandafter\cbRGB\expandafter{\detokenize{255,236,236}}{for\strut} \setlength{\fboxsep}{0pt}\def\cbRGB{\colorbox[RGB]}\expandafter\cbRGB\expandafter{\detokenize{255,229,229}}{youth\strut} \setlength{\fboxsep}{0pt}\def\cbRGB{\colorbox[RGB]}\expandafter\cbRGB\expandafter{\detokenize{255,226,226}}{with\strut} \setlength{\fboxsep}{0pt}\def\cbRGB{\colorbox[RGB]}\expandafter\cbRGB\expandafter{\detokenize{255,225,225}}{unk\strut} \setlength{\fboxsep}{0pt}\def\cbRGB{\colorbox[RGB]}\expandafter\cbRGB\expandafter{\detokenize{255,233,233}}{behavior\strut} \setlength{\fboxsep}{0pt}\def\cbRGB{\colorbox[RGB]}\expandafter\cbRGB\expandafter{\detokenize{255,241,241}}{disorders\strut} \setlength{\fboxsep}{0pt}\def\cbRGB{\colorbox[RGB]}\expandafter\cbRGB\expandafter{\detokenize{255,245,245}}{,\strut} \setlength{\fboxsep}{0pt}\def\cbRGB{\colorbox[RGB]}\expandafter\cbRGB\expandafter{\detokenize{255,244,244}}{participated\strut} \setlength{\fboxsep}{0pt}\def\cbRGB{\colorbox[RGB]}\expandafter\cbRGB\expandafter{\detokenize{255,245,245}}{in\strut} \setlength{\fboxsep}{0pt}\def\cbRGB{\colorbox[RGB]}\expandafter\cbRGB\expandafter{\detokenize{255,246,246}}{a\strut} \setlength{\fboxsep}{0pt}\def\cbRGB{\colorbox[RGB]}\expandafter\cbRGB\expandafter{\detokenize{255,243,243}}{double-blind\strut} \setlength{\fboxsep}{0pt}\def\cbRGB{\colorbox[RGB]}\expandafter\cbRGB\expandafter{\detokenize{255,241,241}}{placebo\strut} \setlength{\fboxsep}{0pt}\def\cbRGB{\colorbox[RGB]}\expandafter\cbRGB\expandafter{\detokenize{255,243,243}}{trial\strut} \setlength{\fboxsep}{0pt}\def\cbRGB{\colorbox[RGB]}\expandafter\cbRGB\expandafter{\detokenize{255,247,247}}{with\strut} \setlength{\fboxsep}{0pt}\def\cbRGB{\colorbox[RGB]}\expandafter\cbRGB\expandafter{\detokenize{255,249,249}}{assessment\strut} \setlength{\fboxsep}{0pt}\def\cbRGB{\colorbox[RGB]}\expandafter\cbRGB\expandafter{\detokenize{255,250,250}}{data\strut} \setlength{\fboxsep}{0pt}\def\cbRGB{\colorbox[RGB]}\expandafter\cbRGB\expandafter{\detokenize{255,252,252}}{obtained\strut} \setlength{\fboxsep}{0pt}\def\cbRGB{\colorbox[RGB]}\expandafter\cbRGB\expandafter{\detokenize{255,252,252}}{from\strut} \setlength{\fboxsep}{0pt}\def\cbRGB{\colorbox[RGB]}\expandafter\cbRGB\expandafter{\detokenize{255,248,248}}{staff\strut} \setlength{\fboxsep}{0pt}\def\cbRGB{\colorbox[RGB]}\expandafter\cbRGB\expandafter{\detokenize{255,246,246}}{in\strut} \setlength{\fboxsep}{0pt}\def\cbRGB{\colorbox[RGB]}\expandafter\cbRGB\expandafter{\detokenize{255,246,246}}{the\strut} \setlength{\fboxsep}{0pt}\def\cbRGB{\colorbox[RGB]}\expandafter\cbRGB\expandafter{\detokenize{255,246,246}}{program\strut} \setlength{\fboxsep}{0pt}\def\cbRGB{\colorbox[RGB]}\expandafter\cbRGB\expandafter{\detokenize{255,248,248}}{and\strut} \setlength{\fboxsep}{0pt}\def\cbRGB{\colorbox[RGB]}\expandafter\cbRGB\expandafter{\detokenize{255,249,249}}{parents\strut} \setlength{\fboxsep}{0pt}\def\cbRGB{\colorbox[RGB]}\expandafter\cbRGB\expandafter{\detokenize{255,250,250}}{at\strut} \setlength{\fboxsep}{0pt}\def\cbRGB{\colorbox[RGB]}\expandafter\cbRGB\expandafter{\detokenize{255,250,250}}{home\strut} \setlength{\fboxsep}{0pt}\def\cbRGB{\colorbox[RGB]}\expandafter\cbRGB\expandafter{\detokenize{255,250,250}}{.\strut} \setlength{\fboxsep}{0pt}\def\cbRGB{\colorbox[RGB]}\expandafter\cbRGB\expandafter{\detokenize{255,251,251}}{based\strut} \setlength{\fboxsep}{0pt}\def\cbRGB{\colorbox[RGB]}\expandafter\cbRGB\expandafter{\detokenize{255,250,250}}{on\strut} \setlength{\fboxsep}{0pt}\def\cbRGB{\colorbox[RGB]}\expandafter\cbRGB\expandafter{\detokenize{255,250,250}}{staff\strut} \setlength{\fboxsep}{0pt}\def\cbRGB{\colorbox[RGB]}\expandafter\cbRGB\expandafter{\detokenize{255,243,243}}{ratings\strut} \setlength{\fboxsep}{0pt}\def\cbRGB{\colorbox[RGB]}\expandafter\cbRGB\expandafter{\detokenize{255,235,235}}{of\strut} \setlength{\fboxsep}{0pt}\def\cbRGB{\colorbox[RGB]}\expandafter\cbRGB\expandafter{\detokenize{255,232,232}}{the\strut} \setlength{\fboxsep}{0pt}\def\cbRGB{\colorbox[RGB]}\expandafter\cbRGB\expandafter{\detokenize{255,230,230}}{children\strut} \setlength{\fboxsep}{0pt}\def\cbRGB{\colorbox[RGB]}\expandafter\cbRGB\expandafter{\detokenize{255,233,233}}{'s\strut} \setlength{\fboxsep}{0pt}\def\cbRGB{\colorbox[RGB]}\expandafter\cbRGB\expandafter{\detokenize{255,232,232}}{behavior\strut} \setlength{\fboxsep}{0pt}\def\cbRGB{\colorbox[RGB]}\expandafter\cbRGB\expandafter{\detokenize{255,239,239}}{in\strut} \setlength{\fboxsep}{0pt}\def\cbRGB{\colorbox[RGB]}\expandafter\cbRGB\expandafter{\detokenize{255,245,245}}{the\strut} \setlength{\fboxsep}{0pt}\def\cbRGB{\colorbox[RGB]}\expandafter\cbRGB\expandafter{\detokenize{255,245,245}}{program\strut} \setlength{\fboxsep}{0pt}\def\cbRGB{\colorbox[RGB]}\expandafter\cbRGB\expandafter{\detokenize{255,243,243}}{and\strut} \setlength{\fboxsep}{0pt}\def\cbRGB{\colorbox[RGB]}\expandafter\cbRGB\expandafter{\detokenize{255,244,244}}{an\strut} \setlength{\fboxsep}{0pt}\def\cbRGB{\colorbox[RGB]}\expandafter\cbRGB\expandafter{\detokenize{255,246,246}}{academic\strut} \setlength{\fboxsep}{0pt}\def\cbRGB{\colorbox[RGB]}\expandafter\cbRGB\expandafter{\detokenize{255,242,242}}{classroom\strut} \setlength{\fboxsep}{0pt}\def\cbRGB{\colorbox[RGB]}\expandafter\cbRGB\expandafter{\detokenize{255,234,234}}{,\strut} \setlength{\fboxsep}{0pt}\def\cbRGB{\colorbox[RGB]}\expandafter\cbRGB\expandafter{\detokenize{255,232,232}}{the\strut} \setlength{\fboxsep}{0pt}\def\cbRGB{\colorbox[RGB]}\expandafter\cbRGB\expandafter{\detokenize{255,225,225}}{children\strut} \setlength{\fboxsep}{0pt}\def\cbRGB{\colorbox[RGB]}\expandafter\cbRGB\expandafter{\detokenize{255,220,220}}{displayed\strut} \setlength{\fboxsep}{0pt}\def\cbRGB{\colorbox[RGB]}\expandafter\cbRGB\expandafter{\detokenize{255,194,194}}{significant\strut} \setlength{\fboxsep}{0pt}\def\cbRGB{\colorbox[RGB]}\expandafter\cbRGB\expandafter{\detokenize{255,191,191}}{improvements\strut} \setlength{\fboxsep}{0pt}\def\cbRGB{\colorbox[RGB]}\expandafter\cbRGB\expandafter{\detokenize{255,207,207}}{in\strut} \setlength{\fboxsep}{0pt}\def\cbRGB{\colorbox[RGB]}\expandafter\cbRGB\expandafter{\detokenize{255,228,228}}{adhd\strut} \setlength{\fboxsep}{0pt}\def\cbRGB{\colorbox[RGB]}\expandafter\cbRGB\expandafter{\detokenize{255,241,241}}{symptoms\strut} \setlength{\fboxsep}{0pt}\def\cbRGB{\colorbox[RGB]}\expandafter\cbRGB\expandafter{\detokenize{255,237,237}}{and\strut} \setlength{\fboxsep}{0pt}\def\cbRGB{\colorbox[RGB]}\expandafter\cbRGB\expandafter{\detokenize{255,241,241}}{aggressive\strut} \setlength{\fboxsep}{0pt}\def\cbRGB{\colorbox[RGB]}\expandafter\cbRGB\expandafter{\detokenize{255,239,239}}{behavior\strut} \setlength{\fboxsep}{0pt}\def\cbRGB{\colorbox[RGB]}\expandafter\cbRGB\expandafter{\detokenize{255,234,234}}{with\strut} \setlength{\fboxsep}{0pt}\def\cbRGB{\colorbox[RGB]}\expandafter\cbRGB\expandafter{\detokenize{255,233,233}}{unk\strut} \setlength{\fboxsep}{0pt}\def\cbRGB{\colorbox[RGB]}\expandafter\cbRGB\expandafter{\detokenize{255,239,239}}{and\strut} \setlength{\fboxsep}{0pt}\def\cbRGB{\colorbox[RGB]}\expandafter\cbRGB\expandafter{\detokenize{255,243,243}}{high-dose\strut} \setlength{\fboxsep}{0pt}\def\cbRGB{\colorbox[RGB]}\expandafter\cbRGB\expandafter{\detokenize{255,249,249}}{unk\strut} \setlength{\fboxsep}{0pt}\def\cbRGB{\colorbox[RGB]}\expandafter\cbRGB\expandafter{\detokenize{255,249,249}}{conditions\strut} \setlength{\fboxsep}{0pt}\def\cbRGB{\colorbox[RGB]}\expandafter\cbRGB\expandafter{\detokenize{255,249,249}}{.\strut} \setlength{\fboxsep}{0pt}\def\cbRGB{\colorbox[RGB]}\expandafter\cbRGB\expandafter{\detokenize{255,248,248}}{at\strut} \setlength{\fboxsep}{0pt}\def\cbRGB{\colorbox[RGB]}\expandafter\cbRGB\expandafter{\detokenize{255,248,248}}{home\strut} \setlength{\fboxsep}{0pt}\def\cbRGB{\colorbox[RGB]}\expandafter\cbRGB\expandafter{\detokenize{255,250,250}}{,\strut} \setlength{\fboxsep}{0pt}\def\cbRGB{\colorbox[RGB]}\expandafter\cbRGB\expandafter{\detokenize{255,250,250}}{parents\strut} \setlength{\fboxsep}{0pt}\def\cbRGB{\colorbox[RGB]}\expandafter\cbRGB\expandafter{\detokenize{255,249,249}}{and\strut} \setlength{\fboxsep}{0pt}\def\cbRGB{\colorbox[RGB]}\expandafter\cbRGB\expandafter{\detokenize{255,246,246}}{unk\strut} \setlength{\fboxsep}{0pt}\def\cbRGB{\colorbox[RGB]}\expandafter\cbRGB\expandafter{\detokenize{255,245,245}}{reported\strut} \setlength{\fboxsep}{0pt}\def\cbRGB{\colorbox[RGB]}\expandafter\cbRGB\expandafter{\detokenize{255,244,244}}{few\strut} \setlength{\fboxsep}{0pt}\def\cbRGB{\colorbox[RGB]}\expandafter\cbRGB\expandafter{\detokenize{255,243,243}}{significant\strut} \setlength{\fboxsep}{0pt}\def\cbRGB{\colorbox[RGB]}\expandafter\cbRGB\expandafter{\detokenize{255,242,242}}{differences\strut} \setlength{\fboxsep}{0pt}\def\cbRGB{\colorbox[RGB]}\expandafter\cbRGB\expandafter{\detokenize{255,243,243}}{between\strut} \setlength{\fboxsep}{0pt}\def\cbRGB{\colorbox[RGB]}\expandafter\cbRGB\expandafter{\detokenize{255,246,246}}{placebo\strut} \setlength{\fboxsep}{0pt}\def\cbRGB{\colorbox[RGB]}\expandafter\cbRGB\expandafter{\detokenize{255,247,247}}{and\strut} \setlength{\fboxsep}{0pt}\def\cbRGB{\colorbox[RGB]}\expandafter\cbRGB\expandafter{\detokenize{255,248,248}}{unk\strut} \setlength{\fboxsep}{0pt}\def\cbRGB{\colorbox[RGB]}\expandafter\cbRGB\expandafter{\detokenize{255,248,248}}{on\strut} \setlength{\fboxsep}{0pt}\def\cbRGB{\colorbox[RGB]}\expandafter\cbRGB\expandafter{\detokenize{255,248,248}}{behavior\strut} \setlength{\fboxsep}{0pt}\def\cbRGB{\colorbox[RGB]}\expandafter\cbRGB\expandafter{\detokenize{255,246,246}}{ratings\strut} \setlength{\fboxsep}{0pt}\def\cbRGB{\colorbox[RGB]}\expandafter\cbRGB\expandafter{\detokenize{255,243,243}}{.\strut} \setlength{\fboxsep}{0pt}\def\cbRGB{\colorbox[RGB]}\expandafter\cbRGB\expandafter{\detokenize{255,240,240}}{in\strut} \setlength{\fboxsep}{0pt}\def\cbRGB{\colorbox[RGB]}\expandafter\cbRGB\expandafter{\detokenize{255,243,243}}{both\strut} \setlength{\fboxsep}{0pt}\def\cbRGB{\colorbox[RGB]}\expandafter\cbRGB\expandafter{\detokenize{255,243,243}}{settings\strut} \setlength{\fboxsep}{0pt}\def\cbRGB{\colorbox[RGB]}\expandafter\cbRGB\expandafter{\detokenize{255,242,242}}{,\strut} \setlength{\fboxsep}{0pt}\def\cbRGB{\colorbox[RGB]}\expandafter\cbRGB\expandafter{\detokenize{255,239,239}}{unk\strut} \setlength{\fboxsep}{0pt}\def\cbRGB{\colorbox[RGB]}\expandafter\cbRGB\expandafter{\detokenize{255,238,238}}{was\strut} \setlength{\fboxsep}{0pt}\def\cbRGB{\colorbox[RGB]}\expandafter\cbRGB\expandafter{\detokenize{255,234,234}}{well\strut} \setlength{\fboxsep}{0pt}\def\cbRGB{\colorbox[RGB]}\expandafter\cbRGB\expandafter{\detokenize{255,228,228}}{tolerated\strut} \setlength{\fboxsep}{0pt}\def\cbRGB{\colorbox[RGB]}\expandafter\cbRGB\expandafter{\detokenize{255,227,227}}{with\strut} \setlength{\fboxsep}{0pt}\def\cbRGB{\colorbox[RGB]}\expandafter\cbRGB\expandafter{\detokenize{255,227,227}}{few\strut} \setlength{\fboxsep}{0pt}\def\cbRGB{\colorbox[RGB]}\expandafter\cbRGB\expandafter{\detokenize{255,232,232}}{side\strut} \setlength{\fboxsep}{0pt}\def\cbRGB{\colorbox[RGB]}\expandafter\cbRGB\expandafter{\detokenize{255,237,237}}{effects\strut} \setlength{\fboxsep}{0pt}\def\cbRGB{\colorbox[RGB]}\expandafter\cbRGB\expandafter{\detokenize{255,241,241}}{found\strut} \setlength{\fboxsep}{0pt}\def\cbRGB{\colorbox[RGB]}\expandafter\cbRGB\expandafter{\detokenize{255,243,243}}{during\strut} \setlength{\fboxsep}{0pt}\def\cbRGB{\colorbox[RGB]}\expandafter\cbRGB\expandafter{\detokenize{255,240,240}}{active\strut} \setlength{\fboxsep}{0pt}\def\cbRGB{\colorbox[RGB]}\expandafter\cbRGB\expandafter{\detokenize{255,242,242}}{drug\strut} \setlength{\fboxsep}{0pt}\def\cbRGB{\colorbox[RGB]}\expandafter\cbRGB\expandafter{\detokenize{255,244,244}}{conditions\strut} \setlength{\fboxsep}{0pt}\def\cbRGB{\colorbox[RGB]}\expandafter\cbRGB\expandafter{\detokenize{255,255,255}}{.\strut} 

\setlength{\fboxsep}{0pt}\def\cbRGB{\colorbox[RGB]}\expandafter\cbRGB\expandafter{\detokenize{251,255,251}}{determined\strut} \setlength{\fboxsep}{0pt}\def\cbRGB{\colorbox[RGB]}\expandafter\cbRGB\expandafter{\detokenize{203,255,203}}{the\strut} \setlength{\fboxsep}{0pt}\def\cbRGB{\colorbox[RGB]}\expandafter\cbRGB\expandafter{\detokenize{203,255,203}}{efficacy\strut} \setlength{\fboxsep}{0pt}\def\cbRGB{\colorbox[RGB]}\expandafter\cbRGB\expandafter{\detokenize{204,255,204}}{of\strut} \setlength{\fboxsep}{0pt}\def\cbRGB{\colorbox[RGB]}\expandafter\cbRGB\expandafter{\detokenize{205,255,205}}{unk\strut} \setlength{\fboxsep}{0pt}\def\cbRGB{\colorbox[RGB]}\expandafter\cbRGB\expandafter{\detokenize{204,255,204}}{(\strut} \setlength{\fboxsep}{0pt}\def\cbRGB{\colorbox[RGB]}\expandafter\cbRGB\expandafter{\detokenize{204,255,204}}{unk\strut} \setlength{\fboxsep}{0pt}\def\cbRGB{\colorbox[RGB]}\expandafter\cbRGB\expandafter{\detokenize{203,255,203}}{)\strut} \setlength{\fboxsep}{0pt}\def\cbRGB{\colorbox[RGB]}\expandafter\cbRGB\expandafter{\detokenize{204,255,204}}{in\strut} \setlength{\fboxsep}{0pt}\def\cbRGB{\colorbox[RGB]}\expandafter\cbRGB\expandafter{\detokenize{204,255,204}}{a\strut} \setlength{\fboxsep}{0pt}\def\cbRGB{\colorbox[RGB]}\expandafter\cbRGB\expandafter{\detokenize{204,255,204}}{clinical\strut} \setlength{\fboxsep}{0pt}\def\cbRGB{\colorbox[RGB]}\expandafter\cbRGB\expandafter{\detokenize{206,255,206}}{population\strut} \setlength{\fboxsep}{0pt}\def\cbRGB{\colorbox[RGB]}\expandafter\cbRGB\expandafter{\detokenize{209,255,209}}{of\strut} \setlength{\fboxsep}{0pt}\def\cbRGB{\colorbox[RGB]}\expandafter\cbRGB\expandafter{\detokenize{211,255,211}}{aggressive\strut} \setlength{\fboxsep}{0pt}\def\cbRGB{\colorbox[RGB]}\expandafter\cbRGB\expandafter{\detokenize{218,255,218}}{,\strut} \setlength{\fboxsep}{0pt}\def\cbRGB{\colorbox[RGB]}\expandafter\cbRGB\expandafter{\detokenize{229,255,229}}{urban\strut} \setlength{\fboxsep}{0pt}\def\cbRGB{\colorbox[RGB]}\expandafter\cbRGB\expandafter{\detokenize{239,255,239}}{children\strut} \setlength{\fboxsep}{0pt}\def\cbRGB{\colorbox[RGB]}\expandafter\cbRGB\expandafter{\detokenize{232,255,232}}{diagnosed\strut} \setlength{\fboxsep}{0pt}\def\cbRGB{\colorbox[RGB]}\expandafter\cbRGB\expandafter{\detokenize{236,255,236}}{with\strut} \setlength{\fboxsep}{0pt}\def\cbRGB{\colorbox[RGB]}\expandafter\cbRGB\expandafter{\detokenize{232,255,232}}{attention\strut} \setlength{\fboxsep}{0pt}\def\cbRGB{\colorbox[RGB]}\expandafter\cbRGB\expandafter{\detokenize{232,255,232}}{deficit\strut} \setlength{\fboxsep}{0pt}\def\cbRGB{\colorbox[RGB]}\expandafter\cbRGB\expandafter{\detokenize{216,255,216}}{unk\strut} \setlength{\fboxsep}{0pt}\def\cbRGB{\colorbox[RGB]}\expandafter\cbRGB\expandafter{\detokenize{205,255,205}}{disorder\strut} \setlength{\fboxsep}{0pt}\def\cbRGB{\colorbox[RGB]}\expandafter\cbRGB\expandafter{\detokenize{198,255,198}}{(\strut} \setlength{\fboxsep}{0pt}\def\cbRGB{\colorbox[RGB]}\expandafter\cbRGB\expandafter{\detokenize{188,255,188}}{adhd\strut} \setlength{\fboxsep}{0pt}\def\cbRGB{\colorbox[RGB]}\expandafter\cbRGB\expandafter{\detokenize{200,255,200}}{)\strut} \setlength{\fboxsep}{0pt}\def\cbRGB{\colorbox[RGB]}\expandafter\cbRGB\expandafter{\detokenize{215,255,215}}{.\strut} \setlength{\fboxsep}{0pt}\def\cbRGB{\colorbox[RGB]}\expandafter\cbRGB\expandafter{\detokenize{235,255,235}}{in\strut} \setlength{\fboxsep}{0pt}\def\cbRGB{\colorbox[RGB]}\expandafter\cbRGB\expandafter{\detokenize{226,255,226}}{previous\strut} \setlength{\fboxsep}{0pt}\def\cbRGB{\colorbox[RGB]}\expandafter\cbRGB\expandafter{\detokenize{233,255,233}}{studies\strut} \setlength{\fboxsep}{0pt}\def\cbRGB{\colorbox[RGB]}\expandafter\cbRGB\expandafter{\detokenize{221,255,221}}{of\strut} \setlength{\fboxsep}{0pt}\def\cbRGB{\colorbox[RGB]}\expandafter\cbRGB\expandafter{\detokenize{222,255,222}}{unk\strut} \setlength{\fboxsep}{0pt}\def\cbRGB{\colorbox[RGB]}\expandafter\cbRGB\expandafter{\detokenize{201,255,201}}{children\strut} \setlength{\fboxsep}{0pt}\def\cbRGB{\colorbox[RGB]}\expandafter\cbRGB\expandafter{\detokenize{195,255,195}}{with\strut} \setlength{\fboxsep}{0pt}\def\cbRGB{\colorbox[RGB]}\expandafter\cbRGB\expandafter{\detokenize{182,255,182}}{adhd\strut} \setlength{\fboxsep}{0pt}\def\cbRGB{\colorbox[RGB]}\expandafter\cbRGB\expandafter{\detokenize{177,255,177}}{,\strut} \setlength{\fboxsep}{0pt}\def\cbRGB{\colorbox[RGB]}\expandafter\cbRGB\expandafter{\detokenize{181,255,181}}{unk\strut} \setlength{\fboxsep}{0pt}\def\cbRGB{\colorbox[RGB]}\expandafter\cbRGB\expandafter{\detokenize{192,255,192}}{has\strut} \setlength{\fboxsep}{0pt}\def\cbRGB{\colorbox[RGB]}\expandafter\cbRGB\expandafter{\detokenize{195,255,195}}{been\strut} \setlength{\fboxsep}{0pt}\def\cbRGB{\colorbox[RGB]}\expandafter\cbRGB\expandafter{\detokenize{188,255,188}}{shown\strut} \setlength{\fboxsep}{0pt}\def\cbRGB{\colorbox[RGB]}\expandafter\cbRGB\expandafter{\detokenize{176,255,176}}{to\strut} \setlength{\fboxsep}{0pt}\def\cbRGB{\colorbox[RGB]}\expandafter\cbRGB\expandafter{\detokenize{180,255,180}}{be\strut} \setlength{\fboxsep}{0pt}\def\cbRGB{\colorbox[RGB]}\expandafter\cbRGB\expandafter{\detokenize{185,255,185}}{effective\strut} \setlength{\fboxsep}{0pt}\def\cbRGB{\colorbox[RGB]}\expandafter\cbRGB\expandafter{\detokenize{163,255,163}}{when\strut} \setlength{\fboxsep}{0pt}\def\cbRGB{\colorbox[RGB]}\expandafter\cbRGB\expandafter{\detokenize{135,255,135}}{compared\strut} \setlength{\fboxsep}{0pt}\def\cbRGB{\colorbox[RGB]}\expandafter\cbRGB\expandafter{\detokenize{126,255,126}}{with\strut} \setlength{\fboxsep}{0pt}\def\cbRGB{\colorbox[RGB]}\expandafter\cbRGB\expandafter{\detokenize{144,255,144}}{placebo\strut} \setlength{\fboxsep}{0pt}\def\cbRGB{\colorbox[RGB]}\expandafter\cbRGB\expandafter{\detokenize{163,255,163}}{.\strut} \setlength{\fboxsep}{0pt}\def\cbRGB{\colorbox[RGB]}\expandafter\cbRGB\expandafter{\detokenize{175,255,175}}{eighteen\strut} \setlength{\fboxsep}{0pt}\def\cbRGB{\colorbox[RGB]}\expandafter\cbRGB\expandafter{\detokenize{196,255,196}}{unk\strut} \setlength{\fboxsep}{0pt}\def\cbRGB{\colorbox[RGB]}\expandafter\cbRGB\expandafter{\detokenize{205,255,205}}{children\strut} \setlength{\fboxsep}{0pt}\def\cbRGB{\colorbox[RGB]}\expandafter\cbRGB\expandafter{\detokenize{206,255,206}}{(\strut} \setlength{\fboxsep}{0pt}\def\cbRGB{\colorbox[RGB]}\expandafter\cbRGB\expandafter{\detokenize{191,255,191}}{ages\strut} \setlength{\fboxsep}{0pt}\def\cbRGB{\colorbox[RGB]}\expandafter\cbRGB\expandafter{\detokenize{198,255,198}}{qqq\strut} \setlength{\fboxsep}{0pt}\def\cbRGB{\colorbox[RGB]}\expandafter\cbRGB\expandafter{\detokenize{199,255,199}}{to\strut} \setlength{\fboxsep}{0pt}\def\cbRGB{\colorbox[RGB]}\expandafter\cbRGB\expandafter{\detokenize{211,255,211}}{qqq\strut} \setlength{\fboxsep}{0pt}\def\cbRGB{\colorbox[RGB]}\expandafter\cbRGB\expandafter{\detokenize{220,255,220}}{years\strut} \setlength{\fboxsep}{0pt}\def\cbRGB{\colorbox[RGB]}\expandafter\cbRGB\expandafter{\detokenize{233,255,233}}{)\strut} \setlength{\fboxsep}{0pt}\def\cbRGB{\colorbox[RGB]}\expandafter\cbRGB\expandafter{\detokenize{246,255,246}}{,\strut} \setlength{\fboxsep}{0pt}\def\cbRGB{\colorbox[RGB]}\expandafter\cbRGB\expandafter{\detokenize{230,255,230}}{diagnosed\strut} \setlength{\fboxsep}{0pt}\def\cbRGB{\colorbox[RGB]}\expandafter\cbRGB\expandafter{\detokenize{235,255,235}}{with\strut} \setlength{\fboxsep}{0pt}\def\cbRGB{\colorbox[RGB]}\expandafter\cbRGB\expandafter{\detokenize{221,255,221}}{adhd\strut} \setlength{\fboxsep}{0pt}\def\cbRGB{\colorbox[RGB]}\expandafter\cbRGB\expandafter{\detokenize{236,255,236}}{and\strut} \setlength{\fboxsep}{0pt}\def\cbRGB{\colorbox[RGB]}\expandafter\cbRGB\expandafter{\detokenize{209,255,209}}{attending\strut} \setlength{\fboxsep}{0pt}\def\cbRGB{\colorbox[RGB]}\expandafter\cbRGB\expandafter{\detokenize{175,255,175}}{a\strut} \setlength{\fboxsep}{0pt}\def\cbRGB{\colorbox[RGB]}\expandafter\cbRGB\expandafter{\detokenize{120,255,120}}{unk\strut} \setlength{\fboxsep}{0pt}\def\cbRGB{\colorbox[RGB]}\expandafter\cbRGB\expandafter{\detokenize{124,255,124}}{treatment\strut} \setlength{\fboxsep}{0pt}\def\cbRGB{\colorbox[RGB]}\expandafter\cbRGB\expandafter{\detokenize{99,255,99}}{program\strut} \setlength{\fboxsep}{0pt}\def\cbRGB{\colorbox[RGB]}\expandafter\cbRGB\expandafter{\detokenize{103,255,103}}{for\strut} \setlength{\fboxsep}{0pt}\def\cbRGB{\colorbox[RGB]}\expandafter\cbRGB\expandafter{\detokenize{102,255,102}}{youth\strut} \setlength{\fboxsep}{0pt}\def\cbRGB{\colorbox[RGB]}\expandafter\cbRGB\expandafter{\detokenize{168,255,168}}{with\strut} \setlength{\fboxsep}{0pt}\def\cbRGB{\colorbox[RGB]}\expandafter\cbRGB\expandafter{\detokenize{214,255,214}}{unk\strut} \setlength{\fboxsep}{0pt}\def\cbRGB{\colorbox[RGB]}\expandafter\cbRGB\expandafter{\detokenize{221,255,221}}{behavior\strut} \setlength{\fboxsep}{0pt}\def\cbRGB{\colorbox[RGB]}\expandafter\cbRGB\expandafter{\detokenize{212,255,212}}{disorders\strut} \setlength{\fboxsep}{0pt}\def\cbRGB{\colorbox[RGB]}\expandafter\cbRGB\expandafter{\detokenize{206,255,206}}{,\strut} \setlength{\fboxsep}{0pt}\def\cbRGB{\colorbox[RGB]}\expandafter\cbRGB\expandafter{\detokenize{170,255,170}}{participated\strut} \setlength{\fboxsep}{0pt}\def\cbRGB{\colorbox[RGB]}\expandafter\cbRGB\expandafter{\detokenize{77,255,77}}{in\strut} \setlength{\fboxsep}{0pt}\def\cbRGB{\colorbox[RGB]}\expandafter\cbRGB\expandafter{\detokenize{19,255,19}}{a\strut} \setlength{\fboxsep}{0pt}\def\cbRGB{\colorbox[RGB]}\expandafter\cbRGB\expandafter{\detokenize{0,255,0}}{double-blind\strut} \setlength{\fboxsep}{0pt}\def\cbRGB{\colorbox[RGB]}\expandafter\cbRGB\expandafter{\detokenize{48,255,48}}{placebo\strut} \setlength{\fboxsep}{0pt}\def\cbRGB{\colorbox[RGB]}\expandafter\cbRGB\expandafter{\detokenize{110,255,110}}{trial\strut} \setlength{\fboxsep}{0pt}\def\cbRGB{\colorbox[RGB]}\expandafter\cbRGB\expandafter{\detokenize{169,255,169}}{with\strut} \setlength{\fboxsep}{0pt}\def\cbRGB{\colorbox[RGB]}\expandafter\cbRGB\expandafter{\detokenize{225,255,225}}{assessment\strut} \setlength{\fboxsep}{0pt}\def\cbRGB{\colorbox[RGB]}\expandafter\cbRGB\expandafter{\detokenize{228,255,228}}{data\strut} \setlength{\fboxsep}{0pt}\def\cbRGB{\colorbox[RGB]}\expandafter\cbRGB\expandafter{\detokenize{233,255,233}}{obtained\strut} \setlength{\fboxsep}{0pt}\def\cbRGB{\colorbox[RGB]}\expandafter\cbRGB\expandafter{\detokenize{223,255,223}}{from\strut} \setlength{\fboxsep}{0pt}\def\cbRGB{\colorbox[RGB]}\expandafter\cbRGB\expandafter{\detokenize{181,255,181}}{staff\strut} \setlength{\fboxsep}{0pt}\def\cbRGB{\colorbox[RGB]}\expandafter\cbRGB\expandafter{\detokenize{120,255,120}}{in\strut} \setlength{\fboxsep}{0pt}\def\cbRGB{\colorbox[RGB]}\expandafter\cbRGB\expandafter{\detokenize{121,255,121}}{the\strut} \setlength{\fboxsep}{0pt}\def\cbRGB{\colorbox[RGB]}\expandafter\cbRGB\expandafter{\detokenize{110,255,110}}{program\strut} \setlength{\fboxsep}{0pt}\def\cbRGB{\colorbox[RGB]}\expandafter\cbRGB\expandafter{\detokenize{117,255,117}}{and\strut} \setlength{\fboxsep}{0pt}\def\cbRGB{\colorbox[RGB]}\expandafter\cbRGB\expandafter{\detokenize{118,255,118}}{parents\strut} \setlength{\fboxsep}{0pt}\def\cbRGB{\colorbox[RGB]}\expandafter\cbRGB\expandafter{\detokenize{166,255,166}}{at\strut} \setlength{\fboxsep}{0pt}\def\cbRGB{\colorbox[RGB]}\expandafter\cbRGB\expandafter{\detokenize{206,255,206}}{home\strut} \setlength{\fboxsep}{0pt}\def\cbRGB{\colorbox[RGB]}\expandafter\cbRGB\expandafter{\detokenize{206,255,206}}{.\strut} \setlength{\fboxsep}{0pt}\def\cbRGB{\colorbox[RGB]}\expandafter\cbRGB\expandafter{\detokenize{219,255,219}}{based\strut} \setlength{\fboxsep}{0pt}\def\cbRGB{\colorbox[RGB]}\expandafter\cbRGB\expandafter{\detokenize{231,255,231}}{on\strut} \setlength{\fboxsep}{0pt}\def\cbRGB{\colorbox[RGB]}\expandafter\cbRGB\expandafter{\detokenize{234,255,234}}{staff\strut} \setlength{\fboxsep}{0pt}\def\cbRGB{\colorbox[RGB]}\expandafter\cbRGB\expandafter{\detokenize{219,255,219}}{ratings\strut} \setlength{\fboxsep}{0pt}\def\cbRGB{\colorbox[RGB]}\expandafter\cbRGB\expandafter{\detokenize{210,255,210}}{of\strut} \setlength{\fboxsep}{0pt}\def\cbRGB{\colorbox[RGB]}\expandafter\cbRGB\expandafter{\detokenize{207,255,207}}{the\strut} \setlength{\fboxsep}{0pt}\def\cbRGB{\colorbox[RGB]}\expandafter\cbRGB\expandafter{\detokenize{205,255,205}}{children\strut} \setlength{\fboxsep}{0pt}\def\cbRGB{\colorbox[RGB]}\expandafter\cbRGB\expandafter{\detokenize{202,255,202}}{'s\strut} \setlength{\fboxsep}{0pt}\def\cbRGB{\colorbox[RGB]}\expandafter\cbRGB\expandafter{\detokenize{163,255,163}}{behavior\strut} \setlength{\fboxsep}{0pt}\def\cbRGB{\colorbox[RGB]}\expandafter\cbRGB\expandafter{\detokenize{114,255,114}}{in\strut} \setlength{\fboxsep}{0pt}\def\cbRGB{\colorbox[RGB]}\expandafter\cbRGB\expandafter{\detokenize{108,255,108}}{the\strut} \setlength{\fboxsep}{0pt}\def\cbRGB{\colorbox[RGB]}\expandafter\cbRGB\expandafter{\detokenize{96,255,96}}{program\strut} \setlength{\fboxsep}{0pt}\def\cbRGB{\colorbox[RGB]}\expandafter\cbRGB\expandafter{\detokenize{105,255,105}}{and\strut} \setlength{\fboxsep}{0pt}\def\cbRGB{\colorbox[RGB]}\expandafter\cbRGB\expandafter{\detokenize{105,255,105}}{an\strut} \setlength{\fboxsep}{0pt}\def\cbRGB{\colorbox[RGB]}\expandafter\cbRGB\expandafter{\detokenize{153,255,153}}{academic\strut} \setlength{\fboxsep}{0pt}\def\cbRGB{\colorbox[RGB]}\expandafter\cbRGB\expandafter{\detokenize{190,255,190}}{classroom\strut} \setlength{\fboxsep}{0pt}\def\cbRGB{\colorbox[RGB]}\expandafter\cbRGB\expandafter{\detokenize{190,255,190}}{,\strut} \setlength{\fboxsep}{0pt}\def\cbRGB{\colorbox[RGB]}\expandafter\cbRGB\expandafter{\detokenize{192,255,192}}{the\strut} \setlength{\fboxsep}{0pt}\def\cbRGB{\colorbox[RGB]}\expandafter\cbRGB\expandafter{\detokenize{192,255,192}}{children\strut} \setlength{\fboxsep}{0pt}\def\cbRGB{\colorbox[RGB]}\expandafter\cbRGB\expandafter{\detokenize{197,255,197}}{displayed\strut} \setlength{\fboxsep}{0pt}\def\cbRGB{\colorbox[RGB]}\expandafter\cbRGB\expandafter{\detokenize{197,255,197}}{significant\strut} \setlength{\fboxsep}{0pt}\def\cbRGB{\colorbox[RGB]}\expandafter\cbRGB\expandafter{\detokenize{198,255,198}}{improvements\strut} \setlength{\fboxsep}{0pt}\def\cbRGB{\colorbox[RGB]}\expandafter\cbRGB\expandafter{\detokenize{192,255,192}}{in\strut} \setlength{\fboxsep}{0pt}\def\cbRGB{\colorbox[RGB]}\expandafter\cbRGB\expandafter{\detokenize{188,255,188}}{adhd\strut} \setlength{\fboxsep}{0pt}\def\cbRGB{\colorbox[RGB]}\expandafter\cbRGB\expandafter{\detokenize{196,255,196}}{symptoms\strut} \setlength{\fboxsep}{0pt}\def\cbRGB{\colorbox[RGB]}\expandafter\cbRGB\expandafter{\detokenize{207,255,207}}{and\strut} \setlength{\fboxsep}{0pt}\def\cbRGB{\colorbox[RGB]}\expandafter\cbRGB\expandafter{\detokenize{211,255,211}}{aggressive\strut} \setlength{\fboxsep}{0pt}\def\cbRGB{\colorbox[RGB]}\expandafter\cbRGB\expandafter{\detokenize{210,255,210}}{behavior\strut} \setlength{\fboxsep}{0pt}\def\cbRGB{\colorbox[RGB]}\expandafter\cbRGB\expandafter{\detokenize{186,255,186}}{with\strut} \setlength{\fboxsep}{0pt}\def\cbRGB{\colorbox[RGB]}\expandafter\cbRGB\expandafter{\detokenize{156,255,156}}{unk\strut} \setlength{\fboxsep}{0pt}\def\cbRGB{\colorbox[RGB]}\expandafter\cbRGB\expandafter{\detokenize{152,255,152}}{and\strut} \setlength{\fboxsep}{0pt}\def\cbRGB{\colorbox[RGB]}\expandafter\cbRGB\expandafter{\detokenize{138,255,138}}{high-dose\strut} \setlength{\fboxsep}{0pt}\def\cbRGB{\colorbox[RGB]}\expandafter\cbRGB\expandafter{\detokenize{154,255,154}}{unk\strut} \setlength{\fboxsep}{0pt}\def\cbRGB{\colorbox[RGB]}\expandafter\cbRGB\expandafter{\detokenize{157,255,157}}{conditions\strut} \setlength{\fboxsep}{0pt}\def\cbRGB{\colorbox[RGB]}\expandafter\cbRGB\expandafter{\detokenize{193,255,193}}{.\strut} \setlength{\fboxsep}{0pt}\def\cbRGB{\colorbox[RGB]}\expandafter\cbRGB\expandafter{\detokenize{209,255,209}}{at\strut} \setlength{\fboxsep}{0pt}\def\cbRGB{\colorbox[RGB]}\expandafter\cbRGB\expandafter{\detokenize{209,255,209}}{home\strut} \setlength{\fboxsep}{0pt}\def\cbRGB{\colorbox[RGB]}\expandafter\cbRGB\expandafter{\detokenize{207,255,207}}{,\strut} \setlength{\fboxsep}{0pt}\def\cbRGB{\colorbox[RGB]}\expandafter\cbRGB\expandafter{\detokenize{207,255,207}}{parents\strut} \setlength{\fboxsep}{0pt}\def\cbRGB{\colorbox[RGB]}\expandafter\cbRGB\expandafter{\detokenize{197,255,197}}{and\strut} \setlength{\fboxsep}{0pt}\def\cbRGB{\colorbox[RGB]}\expandafter\cbRGB\expandafter{\detokenize{202,255,202}}{unk\strut} \setlength{\fboxsep}{0pt}\def\cbRGB{\colorbox[RGB]}\expandafter\cbRGB\expandafter{\detokenize{190,255,190}}{reported\strut} \setlength{\fboxsep}{0pt}\def\cbRGB{\colorbox[RGB]}\expandafter\cbRGB\expandafter{\detokenize{205,255,205}}{few\strut} \setlength{\fboxsep}{0pt}\def\cbRGB{\colorbox[RGB]}\expandafter\cbRGB\expandafter{\detokenize{175,255,175}}{significant\strut} \setlength{\fboxsep}{0pt}\def\cbRGB{\colorbox[RGB]}\expandafter\cbRGB\expandafter{\detokenize{151,255,151}}{differences\strut} \setlength{\fboxsep}{0pt}\def\cbRGB{\colorbox[RGB]}\expandafter\cbRGB\expandafter{\detokenize{137,255,137}}{between\strut} \setlength{\fboxsep}{0pt}\def\cbRGB{\colorbox[RGB]}\expandafter\cbRGB\expandafter{\detokenize{143,255,143}}{placebo\strut} \setlength{\fboxsep}{0pt}\def\cbRGB{\colorbox[RGB]}\expandafter\cbRGB\expandafter{\detokenize{163,255,163}}{and\strut} \setlength{\fboxsep}{0pt}\def\cbRGB{\colorbox[RGB]}\expandafter\cbRGB\expandafter{\detokenize{186,255,186}}{unk\strut} \setlength{\fboxsep}{0pt}\def\cbRGB{\colorbox[RGB]}\expandafter\cbRGB\expandafter{\detokenize{217,255,217}}{on\strut} \setlength{\fboxsep}{0pt}\def\cbRGB{\colorbox[RGB]}\expandafter\cbRGB\expandafter{\detokenize{233,255,233}}{behavior\strut} \setlength{\fboxsep}{0pt}\def\cbRGB{\colorbox[RGB]}\expandafter\cbRGB\expandafter{\detokenize{218,255,218}}{ratings\strut} \setlength{\fboxsep}{0pt}\def\cbRGB{\colorbox[RGB]}\expandafter\cbRGB\expandafter{\detokenize{211,255,211}}{.\strut} \setlength{\fboxsep}{0pt}\def\cbRGB{\colorbox[RGB]}\expandafter\cbRGB\expandafter{\detokenize{208,255,208}}{in\strut} \setlength{\fboxsep}{0pt}\def\cbRGB{\colorbox[RGB]}\expandafter\cbRGB\expandafter{\detokenize{208,255,208}}{both\strut} \setlength{\fboxsep}{0pt}\def\cbRGB{\colorbox[RGB]}\expandafter\cbRGB\expandafter{\detokenize{204,255,204}}{settings\strut} \setlength{\fboxsep}{0pt}\def\cbRGB{\colorbox[RGB]}\expandafter\cbRGB\expandafter{\detokenize{205,255,205}}{,\strut} \setlength{\fboxsep}{0pt}\def\cbRGB{\colorbox[RGB]}\expandafter\cbRGB\expandafter{\detokenize{184,255,184}}{unk\strut} \setlength{\fboxsep}{0pt}\def\cbRGB{\colorbox[RGB]}\expandafter\cbRGB\expandafter{\detokenize{156,255,156}}{was\strut} \setlength{\fboxsep}{0pt}\def\cbRGB{\colorbox[RGB]}\expandafter\cbRGB\expandafter{\detokenize{149,255,149}}{well\strut} \setlength{\fboxsep}{0pt}\def\cbRGB{\colorbox[RGB]}\expandafter\cbRGB\expandafter{\detokenize{174,255,174}}{tolerated\strut} \setlength{\fboxsep}{0pt}\def\cbRGB{\colorbox[RGB]}\expandafter\cbRGB\expandafter{\detokenize{198,255,198}}{with\strut} \setlength{\fboxsep}{0pt}\def\cbRGB{\colorbox[RGB]}\expandafter\cbRGB\expandafter{\detokenize{194,255,194}}{few\strut} \setlength{\fboxsep}{0pt}\def\cbRGB{\colorbox[RGB]}\expandafter\cbRGB\expandafter{\detokenize{188,255,188}}{side\strut} \setlength{\fboxsep}{0pt}\def\cbRGB{\colorbox[RGB]}\expandafter\cbRGB\expandafter{\detokenize{187,255,187}}{effects\strut} \setlength{\fboxsep}{0pt}\def\cbRGB{\colorbox[RGB]}\expandafter\cbRGB\expandafter{\detokenize{202,255,202}}{found\strut} \setlength{\fboxsep}{0pt}\def\cbRGB{\colorbox[RGB]}\expandafter\cbRGB\expandafter{\detokenize{206,255,206}}{during\strut} \setlength{\fboxsep}{0pt}\def\cbRGB{\colorbox[RGB]}\expandafter\cbRGB\expandafter{\detokenize{214,255,214}}{active\strut} \setlength{\fboxsep}{0pt}\def\cbRGB{\colorbox[RGB]}\expandafter\cbRGB\expandafter{\detokenize{209,255,209}}{drug\strut} \setlength{\fboxsep}{0pt}\def\cbRGB{\colorbox[RGB]}\expandafter\cbRGB\expandafter{\detokenize{207,255,207}}{conditions\strut} \setlength{\fboxsep}{0pt}\def\cbRGB{\colorbox[RGB]}\expandafter\cbRGB\expandafter{\detokenize{255,255,255}}{.\strut} 

\setlength{\fboxsep}{0pt}\def\cbRGB{\colorbox[RGB]}\expandafter\cbRGB\expandafter{\detokenize{213,213,255}}{determined\strut} \setlength{\fboxsep}{0pt}\def\cbRGB{\colorbox[RGB]}\expandafter\cbRGB\expandafter{\detokenize{177,177,255}}{the\strut} \setlength{\fboxsep}{0pt}\def\cbRGB{\colorbox[RGB]}\expandafter\cbRGB\expandafter{\detokenize{179,179,255}}{efficacy\strut} \setlength{\fboxsep}{0pt}\def\cbRGB{\colorbox[RGB]}\expandafter\cbRGB\expandafter{\detokenize{185,185,255}}{of\strut} \setlength{\fboxsep}{0pt}\def\cbRGB{\colorbox[RGB]}\expandafter\cbRGB\expandafter{\detokenize{188,188,255}}{unk\strut} \setlength{\fboxsep}{0pt}\def\cbRGB{\colorbox[RGB]}\expandafter\cbRGB\expandafter{\detokenize{191,191,255}}{(\strut} \setlength{\fboxsep}{0pt}\def\cbRGB{\colorbox[RGB]}\expandafter\cbRGB\expandafter{\detokenize{189,189,255}}{unk\strut} \setlength{\fboxsep}{0pt}\def\cbRGB{\colorbox[RGB]}\expandafter\cbRGB\expandafter{\detokenize{186,186,255}}{)\strut} \setlength{\fboxsep}{0pt}\def\cbRGB{\colorbox[RGB]}\expandafter\cbRGB\expandafter{\detokenize{185,185,255}}{in\strut} \setlength{\fboxsep}{0pt}\def\cbRGB{\colorbox[RGB]}\expandafter\cbRGB\expandafter{\detokenize{192,192,255}}{a\strut} \setlength{\fboxsep}{0pt}\def\cbRGB{\colorbox[RGB]}\expandafter\cbRGB\expandafter{\detokenize{193,193,255}}{clinical\strut} \setlength{\fboxsep}{0pt}\def\cbRGB{\colorbox[RGB]}\expandafter\cbRGB\expandafter{\detokenize{200,200,255}}{population\strut} \setlength{\fboxsep}{0pt}\def\cbRGB{\colorbox[RGB]}\expandafter\cbRGB\expandafter{\detokenize{203,203,255}}{of\strut} \setlength{\fboxsep}{0pt}\def\cbRGB{\colorbox[RGB]}\expandafter\cbRGB\expandafter{\detokenize{208,208,255}}{aggressive\strut} \setlength{\fboxsep}{0pt}\def\cbRGB{\colorbox[RGB]}\expandafter\cbRGB\expandafter{\detokenize{213,213,255}}{,\strut} \setlength{\fboxsep}{0pt}\def\cbRGB{\colorbox[RGB]}\expandafter\cbRGB\expandafter{\detokenize{220,220,255}}{urban\strut} \setlength{\fboxsep}{0pt}\def\cbRGB{\colorbox[RGB]}\expandafter\cbRGB\expandafter{\detokenize{223,223,255}}{children\strut} \setlength{\fboxsep}{0pt}\def\cbRGB{\colorbox[RGB]}\expandafter\cbRGB\expandafter{\detokenize{223,223,255}}{diagnosed\strut} \setlength{\fboxsep}{0pt}\def\cbRGB{\colorbox[RGB]}\expandafter\cbRGB\expandafter{\detokenize{225,225,255}}{with\strut} \setlength{\fboxsep}{0pt}\def\cbRGB{\colorbox[RGB]}\expandafter\cbRGB\expandafter{\detokenize{210,210,255}}{attention\strut} \setlength{\fboxsep}{0pt}\def\cbRGB{\colorbox[RGB]}\expandafter\cbRGB\expandafter{\detokenize{183,183,255}}{deficit\strut} \setlength{\fboxsep}{0pt}\def\cbRGB{\colorbox[RGB]}\expandafter\cbRGB\expandafter{\detokenize{164,164,255}}{unk\strut} \setlength{\fboxsep}{0pt}\def\cbRGB{\colorbox[RGB]}\expandafter\cbRGB\expandafter{\detokenize{137,137,255}}{disorder\strut} \setlength{\fboxsep}{0pt}\def\cbRGB{\colorbox[RGB]}\expandafter\cbRGB\expandafter{\detokenize{166,166,255}}{(\strut} \setlength{\fboxsep}{0pt}\def\cbRGB{\colorbox[RGB]}\expandafter\cbRGB\expandafter{\detokenize{147,147,255}}{adhd\strut} \setlength{\fboxsep}{0pt}\def\cbRGB{\colorbox[RGB]}\expandafter\cbRGB\expandafter{\detokenize{182,182,255}}{)\strut} \setlength{\fboxsep}{0pt}\def\cbRGB{\colorbox[RGB]}\expandafter\cbRGB\expandafter{\detokenize{194,194,255}}{.\strut} \setlength{\fboxsep}{0pt}\def\cbRGB{\colorbox[RGB]}\expandafter\cbRGB\expandafter{\detokenize{236,236,255}}{in\strut} \setlength{\fboxsep}{0pt}\def\cbRGB{\colorbox[RGB]}\expandafter\cbRGB\expandafter{\detokenize{255,255,255}}{previous\strut} \setlength{\fboxsep}{0pt}\def\cbRGB{\colorbox[RGB]}\expandafter\cbRGB\expandafter{\detokenize{247,247,255}}{studies\strut} \setlength{\fboxsep}{0pt}\def\cbRGB{\colorbox[RGB]}\expandafter\cbRGB\expandafter{\detokenize{240,240,255}}{of\strut} \setlength{\fboxsep}{0pt}\def\cbRGB{\colorbox[RGB]}\expandafter\cbRGB\expandafter{\detokenize{218,218,255}}{unk\strut} \setlength{\fboxsep}{0pt}\def\cbRGB{\colorbox[RGB]}\expandafter\cbRGB\expandafter{\detokenize{197,197,255}}{children\strut} \setlength{\fboxsep}{0pt}\def\cbRGB{\colorbox[RGB]}\expandafter\cbRGB\expandafter{\detokenize{200,200,255}}{with\strut} \setlength{\fboxsep}{0pt}\def\cbRGB{\colorbox[RGB]}\expandafter\cbRGB\expandafter{\detokenize{160,160,255}}{adhd\strut} \setlength{\fboxsep}{0pt}\def\cbRGB{\colorbox[RGB]}\expandafter\cbRGB\expandafter{\detokenize{127,127,255}}{,\strut} \setlength{\fboxsep}{0pt}\def\cbRGB{\colorbox[RGB]}\expandafter\cbRGB\expandafter{\detokenize{113,113,255}}{unk\strut} \setlength{\fboxsep}{0pt}\def\cbRGB{\colorbox[RGB]}\expandafter\cbRGB\expandafter{\detokenize{151,151,255}}{has\strut} \setlength{\fboxsep}{0pt}\def\cbRGB{\colorbox[RGB]}\expandafter\cbRGB\expandafter{\detokenize{182,182,255}}{been\strut} \setlength{\fboxsep}{0pt}\def\cbRGB{\colorbox[RGB]}\expandafter\cbRGB\expandafter{\detokenize{169,169,255}}{shown\strut} \setlength{\fboxsep}{0pt}\def\cbRGB{\colorbox[RGB]}\expandafter\cbRGB\expandafter{\detokenize{156,156,255}}{to\strut} \setlength{\fboxsep}{0pt}\def\cbRGB{\colorbox[RGB]}\expandafter\cbRGB\expandafter{\detokenize{153,153,255}}{be\strut} \setlength{\fboxsep}{0pt}\def\cbRGB{\colorbox[RGB]}\expandafter\cbRGB\expandafter{\detokenize{152,152,255}}{effective\strut} \setlength{\fboxsep}{0pt}\def\cbRGB{\colorbox[RGB]}\expandafter\cbRGB\expandafter{\detokenize{172,172,255}}{when\strut} \setlength{\fboxsep}{0pt}\def\cbRGB{\colorbox[RGB]}\expandafter\cbRGB\expandafter{\detokenize{188,188,255}}{compared\strut} \setlength{\fboxsep}{0pt}\def\cbRGB{\colorbox[RGB]}\expandafter\cbRGB\expandafter{\detokenize{203,203,255}}{with\strut} \setlength{\fboxsep}{0pt}\def\cbRGB{\colorbox[RGB]}\expandafter\cbRGB\expandafter{\detokenize{208,208,255}}{placebo\strut} \setlength{\fboxsep}{0pt}\def\cbRGB{\colorbox[RGB]}\expandafter\cbRGB\expandafter{\detokenize{208,208,255}}{.\strut} \setlength{\fboxsep}{0pt}\def\cbRGB{\colorbox[RGB]}\expandafter\cbRGB\expandafter{\detokenize{208,208,255}}{eighteen\strut} \setlength{\fboxsep}{0pt}\def\cbRGB{\colorbox[RGB]}\expandafter\cbRGB\expandafter{\detokenize{205,205,255}}{unk\strut} \setlength{\fboxsep}{0pt}\def\cbRGB{\colorbox[RGB]}\expandafter\cbRGB\expandafter{\detokenize{209,209,255}}{children\strut} \setlength{\fboxsep}{0pt}\def\cbRGB{\colorbox[RGB]}\expandafter\cbRGB\expandafter{\detokenize{213,213,255}}{(\strut} \setlength{\fboxsep}{0pt}\def\cbRGB{\colorbox[RGB]}\expandafter\cbRGB\expandafter{\detokenize{216,216,255}}{ages\strut} \setlength{\fboxsep}{0pt}\def\cbRGB{\colorbox[RGB]}\expandafter\cbRGB\expandafter{\detokenize{220,220,255}}{qqq\strut} \setlength{\fboxsep}{0pt}\def\cbRGB{\colorbox[RGB]}\expandafter\cbRGB\expandafter{\detokenize{225,225,255}}{to\strut} \setlength{\fboxsep}{0pt}\def\cbRGB{\colorbox[RGB]}\expandafter\cbRGB\expandafter{\detokenize{225,225,255}}{qqq\strut} \setlength{\fboxsep}{0pt}\def\cbRGB{\colorbox[RGB]}\expandafter\cbRGB\expandafter{\detokenize{224,224,255}}{years\strut} \setlength{\fboxsep}{0pt}\def\cbRGB{\colorbox[RGB]}\expandafter\cbRGB\expandafter{\detokenize{227,227,255}}{)\strut} \setlength{\fboxsep}{0pt}\def\cbRGB{\colorbox[RGB]}\expandafter\cbRGB\expandafter{\detokenize{221,221,255}}{,\strut} \setlength{\fboxsep}{0pt}\def\cbRGB{\colorbox[RGB]}\expandafter\cbRGB\expandafter{\detokenize{212,212,255}}{diagnosed\strut} \setlength{\fboxsep}{0pt}\def\cbRGB{\colorbox[RGB]}\expandafter\cbRGB\expandafter{\detokenize{220,220,255}}{with\strut} \setlength{\fboxsep}{0pt}\def\cbRGB{\colorbox[RGB]}\expandafter\cbRGB\expandafter{\detokenize{214,214,255}}{adhd\strut} \setlength{\fboxsep}{0pt}\def\cbRGB{\colorbox[RGB]}\expandafter\cbRGB\expandafter{\detokenize{202,202,255}}{and\strut} \setlength{\fboxsep}{0pt}\def\cbRGB{\colorbox[RGB]}\expandafter\cbRGB\expandafter{\detokenize{186,186,255}}{attending\strut} \setlength{\fboxsep}{0pt}\def\cbRGB{\colorbox[RGB]}\expandafter\cbRGB\expandafter{\detokenize{193,193,255}}{a\strut} \setlength{\fboxsep}{0pt}\def\cbRGB{\colorbox[RGB]}\expandafter\cbRGB\expandafter{\detokenize{206,206,255}}{unk\strut} \setlength{\fboxsep}{0pt}\def\cbRGB{\colorbox[RGB]}\expandafter\cbRGB\expandafter{\detokenize{209,209,255}}{treatment\strut} \setlength{\fboxsep}{0pt}\def\cbRGB{\colorbox[RGB]}\expandafter\cbRGB\expandafter{\detokenize{215,215,255}}{program\strut} \setlength{\fboxsep}{0pt}\def\cbRGB{\colorbox[RGB]}\expandafter\cbRGB\expandafter{\detokenize{217,217,255}}{for\strut} \setlength{\fboxsep}{0pt}\def\cbRGB{\colorbox[RGB]}\expandafter\cbRGB\expandafter{\detokenize{196,196,255}}{youth\strut} \setlength{\fboxsep}{0pt}\def\cbRGB{\colorbox[RGB]}\expandafter\cbRGB\expandafter{\detokenize{175,175,255}}{with\strut} \setlength{\fboxsep}{0pt}\def\cbRGB{\colorbox[RGB]}\expandafter\cbRGB\expandafter{\detokenize{169,169,255}}{unk\strut} \setlength{\fboxsep}{0pt}\def\cbRGB{\colorbox[RGB]}\expandafter\cbRGB\expandafter{\detokenize{171,171,255}}{behavior\strut} \setlength{\fboxsep}{0pt}\def\cbRGB{\colorbox[RGB]}\expandafter\cbRGB\expandafter{\detokenize{148,148,255}}{disorders\strut} \setlength{\fboxsep}{0pt}\def\cbRGB{\colorbox[RGB]}\expandafter\cbRGB\expandafter{\detokenize{131,131,255}}{,\strut} \setlength{\fboxsep}{0pt}\def\cbRGB{\colorbox[RGB]}\expandafter\cbRGB\expandafter{\detokenize{135,135,255}}{participated\strut} \setlength{\fboxsep}{0pt}\def\cbRGB{\colorbox[RGB]}\expandafter\cbRGB\expandafter{\detokenize{172,172,255}}{in\strut} \setlength{\fboxsep}{0pt}\def\cbRGB{\colorbox[RGB]}\expandafter\cbRGB\expandafter{\detokenize{199,199,255}}{a\strut} \setlength{\fboxsep}{0pt}\def\cbRGB{\colorbox[RGB]}\expandafter\cbRGB\expandafter{\detokenize{210,210,255}}{double-blind\strut} \setlength{\fboxsep}{0pt}\def\cbRGB{\colorbox[RGB]}\expandafter\cbRGB\expandafter{\detokenize{216,216,255}}{placebo\strut} \setlength{\fboxsep}{0pt}\def\cbRGB{\colorbox[RGB]}\expandafter\cbRGB\expandafter{\detokenize{211,211,255}}{trial\strut} \setlength{\fboxsep}{0pt}\def\cbRGB{\colorbox[RGB]}\expandafter\cbRGB\expandafter{\detokenize{206,206,255}}{with\strut} \setlength{\fboxsep}{0pt}\def\cbRGB{\colorbox[RGB]}\expandafter\cbRGB\expandafter{\detokenize{200,200,255}}{assessment\strut} \setlength{\fboxsep}{0pt}\def\cbRGB{\colorbox[RGB]}\expandafter\cbRGB\expandafter{\detokenize{200,200,255}}{data\strut} \setlength{\fboxsep}{0pt}\def\cbRGB{\colorbox[RGB]}\expandafter\cbRGB\expandafter{\detokenize{204,204,255}}{obtained\strut} \setlength{\fboxsep}{0pt}\def\cbRGB{\colorbox[RGB]}\expandafter\cbRGB\expandafter{\detokenize{200,200,255}}{from\strut} \setlength{\fboxsep}{0pt}\def\cbRGB{\colorbox[RGB]}\expandafter\cbRGB\expandafter{\detokenize{197,197,255}}{staff\strut} \setlength{\fboxsep}{0pt}\def\cbRGB{\colorbox[RGB]}\expandafter\cbRGB\expandafter{\detokenize{197,197,255}}{in\strut} \setlength{\fboxsep}{0pt}\def\cbRGB{\colorbox[RGB]}\expandafter\cbRGB\expandafter{\detokenize{209,209,255}}{the\strut} \setlength{\fboxsep}{0pt}\def\cbRGB{\colorbox[RGB]}\expandafter\cbRGB\expandafter{\detokenize{209,209,255}}{program\strut} \setlength{\fboxsep}{0pt}\def\cbRGB{\colorbox[RGB]}\expandafter\cbRGB\expandafter{\detokenize{207,207,255}}{and\strut} \setlength{\fboxsep}{0pt}\def\cbRGB{\colorbox[RGB]}\expandafter\cbRGB\expandafter{\detokenize{207,207,255}}{parents\strut} \setlength{\fboxsep}{0pt}\def\cbRGB{\colorbox[RGB]}\expandafter\cbRGB\expandafter{\detokenize{213,213,255}}{at\strut} \setlength{\fboxsep}{0pt}\def\cbRGB{\colorbox[RGB]}\expandafter\cbRGB\expandafter{\detokenize{221,221,255}}{home\strut} \setlength{\fboxsep}{0pt}\def\cbRGB{\colorbox[RGB]}\expandafter\cbRGB\expandafter{\detokenize{228,228,255}}{.\strut} \setlength{\fboxsep}{0pt}\def\cbRGB{\colorbox[RGB]}\expandafter\cbRGB\expandafter{\detokenize{225,225,255}}{based\strut} \setlength{\fboxsep}{0pt}\def\cbRGB{\colorbox[RGB]}\expandafter\cbRGB\expandafter{\detokenize{208,208,255}}{on\strut} \setlength{\fboxsep}{0pt}\def\cbRGB{\colorbox[RGB]}\expandafter\cbRGB\expandafter{\detokenize{204,204,255}}{staff\strut} \setlength{\fboxsep}{0pt}\def\cbRGB{\colorbox[RGB]}\expandafter\cbRGB\expandafter{\detokenize{186,186,255}}{ratings\strut} \setlength{\fboxsep}{0pt}\def\cbRGB{\colorbox[RGB]}\expandafter\cbRGB\expandafter{\detokenize{182,182,255}}{of\strut} \setlength{\fboxsep}{0pt}\def\cbRGB{\colorbox[RGB]}\expandafter\cbRGB\expandafter{\detokenize{155,155,255}}{the\strut} \setlength{\fboxsep}{0pt}\def\cbRGB{\colorbox[RGB]}\expandafter\cbRGB\expandafter{\detokenize{145,145,255}}{children\strut} \setlength{\fboxsep}{0pt}\def\cbRGB{\colorbox[RGB]}\expandafter\cbRGB\expandafter{\detokenize{153,153,255}}{'s\strut} \setlength{\fboxsep}{0pt}\def\cbRGB{\colorbox[RGB]}\expandafter\cbRGB\expandafter{\detokenize{157,157,255}}{behavior\strut} \setlength{\fboxsep}{0pt}\def\cbRGB{\colorbox[RGB]}\expandafter\cbRGB\expandafter{\detokenize{173,173,255}}{in\strut} \setlength{\fboxsep}{0pt}\def\cbRGB{\colorbox[RGB]}\expandafter\cbRGB\expandafter{\detokenize{180,180,255}}{the\strut} \setlength{\fboxsep}{0pt}\def\cbRGB{\colorbox[RGB]}\expandafter\cbRGB\expandafter{\detokenize{182,182,255}}{program\strut} \setlength{\fboxsep}{0pt}\def\cbRGB{\colorbox[RGB]}\expandafter\cbRGB\expandafter{\detokenize{182,182,255}}{and\strut} \setlength{\fboxsep}{0pt}\def\cbRGB{\colorbox[RGB]}\expandafter\cbRGB\expandafter{\detokenize{185,185,255}}{an\strut} \setlength{\fboxsep}{0pt}\def\cbRGB{\colorbox[RGB]}\expandafter\cbRGB\expandafter{\detokenize{192,192,255}}{academic\strut} \setlength{\fboxsep}{0pt}\def\cbRGB{\colorbox[RGB]}\expandafter\cbRGB\expandafter{\detokenize{196,196,255}}{classroom\strut} \setlength{\fboxsep}{0pt}\def\cbRGB{\colorbox[RGB]}\expandafter\cbRGB\expandafter{\detokenize{170,170,255}}{,\strut} \setlength{\fboxsep}{0pt}\def\cbRGB{\colorbox[RGB]}\expandafter\cbRGB\expandafter{\detokenize{139,139,255}}{the\strut} \setlength{\fboxsep}{0pt}\def\cbRGB{\colorbox[RGB]}\expandafter\cbRGB\expandafter{\detokenize{114,114,255}}{children\strut} \setlength{\fboxsep}{0pt}\def\cbRGB{\colorbox[RGB]}\expandafter\cbRGB\expandafter{\detokenize{62,62,255}}{displayed\strut} \setlength{\fboxsep}{0pt}\def\cbRGB{\colorbox[RGB]}\expandafter\cbRGB\expandafter{\detokenize{47,47,255}}{significant\strut} \setlength{\fboxsep}{0pt}\def\cbRGB{\colorbox[RGB]}\expandafter\cbRGB\expandafter{\detokenize{0,0,255}}{improvements\strut} \setlength{\fboxsep}{0pt}\def\cbRGB{\colorbox[RGB]}\expandafter\cbRGB\expandafter{\detokenize{30,30,255}}{in\strut} \setlength{\fboxsep}{0pt}\def\cbRGB{\colorbox[RGB]}\expandafter\cbRGB\expandafter{\detokenize{25,25,255}}{adhd\strut} \setlength{\fboxsep}{0pt}\def\cbRGB{\colorbox[RGB]}\expandafter\cbRGB\expandafter{\detokenize{24,24,255}}{symptoms\strut} \setlength{\fboxsep}{0pt}\def\cbRGB{\colorbox[RGB]}\expandafter\cbRGB\expandafter{\detokenize{42,42,255}}{and\strut} \setlength{\fboxsep}{0pt}\def\cbRGB{\colorbox[RGB]}\expandafter\cbRGB\expandafter{\detokenize{104,104,255}}{aggressive\strut} \setlength{\fboxsep}{0pt}\def\cbRGB{\colorbox[RGB]}\expandafter\cbRGB\expandafter{\detokenize{173,173,255}}{behavior\strut} \setlength{\fboxsep}{0pt}\def\cbRGB{\colorbox[RGB]}\expandafter\cbRGB\expandafter{\detokenize{191,191,255}}{with\strut} \setlength{\fboxsep}{0pt}\def\cbRGB{\colorbox[RGB]}\expandafter\cbRGB\expandafter{\detokenize{184,184,255}}{unk\strut} \setlength{\fboxsep}{0pt}\def\cbRGB{\colorbox[RGB]}\expandafter\cbRGB\expandafter{\detokenize{187,187,255}}{and\strut} \setlength{\fboxsep}{0pt}\def\cbRGB{\colorbox[RGB]}\expandafter\cbRGB\expandafter{\detokenize{190,190,255}}{high-dose\strut} \setlength{\fboxsep}{0pt}\def\cbRGB{\colorbox[RGB]}\expandafter\cbRGB\expandafter{\detokenize{187,187,255}}{unk\strut} \setlength{\fboxsep}{0pt}\def\cbRGB{\colorbox[RGB]}\expandafter\cbRGB\expandafter{\detokenize{189,189,255}}{conditions\strut} \setlength{\fboxsep}{0pt}\def\cbRGB{\colorbox[RGB]}\expandafter\cbRGB\expandafter{\detokenize{195,195,255}}{.\strut} \setlength{\fboxsep}{0pt}\def\cbRGB{\colorbox[RGB]}\expandafter\cbRGB\expandafter{\detokenize{203,203,255}}{at\strut} \setlength{\fboxsep}{0pt}\def\cbRGB{\colorbox[RGB]}\expandafter\cbRGB\expandafter{\detokenize{208,208,255}}{home\strut} \setlength{\fboxsep}{0pt}\def\cbRGB{\colorbox[RGB]}\expandafter\cbRGB\expandafter{\detokenize{209,209,255}}{,\strut} \setlength{\fboxsep}{0pt}\def\cbRGB{\colorbox[RGB]}\expandafter\cbRGB\expandafter{\detokenize{202,202,255}}{parents\strut} \setlength{\fboxsep}{0pt}\def\cbRGB{\colorbox[RGB]}\expandafter\cbRGB\expandafter{\detokenize{194,194,255}}{and\strut} \setlength{\fboxsep}{0pt}\def\cbRGB{\colorbox[RGB]}\expandafter\cbRGB\expandafter{\detokenize{190,190,255}}{unk\strut} \setlength{\fboxsep}{0pt}\def\cbRGB{\colorbox[RGB]}\expandafter\cbRGB\expandafter{\detokenize{187,187,255}}{reported\strut} \setlength{\fboxsep}{0pt}\def\cbRGB{\colorbox[RGB]}\expandafter\cbRGB\expandafter{\detokenize{186,186,255}}{few\strut} \setlength{\fboxsep}{0pt}\def\cbRGB{\colorbox[RGB]}\expandafter\cbRGB\expandafter{\detokenize{190,190,255}}{significant\strut} \setlength{\fboxsep}{0pt}\def\cbRGB{\colorbox[RGB]}\expandafter\cbRGB\expandafter{\detokenize{200,200,255}}{differences\strut} \setlength{\fboxsep}{0pt}\def\cbRGB{\colorbox[RGB]}\expandafter\cbRGB\expandafter{\detokenize{205,205,255}}{between\strut} \setlength{\fboxsep}{0pt}\def\cbRGB{\colorbox[RGB]}\expandafter\cbRGB\expandafter{\detokenize{204,204,255}}{placebo\strut} \setlength{\fboxsep}{0pt}\def\cbRGB{\colorbox[RGB]}\expandafter\cbRGB\expandafter{\detokenize{186,186,255}}{and\strut} \setlength{\fboxsep}{0pt}\def\cbRGB{\colorbox[RGB]}\expandafter\cbRGB\expandafter{\detokenize{140,140,255}}{unk\strut} \setlength{\fboxsep}{0pt}\def\cbRGB{\colorbox[RGB]}\expandafter\cbRGB\expandafter{\detokenize{116,116,255}}{on\strut} \setlength{\fboxsep}{0pt}\def\cbRGB{\colorbox[RGB]}\expandafter\cbRGB\expandafter{\detokenize{112,112,255}}{behavior\strut} \setlength{\fboxsep}{0pt}\def\cbRGB{\colorbox[RGB]}\expandafter\cbRGB\expandafter{\detokenize{123,123,255}}{ratings\strut} \setlength{\fboxsep}{0pt}\def\cbRGB{\colorbox[RGB]}\expandafter\cbRGB\expandafter{\detokenize{136,136,255}}{.\strut} \setlength{\fboxsep}{0pt}\def\cbRGB{\colorbox[RGB]}\expandafter\cbRGB\expandafter{\detokenize{151,151,255}}{in\strut} \setlength{\fboxsep}{0pt}\def\cbRGB{\colorbox[RGB]}\expandafter\cbRGB\expandafter{\detokenize{187,187,255}}{both\strut} \setlength{\fboxsep}{0pt}\def\cbRGB{\colorbox[RGB]}\expandafter\cbRGB\expandafter{\detokenize{197,197,255}}{settings\strut} \setlength{\fboxsep}{0pt}\def\cbRGB{\colorbox[RGB]}\expandafter\cbRGB\expandafter{\detokenize{184,184,255}}{,\strut} \setlength{\fboxsep}{0pt}\def\cbRGB{\colorbox[RGB]}\expandafter\cbRGB\expandafter{\detokenize{165,165,255}}{unk\strut} \setlength{\fboxsep}{0pt}\def\cbRGB{\colorbox[RGB]}\expandafter\cbRGB\expandafter{\detokenize{160,160,255}}{was\strut} \setlength{\fboxsep}{0pt}\def\cbRGB{\colorbox[RGB]}\expandafter\cbRGB\expandafter{\detokenize{168,168,255}}{well\strut} \setlength{\fboxsep}{0pt}\def\cbRGB{\colorbox[RGB]}\expandafter\cbRGB\expandafter{\detokenize{179,179,255}}{tolerated\strut} \setlength{\fboxsep}{0pt}\def\cbRGB{\colorbox[RGB]}\expandafter\cbRGB\expandafter{\detokenize{181,181,255}}{with\strut} \setlength{\fboxsep}{0pt}\def\cbRGB{\colorbox[RGB]}\expandafter\cbRGB\expandafter{\detokenize{179,179,255}}{few\strut} \setlength{\fboxsep}{0pt}\def\cbRGB{\colorbox[RGB]}\expandafter\cbRGB\expandafter{\detokenize{182,182,255}}{side\strut} \setlength{\fboxsep}{0pt}\def\cbRGB{\colorbox[RGB]}\expandafter\cbRGB\expandafter{\detokenize{181,181,255}}{effects\strut} \setlength{\fboxsep}{0pt}\def\cbRGB{\colorbox[RGB]}\expandafter\cbRGB\expandafter{\detokenize{184,184,255}}{found\strut} \setlength{\fboxsep}{0pt}\def\cbRGB{\colorbox[RGB]}\expandafter\cbRGB\expandafter{\detokenize{186,186,255}}{during\strut} \setlength{\fboxsep}{0pt}\def\cbRGB{\colorbox[RGB]}\expandafter\cbRGB\expandafter{\detokenize{192,192,255}}{active\strut} \setlength{\fboxsep}{0pt}\def\cbRGB{\colorbox[RGB]}\expandafter\cbRGB\expandafter{\detokenize{196,196,255}}{drug\strut} \setlength{\fboxsep}{0pt}\def\cbRGB{\colorbox[RGB]}\expandafter\cbRGB\expandafter{\detokenize{199,199,255}}{conditions\strut} \setlength{\fboxsep}{0pt}\def\cbRGB{\colorbox[RGB]}\expandafter\cbRGB\expandafter{\detokenize{231,231,255}}{.\strut} 

\par
\textbf{Example 2}

\setlength{\fboxsep}{0pt}\def\cbRGB{\colorbox[RGB]}\expandafter\cbRGB\expandafter{\detokenize{255,255,255}}{aims\strut} \setlength{\fboxsep}{0pt}\def\cbRGB{\colorbox[RGB]}\expandafter\cbRGB\expandafter{\detokenize{255,252,252}}{:\strut} \setlength{\fboxsep}{0pt}\def\cbRGB{\colorbox[RGB]}\expandafter\cbRGB\expandafter{\detokenize{255,250,250}}{to\strut} \setlength{\fboxsep}{0pt}\def\cbRGB{\colorbox[RGB]}\expandafter\cbRGB\expandafter{\detokenize{255,249,249}}{assess\strut} \setlength{\fboxsep}{0pt}\def\cbRGB{\colorbox[RGB]}\expandafter\cbRGB\expandafter{\detokenize{255,247,247}}{maternal\strut} \setlength{\fboxsep}{0pt}\def\cbRGB{\colorbox[RGB]}\expandafter\cbRGB\expandafter{\detokenize{255,249,249}}{and\strut} \setlength{\fboxsep}{0pt}\def\cbRGB{\colorbox[RGB]}\expandafter\cbRGB\expandafter{\detokenize{255,248,248}}{neonatal\strut} \setlength{\fboxsep}{0pt}\def\cbRGB{\colorbox[RGB]}\expandafter\cbRGB\expandafter{\detokenize{255,249,249}}{complications\strut} \setlength{\fboxsep}{0pt}\def\cbRGB{\colorbox[RGB]}\expandafter\cbRGB\expandafter{\detokenize{255,247,247}}{in\strut} \setlength{\fboxsep}{0pt}\def\cbRGB{\colorbox[RGB]}\expandafter\cbRGB\expandafter{\detokenize{255,186,186}}{pregnancies\strut} \setlength{\fboxsep}{0pt}\def\cbRGB{\colorbox[RGB]}\expandafter\cbRGB\expandafter{\detokenize{255,141,141}}{of\strut} \setlength{\fboxsep}{0pt}\def\cbRGB{\colorbox[RGB]}\expandafter\cbRGB\expandafter{\detokenize{255,113,113}}{diabetic\strut} \setlength{\fboxsep}{0pt}\def\cbRGB{\colorbox[RGB]}\expandafter\cbRGB\expandafter{\detokenize{255,112,112}}{women\strut} \setlength{\fboxsep}{0pt}\def\cbRGB{\colorbox[RGB]}\expandafter\cbRGB\expandafter{\detokenize{255,141,141}}{treated\strut} \setlength{\fboxsep}{0pt}\def\cbRGB{\colorbox[RGB]}\expandafter\cbRGB\expandafter{\detokenize{255,163,163}}{with\strut} \setlength{\fboxsep}{0pt}\def\cbRGB{\colorbox[RGB]}\expandafter\cbRGB\expandafter{\detokenize{255,224,224}}{oral\strut} \setlength{\fboxsep}{0pt}\def\cbRGB{\colorbox[RGB]}\expandafter\cbRGB\expandafter{\detokenize{255,237,237}}{hypoglycaemic\strut} \setlength{\fboxsep}{0pt}\def\cbRGB{\colorbox[RGB]}\expandafter\cbRGB\expandafter{\detokenize{255,237,237}}{agents\strut} \setlength{\fboxsep}{0pt}\def\cbRGB{\colorbox[RGB]}\expandafter\cbRGB\expandafter{\detokenize{255,238,238}}{during\strut} \setlength{\fboxsep}{0pt}\def\cbRGB{\colorbox[RGB]}\expandafter\cbRGB\expandafter{\detokenize{255,236,236}}{pregnancy\strut} \setlength{\fboxsep}{0pt}\def\cbRGB{\colorbox[RGB]}\expandafter\cbRGB\expandafter{\detokenize{255,237,237}}{.\strut} \setlength{\fboxsep}{0pt}\def\cbRGB{\colorbox[RGB]}\expandafter\cbRGB\expandafter{\detokenize{255,235,235}}{methods\strut} \setlength{\fboxsep}{0pt}\def\cbRGB{\colorbox[RGB]}\expandafter\cbRGB\expandafter{\detokenize{255,238,238}}{:\strut} \setlength{\fboxsep}{0pt}\def\cbRGB{\colorbox[RGB]}\expandafter\cbRGB\expandafter{\detokenize{255,243,243}}{a\strut} \setlength{\fboxsep}{0pt}\def\cbRGB{\colorbox[RGB]}\expandafter\cbRGB\expandafter{\detokenize{255,246,246}}{cohort\strut} \setlength{\fboxsep}{0pt}\def\cbRGB{\colorbox[RGB]}\expandafter\cbRGB\expandafter{\detokenize{255,248,248}}{study\strut} \setlength{\fboxsep}{0pt}\def\cbRGB{\colorbox[RGB]}\expandafter\cbRGB\expandafter{\detokenize{255,248,248}}{including\strut} \setlength{\fboxsep}{0pt}\def\cbRGB{\colorbox[RGB]}\expandafter\cbRGB\expandafter{\detokenize{255,249,249}}{all\strut} \setlength{\fboxsep}{0pt}\def\cbRGB{\colorbox[RGB]}\expandafter\cbRGB\expandafter{\detokenize{255,250,250}}{unk\strut} \setlength{\fboxsep}{0pt}\def\cbRGB{\colorbox[RGB]}\expandafter\cbRGB\expandafter{\detokenize{255,249,249}}{registered\strut} \setlength{\fboxsep}{0pt}\def\cbRGB{\colorbox[RGB]}\expandafter\cbRGB\expandafter{\detokenize{255,247,247}}{,\strut} \setlength{\fboxsep}{0pt}\def\cbRGB{\colorbox[RGB]}\expandafter\cbRGB\expandafter{\detokenize{255,231,231}}{orally\strut} \setlength{\fboxsep}{0pt}\def\cbRGB{\colorbox[RGB]}\expandafter\cbRGB\expandafter{\detokenize{255,185,185}}{treated\strut} \setlength{\fboxsep}{0pt}\def\cbRGB{\colorbox[RGB]}\expandafter\cbRGB\expandafter{\detokenize{255,146,146}}{pregnant\strut} \setlength{\fboxsep}{0pt}\def\cbRGB{\colorbox[RGB]}\expandafter\cbRGB\expandafter{\detokenize{255,106,106}}{diabetic\strut} \setlength{\fboxsep}{0pt}\def\cbRGB{\colorbox[RGB]}\expandafter\cbRGB\expandafter{\detokenize{255,90,90}}{patients\strut} \setlength{\fboxsep}{0pt}\def\cbRGB{\colorbox[RGB]}\expandafter\cbRGB\expandafter{\detokenize{255,113,113}}{set\strut} \setlength{\fboxsep}{0pt}\def\cbRGB{\colorbox[RGB]}\expandafter\cbRGB\expandafter{\detokenize{255,147,147}}{in\strut} \setlength{\fboxsep}{0pt}\def\cbRGB{\colorbox[RGB]}\expandafter\cbRGB\expandafter{\detokenize{255,203,203}}{a\strut} \setlength{\fboxsep}{0pt}\def\cbRGB{\colorbox[RGB]}\expandafter\cbRGB\expandafter{\detokenize{255,208,208}}{diabetic\strut} \setlength{\fboxsep}{0pt}\def\cbRGB{\colorbox[RGB]}\expandafter\cbRGB\expandafter{\detokenize{255,220,220}}{unk\strut} \setlength{\fboxsep}{0pt}\def\cbRGB{\colorbox[RGB]}\expandafter\cbRGB\expandafter{\detokenize{255,225,225}}{service\strut} \setlength{\fboxsep}{0pt}\def\cbRGB{\colorbox[RGB]}\expandafter\cbRGB\expandafter{\detokenize{255,236,236}}{at\strut} \setlength{\fboxsep}{0pt}\def\cbRGB{\colorbox[RGB]}\expandafter\cbRGB\expandafter{\detokenize{255,243,243}}{a\strut} \setlength{\fboxsep}{0pt}\def\cbRGB{\colorbox[RGB]}\expandafter\cbRGB\expandafter{\detokenize{255,245,245}}{university\strut} \setlength{\fboxsep}{0pt}\def\cbRGB{\colorbox[RGB]}\expandafter\cbRGB\expandafter{\detokenize{255,243,243}}{hospital\strut} \setlength{\fboxsep}{0pt}\def\cbRGB{\colorbox[RGB]}\expandafter\cbRGB\expandafter{\detokenize{255,239,239}}{:\strut} \setlength{\fboxsep}{0pt}\def\cbRGB{\colorbox[RGB]}\expandafter\cbRGB\expandafter{\detokenize{255,237,237}}{qqq\strut} \setlength{\fboxsep}{0pt}\def\cbRGB{\colorbox[RGB]}\expandafter\cbRGB\expandafter{\detokenize{255,232,232}}{women\strut} \setlength{\fboxsep}{0pt}\def\cbRGB{\colorbox[RGB]}\expandafter\cbRGB\expandafter{\detokenize{255,229,229}}{treated\strut} \setlength{\fboxsep}{0pt}\def\cbRGB{\colorbox[RGB]}\expandafter\cbRGB\expandafter{\detokenize{255,227,227}}{with\strut} \setlength{\fboxsep}{0pt}\def\cbRGB{\colorbox[RGB]}\expandafter\cbRGB\expandafter{\detokenize{255,228,228}}{metformin\strut} \setlength{\fboxsep}{0pt}\def\cbRGB{\colorbox[RGB]}\expandafter\cbRGB\expandafter{\detokenize{255,230,230}}{,\strut} \setlength{\fboxsep}{0pt}\def\cbRGB{\colorbox[RGB]}\expandafter\cbRGB\expandafter{\detokenize{255,231,231}}{qqq\strut} \setlength{\fboxsep}{0pt}\def\cbRGB{\colorbox[RGB]}\expandafter\cbRGB\expandafter{\detokenize{255,152,152}}{women\strut} \setlength{\fboxsep}{0pt}\def\cbRGB{\colorbox[RGB]}\expandafter\cbRGB\expandafter{\detokenize{255,17,17}}{treated\strut} \setlength{\fboxsep}{0pt}\def\cbRGB{\colorbox[RGB]}\expandafter\cbRGB\expandafter{\detokenize{255,0,0}}{with\strut} \setlength{\fboxsep}{0pt}\def\cbRGB{\colorbox[RGB]}\expandafter\cbRGB\expandafter{\detokenize{255,55,55}}{sulphonylurea\strut} \setlength{\fboxsep}{0pt}\def\cbRGB{\colorbox[RGB]}\expandafter\cbRGB\expandafter{\detokenize{255,186,186}}{during\strut} \setlength{\fboxsep}{0pt}\def\cbRGB{\colorbox[RGB]}\expandafter\cbRGB\expandafter{\detokenize{255,203,203}}{pregnancy\strut} \setlength{\fboxsep}{0pt}\def\cbRGB{\colorbox[RGB]}\expandafter\cbRGB\expandafter{\detokenize{255,232,232}}{and\strut} \setlength{\fboxsep}{0pt}\def\cbRGB{\colorbox[RGB]}\expandafter\cbRGB\expandafter{\detokenize{255,241,241}}{a\strut} \setlength{\fboxsep}{0pt}\def\cbRGB{\colorbox[RGB]}\expandafter\cbRGB\expandafter{\detokenize{255,244,244}}{reference\strut} \setlength{\fboxsep}{0pt}\def\cbRGB{\colorbox[RGB]}\expandafter\cbRGB\expandafter{\detokenize{255,242,242}}{group\strut} \setlength{\fboxsep}{0pt}\def\cbRGB{\colorbox[RGB]}\expandafter\cbRGB\expandafter{\detokenize{255,196,196}}{of\strut} \setlength{\fboxsep}{0pt}\def\cbRGB{\colorbox[RGB]}\expandafter\cbRGB\expandafter{\detokenize{255,166,166}}{qqq\strut} \setlength{\fboxsep}{0pt}\def\cbRGB{\colorbox[RGB]}\expandafter\cbRGB\expandafter{\detokenize{255,140,140}}{diabetic\strut} \setlength{\fboxsep}{0pt}\def\cbRGB{\colorbox[RGB]}\expandafter\cbRGB\expandafter{\detokenize{255,111,111}}{women\strut} \setlength{\fboxsep}{0pt}\def\cbRGB{\colorbox[RGB]}\expandafter\cbRGB\expandafter{\detokenize{255,121,121}}{treated\strut} \setlength{\fboxsep}{0pt}\def\cbRGB{\colorbox[RGB]}\expandafter\cbRGB\expandafter{\detokenize{255,138,138}}{with\strut} \setlength{\fboxsep}{0pt}\def\cbRGB{\colorbox[RGB]}\expandafter\cbRGB\expandafter{\detokenize{255,197,197}}{insulin\strut} \setlength{\fboxsep}{0pt}\def\cbRGB{\colorbox[RGB]}\expandafter\cbRGB\expandafter{\detokenize{255,217,217}}{during\strut} \setlength{\fboxsep}{0pt}\def\cbRGB{\colorbox[RGB]}\expandafter\cbRGB\expandafter{\detokenize{255,227,227}}{pregnancy\strut} \setlength{\fboxsep}{0pt}\def\cbRGB{\colorbox[RGB]}\expandafter\cbRGB\expandafter{\detokenize{255,242,242}}{.\strut} \setlength{\fboxsep}{0pt}\def\cbRGB{\colorbox[RGB]}\expandafter\cbRGB\expandafter{\detokenize{255,242,242}}{results\strut} \setlength{\fboxsep}{0pt}\def\cbRGB{\colorbox[RGB]}\expandafter\cbRGB\expandafter{\detokenize{255,245,245}}{:\strut} \setlength{\fboxsep}{0pt}\def\cbRGB{\colorbox[RGB]}\expandafter\cbRGB\expandafter{\detokenize{255,242,242}}{the\strut} \setlength{\fboxsep}{0pt}\def\cbRGB{\colorbox[RGB]}\expandafter\cbRGB\expandafter{\detokenize{255,221,221}}{prevalence\strut} \setlength{\fboxsep}{0pt}\def\cbRGB{\colorbox[RGB]}\expandafter\cbRGB\expandafter{\detokenize{255,218,218}}{of\strut} \setlength{\fboxsep}{0pt}\def\cbRGB{\colorbox[RGB]}\expandafter\cbRGB\expandafter{\detokenize{255,212,212}}{pre-eclampsia\strut} \setlength{\fboxsep}{0pt}\def\cbRGB{\colorbox[RGB]}\expandafter\cbRGB\expandafter{\detokenize{255,222,222}}{was\strut} \setlength{\fboxsep}{0pt}\def\cbRGB{\colorbox[RGB]}\expandafter\cbRGB\expandafter{\detokenize{255,222,222}}{significantly\strut} \setlength{\fboxsep}{0pt}\def\cbRGB{\colorbox[RGB]}\expandafter\cbRGB\expandafter{\detokenize{255,231,231}}{increased\strut} \setlength{\fboxsep}{0pt}\def\cbRGB{\colorbox[RGB]}\expandafter\cbRGB\expandafter{\detokenize{255,240,240}}{in\strut} \setlength{\fboxsep}{0pt}\def\cbRGB{\colorbox[RGB]}\expandafter\cbRGB\expandafter{\detokenize{255,240,240}}{the\strut} \setlength{\fboxsep}{0pt}\def\cbRGB{\colorbox[RGB]}\expandafter\cbRGB\expandafter{\detokenize{255,237,237}}{group\strut} \setlength{\fboxsep}{0pt}\def\cbRGB{\colorbox[RGB]}\expandafter\cbRGB\expandafter{\detokenize{255,235,235}}{of\strut} \setlength{\fboxsep}{0pt}\def\cbRGB{\colorbox[RGB]}\expandafter\cbRGB\expandafter{\detokenize{255,228,228}}{women\strut} \setlength{\fboxsep}{0pt}\def\cbRGB{\colorbox[RGB]}\expandafter\cbRGB\expandafter{\detokenize{255,225,225}}{treated\strut} \setlength{\fboxsep}{0pt}\def\cbRGB{\colorbox[RGB]}\expandafter\cbRGB\expandafter{\detokenize{255,220,220}}{with\strut} \setlength{\fboxsep}{0pt}\def\cbRGB{\colorbox[RGB]}\expandafter\cbRGB\expandafter{\detokenize{255,223,223}}{metformin\strut} \setlength{\fboxsep}{0pt}\def\cbRGB{\colorbox[RGB]}\expandafter\cbRGB\expandafter{\detokenize{255,225,225}}{compared\strut} \setlength{\fboxsep}{0pt}\def\cbRGB{\colorbox[RGB]}\expandafter\cbRGB\expandafter{\detokenize{255,228,228}}{to\strut} \setlength{\fboxsep}{0pt}\def\cbRGB{\colorbox[RGB]}\expandafter\cbRGB\expandafter{\detokenize{255,162,162}}{women\strut} \setlength{\fboxsep}{0pt}\def\cbRGB{\colorbox[RGB]}\expandafter\cbRGB\expandafter{\detokenize{255,65,65}}{treated\strut} \setlength{\fboxsep}{0pt}\def\cbRGB{\colorbox[RGB]}\expandafter\cbRGB\expandafter{\detokenize{255,63,63}}{with\strut} \setlength{\fboxsep}{0pt}\def\cbRGB{\colorbox[RGB]}\expandafter\cbRGB\expandafter{\detokenize{255,129,129}}{sulphonylurea\strut} \setlength{\fboxsep}{0pt}\def\cbRGB{\colorbox[RGB]}\expandafter\cbRGB\expandafter{\detokenize{255,227,227}}{or\strut} \setlength{\fboxsep}{0pt}\def\cbRGB{\colorbox[RGB]}\expandafter\cbRGB\expandafter{\detokenize{255,234,234}}{insulin\strut} \setlength{\fboxsep}{0pt}\def\cbRGB{\colorbox[RGB]}\expandafter\cbRGB\expandafter{\detokenize{255,243,243}}{(\strut} \setlength{\fboxsep}{0pt}\def\cbRGB{\colorbox[RGB]}\expandafter\cbRGB\expandafter{\detokenize{255,249,249}}{qqq\strut} \setlength{\fboxsep}{0pt}\def\cbRGB{\colorbox[RGB]}\expandafter\cbRGB\expandafter{\detokenize{255,248,248}}{vs.\strut} \setlength{\fboxsep}{0pt}\def\cbRGB{\colorbox[RGB]}\expandafter\cbRGB\expandafter{\detokenize{255,248,248}}{qqq\strut} \setlength{\fboxsep}{0pt}\def\cbRGB{\colorbox[RGB]}\expandafter\cbRGB\expandafter{\detokenize{255,247,247}}{vs.\strut} \setlength{\fboxsep}{0pt}\def\cbRGB{\colorbox[RGB]}\expandafter\cbRGB\expandafter{\detokenize{255,247,247}}{qqq\strut} \setlength{\fboxsep}{0pt}\def\cbRGB{\colorbox[RGB]}\expandafter\cbRGB\expandafter{\detokenize{255,245,245}}{\%\strut} \setlength{\fboxsep}{0pt}\def\cbRGB{\colorbox[RGB]}\expandafter\cbRGB\expandafter{\detokenize{255,245,245}}{,\strut} \setlength{\fboxsep}{0pt}\def\cbRGB{\colorbox[RGB]}\expandafter\cbRGB\expandafter{\detokenize{255,244,244}}{p\strut} \setlength{\fboxsep}{0pt}\def\cbRGB{\colorbox[RGB]}\expandafter\cbRGB\expandafter{\detokenize{255,244,244}}{<\strut} \setlength{\fboxsep}{0pt}\def\cbRGB{\colorbox[RGB]}\expandafter\cbRGB\expandafter{\detokenize{255,244,244}}{qqq\strut} \setlength{\fboxsep}{0pt}\def\cbRGB{\colorbox[RGB]}\expandafter\cbRGB\expandafter{\detokenize{255,245,245}}{)\strut} \setlength{\fboxsep}{0pt}\def\cbRGB{\colorbox[RGB]}\expandafter\cbRGB\expandafter{\detokenize{255,245,245}}{.\strut} \setlength{\fboxsep}{0pt}\def\cbRGB{\colorbox[RGB]}\expandafter\cbRGB\expandafter{\detokenize{255,244,244}}{no\strut} \setlength{\fboxsep}{0pt}\def\cbRGB{\colorbox[RGB]}\expandafter\cbRGB\expandafter{\detokenize{255,244,244}}{difference\strut} \setlength{\fboxsep}{0pt}\def\cbRGB{\colorbox[RGB]}\expandafter\cbRGB\expandafter{\detokenize{255,245,245}}{in\strut} \setlength{\fboxsep}{0pt}\def\cbRGB{\colorbox[RGB]}\expandafter\cbRGB\expandafter{\detokenize{255,245,245}}{neonatal\strut} \setlength{\fboxsep}{0pt}\def\cbRGB{\colorbox[RGB]}\expandafter\cbRGB\expandafter{\detokenize{255,247,247}}{morbidity\strut} \setlength{\fboxsep}{0pt}\def\cbRGB{\colorbox[RGB]}\expandafter\cbRGB\expandafter{\detokenize{255,246,246}}{was\strut} \setlength{\fboxsep}{0pt}\def\cbRGB{\colorbox[RGB]}\expandafter\cbRGB\expandafter{\detokenize{255,246,246}}{observed\strut} \setlength{\fboxsep}{0pt}\def\cbRGB{\colorbox[RGB]}\expandafter\cbRGB\expandafter{\detokenize{255,245,245}}{between\strut} \setlength{\fboxsep}{0pt}\def\cbRGB{\colorbox[RGB]}\expandafter\cbRGB\expandafter{\detokenize{255,245,245}}{the\strut} \setlength{\fboxsep}{0pt}\def\cbRGB{\colorbox[RGB]}\expandafter\cbRGB\expandafter{\detokenize{255,245,245}}{orally\strut} \setlength{\fboxsep}{0pt}\def\cbRGB{\colorbox[RGB]}\expandafter\cbRGB\expandafter{\detokenize{255,246,246}}{treated\strut} \setlength{\fboxsep}{0pt}\def\cbRGB{\colorbox[RGB]}\expandafter\cbRGB\expandafter{\detokenize{255,246,246}}{and\strut} \setlength{\fboxsep}{0pt}\def\cbRGB{\colorbox[RGB]}\expandafter\cbRGB\expandafter{\detokenize{255,246,246}}{unk\strut} \setlength{\fboxsep}{0pt}\def\cbRGB{\colorbox[RGB]}\expandafter\cbRGB\expandafter{\detokenize{255,246,246}}{group\strut} \setlength{\fboxsep}{0pt}\def\cbRGB{\colorbox[RGB]}\expandafter\cbRGB\expandafter{\detokenize{255,246,246}}{;\strut} \setlength{\fboxsep}{0pt}\def\cbRGB{\colorbox[RGB]}\expandafter\cbRGB\expandafter{\detokenize{255,245,245}}{no\strut} \setlength{\fboxsep}{0pt}\def\cbRGB{\colorbox[RGB]}\expandafter\cbRGB\expandafter{\detokenize{255,241,241}}{cases\strut} \setlength{\fboxsep}{0pt}\def\cbRGB{\colorbox[RGB]}\expandafter\cbRGB\expandafter{\detokenize{255,240,240}}{of\strut} \setlength{\fboxsep}{0pt}\def\cbRGB{\colorbox[RGB]}\expandafter\cbRGB\expandafter{\detokenize{255,242,242}}{severe\strut} \setlength{\fboxsep}{0pt}\def\cbRGB{\colorbox[RGB]}\expandafter\cbRGB\expandafter{\detokenize{255,244,244}}{hypoglycaemia\strut} \setlength{\fboxsep}{0pt}\def\cbRGB{\colorbox[RGB]}\expandafter\cbRGB\expandafter{\detokenize{255,247,247}}{or\strut} \setlength{\fboxsep}{0pt}\def\cbRGB{\colorbox[RGB]}\expandafter\cbRGB\expandafter{\detokenize{255,247,247}}{jaundice\strut} \setlength{\fboxsep}{0pt}\def\cbRGB{\colorbox[RGB]}\expandafter\cbRGB\expandafter{\detokenize{255,248,248}}{were\strut} \setlength{\fboxsep}{0pt}\def\cbRGB{\colorbox[RGB]}\expandafter\cbRGB\expandafter{\detokenize{255,245,245}}{seen\strut} \setlength{\fboxsep}{0pt}\def\cbRGB{\colorbox[RGB]}\expandafter\cbRGB\expandafter{\detokenize{255,244,244}}{in\strut} \setlength{\fboxsep}{0pt}\def\cbRGB{\colorbox[RGB]}\expandafter\cbRGB\expandafter{\detokenize{255,244,244}}{the\strut} \setlength{\fboxsep}{0pt}\def\cbRGB{\colorbox[RGB]}\expandafter\cbRGB\expandafter{\detokenize{255,245,245}}{orally\strut} \setlength{\fboxsep}{0pt}\def\cbRGB{\colorbox[RGB]}\expandafter\cbRGB\expandafter{\detokenize{255,245,245}}{treated\strut} \setlength{\fboxsep}{0pt}\def\cbRGB{\colorbox[RGB]}\expandafter\cbRGB\expandafter{\detokenize{255,245,245}}{groups\strut} \setlength{\fboxsep}{0pt}\def\cbRGB{\colorbox[RGB]}\expandafter\cbRGB\expandafter{\detokenize{255,245,245}}{.\strut} \setlength{\fboxsep}{0pt}\def\cbRGB{\colorbox[RGB]}\expandafter\cbRGB\expandafter{\detokenize{255,244,244}}{however\strut} \setlength{\fboxsep}{0pt}\def\cbRGB{\colorbox[RGB]}\expandafter\cbRGB\expandafter{\detokenize{255,244,244}}{,\strut} \setlength{\fboxsep}{0pt}\def\cbRGB{\colorbox[RGB]}\expandafter\cbRGB\expandafter{\detokenize{255,244,244}}{in\strut} \setlength{\fboxsep}{0pt}\def\cbRGB{\colorbox[RGB]}\expandafter\cbRGB\expandafter{\detokenize{255,242,242}}{the\strut} \setlength{\fboxsep}{0pt}\def\cbRGB{\colorbox[RGB]}\expandafter\cbRGB\expandafter{\detokenize{255,237,237}}{group\strut} \setlength{\fboxsep}{0pt}\def\cbRGB{\colorbox[RGB]}\expandafter\cbRGB\expandafter{\detokenize{255,235,235}}{of\strut} \setlength{\fboxsep}{0pt}\def\cbRGB{\colorbox[RGB]}\expandafter\cbRGB\expandafter{\detokenize{255,228,228}}{women\strut} \setlength{\fboxsep}{0pt}\def\cbRGB{\colorbox[RGB]}\expandafter\cbRGB\expandafter{\detokenize{255,225,225}}{treated\strut} \setlength{\fboxsep}{0pt}\def\cbRGB{\colorbox[RGB]}\expandafter\cbRGB\expandafter{\detokenize{255,221,221}}{with\strut} \setlength{\fboxsep}{0pt}\def\cbRGB{\colorbox[RGB]}\expandafter\cbRGB\expandafter{\detokenize{255,225,225}}{metformin\strut} \setlength{\fboxsep}{0pt}\def\cbRGB{\colorbox[RGB]}\expandafter\cbRGB\expandafter{\detokenize{255,228,228}}{in\strut} \setlength{\fboxsep}{0pt}\def\cbRGB{\colorbox[RGB]}\expandafter\cbRGB\expandafter{\detokenize{255,228,228}}{the\strut} \setlength{\fboxsep}{0pt}\def\cbRGB{\colorbox[RGB]}\expandafter\cbRGB\expandafter{\detokenize{255,231,231}}{third\strut} \setlength{\fboxsep}{0pt}\def\cbRGB{\colorbox[RGB]}\expandafter\cbRGB\expandafter{\detokenize{255,220,220}}{trimester\strut} \setlength{\fboxsep}{0pt}\def\cbRGB{\colorbox[RGB]}\expandafter\cbRGB\expandafter{\detokenize{255,209,209}}{,\strut} \setlength{\fboxsep}{0pt}\def\cbRGB{\colorbox[RGB]}\expandafter\cbRGB\expandafter{\detokenize{255,214,214}}{the\strut} \setlength{\fboxsep}{0pt}\def\cbRGB{\colorbox[RGB]}\expandafter\cbRGB\expandafter{\detokenize{255,230,230}}{perinatal\strut} \setlength{\fboxsep}{0pt}\def\cbRGB{\colorbox[RGB]}\expandafter\cbRGB\expandafter{\detokenize{255,242,242}}{mortality\strut} \setlength{\fboxsep}{0pt}\def\cbRGB{\colorbox[RGB]}\expandafter\cbRGB\expandafter{\detokenize{255,241,241}}{was\strut} \setlength{\fboxsep}{0pt}\def\cbRGB{\colorbox[RGB]}\expandafter\cbRGB\expandafter{\detokenize{255,241,241}}{significantly\strut} \setlength{\fboxsep}{0pt}\def\cbRGB{\colorbox[RGB]}\expandafter\cbRGB\expandafter{\detokenize{255,242,242}}{increased\strut} \setlength{\fboxsep}{0pt}\def\cbRGB{\colorbox[RGB]}\expandafter\cbRGB\expandafter{\detokenize{255,235,235}}{compared\strut} \setlength{\fboxsep}{0pt}\def\cbRGB{\colorbox[RGB]}\expandafter\cbRGB\expandafter{\detokenize{255,236,236}}{to\strut} \setlength{\fboxsep}{0pt}\def\cbRGB{\colorbox[RGB]}\expandafter\cbRGB\expandafter{\detokenize{255,236,236}}{women\strut} \setlength{\fboxsep}{0pt}\def\cbRGB{\colorbox[RGB]}\expandafter\cbRGB\expandafter{\detokenize{255,235,235}}{not\strut} \setlength{\fboxsep}{0pt}\def\cbRGB{\colorbox[RGB]}\expandafter\cbRGB\expandafter{\detokenize{255,232,232}}{treated\strut} \setlength{\fboxsep}{0pt}\def\cbRGB{\colorbox[RGB]}\expandafter\cbRGB\expandafter{\detokenize{255,234,234}}{with\strut} \setlength{\fboxsep}{0pt}\def\cbRGB{\colorbox[RGB]}\expandafter\cbRGB\expandafter{\detokenize{255,237,237}}{metformin\strut} \setlength{\fboxsep}{0pt}\def\cbRGB{\colorbox[RGB]}\expandafter\cbRGB\expandafter{\detokenize{255,240,240}}{(\strut} \setlength{\fboxsep}{0pt}\def\cbRGB{\colorbox[RGB]}\expandafter\cbRGB\expandafter{\detokenize{255,243,243}}{qqq\strut} \setlength{\fboxsep}{0pt}\def\cbRGB{\colorbox[RGB]}\expandafter\cbRGB\expandafter{\detokenize{255,246,246}}{vs.\strut} \setlength{\fboxsep}{0pt}\def\cbRGB{\colorbox[RGB]}\expandafter\cbRGB\expandafter{\detokenize{255,246,246}}{qqq\strut} \setlength{\fboxsep}{0pt}\def\cbRGB{\colorbox[RGB]}\expandafter\cbRGB\expandafter{\detokenize{255,245,245}}{\%\strut} \setlength{\fboxsep}{0pt}\def\cbRGB{\colorbox[RGB]}\expandafter\cbRGB\expandafter{\detokenize{255,245,245}}{,\strut} \setlength{\fboxsep}{0pt}\def\cbRGB{\colorbox[RGB]}\expandafter\cbRGB\expandafter{\detokenize{255,244,244}}{p\strut} \setlength{\fboxsep}{0pt}\def\cbRGB{\colorbox[RGB]}\expandafter\cbRGB\expandafter{\detokenize{255,244,244}}{<\strut} \setlength{\fboxsep}{0pt}\def\cbRGB{\colorbox[RGB]}\expandafter\cbRGB\expandafter{\detokenize{255,244,244}}{qqq\strut} \setlength{\fboxsep}{0pt}\def\cbRGB{\colorbox[RGB]}\expandafter\cbRGB\expandafter{\detokenize{255,245,245}}{)\strut} \setlength{\fboxsep}{0pt}\def\cbRGB{\colorbox[RGB]}\expandafter\cbRGB\expandafter{\detokenize{255,246,246}}{.\strut} \setlength{\fboxsep}{0pt}\def\cbRGB{\colorbox[RGB]}\expandafter\cbRGB\expandafter{\detokenize{255,244,244}}{conclusion\strut} \setlength{\fboxsep}{0pt}\def\cbRGB{\colorbox[RGB]}\expandafter\cbRGB\expandafter{\detokenize{255,241,241}}{:\strut} \setlength{\fboxsep}{0pt}\def\cbRGB{\colorbox[RGB]}\expandafter\cbRGB\expandafter{\detokenize{255,237,237}}{treatment\strut} \setlength{\fboxsep}{0pt}\def\cbRGB{\colorbox[RGB]}\expandafter\cbRGB\expandafter{\detokenize{255,228,228}}{with\strut} \setlength{\fboxsep}{0pt}\def\cbRGB{\colorbox[RGB]}\expandafter\cbRGB\expandafter{\detokenize{255,205,205}}{metformin\strut} \setlength{\fboxsep}{0pt}\def\cbRGB{\colorbox[RGB]}\expandafter\cbRGB\expandafter{\detokenize{255,203,203}}{during\strut} \setlength{\fboxsep}{0pt}\def\cbRGB{\colorbox[RGB]}\expandafter\cbRGB\expandafter{\detokenize{255,201,201}}{pregnancy\strut} \setlength{\fboxsep}{0pt}\def\cbRGB{\colorbox[RGB]}\expandafter\cbRGB\expandafter{\detokenize{255,213,213}}{was\strut} \setlength{\fboxsep}{0pt}\def\cbRGB{\colorbox[RGB]}\expandafter\cbRGB\expandafter{\detokenize{255,209,209}}{associated\strut} \setlength{\fboxsep}{0pt}\def\cbRGB{\colorbox[RGB]}\expandafter\cbRGB\expandafter{\detokenize{255,219,219}}{with\strut} \setlength{\fboxsep}{0pt}\def\cbRGB{\colorbox[RGB]}\expandafter\cbRGB\expandafter{\detokenize{255,222,222}}{increased\strut} \setlength{\fboxsep}{0pt}\def\cbRGB{\colorbox[RGB]}\expandafter\cbRGB\expandafter{\detokenize{255,217,217}}{prevalence\strut} \setlength{\fboxsep}{0pt}\def\cbRGB{\colorbox[RGB]}\expandafter\cbRGB\expandafter{\detokenize{255,222,222}}{of\strut} \setlength{\fboxsep}{0pt}\def\cbRGB{\colorbox[RGB]}\expandafter\cbRGB\expandafter{\detokenize{255,229,229}}{pre-eclampsia\strut} \setlength{\fboxsep}{0pt}\def\cbRGB{\colorbox[RGB]}\expandafter\cbRGB\expandafter{\detokenize{255,228,228}}{and\strut} \setlength{\fboxsep}{0pt}\def\cbRGB{\colorbox[RGB]}\expandafter\cbRGB\expandafter{\detokenize{255,217,217}}{a\strut} \setlength{\fboxsep}{0pt}\def\cbRGB{\colorbox[RGB]}\expandafter\cbRGB\expandafter{\detokenize{255,225,225}}{high\strut} \setlength{\fboxsep}{0pt}\def\cbRGB{\colorbox[RGB]}\expandafter\cbRGB\expandafter{\detokenize{255,239,239}}{perinatal\strut} \setlength{\fboxsep}{0pt}\def\cbRGB{\colorbox[RGB]}\expandafter\cbRGB\expandafter{\detokenize{255,248,248}}{mortality\strut} \setlength{\fboxsep}{0pt}\def\cbRGB{\colorbox[RGB]}\expandafter\cbRGB\expandafter{\detokenize{255,249,249}}{.\strut} 

\setlength{\fboxsep}{0pt}\def\cbRGB{\colorbox[RGB]}\expandafter\cbRGB\expandafter{\detokenize{210,255,210}}{aims\strut} \setlength{\fboxsep}{0pt}\def\cbRGB{\colorbox[RGB]}\expandafter\cbRGB\expandafter{\detokenize{145,255,145}}{:\strut} \setlength{\fboxsep}{0pt}\def\cbRGB{\colorbox[RGB]}\expandafter\cbRGB\expandafter{\detokenize{179,255,179}}{to\strut} \setlength{\fboxsep}{0pt}\def\cbRGB{\colorbox[RGB]}\expandafter\cbRGB\expandafter{\detokenize{181,255,181}}{assess\strut} \setlength{\fboxsep}{0pt}\def\cbRGB{\colorbox[RGB]}\expandafter\cbRGB\expandafter{\detokenize{212,255,212}}{maternal\strut} \setlength{\fboxsep}{0pt}\def\cbRGB{\colorbox[RGB]}\expandafter\cbRGB\expandafter{\detokenize{215,255,215}}{and\strut} \setlength{\fboxsep}{0pt}\def\cbRGB{\colorbox[RGB]}\expandafter\cbRGB\expandafter{\detokenize{229,255,229}}{neonatal\strut} \setlength{\fboxsep}{0pt}\def\cbRGB{\colorbox[RGB]}\expandafter\cbRGB\expandafter{\detokenize{219,255,219}}{complications\strut} \setlength{\fboxsep}{0pt}\def\cbRGB{\colorbox[RGB]}\expandafter\cbRGB\expandafter{\detokenize{235,255,235}}{in\strut} \setlength{\fboxsep}{0pt}\def\cbRGB{\colorbox[RGB]}\expandafter\cbRGB\expandafter{\detokenize{227,255,227}}{pregnancies\strut} \setlength{\fboxsep}{0pt}\def\cbRGB{\colorbox[RGB]}\expandafter\cbRGB\expandafter{\detokenize{232,255,232}}{of\strut} \setlength{\fboxsep}{0pt}\def\cbRGB{\colorbox[RGB]}\expandafter\cbRGB\expandafter{\detokenize{204,255,204}}{diabetic\strut} \setlength{\fboxsep}{0pt}\def\cbRGB{\colorbox[RGB]}\expandafter\cbRGB\expandafter{\detokenize{196,255,196}}{women\strut} \setlength{\fboxsep}{0pt}\def\cbRGB{\colorbox[RGB]}\expandafter\cbRGB\expandafter{\detokenize{198,255,198}}{treated\strut} \setlength{\fboxsep}{0pt}\def\cbRGB{\colorbox[RGB]}\expandafter\cbRGB\expandafter{\detokenize{210,255,210}}{with\strut} \setlength{\fboxsep}{0pt}\def\cbRGB{\colorbox[RGB]}\expandafter\cbRGB\expandafter{\detokenize{197,255,197}}{oral\strut} \setlength{\fboxsep}{0pt}\def\cbRGB{\colorbox[RGB]}\expandafter\cbRGB\expandafter{\detokenize{172,255,172}}{hypoglycaemic\strut} \setlength{\fboxsep}{0pt}\def\cbRGB{\colorbox[RGB]}\expandafter\cbRGB\expandafter{\detokenize{180,255,180}}{agents\strut} \setlength{\fboxsep}{0pt}\def\cbRGB{\colorbox[RGB]}\expandafter\cbRGB\expandafter{\detokenize{193,255,193}}{during\strut} \setlength{\fboxsep}{0pt}\def\cbRGB{\colorbox[RGB]}\expandafter\cbRGB\expandafter{\detokenize{215,255,215}}{pregnancy\strut} \setlength{\fboxsep}{0pt}\def\cbRGB{\colorbox[RGB]}\expandafter\cbRGB\expandafter{\detokenize{192,255,192}}{.\strut} \setlength{\fboxsep}{0pt}\def\cbRGB{\colorbox[RGB]}\expandafter\cbRGB\expandafter{\detokenize{201,255,201}}{methods\strut} \setlength{\fboxsep}{0pt}\def\cbRGB{\colorbox[RGB]}\expandafter\cbRGB\expandafter{\detokenize{178,255,178}}{:\strut} \setlength{\fboxsep}{0pt}\def\cbRGB{\colorbox[RGB]}\expandafter\cbRGB\expandafter{\detokenize{191,255,191}}{a\strut} \setlength{\fboxsep}{0pt}\def\cbRGB{\colorbox[RGB]}\expandafter\cbRGB\expandafter{\detokenize{179,255,179}}{cohort\strut} \setlength{\fboxsep}{0pt}\def\cbRGB{\colorbox[RGB]}\expandafter\cbRGB\expandafter{\detokenize{198,255,198}}{study\strut} \setlength{\fboxsep}{0pt}\def\cbRGB{\colorbox[RGB]}\expandafter\cbRGB\expandafter{\detokenize{182,255,182}}{including\strut} \setlength{\fboxsep}{0pt}\def\cbRGB{\colorbox[RGB]}\expandafter\cbRGB\expandafter{\detokenize{168,255,168}}{all\strut} \setlength{\fboxsep}{0pt}\def\cbRGB{\colorbox[RGB]}\expandafter\cbRGB\expandafter{\detokenize{145,255,145}}{unk\strut} \setlength{\fboxsep}{0pt}\def\cbRGB{\colorbox[RGB]}\expandafter\cbRGB\expandafter{\detokenize{126,255,126}}{registered\strut} \setlength{\fboxsep}{0pt}\def\cbRGB{\colorbox[RGB]}\expandafter\cbRGB\expandafter{\detokenize{141,255,141}}{,\strut} \setlength{\fboxsep}{0pt}\def\cbRGB{\colorbox[RGB]}\expandafter\cbRGB\expandafter{\detokenize{173,255,173}}{orally\strut} \setlength{\fboxsep}{0pt}\def\cbRGB{\colorbox[RGB]}\expandafter\cbRGB\expandafter{\detokenize{212,255,212}}{treated\strut} \setlength{\fboxsep}{0pt}\def\cbRGB{\colorbox[RGB]}\expandafter\cbRGB\expandafter{\detokenize{217,255,217}}{pregnant\strut} \setlength{\fboxsep}{0pt}\def\cbRGB{\colorbox[RGB]}\expandafter\cbRGB\expandafter{\detokenize{223,255,223}}{diabetic\strut} \setlength{\fboxsep}{0pt}\def\cbRGB{\colorbox[RGB]}\expandafter\cbRGB\expandafter{\detokenize{222,255,222}}{patients\strut} \setlength{\fboxsep}{0pt}\def\cbRGB{\colorbox[RGB]}\expandafter\cbRGB\expandafter{\detokenize{226,255,226}}{set\strut} \setlength{\fboxsep}{0pt}\def\cbRGB{\colorbox[RGB]}\expandafter\cbRGB\expandafter{\detokenize{213,255,213}}{in\strut} \setlength{\fboxsep}{0pt}\def\cbRGB{\colorbox[RGB]}\expandafter\cbRGB\expandafter{\detokenize{200,255,200}}{a\strut} \setlength{\fboxsep}{0pt}\def\cbRGB{\colorbox[RGB]}\expandafter\cbRGB\expandafter{\detokenize{168,255,168}}{diabetic\strut} \setlength{\fboxsep}{0pt}\def\cbRGB{\colorbox[RGB]}\expandafter\cbRGB\expandafter{\detokenize{173,255,173}}{unk\strut} \setlength{\fboxsep}{0pt}\def\cbRGB{\colorbox[RGB]}\expandafter\cbRGB\expandafter{\detokenize{160,255,160}}{service\strut} \setlength{\fboxsep}{0pt}\def\cbRGB{\colorbox[RGB]}\expandafter\cbRGB\expandafter{\detokenize{169,255,169}}{at\strut} \setlength{\fboxsep}{0pt}\def\cbRGB{\colorbox[RGB]}\expandafter\cbRGB\expandafter{\detokenize{156,255,156}}{a\strut} \setlength{\fboxsep}{0pt}\def\cbRGB{\colorbox[RGB]}\expandafter\cbRGB\expandafter{\detokenize{131,255,131}}{university\strut} \setlength{\fboxsep}{0pt}\def\cbRGB{\colorbox[RGB]}\expandafter\cbRGB\expandafter{\detokenize{149,255,149}}{hospital\strut} \setlength{\fboxsep}{0pt}\def\cbRGB{\colorbox[RGB]}\expandafter\cbRGB\expandafter{\detokenize{131,255,131}}{:\strut} \setlength{\fboxsep}{0pt}\def\cbRGB{\colorbox[RGB]}\expandafter\cbRGB\expandafter{\detokenize{179,255,179}}{qqq\strut} \setlength{\fboxsep}{0pt}\def\cbRGB{\colorbox[RGB]}\expandafter\cbRGB\expandafter{\detokenize{157,255,157}}{women\strut} \setlength{\fboxsep}{0pt}\def\cbRGB{\colorbox[RGB]}\expandafter\cbRGB\expandafter{\detokenize{154,255,154}}{treated\strut} \setlength{\fboxsep}{0pt}\def\cbRGB{\colorbox[RGB]}\expandafter\cbRGB\expandafter{\detokenize{93,255,93}}{with\strut} \setlength{\fboxsep}{0pt}\def\cbRGB{\colorbox[RGB]}\expandafter\cbRGB\expandafter{\detokenize{35,255,35}}{metformin\strut} \setlength{\fboxsep}{0pt}\def\cbRGB{\colorbox[RGB]}\expandafter\cbRGB\expandafter{\detokenize{13,255,13}}{,\strut} \setlength{\fboxsep}{0pt}\def\cbRGB{\colorbox[RGB]}\expandafter\cbRGB\expandafter{\detokenize{62,255,62}}{qqq\strut} \setlength{\fboxsep}{0pt}\def\cbRGB{\colorbox[RGB]}\expandafter\cbRGB\expandafter{\detokenize{128,255,128}}{women\strut} \setlength{\fboxsep}{0pt}\def\cbRGB{\colorbox[RGB]}\expandafter\cbRGB\expandafter{\detokenize{148,255,148}}{treated\strut} \setlength{\fboxsep}{0pt}\def\cbRGB{\colorbox[RGB]}\expandafter\cbRGB\expandafter{\detokenize{148,255,148}}{with\strut} \setlength{\fboxsep}{0pt}\def\cbRGB{\colorbox[RGB]}\expandafter\cbRGB\expandafter{\detokenize{126,255,126}}{sulphonylurea\strut} \setlength{\fboxsep}{0pt}\def\cbRGB{\colorbox[RGB]}\expandafter\cbRGB\expandafter{\detokenize{145,255,145}}{during\strut} \setlength{\fboxsep}{0pt}\def\cbRGB{\colorbox[RGB]}\expandafter\cbRGB\expandafter{\detokenize{147,255,147}}{pregnancy\strut} \setlength{\fboxsep}{0pt}\def\cbRGB{\colorbox[RGB]}\expandafter\cbRGB\expandafter{\detokenize{182,255,182}}{and\strut} \setlength{\fboxsep}{0pt}\def\cbRGB{\colorbox[RGB]}\expandafter\cbRGB\expandafter{\detokenize{193,255,193}}{a\strut} \setlength{\fboxsep}{0pt}\def\cbRGB{\colorbox[RGB]}\expandafter\cbRGB\expandafter{\detokenize{179,255,179}}{reference\strut} \setlength{\fboxsep}{0pt}\def\cbRGB{\colorbox[RGB]}\expandafter\cbRGB\expandafter{\detokenize{172,255,172}}{group\strut} \setlength{\fboxsep}{0pt}\def\cbRGB{\colorbox[RGB]}\expandafter\cbRGB\expandafter{\detokenize{179,255,179}}{of\strut} \setlength{\fboxsep}{0pt}\def\cbRGB{\colorbox[RGB]}\expandafter\cbRGB\expandafter{\detokenize{202,255,202}}{qqq\strut} \setlength{\fboxsep}{0pt}\def\cbRGB{\colorbox[RGB]}\expandafter\cbRGB\expandafter{\detokenize{204,255,204}}{diabetic\strut} \setlength{\fboxsep}{0pt}\def\cbRGB{\colorbox[RGB]}\expandafter\cbRGB\expandafter{\detokenize{212,255,212}}{women\strut} \setlength{\fboxsep}{0pt}\def\cbRGB{\colorbox[RGB]}\expandafter\cbRGB\expandafter{\detokenize{209,255,209}}{treated\strut} \setlength{\fboxsep}{0pt}\def\cbRGB{\colorbox[RGB]}\expandafter\cbRGB\expandafter{\detokenize{214,255,214}}{with\strut} \setlength{\fboxsep}{0pt}\def\cbRGB{\colorbox[RGB]}\expandafter\cbRGB\expandafter{\detokenize{194,255,194}}{insulin\strut} \setlength{\fboxsep}{0pt}\def\cbRGB{\colorbox[RGB]}\expandafter\cbRGB\expandafter{\detokenize{198,255,198}}{during\strut} \setlength{\fboxsep}{0pt}\def\cbRGB{\colorbox[RGB]}\expandafter\cbRGB\expandafter{\detokenize{195,255,195}}{pregnancy\strut} \setlength{\fboxsep}{0pt}\def\cbRGB{\colorbox[RGB]}\expandafter\cbRGB\expandafter{\detokenize{191,255,191}}{.\strut} \setlength{\fboxsep}{0pt}\def\cbRGB{\colorbox[RGB]}\expandafter\cbRGB\expandafter{\detokenize{194,255,194}}{results\strut} \setlength{\fboxsep}{0pt}\def\cbRGB{\colorbox[RGB]}\expandafter\cbRGB\expandafter{\detokenize{176,255,176}}{:\strut} \setlength{\fboxsep}{0pt}\def\cbRGB{\colorbox[RGB]}\expandafter\cbRGB\expandafter{\detokenize{187,255,187}}{the\strut} \setlength{\fboxsep}{0pt}\def\cbRGB{\colorbox[RGB]}\expandafter\cbRGB\expandafter{\detokenize{160,255,160}}{prevalence\strut} \setlength{\fboxsep}{0pt}\def\cbRGB{\colorbox[RGB]}\expandafter\cbRGB\expandafter{\detokenize{179,255,179}}{of\strut} \setlength{\fboxsep}{0pt}\def\cbRGB{\colorbox[RGB]}\expandafter\cbRGB\expandafter{\detokenize{145,255,145}}{pre-eclampsia\strut} \setlength{\fboxsep}{0pt}\def\cbRGB{\colorbox[RGB]}\expandafter\cbRGB\expandafter{\detokenize{160,255,160}}{was\strut} \setlength{\fboxsep}{0pt}\def\cbRGB{\colorbox[RGB]}\expandafter\cbRGB\expandafter{\detokenize{155,255,155}}{significantly\strut} \setlength{\fboxsep}{0pt}\def\cbRGB{\colorbox[RGB]}\expandafter\cbRGB\expandafter{\detokenize{186,255,186}}{increased\strut} \setlength{\fboxsep}{0pt}\def\cbRGB{\colorbox[RGB]}\expandafter\cbRGB\expandafter{\detokenize{185,255,185}}{in\strut} \setlength{\fboxsep}{0pt}\def\cbRGB{\colorbox[RGB]}\expandafter\cbRGB\expandafter{\detokenize{189,255,189}}{the\strut} \setlength{\fboxsep}{0pt}\def\cbRGB{\colorbox[RGB]}\expandafter\cbRGB\expandafter{\detokenize{190,255,190}}{group\strut} \setlength{\fboxsep}{0pt}\def\cbRGB{\colorbox[RGB]}\expandafter\cbRGB\expandafter{\detokenize{194,255,194}}{of\strut} \setlength{\fboxsep}{0pt}\def\cbRGB{\colorbox[RGB]}\expandafter\cbRGB\expandafter{\detokenize{173,255,173}}{women\strut} \setlength{\fboxsep}{0pt}\def\cbRGB{\colorbox[RGB]}\expandafter\cbRGB\expandafter{\detokenize{143,255,143}}{treated\strut} \setlength{\fboxsep}{0pt}\def\cbRGB{\colorbox[RGB]}\expandafter\cbRGB\expandafter{\detokenize{85,255,85}}{with\strut} \setlength{\fboxsep}{0pt}\def\cbRGB{\colorbox[RGB]}\expandafter\cbRGB\expandafter{\detokenize{24,255,24}}{metformin\strut} \setlength{\fboxsep}{0pt}\def\cbRGB{\colorbox[RGB]}\expandafter\cbRGB\expandafter{\detokenize{7,255,7}}{compared\strut} \setlength{\fboxsep}{0pt}\def\cbRGB{\colorbox[RGB]}\expandafter\cbRGB\expandafter{\detokenize{53,255,53}}{to\strut} \setlength{\fboxsep}{0pt}\def\cbRGB{\colorbox[RGB]}\expandafter\cbRGB\expandafter{\detokenize{124,255,124}}{women\strut} \setlength{\fboxsep}{0pt}\def\cbRGB{\colorbox[RGB]}\expandafter\cbRGB\expandafter{\detokenize{143,255,143}}{treated\strut} \setlength{\fboxsep}{0pt}\def\cbRGB{\colorbox[RGB]}\expandafter\cbRGB\expandafter{\detokenize{147,255,147}}{with\strut} \setlength{\fboxsep}{0pt}\def\cbRGB{\colorbox[RGB]}\expandafter\cbRGB\expandafter{\detokenize{101,255,101}}{sulphonylurea\strut} \setlength{\fboxsep}{0pt}\def\cbRGB{\colorbox[RGB]}\expandafter\cbRGB\expandafter{\detokenize{101,255,101}}{or\strut} \setlength{\fboxsep}{0pt}\def\cbRGB{\colorbox[RGB]}\expandafter\cbRGB\expandafter{\detokenize{78,255,78}}{insulin\strut} \setlength{\fboxsep}{0pt}\def\cbRGB{\colorbox[RGB]}\expandafter\cbRGB\expandafter{\detokenize{132,255,132}}{(\strut} \setlength{\fboxsep}{0pt}\def\cbRGB{\colorbox[RGB]}\expandafter\cbRGB\expandafter{\detokenize{153,255,153}}{qqq\strut} \setlength{\fboxsep}{0pt}\def\cbRGB{\colorbox[RGB]}\expandafter\cbRGB\expandafter{\detokenize{184,255,184}}{vs.\strut} \setlength{\fboxsep}{0pt}\def\cbRGB{\colorbox[RGB]}\expandafter\cbRGB\expandafter{\detokenize{182,255,182}}{qqq\strut} \setlength{\fboxsep}{0pt}\def\cbRGB{\colorbox[RGB]}\expandafter\cbRGB\expandafter{\detokenize{187,255,187}}{vs.\strut} \setlength{\fboxsep}{0pt}\def\cbRGB{\colorbox[RGB]}\expandafter\cbRGB\expandafter{\detokenize{180,255,180}}{qqq\strut} \setlength{\fboxsep}{0pt}\def\cbRGB{\colorbox[RGB]}\expandafter\cbRGB\expandafter{\detokenize{185,255,185}}{\%\strut} \setlength{\fboxsep}{0pt}\def\cbRGB{\colorbox[RGB]}\expandafter\cbRGB\expandafter{\detokenize{182,255,182}}{,\strut} \setlength{\fboxsep}{0pt}\def\cbRGB{\colorbox[RGB]}\expandafter\cbRGB\expandafter{\detokenize{188,255,188}}{p\strut} \setlength{\fboxsep}{0pt}\def\cbRGB{\colorbox[RGB]}\expandafter\cbRGB\expandafter{\detokenize{184,255,184}}{<\strut} \setlength{\fboxsep}{0pt}\def\cbRGB{\colorbox[RGB]}\expandafter\cbRGB\expandafter{\detokenize{187,255,187}}{qqq\strut} \setlength{\fboxsep}{0pt}\def\cbRGB{\colorbox[RGB]}\expandafter\cbRGB\expandafter{\detokenize{188,255,188}}{)\strut} \setlength{\fboxsep}{0pt}\def\cbRGB{\colorbox[RGB]}\expandafter\cbRGB\expandafter{\detokenize{191,255,191}}{.\strut} \setlength{\fboxsep}{0pt}\def\cbRGB{\colorbox[RGB]}\expandafter\cbRGB\expandafter{\detokenize{191,255,191}}{no\strut} \setlength{\fboxsep}{0pt}\def\cbRGB{\colorbox[RGB]}\expandafter\cbRGB\expandafter{\detokenize{194,255,194}}{difference\strut} \setlength{\fboxsep}{0pt}\def\cbRGB{\colorbox[RGB]}\expandafter\cbRGB\expandafter{\detokenize{194,255,194}}{in\strut} \setlength{\fboxsep}{0pt}\def\cbRGB{\colorbox[RGB]}\expandafter\cbRGB\expandafter{\detokenize{182,255,182}}{neonatal\strut} \setlength{\fboxsep}{0pt}\def\cbRGB{\colorbox[RGB]}\expandafter\cbRGB\expandafter{\detokenize{162,255,162}}{morbidity\strut} \setlength{\fboxsep}{0pt}\def\cbRGB{\colorbox[RGB]}\expandafter\cbRGB\expandafter{\detokenize{163,255,163}}{was\strut} \setlength{\fboxsep}{0pt}\def\cbRGB{\colorbox[RGB]}\expandafter\cbRGB\expandafter{\detokenize{144,255,144}}{observed\strut} \setlength{\fboxsep}{0pt}\def\cbRGB{\colorbox[RGB]}\expandafter\cbRGB\expandafter{\detokenize{127,255,127}}{between\strut} \setlength{\fboxsep}{0pt}\def\cbRGB{\colorbox[RGB]}\expandafter\cbRGB\expandafter{\detokenize{124,255,124}}{the\strut} \setlength{\fboxsep}{0pt}\def\cbRGB{\colorbox[RGB]}\expandafter\cbRGB\expandafter{\detokenize{145,255,145}}{orally\strut} \setlength{\fboxsep}{0pt}\def\cbRGB{\colorbox[RGB]}\expandafter\cbRGB\expandafter{\detokenize{169,255,169}}{treated\strut} \setlength{\fboxsep}{0pt}\def\cbRGB{\colorbox[RGB]}\expandafter\cbRGB\expandafter{\detokenize{175,255,175}}{and\strut} \setlength{\fboxsep}{0pt}\def\cbRGB{\colorbox[RGB]}\expandafter\cbRGB\expandafter{\detokenize{187,255,187}}{unk\strut} \setlength{\fboxsep}{0pt}\def\cbRGB{\colorbox[RGB]}\expandafter\cbRGB\expandafter{\detokenize{200,255,200}}{group\strut} \setlength{\fboxsep}{0pt}\def\cbRGB{\colorbox[RGB]}\expandafter\cbRGB\expandafter{\detokenize{197,255,197}}{;\strut} \setlength{\fboxsep}{0pt}\def\cbRGB{\colorbox[RGB]}\expandafter\cbRGB\expandafter{\detokenize{199,255,199}}{no\strut} \setlength{\fboxsep}{0pt}\def\cbRGB{\colorbox[RGB]}\expandafter\cbRGB\expandafter{\detokenize{218,255,218}}{cases\strut} \setlength{\fboxsep}{0pt}\def\cbRGB{\colorbox[RGB]}\expandafter\cbRGB\expandafter{\detokenize{239,255,239}}{of\strut} \setlength{\fboxsep}{0pt}\def\cbRGB{\colorbox[RGB]}\expandafter\cbRGB\expandafter{\detokenize{245,255,245}}{severe\strut} \setlength{\fboxsep}{0pt}\def\cbRGB{\colorbox[RGB]}\expandafter\cbRGB\expandafter{\detokenize{217,255,217}}{hypoglycaemia\strut} \setlength{\fboxsep}{0pt}\def\cbRGB{\colorbox[RGB]}\expandafter\cbRGB\expandafter{\detokenize{217,255,217}}{or\strut} \setlength{\fboxsep}{0pt}\def\cbRGB{\colorbox[RGB]}\expandafter\cbRGB\expandafter{\detokenize{204,255,204}}{jaundice\strut} \setlength{\fboxsep}{0pt}\def\cbRGB{\colorbox[RGB]}\expandafter\cbRGB\expandafter{\detokenize{215,255,215}}{were\strut} \setlength{\fboxsep}{0pt}\def\cbRGB{\colorbox[RGB]}\expandafter\cbRGB\expandafter{\detokenize{175,255,175}}{seen\strut} \setlength{\fboxsep}{0pt}\def\cbRGB{\colorbox[RGB]}\expandafter\cbRGB\expandafter{\detokenize{151,255,151}}{in\strut} \setlength{\fboxsep}{0pt}\def\cbRGB{\colorbox[RGB]}\expandafter\cbRGB\expandafter{\detokenize{140,255,140}}{the\strut} \setlength{\fboxsep}{0pt}\def\cbRGB{\colorbox[RGB]}\expandafter\cbRGB\expandafter{\detokenize{147,255,147}}{orally\strut} \setlength{\fboxsep}{0pt}\def\cbRGB{\colorbox[RGB]}\expandafter\cbRGB\expandafter{\detokenize{158,255,158}}{treated\strut} \setlength{\fboxsep}{0pt}\def\cbRGB{\colorbox[RGB]}\expandafter\cbRGB\expandafter{\detokenize{164,255,164}}{groups\strut} \setlength{\fboxsep}{0pt}\def\cbRGB{\colorbox[RGB]}\expandafter\cbRGB\expandafter{\detokenize{170,255,170}}{.\strut} \setlength{\fboxsep}{0pt}\def\cbRGB{\colorbox[RGB]}\expandafter\cbRGB\expandafter{\detokenize{185,255,185}}{however\strut} \setlength{\fboxsep}{0pt}\def\cbRGB{\colorbox[RGB]}\expandafter\cbRGB\expandafter{\detokenize{183,255,183}}{,\strut} \setlength{\fboxsep}{0pt}\def\cbRGB{\colorbox[RGB]}\expandafter\cbRGB\expandafter{\detokenize{187,255,187}}{in\strut} \setlength{\fboxsep}{0pt}\def\cbRGB{\colorbox[RGB]}\expandafter\cbRGB\expandafter{\detokenize{190,255,190}}{the\strut} \setlength{\fboxsep}{0pt}\def\cbRGB{\colorbox[RGB]}\expandafter\cbRGB\expandafter{\detokenize{192,255,192}}{group\strut} \setlength{\fboxsep}{0pt}\def\cbRGB{\colorbox[RGB]}\expandafter\cbRGB\expandafter{\detokenize{194,255,194}}{of\strut} \setlength{\fboxsep}{0pt}\def\cbRGB{\colorbox[RGB]}\expandafter\cbRGB\expandafter{\detokenize{173,255,173}}{women\strut} \setlength{\fboxsep}{0pt}\def\cbRGB{\colorbox[RGB]}\expandafter\cbRGB\expandafter{\detokenize{143,255,143}}{treated\strut} \setlength{\fboxsep}{0pt}\def\cbRGB{\colorbox[RGB]}\expandafter\cbRGB\expandafter{\detokenize{81,255,81}}{with\strut} \setlength{\fboxsep}{0pt}\def\cbRGB{\colorbox[RGB]}\expandafter\cbRGB\expandafter{\detokenize{21,255,21}}{metformin\strut} \setlength{\fboxsep}{0pt}\def\cbRGB{\colorbox[RGB]}\expandafter\cbRGB\expandafter{\detokenize{1,255,1}}{in\strut} \setlength{\fboxsep}{0pt}\def\cbRGB{\colorbox[RGB]}\expandafter\cbRGB\expandafter{\detokenize{55,255,55}}{the\strut} \setlength{\fboxsep}{0pt}\def\cbRGB{\colorbox[RGB]}\expandafter\cbRGB\expandafter{\detokenize{132,255,132}}{third\strut} \setlength{\fboxsep}{0pt}\def\cbRGB{\colorbox[RGB]}\expandafter\cbRGB\expandafter{\detokenize{183,255,183}}{trimester\strut} \setlength{\fboxsep}{0pt}\def\cbRGB{\colorbox[RGB]}\expandafter\cbRGB\expandafter{\detokenize{209,255,209}}{,\strut} \setlength{\fboxsep}{0pt}\def\cbRGB{\colorbox[RGB]}\expandafter\cbRGB\expandafter{\detokenize{230,255,230}}{the\strut} \setlength{\fboxsep}{0pt}\def\cbRGB{\colorbox[RGB]}\expandafter\cbRGB\expandafter{\detokenize{255,255,255}}{perinatal\strut} \setlength{\fboxsep}{0pt}\def\cbRGB{\colorbox[RGB]}\expandafter\cbRGB\expandafter{\detokenize{234,255,234}}{mortality\strut} \setlength{\fboxsep}{0pt}\def\cbRGB{\colorbox[RGB]}\expandafter\cbRGB\expandafter{\detokenize{239,255,239}}{was\strut} \setlength{\fboxsep}{0pt}\def\cbRGB{\colorbox[RGB]}\expandafter\cbRGB\expandafter{\detokenize{222,255,222}}{significantly\strut} \setlength{\fboxsep}{0pt}\def\cbRGB{\colorbox[RGB]}\expandafter\cbRGB\expandafter{\detokenize{221,255,221}}{increased\strut} \setlength{\fboxsep}{0pt}\def\cbRGB{\colorbox[RGB]}\expandafter\cbRGB\expandafter{\detokenize{199,255,199}}{compared\strut} \setlength{\fboxsep}{0pt}\def\cbRGB{\colorbox[RGB]}\expandafter\cbRGB\expandafter{\detokenize{190,255,190}}{to\strut} \setlength{\fboxsep}{0pt}\def\cbRGB{\colorbox[RGB]}\expandafter\cbRGB\expandafter{\detokenize{194,255,194}}{women\strut} \setlength{\fboxsep}{0pt}\def\cbRGB{\colorbox[RGB]}\expandafter\cbRGB\expandafter{\detokenize{175,255,175}}{not\strut} \setlength{\fboxsep}{0pt}\def\cbRGB{\colorbox[RGB]}\expandafter\cbRGB\expandafter{\detokenize{148,255,148}}{treated\strut} \setlength{\fboxsep}{0pt}\def\cbRGB{\colorbox[RGB]}\expandafter\cbRGB\expandafter{\detokenize{87,255,87}}{with\strut} \setlength{\fboxsep}{0pt}\def\cbRGB{\colorbox[RGB]}\expandafter\cbRGB\expandafter{\detokenize{20,255,20}}{metformin\strut} \setlength{\fboxsep}{0pt}\def\cbRGB{\colorbox[RGB]}\expandafter\cbRGB\expandafter{\detokenize{0,255,0}}{(\strut} \setlength{\fboxsep}{0pt}\def\cbRGB{\colorbox[RGB]}\expandafter\cbRGB\expandafter{\detokenize{48,255,48}}{qqq\strut} \setlength{\fboxsep}{0pt}\def\cbRGB{\colorbox[RGB]}\expandafter\cbRGB\expandafter{\detokenize{131,255,131}}{vs.\strut} \setlength{\fboxsep}{0pt}\def\cbRGB{\colorbox[RGB]}\expandafter\cbRGB\expandafter{\detokenize{174,255,174}}{qqq\strut} \setlength{\fboxsep}{0pt}\def\cbRGB{\colorbox[RGB]}\expandafter\cbRGB\expandafter{\detokenize{185,255,185}}{\%\strut} \setlength{\fboxsep}{0pt}\def\cbRGB{\colorbox[RGB]}\expandafter\cbRGB\expandafter{\detokenize{182,255,182}}{,\strut} \setlength{\fboxsep}{0pt}\def\cbRGB{\colorbox[RGB]}\expandafter\cbRGB\expandafter{\detokenize{188,255,188}}{p\strut} \setlength{\fboxsep}{0pt}\def\cbRGB{\colorbox[RGB]}\expandafter\cbRGB\expandafter{\detokenize{184,255,184}}{<\strut} \setlength{\fboxsep}{0pt}\def\cbRGB{\colorbox[RGB]}\expandafter\cbRGB\expandafter{\detokenize{185,255,185}}{qqq\strut} \setlength{\fboxsep}{0pt}\def\cbRGB{\colorbox[RGB]}\expandafter\cbRGB\expandafter{\detokenize{182,255,182}}{)\strut} \setlength{\fboxsep}{0pt}\def\cbRGB{\colorbox[RGB]}\expandafter\cbRGB\expandafter{\detokenize{155,255,155}}{.\strut} \setlength{\fboxsep}{0pt}\def\cbRGB{\colorbox[RGB]}\expandafter\cbRGB\expandafter{\detokenize{170,255,170}}{conclusion\strut} \setlength{\fboxsep}{0pt}\def\cbRGB{\colorbox[RGB]}\expandafter\cbRGB\expandafter{\detokenize{126,255,126}}{:\strut} \setlength{\fboxsep}{0pt}\def\cbRGB{\colorbox[RGB]}\expandafter\cbRGB\expandafter{\detokenize{139,255,139}}{treatment\strut} \setlength{\fboxsep}{0pt}\def\cbRGB{\colorbox[RGB]}\expandafter\cbRGB\expandafter{\detokenize{96,255,96}}{with\strut} \setlength{\fboxsep}{0pt}\def\cbRGB{\colorbox[RGB]}\expandafter\cbRGB\expandafter{\detokenize{95,255,95}}{metformin\strut} \setlength{\fboxsep}{0pt}\def\cbRGB{\colorbox[RGB]}\expandafter\cbRGB\expandafter{\detokenize{74,255,74}}{during\strut} \setlength{\fboxsep}{0pt}\def\cbRGB{\colorbox[RGB]}\expandafter\cbRGB\expandafter{\detokenize{92,255,92}}{pregnancy\strut} \setlength{\fboxsep}{0pt}\def\cbRGB{\colorbox[RGB]}\expandafter\cbRGB\expandafter{\detokenize{152,255,152}}{was\strut} \setlength{\fboxsep}{0pt}\def\cbRGB{\colorbox[RGB]}\expandafter\cbRGB\expandafter{\detokenize{191,255,191}}{associated\strut} \setlength{\fboxsep}{0pt}\def\cbRGB{\colorbox[RGB]}\expandafter\cbRGB\expandafter{\detokenize{194,255,194}}{with\strut} \setlength{\fboxsep}{0pt}\def\cbRGB{\colorbox[RGB]}\expandafter\cbRGB\expandafter{\detokenize{179,255,179}}{increased\strut} \setlength{\fboxsep}{0pt}\def\cbRGB{\colorbox[RGB]}\expandafter\cbRGB\expandafter{\detokenize{158,255,158}}{prevalence\strut} \setlength{\fboxsep}{0pt}\def\cbRGB{\colorbox[RGB]}\expandafter\cbRGB\expandafter{\detokenize{166,255,166}}{of\strut} \setlength{\fboxsep}{0pt}\def\cbRGB{\colorbox[RGB]}\expandafter\cbRGB\expandafter{\detokenize{137,255,137}}{pre-eclampsia\strut} \setlength{\fboxsep}{0pt}\def\cbRGB{\colorbox[RGB]}\expandafter\cbRGB\expandafter{\detokenize{154,255,154}}{and\strut} \setlength{\fboxsep}{0pt}\def\cbRGB{\colorbox[RGB]}\expandafter\cbRGB\expandafter{\detokenize{177,255,177}}{a\strut} \setlength{\fboxsep}{0pt}\def\cbRGB{\colorbox[RGB]}\expandafter\cbRGB\expandafter{\detokenize{222,255,222}}{high\strut} \setlength{\fboxsep}{0pt}\def\cbRGB{\colorbox[RGB]}\expandafter\cbRGB\expandafter{\detokenize{244,255,244}}{perinatal\strut} \setlength{\fboxsep}{0pt}\def\cbRGB{\colorbox[RGB]}\expandafter\cbRGB\expandafter{\detokenize{218,255,218}}{mortality\strut} \setlength{\fboxsep}{0pt}\def\cbRGB{\colorbox[RGB]}\expandafter\cbRGB\expandafter{\detokenize{249,255,249}}{.\strut} 

\setlength{\fboxsep}{0pt}\def\cbRGB{\colorbox[RGB]}\expandafter\cbRGB\expandafter{\detokenize{245,245,255}}{aims\strut} \setlength{\fboxsep}{0pt}\def\cbRGB{\colorbox[RGB]}\expandafter\cbRGB\expandafter{\detokenize{230,230,255}}{:\strut} \setlength{\fboxsep}{0pt}\def\cbRGB{\colorbox[RGB]}\expandafter\cbRGB\expandafter{\detokenize{232,232,255}}{to\strut} \setlength{\fboxsep}{0pt}\def\cbRGB{\colorbox[RGB]}\expandafter\cbRGB\expandafter{\detokenize{227,227,255}}{assess\strut} \setlength{\fboxsep}{0pt}\def\cbRGB{\colorbox[RGB]}\expandafter\cbRGB\expandafter{\detokenize{199,199,255}}{maternal\strut} \setlength{\fboxsep}{0pt}\def\cbRGB{\colorbox[RGB]}\expandafter\cbRGB\expandafter{\detokenize{170,170,255}}{and\strut} \setlength{\fboxsep}{0pt}\def\cbRGB{\colorbox[RGB]}\expandafter\cbRGB\expandafter{\detokenize{145,145,255}}{neonatal\strut} \setlength{\fboxsep}{0pt}\def\cbRGB{\colorbox[RGB]}\expandafter\cbRGB\expandafter{\detokenize{81,81,255}}{complications\strut} \setlength{\fboxsep}{0pt}\def\cbRGB{\colorbox[RGB]}\expandafter\cbRGB\expandafter{\detokenize{101,101,255}}{in\strut} \setlength{\fboxsep}{0pt}\def\cbRGB{\colorbox[RGB]}\expandafter\cbRGB\expandafter{\detokenize{121,121,255}}{pregnancies\strut} \setlength{\fboxsep}{0pt}\def\cbRGB{\colorbox[RGB]}\expandafter\cbRGB\expandafter{\detokenize{197,197,255}}{of\strut} \setlength{\fboxsep}{0pt}\def\cbRGB{\colorbox[RGB]}\expandafter\cbRGB\expandafter{\detokenize{205,205,255}}{diabetic\strut} \setlength{\fboxsep}{0pt}\def\cbRGB{\colorbox[RGB]}\expandafter\cbRGB\expandafter{\detokenize{216,216,255}}{women\strut} \setlength{\fboxsep}{0pt}\def\cbRGB{\colorbox[RGB]}\expandafter\cbRGB\expandafter{\detokenize{240,240,255}}{treated\strut} \setlength{\fboxsep}{0pt}\def\cbRGB{\colorbox[RGB]}\expandafter\cbRGB\expandafter{\detokenize{240,240,255}}{with\strut} \setlength{\fboxsep}{0pt}\def\cbRGB{\colorbox[RGB]}\expandafter\cbRGB\expandafter{\detokenize{244,244,255}}{oral\strut} \setlength{\fboxsep}{0pt}\def\cbRGB{\colorbox[RGB]}\expandafter\cbRGB\expandafter{\detokenize{236,236,255}}{hypoglycaemic\strut} \setlength{\fboxsep}{0pt}\def\cbRGB{\colorbox[RGB]}\expandafter\cbRGB\expandafter{\detokenize{229,229,255}}{agents\strut} \setlength{\fboxsep}{0pt}\def\cbRGB{\colorbox[RGB]}\expandafter\cbRGB\expandafter{\detokenize{190,190,255}}{during\strut} \setlength{\fboxsep}{0pt}\def\cbRGB{\colorbox[RGB]}\expandafter\cbRGB\expandafter{\detokenize{162,162,255}}{pregnancy\strut} \setlength{\fboxsep}{0pt}\def\cbRGB{\colorbox[RGB]}\expandafter\cbRGB\expandafter{\detokenize{122,122,255}}{.\strut} \setlength{\fboxsep}{0pt}\def\cbRGB{\colorbox[RGB]}\expandafter\cbRGB\expandafter{\detokenize{145,145,255}}{methods\strut} \setlength{\fboxsep}{0pt}\def\cbRGB{\colorbox[RGB]}\expandafter\cbRGB\expandafter{\detokenize{171,171,255}}{:\strut} \setlength{\fboxsep}{0pt}\def\cbRGB{\colorbox[RGB]}\expandafter\cbRGB\expandafter{\detokenize{225,225,255}}{a\strut} \setlength{\fboxsep}{0pt}\def\cbRGB{\colorbox[RGB]}\expandafter\cbRGB\expandafter{\detokenize{239,239,255}}{cohort\strut} \setlength{\fboxsep}{0pt}\def\cbRGB{\colorbox[RGB]}\expandafter\cbRGB\expandafter{\detokenize{252,252,255}}{study\strut} \setlength{\fboxsep}{0pt}\def\cbRGB{\colorbox[RGB]}\expandafter\cbRGB\expandafter{\detokenize{253,253,255}}{including\strut} \setlength{\fboxsep}{0pt}\def\cbRGB{\colorbox[RGB]}\expandafter\cbRGB\expandafter{\detokenize{253,253,255}}{all\strut} \setlength{\fboxsep}{0pt}\def\cbRGB{\colorbox[RGB]}\expandafter\cbRGB\expandafter{\detokenize{254,254,255}}{unk\strut} \setlength{\fboxsep}{0pt}\def\cbRGB{\colorbox[RGB]}\expandafter\cbRGB\expandafter{\detokenize{254,254,255}}{registered\strut} \setlength{\fboxsep}{0pt}\def\cbRGB{\colorbox[RGB]}\expandafter\cbRGB\expandafter{\detokenize{254,254,255}}{,\strut} \setlength{\fboxsep}{0pt}\def\cbRGB{\colorbox[RGB]}\expandafter\cbRGB\expandafter{\detokenize{253,253,255}}{orally\strut} \setlength{\fboxsep}{0pt}\def\cbRGB{\colorbox[RGB]}\expandafter\cbRGB\expandafter{\detokenize{250,250,255}}{treated\strut} \setlength{\fboxsep}{0pt}\def\cbRGB{\colorbox[RGB]}\expandafter\cbRGB\expandafter{\detokenize{251,251,255}}{pregnant\strut} \setlength{\fboxsep}{0pt}\def\cbRGB{\colorbox[RGB]}\expandafter\cbRGB\expandafter{\detokenize{250,250,255}}{diabetic\strut} \setlength{\fboxsep}{0pt}\def\cbRGB{\colorbox[RGB]}\expandafter\cbRGB\expandafter{\detokenize{251,251,255}}{patients\strut} \setlength{\fboxsep}{0pt}\def\cbRGB{\colorbox[RGB]}\expandafter\cbRGB\expandafter{\detokenize{249,249,255}}{set\strut} \setlength{\fboxsep}{0pt}\def\cbRGB{\colorbox[RGB]}\expandafter\cbRGB\expandafter{\detokenize{247,247,255}}{in\strut} \setlength{\fboxsep}{0pt}\def\cbRGB{\colorbox[RGB]}\expandafter\cbRGB\expandafter{\detokenize{246,246,255}}{a\strut} \setlength{\fboxsep}{0pt}\def\cbRGB{\colorbox[RGB]}\expandafter\cbRGB\expandafter{\detokenize{243,243,255}}{diabetic\strut} \setlength{\fboxsep}{0pt}\def\cbRGB{\colorbox[RGB]}\expandafter\cbRGB\expandafter{\detokenize{245,245,255}}{unk\strut} \setlength{\fboxsep}{0pt}\def\cbRGB{\colorbox[RGB]}\expandafter\cbRGB\expandafter{\detokenize{246,246,255}}{service\strut} \setlength{\fboxsep}{0pt}\def\cbRGB{\colorbox[RGB]}\expandafter\cbRGB\expandafter{\detokenize{247,247,255}}{at\strut} \setlength{\fboxsep}{0pt}\def\cbRGB{\colorbox[RGB]}\expandafter\cbRGB\expandafter{\detokenize{239,239,255}}{a\strut} \setlength{\fboxsep}{0pt}\def\cbRGB{\colorbox[RGB]}\expandafter\cbRGB\expandafter{\detokenize{230,230,255}}{university\strut} \setlength{\fboxsep}{0pt}\def\cbRGB{\colorbox[RGB]}\expandafter\cbRGB\expandafter{\detokenize{228,228,255}}{hospital\strut} \setlength{\fboxsep}{0pt}\def\cbRGB{\colorbox[RGB]}\expandafter\cbRGB\expandafter{\detokenize{221,221,255}}{:\strut} \setlength{\fboxsep}{0pt}\def\cbRGB{\colorbox[RGB]}\expandafter\cbRGB\expandafter{\detokenize{229,229,255}}{qqq\strut} \setlength{\fboxsep}{0pt}\def\cbRGB{\colorbox[RGB]}\expandafter\cbRGB\expandafter{\detokenize{233,233,255}}{women\strut} \setlength{\fboxsep}{0pt}\def\cbRGB{\colorbox[RGB]}\expandafter\cbRGB\expandafter{\detokenize{248,248,255}}{treated\strut} \setlength{\fboxsep}{0pt}\def\cbRGB{\colorbox[RGB]}\expandafter\cbRGB\expandafter{\detokenize{251,251,255}}{with\strut} \setlength{\fboxsep}{0pt}\def\cbRGB{\colorbox[RGB]}\expandafter\cbRGB\expandafter{\detokenize{252,252,255}}{metformin\strut} \setlength{\fboxsep}{0pt}\def\cbRGB{\colorbox[RGB]}\expandafter\cbRGB\expandafter{\detokenize{249,249,255}}{,\strut} \setlength{\fboxsep}{0pt}\def\cbRGB{\colorbox[RGB]}\expandafter\cbRGB\expandafter{\detokenize{247,247,255}}{qqq\strut} \setlength{\fboxsep}{0pt}\def\cbRGB{\colorbox[RGB]}\expandafter\cbRGB\expandafter{\detokenize{244,244,255}}{women\strut} \setlength{\fboxsep}{0pt}\def\cbRGB{\colorbox[RGB]}\expandafter\cbRGB\expandafter{\detokenize{245,245,255}}{treated\strut} \setlength{\fboxsep}{0pt}\def\cbRGB{\colorbox[RGB]}\expandafter\cbRGB\expandafter{\detokenize{245,245,255}}{with\strut} \setlength{\fboxsep}{0pt}\def\cbRGB{\colorbox[RGB]}\expandafter\cbRGB\expandafter{\detokenize{242,242,255}}{sulphonylurea\strut} \setlength{\fboxsep}{0pt}\def\cbRGB{\colorbox[RGB]}\expandafter\cbRGB\expandafter{\detokenize{243,243,255}}{during\strut} \setlength{\fboxsep}{0pt}\def\cbRGB{\colorbox[RGB]}\expandafter\cbRGB\expandafter{\detokenize{245,245,255}}{pregnancy\strut} \setlength{\fboxsep}{0pt}\def\cbRGB{\colorbox[RGB]}\expandafter\cbRGB\expandafter{\detokenize{245,245,255}}{and\strut} \setlength{\fboxsep}{0pt}\def\cbRGB{\colorbox[RGB]}\expandafter\cbRGB\expandafter{\detokenize{245,245,255}}{a\strut} \setlength{\fboxsep}{0pt}\def\cbRGB{\colorbox[RGB]}\expandafter\cbRGB\expandafter{\detokenize{244,244,255}}{reference\strut} \setlength{\fboxsep}{0pt}\def\cbRGB{\colorbox[RGB]}\expandafter\cbRGB\expandafter{\detokenize{250,250,255}}{group\strut} \setlength{\fboxsep}{0pt}\def\cbRGB{\colorbox[RGB]}\expandafter\cbRGB\expandafter{\detokenize{246,246,255}}{of\strut} \setlength{\fboxsep}{0pt}\def\cbRGB{\colorbox[RGB]}\expandafter\cbRGB\expandafter{\detokenize{241,241,255}}{qqq\strut} \setlength{\fboxsep}{0pt}\def\cbRGB{\colorbox[RGB]}\expandafter\cbRGB\expandafter{\detokenize{236,236,255}}{diabetic\strut} \setlength{\fboxsep}{0pt}\def\cbRGB{\colorbox[RGB]}\expandafter\cbRGB\expandafter{\detokenize{237,237,255}}{women\strut} \setlength{\fboxsep}{0pt}\def\cbRGB{\colorbox[RGB]}\expandafter\cbRGB\expandafter{\detokenize{238,238,255}}{treated\strut} \setlength{\fboxsep}{0pt}\def\cbRGB{\colorbox[RGB]}\expandafter\cbRGB\expandafter{\detokenize{240,240,255}}{with\strut} \setlength{\fboxsep}{0pt}\def\cbRGB{\colorbox[RGB]}\expandafter\cbRGB\expandafter{\detokenize{235,235,255}}{insulin\strut} \setlength{\fboxsep}{0pt}\def\cbRGB{\colorbox[RGB]}\expandafter\cbRGB\expandafter{\detokenize{236,236,255}}{during\strut} \setlength{\fboxsep}{0pt}\def\cbRGB{\colorbox[RGB]}\expandafter\cbRGB\expandafter{\detokenize{233,233,255}}{pregnancy\strut} \setlength{\fboxsep}{0pt}\def\cbRGB{\colorbox[RGB]}\expandafter\cbRGB\expandafter{\detokenize{224,224,255}}{.\strut} \setlength{\fboxsep}{0pt}\def\cbRGB{\colorbox[RGB]}\expandafter\cbRGB\expandafter{\detokenize{224,224,255}}{results\strut} \setlength{\fboxsep}{0pt}\def\cbRGB{\colorbox[RGB]}\expandafter\cbRGB\expandafter{\detokenize{213,213,255}}{:\strut} \setlength{\fboxsep}{0pt}\def\cbRGB{\colorbox[RGB]}\expandafter\cbRGB\expandafter{\detokenize{222,222,255}}{the\strut} \setlength{\fboxsep}{0pt}\def\cbRGB{\colorbox[RGB]}\expandafter\cbRGB\expandafter{\detokenize{199,199,255}}{prevalence\strut} \setlength{\fboxsep}{0pt}\def\cbRGB{\colorbox[RGB]}\expandafter\cbRGB\expandafter{\detokenize{193,193,255}}{of\strut} \setlength{\fboxsep}{0pt}\def\cbRGB{\colorbox[RGB]}\expandafter\cbRGB\expandafter{\detokenize{185,185,255}}{pre-eclampsia\strut} \setlength{\fboxsep}{0pt}\def\cbRGB{\colorbox[RGB]}\expandafter\cbRGB\expandafter{\detokenize{191,191,255}}{was\strut} \setlength{\fboxsep}{0pt}\def\cbRGB{\colorbox[RGB]}\expandafter\cbRGB\expandafter{\detokenize{205,205,255}}{significantly\strut} \setlength{\fboxsep}{0pt}\def\cbRGB{\colorbox[RGB]}\expandafter\cbRGB\expandafter{\detokenize{219,219,255}}{increased\strut} \setlength{\fboxsep}{0pt}\def\cbRGB{\colorbox[RGB]}\expandafter\cbRGB\expandafter{\detokenize{238,238,255}}{in\strut} \setlength{\fboxsep}{0pt}\def\cbRGB{\colorbox[RGB]}\expandafter\cbRGB\expandafter{\detokenize{242,242,255}}{the\strut} \setlength{\fboxsep}{0pt}\def\cbRGB{\colorbox[RGB]}\expandafter\cbRGB\expandafter{\detokenize{239,239,255}}{group\strut} \setlength{\fboxsep}{0pt}\def\cbRGB{\colorbox[RGB]}\expandafter\cbRGB\expandafter{\detokenize{238,238,255}}{of\strut} \setlength{\fboxsep}{0pt}\def\cbRGB{\colorbox[RGB]}\expandafter\cbRGB\expandafter{\detokenize{241,241,255}}{women\strut} \setlength{\fboxsep}{0pt}\def\cbRGB{\colorbox[RGB]}\expandafter\cbRGB\expandafter{\detokenize{245,245,255}}{treated\strut} \setlength{\fboxsep}{0pt}\def\cbRGB{\colorbox[RGB]}\expandafter\cbRGB\expandafter{\detokenize{244,244,255}}{with\strut} \setlength{\fboxsep}{0pt}\def\cbRGB{\colorbox[RGB]}\expandafter\cbRGB\expandafter{\detokenize{244,244,255}}{metformin\strut} \setlength{\fboxsep}{0pt}\def\cbRGB{\colorbox[RGB]}\expandafter\cbRGB\expandafter{\detokenize{239,239,255}}{compared\strut} \setlength{\fboxsep}{0pt}\def\cbRGB{\colorbox[RGB]}\expandafter\cbRGB\expandafter{\detokenize{238,238,255}}{to\strut} \setlength{\fboxsep}{0pt}\def\cbRGB{\colorbox[RGB]}\expandafter\cbRGB\expandafter{\detokenize{237,237,255}}{women\strut} \setlength{\fboxsep}{0pt}\def\cbRGB{\colorbox[RGB]}\expandafter\cbRGB\expandafter{\detokenize{242,242,255}}{treated\strut} \setlength{\fboxsep}{0pt}\def\cbRGB{\colorbox[RGB]}\expandafter\cbRGB\expandafter{\detokenize{248,248,255}}{with\strut} \setlength{\fboxsep}{0pt}\def\cbRGB{\colorbox[RGB]}\expandafter\cbRGB\expandafter{\detokenize{249,249,255}}{sulphonylurea\strut} \setlength{\fboxsep}{0pt}\def\cbRGB{\colorbox[RGB]}\expandafter\cbRGB\expandafter{\detokenize{253,253,255}}{or\strut} \setlength{\fboxsep}{0pt}\def\cbRGB{\colorbox[RGB]}\expandafter\cbRGB\expandafter{\detokenize{252,252,255}}{insulin\strut} \setlength{\fboxsep}{0pt}\def\cbRGB{\colorbox[RGB]}\expandafter\cbRGB\expandafter{\detokenize{255,255,255}}{(\strut} \setlength{\fboxsep}{0pt}\def\cbRGB{\colorbox[RGB]}\expandafter\cbRGB\expandafter{\detokenize{253,253,255}}{qqq\strut} \setlength{\fboxsep}{0pt}\def\cbRGB{\colorbox[RGB]}\expandafter\cbRGB\expandafter{\detokenize{252,252,255}}{vs.\strut} \setlength{\fboxsep}{0pt}\def\cbRGB{\colorbox[RGB]}\expandafter\cbRGB\expandafter{\detokenize{252,252,255}}{qqq\strut} \setlength{\fboxsep}{0pt}\def\cbRGB{\colorbox[RGB]}\expandafter\cbRGB\expandafter{\detokenize{251,251,255}}{vs.\strut} \setlength{\fboxsep}{0pt}\def\cbRGB{\colorbox[RGB]}\expandafter\cbRGB\expandafter{\detokenize{248,248,255}}{qqq\strut} \setlength{\fboxsep}{0pt}\def\cbRGB{\colorbox[RGB]}\expandafter\cbRGB\expandafter{\detokenize{245,245,255}}{\%\strut} \setlength{\fboxsep}{0pt}\def\cbRGB{\colorbox[RGB]}\expandafter\cbRGB\expandafter{\detokenize{245,245,255}}{,\strut} \setlength{\fboxsep}{0pt}\def\cbRGB{\colorbox[RGB]}\expandafter\cbRGB\expandafter{\detokenize{247,247,255}}{p\strut} \setlength{\fboxsep}{0pt}\def\cbRGB{\colorbox[RGB]}\expandafter\cbRGB\expandafter{\detokenize{248,248,255}}{<\strut} \setlength{\fboxsep}{0pt}\def\cbRGB{\colorbox[RGB]}\expandafter\cbRGB\expandafter{\detokenize{249,249,255}}{qqq\strut} \setlength{\fboxsep}{0pt}\def\cbRGB{\colorbox[RGB]}\expandafter\cbRGB\expandafter{\detokenize{249,249,255}}{)\strut} \setlength{\fboxsep}{0pt}\def\cbRGB{\colorbox[RGB]}\expandafter\cbRGB\expandafter{\detokenize{244,244,255}}{.\strut} \setlength{\fboxsep}{0pt}\def\cbRGB{\colorbox[RGB]}\expandafter\cbRGB\expandafter{\detokenize{233,233,255}}{no\strut} \setlength{\fboxsep}{0pt}\def\cbRGB{\colorbox[RGB]}\expandafter\cbRGB\expandafter{\detokenize{196,196,255}}{difference\strut} \setlength{\fboxsep}{0pt}\def\cbRGB{\colorbox[RGB]}\expandafter\cbRGB\expandafter{\detokenize{171,171,255}}{in\strut} \setlength{\fboxsep}{0pt}\def\cbRGB{\colorbox[RGB]}\expandafter\cbRGB\expandafter{\detokenize{164,164,255}}{neonatal\strut} \setlength{\fboxsep}{0pt}\def\cbRGB{\colorbox[RGB]}\expandafter\cbRGB\expandafter{\detokenize{166,166,255}}{morbidity\strut} \setlength{\fboxsep}{0pt}\def\cbRGB{\colorbox[RGB]}\expandafter\cbRGB\expandafter{\detokenize{189,189,255}}{was\strut} \setlength{\fboxsep}{0pt}\def\cbRGB{\colorbox[RGB]}\expandafter\cbRGB\expandafter{\detokenize{206,206,255}}{observed\strut} \setlength{\fboxsep}{0pt}\def\cbRGB{\colorbox[RGB]}\expandafter\cbRGB\expandafter{\detokenize{241,241,255}}{between\strut} \setlength{\fboxsep}{0pt}\def\cbRGB{\colorbox[RGB]}\expandafter\cbRGB\expandafter{\detokenize{246,246,255}}{the\strut} \setlength{\fboxsep}{0pt}\def\cbRGB{\colorbox[RGB]}\expandafter\cbRGB\expandafter{\detokenize{246,246,255}}{orally\strut} \setlength{\fboxsep}{0pt}\def\cbRGB{\colorbox[RGB]}\expandafter\cbRGB\expandafter{\detokenize{246,246,255}}{treated\strut} \setlength{\fboxsep}{0pt}\def\cbRGB{\colorbox[RGB]}\expandafter\cbRGB\expandafter{\detokenize{248,248,255}}{and\strut} \setlength{\fboxsep}{0pt}\def\cbRGB{\colorbox[RGB]}\expandafter\cbRGB\expandafter{\detokenize{249,249,255}}{unk\strut} \setlength{\fboxsep}{0pt}\def\cbRGB{\colorbox[RGB]}\expandafter\cbRGB\expandafter{\detokenize{250,250,255}}{group\strut} \setlength{\fboxsep}{0pt}\def\cbRGB{\colorbox[RGB]}\expandafter\cbRGB\expandafter{\detokenize{250,250,255}}{;\strut} \setlength{\fboxsep}{0pt}\def\cbRGB{\colorbox[RGB]}\expandafter\cbRGB\expandafter{\detokenize{249,249,255}}{no\strut} \setlength{\fboxsep}{0pt}\def\cbRGB{\colorbox[RGB]}\expandafter\cbRGB\expandafter{\detokenize{242,242,255}}{cases\strut} \setlength{\fboxsep}{0pt}\def\cbRGB{\colorbox[RGB]}\expandafter\cbRGB\expandafter{\detokenize{225,225,255}}{of\strut} \setlength{\fboxsep}{0pt}\def\cbRGB{\colorbox[RGB]}\expandafter\cbRGB\expandafter{\detokenize{224,224,255}}{severe\strut} \setlength{\fboxsep}{0pt}\def\cbRGB{\colorbox[RGB]}\expandafter\cbRGB\expandafter{\detokenize{215,215,255}}{hypoglycaemia\strut} \setlength{\fboxsep}{0pt}\def\cbRGB{\colorbox[RGB]}\expandafter\cbRGB\expandafter{\detokenize{230,230,255}}{or\strut} \setlength{\fboxsep}{0pt}\def\cbRGB{\colorbox[RGB]}\expandafter\cbRGB\expandafter{\detokenize{230,230,255}}{jaundice\strut} \setlength{\fboxsep}{0pt}\def\cbRGB{\colorbox[RGB]}\expandafter\cbRGB\expandafter{\detokenize{244,244,255}}{were\strut} \setlength{\fboxsep}{0pt}\def\cbRGB{\colorbox[RGB]}\expandafter\cbRGB\expandafter{\detokenize{245,245,255}}{seen\strut} \setlength{\fboxsep}{0pt}\def\cbRGB{\colorbox[RGB]}\expandafter\cbRGB\expandafter{\detokenize{245,245,255}}{in\strut} \setlength{\fboxsep}{0pt}\def\cbRGB{\colorbox[RGB]}\expandafter\cbRGB\expandafter{\detokenize{245,245,255}}{the\strut} \setlength{\fboxsep}{0pt}\def\cbRGB{\colorbox[RGB]}\expandafter\cbRGB\expandafter{\detokenize{245,245,255}}{orally\strut} \setlength{\fboxsep}{0pt}\def\cbRGB{\colorbox[RGB]}\expandafter\cbRGB\expandafter{\detokenize{245,245,255}}{treated\strut} \setlength{\fboxsep}{0pt}\def\cbRGB{\colorbox[RGB]}\expandafter\cbRGB\expandafter{\detokenize{246,246,255}}{groups\strut} \setlength{\fboxsep}{0pt}\def\cbRGB{\colorbox[RGB]}\expandafter\cbRGB\expandafter{\detokenize{244,244,255}}{.\strut} \setlength{\fboxsep}{0pt}\def\cbRGB{\colorbox[RGB]}\expandafter\cbRGB\expandafter{\detokenize{243,243,255}}{however\strut} \setlength{\fboxsep}{0pt}\def\cbRGB{\colorbox[RGB]}\expandafter\cbRGB\expandafter{\detokenize{243,243,255}}{,\strut} \setlength{\fboxsep}{0pt}\def\cbRGB{\colorbox[RGB]}\expandafter\cbRGB\expandafter{\detokenize{244,244,255}}{in\strut} \setlength{\fboxsep}{0pt}\def\cbRGB{\colorbox[RGB]}\expandafter\cbRGB\expandafter{\detokenize{243,243,255}}{the\strut} \setlength{\fboxsep}{0pt}\def\cbRGB{\colorbox[RGB]}\expandafter\cbRGB\expandafter{\detokenize{239,239,255}}{group\strut} \setlength{\fboxsep}{0pt}\def\cbRGB{\colorbox[RGB]}\expandafter\cbRGB\expandafter{\detokenize{238,238,255}}{of\strut} \setlength{\fboxsep}{0pt}\def\cbRGB{\colorbox[RGB]}\expandafter\cbRGB\expandafter{\detokenize{241,241,255}}{women\strut} \setlength{\fboxsep}{0pt}\def\cbRGB{\colorbox[RGB]}\expandafter\cbRGB\expandafter{\detokenize{245,245,255}}{treated\strut} \setlength{\fboxsep}{0pt}\def\cbRGB{\colorbox[RGB]}\expandafter\cbRGB\expandafter{\detokenize{246,246,255}}{with\strut} \setlength{\fboxsep}{0pt}\def\cbRGB{\colorbox[RGB]}\expandafter\cbRGB\expandafter{\detokenize{246,246,255}}{metformin\strut} \setlength{\fboxsep}{0pt}\def\cbRGB{\colorbox[RGB]}\expandafter\cbRGB\expandafter{\detokenize{246,246,255}}{in\strut} \setlength{\fboxsep}{0pt}\def\cbRGB{\colorbox[RGB]}\expandafter\cbRGB\expandafter{\detokenize{246,246,255}}{the\strut} \setlength{\fboxsep}{0pt}\def\cbRGB{\colorbox[RGB]}\expandafter\cbRGB\expandafter{\detokenize{244,244,255}}{third\strut} \setlength{\fboxsep}{0pt}\def\cbRGB{\colorbox[RGB]}\expandafter\cbRGB\expandafter{\detokenize{230,230,255}}{trimester\strut} \setlength{\fboxsep}{0pt}\def\cbRGB{\colorbox[RGB]}\expandafter\cbRGB\expandafter{\detokenize{143,143,255}}{,\strut} \setlength{\fboxsep}{0pt}\def\cbRGB{\colorbox[RGB]}\expandafter\cbRGB\expandafter{\detokenize{66,66,255}}{the\strut} \setlength{\fboxsep}{0pt}\def\cbRGB{\colorbox[RGB]}\expandafter\cbRGB\expandafter{\detokenize{46,46,255}}{perinatal\strut} \setlength{\fboxsep}{0pt}\def\cbRGB{\colorbox[RGB]}\expandafter\cbRGB\expandafter{\detokenize{0,0,255}}{mortality\strut} \setlength{\fboxsep}{0pt}\def\cbRGB{\colorbox[RGB]}\expandafter\cbRGB\expandafter{\detokenize{57,57,255}}{was\strut} \setlength{\fboxsep}{0pt}\def\cbRGB{\colorbox[RGB]}\expandafter\cbRGB\expandafter{\detokenize{86,86,255}}{significantly\strut} \setlength{\fboxsep}{0pt}\def\cbRGB{\colorbox[RGB]}\expandafter\cbRGB\expandafter{\detokenize{217,217,255}}{increased\strut} \setlength{\fboxsep}{0pt}\def\cbRGB{\colorbox[RGB]}\expandafter\cbRGB\expandafter{\detokenize{234,234,255}}{compared\strut} \setlength{\fboxsep}{0pt}\def\cbRGB{\colorbox[RGB]}\expandafter\cbRGB\expandafter{\detokenize{236,236,255}}{to\strut} \setlength{\fboxsep}{0pt}\def\cbRGB{\colorbox[RGB]}\expandafter\cbRGB\expandafter{\detokenize{237,237,255}}{women\strut} \setlength{\fboxsep}{0pt}\def\cbRGB{\colorbox[RGB]}\expandafter\cbRGB\expandafter{\detokenize{243,243,255}}{not\strut} \setlength{\fboxsep}{0pt}\def\cbRGB{\colorbox[RGB]}\expandafter\cbRGB\expandafter{\detokenize{247,247,255}}{treated\strut} \setlength{\fboxsep}{0pt}\def\cbRGB{\colorbox[RGB]}\expandafter\cbRGB\expandafter{\detokenize{251,251,255}}{with\strut} \setlength{\fboxsep}{0pt}\def\cbRGB{\colorbox[RGB]}\expandafter\cbRGB\expandafter{\detokenize{252,252,255}}{metformin\strut} \setlength{\fboxsep}{0pt}\def\cbRGB{\colorbox[RGB]}\expandafter\cbRGB\expandafter{\detokenize{253,253,255}}{(\strut} \setlength{\fboxsep}{0pt}\def\cbRGB{\colorbox[RGB]}\expandafter\cbRGB\expandafter{\detokenize{254,254,255}}{qqq\strut} \setlength{\fboxsep}{0pt}\def\cbRGB{\colorbox[RGB]}\expandafter\cbRGB\expandafter{\detokenize{252,252,255}}{vs.\strut} \setlength{\fboxsep}{0pt}\def\cbRGB{\colorbox[RGB]}\expandafter\cbRGB\expandafter{\detokenize{248,248,255}}{qqq\strut} \setlength{\fboxsep}{0pt}\def\cbRGB{\colorbox[RGB]}\expandafter\cbRGB\expandafter{\detokenize{245,245,255}}{\%\strut} \setlength{\fboxsep}{0pt}\def\cbRGB{\colorbox[RGB]}\expandafter\cbRGB\expandafter{\detokenize{245,245,255}}{,\strut} \setlength{\fboxsep}{0pt}\def\cbRGB{\colorbox[RGB]}\expandafter\cbRGB\expandafter{\detokenize{247,247,255}}{p\strut} \setlength{\fboxsep}{0pt}\def\cbRGB{\colorbox[RGB]}\expandafter\cbRGB\expandafter{\detokenize{248,248,255}}{<\strut} \setlength{\fboxsep}{0pt}\def\cbRGB{\colorbox[RGB]}\expandafter\cbRGB\expandafter{\detokenize{246,246,255}}{qqq\strut} \setlength{\fboxsep}{0pt}\def\cbRGB{\colorbox[RGB]}\expandafter\cbRGB\expandafter{\detokenize{242,242,255}}{)\strut} \setlength{\fboxsep}{0pt}\def\cbRGB{\colorbox[RGB]}\expandafter\cbRGB\expandafter{\detokenize{235,235,255}}{.\strut} \setlength{\fboxsep}{0pt}\def\cbRGB{\colorbox[RGB]}\expandafter\cbRGB\expandafter{\detokenize{235,235,255}}{conclusion\strut} \setlength{\fboxsep}{0pt}\def\cbRGB{\colorbox[RGB]}\expandafter\cbRGB\expandafter{\detokenize{238,238,255}}{:\strut} \setlength{\fboxsep}{0pt}\def\cbRGB{\colorbox[RGB]}\expandafter\cbRGB\expandafter{\detokenize{246,246,255}}{treatment\strut} \setlength{\fboxsep}{0pt}\def\cbRGB{\colorbox[RGB]}\expandafter\cbRGB\expandafter{\detokenize{247,247,255}}{with\strut} \setlength{\fboxsep}{0pt}\def\cbRGB{\colorbox[RGB]}\expandafter\cbRGB\expandafter{\detokenize{246,246,255}}{metformin\strut} \setlength{\fboxsep}{0pt}\def\cbRGB{\colorbox[RGB]}\expandafter\cbRGB\expandafter{\detokenize{243,243,255}}{during\strut} \setlength{\fboxsep}{0pt}\def\cbRGB{\colorbox[RGB]}\expandafter\cbRGB\expandafter{\detokenize{245,245,255}}{pregnancy\strut} \setlength{\fboxsep}{0pt}\def\cbRGB{\colorbox[RGB]}\expandafter\cbRGB\expandafter{\detokenize{242,242,255}}{was\strut} \setlength{\fboxsep}{0pt}\def\cbRGB{\colorbox[RGB]}\expandafter\cbRGB\expandafter{\detokenize{238,238,255}}{associated\strut} \setlength{\fboxsep}{0pt}\def\cbRGB{\colorbox[RGB]}\expandafter\cbRGB\expandafter{\detokenize{230,230,255}}{with\strut} \setlength{\fboxsep}{0pt}\def\cbRGB{\colorbox[RGB]}\expandafter\cbRGB\expandafter{\detokenize{228,228,255}}{increased\strut} \setlength{\fboxsep}{0pt}\def\cbRGB{\colorbox[RGB]}\expandafter\cbRGB\expandafter{\detokenize{209,209,255}}{prevalence\strut} \setlength{\fboxsep}{0pt}\def\cbRGB{\colorbox[RGB]}\expandafter\cbRGB\expandafter{\detokenize{207,207,255}}{of\strut} \setlength{\fboxsep}{0pt}\def\cbRGB{\colorbox[RGB]}\expandafter\cbRGB\expandafter{\detokenize{208,208,255}}{pre-eclampsia\strut} \setlength{\fboxsep}{0pt}\def\cbRGB{\colorbox[RGB]}\expandafter\cbRGB\expandafter{\detokenize{200,200,255}}{and\strut} \setlength{\fboxsep}{0pt}\def\cbRGB{\colorbox[RGB]}\expandafter\cbRGB\expandafter{\detokenize{128,128,255}}{a\strut} \setlength{\fboxsep}{0pt}\def\cbRGB{\colorbox[RGB]}\expandafter\cbRGB\expandafter{\detokenize{79,79,255}}{high\strut} \setlength{\fboxsep}{0pt}\def\cbRGB{\colorbox[RGB]}\expandafter\cbRGB\expandafter{\detokenize{91,91,255}}{perinatal\strut} \setlength{\fboxsep}{0pt}\def\cbRGB{\colorbox[RGB]}\expandafter\cbRGB\expandafter{\detokenize{83,83,255}}{mortality\strut} \setlength{\fboxsep}{0pt}\def\cbRGB{\colorbox[RGB]}\expandafter\cbRGB\expandafter{\detokenize{147,147,255}}{.\strut} 

\par
\textbf{Example 3}

\setlength{\fboxsep}{0pt}\def\cbRGB{\colorbox[RGB]}\expandafter\cbRGB\expandafter{\detokenize{255,248,248}}{purpose\strut} \setlength{\fboxsep}{0pt}\def\cbRGB{\colorbox[RGB]}\expandafter\cbRGB\expandafter{\detokenize{255,237,237}}{:\strut} \setlength{\fboxsep}{0pt}\def\cbRGB{\colorbox[RGB]}\expandafter\cbRGB\expandafter{\detokenize{255,233,233}}{congestive\strut} \setlength{\fboxsep}{0pt}\def\cbRGB{\colorbox[RGB]}\expandafter\cbRGB\expandafter{\detokenize{255,229,229}}{heart\strut} \setlength{\fboxsep}{0pt}\def\cbRGB{\colorbox[RGB]}\expandafter\cbRGB\expandafter{\detokenize{255,228,228}}{failure\strut} \setlength{\fboxsep}{0pt}\def\cbRGB{\colorbox[RGB]}\expandafter\cbRGB\expandafter{\detokenize{255,227,227}}{is\strut} \setlength{\fboxsep}{0pt}\def\cbRGB{\colorbox[RGB]}\expandafter\cbRGB\expandafter{\detokenize{255,228,228}}{an\strut} \setlength{\fboxsep}{0pt}\def\cbRGB{\colorbox[RGB]}\expandafter\cbRGB\expandafter{\detokenize{255,229,229}}{important\strut} \setlength{\fboxsep}{0pt}\def\cbRGB{\colorbox[RGB]}\expandafter\cbRGB\expandafter{\detokenize{255,232,232}}{cause\strut} \setlength{\fboxsep}{0pt}\def\cbRGB{\colorbox[RGB]}\expandafter\cbRGB\expandafter{\detokenize{255,236,236}}{of\strut} \setlength{\fboxsep}{0pt}\def\cbRGB{\colorbox[RGB]}\expandafter\cbRGB\expandafter{\detokenize{255,238,238}}{patient\strut} \setlength{\fboxsep}{0pt}\def\cbRGB{\colorbox[RGB]}\expandafter\cbRGB\expandafter{\detokenize{255,241,241}}{morbidity\strut} \setlength{\fboxsep}{0pt}\def\cbRGB{\colorbox[RGB]}\expandafter\cbRGB\expandafter{\detokenize{255,245,245}}{and\strut} \setlength{\fboxsep}{0pt}\def\cbRGB{\colorbox[RGB]}\expandafter\cbRGB\expandafter{\detokenize{255,242,242}}{mortality\strut} \setlength{\fboxsep}{0pt}\def\cbRGB{\colorbox[RGB]}\expandafter\cbRGB\expandafter{\detokenize{255,244,244}}{.\strut} \setlength{\fboxsep}{0pt}\def\cbRGB{\colorbox[RGB]}\expandafter\cbRGB\expandafter{\detokenize{255,237,237}}{although\strut} \setlength{\fboxsep}{0pt}\def\cbRGB{\colorbox[RGB]}\expandafter\cbRGB\expandafter{\detokenize{255,237,237}}{several\strut} \setlength{\fboxsep}{0pt}\def\cbRGB{\colorbox[RGB]}\expandafter\cbRGB\expandafter{\detokenize{255,231,231}}{randomized\strut} \setlength{\fboxsep}{0pt}\def\cbRGB{\colorbox[RGB]}\expandafter\cbRGB\expandafter{\detokenize{255,229,229}}{clinical\strut} \setlength{\fboxsep}{0pt}\def\cbRGB{\colorbox[RGB]}\expandafter\cbRGB\expandafter{\detokenize{255,226,226}}{trials\strut} \setlength{\fboxsep}{0pt}\def\cbRGB{\colorbox[RGB]}\expandafter\cbRGB\expandafter{\detokenize{255,218,218}}{have\strut} \setlength{\fboxsep}{0pt}\def\cbRGB{\colorbox[RGB]}\expandafter\cbRGB\expandafter{\detokenize{255,214,214}}{compared\strut} \setlength{\fboxsep}{0pt}\def\cbRGB{\colorbox[RGB]}\expandafter\cbRGB\expandafter{\detokenize{255,211,211}}{beta-blockers\strut} \setlength{\fboxsep}{0pt}\def\cbRGB{\colorbox[RGB]}\expandafter\cbRGB\expandafter{\detokenize{255,214,214}}{with\strut} \setlength{\fboxsep}{0pt}\def\cbRGB{\colorbox[RGB]}\expandafter\cbRGB\expandafter{\detokenize{255,217,217}}{placebo\strut} \setlength{\fboxsep}{0pt}\def\cbRGB{\colorbox[RGB]}\expandafter\cbRGB\expandafter{\detokenize{255,222,222}}{for\strut} \setlength{\fboxsep}{0pt}\def\cbRGB{\colorbox[RGB]}\expandafter\cbRGB\expandafter{\detokenize{255,229,229}}{treatment\strut} \setlength{\fboxsep}{0pt}\def\cbRGB{\colorbox[RGB]}\expandafter\cbRGB\expandafter{\detokenize{255,232,232}}{of\strut} \setlength{\fboxsep}{0pt}\def\cbRGB{\colorbox[RGB]}\expandafter\cbRGB\expandafter{\detokenize{255,232,232}}{congestive\strut} \setlength{\fboxsep}{0pt}\def\cbRGB{\colorbox[RGB]}\expandafter\cbRGB\expandafter{\detokenize{255,234,234}}{heart\strut} \setlength{\fboxsep}{0pt}\def\cbRGB{\colorbox[RGB]}\expandafter\cbRGB\expandafter{\detokenize{255,234,234}}{failure\strut} \setlength{\fboxsep}{0pt}\def\cbRGB{\colorbox[RGB]}\expandafter\cbRGB\expandafter{\detokenize{255,237,237}}{,\strut} \setlength{\fboxsep}{0pt}\def\cbRGB{\colorbox[RGB]}\expandafter\cbRGB\expandafter{\detokenize{255,244,244}}{a\strut} \setlength{\fboxsep}{0pt}\def\cbRGB{\colorbox[RGB]}\expandafter\cbRGB\expandafter{\detokenize{255,246,246}}{meta-analysis\strut} \setlength{\fboxsep}{0pt}\def\cbRGB{\colorbox[RGB]}\expandafter\cbRGB\expandafter{\detokenize{255,248,248}}{unk\strut} \setlength{\fboxsep}{0pt}\def\cbRGB{\colorbox[RGB]}\expandafter\cbRGB\expandafter{\detokenize{255,242,242}}{the\strut} \setlength{\fboxsep}{0pt}\def\cbRGB{\colorbox[RGB]}\expandafter\cbRGB\expandafter{\detokenize{255,243,243}}{effect\strut} \setlength{\fboxsep}{0pt}\def\cbRGB{\colorbox[RGB]}\expandafter\cbRGB\expandafter{\detokenize{255,245,245}}{on\strut} \setlength{\fboxsep}{0pt}\def\cbRGB{\colorbox[RGB]}\expandafter\cbRGB\expandafter{\detokenize{255,242,242}}{mortality\strut} \setlength{\fboxsep}{0pt}\def\cbRGB{\colorbox[RGB]}\expandafter\cbRGB\expandafter{\detokenize{255,246,246}}{and\strut} \setlength{\fboxsep}{0pt}\def\cbRGB{\colorbox[RGB]}\expandafter\cbRGB\expandafter{\detokenize{255,244,244}}{morbidity\strut} \setlength{\fboxsep}{0pt}\def\cbRGB{\colorbox[RGB]}\expandafter\cbRGB\expandafter{\detokenize{255,246,246}}{has\strut} \setlength{\fboxsep}{0pt}\def\cbRGB{\colorbox[RGB]}\expandafter\cbRGB\expandafter{\detokenize{255,241,241}}{not\strut} \setlength{\fboxsep}{0pt}\def\cbRGB{\colorbox[RGB]}\expandafter\cbRGB\expandafter{\detokenize{255,238,238}}{been\strut} \setlength{\fboxsep}{0pt}\def\cbRGB{\colorbox[RGB]}\expandafter\cbRGB\expandafter{\detokenize{255,239,239}}{performed\strut} \setlength{\fboxsep}{0pt}\def\cbRGB{\colorbox[RGB]}\expandafter\cbRGB\expandafter{\detokenize{255,242,242}}{recently\strut} \setlength{\fboxsep}{0pt}\def\cbRGB{\colorbox[RGB]}\expandafter\cbRGB\expandafter{\detokenize{255,244,244}}{.\strut} \setlength{\fboxsep}{0pt}\def\cbRGB{\colorbox[RGB]}\expandafter\cbRGB\expandafter{\detokenize{255,244,244}}{data\strut} \setlength{\fboxsep}{0pt}\def\cbRGB{\colorbox[RGB]}\expandafter\cbRGB\expandafter{\detokenize{255,244,244}}{unk\strut} \setlength{\fboxsep}{0pt}\def\cbRGB{\colorbox[RGB]}\expandafter\cbRGB\expandafter{\detokenize{255,244,244}}{:\strut} \setlength{\fboxsep}{0pt}\def\cbRGB{\colorbox[RGB]}\expandafter\cbRGB\expandafter{\detokenize{255,243,243}}{the\strut} \setlength{\fboxsep}{0pt}\def\cbRGB{\colorbox[RGB]}\expandafter\cbRGB\expandafter{\detokenize{255,242,242}}{unk\strut} \setlength{\fboxsep}{0pt}\def\cbRGB{\colorbox[RGB]}\expandafter\cbRGB\expandafter{\detokenize{255,240,240}}{,\strut} \setlength{\fboxsep}{0pt}\def\cbRGB{\colorbox[RGB]}\expandafter\cbRGB\expandafter{\detokenize{255,240,240}}{unk\strut} \setlength{\fboxsep}{0pt}\def\cbRGB{\colorbox[RGB]}\expandafter\cbRGB\expandafter{\detokenize{255,240,240}}{,\strut} \setlength{\fboxsep}{0pt}\def\cbRGB{\colorbox[RGB]}\expandafter\cbRGB\expandafter{\detokenize{255,240,240}}{and\strut} \setlength{\fboxsep}{0pt}\def\cbRGB{\colorbox[RGB]}\expandafter\cbRGB\expandafter{\detokenize{255,239,239}}{unk\strut} \setlength{\fboxsep}{0pt}\def\cbRGB{\colorbox[RGB]}\expandafter\cbRGB\expandafter{\detokenize{255,239,239}}{of\strut} \setlength{\fboxsep}{0pt}\def\cbRGB{\colorbox[RGB]}\expandafter\cbRGB\expandafter{\detokenize{255,239,239}}{unk\strut} \setlength{\fboxsep}{0pt}\def\cbRGB{\colorbox[RGB]}\expandafter\cbRGB\expandafter{\detokenize{255,238,238}}{electronic\strut} \setlength{\fboxsep}{0pt}\def\cbRGB{\colorbox[RGB]}\expandafter\cbRGB\expandafter{\detokenize{255,238,238}}{unk\strut} \setlength{\fboxsep}{0pt}\def\cbRGB{\colorbox[RGB]}\expandafter\cbRGB\expandafter{\detokenize{255,238,238}}{were\strut} \setlength{\fboxsep}{0pt}\def\cbRGB{\colorbox[RGB]}\expandafter\cbRGB\expandafter{\detokenize{255,238,238}}{unk\strut} \setlength{\fboxsep}{0pt}\def\cbRGB{\colorbox[RGB]}\expandafter\cbRGB\expandafter{\detokenize{255,237,237}}{from\strut} \setlength{\fboxsep}{0pt}\def\cbRGB{\colorbox[RGB]}\expandafter\cbRGB\expandafter{\detokenize{255,237,237}}{qqq\strut} \setlength{\fboxsep}{0pt}\def\cbRGB{\colorbox[RGB]}\expandafter\cbRGB\expandafter{\detokenize{255,240,240}}{to\strut} \setlength{\fboxsep}{0pt}\def\cbRGB{\colorbox[RGB]}\expandafter\cbRGB\expandafter{\detokenize{255,240,240}}{july\strut} \setlength{\fboxsep}{0pt}\def\cbRGB{\colorbox[RGB]}\expandafter\cbRGB\expandafter{\detokenize{255,241,241}}{qqq\strut} \setlength{\fboxsep}{0pt}\def\cbRGB{\colorbox[RGB]}\expandafter\cbRGB\expandafter{\detokenize{255,237,237}}{unk\strut} \setlength{\fboxsep}{0pt}\def\cbRGB{\colorbox[RGB]}\expandafter\cbRGB\expandafter{\detokenize{255,236,236}}{were\strut} \setlength{\fboxsep}{0pt}\def\cbRGB{\colorbox[RGB]}\expandafter\cbRGB\expandafter{\detokenize{255,237,237}}{also\strut} \setlength{\fboxsep}{0pt}\def\cbRGB{\colorbox[RGB]}\expandafter\cbRGB\expandafter{\detokenize{255,236,236}}{identified\strut} \setlength{\fboxsep}{0pt}\def\cbRGB{\colorbox[RGB]}\expandafter\cbRGB\expandafter{\detokenize{255,235,235}}{from\strut} \setlength{\fboxsep}{0pt}\def\cbRGB{\colorbox[RGB]}\expandafter\cbRGB\expandafter{\detokenize{255,235,235}}{unk\strut} \setlength{\fboxsep}{0pt}\def\cbRGB{\colorbox[RGB]}\expandafter\cbRGB\expandafter{\detokenize{255,236,236}}{of\strut} \setlength{\fboxsep}{0pt}\def\cbRGB{\colorbox[RGB]}\expandafter\cbRGB\expandafter{\detokenize{255,238,238}}{unk\strut} \setlength{\fboxsep}{0pt}\def\cbRGB{\colorbox[RGB]}\expandafter\cbRGB\expandafter{\detokenize{255,239,239}}{unk\strut} \setlength{\fboxsep}{0pt}\def\cbRGB{\colorbox[RGB]}\expandafter\cbRGB\expandafter{\detokenize{255,239,239}}{.\strut} \setlength{\fboxsep}{0pt}\def\cbRGB{\colorbox[RGB]}\expandafter\cbRGB\expandafter{\detokenize{255,241,241}}{study\strut} \setlength{\fboxsep}{0pt}\def\cbRGB{\colorbox[RGB]}\expandafter\cbRGB\expandafter{\detokenize{255,240,240}}{selection\strut} \setlength{\fboxsep}{0pt}\def\cbRGB{\colorbox[RGB]}\expandafter\cbRGB\expandafter{\detokenize{255,237,237}}{:\strut} \setlength{\fboxsep}{0pt}\def\cbRGB{\colorbox[RGB]}\expandafter\cbRGB\expandafter{\detokenize{255,237,237}}{all\strut} \setlength{\fboxsep}{0pt}\def\cbRGB{\colorbox[RGB]}\expandafter\cbRGB\expandafter{\detokenize{255,236,236}}{randomized\strut} \setlength{\fboxsep}{0pt}\def\cbRGB{\colorbox[RGB]}\expandafter\cbRGB\expandafter{\detokenize{255,236,236}}{clinical\strut} \setlength{\fboxsep}{0pt}\def\cbRGB{\colorbox[RGB]}\expandafter\cbRGB\expandafter{\detokenize{255,237,237}}{trials\strut} \setlength{\fboxsep}{0pt}\def\cbRGB{\colorbox[RGB]}\expandafter\cbRGB\expandafter{\detokenize{255,245,245}}{of\strut} \setlength{\fboxsep}{0pt}\def\cbRGB{\colorbox[RGB]}\expandafter\cbRGB\expandafter{\detokenize{255,253,253}}{beta-blockers\strut} \setlength{\fboxsep}{0pt}\def\cbRGB{\colorbox[RGB]}\expandafter\cbRGB\expandafter{\detokenize{255,255,255}}{versus\strut} \setlength{\fboxsep}{0pt}\def\cbRGB{\colorbox[RGB]}\expandafter\cbRGB\expandafter{\detokenize{255,241,241}}{placebo\strut} \setlength{\fboxsep}{0pt}\def\cbRGB{\colorbox[RGB]}\expandafter\cbRGB\expandafter{\detokenize{255,194,194}}{in\strut} \setlength{\fboxsep}{0pt}\def\cbRGB{\colorbox[RGB]}\expandafter\cbRGB\expandafter{\detokenize{255,112,112}}{chronic\strut} \setlength{\fboxsep}{0pt}\def\cbRGB{\colorbox[RGB]}\expandafter\cbRGB\expandafter{\detokenize{255,0,0}}{stable\strut} \setlength{\fboxsep}{0pt}\def\cbRGB{\colorbox[RGB]}\expandafter\cbRGB\expandafter{\detokenize{255,7,7}}{congestive\strut} \setlength{\fboxsep}{0pt}\def\cbRGB{\colorbox[RGB]}\expandafter\cbRGB\expandafter{\detokenize{255,80,80}}{heart\strut} \setlength{\fboxsep}{0pt}\def\cbRGB{\colorbox[RGB]}\expandafter\cbRGB\expandafter{\detokenize{255,198,198}}{failure\strut} \setlength{\fboxsep}{0pt}\def\cbRGB{\colorbox[RGB]}\expandafter\cbRGB\expandafter{\detokenize{255,232,232}}{were\strut} \setlength{\fboxsep}{0pt}\def\cbRGB{\colorbox[RGB]}\expandafter\cbRGB\expandafter{\detokenize{255,237,237}}{included\strut} \setlength{\fboxsep}{0pt}\def\cbRGB{\colorbox[RGB]}\expandafter\cbRGB\expandafter{\detokenize{255,239,239}}{.\strut} \setlength{\fboxsep}{0pt}\def\cbRGB{\colorbox[RGB]}\expandafter\cbRGB\expandafter{\detokenize{255,243,243}}{data\strut} \setlength{\fboxsep}{0pt}\def\cbRGB{\colorbox[RGB]}\expandafter\cbRGB\expandafter{\detokenize{255,244,244}}{extraction\strut} \setlength{\fboxsep}{0pt}\def\cbRGB{\colorbox[RGB]}\expandafter\cbRGB\expandafter{\detokenize{255,246,246}}{:\strut} \setlength{\fboxsep}{0pt}\def\cbRGB{\colorbox[RGB]}\expandafter\cbRGB\expandafter{\detokenize{255,245,245}}{a\strut} \setlength{\fboxsep}{0pt}\def\cbRGB{\colorbox[RGB]}\expandafter\cbRGB\expandafter{\detokenize{255,246,246}}{specified\strut} \setlength{\fboxsep}{0pt}\def\cbRGB{\colorbox[RGB]}\expandafter\cbRGB\expandafter{\detokenize{255,247,247}}{protocol\strut} \setlength{\fboxsep}{0pt}\def\cbRGB{\colorbox[RGB]}\expandafter\cbRGB\expandafter{\detokenize{255,247,247}}{was\strut} \setlength{\fboxsep}{0pt}\def\cbRGB{\colorbox[RGB]}\expandafter\cbRGB\expandafter{\detokenize{255,248,248}}{followed\strut} \setlength{\fboxsep}{0pt}\def\cbRGB{\colorbox[RGB]}\expandafter\cbRGB\expandafter{\detokenize{255,246,246}}{to\strut} \setlength{\fboxsep}{0pt}\def\cbRGB{\colorbox[RGB]}\expandafter\cbRGB\expandafter{\detokenize{255,243,243}}{extract\strut} \setlength{\fboxsep}{0pt}\def\cbRGB{\colorbox[RGB]}\expandafter\cbRGB\expandafter{\detokenize{255,241,241}}{data\strut} \setlength{\fboxsep}{0pt}\def\cbRGB{\colorbox[RGB]}\expandafter\cbRGB\expandafter{\detokenize{255,238,238}}{on\strut} \setlength{\fboxsep}{0pt}\def\cbRGB{\colorbox[RGB]}\expandafter\cbRGB\expandafter{\detokenize{255,238,238}}{patient\strut} \setlength{\fboxsep}{0pt}\def\cbRGB{\colorbox[RGB]}\expandafter\cbRGB\expandafter{\detokenize{255,239,239}}{characteristics\strut} \setlength{\fboxsep}{0pt}\def\cbRGB{\colorbox[RGB]}\expandafter\cbRGB\expandafter{\detokenize{255,242,242}}{,\strut} \setlength{\fboxsep}{0pt}\def\cbRGB{\colorbox[RGB]}\expandafter\cbRGB\expandafter{\detokenize{255,243,243}}{unk\strut} \setlength{\fboxsep}{0pt}\def\cbRGB{\colorbox[RGB]}\expandafter\cbRGB\expandafter{\detokenize{255,244,244}}{used\strut} \setlength{\fboxsep}{0pt}\def\cbRGB{\colorbox[RGB]}\expandafter\cbRGB\expandafter{\detokenize{255,247,247}}{,\strut} \setlength{\fboxsep}{0pt}\def\cbRGB{\colorbox[RGB]}\expandafter\cbRGB\expandafter{\detokenize{255,250,250}}{overall\strut} \setlength{\fboxsep}{0pt}\def\cbRGB{\colorbox[RGB]}\expandafter\cbRGB\expandafter{\detokenize{255,249,249}}{mortality\strut} \setlength{\fboxsep}{0pt}\def\cbRGB{\colorbox[RGB]}\expandafter\cbRGB\expandafter{\detokenize{255,251,251}}{,\strut} \setlength{\fboxsep}{0pt}\def\cbRGB{\colorbox[RGB]}\expandafter\cbRGB\expandafter{\detokenize{255,246,246}}{hospitalizations\strut} \setlength{\fboxsep}{0pt}\def\cbRGB{\colorbox[RGB]}\expandafter\cbRGB\expandafter{\detokenize{255,239,239}}{for\strut} \setlength{\fboxsep}{0pt}\def\cbRGB{\colorbox[RGB]}\expandafter\cbRGB\expandafter{\detokenize{255,232,232}}{congestive\strut} \setlength{\fboxsep}{0pt}\def\cbRGB{\colorbox[RGB]}\expandafter\cbRGB\expandafter{\detokenize{255,231,231}}{heart\strut} \setlength{\fboxsep}{0pt}\def\cbRGB{\colorbox[RGB]}\expandafter\cbRGB\expandafter{\detokenize{255,236,236}}{failure\strut} \setlength{\fboxsep}{0pt}\def\cbRGB{\colorbox[RGB]}\expandafter\cbRGB\expandafter{\detokenize{255,239,239}}{,\strut} \setlength{\fboxsep}{0pt}\def\cbRGB{\colorbox[RGB]}\expandafter\cbRGB\expandafter{\detokenize{255,241,241}}{and\strut} \setlength{\fboxsep}{0pt}\def\cbRGB{\colorbox[RGB]}\expandafter\cbRGB\expandafter{\detokenize{255,244,244}}{study\strut} \setlength{\fboxsep}{0pt}\def\cbRGB{\colorbox[RGB]}\expandafter\cbRGB\expandafter{\detokenize{255,245,245}}{quality\strut} \setlength{\fboxsep}{0pt}\def\cbRGB{\colorbox[RGB]}\expandafter\cbRGB\expandafter{\detokenize{255,246,246}}{.\strut} \setlength{\fboxsep}{0pt}\def\cbRGB{\colorbox[RGB]}\expandafter\cbRGB\expandafter{\detokenize{255,247,247}}{data\strut} \setlength{\fboxsep}{0pt}\def\cbRGB{\colorbox[RGB]}\expandafter\cbRGB\expandafter{\detokenize{255,245,245}}{unk\strut} \setlength{\fboxsep}{0pt}\def\cbRGB{\colorbox[RGB]}\expandafter\cbRGB\expandafter{\detokenize{255,244,244}}{:\strut} \setlength{\fboxsep}{0pt}\def\cbRGB{\colorbox[RGB]}\expandafter\cbRGB\expandafter{\detokenize{255,243,243}}{a\strut} \setlength{\fboxsep}{0pt}\def\cbRGB{\colorbox[RGB]}\expandafter\cbRGB\expandafter{\detokenize{255,242,242}}{unk\strut} \setlength{\fboxsep}{0pt}\def\cbRGB{\colorbox[RGB]}\expandafter\cbRGB\expandafter{\detokenize{255,241,241}}{unk\strut} \setlength{\fboxsep}{0pt}\def\cbRGB{\colorbox[RGB]}\expandafter\cbRGB\expandafter{\detokenize{255,242,242}}{model\strut} \setlength{\fboxsep}{0pt}\def\cbRGB{\colorbox[RGB]}\expandafter\cbRGB\expandafter{\detokenize{255,243,243}}{was\strut} \setlength{\fboxsep}{0pt}\def\cbRGB{\colorbox[RGB]}\expandafter\cbRGB\expandafter{\detokenize{255,243,243}}{used\strut} \setlength{\fboxsep}{0pt}\def\cbRGB{\colorbox[RGB]}\expandafter\cbRGB\expandafter{\detokenize{255,242,242}}{to\strut} \setlength{\fboxsep}{0pt}\def\cbRGB{\colorbox[RGB]}\expandafter\cbRGB\expandafter{\detokenize{255,241,241}}{unk\strut} \setlength{\fboxsep}{0pt}\def\cbRGB{\colorbox[RGB]}\expandafter\cbRGB\expandafter{\detokenize{255,239,239}}{the\strut} \setlength{\fboxsep}{0pt}\def\cbRGB{\colorbox[RGB]}\expandafter\cbRGB\expandafter{\detokenize{255,237,237}}{results\strut} \setlength{\fboxsep}{0pt}\def\cbRGB{\colorbox[RGB]}\expandafter\cbRGB\expandafter{\detokenize{255,236,236}}{.\strut} \setlength{\fboxsep}{0pt}\def\cbRGB{\colorbox[RGB]}\expandafter\cbRGB\expandafter{\detokenize{255,235,235}}{a\strut} \setlength{\fboxsep}{0pt}\def\cbRGB{\colorbox[RGB]}\expandafter\cbRGB\expandafter{\detokenize{255,236,236}}{total\strut} \setlength{\fboxsep}{0pt}\def\cbRGB{\colorbox[RGB]}\expandafter\cbRGB\expandafter{\detokenize{255,235,235}}{of\strut} \setlength{\fboxsep}{0pt}\def\cbRGB{\colorbox[RGB]}\expandafter\cbRGB\expandafter{\detokenize{255,235,235}}{qqq\strut} \setlength{\fboxsep}{0pt}\def\cbRGB{\colorbox[RGB]}\expandafter\cbRGB\expandafter{\detokenize{255,236,236}}{trials\strut} \setlength{\fboxsep}{0pt}\def\cbRGB{\colorbox[RGB]}\expandafter\cbRGB\expandafter{\detokenize{255,233,233}}{involving\strut} \setlength{\fboxsep}{0pt}\def\cbRGB{\colorbox[RGB]}\expandafter\cbRGB\expandafter{\detokenize{255,228,228}}{qqq\strut} \setlength{\fboxsep}{0pt}\def\cbRGB{\colorbox[RGB]}\expandafter\cbRGB\expandafter{\detokenize{255,222,222}}{qqq\strut} \setlength{\fboxsep}{0pt}\def\cbRGB{\colorbox[RGB]}\expandafter\cbRGB\expandafter{\detokenize{255,220,220}}{patients\strut} \setlength{\fboxsep}{0pt}\def\cbRGB{\colorbox[RGB]}\expandafter\cbRGB\expandafter{\detokenize{255,220,220}}{were\strut} \setlength{\fboxsep}{0pt}\def\cbRGB{\colorbox[RGB]}\expandafter\cbRGB\expandafter{\detokenize{255,221,221}}{identified\strut} \setlength{\fboxsep}{0pt}\def\cbRGB{\colorbox[RGB]}\expandafter\cbRGB\expandafter{\detokenize{255,225,225}}{.\strut} \setlength{\fboxsep}{0pt}\def\cbRGB{\colorbox[RGB]}\expandafter\cbRGB\expandafter{\detokenize{255,227,227}}{there\strut} \setlength{\fboxsep}{0pt}\def\cbRGB{\colorbox[RGB]}\expandafter\cbRGB\expandafter{\detokenize{255,227,227}}{were\strut} \setlength{\fboxsep}{0pt}\def\cbRGB{\colorbox[RGB]}\expandafter\cbRGB\expandafter{\detokenize{255,229,229}}{qqq\strut} \setlength{\fboxsep}{0pt}\def\cbRGB{\colorbox[RGB]}\expandafter\cbRGB\expandafter{\detokenize{255,224,224}}{deaths\strut} \setlength{\fboxsep}{0pt}\def\cbRGB{\colorbox[RGB]}\expandafter\cbRGB\expandafter{\detokenize{255,218,218}}{among\strut} \setlength{\fboxsep}{0pt}\def\cbRGB{\colorbox[RGB]}\expandafter\cbRGB\expandafter{\detokenize{255,205,205}}{qqq\strut} \setlength{\fboxsep}{0pt}\def\cbRGB{\colorbox[RGB]}\expandafter\cbRGB\expandafter{\detokenize{255,209,209}}{patients\strut} \setlength{\fboxsep}{0pt}\def\cbRGB{\colorbox[RGB]}\expandafter\cbRGB\expandafter{\detokenize{255,214,214}}{randomly\strut} \setlength{\fboxsep}{0pt}\def\cbRGB{\colorbox[RGB]}\expandafter\cbRGB\expandafter{\detokenize{255,225,225}}{assigned\strut} \setlength{\fboxsep}{0pt}\def\cbRGB{\colorbox[RGB]}\expandafter\cbRGB\expandafter{\detokenize{255,231,231}}{to\strut} \setlength{\fboxsep}{0pt}\def\cbRGB{\colorbox[RGB]}\expandafter\cbRGB\expandafter{\detokenize{255,236,236}}{placebo\strut} \setlength{\fboxsep}{0pt}\def\cbRGB{\colorbox[RGB]}\expandafter\cbRGB\expandafter{\detokenize{255,237,237}}{and\strut} \setlength{\fboxsep}{0pt}\def\cbRGB{\colorbox[RGB]}\expandafter\cbRGB\expandafter{\detokenize{255,237,237}}{qqq\strut} \setlength{\fboxsep}{0pt}\def\cbRGB{\colorbox[RGB]}\expandafter\cbRGB\expandafter{\detokenize{255,232,232}}{deaths\strut} \setlength{\fboxsep}{0pt}\def\cbRGB{\colorbox[RGB]}\expandafter\cbRGB\expandafter{\detokenize{255,228,228}}{among\strut} \setlength{\fboxsep}{0pt}\def\cbRGB{\colorbox[RGB]}\expandafter\cbRGB\expandafter{\detokenize{255,223,223}}{qqq\strut} \setlength{\fboxsep}{0pt}\def\cbRGB{\colorbox[RGB]}\expandafter\cbRGB\expandafter{\detokenize{255,225,225}}{patients\strut} \setlength{\fboxsep}{0pt}\def\cbRGB{\colorbox[RGB]}\expandafter\cbRGB\expandafter{\detokenize{255,224,224}}{assigned\strut} \setlength{\fboxsep}{0pt}\def\cbRGB{\colorbox[RGB]}\expandafter\cbRGB\expandafter{\detokenize{255,228,228}}{to\strut} \setlength{\fboxsep}{0pt}\def\cbRGB{\colorbox[RGB]}\expandafter\cbRGB\expandafter{\detokenize{255,231,231}}{unk\strut} \setlength{\fboxsep}{0pt}\def\cbRGB{\colorbox[RGB]}\expandafter\cbRGB\expandafter{\detokenize{255,236,236}}{therapy\strut} \setlength{\fboxsep}{0pt}\def\cbRGB{\colorbox[RGB]}\expandafter\cbRGB\expandafter{\detokenize{255,237,237}}{.\strut} \setlength{\fboxsep}{0pt}\def\cbRGB{\colorbox[RGB]}\expandafter\cbRGB\expandafter{\detokenize{255,237,237}}{in\strut} \setlength{\fboxsep}{0pt}\def\cbRGB{\colorbox[RGB]}\expandafter\cbRGB\expandafter{\detokenize{255,237,237}}{these\strut} \setlength{\fboxsep}{0pt}\def\cbRGB{\colorbox[RGB]}\expandafter\cbRGB\expandafter{\detokenize{255,237,237}}{groups\strut} \setlength{\fboxsep}{0pt}\def\cbRGB{\colorbox[RGB]}\expandafter\cbRGB\expandafter{\detokenize{255,238,238}}{,\strut} \setlength{\fboxsep}{0pt}\def\cbRGB{\colorbox[RGB]}\expandafter\cbRGB\expandafter{\detokenize{255,236,236}}{qqq\strut} \setlength{\fboxsep}{0pt}\def\cbRGB{\colorbox[RGB]}\expandafter\cbRGB\expandafter{\detokenize{255,232,232}}{and\strut} \setlength{\fboxsep}{0pt}\def\cbRGB{\colorbox[RGB]}\expandafter\cbRGB\expandafter{\detokenize{255,227,227}}{qqq\strut} \setlength{\fboxsep}{0pt}\def\cbRGB{\colorbox[RGB]}\expandafter\cbRGB\expandafter{\detokenize{255,219,219}}{patients\strut} \setlength{\fboxsep}{0pt}\def\cbRGB{\colorbox[RGB]}\expandafter\cbRGB\expandafter{\detokenize{255,221,221}}{,\strut} \setlength{\fboxsep}{0pt}\def\cbRGB{\colorbox[RGB]}\expandafter\cbRGB\expandafter{\detokenize{255,224,224}}{respectively\strut} \setlength{\fboxsep}{0pt}\def\cbRGB{\colorbox[RGB]}\expandafter\cbRGB\expandafter{\detokenize{255,232,232}}{,\strut} \setlength{\fboxsep}{0pt}\def\cbRGB{\colorbox[RGB]}\expandafter\cbRGB\expandafter{\detokenize{255,234,234}}{required\strut} \setlength{\fboxsep}{0pt}\def\cbRGB{\colorbox[RGB]}\expandafter\cbRGB\expandafter{\detokenize{255,234,234}}{hospitalization\strut} \setlength{\fboxsep}{0pt}\def\cbRGB{\colorbox[RGB]}\expandafter\cbRGB\expandafter{\detokenize{255,233,233}}{for\strut} \setlength{\fboxsep}{0pt}\def\cbRGB{\colorbox[RGB]}\expandafter\cbRGB\expandafter{\detokenize{255,231,231}}{congestive\strut} \setlength{\fboxsep}{0pt}\def\cbRGB{\colorbox[RGB]}\expandafter\cbRGB\expandafter{\detokenize{255,232,232}}{heart\strut} \setlength{\fboxsep}{0pt}\def\cbRGB{\colorbox[RGB]}\expandafter\cbRGB\expandafter{\detokenize{255,234,234}}{failure\strut} \setlength{\fboxsep}{0pt}\def\cbRGB{\colorbox[RGB]}\expandafter\cbRGB\expandafter{\detokenize{255,236,236}}{.\strut} \setlength{\fboxsep}{0pt}\def\cbRGB{\colorbox[RGB]}\expandafter\cbRGB\expandafter{\detokenize{255,237,237}}{the\strut} \setlength{\fboxsep}{0pt}\def\cbRGB{\colorbox[RGB]}\expandafter\cbRGB\expandafter{\detokenize{255,239,239}}{probability\strut} \setlength{\fboxsep}{0pt}\def\cbRGB{\colorbox[RGB]}\expandafter\cbRGB\expandafter{\detokenize{255,237,237}}{that\strut} \setlength{\fboxsep}{0pt}\def\cbRGB{\colorbox[RGB]}\expandafter\cbRGB\expandafter{\detokenize{255,235,235}}{unk\strut} \setlength{\fboxsep}{0pt}\def\cbRGB{\colorbox[RGB]}\expandafter\cbRGB\expandafter{\detokenize{255,232,232}}{therapy\strut} \setlength{\fboxsep}{0pt}\def\cbRGB{\colorbox[RGB]}\expandafter\cbRGB\expandafter{\detokenize{255,233,233}}{reduced\strut} \setlength{\fboxsep}{0pt}\def\cbRGB{\colorbox[RGB]}\expandafter\cbRGB\expandafter{\detokenize{255,239,239}}{total\strut} \setlength{\fboxsep}{0pt}\def\cbRGB{\colorbox[RGB]}\expandafter\cbRGB\expandafter{\detokenize{255,241,241}}{mortality\strut} \setlength{\fboxsep}{0pt}\def\cbRGB{\colorbox[RGB]}\expandafter\cbRGB\expandafter{\detokenize{255,246,246}}{and\strut} \setlength{\fboxsep}{0pt}\def\cbRGB{\colorbox[RGB]}\expandafter\cbRGB\expandafter{\detokenize{255,242,242}}{hospitalizations\strut} \setlength{\fboxsep}{0pt}\def\cbRGB{\colorbox[RGB]}\expandafter\cbRGB\expandafter{\detokenize{255,238,238}}{for\strut} \setlength{\fboxsep}{0pt}\def\cbRGB{\colorbox[RGB]}\expandafter\cbRGB\expandafter{\detokenize{255,230,230}}{congestive\strut} \setlength{\fboxsep}{0pt}\def\cbRGB{\colorbox[RGB]}\expandafter\cbRGB\expandafter{\detokenize{255,228,228}}{heart\strut} \setlength{\fboxsep}{0pt}\def\cbRGB{\colorbox[RGB]}\expandafter\cbRGB\expandafter{\detokenize{255,230,230}}{failure\strut} \setlength{\fboxsep}{0pt}\def\cbRGB{\colorbox[RGB]}\expandafter\cbRGB\expandafter{\detokenize{255,232,232}}{was\strut} \setlength{\fboxsep}{0pt}\def\cbRGB{\colorbox[RGB]}\expandafter\cbRGB\expandafter{\detokenize{255,233,233}}{almost\strut} \setlength{\fboxsep}{0pt}\def\cbRGB{\colorbox[RGB]}\expandafter\cbRGB\expandafter{\detokenize{255,235,235}}{qqq\strut} \setlength{\fboxsep}{0pt}\def\cbRGB{\colorbox[RGB]}\expandafter\cbRGB\expandafter{\detokenize{255,235,235}}{\%\strut} \setlength{\fboxsep}{0pt}\def\cbRGB{\colorbox[RGB]}\expandafter\cbRGB\expandafter{\detokenize{255,235,235}}{.\strut} \setlength{\fboxsep}{0pt}\def\cbRGB{\colorbox[RGB]}\expandafter\cbRGB\expandafter{\detokenize{255,233,233}}{the\strut} \setlength{\fboxsep}{0pt}\def\cbRGB{\colorbox[RGB]}\expandafter\cbRGB\expandafter{\detokenize{255,233,233}}{best\strut} \setlength{\fboxsep}{0pt}\def\cbRGB{\colorbox[RGB]}\expandafter\cbRGB\expandafter{\detokenize{255,232,232}}{estimates\strut} \setlength{\fboxsep}{0pt}\def\cbRGB{\colorbox[RGB]}\expandafter\cbRGB\expandafter{\detokenize{255,231,231}}{of\strut} \setlength{\fboxsep}{0pt}\def\cbRGB{\colorbox[RGB]}\expandafter\cbRGB\expandafter{\detokenize{255,228,228}}{these\strut} \setlength{\fboxsep}{0pt}\def\cbRGB{\colorbox[RGB]}\expandafter\cbRGB\expandafter{\detokenize{255,231,231}}{advantages\strut} \setlength{\fboxsep}{0pt}\def\cbRGB{\colorbox[RGB]}\expandafter\cbRGB\expandafter{\detokenize{255,233,233}}{are\strut} \setlength{\fboxsep}{0pt}\def\cbRGB{\colorbox[RGB]}\expandafter\cbRGB\expandafter{\detokenize{255,235,235}}{qqq\strut} \setlength{\fboxsep}{0pt}\def\cbRGB{\colorbox[RGB]}\expandafter\cbRGB\expandafter{\detokenize{255,237,237}}{unk\strut} \setlength{\fboxsep}{0pt}\def\cbRGB{\colorbox[RGB]}\expandafter\cbRGB\expandafter{\detokenize{255,237,237}}{unk\strut} \setlength{\fboxsep}{0pt}\def\cbRGB{\colorbox[RGB]}\expandafter\cbRGB\expandafter{\detokenize{255,233,233}}{and\strut} \setlength{\fboxsep}{0pt}\def\cbRGB{\colorbox[RGB]}\expandafter\cbRGB\expandafter{\detokenize{255,233,233}}{qqq\strut} \setlength{\fboxsep}{0pt}\def\cbRGB{\colorbox[RGB]}\expandafter\cbRGB\expandafter{\detokenize{255,234,234}}{fewer\strut} \setlength{\fboxsep}{0pt}\def\cbRGB{\colorbox[RGB]}\expandafter\cbRGB\expandafter{\detokenize{255,231,231}}{hospitalizations\strut} \setlength{\fboxsep}{0pt}\def\cbRGB{\colorbox[RGB]}\expandafter\cbRGB\expandafter{\detokenize{255,226,226}}{per\strut} \setlength{\fboxsep}{0pt}\def\cbRGB{\colorbox[RGB]}\expandafter\cbRGB\expandafter{\detokenize{255,222,222}}{qqq\strut} \setlength{\fboxsep}{0pt}\def\cbRGB{\colorbox[RGB]}\expandafter\cbRGB\expandafter{\detokenize{255,223,223}}{patients\strut} \setlength{\fboxsep}{0pt}\def\cbRGB{\colorbox[RGB]}\expandafter\cbRGB\expandafter{\detokenize{255,221,221}}{treated\strut} \setlength{\fboxsep}{0pt}\def\cbRGB{\colorbox[RGB]}\expandafter\cbRGB\expandafter{\detokenize{255,223,223}}{in\strut} \setlength{\fboxsep}{0pt}\def\cbRGB{\colorbox[RGB]}\expandafter\cbRGB\expandafter{\detokenize{255,229,229}}{the\strut} \setlength{\fboxsep}{0pt}\def\cbRGB{\colorbox[RGB]}\expandafter\cbRGB\expandafter{\detokenize{255,235,235}}{first\strut} \setlength{\fboxsep}{0pt}\def\cbRGB{\colorbox[RGB]}\expandafter\cbRGB\expandafter{\detokenize{255,237,237}}{year\strut} \setlength{\fboxsep}{0pt}\def\cbRGB{\colorbox[RGB]}\expandafter\cbRGB\expandafter{\detokenize{255,237,237}}{after\strut} \setlength{\fboxsep}{0pt}\def\cbRGB{\colorbox[RGB]}\expandafter\cbRGB\expandafter{\detokenize{255,238,238}}{therapy\strut} \setlength{\fboxsep}{0pt}\def\cbRGB{\colorbox[RGB]}\expandafter\cbRGB\expandafter{\detokenize{255,238,238}}{.\strut} \setlength{\fboxsep}{0pt}\def\cbRGB{\colorbox[RGB]}\expandafter\cbRGB\expandafter{\detokenize{255,237,237}}{the\strut} \setlength{\fboxsep}{0pt}\def\cbRGB{\colorbox[RGB]}\expandafter\cbRGB\expandafter{\detokenize{255,236,236}}{probability\strut} \setlength{\fboxsep}{0pt}\def\cbRGB{\colorbox[RGB]}\expandafter\cbRGB\expandafter{\detokenize{255,235,235}}{that\strut} \setlength{\fboxsep}{0pt}\def\cbRGB{\colorbox[RGB]}\expandafter\cbRGB\expandafter{\detokenize{255,233,233}}{these\strut} \setlength{\fboxsep}{0pt}\def\cbRGB{\colorbox[RGB]}\expandafter\cbRGB\expandafter{\detokenize{255,233,233}}{benefits\strut} \setlength{\fboxsep}{0pt}\def\cbRGB{\colorbox[RGB]}\expandafter\cbRGB\expandafter{\detokenize{255,232,232}}{are\strut} \setlength{\fboxsep}{0pt}\def\cbRGB{\colorbox[RGB]}\expandafter\cbRGB\expandafter{\detokenize{255,233,233}}{clinically\strut} \setlength{\fboxsep}{0pt}\def\cbRGB{\colorbox[RGB]}\expandafter\cbRGB\expandafter{\detokenize{255,232,232}}{significant\strut} \setlength{\fboxsep}{0pt}\def\cbRGB{\colorbox[RGB]}\expandafter\cbRGB\expandafter{\detokenize{255,234,234}}{(\strut} \setlength{\fboxsep}{0pt}\def\cbRGB{\colorbox[RGB]}\expandafter\cbRGB\expandafter{\detokenize{255,235,235}}{>\strut} \setlength{\fboxsep}{0pt}\def\cbRGB{\colorbox[RGB]}\expandafter\cbRGB\expandafter{\detokenize{255,237,237}}{qqq\strut} \setlength{\fboxsep}{0pt}\def\cbRGB{\colorbox[RGB]}\expandafter\cbRGB\expandafter{\detokenize{255,237,237}}{unk\strut} \setlength{\fboxsep}{0pt}\def\cbRGB{\colorbox[RGB]}\expandafter\cbRGB\expandafter{\detokenize{255,237,237}}{unk\strut} \setlength{\fboxsep}{0pt}\def\cbRGB{\colorbox[RGB]}\expandafter\cbRGB\expandafter{\detokenize{255,237,237}}{or\strut} \setlength{\fboxsep}{0pt}\def\cbRGB{\colorbox[RGB]}\expandafter\cbRGB\expandafter{\detokenize{255,230,230}}{>\strut} \setlength{\fboxsep}{0pt}\def\cbRGB{\colorbox[RGB]}\expandafter\cbRGB\expandafter{\detokenize{255,230,230}}{qqq\strut} \setlength{\fboxsep}{0pt}\def\cbRGB{\colorbox[RGB]}\expandafter\cbRGB\expandafter{\detokenize{255,231,231}}{fewer\strut} \setlength{\fboxsep}{0pt}\def\cbRGB{\colorbox[RGB]}\expandafter\cbRGB\expandafter{\detokenize{255,230,230}}{hospitalizations\strut} \setlength{\fboxsep}{0pt}\def\cbRGB{\colorbox[RGB]}\expandafter\cbRGB\expandafter{\detokenize{255,226,226}}{per\strut} \setlength{\fboxsep}{0pt}\def\cbRGB{\colorbox[RGB]}\expandafter\cbRGB\expandafter{\detokenize{255,224,224}}{qqq\strut} \setlength{\fboxsep}{0pt}\def\cbRGB{\colorbox[RGB]}\expandafter\cbRGB\expandafter{\detokenize{255,225,225}}{patients\strut} \setlength{\fboxsep}{0pt}\def\cbRGB{\colorbox[RGB]}\expandafter\cbRGB\expandafter{\detokenize{255,219,219}}{treated\strut} \setlength{\fboxsep}{0pt}\def\cbRGB{\colorbox[RGB]}\expandafter\cbRGB\expandafter{\detokenize{255,221,221}}{)\strut} \setlength{\fboxsep}{0pt}\def\cbRGB{\colorbox[RGB]}\expandafter\cbRGB\expandafter{\detokenize{255,227,227}}{is\strut} \setlength{\fboxsep}{0pt}\def\cbRGB{\colorbox[RGB]}\expandafter\cbRGB\expandafter{\detokenize{255,235,235}}{qqq\strut} \setlength{\fboxsep}{0pt}\def\cbRGB{\colorbox[RGB]}\expandafter\cbRGB\expandafter{\detokenize{255,234,234}}{\%\strut} \setlength{\fboxsep}{0pt}\def\cbRGB{\colorbox[RGB]}\expandafter\cbRGB\expandafter{\detokenize{255,236,236}}{.\strut} \setlength{\fboxsep}{0pt}\def\cbRGB{\colorbox[RGB]}\expandafter\cbRGB\expandafter{\detokenize{255,238,238}}{both\strut} \setlength{\fboxsep}{0pt}\def\cbRGB{\colorbox[RGB]}\expandafter\cbRGB\expandafter{\detokenize{255,239,239}}{selective\strut} \setlength{\fboxsep}{0pt}\def\cbRGB{\colorbox[RGB]}\expandafter\cbRGB\expandafter{\detokenize{255,239,239}}{and\strut} \setlength{\fboxsep}{0pt}\def\cbRGB{\colorbox[RGB]}\expandafter\cbRGB\expandafter{\detokenize{255,237,237}}{unk\strut} \setlength{\fboxsep}{0pt}\def\cbRGB{\colorbox[RGB]}\expandafter\cbRGB\expandafter{\detokenize{255,237,237}}{agents\strut} \setlength{\fboxsep}{0pt}\def\cbRGB{\colorbox[RGB]}\expandafter\cbRGB\expandafter{\detokenize{255,236,236}}{produced\strut} \setlength{\fboxsep}{0pt}\def\cbRGB{\colorbox[RGB]}\expandafter\cbRGB\expandafter{\detokenize{255,236,236}}{these\strut} \setlength{\fboxsep}{0pt}\def\cbRGB{\colorbox[RGB]}\expandafter\cbRGB\expandafter{\detokenize{255,234,234}}{unk\strut} \setlength{\fboxsep}{0pt}\def\cbRGB{\colorbox[RGB]}\expandafter\cbRGB\expandafter{\detokenize{255,235,235}}{effects\strut} \setlength{\fboxsep}{0pt}\def\cbRGB{\colorbox[RGB]}\expandafter\cbRGB\expandafter{\detokenize{255,236,236}}{.\strut} \setlength{\fboxsep}{0pt}\def\cbRGB{\colorbox[RGB]}\expandafter\cbRGB\expandafter{\detokenize{255,235,235}}{the\strut} \setlength{\fboxsep}{0pt}\def\cbRGB{\colorbox[RGB]}\expandafter\cbRGB\expandafter{\detokenize{255,236,236}}{results\strut} \setlength{\fboxsep}{0pt}\def\cbRGB{\colorbox[RGB]}\expandafter\cbRGB\expandafter{\detokenize{255,236,236}}{are\strut} \setlength{\fboxsep}{0pt}\def\cbRGB{\colorbox[RGB]}\expandafter\cbRGB\expandafter{\detokenize{255,237,237}}{unk\strut} \setlength{\fboxsep}{0pt}\def\cbRGB{\colorbox[RGB]}\expandafter\cbRGB\expandafter{\detokenize{255,236,236}}{to\strut} \setlength{\fboxsep}{0pt}\def\cbRGB{\colorbox[RGB]}\expandafter\cbRGB\expandafter{\detokenize{255,239,239}}{any\strut} \setlength{\fboxsep}{0pt}\def\cbRGB{\colorbox[RGB]}\expandafter\cbRGB\expandafter{\detokenize{255,243,243}}{unk\strut} \setlength{\fboxsep}{0pt}\def\cbRGB{\colorbox[RGB]}\expandafter\cbRGB\expandafter{\detokenize{255,245,245}}{publication\strut} \setlength{\fboxsep}{0pt}\def\cbRGB{\colorbox[RGB]}\expandafter\cbRGB\expandafter{\detokenize{255,244,244}}{unk\strut} \setlength{\fboxsep}{0pt}\def\cbRGB{\colorbox[RGB]}\expandafter\cbRGB\expandafter{\detokenize{255,242,242}}{.\strut} \setlength{\fboxsep}{0pt}\def\cbRGB{\colorbox[RGB]}\expandafter\cbRGB\expandafter{\detokenize{255,240,240}}{conclusions\strut} \setlength{\fboxsep}{0pt}\def\cbRGB{\colorbox[RGB]}\expandafter\cbRGB\expandafter{\detokenize{255,238,238}}{:\strut} \setlength{\fboxsep}{0pt}\def\cbRGB{\colorbox[RGB]}\expandafter\cbRGB\expandafter{\detokenize{255,233,233}}{unk\strut} \setlength{\fboxsep}{0pt}\def\cbRGB{\colorbox[RGB]}\expandafter\cbRGB\expandafter{\detokenize{255,228,228}}{therapy\strut} \setlength{\fboxsep}{0pt}\def\cbRGB{\colorbox[RGB]}\expandafter\cbRGB\expandafter{\detokenize{255,221,221}}{is\strut} \setlength{\fboxsep}{0pt}\def\cbRGB{\colorbox[RGB]}\expandafter\cbRGB\expandafter{\detokenize{255,214,214}}{associated\strut} \setlength{\fboxsep}{0pt}\def\cbRGB{\colorbox[RGB]}\expandafter\cbRGB\expandafter{\detokenize{255,207,207}}{with\strut} \setlength{\fboxsep}{0pt}\def\cbRGB{\colorbox[RGB]}\expandafter\cbRGB\expandafter{\detokenize{255,204,204}}{clinically\strut} \setlength{\fboxsep}{0pt}\def\cbRGB{\colorbox[RGB]}\expandafter\cbRGB\expandafter{\detokenize{255,211,211}}{meaningful\strut} \setlength{\fboxsep}{0pt}\def\cbRGB{\colorbox[RGB]}\expandafter\cbRGB\expandafter{\detokenize{255,222,222}}{reductions\strut} \setlength{\fboxsep}{0pt}\def\cbRGB{\colorbox[RGB]}\expandafter\cbRGB\expandafter{\detokenize{255,239,239}}{in\strut} \setlength{\fboxsep}{0pt}\def\cbRGB{\colorbox[RGB]}\expandafter\cbRGB\expandafter{\detokenize{255,242,242}}{mortality\strut} \setlength{\fboxsep}{0pt}\def\cbRGB{\colorbox[RGB]}\expandafter\cbRGB\expandafter{\detokenize{255,245,245}}{and\strut} \setlength{\fboxsep}{0pt}\def\cbRGB{\colorbox[RGB]}\expandafter\cbRGB\expandafter{\detokenize{255,234,234}}{morbidity\strut} \setlength{\fboxsep}{0pt}\def\cbRGB{\colorbox[RGB]}\expandafter\cbRGB\expandafter{\detokenize{255,204,204}}{in\strut} \setlength{\fboxsep}{0pt}\def\cbRGB{\colorbox[RGB]}\expandafter\cbRGB\expandafter{\detokenize{255,136,136}}{patients\strut} \setlength{\fboxsep}{0pt}\def\cbRGB{\colorbox[RGB]}\expandafter\cbRGB\expandafter{\detokenize{255,69,69}}{with\strut} \setlength{\fboxsep}{0pt}\def\cbRGB{\colorbox[RGB]}\expandafter\cbRGB\expandafter{\detokenize{255,29,29}}{stable\strut} \setlength{\fboxsep}{0pt}\def\cbRGB{\colorbox[RGB]}\expandafter\cbRGB\expandafter{\detokenize{255,71,71}}{congestive\strut} \setlength{\fboxsep}{0pt}\def\cbRGB{\colorbox[RGB]}\expandafter\cbRGB\expandafter{\detokenize{255,140,140}}{heart\strut} \setlength{\fboxsep}{0pt}\def\cbRGB{\colorbox[RGB]}\expandafter\cbRGB\expandafter{\detokenize{255,205,205}}{failure\strut} \setlength{\fboxsep}{0pt}\def\cbRGB{\colorbox[RGB]}\expandafter\cbRGB\expandafter{\detokenize{255,228,228}}{and\strut} \setlength{\fboxsep}{0pt}\def\cbRGB{\colorbox[RGB]}\expandafter\cbRGB\expandafter{\detokenize{255,235,235}}{should\strut} \setlength{\fboxsep}{0pt}\def\cbRGB{\colorbox[RGB]}\expandafter\cbRGB\expandafter{\detokenize{255,245,245}}{be\strut} \setlength{\fboxsep}{0pt}\def\cbRGB{\colorbox[RGB]}\expandafter\cbRGB\expandafter{\detokenize{255,249,249}}{routinely\strut} \setlength{\fboxsep}{0pt}\def\cbRGB{\colorbox[RGB]}\expandafter\cbRGB\expandafter{\detokenize{255,251,251}}{offered\strut} \setlength{\fboxsep}{0pt}\def\cbRGB{\colorbox[RGB]}\expandafter\cbRGB\expandafter{\detokenize{255,248,248}}{to\strut} \setlength{\fboxsep}{0pt}\def\cbRGB{\colorbox[RGB]}\expandafter\cbRGB\expandafter{\detokenize{255,241,241}}{all\strut} \setlength{\fboxsep}{0pt}\def\cbRGB{\colorbox[RGB]}\expandafter\cbRGB\expandafter{\detokenize{255,233,233}}{patients\strut} \setlength{\fboxsep}{0pt}\def\cbRGB{\colorbox[RGB]}\expandafter\cbRGB\expandafter{\detokenize{255,223,223}}{similar\strut} \setlength{\fboxsep}{0pt}\def\cbRGB{\colorbox[RGB]}\expandafter\cbRGB\expandafter{\detokenize{255,222,222}}{to\strut} \setlength{\fboxsep}{0pt}\def\cbRGB{\colorbox[RGB]}\expandafter\cbRGB\expandafter{\detokenize{255,226,226}}{those\strut} \setlength{\fboxsep}{0pt}\def\cbRGB{\colorbox[RGB]}\expandafter\cbRGB\expandafter{\detokenize{255,235,235}}{included\strut} \setlength{\fboxsep}{0pt}\def\cbRGB{\colorbox[RGB]}\expandafter\cbRGB\expandafter{\detokenize{255,237,237}}{in\strut} \setlength{\fboxsep}{0pt}\def\cbRGB{\colorbox[RGB]}\expandafter\cbRGB\expandafter{\detokenize{255,237,237}}{trials\strut} \setlength{\fboxsep}{0pt}\def\cbRGB{\colorbox[RGB]}\expandafter\cbRGB\expandafter{\detokenize{255,246,246}}{.\strut} 

\setlength{\fboxsep}{0pt}\def\cbRGB{\colorbox[RGB]}\expandafter\cbRGB\expandafter{\detokenize{233,255,233}}{purpose\strut} \setlength{\fboxsep}{0pt}\def\cbRGB{\colorbox[RGB]}\expandafter\cbRGB\expandafter{\detokenize{208,255,208}}{:\strut} \setlength{\fboxsep}{0pt}\def\cbRGB{\colorbox[RGB]}\expandafter\cbRGB\expandafter{\detokenize{222,255,222}}{congestive\strut} \setlength{\fboxsep}{0pt}\def\cbRGB{\colorbox[RGB]}\expandafter\cbRGB\expandafter{\detokenize{223,255,223}}{heart\strut} \setlength{\fboxsep}{0pt}\def\cbRGB{\colorbox[RGB]}\expandafter\cbRGB\expandafter{\detokenize{233,255,233}}{failure\strut} \setlength{\fboxsep}{0pt}\def\cbRGB{\colorbox[RGB]}\expandafter\cbRGB\expandafter{\detokenize{234,255,234}}{is\strut} \setlength{\fboxsep}{0pt}\def\cbRGB{\colorbox[RGB]}\expandafter\cbRGB\expandafter{\detokenize{227,255,227}}{an\strut} \setlength{\fboxsep}{0pt}\def\cbRGB{\colorbox[RGB]}\expandafter\cbRGB\expandafter{\detokenize{224,255,224}}{important\strut} \setlength{\fboxsep}{0pt}\def\cbRGB{\colorbox[RGB]}\expandafter\cbRGB\expandafter{\detokenize{223,255,223}}{cause\strut} \setlength{\fboxsep}{0pt}\def\cbRGB{\colorbox[RGB]}\expandafter\cbRGB\expandafter{\detokenize{225,255,225}}{of\strut} \setlength{\fboxsep}{0pt}\def\cbRGB{\colorbox[RGB]}\expandafter\cbRGB\expandafter{\detokenize{232,255,232}}{patient\strut} \setlength{\fboxsep}{0pt}\def\cbRGB{\colorbox[RGB]}\expandafter\cbRGB\expandafter{\detokenize{238,255,238}}{morbidity\strut} \setlength{\fboxsep}{0pt}\def\cbRGB{\colorbox[RGB]}\expandafter\cbRGB\expandafter{\detokenize{246,255,246}}{and\strut} \setlength{\fboxsep}{0pt}\def\cbRGB{\colorbox[RGB]}\expandafter\cbRGB\expandafter{\detokenize{238,255,238}}{mortality\strut} \setlength{\fboxsep}{0pt}\def\cbRGB{\colorbox[RGB]}\expandafter\cbRGB\expandafter{\detokenize{239,255,239}}{.\strut} \setlength{\fboxsep}{0pt}\def\cbRGB{\colorbox[RGB]}\expandafter\cbRGB\expandafter{\detokenize{234,255,234}}{although\strut} \setlength{\fboxsep}{0pt}\def\cbRGB{\colorbox[RGB]}\expandafter\cbRGB\expandafter{\detokenize{225,255,225}}{several\strut} \setlength{\fboxsep}{0pt}\def\cbRGB{\colorbox[RGB]}\expandafter\cbRGB\expandafter{\detokenize{207,255,207}}{randomized\strut} \setlength{\fboxsep}{0pt}\def\cbRGB{\colorbox[RGB]}\expandafter\cbRGB\expandafter{\detokenize{202,255,202}}{clinical\strut} \setlength{\fboxsep}{0pt}\def\cbRGB{\colorbox[RGB]}\expandafter\cbRGB\expandafter{\detokenize{199,255,199}}{trials\strut} \setlength{\fboxsep}{0pt}\def\cbRGB{\colorbox[RGB]}\expandafter\cbRGB\expandafter{\detokenize{191,255,191}}{have\strut} \setlength{\fboxsep}{0pt}\def\cbRGB{\colorbox[RGB]}\expandafter\cbRGB\expandafter{\detokenize{171,255,171}}{compared\strut} \setlength{\fboxsep}{0pt}\def\cbRGB{\colorbox[RGB]}\expandafter\cbRGB\expandafter{\detokenize{142,255,142}}{beta-blockers\strut} \setlength{\fboxsep}{0pt}\def\cbRGB{\colorbox[RGB]}\expandafter\cbRGB\expandafter{\detokenize{149,255,149}}{with\strut} \setlength{\fboxsep}{0pt}\def\cbRGB{\colorbox[RGB]}\expandafter\cbRGB\expandafter{\detokenize{164,255,164}}{placebo\strut} \setlength{\fboxsep}{0pt}\def\cbRGB{\colorbox[RGB]}\expandafter\cbRGB\expandafter{\detokenize{206,255,206}}{for\strut} \setlength{\fboxsep}{0pt}\def\cbRGB{\colorbox[RGB]}\expandafter\cbRGB\expandafter{\detokenize{224,255,224}}{treatment\strut} \setlength{\fboxsep}{0pt}\def\cbRGB{\colorbox[RGB]}\expandafter\cbRGB\expandafter{\detokenize{238,255,238}}{of\strut} \setlength{\fboxsep}{0pt}\def\cbRGB{\colorbox[RGB]}\expandafter\cbRGB\expandafter{\detokenize{237,255,237}}{congestive\strut} \setlength{\fboxsep}{0pt}\def\cbRGB{\colorbox[RGB]}\expandafter\cbRGB\expandafter{\detokenize{235,255,235}}{heart\strut} \setlength{\fboxsep}{0pt}\def\cbRGB{\colorbox[RGB]}\expandafter\cbRGB\expandafter{\detokenize{241,255,241}}{failure\strut} \setlength{\fboxsep}{0pt}\def\cbRGB{\colorbox[RGB]}\expandafter\cbRGB\expandafter{\detokenize{250,255,250}}{,\strut} \setlength{\fboxsep}{0pt}\def\cbRGB{\colorbox[RGB]}\expandafter\cbRGB\expandafter{\detokenize{255,255,255}}{a\strut} \setlength{\fboxsep}{0pt}\def\cbRGB{\colorbox[RGB]}\expandafter\cbRGB\expandafter{\detokenize{238,255,238}}{meta-analysis\strut} \setlength{\fboxsep}{0pt}\def\cbRGB{\colorbox[RGB]}\expandafter\cbRGB\expandafter{\detokenize{234,255,234}}{unk\strut} \setlength{\fboxsep}{0pt}\def\cbRGB{\colorbox[RGB]}\expandafter\cbRGB\expandafter{\detokenize{234,255,234}}{the\strut} \setlength{\fboxsep}{0pt}\def\cbRGB{\colorbox[RGB]}\expandafter\cbRGB\expandafter{\detokenize{245,255,245}}{effect\strut} \setlength{\fboxsep}{0pt}\def\cbRGB{\colorbox[RGB]}\expandafter\cbRGB\expandafter{\detokenize{248,255,248}}{on\strut} \setlength{\fboxsep}{0pt}\def\cbRGB{\colorbox[RGB]}\expandafter\cbRGB\expandafter{\detokenize{235,255,235}}{mortality\strut} \setlength{\fboxsep}{0pt}\def\cbRGB{\colorbox[RGB]}\expandafter\cbRGB\expandafter{\detokenize{236,255,236}}{and\strut} \setlength{\fboxsep}{0pt}\def\cbRGB{\colorbox[RGB]}\expandafter\cbRGB\expandafter{\detokenize{230,255,230}}{morbidity\strut} \setlength{\fboxsep}{0pt}\def\cbRGB{\colorbox[RGB]}\expandafter\cbRGB\expandafter{\detokenize{231,255,231}}{has\strut} \setlength{\fboxsep}{0pt}\def\cbRGB{\colorbox[RGB]}\expandafter\cbRGB\expandafter{\detokenize{222,255,222}}{not\strut} \setlength{\fboxsep}{0pt}\def\cbRGB{\colorbox[RGB]}\expandafter\cbRGB\expandafter{\detokenize{220,255,220}}{been\strut} \setlength{\fboxsep}{0pt}\def\cbRGB{\colorbox[RGB]}\expandafter\cbRGB\expandafter{\detokenize{224,255,224}}{performed\strut} \setlength{\fboxsep}{0pt}\def\cbRGB{\colorbox[RGB]}\expandafter\cbRGB\expandafter{\detokenize{229,255,229}}{recently\strut} \setlength{\fboxsep}{0pt}\def\cbRGB{\colorbox[RGB]}\expandafter\cbRGB\expandafter{\detokenize{226,255,226}}{.\strut} \setlength{\fboxsep}{0pt}\def\cbRGB{\colorbox[RGB]}\expandafter\cbRGB\expandafter{\detokenize{213,255,213}}{data\strut} \setlength{\fboxsep}{0pt}\def\cbRGB{\colorbox[RGB]}\expandafter\cbRGB\expandafter{\detokenize{216,255,216}}{unk\strut} \setlength{\fboxsep}{0pt}\def\cbRGB{\colorbox[RGB]}\expandafter\cbRGB\expandafter{\detokenize{212,255,212}}{:\strut} \setlength{\fboxsep}{0pt}\def\cbRGB{\colorbox[RGB]}\expandafter\cbRGB\expandafter{\detokenize{225,255,225}}{the\strut} \setlength{\fboxsep}{0pt}\def\cbRGB{\colorbox[RGB]}\expandafter\cbRGB\expandafter{\detokenize{221,255,221}}{unk\strut} \setlength{\fboxsep}{0pt}\def\cbRGB{\colorbox[RGB]}\expandafter\cbRGB\expandafter{\detokenize{225,255,225}}{,\strut} \setlength{\fboxsep}{0pt}\def\cbRGB{\colorbox[RGB]}\expandafter\cbRGB\expandafter{\detokenize{224,255,224}}{unk\strut} \setlength{\fboxsep}{0pt}\def\cbRGB{\colorbox[RGB]}\expandafter\cbRGB\expandafter{\detokenize{222,255,222}}{,\strut} \setlength{\fboxsep}{0pt}\def\cbRGB{\colorbox[RGB]}\expandafter\cbRGB\expandafter{\detokenize{222,255,222}}{and\strut} \setlength{\fboxsep}{0pt}\def\cbRGB{\colorbox[RGB]}\expandafter\cbRGB\expandafter{\detokenize{221,255,221}}{unk\strut} \setlength{\fboxsep}{0pt}\def\cbRGB{\colorbox[RGB]}\expandafter\cbRGB\expandafter{\detokenize{222,255,222}}{of\strut} \setlength{\fboxsep}{0pt}\def\cbRGB{\colorbox[RGB]}\expandafter\cbRGB\expandafter{\detokenize{221,255,221}}{unk\strut} \setlength{\fboxsep}{0pt}\def\cbRGB{\colorbox[RGB]}\expandafter\cbRGB\expandafter{\detokenize{222,255,222}}{electronic\strut} \setlength{\fboxsep}{0pt}\def\cbRGB{\colorbox[RGB]}\expandafter\cbRGB\expandafter{\detokenize{220,255,220}}{unk\strut} \setlength{\fboxsep}{0pt}\def\cbRGB{\colorbox[RGB]}\expandafter\cbRGB\expandafter{\detokenize{222,255,222}}{were\strut} \setlength{\fboxsep}{0pt}\def\cbRGB{\colorbox[RGB]}\expandafter\cbRGB\expandafter{\detokenize{219,255,219}}{unk\strut} \setlength{\fboxsep}{0pt}\def\cbRGB{\colorbox[RGB]}\expandafter\cbRGB\expandafter{\detokenize{219,255,219}}{from\strut} \setlength{\fboxsep}{0pt}\def\cbRGB{\colorbox[RGB]}\expandafter\cbRGB\expandafter{\detokenize{220,255,220}}{qqq\strut} \setlength{\fboxsep}{0pt}\def\cbRGB{\colorbox[RGB]}\expandafter\cbRGB\expandafter{\detokenize{220,255,220}}{to\strut} \setlength{\fboxsep}{0pt}\def\cbRGB{\colorbox[RGB]}\expandafter\cbRGB\expandafter{\detokenize{220,255,220}}{july\strut} \setlength{\fboxsep}{0pt}\def\cbRGB{\colorbox[RGB]}\expandafter\cbRGB\expandafter{\detokenize{220,255,220}}{qqq\strut} \setlength{\fboxsep}{0pt}\def\cbRGB{\colorbox[RGB]}\expandafter\cbRGB\expandafter{\detokenize{220,255,220}}{unk\strut} \setlength{\fboxsep}{0pt}\def\cbRGB{\colorbox[RGB]}\expandafter\cbRGB\expandafter{\detokenize{220,255,220}}{were\strut} \setlength{\fboxsep}{0pt}\def\cbRGB{\colorbox[RGB]}\expandafter\cbRGB\expandafter{\detokenize{220,255,220}}{also\strut} \setlength{\fboxsep}{0pt}\def\cbRGB{\colorbox[RGB]}\expandafter\cbRGB\expandafter{\detokenize{221,255,221}}{identified\strut} \setlength{\fboxsep}{0pt}\def\cbRGB{\colorbox[RGB]}\expandafter\cbRGB\expandafter{\detokenize{221,255,221}}{from\strut} \setlength{\fboxsep}{0pt}\def\cbRGB{\colorbox[RGB]}\expandafter\cbRGB\expandafter{\detokenize{223,255,223}}{unk\strut} \setlength{\fboxsep}{0pt}\def\cbRGB{\colorbox[RGB]}\expandafter\cbRGB\expandafter{\detokenize{223,255,223}}{of\strut} \setlength{\fboxsep}{0pt}\def\cbRGB{\colorbox[RGB]}\expandafter\cbRGB\expandafter{\detokenize{223,255,223}}{unk\strut} \setlength{\fboxsep}{0pt}\def\cbRGB{\colorbox[RGB]}\expandafter\cbRGB\expandafter{\detokenize{221,255,221}}{unk\strut} \setlength{\fboxsep}{0pt}\def\cbRGB{\colorbox[RGB]}\expandafter\cbRGB\expandafter{\detokenize{219,255,219}}{.\strut} \setlength{\fboxsep}{0pt}\def\cbRGB{\colorbox[RGB]}\expandafter\cbRGB\expandafter{\detokenize{209,255,209}}{study\strut} \setlength{\fboxsep}{0pt}\def\cbRGB{\colorbox[RGB]}\expandafter\cbRGB\expandafter{\detokenize{214,255,214}}{selection\strut} \setlength{\fboxsep}{0pt}\def\cbRGB{\colorbox[RGB]}\expandafter\cbRGB\expandafter{\detokenize{208,255,208}}{:\strut} \setlength{\fboxsep}{0pt}\def\cbRGB{\colorbox[RGB]}\expandafter\cbRGB\expandafter{\detokenize{213,255,213}}{all\strut} \setlength{\fboxsep}{0pt}\def\cbRGB{\colorbox[RGB]}\expandafter\cbRGB\expandafter{\detokenize{204,255,204}}{randomized\strut} \setlength{\fboxsep}{0pt}\def\cbRGB{\colorbox[RGB]}\expandafter\cbRGB\expandafter{\detokenize{200,255,200}}{clinical\strut} \setlength{\fboxsep}{0pt}\def\cbRGB{\colorbox[RGB]}\expandafter\cbRGB\expandafter{\detokenize{188,255,188}}{trials\strut} \setlength{\fboxsep}{0pt}\def\cbRGB{\colorbox[RGB]}\expandafter\cbRGB\expandafter{\detokenize{166,255,166}}{of\strut} \setlength{\fboxsep}{0pt}\def\cbRGB{\colorbox[RGB]}\expandafter\cbRGB\expandafter{\detokenize{136,255,136}}{beta-blockers\strut} \setlength{\fboxsep}{0pt}\def\cbRGB{\colorbox[RGB]}\expandafter\cbRGB\expandafter{\detokenize{143,255,143}}{versus\strut} \setlength{\fboxsep}{0pt}\def\cbRGB{\colorbox[RGB]}\expandafter\cbRGB\expandafter{\detokenize{164,255,164}}{placebo\strut} \setlength{\fboxsep}{0pt}\def\cbRGB{\colorbox[RGB]}\expandafter\cbRGB\expandafter{\detokenize{208,255,208}}{in\strut} \setlength{\fboxsep}{0pt}\def\cbRGB{\colorbox[RGB]}\expandafter\cbRGB\expandafter{\detokenize{225,255,225}}{chronic\strut} \setlength{\fboxsep}{0pt}\def\cbRGB{\colorbox[RGB]}\expandafter\cbRGB\expandafter{\detokenize{238,255,238}}{stable\strut} \setlength{\fboxsep}{0pt}\def\cbRGB{\colorbox[RGB]}\expandafter\cbRGB\expandafter{\detokenize{237,255,237}}{congestive\strut} \setlength{\fboxsep}{0pt}\def\cbRGB{\colorbox[RGB]}\expandafter\cbRGB\expandafter{\detokenize{237,255,237}}{heart\strut} \setlength{\fboxsep}{0pt}\def\cbRGB{\colorbox[RGB]}\expandafter\cbRGB\expandafter{\detokenize{235,255,235}}{failure\strut} \setlength{\fboxsep}{0pt}\def\cbRGB{\colorbox[RGB]}\expandafter\cbRGB\expandafter{\detokenize{238,255,238}}{were\strut} \setlength{\fboxsep}{0pt}\def\cbRGB{\colorbox[RGB]}\expandafter\cbRGB\expandafter{\detokenize{234,255,234}}{included\strut} \setlength{\fboxsep}{0pt}\def\cbRGB{\colorbox[RGB]}\expandafter\cbRGB\expandafter{\detokenize{222,255,222}}{.\strut} \setlength{\fboxsep}{0pt}\def\cbRGB{\colorbox[RGB]}\expandafter\cbRGB\expandafter{\detokenize{209,255,209}}{data\strut} \setlength{\fboxsep}{0pt}\def\cbRGB{\colorbox[RGB]}\expandafter\cbRGB\expandafter{\detokenize{201,255,201}}{extraction\strut} \setlength{\fboxsep}{0pt}\def\cbRGB{\colorbox[RGB]}\expandafter\cbRGB\expandafter{\detokenize{194,255,194}}{:\strut} \setlength{\fboxsep}{0pt}\def\cbRGB{\colorbox[RGB]}\expandafter\cbRGB\expandafter{\detokenize{204,255,204}}{a\strut} \setlength{\fboxsep}{0pt}\def\cbRGB{\colorbox[RGB]}\expandafter\cbRGB\expandafter{\detokenize{207,255,207}}{specified\strut} \setlength{\fboxsep}{0pt}\def\cbRGB{\colorbox[RGB]}\expandafter\cbRGB\expandafter{\detokenize{223,255,223}}{protocol\strut} \setlength{\fboxsep}{0pt}\def\cbRGB{\colorbox[RGB]}\expandafter\cbRGB\expandafter{\detokenize{193,255,193}}{was\strut} \setlength{\fboxsep}{0pt}\def\cbRGB{\colorbox[RGB]}\expandafter\cbRGB\expandafter{\detokenize{167,255,167}}{followed\strut} \setlength{\fboxsep}{0pt}\def\cbRGB{\colorbox[RGB]}\expandafter\cbRGB\expandafter{\detokenize{169,255,169}}{to\strut} \setlength{\fboxsep}{0pt}\def\cbRGB{\colorbox[RGB]}\expandafter\cbRGB\expandafter{\detokenize{184,255,184}}{extract\strut} \setlength{\fboxsep}{0pt}\def\cbRGB{\colorbox[RGB]}\expandafter\cbRGB\expandafter{\detokenize{207,255,207}}{data\strut} \setlength{\fboxsep}{0pt}\def\cbRGB{\colorbox[RGB]}\expandafter\cbRGB\expandafter{\detokenize{211,255,211}}{on\strut} \setlength{\fboxsep}{0pt}\def\cbRGB{\colorbox[RGB]}\expandafter\cbRGB\expandafter{\detokenize{224,255,224}}{patient\strut} \setlength{\fboxsep}{0pt}\def\cbRGB{\colorbox[RGB]}\expandafter\cbRGB\expandafter{\detokenize{229,255,229}}{characteristics\strut} \setlength{\fboxsep}{0pt}\def\cbRGB{\colorbox[RGB]}\expandafter\cbRGB\expandafter{\detokenize{226,255,226}}{,\strut} \setlength{\fboxsep}{0pt}\def\cbRGB{\colorbox[RGB]}\expandafter\cbRGB\expandafter{\detokenize{226,255,226}}{unk\strut} \setlength{\fboxsep}{0pt}\def\cbRGB{\colorbox[RGB]}\expandafter\cbRGB\expandafter{\detokenize{233,255,233}}{used\strut} \setlength{\fboxsep}{0pt}\def\cbRGB{\colorbox[RGB]}\expandafter\cbRGB\expandafter{\detokenize{239,255,239}}{,\strut} \setlength{\fboxsep}{0pt}\def\cbRGB{\colorbox[RGB]}\expandafter\cbRGB\expandafter{\detokenize{246,255,246}}{overall\strut} \setlength{\fboxsep}{0pt}\def\cbRGB{\colorbox[RGB]}\expandafter\cbRGB\expandafter{\detokenize{239,255,239}}{mortality\strut} \setlength{\fboxsep}{0pt}\def\cbRGB{\colorbox[RGB]}\expandafter\cbRGB\expandafter{\detokenize{243,255,243}}{,\strut} \setlength{\fboxsep}{0pt}\def\cbRGB{\colorbox[RGB]}\expandafter\cbRGB\expandafter{\detokenize{245,255,245}}{hospitalizations\strut} \setlength{\fboxsep}{0pt}\def\cbRGB{\colorbox[RGB]}\expandafter\cbRGB\expandafter{\detokenize{250,255,250}}{for\strut} \setlength{\fboxsep}{0pt}\def\cbRGB{\colorbox[RGB]}\expandafter\cbRGB\expandafter{\detokenize{239,255,239}}{congestive\strut} \setlength{\fboxsep}{0pt}\def\cbRGB{\colorbox[RGB]}\expandafter\cbRGB\expandafter{\detokenize{236,255,236}}{heart\strut} \setlength{\fboxsep}{0pt}\def\cbRGB{\colorbox[RGB]}\expandafter\cbRGB\expandafter{\detokenize{233,255,233}}{failure\strut} \setlength{\fboxsep}{0pt}\def\cbRGB{\colorbox[RGB]}\expandafter\cbRGB\expandafter{\detokenize{233,255,233}}{,\strut} \setlength{\fboxsep}{0pt}\def\cbRGB{\colorbox[RGB]}\expandafter\cbRGB\expandafter{\detokenize{228,255,228}}{and\strut} \setlength{\fboxsep}{0pt}\def\cbRGB{\colorbox[RGB]}\expandafter\cbRGB\expandafter{\detokenize{227,255,227}}{study\strut} \setlength{\fboxsep}{0pt}\def\cbRGB{\colorbox[RGB]}\expandafter\cbRGB\expandafter{\detokenize{230,255,230}}{quality\strut} \setlength{\fboxsep}{0pt}\def\cbRGB{\colorbox[RGB]}\expandafter\cbRGB\expandafter{\detokenize{228,255,228}}{.\strut} \setlength{\fboxsep}{0pt}\def\cbRGB{\colorbox[RGB]}\expandafter\cbRGB\expandafter{\detokenize{215,255,215}}{data\strut} \setlength{\fboxsep}{0pt}\def\cbRGB{\colorbox[RGB]}\expandafter\cbRGB\expandafter{\detokenize{216,255,216}}{unk\strut} \setlength{\fboxsep}{0pt}\def\cbRGB{\colorbox[RGB]}\expandafter\cbRGB\expandafter{\detokenize{212,255,212}}{:\strut} \setlength{\fboxsep}{0pt}\def\cbRGB{\colorbox[RGB]}\expandafter\cbRGB\expandafter{\detokenize{221,255,221}}{a\strut} \setlength{\fboxsep}{0pt}\def\cbRGB{\colorbox[RGB]}\expandafter\cbRGB\expandafter{\detokenize{211,255,211}}{unk\strut} \setlength{\fboxsep}{0pt}\def\cbRGB{\colorbox[RGB]}\expandafter\cbRGB\expandafter{\detokenize{217,255,217}}{unk\strut} \setlength{\fboxsep}{0pt}\def\cbRGB{\colorbox[RGB]}\expandafter\cbRGB\expandafter{\detokenize{211,255,211}}{model\strut} \setlength{\fboxsep}{0pt}\def\cbRGB{\colorbox[RGB]}\expandafter\cbRGB\expandafter{\detokenize{210,255,210}}{was\strut} \setlength{\fboxsep}{0pt}\def\cbRGB{\colorbox[RGB]}\expandafter\cbRGB\expandafter{\detokenize{209,255,209}}{used\strut} \setlength{\fboxsep}{0pt}\def\cbRGB{\colorbox[RGB]}\expandafter\cbRGB\expandafter{\detokenize{217,255,217}}{to\strut} \setlength{\fboxsep}{0pt}\def\cbRGB{\colorbox[RGB]}\expandafter\cbRGB\expandafter{\detokenize{223,255,223}}{unk\strut} \setlength{\fboxsep}{0pt}\def\cbRGB{\colorbox[RGB]}\expandafter\cbRGB\expandafter{\detokenize{223,255,223}}{the\strut} \setlength{\fboxsep}{0pt}\def\cbRGB{\colorbox[RGB]}\expandafter\cbRGB\expandafter{\detokenize{222,255,222}}{results\strut} \setlength{\fboxsep}{0pt}\def\cbRGB{\colorbox[RGB]}\expandafter\cbRGB\expandafter{\detokenize{222,255,222}}{.\strut} \setlength{\fboxsep}{0pt}\def\cbRGB{\colorbox[RGB]}\expandafter\cbRGB\expandafter{\detokenize{223,255,223}}{a\strut} \setlength{\fboxsep}{0pt}\def\cbRGB{\colorbox[RGB]}\expandafter\cbRGB\expandafter{\detokenize{216,255,216}}{total\strut} \setlength{\fboxsep}{0pt}\def\cbRGB{\colorbox[RGB]}\expandafter\cbRGB\expandafter{\detokenize{205,255,205}}{of\strut} \setlength{\fboxsep}{0pt}\def\cbRGB{\colorbox[RGB]}\expandafter\cbRGB\expandafter{\detokenize{203,255,203}}{qqq\strut} \setlength{\fboxsep}{0pt}\def\cbRGB{\colorbox[RGB]}\expandafter\cbRGB\expandafter{\detokenize{206,255,206}}{trials\strut} \setlength{\fboxsep}{0pt}\def\cbRGB{\colorbox[RGB]}\expandafter\cbRGB\expandafter{\detokenize{212,255,212}}{involving\strut} \setlength{\fboxsep}{0pt}\def\cbRGB{\colorbox[RGB]}\expandafter\cbRGB\expandafter{\detokenize{210,255,210}}{qqq\strut} \setlength{\fboxsep}{0pt}\def\cbRGB{\colorbox[RGB]}\expandafter\cbRGB\expandafter{\detokenize{214,255,214}}{qqq\strut} \setlength{\fboxsep}{0pt}\def\cbRGB{\colorbox[RGB]}\expandafter\cbRGB\expandafter{\detokenize{217,255,217}}{patients\strut} \setlength{\fboxsep}{0pt}\def\cbRGB{\colorbox[RGB]}\expandafter\cbRGB\expandafter{\detokenize{220,255,220}}{were\strut} \setlength{\fboxsep}{0pt}\def\cbRGB{\colorbox[RGB]}\expandafter\cbRGB\expandafter{\detokenize{220,255,220}}{identified\strut} \setlength{\fboxsep}{0pt}\def\cbRGB{\colorbox[RGB]}\expandafter\cbRGB\expandafter{\detokenize{221,255,221}}{.\strut} \setlength{\fboxsep}{0pt}\def\cbRGB{\colorbox[RGB]}\expandafter\cbRGB\expandafter{\detokenize{224,255,224}}{there\strut} \setlength{\fboxsep}{0pt}\def\cbRGB{\colorbox[RGB]}\expandafter\cbRGB\expandafter{\detokenize{224,255,224}}{were\strut} \setlength{\fboxsep}{0pt}\def\cbRGB{\colorbox[RGB]}\expandafter\cbRGB\expandafter{\detokenize{224,255,224}}{qqq\strut} \setlength{\fboxsep}{0pt}\def\cbRGB{\colorbox[RGB]}\expandafter\cbRGB\expandafter{\detokenize{222,255,222}}{deaths\strut} \setlength{\fboxsep}{0pt}\def\cbRGB{\colorbox[RGB]}\expandafter\cbRGB\expandafter{\detokenize{222,255,222}}{among\strut} \setlength{\fboxsep}{0pt}\def\cbRGB{\colorbox[RGB]}\expandafter\cbRGB\expandafter{\detokenize{222,255,222}}{qqq\strut} \setlength{\fboxsep}{0pt}\def\cbRGB{\colorbox[RGB]}\expandafter\cbRGB\expandafter{\detokenize{222,255,222}}{patients\strut} \setlength{\fboxsep}{0pt}\def\cbRGB{\colorbox[RGB]}\expandafter\cbRGB\expandafter{\detokenize{212,255,212}}{randomly\strut} \setlength{\fboxsep}{0pt}\def\cbRGB{\colorbox[RGB]}\expandafter\cbRGB\expandafter{\detokenize{202,255,202}}{assigned\strut} \setlength{\fboxsep}{0pt}\def\cbRGB{\colorbox[RGB]}\expandafter\cbRGB\expandafter{\detokenize{198,255,198}}{to\strut} \setlength{\fboxsep}{0pt}\def\cbRGB{\colorbox[RGB]}\expandafter\cbRGB\expandafter{\detokenize{202,255,202}}{placebo\strut} \setlength{\fboxsep}{0pt}\def\cbRGB{\colorbox[RGB]}\expandafter\cbRGB\expandafter{\detokenize{209,255,209}}{and\strut} \setlength{\fboxsep}{0pt}\def\cbRGB{\colorbox[RGB]}\expandafter\cbRGB\expandafter{\detokenize{213,255,213}}{qqq\strut} \setlength{\fboxsep}{0pt}\def\cbRGB{\colorbox[RGB]}\expandafter\cbRGB\expandafter{\detokenize{218,255,218}}{deaths\strut} \setlength{\fboxsep}{0pt}\def\cbRGB{\colorbox[RGB]}\expandafter\cbRGB\expandafter{\detokenize{222,255,222}}{among\strut} \setlength{\fboxsep}{0pt}\def\cbRGB{\colorbox[RGB]}\expandafter\cbRGB\expandafter{\detokenize{220,255,220}}{qqq\strut} \setlength{\fboxsep}{0pt}\def\cbRGB{\colorbox[RGB]}\expandafter\cbRGB\expandafter{\detokenize{219,255,219}}{patients\strut} \setlength{\fboxsep}{0pt}\def\cbRGB{\colorbox[RGB]}\expandafter\cbRGB\expandafter{\detokenize{218,255,218}}{assigned\strut} \setlength{\fboxsep}{0pt}\def\cbRGB{\colorbox[RGB]}\expandafter\cbRGB\expandafter{\detokenize{218,255,218}}{to\strut} \setlength{\fboxsep}{0pt}\def\cbRGB{\colorbox[RGB]}\expandafter\cbRGB\expandafter{\detokenize{218,255,218}}{unk\strut} \setlength{\fboxsep}{0pt}\def\cbRGB{\colorbox[RGB]}\expandafter\cbRGB\expandafter{\detokenize{217,255,217}}{therapy\strut} \setlength{\fboxsep}{0pt}\def\cbRGB{\colorbox[RGB]}\expandafter\cbRGB\expandafter{\detokenize{218,255,218}}{.\strut} \setlength{\fboxsep}{0pt}\def\cbRGB{\colorbox[RGB]}\expandafter\cbRGB\expandafter{\detokenize{219,255,219}}{in\strut} \setlength{\fboxsep}{0pt}\def\cbRGB{\colorbox[RGB]}\expandafter\cbRGB\expandafter{\detokenize{220,255,220}}{these\strut} \setlength{\fboxsep}{0pt}\def\cbRGB{\colorbox[RGB]}\expandafter\cbRGB\expandafter{\detokenize{221,255,221}}{groups\strut} \setlength{\fboxsep}{0pt}\def\cbRGB{\colorbox[RGB]}\expandafter\cbRGB\expandafter{\detokenize{221,255,221}}{,\strut} \setlength{\fboxsep}{0pt}\def\cbRGB{\colorbox[RGB]}\expandafter\cbRGB\expandafter{\detokenize{221,255,221}}{qqq\strut} \setlength{\fboxsep}{0pt}\def\cbRGB{\colorbox[RGB]}\expandafter\cbRGB\expandafter{\detokenize{221,255,221}}{and\strut} \setlength{\fboxsep}{0pt}\def\cbRGB{\colorbox[RGB]}\expandafter\cbRGB\expandafter{\detokenize{221,255,221}}{qqq\strut} \setlength{\fboxsep}{0pt}\def\cbRGB{\colorbox[RGB]}\expandafter\cbRGB\expandafter{\detokenize{221,255,221}}{patients\strut} \setlength{\fboxsep}{0pt}\def\cbRGB{\colorbox[RGB]}\expandafter\cbRGB\expandafter{\detokenize{222,255,222}}{,\strut} \setlength{\fboxsep}{0pt}\def\cbRGB{\colorbox[RGB]}\expandafter\cbRGB\expandafter{\detokenize{224,255,224}}{respectively\strut} \setlength{\fboxsep}{0pt}\def\cbRGB{\colorbox[RGB]}\expandafter\cbRGB\expandafter{\detokenize{223,255,223}}{,\strut} \setlength{\fboxsep}{0pt}\def\cbRGB{\colorbox[RGB]}\expandafter\cbRGB\expandafter{\detokenize{229,255,229}}{required\strut} \setlength{\fboxsep}{0pt}\def\cbRGB{\colorbox[RGB]}\expandafter\cbRGB\expandafter{\detokenize{235,255,235}}{hospitalization\strut} \setlength{\fboxsep}{0pt}\def\cbRGB{\colorbox[RGB]}\expandafter\cbRGB\expandafter{\detokenize{242,255,242}}{for\strut} \setlength{\fboxsep}{0pt}\def\cbRGB{\colorbox[RGB]}\expandafter\cbRGB\expandafter{\detokenize{238,255,238}}{congestive\strut} \setlength{\fboxsep}{0pt}\def\cbRGB{\colorbox[RGB]}\expandafter\cbRGB\expandafter{\detokenize{236,255,236}}{heart\strut} \setlength{\fboxsep}{0pt}\def\cbRGB{\colorbox[RGB]}\expandafter\cbRGB\expandafter{\detokenize{232,255,232}}{failure\strut} \setlength{\fboxsep}{0pt}\def\cbRGB{\colorbox[RGB]}\expandafter\cbRGB\expandafter{\detokenize{233,255,233}}{.\strut} \setlength{\fboxsep}{0pt}\def\cbRGB{\colorbox[RGB]}\expandafter\cbRGB\expandafter{\detokenize{226,255,226}}{the\strut} \setlength{\fboxsep}{0pt}\def\cbRGB{\colorbox[RGB]}\expandafter\cbRGB\expandafter{\detokenize{222,255,222}}{probability\strut} \setlength{\fboxsep}{0pt}\def\cbRGB{\colorbox[RGB]}\expandafter\cbRGB\expandafter{\detokenize{219,255,219}}{that\strut} \setlength{\fboxsep}{0pt}\def\cbRGB{\colorbox[RGB]}\expandafter\cbRGB\expandafter{\detokenize{220,255,220}}{unk\strut} \setlength{\fboxsep}{0pt}\def\cbRGB{\colorbox[RGB]}\expandafter\cbRGB\expandafter{\detokenize{228,255,228}}{therapy\strut} \setlength{\fboxsep}{0pt}\def\cbRGB{\colorbox[RGB]}\expandafter\cbRGB\expandafter{\detokenize{235,255,235}}{reduced\strut} \setlength{\fboxsep}{0pt}\def\cbRGB{\colorbox[RGB]}\expandafter\cbRGB\expandafter{\detokenize{244,255,244}}{total\strut} \setlength{\fboxsep}{0pt}\def\cbRGB{\colorbox[RGB]}\expandafter\cbRGB\expandafter{\detokenize{238,255,238}}{mortality\strut} \setlength{\fboxsep}{0pt}\def\cbRGB{\colorbox[RGB]}\expandafter\cbRGB\expandafter{\detokenize{244,255,244}}{and\strut} \setlength{\fboxsep}{0pt}\def\cbRGB{\colorbox[RGB]}\expandafter\cbRGB\expandafter{\detokenize{245,255,245}}{hospitalizations\strut} \setlength{\fboxsep}{0pt}\def\cbRGB{\colorbox[RGB]}\expandafter\cbRGB\expandafter{\detokenize{250,255,250}}{for\strut} \setlength{\fboxsep}{0pt}\def\cbRGB{\colorbox[RGB]}\expandafter\cbRGB\expandafter{\detokenize{238,255,238}}{congestive\strut} \setlength{\fboxsep}{0pt}\def\cbRGB{\colorbox[RGB]}\expandafter\cbRGB\expandafter{\detokenize{236,255,236}}{heart\strut} \setlength{\fboxsep}{0pt}\def\cbRGB{\colorbox[RGB]}\expandafter\cbRGB\expandafter{\detokenize{233,255,233}}{failure\strut} \setlength{\fboxsep}{0pt}\def\cbRGB{\colorbox[RGB]}\expandafter\cbRGB\expandafter{\detokenize{234,255,234}}{was\strut} \setlength{\fboxsep}{0pt}\def\cbRGB{\colorbox[RGB]}\expandafter\cbRGB\expandafter{\detokenize{226,255,226}}{almost\strut} \setlength{\fboxsep}{0pt}\def\cbRGB{\colorbox[RGB]}\expandafter\cbRGB\expandafter{\detokenize{222,255,222}}{qqq\strut} \setlength{\fboxsep}{0pt}\def\cbRGB{\colorbox[RGB]}\expandafter\cbRGB\expandafter{\detokenize{221,255,221}}{\%\strut} \setlength{\fboxsep}{0pt}\def\cbRGB{\colorbox[RGB]}\expandafter\cbRGB\expandafter{\detokenize{222,255,222}}{.\strut} \setlength{\fboxsep}{0pt}\def\cbRGB{\colorbox[RGB]}\expandafter\cbRGB\expandafter{\detokenize{223,255,223}}{the\strut} \setlength{\fboxsep}{0pt}\def\cbRGB{\colorbox[RGB]}\expandafter\cbRGB\expandafter{\detokenize{221,255,221}}{best\strut} \setlength{\fboxsep}{0pt}\def\cbRGB{\colorbox[RGB]}\expandafter\cbRGB\expandafter{\detokenize{223,255,223}}{estimates\strut} \setlength{\fboxsep}{0pt}\def\cbRGB{\colorbox[RGB]}\expandafter\cbRGB\expandafter{\detokenize{225,255,225}}{of\strut} \setlength{\fboxsep}{0pt}\def\cbRGB{\colorbox[RGB]}\expandafter\cbRGB\expandafter{\detokenize{226,255,226}}{these\strut} \setlength{\fboxsep}{0pt}\def\cbRGB{\colorbox[RGB]}\expandafter\cbRGB\expandafter{\detokenize{222,255,222}}{advantages\strut} \setlength{\fboxsep}{0pt}\def\cbRGB{\colorbox[RGB]}\expandafter\cbRGB\expandafter{\detokenize{221,255,221}}{are\strut} \setlength{\fboxsep}{0pt}\def\cbRGB{\colorbox[RGB]}\expandafter\cbRGB\expandafter{\detokenize{221,255,221}}{qqq\strut} \setlength{\fboxsep}{0pt}\def\cbRGB{\colorbox[RGB]}\expandafter\cbRGB\expandafter{\detokenize{221,255,221}}{unk\strut} \setlength{\fboxsep}{0pt}\def\cbRGB{\colorbox[RGB]}\expandafter\cbRGB\expandafter{\detokenize{219,255,219}}{unk\strut} \setlength{\fboxsep}{0pt}\def\cbRGB{\colorbox[RGB]}\expandafter\cbRGB\expandafter{\detokenize{221,255,221}}{and\strut} \setlength{\fboxsep}{0pt}\def\cbRGB{\colorbox[RGB]}\expandafter\cbRGB\expandafter{\detokenize{221,255,221}}{qqq\strut} \setlength{\fboxsep}{0pt}\def\cbRGB{\colorbox[RGB]}\expandafter\cbRGB\expandafter{\detokenize{223,255,223}}{fewer\strut} \setlength{\fboxsep}{0pt}\def\cbRGB{\colorbox[RGB]}\expandafter\cbRGB\expandafter{\detokenize{221,255,221}}{hospitalizations\strut} \setlength{\fboxsep}{0pt}\def\cbRGB{\colorbox[RGB]}\expandafter\cbRGB\expandafter{\detokenize{221,255,221}}{per\strut} \setlength{\fboxsep}{0pt}\def\cbRGB{\colorbox[RGB]}\expandafter\cbRGB\expandafter{\detokenize{221,255,221}}{qqq\strut} \setlength{\fboxsep}{0pt}\def\cbRGB{\colorbox[RGB]}\expandafter\cbRGB\expandafter{\detokenize{221,255,221}}{patients\strut} \setlength{\fboxsep}{0pt}\def\cbRGB{\colorbox[RGB]}\expandafter\cbRGB\expandafter{\detokenize{222,255,222}}{treated\strut} \setlength{\fboxsep}{0pt}\def\cbRGB{\colorbox[RGB]}\expandafter\cbRGB\expandafter{\detokenize{222,255,222}}{in\strut} \setlength{\fboxsep}{0pt}\def\cbRGB{\colorbox[RGB]}\expandafter\cbRGB\expandafter{\detokenize{223,255,223}}{the\strut} \setlength{\fboxsep}{0pt}\def\cbRGB{\colorbox[RGB]}\expandafter\cbRGB\expandafter{\detokenize{220,255,220}}{first\strut} \setlength{\fboxsep}{0pt}\def\cbRGB{\colorbox[RGB]}\expandafter\cbRGB\expandafter{\detokenize{219,255,219}}{year\strut} \setlength{\fboxsep}{0pt}\def\cbRGB{\colorbox[RGB]}\expandafter\cbRGB\expandafter{\detokenize{218,255,218}}{after\strut} \setlength{\fboxsep}{0pt}\def\cbRGB{\colorbox[RGB]}\expandafter\cbRGB\expandafter{\detokenize{219,255,219}}{therapy\strut} \setlength{\fboxsep}{0pt}\def\cbRGB{\colorbox[RGB]}\expandafter\cbRGB\expandafter{\detokenize{218,255,218}}{.\strut} \setlength{\fboxsep}{0pt}\def\cbRGB{\colorbox[RGB]}\expandafter\cbRGB\expandafter{\detokenize{217,255,217}}{the\strut} \setlength{\fboxsep}{0pt}\def\cbRGB{\colorbox[RGB]}\expandafter\cbRGB\expandafter{\detokenize{218,255,218}}{probability\strut} \setlength{\fboxsep}{0pt}\def\cbRGB{\colorbox[RGB]}\expandafter\cbRGB\expandafter{\detokenize{219,255,219}}{that\strut} \setlength{\fboxsep}{0pt}\def\cbRGB{\colorbox[RGB]}\expandafter\cbRGB\expandafter{\detokenize{216,255,216}}{these\strut} \setlength{\fboxsep}{0pt}\def\cbRGB{\colorbox[RGB]}\expandafter\cbRGB\expandafter{\detokenize{210,255,210}}{benefits\strut} \setlength{\fboxsep}{0pt}\def\cbRGB{\colorbox[RGB]}\expandafter\cbRGB\expandafter{\detokenize{210,255,210}}{are\strut} \setlength{\fboxsep}{0pt}\def\cbRGB{\colorbox[RGB]}\expandafter\cbRGB\expandafter{\detokenize{213,255,213}}{clinically\strut} \setlength{\fboxsep}{0pt}\def\cbRGB{\colorbox[RGB]}\expandafter\cbRGB\expandafter{\detokenize{218,255,218}}{significant\strut} \setlength{\fboxsep}{0pt}\def\cbRGB{\colorbox[RGB]}\expandafter\cbRGB\expandafter{\detokenize{219,255,219}}{(\strut} \setlength{\fboxsep}{0pt}\def\cbRGB{\colorbox[RGB]}\expandafter\cbRGB\expandafter{\detokenize{222,255,222}}{>\strut} \setlength{\fboxsep}{0pt}\def\cbRGB{\colorbox[RGB]}\expandafter\cbRGB\expandafter{\detokenize{224,255,224}}{qqq\strut} \setlength{\fboxsep}{0pt}\def\cbRGB{\colorbox[RGB]}\expandafter\cbRGB\expandafter{\detokenize{225,255,225}}{unk\strut} \setlength{\fboxsep}{0pt}\def\cbRGB{\colorbox[RGB]}\expandafter\cbRGB\expandafter{\detokenize{225,255,225}}{unk\strut} \setlength{\fboxsep}{0pt}\def\cbRGB{\colorbox[RGB]}\expandafter\cbRGB\expandafter{\detokenize{222,255,222}}{or\strut} \setlength{\fboxsep}{0pt}\def\cbRGB{\colorbox[RGB]}\expandafter\cbRGB\expandafter{\detokenize{222,255,222}}{>\strut} \setlength{\fboxsep}{0pt}\def\cbRGB{\colorbox[RGB]}\expandafter\cbRGB\expandafter{\detokenize{222,255,222}}{qqq\strut} \setlength{\fboxsep}{0pt}\def\cbRGB{\colorbox[RGB]}\expandafter\cbRGB\expandafter{\detokenize{224,255,224}}{fewer\strut} \setlength{\fboxsep}{0pt}\def\cbRGB{\colorbox[RGB]}\expandafter\cbRGB\expandafter{\detokenize{222,255,222}}{hospitalizations\strut} \setlength{\fboxsep}{0pt}\def\cbRGB{\colorbox[RGB]}\expandafter\cbRGB\expandafter{\detokenize{221,255,221}}{per\strut} \setlength{\fboxsep}{0pt}\def\cbRGB{\colorbox[RGB]}\expandafter\cbRGB\expandafter{\detokenize{221,255,221}}{qqq\strut} \setlength{\fboxsep}{0pt}\def\cbRGB{\colorbox[RGB]}\expandafter\cbRGB\expandafter{\detokenize{220,255,220}}{patients\strut} \setlength{\fboxsep}{0pt}\def\cbRGB{\colorbox[RGB]}\expandafter\cbRGB\expandafter{\detokenize{221,255,221}}{treated\strut} \setlength{\fboxsep}{0pt}\def\cbRGB{\colorbox[RGB]}\expandafter\cbRGB\expandafter{\detokenize{220,255,220}}{)\strut} \setlength{\fboxsep}{0pt}\def\cbRGB{\colorbox[RGB]}\expandafter\cbRGB\expandafter{\detokenize{221,255,221}}{is\strut} \setlength{\fboxsep}{0pt}\def\cbRGB{\colorbox[RGB]}\expandafter\cbRGB\expandafter{\detokenize{219,255,219}}{qqq\strut} \setlength{\fboxsep}{0pt}\def\cbRGB{\colorbox[RGB]}\expandafter\cbRGB\expandafter{\detokenize{220,255,220}}{\%\strut} \setlength{\fboxsep}{0pt}\def\cbRGB{\colorbox[RGB]}\expandafter\cbRGB\expandafter{\detokenize{219,255,219}}{.\strut} \setlength{\fboxsep}{0pt}\def\cbRGB{\colorbox[RGB]}\expandafter\cbRGB\expandafter{\detokenize{221,255,221}}{both\strut} \setlength{\fboxsep}{0pt}\def\cbRGB{\colorbox[RGB]}\expandafter\cbRGB\expandafter{\detokenize{222,255,222}}{selective\strut} \setlength{\fboxsep}{0pt}\def\cbRGB{\colorbox[RGB]}\expandafter\cbRGB\expandafter{\detokenize{220,255,220}}{and\strut} \setlength{\fboxsep}{0pt}\def\cbRGB{\colorbox[RGB]}\expandafter\cbRGB\expandafter{\detokenize{217,255,217}}{unk\strut} \setlength{\fboxsep}{0pt}\def\cbRGB{\colorbox[RGB]}\expandafter\cbRGB\expandafter{\detokenize{212,255,212}}{agents\strut} \setlength{\fboxsep}{0pt}\def\cbRGB{\colorbox[RGB]}\expandafter\cbRGB\expandafter{\detokenize{213,255,213}}{produced\strut} \setlength{\fboxsep}{0pt}\def\cbRGB{\colorbox[RGB]}\expandafter\cbRGB\expandafter{\detokenize{212,255,212}}{these\strut} \setlength{\fboxsep}{0pt}\def\cbRGB{\colorbox[RGB]}\expandafter\cbRGB\expandafter{\detokenize{217,255,217}}{unk\strut} \setlength{\fboxsep}{0pt}\def\cbRGB{\colorbox[RGB]}\expandafter\cbRGB\expandafter{\detokenize{215,255,215}}{effects\strut} \setlength{\fboxsep}{0pt}\def\cbRGB{\colorbox[RGB]}\expandafter\cbRGB\expandafter{\detokenize{219,255,219}}{.\strut} \setlength{\fboxsep}{0pt}\def\cbRGB{\colorbox[RGB]}\expandafter\cbRGB\expandafter{\detokenize{219,255,219}}{the\strut} \setlength{\fboxsep}{0pt}\def\cbRGB{\colorbox[RGB]}\expandafter\cbRGB\expandafter{\detokenize{222,255,222}}{results\strut} \setlength{\fboxsep}{0pt}\def\cbRGB{\colorbox[RGB]}\expandafter\cbRGB\expandafter{\detokenize{224,255,224}}{are\strut} \setlength{\fboxsep}{0pt}\def\cbRGB{\colorbox[RGB]}\expandafter\cbRGB\expandafter{\detokenize{222,255,222}}{unk\strut} \setlength{\fboxsep}{0pt}\def\cbRGB{\colorbox[RGB]}\expandafter\cbRGB\expandafter{\detokenize{226,255,226}}{to\strut} \setlength{\fboxsep}{0pt}\def\cbRGB{\colorbox[RGB]}\expandafter\cbRGB\expandafter{\detokenize{143,255,143}}{any\strut} \setlength{\fboxsep}{0pt}\def\cbRGB{\colorbox[RGB]}\expandafter\cbRGB\expandafter{\detokenize{146,255,146}}{unk\strut} \setlength{\fboxsep}{0pt}\def\cbRGB{\colorbox[RGB]}\expandafter\cbRGB\expandafter{\detokenize{0,255,0}}{publication\strut} \setlength{\fboxsep}{0pt}\def\cbRGB{\colorbox[RGB]}\expandafter\cbRGB\expandafter{\detokenize{60,255,60}}{unk\strut} \setlength{\fboxsep}{0pt}\def\cbRGB{\colorbox[RGB]}\expandafter\cbRGB\expandafter{\detokenize{50,255,50}}{.\strut} \setlength{\fboxsep}{0pt}\def\cbRGB{\colorbox[RGB]}\expandafter\cbRGB\expandafter{\detokenize{197,255,197}}{conclusions\strut} \setlength{\fboxsep}{0pt}\def\cbRGB{\colorbox[RGB]}\expandafter\cbRGB\expandafter{\detokenize{211,255,211}}{:\strut} \setlength{\fboxsep}{0pt}\def\cbRGB{\colorbox[RGB]}\expandafter\cbRGB\expandafter{\detokenize{217,255,217}}{unk\strut} \setlength{\fboxsep}{0pt}\def\cbRGB{\colorbox[RGB]}\expandafter\cbRGB\expandafter{\detokenize{209,255,209}}{therapy\strut} \setlength{\fboxsep}{0pt}\def\cbRGB{\colorbox[RGB]}\expandafter\cbRGB\expandafter{\detokenize{211,255,211}}{is\strut} \setlength{\fboxsep}{0pt}\def\cbRGB{\colorbox[RGB]}\expandafter\cbRGB\expandafter{\detokenize{196,255,196}}{associated\strut} \setlength{\fboxsep}{0pt}\def\cbRGB{\colorbox[RGB]}\expandafter\cbRGB\expandafter{\detokenize{188,255,188}}{with\strut} \setlength{\fboxsep}{0pt}\def\cbRGB{\colorbox[RGB]}\expandafter\cbRGB\expandafter{\detokenize{189,255,189}}{clinically\strut} \setlength{\fboxsep}{0pt}\def\cbRGB{\colorbox[RGB]}\expandafter\cbRGB\expandafter{\detokenize{211,255,211}}{meaningful\strut} \setlength{\fboxsep}{0pt}\def\cbRGB{\colorbox[RGB]}\expandafter\cbRGB\expandafter{\detokenize{226,255,226}}{reductions\strut} \setlength{\fboxsep}{0pt}\def\cbRGB{\colorbox[RGB]}\expandafter\cbRGB\expandafter{\detokenize{239,255,239}}{in\strut} \setlength{\fboxsep}{0pt}\def\cbRGB{\colorbox[RGB]}\expandafter\cbRGB\expandafter{\detokenize{234,255,234}}{mortality\strut} \setlength{\fboxsep}{0pt}\def\cbRGB{\colorbox[RGB]}\expandafter\cbRGB\expandafter{\detokenize{239,255,239}}{and\strut} \setlength{\fboxsep}{0pt}\def\cbRGB{\colorbox[RGB]}\expandafter\cbRGB\expandafter{\detokenize{234,255,234}}{morbidity\strut} \setlength{\fboxsep}{0pt}\def\cbRGB{\colorbox[RGB]}\expandafter\cbRGB\expandafter{\detokenize{235,255,235}}{in\strut} \setlength{\fboxsep}{0pt}\def\cbRGB{\colorbox[RGB]}\expandafter\cbRGB\expandafter{\detokenize{232,255,232}}{patients\strut} \setlength{\fboxsep}{0pt}\def\cbRGB{\colorbox[RGB]}\expandafter\cbRGB\expandafter{\detokenize{237,255,237}}{with\strut} \setlength{\fboxsep}{0pt}\def\cbRGB{\colorbox[RGB]}\expandafter\cbRGB\expandafter{\detokenize{242,255,242}}{stable\strut} \setlength{\fboxsep}{0pt}\def\cbRGB{\colorbox[RGB]}\expandafter\cbRGB\expandafter{\detokenize{238,255,238}}{congestive\strut} \setlength{\fboxsep}{0pt}\def\cbRGB{\colorbox[RGB]}\expandafter\cbRGB\expandafter{\detokenize{237,255,237}}{heart\strut} \setlength{\fboxsep}{0pt}\def\cbRGB{\colorbox[RGB]}\expandafter\cbRGB\expandafter{\detokenize{234,255,234}}{failure\strut} \setlength{\fboxsep}{0pt}\def\cbRGB{\colorbox[RGB]}\expandafter\cbRGB\expandafter{\detokenize{232,255,232}}{and\strut} \setlength{\fboxsep}{0pt}\def\cbRGB{\colorbox[RGB]}\expandafter\cbRGB\expandafter{\detokenize{226,255,226}}{should\strut} \setlength{\fboxsep}{0pt}\def\cbRGB{\colorbox[RGB]}\expandafter\cbRGB\expandafter{\detokenize{220,255,220}}{be\strut} \setlength{\fboxsep}{0pt}\def\cbRGB{\colorbox[RGB]}\expandafter\cbRGB\expandafter{\detokenize{220,255,220}}{routinely\strut} \setlength{\fboxsep}{0pt}\def\cbRGB{\colorbox[RGB]}\expandafter\cbRGB\expandafter{\detokenize{216,255,216}}{offered\strut} \setlength{\fboxsep}{0pt}\def\cbRGB{\colorbox[RGB]}\expandafter\cbRGB\expandafter{\detokenize{218,255,218}}{to\strut} \setlength{\fboxsep}{0pt}\def\cbRGB{\colorbox[RGB]}\expandafter\cbRGB\expandafter{\detokenize{219,255,219}}{all\strut} \setlength{\fboxsep}{0pt}\def\cbRGB{\colorbox[RGB]}\expandafter\cbRGB\expandafter{\detokenize{219,255,219}}{patients\strut} \setlength{\fboxsep}{0pt}\def\cbRGB{\colorbox[RGB]}\expandafter\cbRGB\expandafter{\detokenize{220,255,220}}{similar\strut} \setlength{\fboxsep}{0pt}\def\cbRGB{\colorbox[RGB]}\expandafter\cbRGB\expandafter{\detokenize{220,255,220}}{to\strut} \setlength{\fboxsep}{0pt}\def\cbRGB{\colorbox[RGB]}\expandafter\cbRGB\expandafter{\detokenize{213,255,213}}{those\strut} \setlength{\fboxsep}{0pt}\def\cbRGB{\colorbox[RGB]}\expandafter\cbRGB\expandafter{\detokenize{206,255,206}}{included\strut} \setlength{\fboxsep}{0pt}\def\cbRGB{\colorbox[RGB]}\expandafter\cbRGB\expandafter{\detokenize{205,255,205}}{in\strut} \setlength{\fboxsep}{0pt}\def\cbRGB{\colorbox[RGB]}\expandafter\cbRGB\expandafter{\detokenize{209,255,209}}{trials\strut} \setlength{\fboxsep}{0pt}\def\cbRGB{\colorbox[RGB]}\expandafter\cbRGB\expandafter{\detokenize{235,255,235}}{.\strut} 

\setlength{\fboxsep}{0pt}\def\cbRGB{\colorbox[RGB]}\expandafter\cbRGB\expandafter{\detokenize{223,223,255}}{purpose\strut} \setlength{\fboxsep}{0pt}\def\cbRGB{\colorbox[RGB]}\expandafter\cbRGB\expandafter{\detokenize{178,178,255}}{:\strut} \setlength{\fboxsep}{0pt}\def\cbRGB{\colorbox[RGB]}\expandafter\cbRGB\expandafter{\detokenize{177,177,255}}{congestive\strut} \setlength{\fboxsep}{0pt}\def\cbRGB{\colorbox[RGB]}\expandafter\cbRGB\expandafter{\detokenize{166,166,255}}{heart\strut} \setlength{\fboxsep}{0pt}\def\cbRGB{\colorbox[RGB]}\expandafter\cbRGB\expandafter{\detokenize{173,173,255}}{failure\strut} \setlength{\fboxsep}{0pt}\def\cbRGB{\colorbox[RGB]}\expandafter\cbRGB\expandafter{\detokenize{182,182,255}}{is\strut} \setlength{\fboxsep}{0pt}\def\cbRGB{\colorbox[RGB]}\expandafter\cbRGB\expandafter{\detokenize{184,184,255}}{an\strut} \setlength{\fboxsep}{0pt}\def\cbRGB{\colorbox[RGB]}\expandafter\cbRGB\expandafter{\detokenize{199,199,255}}{important\strut} \setlength{\fboxsep}{0pt}\def\cbRGB{\colorbox[RGB]}\expandafter\cbRGB\expandafter{\detokenize{194,194,255}}{cause\strut} \setlength{\fboxsep}{0pt}\def\cbRGB{\colorbox[RGB]}\expandafter\cbRGB\expandafter{\detokenize{185,185,255}}{of\strut} \setlength{\fboxsep}{0pt}\def\cbRGB{\colorbox[RGB]}\expandafter\cbRGB\expandafter{\detokenize{154,154,255}}{patient\strut} \setlength{\fboxsep}{0pt}\def\cbRGB{\colorbox[RGB]}\expandafter\cbRGB\expandafter{\detokenize{92,92,255}}{morbidity\strut} \setlength{\fboxsep}{0pt}\def\cbRGB{\colorbox[RGB]}\expandafter\cbRGB\expandafter{\detokenize{91,91,255}}{and\strut} \setlength{\fboxsep}{0pt}\def\cbRGB{\colorbox[RGB]}\expandafter\cbRGB\expandafter{\detokenize{93,93,255}}{mortality\strut} \setlength{\fboxsep}{0pt}\def\cbRGB{\colorbox[RGB]}\expandafter\cbRGB\expandafter{\detokenize{145,145,255}}{.\strut} \setlength{\fboxsep}{0pt}\def\cbRGB{\colorbox[RGB]}\expandafter\cbRGB\expandafter{\detokenize{152,152,255}}{although\strut} \setlength{\fboxsep}{0pt}\def\cbRGB{\colorbox[RGB]}\expandafter\cbRGB\expandafter{\detokenize{176,176,255}}{several\strut} \setlength{\fboxsep}{0pt}\def\cbRGB{\colorbox[RGB]}\expandafter\cbRGB\expandafter{\detokenize{197,197,255}}{randomized\strut} \setlength{\fboxsep}{0pt}\def\cbRGB{\colorbox[RGB]}\expandafter\cbRGB\expandafter{\detokenize{197,197,255}}{clinical\strut} \setlength{\fboxsep}{0pt}\def\cbRGB{\colorbox[RGB]}\expandafter\cbRGB\expandafter{\detokenize{200,200,255}}{trials\strut} \setlength{\fboxsep}{0pt}\def\cbRGB{\colorbox[RGB]}\expandafter\cbRGB\expandafter{\detokenize{211,211,255}}{have\strut} \setlength{\fboxsep}{0pt}\def\cbRGB{\colorbox[RGB]}\expandafter\cbRGB\expandafter{\detokenize{229,229,255}}{compared\strut} \setlength{\fboxsep}{0pt}\def\cbRGB{\colorbox[RGB]}\expandafter\cbRGB\expandafter{\detokenize{239,239,255}}{beta-blockers\strut} \setlength{\fboxsep}{0pt}\def\cbRGB{\colorbox[RGB]}\expandafter\cbRGB\expandafter{\detokenize{240,240,255}}{with\strut} \setlength{\fboxsep}{0pt}\def\cbRGB{\colorbox[RGB]}\expandafter\cbRGB\expandafter{\detokenize{240,240,255}}{placebo\strut} \setlength{\fboxsep}{0pt}\def\cbRGB{\colorbox[RGB]}\expandafter\cbRGB\expandafter{\detokenize{235,235,255}}{for\strut} \setlength{\fboxsep}{0pt}\def\cbRGB{\colorbox[RGB]}\expandafter\cbRGB\expandafter{\detokenize{228,228,255}}{treatment\strut} \setlength{\fboxsep}{0pt}\def\cbRGB{\colorbox[RGB]}\expandafter\cbRGB\expandafter{\detokenize{215,215,255}}{of\strut} \setlength{\fboxsep}{0pt}\def\cbRGB{\colorbox[RGB]}\expandafter\cbRGB\expandafter{\detokenize{199,199,255}}{congestive\strut} \setlength{\fboxsep}{0pt}\def\cbRGB{\colorbox[RGB]}\expandafter\cbRGB\expandafter{\detokenize{197,197,255}}{heart\strut} \setlength{\fboxsep}{0pt}\def\cbRGB{\colorbox[RGB]}\expandafter\cbRGB\expandafter{\detokenize{190,190,255}}{failure\strut} \setlength{\fboxsep}{0pt}\def\cbRGB{\colorbox[RGB]}\expandafter\cbRGB\expandafter{\detokenize{203,203,255}}{,\strut} \setlength{\fboxsep}{0pt}\def\cbRGB{\colorbox[RGB]}\expandafter\cbRGB\expandafter{\detokenize{219,219,255}}{a\strut} \setlength{\fboxsep}{0pt}\def\cbRGB{\colorbox[RGB]}\expandafter\cbRGB\expandafter{\detokenize{229,229,255}}{meta-analysis\strut} \setlength{\fboxsep}{0pt}\def\cbRGB{\colorbox[RGB]}\expandafter\cbRGB\expandafter{\detokenize{224,224,255}}{unk\strut} \setlength{\fboxsep}{0pt}\def\cbRGB{\colorbox[RGB]}\expandafter\cbRGB\expandafter{\detokenize{185,185,255}}{the\strut} \setlength{\fboxsep}{0pt}\def\cbRGB{\colorbox[RGB]}\expandafter\cbRGB\expandafter{\detokenize{130,130,255}}{effect\strut} \setlength{\fboxsep}{0pt}\def\cbRGB{\colorbox[RGB]}\expandafter\cbRGB\expandafter{\detokenize{102,102,255}}{on\strut} \setlength{\fboxsep}{0pt}\def\cbRGB{\colorbox[RGB]}\expandafter\cbRGB\expandafter{\detokenize{40,40,255}}{mortality\strut} \setlength{\fboxsep}{0pt}\def\cbRGB{\colorbox[RGB]}\expandafter\cbRGB\expandafter{\detokenize{63,63,255}}{and\strut} \setlength{\fboxsep}{0pt}\def\cbRGB{\colorbox[RGB]}\expandafter\cbRGB\expandafter{\detokenize{84,84,255}}{morbidity\strut} \setlength{\fboxsep}{0pt}\def\cbRGB{\colorbox[RGB]}\expandafter\cbRGB\expandafter{\detokenize{161,161,255}}{has\strut} \setlength{\fboxsep}{0pt}\def\cbRGB{\colorbox[RGB]}\expandafter\cbRGB\expandafter{\detokenize{197,197,255}}{not\strut} \setlength{\fboxsep}{0pt}\def\cbRGB{\colorbox[RGB]}\expandafter\cbRGB\expandafter{\detokenize{214,214,255}}{been\strut} \setlength{\fboxsep}{0pt}\def\cbRGB{\colorbox[RGB]}\expandafter\cbRGB\expandafter{\detokenize{227,227,255}}{performed\strut} \setlength{\fboxsep}{0pt}\def\cbRGB{\colorbox[RGB]}\expandafter\cbRGB\expandafter{\detokenize{224,224,255}}{recently\strut} \setlength{\fboxsep}{0pt}\def\cbRGB{\colorbox[RGB]}\expandafter\cbRGB\expandafter{\detokenize{205,205,255}}{.\strut} \setlength{\fboxsep}{0pt}\def\cbRGB{\colorbox[RGB]}\expandafter\cbRGB\expandafter{\detokenize{180,180,255}}{data\strut} \setlength{\fboxsep}{0pt}\def\cbRGB{\colorbox[RGB]}\expandafter\cbRGB\expandafter{\detokenize{179,179,255}}{unk\strut} \setlength{\fboxsep}{0pt}\def\cbRGB{\colorbox[RGB]}\expandafter\cbRGB\expandafter{\detokenize{180,180,255}}{:\strut} \setlength{\fboxsep}{0pt}\def\cbRGB{\colorbox[RGB]}\expandafter\cbRGB\expandafter{\detokenize{206,206,255}}{the\strut} \setlength{\fboxsep}{0pt}\def\cbRGB{\colorbox[RGB]}\expandafter\cbRGB\expandafter{\detokenize{210,210,255}}{unk\strut} \setlength{\fboxsep}{0pt}\def\cbRGB{\colorbox[RGB]}\expandafter\cbRGB\expandafter{\detokenize{224,224,255}}{,\strut} \setlength{\fboxsep}{0pt}\def\cbRGB{\colorbox[RGB]}\expandafter\cbRGB\expandafter{\detokenize{223,223,255}}{unk\strut} \setlength{\fboxsep}{0pt}\def\cbRGB{\colorbox[RGB]}\expandafter\cbRGB\expandafter{\detokenize{223,223,255}}{,\strut} \setlength{\fboxsep}{0pt}\def\cbRGB{\colorbox[RGB]}\expandafter\cbRGB\expandafter{\detokenize{222,222,255}}{and\strut} \setlength{\fboxsep}{0pt}\def\cbRGB{\colorbox[RGB]}\expandafter\cbRGB\expandafter{\detokenize{222,222,255}}{unk\strut} \setlength{\fboxsep}{0pt}\def\cbRGB{\colorbox[RGB]}\expandafter\cbRGB\expandafter{\detokenize{223,223,255}}{of\strut} \setlength{\fboxsep}{0pt}\def\cbRGB{\colorbox[RGB]}\expandafter\cbRGB\expandafter{\detokenize{222,222,255}}{unk\strut} \setlength{\fboxsep}{0pt}\def\cbRGB{\colorbox[RGB]}\expandafter\cbRGB\expandafter{\detokenize{222,222,255}}{electronic\strut} \setlength{\fboxsep}{0pt}\def\cbRGB{\colorbox[RGB]}\expandafter\cbRGB\expandafter{\detokenize{222,222,255}}{unk\strut} \setlength{\fboxsep}{0pt}\def\cbRGB{\colorbox[RGB]}\expandafter\cbRGB\expandafter{\detokenize{224,224,255}}{were\strut} \setlength{\fboxsep}{0pt}\def\cbRGB{\colorbox[RGB]}\expandafter\cbRGB\expandafter{\detokenize{227,227,255}}{unk\strut} \setlength{\fboxsep}{0pt}\def\cbRGB{\colorbox[RGB]}\expandafter\cbRGB\expandafter{\detokenize{231,231,255}}{from\strut} \setlength{\fboxsep}{0pt}\def\cbRGB{\colorbox[RGB]}\expandafter\cbRGB\expandafter{\detokenize{238,238,255}}{qqq\strut} \setlength{\fboxsep}{0pt}\def\cbRGB{\colorbox[RGB]}\expandafter\cbRGB\expandafter{\detokenize{246,246,255}}{to\strut} \setlength{\fboxsep}{0pt}\def\cbRGB{\colorbox[RGB]}\expandafter\cbRGB\expandafter{\detokenize{244,244,255}}{july\strut} \setlength{\fboxsep}{0pt}\def\cbRGB{\colorbox[RGB]}\expandafter\cbRGB\expandafter{\detokenize{243,243,255}}{qqq\strut} \setlength{\fboxsep}{0pt}\def\cbRGB{\colorbox[RGB]}\expandafter\cbRGB\expandafter{\detokenize{240,240,255}}{unk\strut} \setlength{\fboxsep}{0pt}\def\cbRGB{\colorbox[RGB]}\expandafter\cbRGB\expandafter{\detokenize{242,242,255}}{were\strut} \setlength{\fboxsep}{0pt}\def\cbRGB{\colorbox[RGB]}\expandafter\cbRGB\expandafter{\detokenize{238,238,255}}{also\strut} \setlength{\fboxsep}{0pt}\def\cbRGB{\colorbox[RGB]}\expandafter\cbRGB\expandafter{\detokenize{234,234,255}}{identified\strut} \setlength{\fboxsep}{0pt}\def\cbRGB{\colorbox[RGB]}\expandafter\cbRGB\expandafter{\detokenize{231,231,255}}{from\strut} \setlength{\fboxsep}{0pt}\def\cbRGB{\colorbox[RGB]}\expandafter\cbRGB\expandafter{\detokenize{228,228,255}}{unk\strut} \setlength{\fboxsep}{0pt}\def\cbRGB{\colorbox[RGB]}\expandafter\cbRGB\expandafter{\detokenize{224,224,255}}{of\strut} \setlength{\fboxsep}{0pt}\def\cbRGB{\colorbox[RGB]}\expandafter\cbRGB\expandafter{\detokenize{224,224,255}}{unk\strut} \setlength{\fboxsep}{0pt}\def\cbRGB{\colorbox[RGB]}\expandafter\cbRGB\expandafter{\detokenize{220,220,255}}{unk\strut} \setlength{\fboxsep}{0pt}\def\cbRGB{\colorbox[RGB]}\expandafter\cbRGB\expandafter{\detokenize{205,205,255}}{.\strut} \setlength{\fboxsep}{0pt}\def\cbRGB{\colorbox[RGB]}\expandafter\cbRGB\expandafter{\detokenize{187,187,255}}{study\strut} \setlength{\fboxsep}{0pt}\def\cbRGB{\colorbox[RGB]}\expandafter\cbRGB\expandafter{\detokenize{185,185,255}}{selection\strut} \setlength{\fboxsep}{0pt}\def\cbRGB{\colorbox[RGB]}\expandafter\cbRGB\expandafter{\detokenize{173,173,255}}{:\strut} \setlength{\fboxsep}{0pt}\def\cbRGB{\colorbox[RGB]}\expandafter\cbRGB\expandafter{\detokenize{190,190,255}}{all\strut} \setlength{\fboxsep}{0pt}\def\cbRGB{\colorbox[RGB]}\expandafter\cbRGB\expandafter{\detokenize{192,192,255}}{randomized\strut} \setlength{\fboxsep}{0pt}\def\cbRGB{\colorbox[RGB]}\expandafter\cbRGB\expandafter{\detokenize{217,217,255}}{clinical\strut} \setlength{\fboxsep}{0pt}\def\cbRGB{\colorbox[RGB]}\expandafter\cbRGB\expandafter{\detokenize{227,227,255}}{trials\strut} \setlength{\fboxsep}{0pt}\def\cbRGB{\colorbox[RGB]}\expandafter\cbRGB\expandafter{\detokenize{239,239,255}}{of\strut} \setlength{\fboxsep}{0pt}\def\cbRGB{\colorbox[RGB]}\expandafter\cbRGB\expandafter{\detokenize{250,250,255}}{beta-blockers\strut} \setlength{\fboxsep}{0pt}\def\cbRGB{\colorbox[RGB]}\expandafter\cbRGB\expandafter{\detokenize{248,248,255}}{versus\strut} \setlength{\fboxsep}{0pt}\def\cbRGB{\colorbox[RGB]}\expandafter\cbRGB\expandafter{\detokenize{249,249,255}}{placebo\strut} \setlength{\fboxsep}{0pt}\def\cbRGB{\colorbox[RGB]}\expandafter\cbRGB\expandafter{\detokenize{249,249,255}}{in\strut} \setlength{\fboxsep}{0pt}\def\cbRGB{\colorbox[RGB]}\expandafter\cbRGB\expandafter{\detokenize{247,247,255}}{chronic\strut} \setlength{\fboxsep}{0pt}\def\cbRGB{\colorbox[RGB]}\expandafter\cbRGB\expandafter{\detokenize{239,239,255}}{stable\strut} \setlength{\fboxsep}{0pt}\def\cbRGB{\colorbox[RGB]}\expandafter\cbRGB\expandafter{\detokenize{228,228,255}}{congestive\strut} \setlength{\fboxsep}{0pt}\def\cbRGB{\colorbox[RGB]}\expandafter\cbRGB\expandafter{\detokenize{225,225,255}}{heart\strut} \setlength{\fboxsep}{0pt}\def\cbRGB{\colorbox[RGB]}\expandafter\cbRGB\expandafter{\detokenize{217,217,255}}{failure\strut} \setlength{\fboxsep}{0pt}\def\cbRGB{\colorbox[RGB]}\expandafter\cbRGB\expandafter{\detokenize{217,217,255}}{were\strut} \setlength{\fboxsep}{0pt}\def\cbRGB{\colorbox[RGB]}\expandafter\cbRGB\expandafter{\detokenize{215,215,255}}{included\strut} \setlength{\fboxsep}{0pt}\def\cbRGB{\colorbox[RGB]}\expandafter\cbRGB\expandafter{\detokenize{212,212,255}}{.\strut} \setlength{\fboxsep}{0pt}\def\cbRGB{\colorbox[RGB]}\expandafter\cbRGB\expandafter{\detokenize{184,184,255}}{data\strut} \setlength{\fboxsep}{0pt}\def\cbRGB{\colorbox[RGB]}\expandafter\cbRGB\expandafter{\detokenize{169,169,255}}{extraction\strut} \setlength{\fboxsep}{0pt}\def\cbRGB{\colorbox[RGB]}\expandafter\cbRGB\expandafter{\detokenize{144,144,255}}{:\strut} \setlength{\fboxsep}{0pt}\def\cbRGB{\colorbox[RGB]}\expandafter\cbRGB\expandafter{\detokenize{176,176,255}}{a\strut} \setlength{\fboxsep}{0pt}\def\cbRGB{\colorbox[RGB]}\expandafter\cbRGB\expandafter{\detokenize{185,185,255}}{specified\strut} \setlength{\fboxsep}{0pt}\def\cbRGB{\colorbox[RGB]}\expandafter\cbRGB\expandafter{\detokenize{222,222,255}}{protocol\strut} \setlength{\fboxsep}{0pt}\def\cbRGB{\colorbox[RGB]}\expandafter\cbRGB\expandafter{\detokenize{228,228,255}}{was\strut} \setlength{\fboxsep}{0pt}\def\cbRGB{\colorbox[RGB]}\expandafter\cbRGB\expandafter{\detokenize{240,240,255}}{followed\strut} \setlength{\fboxsep}{0pt}\def\cbRGB{\colorbox[RGB]}\expandafter\cbRGB\expandafter{\detokenize{238,238,255}}{to\strut} \setlength{\fboxsep}{0pt}\def\cbRGB{\colorbox[RGB]}\expandafter\cbRGB\expandafter{\detokenize{235,235,255}}{extract\strut} \setlength{\fboxsep}{0pt}\def\cbRGB{\colorbox[RGB]}\expandafter\cbRGB\expandafter{\detokenize{228,228,255}}{data\strut} \setlength{\fboxsep}{0pt}\def\cbRGB{\colorbox[RGB]}\expandafter\cbRGB\expandafter{\detokenize{220,220,255}}{on\strut} \setlength{\fboxsep}{0pt}\def\cbRGB{\colorbox[RGB]}\expandafter\cbRGB\expandafter{\detokenize{218,218,255}}{patient\strut} \setlength{\fboxsep}{0pt}\def\cbRGB{\colorbox[RGB]}\expandafter\cbRGB\expandafter{\detokenize{217,217,255}}{characteristics\strut} \setlength{\fboxsep}{0pt}\def\cbRGB{\colorbox[RGB]}\expandafter\cbRGB\expandafter{\detokenize{227,227,255}}{,\strut} \setlength{\fboxsep}{0pt}\def\cbRGB{\colorbox[RGB]}\expandafter\cbRGB\expandafter{\detokenize{215,215,255}}{unk\strut} \setlength{\fboxsep}{0pt}\def\cbRGB{\colorbox[RGB]}\expandafter\cbRGB\expandafter{\detokenize{144,144,255}}{used\strut} \setlength{\fboxsep}{0pt}\def\cbRGB{\colorbox[RGB]}\expandafter\cbRGB\expandafter{\detokenize{83,83,255}}{,\strut} \setlength{\fboxsep}{0pt}\def\cbRGB{\colorbox[RGB]}\expandafter\cbRGB\expandafter{\detokenize{53,53,255}}{overall\strut} \setlength{\fboxsep}{0pt}\def\cbRGB{\colorbox[RGB]}\expandafter\cbRGB\expandafter{\detokenize{48,48,255}}{mortality\strut} \setlength{\fboxsep}{0pt}\def\cbRGB{\colorbox[RGB]}\expandafter\cbRGB\expandafter{\detokenize{91,91,255}}{,\strut} \setlength{\fboxsep}{0pt}\def\cbRGB{\colorbox[RGB]}\expandafter\cbRGB\expandafter{\detokenize{123,123,255}}{hospitalizations\strut} \setlength{\fboxsep}{0pt}\def\cbRGB{\colorbox[RGB]}\expandafter\cbRGB\expandafter{\detokenize{186,186,255}}{for\strut} \setlength{\fboxsep}{0pt}\def\cbRGB{\colorbox[RGB]}\expandafter\cbRGB\expandafter{\detokenize{185,185,255}}{congestive\strut} \setlength{\fboxsep}{0pt}\def\cbRGB{\colorbox[RGB]}\expandafter\cbRGB\expandafter{\detokenize{191,191,255}}{heart\strut} \setlength{\fboxsep}{0pt}\def\cbRGB{\colorbox[RGB]}\expandafter\cbRGB\expandafter{\detokenize{196,196,255}}{failure\strut} \setlength{\fboxsep}{0pt}\def\cbRGB{\colorbox[RGB]}\expandafter\cbRGB\expandafter{\detokenize{201,201,255}}{,\strut} \setlength{\fboxsep}{0pt}\def\cbRGB{\colorbox[RGB]}\expandafter\cbRGB\expandafter{\detokenize{197,197,255}}{and\strut} \setlength{\fboxsep}{0pt}\def\cbRGB{\colorbox[RGB]}\expandafter\cbRGB\expandafter{\detokenize{207,207,255}}{study\strut} \setlength{\fboxsep}{0pt}\def\cbRGB{\colorbox[RGB]}\expandafter\cbRGB\expandafter{\detokenize{207,207,255}}{quality\strut} \setlength{\fboxsep}{0pt}\def\cbRGB{\colorbox[RGB]}\expandafter\cbRGB\expandafter{\detokenize{192,192,255}}{.\strut} \setlength{\fboxsep}{0pt}\def\cbRGB{\colorbox[RGB]}\expandafter\cbRGB\expandafter{\detokenize{168,168,255}}{data\strut} \setlength{\fboxsep}{0pt}\def\cbRGB{\colorbox[RGB]}\expandafter\cbRGB\expandafter{\detokenize{174,174,255}}{unk\strut} \setlength{\fboxsep}{0pt}\def\cbRGB{\colorbox[RGB]}\expandafter\cbRGB\expandafter{\detokenize{182,182,255}}{:\strut} \setlength{\fboxsep}{0pt}\def\cbRGB{\colorbox[RGB]}\expandafter\cbRGB\expandafter{\detokenize{210,210,255}}{a\strut} \setlength{\fboxsep}{0pt}\def\cbRGB{\colorbox[RGB]}\expandafter\cbRGB\expandafter{\detokenize{220,220,255}}{unk\strut} \setlength{\fboxsep}{0pt}\def\cbRGB{\colorbox[RGB]}\expandafter\cbRGB\expandafter{\detokenize{236,236,255}}{unk\strut} \setlength{\fboxsep}{0pt}\def\cbRGB{\colorbox[RGB]}\expandafter\cbRGB\expandafter{\detokenize{237,237,255}}{model\strut} \setlength{\fboxsep}{0pt}\def\cbRGB{\colorbox[RGB]}\expandafter\cbRGB\expandafter{\detokenize{235,235,255}}{was\strut} \setlength{\fboxsep}{0pt}\def\cbRGB{\colorbox[RGB]}\expandafter\cbRGB\expandafter{\detokenize{232,232,255}}{used\strut} \setlength{\fboxsep}{0pt}\def\cbRGB{\colorbox[RGB]}\expandafter\cbRGB\expandafter{\detokenize{226,226,255}}{to\strut} \setlength{\fboxsep}{0pt}\def\cbRGB{\colorbox[RGB]}\expandafter\cbRGB\expandafter{\detokenize{223,223,255}}{unk\strut} \setlength{\fboxsep}{0pt}\def\cbRGB{\colorbox[RGB]}\expandafter\cbRGB\expandafter{\detokenize{221,221,255}}{the\strut} \setlength{\fboxsep}{0pt}\def\cbRGB{\colorbox[RGB]}\expandafter\cbRGB\expandafter{\detokenize{219,219,255}}{results\strut} \setlength{\fboxsep}{0pt}\def\cbRGB{\colorbox[RGB]}\expandafter\cbRGB\expandafter{\detokenize{218,218,255}}{.\strut} \setlength{\fboxsep}{0pt}\def\cbRGB{\colorbox[RGB]}\expandafter\cbRGB\expandafter{\detokenize{219,219,255}}{a\strut} \setlength{\fboxsep}{0pt}\def\cbRGB{\colorbox[RGB]}\expandafter\cbRGB\expandafter{\detokenize{222,222,255}}{total\strut} \setlength{\fboxsep}{0pt}\def\cbRGB{\colorbox[RGB]}\expandafter\cbRGB\expandafter{\detokenize{225,225,255}}{of\strut} \setlength{\fboxsep}{0pt}\def\cbRGB{\colorbox[RGB]}\expandafter\cbRGB\expandafter{\detokenize{228,228,255}}{qqq\strut} \setlength{\fboxsep}{0pt}\def\cbRGB{\colorbox[RGB]}\expandafter\cbRGB\expandafter{\detokenize{233,233,255}}{trials\strut} \setlength{\fboxsep}{0pt}\def\cbRGB{\colorbox[RGB]}\expandafter\cbRGB\expandafter{\detokenize{236,236,255}}{involving\strut} \setlength{\fboxsep}{0pt}\def\cbRGB{\colorbox[RGB]}\expandafter\cbRGB\expandafter{\detokenize{234,234,255}}{qqq\strut} \setlength{\fboxsep}{0pt}\def\cbRGB{\colorbox[RGB]}\expandafter\cbRGB\expandafter{\detokenize{232,232,255}}{qqq\strut} \setlength{\fboxsep}{0pt}\def\cbRGB{\colorbox[RGB]}\expandafter\cbRGB\expandafter{\detokenize{232,232,255}}{patients\strut} \setlength{\fboxsep}{0pt}\def\cbRGB{\colorbox[RGB]}\expandafter\cbRGB\expandafter{\detokenize{228,228,255}}{were\strut} \setlength{\fboxsep}{0pt}\def\cbRGB{\colorbox[RGB]}\expandafter\cbRGB\expandafter{\detokenize{221,221,255}}{identified\strut} \setlength{\fboxsep}{0pt}\def\cbRGB{\colorbox[RGB]}\expandafter\cbRGB\expandafter{\detokenize{219,219,255}}{.\strut} \setlength{\fboxsep}{0pt}\def\cbRGB{\colorbox[RGB]}\expandafter\cbRGB\expandafter{\detokenize{215,215,255}}{there\strut} \setlength{\fboxsep}{0pt}\def\cbRGB{\colorbox[RGB]}\expandafter\cbRGB\expandafter{\detokenize{192,192,255}}{were\strut} \setlength{\fboxsep}{0pt}\def\cbRGB{\colorbox[RGB]}\expandafter\cbRGB\expandafter{\detokenize{183,183,255}}{qqq\strut} \setlength{\fboxsep}{0pt}\def\cbRGB{\colorbox[RGB]}\expandafter\cbRGB\expandafter{\detokenize{184,184,255}}{deaths\strut} \setlength{\fboxsep}{0pt}\def\cbRGB{\colorbox[RGB]}\expandafter\cbRGB\expandafter{\detokenize{178,178,255}}{among\strut} \setlength{\fboxsep}{0pt}\def\cbRGB{\colorbox[RGB]}\expandafter\cbRGB\expandafter{\detokenize{161,161,255}}{qqq\strut} \setlength{\fboxsep}{0pt}\def\cbRGB{\colorbox[RGB]}\expandafter\cbRGB\expandafter{\detokenize{164,164,255}}{patients\strut} \setlength{\fboxsep}{0pt}\def\cbRGB{\colorbox[RGB]}\expandafter\cbRGB\expandafter{\detokenize{190,190,255}}{randomly\strut} \setlength{\fboxsep}{0pt}\def\cbRGB{\colorbox[RGB]}\expandafter\cbRGB\expandafter{\detokenize{212,212,255}}{assigned\strut} \setlength{\fboxsep}{0pt}\def\cbRGB{\colorbox[RGB]}\expandafter\cbRGB\expandafter{\detokenize{218,218,255}}{to\strut} \setlength{\fboxsep}{0pt}\def\cbRGB{\colorbox[RGB]}\expandafter\cbRGB\expandafter{\detokenize{223,223,255}}{placebo\strut} \setlength{\fboxsep}{0pt}\def\cbRGB{\colorbox[RGB]}\expandafter\cbRGB\expandafter{\detokenize{209,209,255}}{and\strut} \setlength{\fboxsep}{0pt}\def\cbRGB{\colorbox[RGB]}\expandafter\cbRGB\expandafter{\detokenize{200,200,255}}{qqq\strut} \setlength{\fboxsep}{0pt}\def\cbRGB{\colorbox[RGB]}\expandafter\cbRGB\expandafter{\detokenize{195,195,255}}{deaths\strut} \setlength{\fboxsep}{0pt}\def\cbRGB{\colorbox[RGB]}\expandafter\cbRGB\expandafter{\detokenize{199,199,255}}{among\strut} \setlength{\fboxsep}{0pt}\def\cbRGB{\colorbox[RGB]}\expandafter\cbRGB\expandafter{\detokenize{203,203,255}}{qqq\strut} \setlength{\fboxsep}{0pt}\def\cbRGB{\colorbox[RGB]}\expandafter\cbRGB\expandafter{\detokenize{209,209,255}}{patients\strut} \setlength{\fboxsep}{0pt}\def\cbRGB{\colorbox[RGB]}\expandafter\cbRGB\expandafter{\detokenize{221,221,255}}{assigned\strut} \setlength{\fboxsep}{0pt}\def\cbRGB{\colorbox[RGB]}\expandafter\cbRGB\expandafter{\detokenize{225,225,255}}{to\strut} \setlength{\fboxsep}{0pt}\def\cbRGB{\colorbox[RGB]}\expandafter\cbRGB\expandafter{\detokenize{221,221,255}}{unk\strut} \setlength{\fboxsep}{0pt}\def\cbRGB{\colorbox[RGB]}\expandafter\cbRGB\expandafter{\detokenize{217,217,255}}{therapy\strut} \setlength{\fboxsep}{0pt}\def\cbRGB{\colorbox[RGB]}\expandafter\cbRGB\expandafter{\detokenize{216,216,255}}{.\strut} \setlength{\fboxsep}{0pt}\def\cbRGB{\colorbox[RGB]}\expandafter\cbRGB\expandafter{\detokenize{217,217,255}}{in\strut} \setlength{\fboxsep}{0pt}\def\cbRGB{\colorbox[RGB]}\expandafter\cbRGB\expandafter{\detokenize{220,220,255}}{these\strut} \setlength{\fboxsep}{0pt}\def\cbRGB{\colorbox[RGB]}\expandafter\cbRGB\expandafter{\detokenize{223,223,255}}{groups\strut} \setlength{\fboxsep}{0pt}\def\cbRGB{\colorbox[RGB]}\expandafter\cbRGB\expandafter{\detokenize{228,228,255}}{,\strut} \setlength{\fboxsep}{0pt}\def\cbRGB{\colorbox[RGB]}\expandafter\cbRGB\expandafter{\detokenize{230,230,255}}{qqq\strut} \setlength{\fboxsep}{0pt}\def\cbRGB{\colorbox[RGB]}\expandafter\cbRGB\expandafter{\detokenize{230,230,255}}{and\strut} \setlength{\fboxsep}{0pt}\def\cbRGB{\colorbox[RGB]}\expandafter\cbRGB\expandafter{\detokenize{229,229,255}}{qqq\strut} \setlength{\fboxsep}{0pt}\def\cbRGB{\colorbox[RGB]}\expandafter\cbRGB\expandafter{\detokenize{229,229,255}}{patients\strut} \setlength{\fboxsep}{0pt}\def\cbRGB{\colorbox[RGB]}\expandafter\cbRGB\expandafter{\detokenize{230,230,255}}{,\strut} \setlength{\fboxsep}{0pt}\def\cbRGB{\colorbox[RGB]}\expandafter\cbRGB\expandafter{\detokenize{227,227,255}}{respectively\strut} \setlength{\fboxsep}{0pt}\def\cbRGB{\colorbox[RGB]}\expandafter\cbRGB\expandafter{\detokenize{223,223,255}}{,\strut} \setlength{\fboxsep}{0pt}\def\cbRGB{\colorbox[RGB]}\expandafter\cbRGB\expandafter{\detokenize{220,220,255}}{required\strut} \setlength{\fboxsep}{0pt}\def\cbRGB{\colorbox[RGB]}\expandafter\cbRGB\expandafter{\detokenize{216,216,255}}{hospitalization\strut} \setlength{\fboxsep}{0pt}\def\cbRGB{\colorbox[RGB]}\expandafter\cbRGB\expandafter{\detokenize{208,208,255}}{for\strut} \setlength{\fboxsep}{0pt}\def\cbRGB{\colorbox[RGB]}\expandafter\cbRGB\expandafter{\detokenize{194,194,255}}{congestive\strut} \setlength{\fboxsep}{0pt}\def\cbRGB{\colorbox[RGB]}\expandafter\cbRGB\expandafter{\detokenize{195,195,255}}{heart\strut} \setlength{\fboxsep}{0pt}\def\cbRGB{\colorbox[RGB]}\expandafter\cbRGB\expandafter{\detokenize{181,181,255}}{failure\strut} \setlength{\fboxsep}{0pt}\def\cbRGB{\colorbox[RGB]}\expandafter\cbRGB\expandafter{\detokenize{175,175,255}}{.\strut} \setlength{\fboxsep}{0pt}\def\cbRGB{\colorbox[RGB]}\expandafter\cbRGB\expandafter{\detokenize{176,176,255}}{the\strut} \setlength{\fboxsep}{0pt}\def\cbRGB{\colorbox[RGB]}\expandafter\cbRGB\expandafter{\detokenize{195,195,255}}{probability\strut} \setlength{\fboxsep}{0pt}\def\cbRGB{\colorbox[RGB]}\expandafter\cbRGB\expandafter{\detokenize{203,203,255}}{that\strut} \setlength{\fboxsep}{0pt}\def\cbRGB{\colorbox[RGB]}\expandafter\cbRGB\expandafter{\detokenize{198,198,255}}{unk\strut} \setlength{\fboxsep}{0pt}\def\cbRGB{\colorbox[RGB]}\expandafter\cbRGB\expandafter{\detokenize{176,176,255}}{therapy\strut} \setlength{\fboxsep}{0pt}\def\cbRGB{\colorbox[RGB]}\expandafter\cbRGB\expandafter{\detokenize{133,133,255}}{reduced\strut} \setlength{\fboxsep}{0pt}\def\cbRGB{\colorbox[RGB]}\expandafter\cbRGB\expandafter{\detokenize{118,118,255}}{total\strut} \setlength{\fboxsep}{0pt}\def\cbRGB{\colorbox[RGB]}\expandafter\cbRGB\expandafter{\detokenize{99,99,255}}{mortality\strut} \setlength{\fboxsep}{0pt}\def\cbRGB{\colorbox[RGB]}\expandafter\cbRGB\expandafter{\detokenize{143,143,255}}{and\strut} \setlength{\fboxsep}{0pt}\def\cbRGB{\colorbox[RGB]}\expandafter\cbRGB\expandafter{\detokenize{156,156,255}}{hospitalizations\strut} \setlength{\fboxsep}{0pt}\def\cbRGB{\colorbox[RGB]}\expandafter\cbRGB\expandafter{\detokenize{188,188,255}}{for\strut} \setlength{\fboxsep}{0pt}\def\cbRGB{\colorbox[RGB]}\expandafter\cbRGB\expandafter{\detokenize{182,182,255}}{congestive\strut} \setlength{\fboxsep}{0pt}\def\cbRGB{\colorbox[RGB]}\expandafter\cbRGB\expandafter{\detokenize{171,171,255}}{heart\strut} \setlength{\fboxsep}{0pt}\def\cbRGB{\colorbox[RGB]}\expandafter\cbRGB\expandafter{\detokenize{154,154,255}}{failure\strut} \setlength{\fboxsep}{0pt}\def\cbRGB{\colorbox[RGB]}\expandafter\cbRGB\expandafter{\detokenize{166,166,255}}{was\strut} \setlength{\fboxsep}{0pt}\def\cbRGB{\colorbox[RGB]}\expandafter\cbRGB\expandafter{\detokenize{179,179,255}}{almost\strut} \setlength{\fboxsep}{0pt}\def\cbRGB{\colorbox[RGB]}\expandafter\cbRGB\expandafter{\detokenize{205,205,255}}{qqq\strut} \setlength{\fboxsep}{0pt}\def\cbRGB{\colorbox[RGB]}\expandafter\cbRGB\expandafter{\detokenize{211,211,255}}{\%\strut} \setlength{\fboxsep}{0pt}\def\cbRGB{\colorbox[RGB]}\expandafter\cbRGB\expandafter{\detokenize{218,218,255}}{.\strut} \setlength{\fboxsep}{0pt}\def\cbRGB{\colorbox[RGB]}\expandafter\cbRGB\expandafter{\detokenize{224,224,255}}{the\strut} \setlength{\fboxsep}{0pt}\def\cbRGB{\colorbox[RGB]}\expandafter\cbRGB\expandafter{\detokenize{220,220,255}}{best\strut} \setlength{\fboxsep}{0pt}\def\cbRGB{\colorbox[RGB]}\expandafter\cbRGB\expandafter{\detokenize{218,218,255}}{estimates\strut} \setlength{\fboxsep}{0pt}\def\cbRGB{\colorbox[RGB]}\expandafter\cbRGB\expandafter{\detokenize{212,212,255}}{of\strut} \setlength{\fboxsep}{0pt}\def\cbRGB{\colorbox[RGB]}\expandafter\cbRGB\expandafter{\detokenize{209,209,255}}{these\strut} \setlength{\fboxsep}{0pt}\def\cbRGB{\colorbox[RGB]}\expandafter\cbRGB\expandafter{\detokenize{212,212,255}}{advantages\strut} \setlength{\fboxsep}{0pt}\def\cbRGB{\colorbox[RGB]}\expandafter\cbRGB\expandafter{\detokenize{215,215,255}}{are\strut} \setlength{\fboxsep}{0pt}\def\cbRGB{\colorbox[RGB]}\expandafter\cbRGB\expandafter{\detokenize{219,219,255}}{qqq\strut} \setlength{\fboxsep}{0pt}\def\cbRGB{\colorbox[RGB]}\expandafter\cbRGB\expandafter{\detokenize{220,220,255}}{unk\strut} \setlength{\fboxsep}{0pt}\def\cbRGB{\colorbox[RGB]}\expandafter\cbRGB\expandafter{\detokenize{211,211,255}}{unk\strut} \setlength{\fboxsep}{0pt}\def\cbRGB{\colorbox[RGB]}\expandafter\cbRGB\expandafter{\detokenize{180,180,255}}{and\strut} \setlength{\fboxsep}{0pt}\def\cbRGB{\colorbox[RGB]}\expandafter\cbRGB\expandafter{\detokenize{166,166,255}}{qqq\strut} \setlength{\fboxsep}{0pt}\def\cbRGB{\colorbox[RGB]}\expandafter\cbRGB\expandafter{\detokenize{174,174,255}}{fewer\strut} \setlength{\fboxsep}{0pt}\def\cbRGB{\colorbox[RGB]}\expandafter\cbRGB\expandafter{\detokenize{191,191,255}}{hospitalizations\strut} \setlength{\fboxsep}{0pt}\def\cbRGB{\colorbox[RGB]}\expandafter\cbRGB\expandafter{\detokenize{195,195,255}}{per\strut} \setlength{\fboxsep}{0pt}\def\cbRGB{\colorbox[RGB]}\expandafter\cbRGB\expandafter{\detokenize{193,193,255}}{qqq\strut} \setlength{\fboxsep}{0pt}\def\cbRGB{\colorbox[RGB]}\expandafter\cbRGB\expandafter{\detokenize{205,205,255}}{patients\strut} \setlength{\fboxsep}{0pt}\def\cbRGB{\colorbox[RGB]}\expandafter\cbRGB\expandafter{\detokenize{211,211,255}}{treated\strut} \setlength{\fboxsep}{0pt}\def\cbRGB{\colorbox[RGB]}\expandafter\cbRGB\expandafter{\detokenize{211,211,255}}{in\strut} \setlength{\fboxsep}{0pt}\def\cbRGB{\colorbox[RGB]}\expandafter\cbRGB\expandafter{\detokenize{208,208,255}}{the\strut} \setlength{\fboxsep}{0pt}\def\cbRGB{\colorbox[RGB]}\expandafter\cbRGB\expandafter{\detokenize{206,206,255}}{first\strut} \setlength{\fboxsep}{0pt}\def\cbRGB{\colorbox[RGB]}\expandafter\cbRGB\expandafter{\detokenize{210,210,255}}{year\strut} \setlength{\fboxsep}{0pt}\def\cbRGB{\colorbox[RGB]}\expandafter\cbRGB\expandafter{\detokenize{214,214,255}}{after\strut} \setlength{\fboxsep}{0pt}\def\cbRGB{\colorbox[RGB]}\expandafter\cbRGB\expandafter{\detokenize{206,206,255}}{therapy\strut} \setlength{\fboxsep}{0pt}\def\cbRGB{\colorbox[RGB]}\expandafter\cbRGB\expandafter{\detokenize{194,194,255}}{.\strut} \setlength{\fboxsep}{0pt}\def\cbRGB{\colorbox[RGB]}\expandafter\cbRGB\expandafter{\detokenize{194,194,255}}{the\strut} \setlength{\fboxsep}{0pt}\def\cbRGB{\colorbox[RGB]}\expandafter\cbRGB\expandafter{\detokenize{199,199,255}}{probability\strut} \setlength{\fboxsep}{0pt}\def\cbRGB{\colorbox[RGB]}\expandafter\cbRGB\expandafter{\detokenize{200,200,255}}{that\strut} \setlength{\fboxsep}{0pt}\def\cbRGB{\colorbox[RGB]}\expandafter\cbRGB\expandafter{\detokenize{195,195,255}}{these\strut} \setlength{\fboxsep}{0pt}\def\cbRGB{\colorbox[RGB]}\expandafter\cbRGB\expandafter{\detokenize{199,199,255}}{benefits\strut} \setlength{\fboxsep}{0pt}\def\cbRGB{\colorbox[RGB]}\expandafter\cbRGB\expandafter{\detokenize{207,207,255}}{are\strut} \setlength{\fboxsep}{0pt}\def\cbRGB{\colorbox[RGB]}\expandafter\cbRGB\expandafter{\detokenize{218,218,255}}{clinically\strut} \setlength{\fboxsep}{0pt}\def\cbRGB{\colorbox[RGB]}\expandafter\cbRGB\expandafter{\detokenize{232,232,255}}{significant\strut} \setlength{\fboxsep}{0pt}\def\cbRGB{\colorbox[RGB]}\expandafter\cbRGB\expandafter{\detokenize{246,246,255}}{(\strut} \setlength{\fboxsep}{0pt}\def\cbRGB{\colorbox[RGB]}\expandafter\cbRGB\expandafter{\detokenize{251,251,255}}{>\strut} \setlength{\fboxsep}{0pt}\def\cbRGB{\colorbox[RGB]}\expandafter\cbRGB\expandafter{\detokenize{247,247,255}}{qqq\strut} \setlength{\fboxsep}{0pt}\def\cbRGB{\colorbox[RGB]}\expandafter\cbRGB\expandafter{\detokenize{248,248,255}}{unk\strut} \setlength{\fboxsep}{0pt}\def\cbRGB{\colorbox[RGB]}\expandafter\cbRGB\expandafter{\detokenize{254,254,255}}{unk\strut} \setlength{\fboxsep}{0pt}\def\cbRGB{\colorbox[RGB]}\expandafter\cbRGB\expandafter{\detokenize{255,255,255}}{or\strut} \setlength{\fboxsep}{0pt}\def\cbRGB{\colorbox[RGB]}\expandafter\cbRGB\expandafter{\detokenize{242,242,255}}{>\strut} \setlength{\fboxsep}{0pt}\def\cbRGB{\colorbox[RGB]}\expandafter\cbRGB\expandafter{\detokenize{225,225,255}}{qqq\strut} \setlength{\fboxsep}{0pt}\def\cbRGB{\colorbox[RGB]}\expandafter\cbRGB\expandafter{\detokenize{214,214,255}}{fewer\strut} \setlength{\fboxsep}{0pt}\def\cbRGB{\colorbox[RGB]}\expandafter\cbRGB\expandafter{\detokenize{201,201,255}}{hospitalizations\strut} \setlength{\fboxsep}{0pt}\def\cbRGB{\colorbox[RGB]}\expandafter\cbRGB\expandafter{\detokenize{196,196,255}}{per\strut} \setlength{\fboxsep}{0pt}\def\cbRGB{\colorbox[RGB]}\expandafter\cbRGB\expandafter{\detokenize{197,197,255}}{qqq\strut} \setlength{\fboxsep}{0pt}\def\cbRGB{\colorbox[RGB]}\expandafter\cbRGB\expandafter{\detokenize{206,206,255}}{patients\strut} \setlength{\fboxsep}{0pt}\def\cbRGB{\colorbox[RGB]}\expandafter\cbRGB\expandafter{\detokenize{204,204,255}}{treated\strut} \setlength{\fboxsep}{0pt}\def\cbRGB{\colorbox[RGB]}\expandafter\cbRGB\expandafter{\detokenize{205,205,255}}{)\strut} \setlength{\fboxsep}{0pt}\def\cbRGB{\colorbox[RGB]}\expandafter\cbRGB\expandafter{\detokenize{209,209,255}}{is\strut} \setlength{\fboxsep}{0pt}\def\cbRGB{\colorbox[RGB]}\expandafter\cbRGB\expandafter{\detokenize{218,218,255}}{qqq\strut} \setlength{\fboxsep}{0pt}\def\cbRGB{\colorbox[RGB]}\expandafter\cbRGB\expandafter{\detokenize{220,220,255}}{\%\strut} \setlength{\fboxsep}{0pt}\def\cbRGB{\colorbox[RGB]}\expandafter\cbRGB\expandafter{\detokenize{222,222,255}}{.\strut} \setlength{\fboxsep}{0pt}\def\cbRGB{\colorbox[RGB]}\expandafter\cbRGB\expandafter{\detokenize{224,224,255}}{both\strut} \setlength{\fboxsep}{0pt}\def\cbRGB{\colorbox[RGB]}\expandafter\cbRGB\expandafter{\detokenize{223,223,255}}{selective\strut} \setlength{\fboxsep}{0pt}\def\cbRGB{\colorbox[RGB]}\expandafter\cbRGB\expandafter{\detokenize{223,223,255}}{and\strut} \setlength{\fboxsep}{0pt}\def\cbRGB{\colorbox[RGB]}\expandafter\cbRGB\expandafter{\detokenize{222,222,255}}{unk\strut} \setlength{\fboxsep}{0pt}\def\cbRGB{\colorbox[RGB]}\expandafter\cbRGB\expandafter{\detokenize{221,221,255}}{agents\strut} \setlength{\fboxsep}{0pt}\def\cbRGB{\colorbox[RGB]}\expandafter\cbRGB\expandafter{\detokenize{219,219,255}}{produced\strut} \setlength{\fboxsep}{0pt}\def\cbRGB{\colorbox[RGB]}\expandafter\cbRGB\expandafter{\detokenize{216,216,255}}{these\strut} \setlength{\fboxsep}{0pt}\def\cbRGB{\colorbox[RGB]}\expandafter\cbRGB\expandafter{\detokenize{215,215,255}}{unk\strut} \setlength{\fboxsep}{0pt}\def\cbRGB{\colorbox[RGB]}\expandafter\cbRGB\expandafter{\detokenize{214,214,255}}{effects\strut} \setlength{\fboxsep}{0pt}\def\cbRGB{\colorbox[RGB]}\expandafter\cbRGB\expandafter{\detokenize{210,210,255}}{.\strut} \setlength{\fboxsep}{0pt}\def\cbRGB{\colorbox[RGB]}\expandafter\cbRGB\expandafter{\detokenize{203,203,255}}{the\strut} \setlength{\fboxsep}{0pt}\def\cbRGB{\colorbox[RGB]}\expandafter\cbRGB\expandafter{\detokenize{204,204,255}}{results\strut} \setlength{\fboxsep}{0pt}\def\cbRGB{\colorbox[RGB]}\expandafter\cbRGB\expandafter{\detokenize{208,208,255}}{are\strut} \setlength{\fboxsep}{0pt}\def\cbRGB{\colorbox[RGB]}\expandafter\cbRGB\expandafter{\detokenize{214,214,255}}{unk\strut} \setlength{\fboxsep}{0pt}\def\cbRGB{\colorbox[RGB]}\expandafter\cbRGB\expandafter{\detokenize{217,217,255}}{to\strut} \setlength{\fboxsep}{0pt}\def\cbRGB{\colorbox[RGB]}\expandafter\cbRGB\expandafter{\detokenize{210,210,255}}{any\strut} \setlength{\fboxsep}{0pt}\def\cbRGB{\colorbox[RGB]}\expandafter\cbRGB\expandafter{\detokenize{211,211,255}}{unk\strut} \setlength{\fboxsep}{0pt}\def\cbRGB{\colorbox[RGB]}\expandafter\cbRGB\expandafter{\detokenize{195,195,255}}{publication\strut} \setlength{\fboxsep}{0pt}\def\cbRGB{\colorbox[RGB]}\expandafter\cbRGB\expandafter{\detokenize{186,186,255}}{unk\strut} \setlength{\fboxsep}{0pt}\def\cbRGB{\colorbox[RGB]}\expandafter\cbRGB\expandafter{\detokenize{164,164,255}}{.\strut} \setlength{\fboxsep}{0pt}\def\cbRGB{\colorbox[RGB]}\expandafter\cbRGB\expandafter{\detokenize{175,175,255}}{conclusions\strut} \setlength{\fboxsep}{0pt}\def\cbRGB{\colorbox[RGB]}\expandafter\cbRGB\expandafter{\detokenize{167,167,255}}{:\strut} \setlength{\fboxsep}{0pt}\def\cbRGB{\colorbox[RGB]}\expandafter\cbRGB\expandafter{\detokenize{181,181,255}}{unk\strut} \setlength{\fboxsep}{0pt}\def\cbRGB{\colorbox[RGB]}\expandafter\cbRGB\expandafter{\detokenize{185,185,255}}{therapy\strut} \setlength{\fboxsep}{0pt}\def\cbRGB{\colorbox[RGB]}\expandafter\cbRGB\expandafter{\detokenize{209,209,255}}{is\strut} \setlength{\fboxsep}{0pt}\def\cbRGB{\colorbox[RGB]}\expandafter\cbRGB\expandafter{\detokenize{212,212,255}}{associated\strut} \setlength{\fboxsep}{0pt}\def\cbRGB{\colorbox[RGB]}\expandafter\cbRGB\expandafter{\detokenize{197,197,255}}{with\strut} \setlength{\fboxsep}{0pt}\def\cbRGB{\colorbox[RGB]}\expandafter\cbRGB\expandafter{\detokenize{174,174,255}}{clinically\strut} \setlength{\fboxsep}{0pt}\def\cbRGB{\colorbox[RGB]}\expandafter\cbRGB\expandafter{\detokenize{135,135,255}}{meaningful\strut} \setlength{\fboxsep}{0pt}\def\cbRGB{\colorbox[RGB]}\expandafter\cbRGB\expandafter{\detokenize{58,58,255}}{reductions\strut} \setlength{\fboxsep}{0pt}\def\cbRGB{\colorbox[RGB]}\expandafter\cbRGB\expandafter{\detokenize{34,34,255}}{in\strut} \setlength{\fboxsep}{0pt}\def\cbRGB{\colorbox[RGB]}\expandafter\cbRGB\expandafter{\detokenize{0,0,255}}{mortality\strut} \setlength{\fboxsep}{0pt}\def\cbRGB{\colorbox[RGB]}\expandafter\cbRGB\expandafter{\detokenize{63,63,255}}{and\strut} \setlength{\fboxsep}{0pt}\def\cbRGB{\colorbox[RGB]}\expandafter\cbRGB\expandafter{\detokenize{104,104,255}}{morbidity\strut} \setlength{\fboxsep}{0pt}\def\cbRGB{\colorbox[RGB]}\expandafter\cbRGB\expandafter{\detokenize{182,182,255}}{in\strut} \setlength{\fboxsep}{0pt}\def\cbRGB{\colorbox[RGB]}\expandafter\cbRGB\expandafter{\detokenize{217,217,255}}{patients\strut} \setlength{\fboxsep}{0pt}\def\cbRGB{\colorbox[RGB]}\expandafter\cbRGB\expandafter{\detokenize{235,235,255}}{with\strut} \setlength{\fboxsep}{0pt}\def\cbRGB{\colorbox[RGB]}\expandafter\cbRGB\expandafter{\detokenize{239,239,255}}{stable\strut} \setlength{\fboxsep}{0pt}\def\cbRGB{\colorbox[RGB]}\expandafter\cbRGB\expandafter{\detokenize{228,228,255}}{congestive\strut} \setlength{\fboxsep}{0pt}\def\cbRGB{\colorbox[RGB]}\expandafter\cbRGB\expandafter{\detokenize{213,213,255}}{heart\strut} \setlength{\fboxsep}{0pt}\def\cbRGB{\colorbox[RGB]}\expandafter\cbRGB\expandafter{\detokenize{185,185,255}}{failure\strut} \setlength{\fboxsep}{0pt}\def\cbRGB{\colorbox[RGB]}\expandafter\cbRGB\expandafter{\detokenize{183,183,255}}{and\strut} \setlength{\fboxsep}{0pt}\def\cbRGB{\colorbox[RGB]}\expandafter\cbRGB\expandafter{\detokenize{190,190,255}}{should\strut} \setlength{\fboxsep}{0pt}\def\cbRGB{\colorbox[RGB]}\expandafter\cbRGB\expandafter{\detokenize{216,216,255}}{be\strut} \setlength{\fboxsep}{0pt}\def\cbRGB{\colorbox[RGB]}\expandafter\cbRGB\expandafter{\detokenize{228,228,255}}{routinely\strut} \setlength{\fboxsep}{0pt}\def\cbRGB{\colorbox[RGB]}\expandafter\cbRGB\expandafter{\detokenize{235,235,255}}{offered\strut} \setlength{\fboxsep}{0pt}\def\cbRGB{\colorbox[RGB]}\expandafter\cbRGB\expandafter{\detokenize{237,237,255}}{to\strut} \setlength{\fboxsep}{0pt}\def\cbRGB{\colorbox[RGB]}\expandafter\cbRGB\expandafter{\detokenize{235,235,255}}{all\strut} \setlength{\fboxsep}{0pt}\def\cbRGB{\colorbox[RGB]}\expandafter\cbRGB\expandafter{\detokenize{229,229,255}}{patients\strut} \setlength{\fboxsep}{0pt}\def\cbRGB{\colorbox[RGB]}\expandafter\cbRGB\expandafter{\detokenize{227,227,255}}{similar\strut} \setlength{\fboxsep}{0pt}\def\cbRGB{\colorbox[RGB]}\expandafter\cbRGB\expandafter{\detokenize{224,224,255}}{to\strut} \setlength{\fboxsep}{0pt}\def\cbRGB{\colorbox[RGB]}\expandafter\cbRGB\expandafter{\detokenize{220,220,255}}{those\strut} \setlength{\fboxsep}{0pt}\def\cbRGB{\colorbox[RGB]}\expandafter\cbRGB\expandafter{\detokenize{219,219,255}}{included\strut} \setlength{\fboxsep}{0pt}\def\cbRGB{\colorbox[RGB]}\expandafter\cbRGB\expandafter{\detokenize{219,219,255}}{in\strut} \setlength{\fboxsep}{0pt}\def\cbRGB{\colorbox[RGB]}\expandafter\cbRGB\expandafter{\detokenize{221,221,255}}{trials\strut} \setlength{\fboxsep}{0pt}\def\cbRGB{\colorbox[RGB]}\expandafter\cbRGB\expandafter{\detokenize{242,242,255}}{.\strut} 

\subsection{BeerAdvocate}
Color Legend : \setlength{\fboxsep}{0pt}\def\cbRGB{\colorbox[RGB]}\expandafter\cbRGB\expandafter{\detokenize{255,0,0}}{Look\strut} 
\setlength{\fboxsep}{0pt}\def\cbRGB{\colorbox[RGB]}\expandafter\cbRGB\expandafter{\detokenize{0,255,0}}{Aroma\strut} 
\setlength{\fboxsep}{0pt}\def\cbRGB{\colorbox[RGB]}\expandafter\cbRGB\expandafter{\detokenize{0,0,255}}{Palate\strut} 
\setlength{\fboxsep}{0pt}\def\cbRGB{\colorbox[RGB]}\expandafter\cbRGB\expandafter{\detokenize{255,255,0}}{Taste\strut}

\textbf{Example 1}

\setlength{\fboxsep}{0pt}\def\cbRGB{\colorbox[RGB]}\expandafter\cbRGB\expandafter{\detokenize{255,109,109}}{appearance\strut} \setlength{\fboxsep}{0pt}\def\cbRGB{\colorbox[RGB]}\expandafter\cbRGB\expandafter{\detokenize{255,54,54}}{was\strut} \setlength{\fboxsep}{0pt}\def\cbRGB{\colorbox[RGB]}\expandafter\cbRGB\expandafter{\detokenize{255,0,0}}{gold\strut} \setlength{\fboxsep}{0pt}\def\cbRGB{\colorbox[RGB]}\expandafter\cbRGB\expandafter{\detokenize{255,54,54}}{,\strut} \setlength{\fboxsep}{0pt}\def\cbRGB{\colorbox[RGB]}\expandafter\cbRGB\expandafter{\detokenize{255,91,91}}{clear\strut} \setlength{\fboxsep}{0pt}\def\cbRGB{\colorbox[RGB]}\expandafter\cbRGB\expandafter{\detokenize{255,109,109}}{,\strut} \setlength{\fboxsep}{0pt}\def\cbRGB{\colorbox[RGB]}\expandafter\cbRGB\expandafter{\detokenize{255,128,128}}{no\strut} \setlength{\fboxsep}{0pt}\def\cbRGB{\colorbox[RGB]}\expandafter\cbRGB\expandafter{\detokenize{255,165,165}}{head\strut} \setlength{\fboxsep}{0pt}\def\cbRGB{\colorbox[RGB]}\expandafter\cbRGB\expandafter{\detokenize{255,183,183}}{,\strut} \setlength{\fboxsep}{0pt}\def\cbRGB{\colorbox[RGB]}\expandafter\cbRGB\expandafter{\detokenize{255,184,184}}{and\strut} \setlength{\fboxsep}{0pt}\def\cbRGB{\colorbox[RGB]}\expandafter\cbRGB\expandafter{\detokenize{255,166,166}}{fizzy\strut} \setlength{\fboxsep}{0pt}\def\cbRGB{\colorbox[RGB]}\expandafter\cbRGB\expandafter{\detokenize{255,165,165}}{.\strut} \setlength{\fboxsep}{0pt}\def\cbRGB{\colorbox[RGB]}\expandafter\cbRGB\expandafter{\detokenize{255,146,146}}{almost\strut} \setlength{\fboxsep}{0pt}\def\cbRGB{\colorbox[RGB]}\expandafter\cbRGB\expandafter{\detokenize{255,127,127}}{looked\strut} \setlength{\fboxsep}{0pt}\def\cbRGB{\colorbox[RGB]}\expandafter\cbRGB\expandafter{\detokenize{255,128,128}}{like\strut} \setlength{\fboxsep}{0pt}\def\cbRGB{\colorbox[RGB]}\expandafter\cbRGB\expandafter{\detokenize{255,162,162}}{champagne\strut} \setlength{\fboxsep}{0pt}\def\cbRGB{\colorbox[RGB]}\expandafter\cbRGB\expandafter{\detokenize{255,167,167}}{.\strut} \setlength{\fboxsep}{0pt}\def\cbRGB{\colorbox[RGB]}\expandafter\cbRGB\expandafter{\detokenize{255,172,172}}{aroma\strut} \setlength{\fboxsep}{0pt}\def\cbRGB{\colorbox[RGB]}\expandafter\cbRGB\expandafter{\detokenize{255,178,178}}{was\strut} \setlength{\fboxsep}{0pt}\def\cbRGB{\colorbox[RGB]}\expandafter\cbRGB\expandafter{\detokenize{255,199,199}}{actually\strut} \setlength{\fboxsep}{0pt}\def\cbRGB{\colorbox[RGB]}\expandafter\cbRGB\expandafter{\detokenize{255,208,208}}{real\strut} \setlength{\fboxsep}{0pt}\def\cbRGB{\colorbox[RGB]}\expandafter\cbRGB\expandafter{\detokenize{255,240,240}}{good\strut} \setlength{\fboxsep}{0pt}\def\cbRGB{\colorbox[RGB]}\expandafter\cbRGB\expandafter{\detokenize{255,255,255}}{.\strut} \setlength{\fboxsep}{0pt}\def\cbRGB{\colorbox[RGB]}\expandafter\cbRGB\expandafter{\detokenize{255,255,255}}{sweet\strut} \setlength{\fboxsep}{0pt}\def\cbRGB{\colorbox[RGB]}\expandafter\cbRGB\expandafter{\detokenize{255,255,255}}{apricot\strut} \setlength{\fboxsep}{0pt}\def\cbRGB{\colorbox[RGB]}\expandafter\cbRGB\expandafter{\detokenize{255,255,255}}{,\strut} \setlength{\fboxsep}{0pt}\def\cbRGB{\colorbox[RGB]}\expandafter\cbRGB\expandafter{\detokenize{255,255,255}}{unk\strut} \setlength{\fboxsep}{0pt}\def\cbRGB{\colorbox[RGB]}\expandafter\cbRGB\expandafter{\detokenize{255,255,255}}{smith\strut} \setlength{\fboxsep}{0pt}\def\cbRGB{\colorbox[RGB]}\expandafter\cbRGB\expandafter{\detokenize{255,255,255}}{apple\strut} \setlength{\fboxsep}{0pt}\def\cbRGB{\colorbox[RGB]}\expandafter\cbRGB\expandafter{\detokenize{255,255,255}}{,\strut} \setlength{\fboxsep}{0pt}\def\cbRGB{\colorbox[RGB]}\expandafter\cbRGB\expandafter{\detokenize{255,255,255}}{almost\strut} \setlength{\fboxsep}{0pt}\def\cbRGB{\colorbox[RGB]}\expandafter\cbRGB\expandafter{\detokenize{255,255,255}}{like\strut} \setlength{\fboxsep}{0pt}\def\cbRGB{\colorbox[RGB]}\expandafter\cbRGB\expandafter{\detokenize{255,255,255}}{a\strut} \setlength{\fboxsep}{0pt}\def\cbRGB{\colorbox[RGB]}\expandafter\cbRGB\expandafter{\detokenize{255,255,255}}{unk\strut} \setlength{\fboxsep}{0pt}\def\cbRGB{\colorbox[RGB]}\expandafter\cbRGB\expandafter{\detokenize{255,255,255}}{wine\strut} \setlength{\fboxsep}{0pt}\def\cbRGB{\colorbox[RGB]}\expandafter\cbRGB\expandafter{\detokenize{255,255,255}}{.\strut} \setlength{\fboxsep}{0pt}\def\cbRGB{\colorbox[RGB]}\expandafter\cbRGB\expandafter{\detokenize{255,255,255}}{taste\strut} \setlength{\fboxsep}{0pt}\def\cbRGB{\colorbox[RGB]}\expandafter\cbRGB\expandafter{\detokenize{255,255,255}}{was\strut} \setlength{\fboxsep}{0pt}\def\cbRGB{\colorbox[RGB]}\expandafter\cbRGB\expandafter{\detokenize{255,255,255}}{unk\strut} \setlength{\fboxsep}{0pt}\def\cbRGB{\colorbox[RGB]}\expandafter\cbRGB\expandafter{\detokenize{255,255,255}}{.\strut} \setlength{\fboxsep}{0pt}\def\cbRGB{\colorbox[RGB]}\expandafter\cbRGB\expandafter{\detokenize{255,255,255}}{i\strut} \setlength{\fboxsep}{0pt}\def\cbRGB{\colorbox[RGB]}\expandafter\cbRGB\expandafter{\detokenize{255,255,255}}{unk\strut} \setlength{\fboxsep}{0pt}\def\cbRGB{\colorbox[RGB]}\expandafter\cbRGB\expandafter{\detokenize{255,255,255}}{a\strut} \setlength{\fboxsep}{0pt}\def\cbRGB{\colorbox[RGB]}\expandafter\cbRGB\expandafter{\detokenize{255,255,255}}{couple\strut} \setlength{\fboxsep}{0pt}\def\cbRGB{\colorbox[RGB]}\expandafter\cbRGB\expandafter{\detokenize{255,255,255}}{of\strut} \setlength{\fboxsep}{0pt}\def\cbRGB{\colorbox[RGB]}\expandafter\cbRGB\expandafter{\detokenize{255,255,255}}{bottles\strut} \setlength{\fboxsep}{0pt}\def\cbRGB{\colorbox[RGB]}\expandafter\cbRGB\expandafter{\detokenize{255,255,255}}{an\strut} \setlength{\fboxsep}{0pt}\def\cbRGB{\colorbox[RGB]}\expandafter\cbRGB\expandafter{\detokenize{255,255,255}}{tried\strut} \setlength{\fboxsep}{0pt}\def\cbRGB{\colorbox[RGB]}\expandafter\cbRGB\expandafter{\detokenize{255,255,255}}{them\strut} \setlength{\fboxsep}{0pt}\def\cbRGB{\colorbox[RGB]}\expandafter\cbRGB\expandafter{\detokenize{255,255,255}}{so\strut} \setlength{\fboxsep}{0pt}\def\cbRGB{\colorbox[RGB]}\expandafter\cbRGB\expandafter{\detokenize{255,255,255}}{if\strut} \setlength{\fboxsep}{0pt}\def\cbRGB{\colorbox[RGB]}\expandafter\cbRGB\expandafter{\detokenize{255,255,255}}{they\strut} \setlength{\fboxsep}{0pt}\def\cbRGB{\colorbox[RGB]}\expandafter\cbRGB\expandafter{\detokenize{255,255,255}}{were\strut} \setlength{\fboxsep}{0pt}\def\cbRGB{\colorbox[RGB]}\expandafter\cbRGB\expandafter{\detokenize{255,255,255}}{skunked\strut} \setlength{\fboxsep}{0pt}\def\cbRGB{\colorbox[RGB]}\expandafter\cbRGB\expandafter{\detokenize{255,255,255}}{,\strut} \setlength{\fboxsep}{0pt}\def\cbRGB{\colorbox[RGB]}\expandafter\cbRGB\expandafter{\detokenize{255,255,255}}{it\strut} \setlength{\fboxsep}{0pt}\def\cbRGB{\colorbox[RGB]}\expandafter\cbRGB\expandafter{\detokenize{255,255,255}}{was\strut} \setlength{\fboxsep}{0pt}\def\cbRGB{\colorbox[RGB]}\expandafter\cbRGB\expandafter{\detokenize{255,255,255}}{unk\strut} \setlength{\fboxsep}{0pt}\def\cbRGB{\colorbox[RGB]}\expandafter\cbRGB\expandafter{\detokenize{255,255,255}}{unk\strut} \setlength{\fboxsep}{0pt}\def\cbRGB{\colorbox[RGB]}\expandafter\cbRGB\expandafter{\detokenize{255,255,255}}{that\strut} \setlength{\fboxsep}{0pt}\def\cbRGB{\colorbox[RGB]}\expandafter\cbRGB\expandafter{\detokenize{255,255,255}}{were\strut} \setlength{\fboxsep}{0pt}\def\cbRGB{\colorbox[RGB]}\expandafter\cbRGB\expandafter{\detokenize{255,229,229}}{bad\strut} \setlength{\fboxsep}{0pt}\def\cbRGB{\colorbox[RGB]}\expandafter\cbRGB\expandafter{\detokenize{255,224,224}}{.\strut} \setlength{\fboxsep}{0pt}\def\cbRGB{\colorbox[RGB]}\expandafter\cbRGB\expandafter{\detokenize{255,218,218}}{there\strut} \setlength{\fboxsep}{0pt}\def\cbRGB{\colorbox[RGB]}\expandafter\cbRGB\expandafter{\detokenize{255,213,213}}{was\strut} \setlength{\fboxsep}{0pt}\def\cbRGB{\colorbox[RGB]}\expandafter\cbRGB\expandafter{\detokenize{255,207,207}}{no\strut} \setlength{\fboxsep}{0pt}\def\cbRGB{\colorbox[RGB]}\expandafter\cbRGB\expandafter{\detokenize{255,239,239}}{real\strut} \setlength{\fboxsep}{0pt}\def\cbRGB{\colorbox[RGB]}\expandafter\cbRGB\expandafter{\detokenize{255,242,242}}{apricot\strut} \setlength{\fboxsep}{0pt}\def\cbRGB{\colorbox[RGB]}\expandafter\cbRGB\expandafter{\detokenize{255,245,245}}{taste\strut} \setlength{\fboxsep}{0pt}\def\cbRGB{\colorbox[RGB]}\expandafter\cbRGB\expandafter{\detokenize{255,249,249}}{,\strut} \setlength{\fboxsep}{0pt}\def\cbRGB{\colorbox[RGB]}\expandafter\cbRGB\expandafter{\detokenize{255,252,252}}{and\strut} \setlength{\fboxsep}{0pt}\def\cbRGB{\colorbox[RGB]}\expandafter\cbRGB\expandafter{\detokenize{255,237,237}}{no\strut} \setlength{\fboxsep}{0pt}\def\cbRGB{\colorbox[RGB]}\expandafter\cbRGB\expandafter{\detokenize{255,255,255}}{sweetness\strut} \setlength{\fboxsep}{0pt}\def\cbRGB{\colorbox[RGB]}\expandafter\cbRGB\expandafter{\detokenize{255,255,255}}{.\strut} \setlength{\fboxsep}{0pt}\def\cbRGB{\colorbox[RGB]}\expandafter\cbRGB\expandafter{\detokenize{255,255,255}}{it\strut} \setlength{\fboxsep}{0pt}\def\cbRGB{\colorbox[RGB]}\expandafter\cbRGB\expandafter{\detokenize{255,255,255}}{tasted\strut} \setlength{\fboxsep}{0pt}\def\cbRGB{\colorbox[RGB]}\expandafter\cbRGB\expandafter{\detokenize{255,255,255}}{like\strut} \setlength{\fboxsep}{0pt}\def\cbRGB{\colorbox[RGB]}\expandafter\cbRGB\expandafter{\detokenize{255,255,255}}{hay\strut} \setlength{\fboxsep}{0pt}\def\cbRGB{\colorbox[RGB]}\expandafter\cbRGB\expandafter{\detokenize{255,255,255}}{but\strut} \setlength{\fboxsep}{0pt}\def\cbRGB{\colorbox[RGB]}\expandafter\cbRGB\expandafter{\detokenize{255,222,222}}{not\strut} \setlength{\fboxsep}{0pt}\def\cbRGB{\colorbox[RGB]}\expandafter\cbRGB\expandafter{\detokenize{255,221,221}}{in\strut} \setlength{\fboxsep}{0pt}\def\cbRGB{\colorbox[RGB]}\expandafter\cbRGB\expandafter{\detokenize{255,220,220}}{the\strut} \setlength{\fboxsep}{0pt}\def\cbRGB{\colorbox[RGB]}\expandafter\cbRGB\expandafter{\detokenize{255,220,220}}{good\strut} \setlength{\fboxsep}{0pt}\def\cbRGB{\colorbox[RGB]}\expandafter\cbRGB\expandafter{\detokenize{255,219,219}}{wheat\strut} \setlength{\fboxsep}{0pt}\def\cbRGB{\colorbox[RGB]}\expandafter\cbRGB\expandafter{\detokenize{255,255,255}}{beer\strut} \setlength{\fboxsep}{0pt}\def\cbRGB{\colorbox[RGB]}\expandafter\cbRGB\expandafter{\detokenize{255,255,255}}{sense\strut} \setlength{\fboxsep}{0pt}\def\cbRGB{\colorbox[RGB]}\expandafter\cbRGB\expandafter{\detokenize{255,254,254}}{,\strut} \setlength{\fboxsep}{0pt}\def\cbRGB{\colorbox[RGB]}\expandafter\cbRGB\expandafter{\detokenize{255,254,254}}{more\strut} \setlength{\fboxsep}{0pt}\def\cbRGB{\colorbox[RGB]}\expandafter\cbRGB\expandafter{\detokenize{255,254,254}}{like\strut} \setlength{\fboxsep}{0pt}\def\cbRGB{\colorbox[RGB]}\expandafter\cbRGB\expandafter{\detokenize{255,235,235}}{dirty\strut} \setlength{\fboxsep}{0pt}\def\cbRGB{\colorbox[RGB]}\expandafter\cbRGB\expandafter{\detokenize{255,254,254}}{hay\strut} \setlength{\fboxsep}{0pt}\def\cbRGB{\colorbox[RGB]}\expandafter\cbRGB\expandafter{\detokenize{255,255,255}}{thats\strut} \setlength{\fboxsep}{0pt}\def\cbRGB{\colorbox[RGB]}\expandafter\cbRGB\expandafter{\detokenize{255,255,255}}{been\strut} \setlength{\fboxsep}{0pt}\def\cbRGB{\colorbox[RGB]}\expandafter\cbRGB\expandafter{\detokenize{255,255,255}}{under\strut} \setlength{\fboxsep}{0pt}\def\cbRGB{\colorbox[RGB]}\expandafter\cbRGB\expandafter{\detokenize{255,255,255}}{the\strut} \setlength{\fboxsep}{0pt}\def\cbRGB{\colorbox[RGB]}\expandafter\cbRGB\expandafter{\detokenize{255,255,255}}{budweiser\strut} \setlength{\fboxsep}{0pt}\def\cbRGB{\colorbox[RGB]}\expandafter\cbRGB\expandafter{\detokenize{255,255,255}}{unk\strut} \setlength{\fboxsep}{0pt}\def\cbRGB{\colorbox[RGB]}\expandafter\cbRGB\expandafter{\detokenize{255,255,255}}{.\strut} \setlength{\fboxsep}{0pt}\def\cbRGB{\colorbox[RGB]}\expandafter\cbRGB\expandafter{\detokenize{255,255,255}}{mouthfeel\strut} \setlength{\fboxsep}{0pt}\def\cbRGB{\colorbox[RGB]}\expandafter\cbRGB\expandafter{\detokenize{255,255,255}}{was\strut} \setlength{\fboxsep}{0pt}\def\cbRGB{\colorbox[RGB]}\expandafter\cbRGB\expandafter{\detokenize{255,255,255}}{thin\strut} \setlength{\fboxsep}{0pt}\def\cbRGB{\colorbox[RGB]}\expandafter\cbRGB\expandafter{\detokenize{255,255,255}}{and\strut} \setlength{\fboxsep}{0pt}\def\cbRGB{\colorbox[RGB]}\expandafter\cbRGB\expandafter{\detokenize{255,255,255}}{carbonated\strut} \setlength{\fboxsep}{0pt}\def\cbRGB{\colorbox[RGB]}\expandafter\cbRGB\expandafter{\detokenize{255,255,255}}{.\strut} \setlength{\fboxsep}{0pt}\def\cbRGB{\colorbox[RGB]}\expandafter\cbRGB\expandafter{\detokenize{255,255,255}}{drinkability\strut} \setlength{\fboxsep}{0pt}\def\cbRGB{\colorbox[RGB]}\expandafter\cbRGB\expandafter{\detokenize{255,213,213}}{{\ldots}\strut} \setlength{\fboxsep}{0pt}\def\cbRGB{\colorbox[RGB]}\expandafter\cbRGB\expandafter{\detokenize{255,144,144}}{well\strut} \setlength{\fboxsep}{0pt}\def\cbRGB{\colorbox[RGB]}\expandafter\cbRGB\expandafter{\detokenize{255,75,75}}{this\strut} \setlength{\fboxsep}{0pt}\def\cbRGB{\colorbox[RGB]}\expandafter\cbRGB\expandafter{\detokenize{255,42,42}}{is\strut} \setlength{\fboxsep}{0pt}\def\cbRGB{\colorbox[RGB]}\expandafter\cbRGB\expandafter{\detokenize{255,45,45}}{the\strut} \setlength{\fboxsep}{0pt}\def\cbRGB{\colorbox[RGB]}\expandafter\cbRGB\expandafter{\detokenize{255,84,84}}{second\strut} \setlength{\fboxsep}{0pt}\def\cbRGB{\colorbox[RGB]}\expandafter\cbRGB\expandafter{\detokenize{255,145,145}}{beer\strut} \setlength{\fboxsep}{0pt}\def\cbRGB{\colorbox[RGB]}\expandafter\cbRGB\expandafter{\detokenize{255,205,205}}{in\strut} \setlength{\fboxsep}{0pt}\def\cbRGB{\colorbox[RGB]}\expandafter\cbRGB\expandafter{\detokenize{255,248,248}}{my\strut} \setlength{\fboxsep}{0pt}\def\cbRGB{\colorbox[RGB]}\expandafter\cbRGB\expandafter{\detokenize{255,255,255}}{long\strut} \setlength{\fboxsep}{0pt}\def\cbRGB{\colorbox[RGB]}\expandafter\cbRGB\expandafter{\detokenize{255,255,255}}{list\strut} \setlength{\fboxsep}{0pt}\def\cbRGB{\colorbox[RGB]}\expandafter\cbRGB\expandafter{\detokenize{255,255,255}}{that\strut} \setlength{\fboxsep}{0pt}\def\cbRGB{\colorbox[RGB]}\expandafter\cbRGB\expandafter{\detokenize{255,255,255}}{i\strut} \setlength{\fboxsep}{0pt}\def\cbRGB{\colorbox[RGB]}\expandafter\cbRGB\expandafter{\detokenize{255,255,255}}{'ve\strut} \setlength{\fboxsep}{0pt}\def\cbRGB{\colorbox[RGB]}\expandafter\cbRGB\expandafter{\detokenize{255,255,255}}{actually\strut} \setlength{\fboxsep}{0pt}\def\cbRGB{\colorbox[RGB]}\expandafter\cbRGB\expandafter{\detokenize{255,255,255}}{had\strut} \setlength{\fboxsep}{0pt}\def\cbRGB{\colorbox[RGB]}\expandafter\cbRGB\expandafter{\detokenize{255,255,255}}{to\strut} \setlength{\fboxsep}{0pt}\def\cbRGB{\colorbox[RGB]}\expandafter\cbRGB\expandafter{\detokenize{255,255,255}}{pour\strut} \setlength{\fboxsep}{0pt}\def\cbRGB{\colorbox[RGB]}\expandafter\cbRGB\expandafter{\detokenize{255,255,255}}{out\strut} \setlength{\fboxsep}{0pt}\def\cbRGB{\colorbox[RGB]}\expandafter\cbRGB\expandafter{\detokenize{255,255,255}}{.\strut} 

\setlength{\fboxsep}{0pt}\def\cbRGB{\colorbox[RGB]}\expandafter\cbRGB\expandafter{\detokenize{81,255,81}}{appearance\strut} \setlength{\fboxsep}{0pt}\def\cbRGB{\colorbox[RGB]}\expandafter\cbRGB\expandafter{\detokenize{193,255,193}}{was\strut} \setlength{\fboxsep}{0pt}\def\cbRGB{\colorbox[RGB]}\expandafter\cbRGB\expandafter{\detokenize{255,255,255}}{gold\strut} \setlength{\fboxsep}{0pt}\def\cbRGB{\colorbox[RGB]}\expandafter\cbRGB\expandafter{\detokenize{255,255,255}}{,\strut} \setlength{\fboxsep}{0pt}\def\cbRGB{\colorbox[RGB]}\expandafter\cbRGB\expandafter{\detokenize{255,255,255}}{clear\strut} \setlength{\fboxsep}{0pt}\def\cbRGB{\colorbox[RGB]}\expandafter\cbRGB\expandafter{\detokenize{255,255,255}}{,\strut} \setlength{\fboxsep}{0pt}\def\cbRGB{\colorbox[RGB]}\expandafter\cbRGB\expandafter{\detokenize{255,255,255}}{no\strut} \setlength{\fboxsep}{0pt}\def\cbRGB{\colorbox[RGB]}\expandafter\cbRGB\expandafter{\detokenize{255,255,255}}{head\strut} \setlength{\fboxsep}{0pt}\def\cbRGB{\colorbox[RGB]}\expandafter\cbRGB\expandafter{\detokenize{255,255,255}}{,\strut} \setlength{\fboxsep}{0pt}\def\cbRGB{\colorbox[RGB]}\expandafter\cbRGB\expandafter{\detokenize{255,255,255}}{and\strut} \setlength{\fboxsep}{0pt}\def\cbRGB{\colorbox[RGB]}\expandafter\cbRGB\expandafter{\detokenize{255,255,255}}{fizzy\strut} \setlength{\fboxsep}{0pt}\def\cbRGB{\colorbox[RGB]}\expandafter\cbRGB\expandafter{\detokenize{255,255,255}}{.\strut} \setlength{\fboxsep}{0pt}\def\cbRGB{\colorbox[RGB]}\expandafter\cbRGB\expandafter{\detokenize{255,255,255}}{almost\strut} \setlength{\fboxsep}{0pt}\def\cbRGB{\colorbox[RGB]}\expandafter\cbRGB\expandafter{\detokenize{255,255,255}}{looked\strut} \setlength{\fboxsep}{0pt}\def\cbRGB{\colorbox[RGB]}\expandafter\cbRGB\expandafter{\detokenize{255,255,255}}{like\strut} \setlength{\fboxsep}{0pt}\def\cbRGB{\colorbox[RGB]}\expandafter\cbRGB\expandafter{\detokenize{255,255,255}}{champagne\strut} \setlength{\fboxsep}{0pt}\def\cbRGB{\colorbox[RGB]}\expandafter\cbRGB\expandafter{\detokenize{237,255,237}}{.\strut} \setlength{\fboxsep}{0pt}\def\cbRGB{\colorbox[RGB]}\expandafter\cbRGB\expandafter{\detokenize{164,255,164}}{aroma\strut} \setlength{\fboxsep}{0pt}\def\cbRGB{\colorbox[RGB]}\expandafter\cbRGB\expandafter{\detokenize{91,255,91}}{was\strut} \setlength{\fboxsep}{0pt}\def\cbRGB{\colorbox[RGB]}\expandafter\cbRGB\expandafter{\detokenize{21,255,21}}{actually\strut} \setlength{\fboxsep}{0pt}\def\cbRGB{\colorbox[RGB]}\expandafter\cbRGB\expandafter{\detokenize{9,255,9}}{real\strut} \setlength{\fboxsep}{0pt}\def\cbRGB{\colorbox[RGB]}\expandafter\cbRGB\expandafter{\detokenize{51,255,51}}{good\strut} \setlength{\fboxsep}{0pt}\def\cbRGB{\colorbox[RGB]}\expandafter\cbRGB\expandafter{\detokenize{64,255,64}}{.\strut} \setlength{\fboxsep}{0pt}\def\cbRGB{\colorbox[RGB]}\expandafter\cbRGB\expandafter{\detokenize{49,255,49}}{sweet\strut} \setlength{\fboxsep}{0pt}\def\cbRGB{\colorbox[RGB]}\expandafter\cbRGB\expandafter{\detokenize{75,255,75}}{apricot\strut} \setlength{\fboxsep}{0pt}\def\cbRGB{\colorbox[RGB]}\expandafter\cbRGB\expandafter{\detokenize{72,255,72}}{,\strut} \setlength{\fboxsep}{0pt}\def\cbRGB{\colorbox[RGB]}\expandafter\cbRGB\expandafter{\detokenize{46,255,46}}{unk\strut} \setlength{\fboxsep}{0pt}\def\cbRGB{\colorbox[RGB]}\expandafter\cbRGB\expandafter{\detokenize{77,255,77}}{smith\strut} \setlength{\fboxsep}{0pt}\def\cbRGB{\colorbox[RGB]}\expandafter\cbRGB\expandafter{\detokenize{166,255,166}}{apple\strut} \setlength{\fboxsep}{0pt}\def\cbRGB{\colorbox[RGB]}\expandafter\cbRGB\expandafter{\detokenize{201,255,201}}{,\strut} \setlength{\fboxsep}{0pt}\def\cbRGB{\colorbox[RGB]}\expandafter\cbRGB\expandafter{\detokenize{236,255,236}}{almost\strut} \setlength{\fboxsep}{0pt}\def\cbRGB{\colorbox[RGB]}\expandafter\cbRGB\expandafter{\detokenize{255,255,255}}{like\strut} \setlength{\fboxsep}{0pt}\def\cbRGB{\colorbox[RGB]}\expandafter\cbRGB\expandafter{\detokenize{255,255,255}}{a\strut} \setlength{\fboxsep}{0pt}\def\cbRGB{\colorbox[RGB]}\expandafter\cbRGB\expandafter{\detokenize{230,255,230}}{unk\strut} \setlength{\fboxsep}{0pt}\def\cbRGB{\colorbox[RGB]}\expandafter\cbRGB\expandafter{\detokenize{246,255,246}}{wine\strut} \setlength{\fboxsep}{0pt}\def\cbRGB{\colorbox[RGB]}\expandafter\cbRGB\expandafter{\detokenize{255,255,255}}{.\strut} \setlength{\fboxsep}{0pt}\def\cbRGB{\colorbox[RGB]}\expandafter\cbRGB\expandafter{\detokenize{255,255,255}}{taste\strut} \setlength{\fboxsep}{0pt}\def\cbRGB{\colorbox[RGB]}\expandafter\cbRGB\expandafter{\detokenize{255,255,255}}{was\strut} \setlength{\fboxsep}{0pt}\def\cbRGB{\colorbox[RGB]}\expandafter\cbRGB\expandafter{\detokenize{255,255,255}}{unk\strut} \setlength{\fboxsep}{0pt}\def\cbRGB{\colorbox[RGB]}\expandafter\cbRGB\expandafter{\detokenize{255,255,255}}{.\strut} \setlength{\fboxsep}{0pt}\def\cbRGB{\colorbox[RGB]}\expandafter\cbRGB\expandafter{\detokenize{255,255,255}}{i\strut} \setlength{\fboxsep}{0pt}\def\cbRGB{\colorbox[RGB]}\expandafter\cbRGB\expandafter{\detokenize{255,255,255}}{unk\strut} \setlength{\fboxsep}{0pt}\def\cbRGB{\colorbox[RGB]}\expandafter\cbRGB\expandafter{\detokenize{255,255,255}}{a\strut} \setlength{\fboxsep}{0pt}\def\cbRGB{\colorbox[RGB]}\expandafter\cbRGB\expandafter{\detokenize{255,255,255}}{couple\strut} \setlength{\fboxsep}{0pt}\def\cbRGB{\colorbox[RGB]}\expandafter\cbRGB\expandafter{\detokenize{255,255,255}}{of\strut} \setlength{\fboxsep}{0pt}\def\cbRGB{\colorbox[RGB]}\expandafter\cbRGB\expandafter{\detokenize{255,255,255}}{bottles\strut} \setlength{\fboxsep}{0pt}\def\cbRGB{\colorbox[RGB]}\expandafter\cbRGB\expandafter{\detokenize{255,255,255}}{an\strut} \setlength{\fboxsep}{0pt}\def\cbRGB{\colorbox[RGB]}\expandafter\cbRGB\expandafter{\detokenize{255,255,255}}{tried\strut} \setlength{\fboxsep}{0pt}\def\cbRGB{\colorbox[RGB]}\expandafter\cbRGB\expandafter{\detokenize{255,255,255}}{them\strut} \setlength{\fboxsep}{0pt}\def\cbRGB{\colorbox[RGB]}\expandafter\cbRGB\expandafter{\detokenize{255,255,255}}{so\strut} \setlength{\fboxsep}{0pt}\def\cbRGB{\colorbox[RGB]}\expandafter\cbRGB\expandafter{\detokenize{255,255,255}}{if\strut} \setlength{\fboxsep}{0pt}\def\cbRGB{\colorbox[RGB]}\expandafter\cbRGB\expandafter{\detokenize{255,255,255}}{they\strut} \setlength{\fboxsep}{0pt}\def\cbRGB{\colorbox[RGB]}\expandafter\cbRGB\expandafter{\detokenize{255,255,255}}{were\strut} \setlength{\fboxsep}{0pt}\def\cbRGB{\colorbox[RGB]}\expandafter\cbRGB\expandafter{\detokenize{255,255,255}}{skunked\strut} \setlength{\fboxsep}{0pt}\def\cbRGB{\colorbox[RGB]}\expandafter\cbRGB\expandafter{\detokenize{255,255,255}}{,\strut} \setlength{\fboxsep}{0pt}\def\cbRGB{\colorbox[RGB]}\expandafter\cbRGB\expandafter{\detokenize{255,255,255}}{it\strut} \setlength{\fboxsep}{0pt}\def\cbRGB{\colorbox[RGB]}\expandafter\cbRGB\expandafter{\detokenize{255,255,255}}{was\strut} \setlength{\fboxsep}{0pt}\def\cbRGB{\colorbox[RGB]}\expandafter\cbRGB\expandafter{\detokenize{255,255,255}}{unk\strut} \setlength{\fboxsep}{0pt}\def\cbRGB{\colorbox[RGB]}\expandafter\cbRGB\expandafter{\detokenize{255,255,255}}{unk\strut} \setlength{\fboxsep}{0pt}\def\cbRGB{\colorbox[RGB]}\expandafter\cbRGB\expandafter{\detokenize{255,255,255}}{that\strut} \setlength{\fboxsep}{0pt}\def\cbRGB{\colorbox[RGB]}\expandafter\cbRGB\expandafter{\detokenize{255,255,255}}{were\strut} \setlength{\fboxsep}{0pt}\def\cbRGB{\colorbox[RGB]}\expandafter\cbRGB\expandafter{\detokenize{216,255,216}}{bad\strut} \setlength{\fboxsep}{0pt}\def\cbRGB{\colorbox[RGB]}\expandafter\cbRGB\expandafter{\detokenize{174,255,174}}{.\strut} \setlength{\fboxsep}{0pt}\def\cbRGB{\colorbox[RGB]}\expandafter\cbRGB\expandafter{\detokenize{132,255,132}}{there\strut} \setlength{\fboxsep}{0pt}\def\cbRGB{\colorbox[RGB]}\expandafter\cbRGB\expandafter{\detokenize{138,255,138}}{was\strut} \setlength{\fboxsep}{0pt}\def\cbRGB{\colorbox[RGB]}\expandafter\cbRGB\expandafter{\detokenize{144,255,144}}{no\strut} \setlength{\fboxsep}{0pt}\def\cbRGB{\colorbox[RGB]}\expandafter\cbRGB\expandafter{\detokenize{150,255,150}}{real\strut} \setlength{\fboxsep}{0pt}\def\cbRGB{\colorbox[RGB]}\expandafter\cbRGB\expandafter{\detokenize{199,255,199}}{apricot\strut} \setlength{\fboxsep}{0pt}\def\cbRGB{\colorbox[RGB]}\expandafter\cbRGB\expandafter{\detokenize{247,255,247}}{taste\strut} \setlength{\fboxsep}{0pt}\def\cbRGB{\colorbox[RGB]}\expandafter\cbRGB\expandafter{\detokenize{255,255,255}}{,\strut} \setlength{\fboxsep}{0pt}\def\cbRGB{\colorbox[RGB]}\expandafter\cbRGB\expandafter{\detokenize{255,255,255}}{and\strut} \setlength{\fboxsep}{0pt}\def\cbRGB{\colorbox[RGB]}\expandafter\cbRGB\expandafter{\detokenize{255,255,255}}{no\strut} \setlength{\fboxsep}{0pt}\def\cbRGB{\colorbox[RGB]}\expandafter\cbRGB\expandafter{\detokenize{255,255,255}}{sweetness\strut} \setlength{\fboxsep}{0pt}\def\cbRGB{\colorbox[RGB]}\expandafter\cbRGB\expandafter{\detokenize{255,255,255}}{.\strut} \setlength{\fboxsep}{0pt}\def\cbRGB{\colorbox[RGB]}\expandafter\cbRGB\expandafter{\detokenize{255,255,255}}{it\strut} \setlength{\fboxsep}{0pt}\def\cbRGB{\colorbox[RGB]}\expandafter\cbRGB\expandafter{\detokenize{255,255,255}}{tasted\strut} \setlength{\fboxsep}{0pt}\def\cbRGB{\colorbox[RGB]}\expandafter\cbRGB\expandafter{\detokenize{255,255,255}}{like\strut} \setlength{\fboxsep}{0pt}\def\cbRGB{\colorbox[RGB]}\expandafter\cbRGB\expandafter{\detokenize{255,255,255}}{hay\strut} \setlength{\fboxsep}{0pt}\def\cbRGB{\colorbox[RGB]}\expandafter\cbRGB\expandafter{\detokenize{255,255,255}}{but\strut} \setlength{\fboxsep}{0pt}\def\cbRGB{\colorbox[RGB]}\expandafter\cbRGB\expandafter{\detokenize{255,255,255}}{not\strut} \setlength{\fboxsep}{0pt}\def\cbRGB{\colorbox[RGB]}\expandafter\cbRGB\expandafter{\detokenize{255,255,255}}{in\strut} \setlength{\fboxsep}{0pt}\def\cbRGB{\colorbox[RGB]}\expandafter\cbRGB\expandafter{\detokenize{255,255,255}}{the\strut} \setlength{\fboxsep}{0pt}\def\cbRGB{\colorbox[RGB]}\expandafter\cbRGB\expandafter{\detokenize{255,255,255}}{good\strut} \setlength{\fboxsep}{0pt}\def\cbRGB{\colorbox[RGB]}\expandafter\cbRGB\expandafter{\detokenize{255,255,255}}{wheat\strut} \setlength{\fboxsep}{0pt}\def\cbRGB{\colorbox[RGB]}\expandafter\cbRGB\expandafter{\detokenize{233,255,233}}{beer\strut} \setlength{\fboxsep}{0pt}\def\cbRGB{\colorbox[RGB]}\expandafter\cbRGB\expandafter{\detokenize{233,255,233}}{sense\strut} \setlength{\fboxsep}{0pt}\def\cbRGB{\colorbox[RGB]}\expandafter\cbRGB\expandafter{\detokenize{232,255,232}}{,\strut} \setlength{\fboxsep}{0pt}\def\cbRGB{\colorbox[RGB]}\expandafter\cbRGB\expandafter{\detokenize{232,255,232}}{more\strut} \setlength{\fboxsep}{0pt}\def\cbRGB{\colorbox[RGB]}\expandafter\cbRGB\expandafter{\detokenize{232,255,232}}{like\strut} \setlength{\fboxsep}{0pt}\def\cbRGB{\colorbox[RGB]}\expandafter\cbRGB\expandafter{\detokenize{229,255,229}}{dirty\strut} \setlength{\fboxsep}{0pt}\def\cbRGB{\colorbox[RGB]}\expandafter\cbRGB\expandafter{\detokenize{255,255,255}}{hay\strut} \setlength{\fboxsep}{0pt}\def\cbRGB{\colorbox[RGB]}\expandafter\cbRGB\expandafter{\detokenize{255,255,255}}{thats\strut} \setlength{\fboxsep}{0pt}\def\cbRGB{\colorbox[RGB]}\expandafter\cbRGB\expandafter{\detokenize{255,255,255}}{been\strut} \setlength{\fboxsep}{0pt}\def\cbRGB{\colorbox[RGB]}\expandafter\cbRGB\expandafter{\detokenize{255,255,255}}{under\strut} \setlength{\fboxsep}{0pt}\def\cbRGB{\colorbox[RGB]}\expandafter\cbRGB\expandafter{\detokenize{255,255,255}}{the\strut} \setlength{\fboxsep}{0pt}\def\cbRGB{\colorbox[RGB]}\expandafter\cbRGB\expandafter{\detokenize{255,255,255}}{budweiser\strut} \setlength{\fboxsep}{0pt}\def\cbRGB{\colorbox[RGB]}\expandafter\cbRGB\expandafter{\detokenize{255,255,255}}{unk\strut} \setlength{\fboxsep}{0pt}\def\cbRGB{\colorbox[RGB]}\expandafter\cbRGB\expandafter{\detokenize{255,255,255}}{.\strut} \setlength{\fboxsep}{0pt}\def\cbRGB{\colorbox[RGB]}\expandafter\cbRGB\expandafter{\detokenize{255,255,255}}{mouthfeel\strut} \setlength{\fboxsep}{0pt}\def\cbRGB{\colorbox[RGB]}\expandafter\cbRGB\expandafter{\detokenize{255,255,255}}{was\strut} \setlength{\fboxsep}{0pt}\def\cbRGB{\colorbox[RGB]}\expandafter\cbRGB\expandafter{\detokenize{207,255,207}}{thin\strut} \setlength{\fboxsep}{0pt}\def\cbRGB{\colorbox[RGB]}\expandafter\cbRGB\expandafter{\detokenize{123,255,123}}{and\strut} \setlength{\fboxsep}{0pt}\def\cbRGB{\colorbox[RGB]}\expandafter\cbRGB\expandafter{\detokenize{36,255,36}}{carbonated\strut} \setlength{\fboxsep}{0pt}\def\cbRGB{\colorbox[RGB]}\expandafter\cbRGB\expandafter{\detokenize{3,255,3}}{.\strut} \setlength{\fboxsep}{0pt}\def\cbRGB{\colorbox[RGB]}\expandafter\cbRGB\expandafter{\detokenize{0,255,0}}{drinkability\strut} \setlength{\fboxsep}{0pt}\def\cbRGB{\colorbox[RGB]}\expandafter\cbRGB\expandafter{\detokenize{48,255,48}}{{\ldots}\strut} \setlength{\fboxsep}{0pt}\def\cbRGB{\colorbox[RGB]}\expandafter\cbRGB\expandafter{\detokenize{69,255,69}}{well\strut} \setlength{\fboxsep}{0pt}\def\cbRGB{\colorbox[RGB]}\expandafter\cbRGB\expandafter{\detokenize{96,255,96}}{this\strut} \setlength{\fboxsep}{0pt}\def\cbRGB{\colorbox[RGB]}\expandafter\cbRGB\expandafter{\detokenize{93,255,93}}{is\strut} \setlength{\fboxsep}{0pt}\def\cbRGB{\colorbox[RGB]}\expandafter\cbRGB\expandafter{\detokenize{119,255,119}}{the\strut} \setlength{\fboxsep}{0pt}\def\cbRGB{\colorbox[RGB]}\expandafter\cbRGB\expandafter{\detokenize{119,255,119}}{second\strut} \setlength{\fboxsep}{0pt}\def\cbRGB{\colorbox[RGB]}\expandafter\cbRGB\expandafter{\detokenize{148,255,148}}{beer\strut} \setlength{\fboxsep}{0pt}\def\cbRGB{\colorbox[RGB]}\expandafter\cbRGB\expandafter{\detokenize{173,255,173}}{in\strut} \setlength{\fboxsep}{0pt}\def\cbRGB{\colorbox[RGB]}\expandafter\cbRGB\expandafter{\detokenize{204,255,204}}{my\strut} \setlength{\fboxsep}{0pt}\def\cbRGB{\colorbox[RGB]}\expandafter\cbRGB\expandafter{\detokenize{206,255,206}}{long\strut} \setlength{\fboxsep}{0pt}\def\cbRGB{\colorbox[RGB]}\expandafter\cbRGB\expandafter{\detokenize{211,255,211}}{list\strut} \setlength{\fboxsep}{0pt}\def\cbRGB{\colorbox[RGB]}\expandafter\cbRGB\expandafter{\detokenize{244,255,244}}{that\strut} \setlength{\fboxsep}{0pt}\def\cbRGB{\colorbox[RGB]}\expandafter\cbRGB\expandafter{\detokenize{255,255,255}}{i\strut} \setlength{\fboxsep}{0pt}\def\cbRGB{\colorbox[RGB]}\expandafter\cbRGB\expandafter{\detokenize{255,255,255}}{'ve\strut} \setlength{\fboxsep}{0pt}\def\cbRGB{\colorbox[RGB]}\expandafter\cbRGB\expandafter{\detokenize{255,255,255}}{actually\strut} \setlength{\fboxsep}{0pt}\def\cbRGB{\colorbox[RGB]}\expandafter\cbRGB\expandafter{\detokenize{255,255,255}}{had\strut} \setlength{\fboxsep}{0pt}\def\cbRGB{\colorbox[RGB]}\expandafter\cbRGB\expandafter{\detokenize{255,255,255}}{to\strut} \setlength{\fboxsep}{0pt}\def\cbRGB{\colorbox[RGB]}\expandafter\cbRGB\expandafter{\detokenize{255,255,255}}{pour\strut} \setlength{\fboxsep}{0pt}\def\cbRGB{\colorbox[RGB]}\expandafter\cbRGB\expandafter{\detokenize{255,255,255}}{out\strut} \setlength{\fboxsep}{0pt}\def\cbRGB{\colorbox[RGB]}\expandafter\cbRGB\expandafter{\detokenize{255,255,255}}{.\strut} 

\setlength{\fboxsep}{0pt}\def\cbRGB{\colorbox[RGB]}\expandafter\cbRGB\expandafter{\detokenize{83,83,255}}{appearance\strut} \setlength{\fboxsep}{0pt}\def\cbRGB{\colorbox[RGB]}\expandafter\cbRGB\expandafter{\detokenize{143,143,255}}{was\strut} \setlength{\fboxsep}{0pt}\def\cbRGB{\colorbox[RGB]}\expandafter\cbRGB\expandafter{\detokenize{202,202,255}}{gold\strut} \setlength{\fboxsep}{0pt}\def\cbRGB{\colorbox[RGB]}\expandafter\cbRGB\expandafter{\detokenize{244,244,255}}{,\strut} \setlength{\fboxsep}{0pt}\def\cbRGB{\colorbox[RGB]}\expandafter\cbRGB\expandafter{\detokenize{255,255,255}}{clear\strut} \setlength{\fboxsep}{0pt}\def\cbRGB{\colorbox[RGB]}\expandafter\cbRGB\expandafter{\detokenize{255,255,255}}{,\strut} \setlength{\fboxsep}{0pt}\def\cbRGB{\colorbox[RGB]}\expandafter\cbRGB\expandafter{\detokenize{255,255,255}}{no\strut} \setlength{\fboxsep}{0pt}\def\cbRGB{\colorbox[RGB]}\expandafter\cbRGB\expandafter{\detokenize{255,255,255}}{head\strut} \setlength{\fboxsep}{0pt}\def\cbRGB{\colorbox[RGB]}\expandafter\cbRGB\expandafter{\detokenize{242,242,255}}{,\strut} \setlength{\fboxsep}{0pt}\def\cbRGB{\colorbox[RGB]}\expandafter\cbRGB\expandafter{\detokenize{204,204,255}}{and\strut} \setlength{\fboxsep}{0pt}\def\cbRGB{\colorbox[RGB]}\expandafter\cbRGB\expandafter{\detokenize{167,167,255}}{fizzy\strut} \setlength{\fboxsep}{0pt}\def\cbRGB{\colorbox[RGB]}\expandafter\cbRGB\expandafter{\detokenize{151,151,255}}{.\strut} \setlength{\fboxsep}{0pt}\def\cbRGB{\colorbox[RGB]}\expandafter\cbRGB\expandafter{\detokenize{152,152,255}}{almost\strut} \setlength{\fboxsep}{0pt}\def\cbRGB{\colorbox[RGB]}\expandafter\cbRGB\expandafter{\detokenize{159,159,255}}{looked\strut} \setlength{\fboxsep}{0pt}\def\cbRGB{\colorbox[RGB]}\expandafter\cbRGB\expandafter{\detokenize{184,184,255}}{like\strut} \setlength{\fboxsep}{0pt}\def\cbRGB{\colorbox[RGB]}\expandafter\cbRGB\expandafter{\detokenize{191,191,255}}{champagne\strut} \setlength{\fboxsep}{0pt}\def\cbRGB{\colorbox[RGB]}\expandafter\cbRGB\expandafter{\detokenize{181,181,255}}{.\strut} \setlength{\fboxsep}{0pt}\def\cbRGB{\colorbox[RGB]}\expandafter\cbRGB\expandafter{\detokenize{139,139,255}}{aroma\strut} \setlength{\fboxsep}{0pt}\def\cbRGB{\colorbox[RGB]}\expandafter\cbRGB\expandafter{\detokenize{109,109,255}}{was\strut} \setlength{\fboxsep}{0pt}\def\cbRGB{\colorbox[RGB]}\expandafter\cbRGB\expandafter{\detokenize{93,93,255}}{actually\strut} \setlength{\fboxsep}{0pt}\def\cbRGB{\colorbox[RGB]}\expandafter\cbRGB\expandafter{\detokenize{111,111,255}}{real\strut} \setlength{\fboxsep}{0pt}\def\cbRGB{\colorbox[RGB]}\expandafter\cbRGB\expandafter{\detokenize{133,133,255}}{good\strut} \setlength{\fboxsep}{0pt}\def\cbRGB{\colorbox[RGB]}\expandafter\cbRGB\expandafter{\detokenize{180,180,255}}{.\strut} \setlength{\fboxsep}{0pt}\def\cbRGB{\colorbox[RGB]}\expandafter\cbRGB\expandafter{\detokenize{227,227,255}}{sweet\strut} \setlength{\fboxsep}{0pt}\def\cbRGB{\colorbox[RGB]}\expandafter\cbRGB\expandafter{\detokenize{255,255,255}}{apricot\strut} \setlength{\fboxsep}{0pt}\def\cbRGB{\colorbox[RGB]}\expandafter\cbRGB\expandafter{\detokenize{255,255,255}}{,\strut} \setlength{\fboxsep}{0pt}\def\cbRGB{\colorbox[RGB]}\expandafter\cbRGB\expandafter{\detokenize{255,255,255}}{unk\strut} \setlength{\fboxsep}{0pt}\def\cbRGB{\colorbox[RGB]}\expandafter\cbRGB\expandafter{\detokenize{255,255,255}}{smith\strut} \setlength{\fboxsep}{0pt}\def\cbRGB{\colorbox[RGB]}\expandafter\cbRGB\expandafter{\detokenize{255,255,255}}{apple\strut} \setlength{\fboxsep}{0pt}\def\cbRGB{\colorbox[RGB]}\expandafter\cbRGB\expandafter{\detokenize{255,255,255}}{,\strut} \setlength{\fboxsep}{0pt}\def\cbRGB{\colorbox[RGB]}\expandafter\cbRGB\expandafter{\detokenize{255,255,255}}{almost\strut} \setlength{\fboxsep}{0pt}\def\cbRGB{\colorbox[RGB]}\expandafter\cbRGB\expandafter{\detokenize{255,255,255}}{like\strut} \setlength{\fboxsep}{0pt}\def\cbRGB{\colorbox[RGB]}\expandafter\cbRGB\expandafter{\detokenize{255,255,255}}{a\strut} \setlength{\fboxsep}{0pt}\def\cbRGB{\colorbox[RGB]}\expandafter\cbRGB\expandafter{\detokenize{255,255,255}}{unk\strut} \setlength{\fboxsep}{0pt}\def\cbRGB{\colorbox[RGB]}\expandafter\cbRGB\expandafter{\detokenize{231,231,255}}{wine\strut} \setlength{\fboxsep}{0pt}\def\cbRGB{\colorbox[RGB]}\expandafter\cbRGB\expandafter{\detokenize{218,218,255}}{.\strut} \setlength{\fboxsep}{0pt}\def\cbRGB{\colorbox[RGB]}\expandafter\cbRGB\expandafter{\detokenize{214,214,255}}{taste\strut} \setlength{\fboxsep}{0pt}\def\cbRGB{\colorbox[RGB]}\expandafter\cbRGB\expandafter{\detokenize{241,241,255}}{was\strut} \setlength{\fboxsep}{0pt}\def\cbRGB{\colorbox[RGB]}\expandafter\cbRGB\expandafter{\detokenize{255,255,255}}{unk\strut} \setlength{\fboxsep}{0pt}\def\cbRGB{\colorbox[RGB]}\expandafter\cbRGB\expandafter{\detokenize{255,255,255}}{.\strut} \setlength{\fboxsep}{0pt}\def\cbRGB{\colorbox[RGB]}\expandafter\cbRGB\expandafter{\detokenize{255,255,255}}{i\strut} \setlength{\fboxsep}{0pt}\def\cbRGB{\colorbox[RGB]}\expandafter\cbRGB\expandafter{\detokenize{255,255,255}}{unk\strut} \setlength{\fboxsep}{0pt}\def\cbRGB{\colorbox[RGB]}\expandafter\cbRGB\expandafter{\detokenize{255,255,255}}{a\strut} \setlength{\fboxsep}{0pt}\def\cbRGB{\colorbox[RGB]}\expandafter\cbRGB\expandafter{\detokenize{255,255,255}}{couple\strut} \setlength{\fboxsep}{0pt}\def\cbRGB{\colorbox[RGB]}\expandafter\cbRGB\expandafter{\detokenize{255,255,255}}{of\strut} \setlength{\fboxsep}{0pt}\def\cbRGB{\colorbox[RGB]}\expandafter\cbRGB\expandafter{\detokenize{255,255,255}}{bottles\strut} \setlength{\fboxsep}{0pt}\def\cbRGB{\colorbox[RGB]}\expandafter\cbRGB\expandafter{\detokenize{255,255,255}}{an\strut} \setlength{\fboxsep}{0pt}\def\cbRGB{\colorbox[RGB]}\expandafter\cbRGB\expandafter{\detokenize{255,255,255}}{tried\strut} \setlength{\fboxsep}{0pt}\def\cbRGB{\colorbox[RGB]}\expandafter\cbRGB\expandafter{\detokenize{255,255,255}}{them\strut} \setlength{\fboxsep}{0pt}\def\cbRGB{\colorbox[RGB]}\expandafter\cbRGB\expandafter{\detokenize{255,255,255}}{so\strut} \setlength{\fboxsep}{0pt}\def\cbRGB{\colorbox[RGB]}\expandafter\cbRGB\expandafter{\detokenize{255,255,255}}{if\strut} \setlength{\fboxsep}{0pt}\def\cbRGB{\colorbox[RGB]}\expandafter\cbRGB\expandafter{\detokenize{255,255,255}}{they\strut} \setlength{\fboxsep}{0pt}\def\cbRGB{\colorbox[RGB]}\expandafter\cbRGB\expandafter{\detokenize{255,255,255}}{were\strut} \setlength{\fboxsep}{0pt}\def\cbRGB{\colorbox[RGB]}\expandafter\cbRGB\expandafter{\detokenize{255,255,255}}{skunked\strut} \setlength{\fboxsep}{0pt}\def\cbRGB{\colorbox[RGB]}\expandafter\cbRGB\expandafter{\detokenize{255,255,255}}{,\strut} \setlength{\fboxsep}{0pt}\def\cbRGB{\colorbox[RGB]}\expandafter\cbRGB\expandafter{\detokenize{255,255,255}}{it\strut} \setlength{\fboxsep}{0pt}\def\cbRGB{\colorbox[RGB]}\expandafter\cbRGB\expandafter{\detokenize{255,255,255}}{was\strut} \setlength{\fboxsep}{0pt}\def\cbRGB{\colorbox[RGB]}\expandafter\cbRGB\expandafter{\detokenize{255,255,255}}{unk\strut} \setlength{\fboxsep}{0pt}\def\cbRGB{\colorbox[RGB]}\expandafter\cbRGB\expandafter{\detokenize{255,255,255}}{unk\strut} \setlength{\fboxsep}{0pt}\def\cbRGB{\colorbox[RGB]}\expandafter\cbRGB\expandafter{\detokenize{255,255,255}}{that\strut} \setlength{\fboxsep}{0pt}\def\cbRGB{\colorbox[RGB]}\expandafter\cbRGB\expandafter{\detokenize{255,255,255}}{were\strut} \setlength{\fboxsep}{0pt}\def\cbRGB{\colorbox[RGB]}\expandafter\cbRGB\expandafter{\detokenize{255,255,255}}{bad\strut} \setlength{\fboxsep}{0pt}\def\cbRGB{\colorbox[RGB]}\expandafter\cbRGB\expandafter{\detokenize{255,255,255}}{.\strut} \setlength{\fboxsep}{0pt}\def\cbRGB{\colorbox[RGB]}\expandafter\cbRGB\expandafter{\detokenize{255,255,255}}{there\strut} \setlength{\fboxsep}{0pt}\def\cbRGB{\colorbox[RGB]}\expandafter\cbRGB\expandafter{\detokenize{222,222,255}}{was\strut} \setlength{\fboxsep}{0pt}\def\cbRGB{\colorbox[RGB]}\expandafter\cbRGB\expandafter{\detokenize{219,219,255}}{no\strut} \setlength{\fboxsep}{0pt}\def\cbRGB{\colorbox[RGB]}\expandafter\cbRGB\expandafter{\detokenize{233,233,255}}{real\strut} \setlength{\fboxsep}{0pt}\def\cbRGB{\colorbox[RGB]}\expandafter\cbRGB\expandafter{\detokenize{236,236,255}}{apricot\strut} \setlength{\fboxsep}{0pt}\def\cbRGB{\colorbox[RGB]}\expandafter\cbRGB\expandafter{\detokenize{239,239,255}}{taste\strut} \setlength{\fboxsep}{0pt}\def\cbRGB{\colorbox[RGB]}\expandafter\cbRGB\expandafter{\detokenize{242,242,255}}{,\strut} \setlength{\fboxsep}{0pt}\def\cbRGB{\colorbox[RGB]}\expandafter\cbRGB\expandafter{\detokenize{228,228,255}}{and\strut} \setlength{\fboxsep}{0pt}\def\cbRGB{\colorbox[RGB]}\expandafter\cbRGB\expandafter{\detokenize{231,231,255}}{no\strut} \setlength{\fboxsep}{0pt}\def\cbRGB{\colorbox[RGB]}\expandafter\cbRGB\expandafter{\detokenize{255,255,255}}{sweetness\strut} \setlength{\fboxsep}{0pt}\def\cbRGB{\colorbox[RGB]}\expandafter\cbRGB\expandafter{\detokenize{255,255,255}}{.\strut} \setlength{\fboxsep}{0pt}\def\cbRGB{\colorbox[RGB]}\expandafter\cbRGB\expandafter{\detokenize{255,255,255}}{it\strut} \setlength{\fboxsep}{0pt}\def\cbRGB{\colorbox[RGB]}\expandafter\cbRGB\expandafter{\detokenize{255,255,255}}{tasted\strut} \setlength{\fboxsep}{0pt}\def\cbRGB{\colorbox[RGB]}\expandafter\cbRGB\expandafter{\detokenize{255,255,255}}{like\strut} \setlength{\fboxsep}{0pt}\def\cbRGB{\colorbox[RGB]}\expandafter\cbRGB\expandafter{\detokenize{255,255,255}}{hay\strut} \setlength{\fboxsep}{0pt}\def\cbRGB{\colorbox[RGB]}\expandafter\cbRGB\expandafter{\detokenize{255,255,255}}{but\strut} \setlength{\fboxsep}{0pt}\def\cbRGB{\colorbox[RGB]}\expandafter\cbRGB\expandafter{\detokenize{255,255,255}}{not\strut} \setlength{\fboxsep}{0pt}\def\cbRGB{\colorbox[RGB]}\expandafter\cbRGB\expandafter{\detokenize{255,255,255}}{in\strut} \setlength{\fboxsep}{0pt}\def\cbRGB{\colorbox[RGB]}\expandafter\cbRGB\expandafter{\detokenize{255,255,255}}{the\strut} \setlength{\fboxsep}{0pt}\def\cbRGB{\colorbox[RGB]}\expandafter\cbRGB\expandafter{\detokenize{255,255,255}}{good\strut} \setlength{\fboxsep}{0pt}\def\cbRGB{\colorbox[RGB]}\expandafter\cbRGB\expandafter{\detokenize{255,255,255}}{wheat\strut} \setlength{\fboxsep}{0pt}\def\cbRGB{\colorbox[RGB]}\expandafter\cbRGB\expandafter{\detokenize{255,255,255}}{beer\strut} \setlength{\fboxsep}{0pt}\def\cbRGB{\colorbox[RGB]}\expandafter\cbRGB\expandafter{\detokenize{255,255,255}}{sense\strut} \setlength{\fboxsep}{0pt}\def\cbRGB{\colorbox[RGB]}\expandafter\cbRGB\expandafter{\detokenize{255,255,255}}{,\strut} \setlength{\fboxsep}{0pt}\def\cbRGB{\colorbox[RGB]}\expandafter\cbRGB\expandafter{\detokenize{255,255,255}}{more\strut} \setlength{\fboxsep}{0pt}\def\cbRGB{\colorbox[RGB]}\expandafter\cbRGB\expandafter{\detokenize{255,255,255}}{like\strut} \setlength{\fboxsep}{0pt}\def\cbRGB{\colorbox[RGB]}\expandafter\cbRGB\expandafter{\detokenize{226,226,255}}{dirty\strut} \setlength{\fboxsep}{0pt}\def\cbRGB{\colorbox[RGB]}\expandafter\cbRGB\expandafter{\detokenize{206,206,255}}{hay\strut} \setlength{\fboxsep}{0pt}\def\cbRGB{\colorbox[RGB]}\expandafter\cbRGB\expandafter{\detokenize{204,204,255}}{thats\strut} \setlength{\fboxsep}{0pt}\def\cbRGB{\colorbox[RGB]}\expandafter\cbRGB\expandafter{\detokenize{212,212,255}}{been\strut} \setlength{\fboxsep}{0pt}\def\cbRGB{\colorbox[RGB]}\expandafter\cbRGB\expandafter{\detokenize{220,220,255}}{under\strut} \setlength{\fboxsep}{0pt}\def\cbRGB{\colorbox[RGB]}\expandafter\cbRGB\expandafter{\detokenize{227,227,255}}{the\strut} \setlength{\fboxsep}{0pt}\def\cbRGB{\colorbox[RGB]}\expandafter\cbRGB\expandafter{\detokenize{209,209,255}}{budweiser\strut} \setlength{\fboxsep}{0pt}\def\cbRGB{\colorbox[RGB]}\expandafter\cbRGB\expandafter{\detokenize{187,187,255}}{unk\strut} \setlength{\fboxsep}{0pt}\def\cbRGB{\colorbox[RGB]}\expandafter\cbRGB\expandafter{\detokenize{169,169,255}}{.\strut} \setlength{\fboxsep}{0pt}\def\cbRGB{\colorbox[RGB]}\expandafter\cbRGB\expandafter{\detokenize{134,134,255}}{mouthfeel\strut} \setlength{\fboxsep}{0pt}\def\cbRGB{\colorbox[RGB]}\expandafter\cbRGB\expandafter{\detokenize{82,82,255}}{was\strut} \setlength{\fboxsep}{0pt}\def\cbRGB{\colorbox[RGB]}\expandafter\cbRGB\expandafter{\detokenize{47,47,255}}{thin\strut} \setlength{\fboxsep}{0pt}\def\cbRGB{\colorbox[RGB]}\expandafter\cbRGB\expandafter{\detokenize{16,16,255}}{and\strut} \setlength{\fboxsep}{0pt}\def\cbRGB{\colorbox[RGB]}\expandafter\cbRGB\expandafter{\detokenize{0,0,255}}{carbonated\strut} \setlength{\fboxsep}{0pt}\def\cbRGB{\colorbox[RGB]}\expandafter\cbRGB\expandafter{\detokenize{16,16,255}}{.\strut} \setlength{\fboxsep}{0pt}\def\cbRGB{\colorbox[RGB]}\expandafter\cbRGB\expandafter{\detokenize{66,66,255}}{drinkability\strut} \setlength{\fboxsep}{0pt}\def\cbRGB{\colorbox[RGB]}\expandafter\cbRGB\expandafter{\detokenize{116,116,255}}{{\ldots}\strut} \setlength{\fboxsep}{0pt}\def\cbRGB{\colorbox[RGB]}\expandafter\cbRGB\expandafter{\detokenize{162,162,255}}{well\strut} \setlength{\fboxsep}{0pt}\def\cbRGB{\colorbox[RGB]}\expandafter\cbRGB\expandafter{\detokenize{182,182,255}}{this\strut} \setlength{\fboxsep}{0pt}\def\cbRGB{\colorbox[RGB]}\expandafter\cbRGB\expandafter{\detokenize{180,180,255}}{is\strut} \setlength{\fboxsep}{0pt}\def\cbRGB{\colorbox[RGB]}\expandafter\cbRGB\expandafter{\detokenize{153,153,255}}{the\strut} \setlength{\fboxsep}{0pt}\def\cbRGB{\colorbox[RGB]}\expandafter\cbRGB\expandafter{\detokenize{132,132,255}}{second\strut} \setlength{\fboxsep}{0pt}\def\cbRGB{\colorbox[RGB]}\expandafter\cbRGB\expandafter{\detokenize{129,129,255}}{beer\strut} \setlength{\fboxsep}{0pt}\def\cbRGB{\colorbox[RGB]}\expandafter\cbRGB\expandafter{\detokenize{152,152,255}}{in\strut} \setlength{\fboxsep}{0pt}\def\cbRGB{\colorbox[RGB]}\expandafter\cbRGB\expandafter{\detokenize{184,184,255}}{my\strut} \setlength{\fboxsep}{0pt}\def\cbRGB{\colorbox[RGB]}\expandafter\cbRGB\expandafter{\detokenize{232,232,255}}{long\strut} \setlength{\fboxsep}{0pt}\def\cbRGB{\colorbox[RGB]}\expandafter\cbRGB\expandafter{\detokenize{255,255,255}}{list\strut} \setlength{\fboxsep}{0pt}\def\cbRGB{\colorbox[RGB]}\expandafter\cbRGB\expandafter{\detokenize{255,255,255}}{that\strut} \setlength{\fboxsep}{0pt}\def\cbRGB{\colorbox[RGB]}\expandafter\cbRGB\expandafter{\detokenize{255,255,255}}{i\strut} \setlength{\fboxsep}{0pt}\def\cbRGB{\colorbox[RGB]}\expandafter\cbRGB\expandafter{\detokenize{255,255,255}}{'ve\strut} \setlength{\fboxsep}{0pt}\def\cbRGB{\colorbox[RGB]}\expandafter\cbRGB\expandafter{\detokenize{255,255,255}}{actually\strut} \setlength{\fboxsep}{0pt}\def\cbRGB{\colorbox[RGB]}\expandafter\cbRGB\expandafter{\detokenize{255,255,255}}{had\strut} \setlength{\fboxsep}{0pt}\def\cbRGB{\colorbox[RGB]}\expandafter\cbRGB\expandafter{\detokenize{255,255,255}}{to\strut} \setlength{\fboxsep}{0pt}\def\cbRGB{\colorbox[RGB]}\expandafter\cbRGB\expandafter{\detokenize{255,255,255}}{pour\strut} \setlength{\fboxsep}{0pt}\def\cbRGB{\colorbox[RGB]}\expandafter\cbRGB\expandafter{\detokenize{255,255,255}}{out\strut} \setlength{\fboxsep}{0pt}\def\cbRGB{\colorbox[RGB]}\expandafter\cbRGB\expandafter{\detokenize{255,255,255}}{.\strut} 

\setlength{\fboxsep}{0pt}\def\cbRGB{\colorbox[RGB]}\expandafter\cbRGB\expandafter{\detokenize{255,255,44}}{appearance\strut} \setlength{\fboxsep}{0pt}\def\cbRGB{\colorbox[RGB]}\expandafter\cbRGB\expandafter{\detokenize{255,255,171}}{was\strut} \setlength{\fboxsep}{0pt}\def\cbRGB{\colorbox[RGB]}\expandafter\cbRGB\expandafter{\detokenize{255,255,253}}{gold\strut} \setlength{\fboxsep}{0pt}\def\cbRGB{\colorbox[RGB]}\expandafter\cbRGB\expandafter{\detokenize{255,255,255}}{,\strut} \setlength{\fboxsep}{0pt}\def\cbRGB{\colorbox[RGB]}\expandafter\cbRGB\expandafter{\detokenize{255,255,255}}{clear\strut} \setlength{\fboxsep}{0pt}\def\cbRGB{\colorbox[RGB]}\expandafter\cbRGB\expandafter{\detokenize{255,255,255}}{,\strut} \setlength{\fboxsep}{0pt}\def\cbRGB{\colorbox[RGB]}\expandafter\cbRGB\expandafter{\detokenize{255,255,255}}{no\strut} \setlength{\fboxsep}{0pt}\def\cbRGB{\colorbox[RGB]}\expandafter\cbRGB\expandafter{\detokenize{255,255,255}}{head\strut} \setlength{\fboxsep}{0pt}\def\cbRGB{\colorbox[RGB]}\expandafter\cbRGB\expandafter{\detokenize{255,255,255}}{,\strut} \setlength{\fboxsep}{0pt}\def\cbRGB{\colorbox[RGB]}\expandafter\cbRGB\expandafter{\detokenize{255,255,255}}{and\strut} \setlength{\fboxsep}{0pt}\def\cbRGB{\colorbox[RGB]}\expandafter\cbRGB\expandafter{\detokenize{255,255,255}}{fizzy\strut} \setlength{\fboxsep}{0pt}\def\cbRGB{\colorbox[RGB]}\expandafter\cbRGB\expandafter{\detokenize{255,255,255}}{.\strut} \setlength{\fboxsep}{0pt}\def\cbRGB{\colorbox[RGB]}\expandafter\cbRGB\expandafter{\detokenize{255,255,255}}{almost\strut} \setlength{\fboxsep}{0pt}\def\cbRGB{\colorbox[RGB]}\expandafter\cbRGB\expandafter{\detokenize{255,255,255}}{looked\strut} \setlength{\fboxsep}{0pt}\def\cbRGB{\colorbox[RGB]}\expandafter\cbRGB\expandafter{\detokenize{255,255,255}}{like\strut} \setlength{\fboxsep}{0pt}\def\cbRGB{\colorbox[RGB]}\expandafter\cbRGB\expandafter{\detokenize{255,255,255}}{champagne\strut} \setlength{\fboxsep}{0pt}\def\cbRGB{\colorbox[RGB]}\expandafter\cbRGB\expandafter{\detokenize{255,255,255}}{.\strut} \setlength{\fboxsep}{0pt}\def\cbRGB{\colorbox[RGB]}\expandafter\cbRGB\expandafter{\detokenize{255,255,238}}{aroma\strut} \setlength{\fboxsep}{0pt}\def\cbRGB{\colorbox[RGB]}\expandafter\cbRGB\expandafter{\detokenize{255,255,160}}{was\strut} \setlength{\fboxsep}{0pt}\def\cbRGB{\colorbox[RGB]}\expandafter\cbRGB\expandafter{\detokenize{255,255,82}}{actually\strut} \setlength{\fboxsep}{0pt}\def\cbRGB{\colorbox[RGB]}\expandafter\cbRGB\expandafter{\detokenize{255,255,67}}{real\strut} \setlength{\fboxsep}{0pt}\def\cbRGB{\colorbox[RGB]}\expandafter\cbRGB\expandafter{\detokenize{255,255,114}}{good\strut} \setlength{\fboxsep}{0pt}\def\cbRGB{\colorbox[RGB]}\expandafter\cbRGB\expandafter{\detokenize{255,255,191}}{.\strut} \setlength{\fboxsep}{0pt}\def\cbRGB{\colorbox[RGB]}\expandafter\cbRGB\expandafter{\detokenize{255,255,233}}{sweet\strut} \setlength{\fboxsep}{0pt}\def\cbRGB{\colorbox[RGB]}\expandafter\cbRGB\expandafter{\detokenize{255,255,255}}{apricot\strut} \setlength{\fboxsep}{0pt}\def\cbRGB{\colorbox[RGB]}\expandafter\cbRGB\expandafter{\detokenize{255,255,255}}{,\strut} \setlength{\fboxsep}{0pt}\def\cbRGB{\colorbox[RGB]}\expandafter\cbRGB\expandafter{\detokenize{255,255,238}}{unk\strut} \setlength{\fboxsep}{0pt}\def\cbRGB{\colorbox[RGB]}\expandafter\cbRGB\expandafter{\detokenize{255,255,208}}{smith\strut} \setlength{\fboxsep}{0pt}\def\cbRGB{\colorbox[RGB]}\expandafter\cbRGB\expandafter{\detokenize{255,255,237}}{apple\strut} \setlength{\fboxsep}{0pt}\def\cbRGB{\colorbox[RGB]}\expandafter\cbRGB\expandafter{\detokenize{255,255,255}}{,\strut} \setlength{\fboxsep}{0pt}\def\cbRGB{\colorbox[RGB]}\expandafter\cbRGB\expandafter{\detokenize{255,255,255}}{almost\strut} \setlength{\fboxsep}{0pt}\def\cbRGB{\colorbox[RGB]}\expandafter\cbRGB\expandafter{\detokenize{255,255,255}}{like\strut} \setlength{\fboxsep}{0pt}\def\cbRGB{\colorbox[RGB]}\expandafter\cbRGB\expandafter{\detokenize{255,255,255}}{a\strut} \setlength{\fboxsep}{0pt}\def\cbRGB{\colorbox[RGB]}\expandafter\cbRGB\expandafter{\detokenize{255,255,255}}{unk\strut} \setlength{\fboxsep}{0pt}\def\cbRGB{\colorbox[RGB]}\expandafter\cbRGB\expandafter{\detokenize{255,255,236}}{wine\strut} \setlength{\fboxsep}{0pt}\def\cbRGB{\colorbox[RGB]}\expandafter\cbRGB\expandafter{\detokenize{255,255,208}}{.\strut} \setlength{\fboxsep}{0pt}\def\cbRGB{\colorbox[RGB]}\expandafter\cbRGB\expandafter{\detokenize{255,255,169}}{taste\strut} \setlength{\fboxsep}{0pt}\def\cbRGB{\colorbox[RGB]}\expandafter\cbRGB\expandafter{\detokenize{255,255,134}}{was\strut} \setlength{\fboxsep}{0pt}\def\cbRGB{\colorbox[RGB]}\expandafter\cbRGB\expandafter{\detokenize{255,255,102}}{unk\strut} \setlength{\fboxsep}{0pt}\def\cbRGB{\colorbox[RGB]}\expandafter\cbRGB\expandafter{\detokenize{255,255,126}}{.\strut} \setlength{\fboxsep}{0pt}\def\cbRGB{\colorbox[RGB]}\expandafter\cbRGB\expandafter{\detokenize{255,255,166}}{i\strut} \setlength{\fboxsep}{0pt}\def\cbRGB{\colorbox[RGB]}\expandafter\cbRGB\expandafter{\detokenize{255,255,227}}{unk\strut} \setlength{\fboxsep}{0pt}\def\cbRGB{\colorbox[RGB]}\expandafter\cbRGB\expandafter{\detokenize{255,255,255}}{a\strut} \setlength{\fboxsep}{0pt}\def\cbRGB{\colorbox[RGB]}\expandafter\cbRGB\expandafter{\detokenize{255,255,255}}{couple\strut} \setlength{\fboxsep}{0pt}\def\cbRGB{\colorbox[RGB]}\expandafter\cbRGB\expandafter{\detokenize{255,255,255}}{of\strut} \setlength{\fboxsep}{0pt}\def\cbRGB{\colorbox[RGB]}\expandafter\cbRGB\expandafter{\detokenize{255,255,255}}{bottles\strut} \setlength{\fboxsep}{0pt}\def\cbRGB{\colorbox[RGB]}\expandafter\cbRGB\expandafter{\detokenize{255,255,255}}{an\strut} \setlength{\fboxsep}{0pt}\def\cbRGB{\colorbox[RGB]}\expandafter\cbRGB\expandafter{\detokenize{255,255,255}}{tried\strut} \setlength{\fboxsep}{0pt}\def\cbRGB{\colorbox[RGB]}\expandafter\cbRGB\expandafter{\detokenize{255,255,255}}{them\strut} \setlength{\fboxsep}{0pt}\def\cbRGB{\colorbox[RGB]}\expandafter\cbRGB\expandafter{\detokenize{255,255,255}}{so\strut} \setlength{\fboxsep}{0pt}\def\cbRGB{\colorbox[RGB]}\expandafter\cbRGB\expandafter{\detokenize{255,255,255}}{if\strut} \setlength{\fboxsep}{0pt}\def\cbRGB{\colorbox[RGB]}\expandafter\cbRGB\expandafter{\detokenize{255,255,255}}{they\strut} \setlength{\fboxsep}{0pt}\def\cbRGB{\colorbox[RGB]}\expandafter\cbRGB\expandafter{\detokenize{255,255,255}}{were\strut} \setlength{\fboxsep}{0pt}\def\cbRGB{\colorbox[RGB]}\expandafter\cbRGB\expandafter{\detokenize{255,255,255}}{skunked\strut} \setlength{\fboxsep}{0pt}\def\cbRGB{\colorbox[RGB]}\expandafter\cbRGB\expandafter{\detokenize{255,255,255}}{,\strut} \setlength{\fboxsep}{0pt}\def\cbRGB{\colorbox[RGB]}\expandafter\cbRGB\expandafter{\detokenize{255,255,255}}{it\strut} \setlength{\fboxsep}{0pt}\def\cbRGB{\colorbox[RGB]}\expandafter\cbRGB\expandafter{\detokenize{255,255,255}}{was\strut} \setlength{\fboxsep}{0pt}\def\cbRGB{\colorbox[RGB]}\expandafter\cbRGB\expandafter{\detokenize{255,255,255}}{unk\strut} \setlength{\fboxsep}{0pt}\def\cbRGB{\colorbox[RGB]}\expandafter\cbRGB\expandafter{\detokenize{255,255,255}}{unk\strut} \setlength{\fboxsep}{0pt}\def\cbRGB{\colorbox[RGB]}\expandafter\cbRGB\expandafter{\detokenize{255,255,255}}{that\strut} \setlength{\fboxsep}{0pt}\def\cbRGB{\colorbox[RGB]}\expandafter\cbRGB\expandafter{\detokenize{255,255,255}}{were\strut} \setlength{\fboxsep}{0pt}\def\cbRGB{\colorbox[RGB]}\expandafter\cbRGB\expandafter{\detokenize{255,255,255}}{bad\strut} \setlength{\fboxsep}{0pt}\def\cbRGB{\colorbox[RGB]}\expandafter\cbRGB\expandafter{\detokenize{255,255,255}}{.\strut} \setlength{\fboxsep}{0pt}\def\cbRGB{\colorbox[RGB]}\expandafter\cbRGB\expandafter{\detokenize{255,255,242}}{there\strut} \setlength{\fboxsep}{0pt}\def\cbRGB{\colorbox[RGB]}\expandafter\cbRGB\expandafter{\detokenize{255,255,183}}{was\strut} \setlength{\fboxsep}{0pt}\def\cbRGB{\colorbox[RGB]}\expandafter\cbRGB\expandafter{\detokenize{255,255,146}}{no\strut} \setlength{\fboxsep}{0pt}\def\cbRGB{\colorbox[RGB]}\expandafter\cbRGB\expandafter{\detokenize{255,255,145}}{real\strut} \setlength{\fboxsep}{0pt}\def\cbRGB{\colorbox[RGB]}\expandafter\cbRGB\expandafter{\detokenize{255,255,132}}{apricot\strut} \setlength{\fboxsep}{0pt}\def\cbRGB{\colorbox[RGB]}\expandafter\cbRGB\expandafter{\detokenize{255,255,118}}{taste\strut} \setlength{\fboxsep}{0pt}\def\cbRGB{\colorbox[RGB]}\expandafter\cbRGB\expandafter{\detokenize{255,255,103}}{,\strut} \setlength{\fboxsep}{0pt}\def\cbRGB{\colorbox[RGB]}\expandafter\cbRGB\expandafter{\detokenize{255,255,103}}{and\strut} \setlength{\fboxsep}{0pt}\def\cbRGB{\colorbox[RGB]}\expandafter\cbRGB\expandafter{\detokenize{255,255,98}}{no\strut} \setlength{\fboxsep}{0pt}\def\cbRGB{\colorbox[RGB]}\expandafter\cbRGB\expandafter{\detokenize{255,255,112}}{sweetness\strut} \setlength{\fboxsep}{0pt}\def\cbRGB{\colorbox[RGB]}\expandafter\cbRGB\expandafter{\detokenize{255,255,102}}{.\strut} \setlength{\fboxsep}{0pt}\def\cbRGB{\colorbox[RGB]}\expandafter\cbRGB\expandafter{\detokenize{255,255,87}}{it\strut} \setlength{\fboxsep}{0pt}\def\cbRGB{\colorbox[RGB]}\expandafter\cbRGB\expandafter{\detokenize{255,255,53}}{tasted\strut} \setlength{\fboxsep}{0pt}\def\cbRGB{\colorbox[RGB]}\expandafter\cbRGB\expandafter{\detokenize{255,255,55}}{like\strut} \setlength{\fboxsep}{0pt}\def\cbRGB{\colorbox[RGB]}\expandafter\cbRGB\expandafter{\detokenize{255,255,88}}{hay\strut} \setlength{\fboxsep}{0pt}\def\cbRGB{\colorbox[RGB]}\expandafter\cbRGB\expandafter{\detokenize{255,255,172}}{but\strut} \setlength{\fboxsep}{0pt}\def\cbRGB{\colorbox[RGB]}\expandafter\cbRGB\expandafter{\detokenize{255,255,255}}{not\strut} \setlength{\fboxsep}{0pt}\def\cbRGB{\colorbox[RGB]}\expandafter\cbRGB\expandafter{\detokenize{255,255,255}}{in\strut} \setlength{\fboxsep}{0pt}\def\cbRGB{\colorbox[RGB]}\expandafter\cbRGB\expandafter{\detokenize{255,255,255}}{the\strut} \setlength{\fboxsep}{0pt}\def\cbRGB{\colorbox[RGB]}\expandafter\cbRGB\expandafter{\detokenize{255,255,255}}{good\strut} \setlength{\fboxsep}{0pt}\def\cbRGB{\colorbox[RGB]}\expandafter\cbRGB\expandafter{\detokenize{255,255,255}}{wheat\strut} \setlength{\fboxsep}{0pt}\def\cbRGB{\colorbox[RGB]}\expandafter\cbRGB\expandafter{\detokenize{255,255,255}}{beer\strut} \setlength{\fboxsep}{0pt}\def\cbRGB{\colorbox[RGB]}\expandafter\cbRGB\expandafter{\detokenize{255,255,255}}{sense\strut} \setlength{\fboxsep}{0pt}\def\cbRGB{\colorbox[RGB]}\expandafter\cbRGB\expandafter{\detokenize{255,255,255}}{,\strut} \setlength{\fboxsep}{0pt}\def\cbRGB{\colorbox[RGB]}\expandafter\cbRGB\expandafter{\detokenize{255,255,236}}{more\strut} \setlength{\fboxsep}{0pt}\def\cbRGB{\colorbox[RGB]}\expandafter\cbRGB\expandafter{\detokenize{255,255,189}}{like\strut} \setlength{\fboxsep}{0pt}\def\cbRGB{\colorbox[RGB]}\expandafter\cbRGB\expandafter{\detokenize{255,255,139}}{dirty\strut} \setlength{\fboxsep}{0pt}\def\cbRGB{\colorbox[RGB]}\expandafter\cbRGB\expandafter{\detokenize{255,255,130}}{hay\strut} \setlength{\fboxsep}{0pt}\def\cbRGB{\colorbox[RGB]}\expandafter\cbRGB\expandafter{\detokenize{255,255,130}}{thats\strut} \setlength{\fboxsep}{0pt}\def\cbRGB{\colorbox[RGB]}\expandafter\cbRGB\expandafter{\detokenize{255,255,133}}{been\strut} \setlength{\fboxsep}{0pt}\def\cbRGB{\colorbox[RGB]}\expandafter\cbRGB\expandafter{\detokenize{255,255,136}}{under\strut} \setlength{\fboxsep}{0pt}\def\cbRGB{\colorbox[RGB]}\expandafter\cbRGB\expandafter{\detokenize{255,255,156}}{the\strut} \setlength{\fboxsep}{0pt}\def\cbRGB{\colorbox[RGB]}\expandafter\cbRGB\expandafter{\detokenize{255,255,174}}{budweiser\strut} \setlength{\fboxsep}{0pt}\def\cbRGB{\colorbox[RGB]}\expandafter\cbRGB\expandafter{\detokenize{255,255,192}}{unk\strut} \setlength{\fboxsep}{0pt}\def\cbRGB{\colorbox[RGB]}\expandafter\cbRGB\expandafter{\detokenize{255,255,200}}{.\strut} \setlength{\fboxsep}{0pt}\def\cbRGB{\colorbox[RGB]}\expandafter\cbRGB\expandafter{\detokenize{255,255,202}}{mouthfeel\strut} \setlength{\fboxsep}{0pt}\def\cbRGB{\colorbox[RGB]}\expandafter\cbRGB\expandafter{\detokenize{255,255,163}}{was\strut} \setlength{\fboxsep}{0pt}\def\cbRGB{\colorbox[RGB]}\expandafter\cbRGB\expandafter{\detokenize{255,255,124}}{thin\strut} \setlength{\fboxsep}{0pt}\def\cbRGB{\colorbox[RGB]}\expandafter\cbRGB\expandafter{\detokenize{255,255,117}}{and\strut} \setlength{\fboxsep}{0pt}\def\cbRGB{\colorbox[RGB]}\expandafter\cbRGB\expandafter{\detokenize{255,255,98}}{carbonated\strut} \setlength{\fboxsep}{0pt}\def\cbRGB{\colorbox[RGB]}\expandafter\cbRGB\expandafter{\detokenize{255,255,89}}{.\strut} \setlength{\fboxsep}{0pt}\def\cbRGB{\colorbox[RGB]}\expandafter\cbRGB\expandafter{\detokenize{255,255,142}}{drinkability\strut} \setlength{\fboxsep}{0pt}\def\cbRGB{\colorbox[RGB]}\expandafter\cbRGB\expandafter{\detokenize{255,255,163}}{{\ldots}\strut} \setlength{\fboxsep}{0pt}\def\cbRGB{\colorbox[RGB]}\expandafter\cbRGB\expandafter{\detokenize{255,255,120}}{well\strut} \setlength{\fboxsep}{0pt}\def\cbRGB{\colorbox[RGB]}\expandafter\cbRGB\expandafter{\detokenize{255,255,108}}{this\strut} \setlength{\fboxsep}{0pt}\def\cbRGB{\colorbox[RGB]}\expandafter\cbRGB\expandafter{\detokenize{255,255,82}}{is\strut} \setlength{\fboxsep}{0pt}\def\cbRGB{\colorbox[RGB]}\expandafter\cbRGB\expandafter{\detokenize{255,255,25}}{the\strut} \setlength{\fboxsep}{0pt}\def\cbRGB{\colorbox[RGB]}\expandafter\cbRGB\expandafter{\detokenize{255,255,0}}{second\strut} \setlength{\fboxsep}{0pt}\def\cbRGB{\colorbox[RGB]}\expandafter\cbRGB\expandafter{\detokenize{255,255,15}}{beer\strut} \setlength{\fboxsep}{0pt}\def\cbRGB{\colorbox[RGB]}\expandafter\cbRGB\expandafter{\detokenize{255,255,43}}{in\strut} \setlength{\fboxsep}{0pt}\def\cbRGB{\colorbox[RGB]}\expandafter\cbRGB\expandafter{\detokenize{255,255,86}}{my\strut} \setlength{\fboxsep}{0pt}\def\cbRGB{\colorbox[RGB]}\expandafter\cbRGB\expandafter{\detokenize{255,255,135}}{long\strut} \setlength{\fboxsep}{0pt}\def\cbRGB{\colorbox[RGB]}\expandafter\cbRGB\expandafter{\detokenize{255,255,194}}{list\strut} \setlength{\fboxsep}{0pt}\def\cbRGB{\colorbox[RGB]}\expandafter\cbRGB\expandafter{\detokenize{255,255,255}}{that\strut} \setlength{\fboxsep}{0pt}\def\cbRGB{\colorbox[RGB]}\expandafter\cbRGB\expandafter{\detokenize{255,255,255}}{i\strut} \setlength{\fboxsep}{0pt}\def\cbRGB{\colorbox[RGB]}\expandafter\cbRGB\expandafter{\detokenize{255,255,255}}{'ve\strut} \setlength{\fboxsep}{0pt}\def\cbRGB{\colorbox[RGB]}\expandafter\cbRGB\expandafter{\detokenize{255,255,255}}{actually\strut} \setlength{\fboxsep}{0pt}\def\cbRGB{\colorbox[RGB]}\expandafter\cbRGB\expandafter{\detokenize{255,255,255}}{had\strut} \setlength{\fboxsep}{0pt}\def\cbRGB{\colorbox[RGB]}\expandafter\cbRGB\expandafter{\detokenize{255,255,255}}{to\strut} \setlength{\fboxsep}{0pt}\def\cbRGB{\colorbox[RGB]}\expandafter\cbRGB\expandafter{\detokenize{255,255,255}}{pour\strut} \setlength{\fboxsep}{0pt}\def\cbRGB{\colorbox[RGB]}\expandafter\cbRGB\expandafter{\detokenize{255,255,255}}{out\strut} \setlength{\fboxsep}{0pt}\def\cbRGB{\colorbox[RGB]}\expandafter\cbRGB\expandafter{\detokenize{255,255,255}}{.\strut} 

\par
\textbf{Example 2}

\setlength{\fboxsep}{0pt}\def\cbRGB{\colorbox[RGB]}\expandafter\cbRGB\expandafter{\detokenize{255,171,171}}{poured\strut} \setlength{\fboxsep}{0pt}\def\cbRGB{\colorbox[RGB]}\expandafter\cbRGB\expandafter{\detokenize{255,144,144}}{into\strut} \setlength{\fboxsep}{0pt}\def\cbRGB{\colorbox[RGB]}\expandafter\cbRGB\expandafter{\detokenize{255,117,117}}{a\strut} \setlength{\fboxsep}{0pt}\def\cbRGB{\colorbox[RGB]}\expandafter\cbRGB\expandafter{\detokenize{255,139,139}}{nonic\strut} \setlength{\fboxsep}{0pt}\def\cbRGB{\colorbox[RGB]}\expandafter\cbRGB\expandafter{\detokenize{255,174,174}}{pint\strut} \setlength{\fboxsep}{0pt}\def\cbRGB{\colorbox[RGB]}\expandafter\cbRGB\expandafter{\detokenize{255,196,196}}{glass\strut} \setlength{\fboxsep}{0pt}\def\cbRGB{\colorbox[RGB]}\expandafter\cbRGB\expandafter{\detokenize{255,209,209}}{{\ldots}\strut} \setlength{\fboxsep}{0pt}\def\cbRGB{\colorbox[RGB]}\expandafter\cbRGB\expandafter{\detokenize{255,223,223}}{jet\strut} \setlength{\fboxsep}{0pt}\def\cbRGB{\colorbox[RGB]}\expandafter\cbRGB\expandafter{\detokenize{255,236,236}}{black\strut} \setlength{\fboxsep}{0pt}\def\cbRGB{\colorbox[RGB]}\expandafter\cbRGB\expandafter{\detokenize{255,237,237}}{,\strut} \setlength{\fboxsep}{0pt}\def\cbRGB{\colorbox[RGB]}\expandafter\cbRGB\expandafter{\detokenize{255,237,237}}{unk\strut} \setlength{\fboxsep}{0pt}\def\cbRGB{\colorbox[RGB]}\expandafter\cbRGB\expandafter{\detokenize{255,237,237}}{no\strut} \setlength{\fboxsep}{0pt}\def\cbRGB{\colorbox[RGB]}\expandafter\cbRGB\expandafter{\detokenize{255,250,250}}{highlights\strut} \setlength{\fboxsep}{0pt}\def\cbRGB{\colorbox[RGB]}\expandafter\cbRGB\expandafter{\detokenize{255,255,255}}{around\strut} \setlength{\fboxsep}{0pt}\def\cbRGB{\colorbox[RGB]}\expandafter\cbRGB\expandafter{\detokenize{255,255,255}}{the\strut} \setlength{\fboxsep}{0pt}\def\cbRGB{\colorbox[RGB]}\expandafter\cbRGB\expandafter{\detokenize{255,255,255}}{edges\strut} \setlength{\fboxsep}{0pt}\def\cbRGB{\colorbox[RGB]}\expandafter\cbRGB\expandafter{\detokenize{255,255,255}}{at\strut} \setlength{\fboxsep}{0pt}\def\cbRGB{\colorbox[RGB]}\expandafter\cbRGB\expandafter{\detokenize{255,223,223}}{all\strut} \setlength{\fboxsep}{0pt}\def\cbRGB{\colorbox[RGB]}\expandafter\cbRGB\expandafter{\detokenize{255,170,170}}{,\strut} \setlength{\fboxsep}{0pt}\def\cbRGB{\colorbox[RGB]}\expandafter\cbRGB\expandafter{\detokenize{255,104,104}}{dense\strut} \setlength{\fboxsep}{0pt}\def\cbRGB{\colorbox[RGB]}\expandafter\cbRGB\expandafter{\detokenize{255,52,52}}{tan\strut} \setlength{\fboxsep}{0pt}\def\cbRGB{\colorbox[RGB]}\expandafter\cbRGB\expandafter{\detokenize{255,12,12}}{unk\strut} \setlength{\fboxsep}{0pt}\def\cbRGB{\colorbox[RGB]}\expandafter\cbRGB\expandafter{\detokenize{255,0,0}}{head\strut} \setlength{\fboxsep}{0pt}\def\cbRGB{\colorbox[RGB]}\expandafter\cbRGB\expandafter{\detokenize{255,14,14}}{{\ldots}\strut} \setlength{\fboxsep}{0pt}\def\cbRGB{\colorbox[RGB]}\expandafter\cbRGB\expandafter{\detokenize{255,53,53}}{looks\strut} \setlength{\fboxsep}{0pt}\def\cbRGB{\colorbox[RGB]}\expandafter\cbRGB\expandafter{\detokenize{255,105,105}}{great\strut} \setlength{\fboxsep}{0pt}\def\cbRGB{\colorbox[RGB]}\expandafter\cbRGB\expandafter{\detokenize{255,170,170}}{!\strut} \setlength{\fboxsep}{0pt}\def\cbRGB{\colorbox[RGB]}\expandafter\cbRGB\expandafter{\detokenize{255,221,221}}{a\strut} \setlength{\fboxsep}{0pt}\def\cbRGB{\colorbox[RGB]}\expandafter\cbRGB\expandafter{\detokenize{255,255,255}}{bit\strut} \setlength{\fboxsep}{0pt}\def\cbRGB{\colorbox[RGB]}\expandafter\cbRGB\expandafter{\detokenize{255,255,255}}{of\strut} \setlength{\fboxsep}{0pt}\def\cbRGB{\colorbox[RGB]}\expandafter\cbRGB\expandafter{\detokenize{255,255,255}}{sediment\strut} \setlength{\fboxsep}{0pt}\def\cbRGB{\colorbox[RGB]}\expandafter\cbRGB\expandafter{\detokenize{255,255,255}}{in\strut} \setlength{\fboxsep}{0pt}\def\cbRGB{\colorbox[RGB]}\expandafter\cbRGB\expandafter{\detokenize{255,255,255}}{the\strut} \setlength{\fboxsep}{0pt}\def\cbRGB{\colorbox[RGB]}\expandafter\cbRGB\expandafter{\detokenize{255,255,255}}{bottom\strut} \setlength{\fboxsep}{0pt}\def\cbRGB{\colorbox[RGB]}\expandafter\cbRGB\expandafter{\detokenize{255,255,255}}{of\strut} \setlength{\fboxsep}{0pt}\def\cbRGB{\colorbox[RGB]}\expandafter\cbRGB\expandafter{\detokenize{255,255,255}}{the\strut} \setlength{\fboxsep}{0pt}\def\cbRGB{\colorbox[RGB]}\expandafter\cbRGB\expandafter{\detokenize{255,255,255}}{bottle\strut} \setlength{\fboxsep}{0pt}\def\cbRGB{\colorbox[RGB]}\expandafter\cbRGB\expandafter{\detokenize{255,255,255}}{.\strut} \setlength{\fboxsep}{0pt}\def\cbRGB{\colorbox[RGB]}\expandafter\cbRGB\expandafter{\detokenize{255,255,255}}{bottle\strut} \setlength{\fboxsep}{0pt}\def\cbRGB{\colorbox[RGB]}\expandafter\cbRGB\expandafter{\detokenize{255,255,255}}{read\strut} \setlength{\fboxsep}{0pt}\def\cbRGB{\colorbox[RGB]}\expandafter\cbRGB\expandafter{\detokenize{255,255,255}}{``\strut} \setlength{\fboxsep}{0pt}\def\cbRGB{\colorbox[RGB]}\expandafter\cbRGB\expandafter{\detokenize{255,255,255}}{live\strut} \setlength{\fboxsep}{0pt}\def\cbRGB{\colorbox[RGB]}\expandafter\cbRGB\expandafter{\detokenize{255,255,255}}{ale\strut} \setlength{\fboxsep}{0pt}\def\cbRGB{\colorbox[RGB]}\expandafter\cbRGB\expandafter{\detokenize{255,255,255}}{,\strut} \setlength{\fboxsep}{0pt}\def\cbRGB{\colorbox[RGB]}\expandafter\cbRGB\expandafter{\detokenize{255,250,250}}{keep\strut} \setlength{\fboxsep}{0pt}\def\cbRGB{\colorbox[RGB]}\expandafter\cbRGB\expandafter{\detokenize{255,237,237}}{unk\strut} \setlength{\fboxsep}{0pt}\def\cbRGB{\colorbox[RGB]}\expandafter\cbRGB\expandafter{\detokenize{255,250,250}}{.\strut} \setlength{\fboxsep}{0pt}\def\cbRGB{\colorbox[RGB]}\expandafter\cbRGB\expandafter{\detokenize{255,255,255}}{''\strut} \setlength{\fboxsep}{0pt}\def\cbRGB{\colorbox[RGB]}\expandafter\cbRGB\expandafter{\detokenize{255,255,255}}{the\strut} \setlength{\fboxsep}{0pt}\def\cbRGB{\colorbox[RGB]}\expandafter\cbRGB\expandafter{\detokenize{255,255,255}}{store\strut} \setlength{\fboxsep}{0pt}\def\cbRGB{\colorbox[RGB]}\expandafter\cbRGB\expandafter{\detokenize{255,255,255}}{where\strut} \setlength{\fboxsep}{0pt}\def\cbRGB{\colorbox[RGB]}\expandafter\cbRGB\expandafter{\detokenize{255,255,255}}{i\strut} \setlength{\fboxsep}{0pt}\def\cbRGB{\colorbox[RGB]}\expandafter\cbRGB\expandafter{\detokenize{255,255,255}}{bought\strut} \setlength{\fboxsep}{0pt}\def\cbRGB{\colorbox[RGB]}\expandafter\cbRGB\expandafter{\detokenize{255,255,255}}{it\strut} \setlength{\fboxsep}{0pt}\def\cbRGB{\colorbox[RGB]}\expandafter\cbRGB\expandafter{\detokenize{255,255,255}}{from\strut} \setlength{\fboxsep}{0pt}\def\cbRGB{\colorbox[RGB]}\expandafter\cbRGB\expandafter{\detokenize{255,255,255}}{had\strut} \setlength{\fboxsep}{0pt}\def\cbRGB{\colorbox[RGB]}\expandafter\cbRGB\expandafter{\detokenize{255,255,255}}{it\strut} \setlength{\fboxsep}{0pt}\def\cbRGB{\colorbox[RGB]}\expandafter\cbRGB\expandafter{\detokenize{255,255,255}}{on\strut} \setlength{\fboxsep}{0pt}\def\cbRGB{\colorbox[RGB]}\expandafter\cbRGB\expandafter{\detokenize{255,246,246}}{the\strut} \setlength{\fboxsep}{0pt}\def\cbRGB{\colorbox[RGB]}\expandafter\cbRGB\expandafter{\detokenize{255,255,255}}{shelf\strut} \setlength{\fboxsep}{0pt}\def\cbRGB{\colorbox[RGB]}\expandafter\cbRGB\expandafter{\detokenize{255,255,255}}{at\strut} \setlength{\fboxsep}{0pt}\def\cbRGB{\colorbox[RGB]}\expandafter\cbRGB\expandafter{\detokenize{255,255,255}}{room\strut} \setlength{\fboxsep}{0pt}\def\cbRGB{\colorbox[RGB]}\expandafter\cbRGB\expandafter{\detokenize{255,255,255}}{temp\strut} \setlength{\fboxsep}{0pt}\def\cbRGB{\colorbox[RGB]}\expandafter\cbRGB\expandafter{\detokenize{255,255,255}}{:\strut} \setlength{\fboxsep}{0pt}\def\cbRGB{\colorbox[RGB]}\expandafter\cbRGB\expandafter{\detokenize{255,255,255}}{(\strut} \setlength{\fboxsep}{0pt}\def\cbRGB{\colorbox[RGB]}\expandafter\cbRGB\expandafter{\detokenize{255,255,255}}{the\strut} \setlength{\fboxsep}{0pt}\def\cbRGB{\colorbox[RGB]}\expandafter\cbRGB\expandafter{\detokenize{255,255,255}}{smell\strut} \setlength{\fboxsep}{0pt}\def\cbRGB{\colorbox[RGB]}\expandafter\cbRGB\expandafter{\detokenize{255,255,255}}{was\strut} \setlength{\fboxsep}{0pt}\def\cbRGB{\colorbox[RGB]}\expandafter\cbRGB\expandafter{\detokenize{255,255,255}}{surprising\strut} \setlength{\fboxsep}{0pt}\def\cbRGB{\colorbox[RGB]}\expandafter\cbRGB\expandafter{\detokenize{255,255,255}}{{\ldots}\strut} \setlength{\fboxsep}{0pt}\def\cbRGB{\colorbox[RGB]}\expandafter\cbRGB\expandafter{\detokenize{255,255,255}}{an\strut} \setlength{\fboxsep}{0pt}\def\cbRGB{\colorbox[RGB]}\expandafter\cbRGB\expandafter{\detokenize{255,255,255}}{earthy\strut} \setlength{\fboxsep}{0pt}\def\cbRGB{\colorbox[RGB]}\expandafter\cbRGB\expandafter{\detokenize{255,255,255}}{roasted\strut} \setlength{\fboxsep}{0pt}\def\cbRGB{\colorbox[RGB]}\expandafter\cbRGB\expandafter{\detokenize{255,255,255}}{smell\strut} \setlength{\fboxsep}{0pt}\def\cbRGB{\colorbox[RGB]}\expandafter\cbRGB\expandafter{\detokenize{255,255,255}}{,\strut} \setlength{\fboxsep}{0pt}\def\cbRGB{\colorbox[RGB]}\expandafter\cbRGB\expandafter{\detokenize{255,255,255}}{mixed\strut} \setlength{\fboxsep}{0pt}\def\cbRGB{\colorbox[RGB]}\expandafter\cbRGB\expandafter{\detokenize{255,255,255}}{with\strut} \setlength{\fboxsep}{0pt}\def\cbRGB{\colorbox[RGB]}\expandafter\cbRGB\expandafter{\detokenize{255,255,255}}{day\strut} \setlength{\fboxsep}{0pt}\def\cbRGB{\colorbox[RGB]}\expandafter\cbRGB\expandafter{\detokenize{255,255,255}}{old\strut} \setlength{\fboxsep}{0pt}\def\cbRGB{\colorbox[RGB]}\expandafter\cbRGB\expandafter{\detokenize{255,255,255}}{coffee\strut} \setlength{\fboxsep}{0pt}\def\cbRGB{\colorbox[RGB]}\expandafter\cbRGB\expandafter{\detokenize{255,255,255}}{.\strut} \setlength{\fboxsep}{0pt}\def\cbRGB{\colorbox[RGB]}\expandafter\cbRGB\expandafter{\detokenize{255,255,255}}{also\strut} \setlength{\fboxsep}{0pt}\def\cbRGB{\colorbox[RGB]}\expandafter\cbRGB\expandafter{\detokenize{255,255,255}}{some\strut} \setlength{\fboxsep}{0pt}\def\cbRGB{\colorbox[RGB]}\expandafter\cbRGB\expandafter{\detokenize{255,255,255}}{weird\strut} \setlength{\fboxsep}{0pt}\def\cbRGB{\colorbox[RGB]}\expandafter\cbRGB\expandafter{\detokenize{255,255,255}}{``\strut} \setlength{\fboxsep}{0pt}\def\cbRGB{\colorbox[RGB]}\expandafter\cbRGB\expandafter{\detokenize{255,255,255}}{off\strut} \setlength{\fboxsep}{0pt}\def\cbRGB{\colorbox[RGB]}\expandafter\cbRGB\expandafter{\detokenize{255,255,255}}{''\strut} \setlength{\fboxsep}{0pt}\def\cbRGB{\colorbox[RGB]}\expandafter\cbRGB\expandafter{\detokenize{255,255,255}}{licorice\strut} \setlength{\fboxsep}{0pt}\def\cbRGB{\colorbox[RGB]}\expandafter\cbRGB\expandafter{\detokenize{255,255,255}}{notes\strut} \setlength{\fboxsep}{0pt}\def\cbRGB{\colorbox[RGB]}\expandafter\cbRGB\expandafter{\detokenize{255,255,255}}{.\strut} \setlength{\fboxsep}{0pt}\def\cbRGB{\colorbox[RGB]}\expandafter\cbRGB\expandafter{\detokenize{255,255,255}}{taste\strut} \setlength{\fboxsep}{0pt}\def\cbRGB{\colorbox[RGB]}\expandafter\cbRGB\expandafter{\detokenize{255,255,255}}{was\strut} \setlength{\fboxsep}{0pt}\def\cbRGB{\colorbox[RGB]}\expandafter\cbRGB\expandafter{\detokenize{255,255,255}}{dry\strut} \setlength{\fboxsep}{0pt}\def\cbRGB{\colorbox[RGB]}\expandafter\cbRGB\expandafter{\detokenize{255,255,255}}{at\strut} \setlength{\fboxsep}{0pt}\def\cbRGB{\colorbox[RGB]}\expandafter\cbRGB\expandafter{\detokenize{255,255,255}}{the\strut} \setlength{\fboxsep}{0pt}\def\cbRGB{\colorbox[RGB]}\expandafter\cbRGB\expandafter{\detokenize{255,255,255}}{start\strut} \setlength{\fboxsep}{0pt}\def\cbRGB{\colorbox[RGB]}\expandafter\cbRGB\expandafter{\detokenize{255,255,255}}{,\strut} \setlength{\fboxsep}{0pt}\def\cbRGB{\colorbox[RGB]}\expandafter\cbRGB\expandafter{\detokenize{255,255,255}}{and\strut} \setlength{\fboxsep}{0pt}\def\cbRGB{\colorbox[RGB]}\expandafter\cbRGB\expandafter{\detokenize{255,255,255}}{very\strut} \setlength{\fboxsep}{0pt}\def\cbRGB{\colorbox[RGB]}\expandafter\cbRGB\expandafter{\detokenize{255,255,255}}{dry\strut} \setlength{\fboxsep}{0pt}\def\cbRGB{\colorbox[RGB]}\expandafter\cbRGB\expandafter{\detokenize{255,255,255}}{on\strut} \setlength{\fboxsep}{0pt}\def\cbRGB{\colorbox[RGB]}\expandafter\cbRGB\expandafter{\detokenize{255,255,255}}{the\strut} \setlength{\fboxsep}{0pt}\def\cbRGB{\colorbox[RGB]}\expandafter\cbRGB\expandafter{\detokenize{255,255,255}}{finish\strut} \setlength{\fboxsep}{0pt}\def\cbRGB{\colorbox[RGB]}\expandafter\cbRGB\expandafter{\detokenize{255,255,255}}{.\strut} \setlength{\fboxsep}{0pt}\def\cbRGB{\colorbox[RGB]}\expandafter\cbRGB\expandafter{\detokenize{255,255,255}}{bitter\strut} \setlength{\fboxsep}{0pt}\def\cbRGB{\colorbox[RGB]}\expandafter\cbRGB\expandafter{\detokenize{255,255,255}}{roast\strut} \setlength{\fboxsep}{0pt}\def\cbRGB{\colorbox[RGB]}\expandafter\cbRGB\expandafter{\detokenize{255,255,255}}{notes\strut} \setlength{\fboxsep}{0pt}\def\cbRGB{\colorbox[RGB]}\expandafter\cbRGB\expandafter{\detokenize{255,255,255}}{with\strut} \setlength{\fboxsep}{0pt}\def\cbRGB{\colorbox[RGB]}\expandafter\cbRGB\expandafter{\detokenize{255,255,255}}{a\strut} \setlength{\fboxsep}{0pt}\def\cbRGB{\colorbox[RGB]}\expandafter\cbRGB\expandafter{\detokenize{255,255,255}}{sort\strut} \setlength{\fboxsep}{0pt}\def\cbRGB{\colorbox[RGB]}\expandafter\cbRGB\expandafter{\detokenize{255,255,255}}{of\strut} \setlength{\fboxsep}{0pt}\def\cbRGB{\colorbox[RGB]}\expandafter\cbRGB\expandafter{\detokenize{255,243,243}}{unk\strut} \setlength{\fboxsep}{0pt}\def\cbRGB{\colorbox[RGB]}\expandafter\cbRGB\expandafter{\detokenize{255,215,215}}{.\strut} \setlength{\fboxsep}{0pt}\def\cbRGB{\colorbox[RGB]}\expandafter\cbRGB\expandafter{\detokenize{255,174,174}}{full\strut} \setlength{\fboxsep}{0pt}\def\cbRGB{\colorbox[RGB]}\expandafter\cbRGB\expandafter{\detokenize{255,154,154}}{bodied\strut} \setlength{\fboxsep}{0pt}\def\cbRGB{\colorbox[RGB]}\expandafter\cbRGB\expandafter{\detokenize{255,135,135}}{with\strut} \setlength{\fboxsep}{0pt}\def\cbRGB{\colorbox[RGB]}\expandafter\cbRGB\expandafter{\detokenize{255,131,131}}{appropriate\strut} \setlength{\fboxsep}{0pt}\def\cbRGB{\colorbox[RGB]}\expandafter\cbRGB\expandafter{\detokenize{255,136,136}}{carbonation\strut} \setlength{\fboxsep}{0pt}\def\cbRGB{\colorbox[RGB]}\expandafter\cbRGB\expandafter{\detokenize{255,159,159}}{.\strut} \setlength{\fboxsep}{0pt}\def\cbRGB{\colorbox[RGB]}\expandafter\cbRGB\expandafter{\detokenize{255,172,172}}{i\strut} \setlength{\fboxsep}{0pt}\def\cbRGB{\colorbox[RGB]}\expandafter\cbRGB\expandafter{\detokenize{255,181,181}}{was\strut} \setlength{\fboxsep}{0pt}\def\cbRGB{\colorbox[RGB]}\expandafter\cbRGB\expandafter{\detokenize{255,184,184}}{very\strut} \setlength{\fboxsep}{0pt}\def\cbRGB{\colorbox[RGB]}\expandafter\cbRGB\expandafter{\detokenize{255,200,200}}{excited\strut} \setlength{\fboxsep}{0pt}\def\cbRGB{\colorbox[RGB]}\expandafter\cbRGB\expandafter{\detokenize{255,219,219}}{to\strut} \setlength{\fboxsep}{0pt}\def\cbRGB{\colorbox[RGB]}\expandafter\cbRGB\expandafter{\detokenize{255,238,238}}{try\strut} \setlength{\fboxsep}{0pt}\def\cbRGB{\colorbox[RGB]}\expandafter\cbRGB\expandafter{\detokenize{255,255,255}}{this\strut} \setlength{\fboxsep}{0pt}\def\cbRGB{\colorbox[RGB]}\expandafter\cbRGB\expandafter{\detokenize{255,255,255}}{beer\strut} \setlength{\fboxsep}{0pt}\def\cbRGB{\colorbox[RGB]}\expandafter\cbRGB\expandafter{\detokenize{255,255,255}}{,\strut} \setlength{\fboxsep}{0pt}\def\cbRGB{\colorbox[RGB]}\expandafter\cbRGB\expandafter{\detokenize{255,255,255}}{and\strut} \setlength{\fboxsep}{0pt}\def\cbRGB{\colorbox[RGB]}\expandafter\cbRGB\expandafter{\detokenize{255,255,255}}{i\strut} \setlength{\fboxsep}{0pt}\def\cbRGB{\colorbox[RGB]}\expandafter\cbRGB\expandafter{\detokenize{255,255,255}}{was\strut} \setlength{\fboxsep}{0pt}\def\cbRGB{\colorbox[RGB]}\expandafter\cbRGB\expandafter{\detokenize{255,255,255}}{pretty\strut} \setlength{\fboxsep}{0pt}\def\cbRGB{\colorbox[RGB]}\expandafter\cbRGB\expandafter{\detokenize{255,246,246}}{disappointed\strut} \setlength{\fboxsep}{0pt}\def\cbRGB{\colorbox[RGB]}\expandafter\cbRGB\expandafter{\detokenize{255,255,255}}{.\strut} \setlength{\fboxsep}{0pt}\def\cbRGB{\colorbox[RGB]}\expandafter\cbRGB\expandafter{\detokenize{255,255,255}}{this\strut} \setlength{\fboxsep}{0pt}\def\cbRGB{\colorbox[RGB]}\expandafter\cbRGB\expandafter{\detokenize{255,255,255}}{bottle\strut} \setlength{\fboxsep}{0pt}\def\cbRGB{\colorbox[RGB]}\expandafter\cbRGB\expandafter{\detokenize{255,255,255}}{could\strut} \setlength{\fboxsep}{0pt}\def\cbRGB{\colorbox[RGB]}\expandafter\cbRGB\expandafter{\detokenize{255,255,255}}{be\strut} \setlength{\fboxsep}{0pt}\def\cbRGB{\colorbox[RGB]}\expandafter\cbRGB\expandafter{\detokenize{255,255,255}}{old\strut} \setlength{\fboxsep}{0pt}\def\cbRGB{\colorbox[RGB]}\expandafter\cbRGB\expandafter{\detokenize{255,255,255}}{,\strut} \setlength{\fboxsep}{0pt}\def\cbRGB{\colorbox[RGB]}\expandafter\cbRGB\expandafter{\detokenize{255,255,255}}{or\strut} \setlength{\fboxsep}{0pt}\def\cbRGB{\colorbox[RGB]}\expandafter\cbRGB\expandafter{\detokenize{255,255,255}}{the\strut} \setlength{\fboxsep}{0pt}\def\cbRGB{\colorbox[RGB]}\expandafter\cbRGB\expandafter{\detokenize{255,255,255}}{flavors\strut} \setlength{\fboxsep}{0pt}\def\cbRGB{\colorbox[RGB]}\expandafter\cbRGB\expandafter{\detokenize{255,255,255}}{could\strut} \setlength{\fboxsep}{0pt}\def\cbRGB{\colorbox[RGB]}\expandafter\cbRGB\expandafter{\detokenize{255,255,255}}{be\strut} \setlength{\fboxsep}{0pt}\def\cbRGB{\colorbox[RGB]}\expandafter\cbRGB\expandafter{\detokenize{255,255,255}}{``\strut} \setlength{\fboxsep}{0pt}\def\cbRGB{\colorbox[RGB]}\expandafter\cbRGB\expandafter{\detokenize{255,255,255}}{off\strut} \setlength{\fboxsep}{0pt}\def\cbRGB{\colorbox[RGB]}\expandafter\cbRGB\expandafter{\detokenize{255,255,255}}{''\strut} \setlength{\fboxsep}{0pt}\def\cbRGB{\colorbox[RGB]}\expandafter\cbRGB\expandafter{\detokenize{255,255,255}}{from\strut} \setlength{\fboxsep}{0pt}\def\cbRGB{\colorbox[RGB]}\expandafter\cbRGB\expandafter{\detokenize{255,250,250}}{the\strut} \setlength{\fboxsep}{0pt}\def\cbRGB{\colorbox[RGB]}\expandafter\cbRGB\expandafter{\detokenize{255,236,236}}{room\strut} \setlength{\fboxsep}{0pt}\def\cbRGB{\colorbox[RGB]}\expandafter\cbRGB\expandafter{\detokenize{255,250,250}}{temp\strut} \setlength{\fboxsep}{0pt}\def\cbRGB{\colorbox[RGB]}\expandafter\cbRGB\expandafter{\detokenize{255,255,255}}{unk\strut} \setlength{\fboxsep}{0pt}\def\cbRGB{\colorbox[RGB]}\expandafter\cbRGB\expandafter{\detokenize{255,255,255}}{.\strut} 

\setlength{\fboxsep}{0pt}\def\cbRGB{\colorbox[RGB]}\expandafter\cbRGB\expandafter{\detokenize{174,255,174}}{poured\strut} \setlength{\fboxsep}{0pt}\def\cbRGB{\colorbox[RGB]}\expandafter\cbRGB\expandafter{\detokenize{221,255,221}}{into\strut} \setlength{\fboxsep}{0pt}\def\cbRGB{\colorbox[RGB]}\expandafter\cbRGB\expandafter{\detokenize{254,255,254}}{a\strut} \setlength{\fboxsep}{0pt}\def\cbRGB{\colorbox[RGB]}\expandafter\cbRGB\expandafter{\detokenize{255,255,255}}{nonic\strut} \setlength{\fboxsep}{0pt}\def\cbRGB{\colorbox[RGB]}\expandafter\cbRGB\expandafter{\detokenize{255,255,255}}{pint\strut} \setlength{\fboxsep}{0pt}\def\cbRGB{\colorbox[RGB]}\expandafter\cbRGB\expandafter{\detokenize{255,255,255}}{glass\strut} \setlength{\fboxsep}{0pt}\def\cbRGB{\colorbox[RGB]}\expandafter\cbRGB\expandafter{\detokenize{255,255,255}}{{\ldots}\strut} \setlength{\fboxsep}{0pt}\def\cbRGB{\colorbox[RGB]}\expandafter\cbRGB\expandafter{\detokenize{255,255,255}}{jet\strut} \setlength{\fboxsep}{0pt}\def\cbRGB{\colorbox[RGB]}\expandafter\cbRGB\expandafter{\detokenize{255,255,255}}{black\strut} \setlength{\fboxsep}{0pt}\def\cbRGB{\colorbox[RGB]}\expandafter\cbRGB\expandafter{\detokenize{255,255,255}}{,\strut} \setlength{\fboxsep}{0pt}\def\cbRGB{\colorbox[RGB]}\expandafter\cbRGB\expandafter{\detokenize{255,255,255}}{unk\strut} \setlength{\fboxsep}{0pt}\def\cbRGB{\colorbox[RGB]}\expandafter\cbRGB\expandafter{\detokenize{255,255,255}}{no\strut} \setlength{\fboxsep}{0pt}\def\cbRGB{\colorbox[RGB]}\expandafter\cbRGB\expandafter{\detokenize{255,255,255}}{highlights\strut} \setlength{\fboxsep}{0pt}\def\cbRGB{\colorbox[RGB]}\expandafter\cbRGB\expandafter{\detokenize{255,255,255}}{around\strut} \setlength{\fboxsep}{0pt}\def\cbRGB{\colorbox[RGB]}\expandafter\cbRGB\expandafter{\detokenize{255,255,255}}{the\strut} \setlength{\fboxsep}{0pt}\def\cbRGB{\colorbox[RGB]}\expandafter\cbRGB\expandafter{\detokenize{255,255,255}}{edges\strut} \setlength{\fboxsep}{0pt}\def\cbRGB{\colorbox[RGB]}\expandafter\cbRGB\expandafter{\detokenize{255,255,255}}{at\strut} \setlength{\fboxsep}{0pt}\def\cbRGB{\colorbox[RGB]}\expandafter\cbRGB\expandafter{\detokenize{255,255,255}}{all\strut} \setlength{\fboxsep}{0pt}\def\cbRGB{\colorbox[RGB]}\expandafter\cbRGB\expandafter{\detokenize{255,255,255}}{,\strut} \setlength{\fboxsep}{0pt}\def\cbRGB{\colorbox[RGB]}\expandafter\cbRGB\expandafter{\detokenize{255,255,255}}{dense\strut} \setlength{\fboxsep}{0pt}\def\cbRGB{\colorbox[RGB]}\expandafter\cbRGB\expandafter{\detokenize{255,255,255}}{tan\strut} \setlength{\fboxsep}{0pt}\def\cbRGB{\colorbox[RGB]}\expandafter\cbRGB\expandafter{\detokenize{255,255,255}}{unk\strut} \setlength{\fboxsep}{0pt}\def\cbRGB{\colorbox[RGB]}\expandafter\cbRGB\expandafter{\detokenize{230,255,230}}{head\strut} \setlength{\fboxsep}{0pt}\def\cbRGB{\colorbox[RGB]}\expandafter\cbRGB\expandafter{\detokenize{195,255,195}}{{\ldots}\strut} \setlength{\fboxsep}{0pt}\def\cbRGB{\colorbox[RGB]}\expandafter\cbRGB\expandafter{\detokenize{175,255,175}}{looks\strut} \setlength{\fboxsep}{0pt}\def\cbRGB{\colorbox[RGB]}\expandafter\cbRGB\expandafter{\detokenize{182,255,182}}{great\strut} \setlength{\fboxsep}{0pt}\def\cbRGB{\colorbox[RGB]}\expandafter\cbRGB\expandafter{\detokenize{216,255,216}}{!\strut} \setlength{\fboxsep}{0pt}\def\cbRGB{\colorbox[RGB]}\expandafter\cbRGB\expandafter{\detokenize{249,255,249}}{a\strut} \setlength{\fboxsep}{0pt}\def\cbRGB{\colorbox[RGB]}\expandafter\cbRGB\expandafter{\detokenize{255,255,255}}{bit\strut} \setlength{\fboxsep}{0pt}\def\cbRGB{\colorbox[RGB]}\expandafter\cbRGB\expandafter{\detokenize{255,255,255}}{of\strut} \setlength{\fboxsep}{0pt}\def\cbRGB{\colorbox[RGB]}\expandafter\cbRGB\expandafter{\detokenize{255,255,255}}{sediment\strut} \setlength{\fboxsep}{0pt}\def\cbRGB{\colorbox[RGB]}\expandafter\cbRGB\expandafter{\detokenize{255,255,255}}{in\strut} \setlength{\fboxsep}{0pt}\def\cbRGB{\colorbox[RGB]}\expandafter\cbRGB\expandafter{\detokenize{255,255,255}}{the\strut} \setlength{\fboxsep}{0pt}\def\cbRGB{\colorbox[RGB]}\expandafter\cbRGB\expandafter{\detokenize{255,255,255}}{bottom\strut} \setlength{\fboxsep}{0pt}\def\cbRGB{\colorbox[RGB]}\expandafter\cbRGB\expandafter{\detokenize{255,255,255}}{of\strut} \setlength{\fboxsep}{0pt}\def\cbRGB{\colorbox[RGB]}\expandafter\cbRGB\expandafter{\detokenize{255,255,255}}{the\strut} \setlength{\fboxsep}{0pt}\def\cbRGB{\colorbox[RGB]}\expandafter\cbRGB\expandafter{\detokenize{255,255,255}}{bottle\strut} \setlength{\fboxsep}{0pt}\def\cbRGB{\colorbox[RGB]}\expandafter\cbRGB\expandafter{\detokenize{255,255,255}}{.\strut} \setlength{\fboxsep}{0pt}\def\cbRGB{\colorbox[RGB]}\expandafter\cbRGB\expandafter{\detokenize{255,255,255}}{bottle\strut} \setlength{\fboxsep}{0pt}\def\cbRGB{\colorbox[RGB]}\expandafter\cbRGB\expandafter{\detokenize{255,255,255}}{read\strut} \setlength{\fboxsep}{0pt}\def\cbRGB{\colorbox[RGB]}\expandafter\cbRGB\expandafter{\detokenize{255,255,255}}{``\strut} \setlength{\fboxsep}{0pt}\def\cbRGB{\colorbox[RGB]}\expandafter\cbRGB\expandafter{\detokenize{255,255,255}}{live\strut} \setlength{\fboxsep}{0pt}\def\cbRGB{\colorbox[RGB]}\expandafter\cbRGB\expandafter{\detokenize{255,255,255}}{ale\strut} \setlength{\fboxsep}{0pt}\def\cbRGB{\colorbox[RGB]}\expandafter\cbRGB\expandafter{\detokenize{255,255,255}}{,\strut} \setlength{\fboxsep}{0pt}\def\cbRGB{\colorbox[RGB]}\expandafter\cbRGB\expandafter{\detokenize{255,255,255}}{keep\strut} \setlength{\fboxsep}{0pt}\def\cbRGB{\colorbox[RGB]}\expandafter\cbRGB\expandafter{\detokenize{255,255,255}}{unk\strut} \setlength{\fboxsep}{0pt}\def\cbRGB{\colorbox[RGB]}\expandafter\cbRGB\expandafter{\detokenize{255,255,255}}{.\strut} \setlength{\fboxsep}{0pt}\def\cbRGB{\colorbox[RGB]}\expandafter\cbRGB\expandafter{\detokenize{255,255,255}}{''\strut} \setlength{\fboxsep}{0pt}\def\cbRGB{\colorbox[RGB]}\expandafter\cbRGB\expandafter{\detokenize{254,255,254}}{the\strut} \setlength{\fboxsep}{0pt}\def\cbRGB{\colorbox[RGB]}\expandafter\cbRGB\expandafter{\detokenize{224,255,224}}{store\strut} \setlength{\fboxsep}{0pt}\def\cbRGB{\colorbox[RGB]}\expandafter\cbRGB\expandafter{\detokenize{195,255,195}}{where\strut} \setlength{\fboxsep}{0pt}\def\cbRGB{\colorbox[RGB]}\expandafter\cbRGB\expandafter{\detokenize{190,255,190}}{i\strut} \setlength{\fboxsep}{0pt}\def\cbRGB{\colorbox[RGB]}\expandafter\cbRGB\expandafter{\detokenize{212,255,212}}{bought\strut} \setlength{\fboxsep}{0pt}\def\cbRGB{\colorbox[RGB]}\expandafter\cbRGB\expandafter{\detokenize{239,255,239}}{it\strut} \setlength{\fboxsep}{0pt}\def\cbRGB{\colorbox[RGB]}\expandafter\cbRGB\expandafter{\detokenize{255,255,255}}{from\strut} \setlength{\fboxsep}{0pt}\def\cbRGB{\colorbox[RGB]}\expandafter\cbRGB\expandafter{\detokenize{255,255,255}}{had\strut} \setlength{\fboxsep}{0pt}\def\cbRGB{\colorbox[RGB]}\expandafter\cbRGB\expandafter{\detokenize{255,255,255}}{it\strut} \setlength{\fboxsep}{0pt}\def\cbRGB{\colorbox[RGB]}\expandafter\cbRGB\expandafter{\detokenize{255,255,255}}{on\strut} \setlength{\fboxsep}{0pt}\def\cbRGB{\colorbox[RGB]}\expandafter\cbRGB\expandafter{\detokenize{255,255,255}}{the\strut} \setlength{\fboxsep}{0pt}\def\cbRGB{\colorbox[RGB]}\expandafter\cbRGB\expandafter{\detokenize{238,255,238}}{shelf\strut} \setlength{\fboxsep}{0pt}\def\cbRGB{\colorbox[RGB]}\expandafter\cbRGB\expandafter{\detokenize{207,255,207}}{at\strut} \setlength{\fboxsep}{0pt}\def\cbRGB{\colorbox[RGB]}\expandafter\cbRGB\expandafter{\detokenize{200,255,200}}{room\strut} \setlength{\fboxsep}{0pt}\def\cbRGB{\colorbox[RGB]}\expandafter\cbRGB\expandafter{\detokenize{180,255,180}}{temp\strut} \setlength{\fboxsep}{0pt}\def\cbRGB{\colorbox[RGB]}\expandafter\cbRGB\expandafter{\detokenize{147,255,147}}{:\strut} \setlength{\fboxsep}{0pt}\def\cbRGB{\colorbox[RGB]}\expandafter\cbRGB\expandafter{\detokenize{128,255,128}}{(\strut} \setlength{\fboxsep}{0pt}\def\cbRGB{\colorbox[RGB]}\expandafter\cbRGB\expandafter{\detokenize{111,255,111}}{the\strut} \setlength{\fboxsep}{0pt}\def\cbRGB{\colorbox[RGB]}\expandafter\cbRGB\expandafter{\detokenize{67,255,67}}{smell\strut} \setlength{\fboxsep}{0pt}\def\cbRGB{\colorbox[RGB]}\expandafter\cbRGB\expandafter{\detokenize{37,255,37}}{was\strut} \setlength{\fboxsep}{0pt}\def\cbRGB{\colorbox[RGB]}\expandafter\cbRGB\expandafter{\detokenize{19,255,19}}{surprising\strut} \setlength{\fboxsep}{0pt}\def\cbRGB{\colorbox[RGB]}\expandafter\cbRGB\expandafter{\detokenize{0,255,0}}{{\ldots}\strut} \setlength{\fboxsep}{0pt}\def\cbRGB{\colorbox[RGB]}\expandafter\cbRGB\expandafter{\detokenize{0,255,0}}{an\strut} \setlength{\fboxsep}{0pt}\def\cbRGB{\colorbox[RGB]}\expandafter\cbRGB\expandafter{\detokenize{25,255,25}}{earthy\strut} \setlength{\fboxsep}{0pt}\def\cbRGB{\colorbox[RGB]}\expandafter\cbRGB\expandafter{\detokenize{61,255,61}}{roasted\strut} \setlength{\fboxsep}{0pt}\def\cbRGB{\colorbox[RGB]}\expandafter\cbRGB\expandafter{\detokenize{111,255,111}}{smell\strut} \setlength{\fboxsep}{0pt}\def\cbRGB{\colorbox[RGB]}\expandafter\cbRGB\expandafter{\detokenize{162,255,162}}{,\strut} \setlength{\fboxsep}{0pt}\def\cbRGB{\colorbox[RGB]}\expandafter\cbRGB\expandafter{\detokenize{199,255,199}}{mixed\strut} \setlength{\fboxsep}{0pt}\def\cbRGB{\colorbox[RGB]}\expandafter\cbRGB\expandafter{\detokenize{224,255,224}}{with\strut} \setlength{\fboxsep}{0pt}\def\cbRGB{\colorbox[RGB]}\expandafter\cbRGB\expandafter{\detokenize{238,255,238}}{day\strut} \setlength{\fboxsep}{0pt}\def\cbRGB{\colorbox[RGB]}\expandafter\cbRGB\expandafter{\detokenize{238,255,238}}{old\strut} \setlength{\fboxsep}{0pt}\def\cbRGB{\colorbox[RGB]}\expandafter\cbRGB\expandafter{\detokenize{249,255,249}}{coffee\strut} \setlength{\fboxsep}{0pt}\def\cbRGB{\colorbox[RGB]}\expandafter\cbRGB\expandafter{\detokenize{255,255,255}}{.\strut} \setlength{\fboxsep}{0pt}\def\cbRGB{\colorbox[RGB]}\expandafter\cbRGB\expandafter{\detokenize{255,255,255}}{also\strut} \setlength{\fboxsep}{0pt}\def\cbRGB{\colorbox[RGB]}\expandafter\cbRGB\expandafter{\detokenize{255,255,255}}{some\strut} \setlength{\fboxsep}{0pt}\def\cbRGB{\colorbox[RGB]}\expandafter\cbRGB\expandafter{\detokenize{255,255,255}}{weird\strut} \setlength{\fboxsep}{0pt}\def\cbRGB{\colorbox[RGB]}\expandafter\cbRGB\expandafter{\detokenize{255,255,255}}{``\strut} \setlength{\fboxsep}{0pt}\def\cbRGB{\colorbox[RGB]}\expandafter\cbRGB\expandafter{\detokenize{255,255,255}}{off\strut} \setlength{\fboxsep}{0pt}\def\cbRGB{\colorbox[RGB]}\expandafter\cbRGB\expandafter{\detokenize{255,255,255}}{''\strut} \setlength{\fboxsep}{0pt}\def\cbRGB{\colorbox[RGB]}\expandafter\cbRGB\expandafter{\detokenize{255,255,255}}{licorice\strut} \setlength{\fboxsep}{0pt}\def\cbRGB{\colorbox[RGB]}\expandafter\cbRGB\expandafter{\detokenize{255,255,255}}{notes\strut} \setlength{\fboxsep}{0pt}\def\cbRGB{\colorbox[RGB]}\expandafter\cbRGB\expandafter{\detokenize{255,255,255}}{.\strut} \setlength{\fboxsep}{0pt}\def\cbRGB{\colorbox[RGB]}\expandafter\cbRGB\expandafter{\detokenize{255,255,255}}{taste\strut} \setlength{\fboxsep}{0pt}\def\cbRGB{\colorbox[RGB]}\expandafter\cbRGB\expandafter{\detokenize{255,255,255}}{was\strut} \setlength{\fboxsep}{0pt}\def\cbRGB{\colorbox[RGB]}\expandafter\cbRGB\expandafter{\detokenize{255,255,255}}{dry\strut} \setlength{\fboxsep}{0pt}\def\cbRGB{\colorbox[RGB]}\expandafter\cbRGB\expandafter{\detokenize{255,255,255}}{at\strut} \setlength{\fboxsep}{0pt}\def\cbRGB{\colorbox[RGB]}\expandafter\cbRGB\expandafter{\detokenize{255,255,255}}{the\strut} \setlength{\fboxsep}{0pt}\def\cbRGB{\colorbox[RGB]}\expandafter\cbRGB\expandafter{\detokenize{255,255,255}}{start\strut} \setlength{\fboxsep}{0pt}\def\cbRGB{\colorbox[RGB]}\expandafter\cbRGB\expandafter{\detokenize{255,255,255}}{,\strut} \setlength{\fboxsep}{0pt}\def\cbRGB{\colorbox[RGB]}\expandafter\cbRGB\expandafter{\detokenize{255,255,255}}{and\strut} \setlength{\fboxsep}{0pt}\def\cbRGB{\colorbox[RGB]}\expandafter\cbRGB\expandafter{\detokenize{255,255,255}}{very\strut} \setlength{\fboxsep}{0pt}\def\cbRGB{\colorbox[RGB]}\expandafter\cbRGB\expandafter{\detokenize{255,255,255}}{dry\strut} \setlength{\fboxsep}{0pt}\def\cbRGB{\colorbox[RGB]}\expandafter\cbRGB\expandafter{\detokenize{255,255,255}}{on\strut} \setlength{\fboxsep}{0pt}\def\cbRGB{\colorbox[RGB]}\expandafter\cbRGB\expandafter{\detokenize{255,255,255}}{the\strut} \setlength{\fboxsep}{0pt}\def\cbRGB{\colorbox[RGB]}\expandafter\cbRGB\expandafter{\detokenize{255,255,255}}{finish\strut} \setlength{\fboxsep}{0pt}\def\cbRGB{\colorbox[RGB]}\expandafter\cbRGB\expandafter{\detokenize{255,255,255}}{.\strut} \setlength{\fboxsep}{0pt}\def\cbRGB{\colorbox[RGB]}\expandafter\cbRGB\expandafter{\detokenize{255,255,255}}{bitter\strut} \setlength{\fboxsep}{0pt}\def\cbRGB{\colorbox[RGB]}\expandafter\cbRGB\expandafter{\detokenize{255,255,255}}{roast\strut} \setlength{\fboxsep}{0pt}\def\cbRGB{\colorbox[RGB]}\expandafter\cbRGB\expandafter{\detokenize{255,255,255}}{notes\strut} \setlength{\fboxsep}{0pt}\def\cbRGB{\colorbox[RGB]}\expandafter\cbRGB\expandafter{\detokenize{255,255,255}}{with\strut} \setlength{\fboxsep}{0pt}\def\cbRGB{\colorbox[RGB]}\expandafter\cbRGB\expandafter{\detokenize{255,255,255}}{a\strut} \setlength{\fboxsep}{0pt}\def\cbRGB{\colorbox[RGB]}\expandafter\cbRGB\expandafter{\detokenize{255,255,255}}{sort\strut} \setlength{\fboxsep}{0pt}\def\cbRGB{\colorbox[RGB]}\expandafter\cbRGB\expandafter{\detokenize{255,255,255}}{of\strut} \setlength{\fboxsep}{0pt}\def\cbRGB{\colorbox[RGB]}\expandafter\cbRGB\expandafter{\detokenize{255,255,255}}{unk\strut} \setlength{\fboxsep}{0pt}\def\cbRGB{\colorbox[RGB]}\expandafter\cbRGB\expandafter{\detokenize{255,255,255}}{.\strut} \setlength{\fboxsep}{0pt}\def\cbRGB{\colorbox[RGB]}\expandafter\cbRGB\expandafter{\detokenize{236,255,236}}{full\strut} \setlength{\fboxsep}{0pt}\def\cbRGB{\colorbox[RGB]}\expandafter\cbRGB\expandafter{\detokenize{230,255,230}}{bodied\strut} \setlength{\fboxsep}{0pt}\def\cbRGB{\colorbox[RGB]}\expandafter\cbRGB\expandafter{\detokenize{251,255,251}}{with\strut} \setlength{\fboxsep}{0pt}\def\cbRGB{\colorbox[RGB]}\expandafter\cbRGB\expandafter{\detokenize{255,255,255}}{appropriate\strut} \setlength{\fboxsep}{0pt}\def\cbRGB{\colorbox[RGB]}\expandafter\cbRGB\expandafter{\detokenize{251,255,251}}{carbonation\strut} \setlength{\fboxsep}{0pt}\def\cbRGB{\colorbox[RGB]}\expandafter\cbRGB\expandafter{\detokenize{233,255,233}}{.\strut} \setlength{\fboxsep}{0pt}\def\cbRGB{\colorbox[RGB]}\expandafter\cbRGB\expandafter{\detokenize{194,255,194}}{i\strut} \setlength{\fboxsep}{0pt}\def\cbRGB{\colorbox[RGB]}\expandafter\cbRGB\expandafter{\detokenize{142,255,142}}{was\strut} \setlength{\fboxsep}{0pt}\def\cbRGB{\colorbox[RGB]}\expandafter\cbRGB\expandafter{\detokenize{117,255,117}}{very\strut} \setlength{\fboxsep}{0pt}\def\cbRGB{\colorbox[RGB]}\expandafter\cbRGB\expandafter{\detokenize{117,255,117}}{excited\strut} \setlength{\fboxsep}{0pt}\def\cbRGB{\colorbox[RGB]}\expandafter\cbRGB\expandafter{\detokenize{139,255,139}}{to\strut} \setlength{\fboxsep}{0pt}\def\cbRGB{\colorbox[RGB]}\expandafter\cbRGB\expandafter{\detokenize{188,255,188}}{try\strut} \setlength{\fboxsep}{0pt}\def\cbRGB{\colorbox[RGB]}\expandafter\cbRGB\expandafter{\detokenize{237,255,237}}{this\strut} \setlength{\fboxsep}{0pt}\def\cbRGB{\colorbox[RGB]}\expandafter\cbRGB\expandafter{\detokenize{255,255,255}}{beer\strut} \setlength{\fboxsep}{0pt}\def\cbRGB{\colorbox[RGB]}\expandafter\cbRGB\expandafter{\detokenize{255,255,255}}{,\strut} \setlength{\fboxsep}{0pt}\def\cbRGB{\colorbox[RGB]}\expandafter\cbRGB\expandafter{\detokenize{255,255,255}}{and\strut} \setlength{\fboxsep}{0pt}\def\cbRGB{\colorbox[RGB]}\expandafter\cbRGB\expandafter{\detokenize{255,255,255}}{i\strut} \setlength{\fboxsep}{0pt}\def\cbRGB{\colorbox[RGB]}\expandafter\cbRGB\expandafter{\detokenize{255,255,255}}{was\strut} \setlength{\fboxsep}{0pt}\def\cbRGB{\colorbox[RGB]}\expandafter\cbRGB\expandafter{\detokenize{246,255,246}}{pretty\strut} \setlength{\fboxsep}{0pt}\def\cbRGB{\colorbox[RGB]}\expandafter\cbRGB\expandafter{\detokenize{218,255,218}}{disappointed\strut} \setlength{\fboxsep}{0pt}\def\cbRGB{\colorbox[RGB]}\expandafter\cbRGB\expandafter{\detokenize{216,255,216}}{.\strut} \setlength{\fboxsep}{0pt}\def\cbRGB{\colorbox[RGB]}\expandafter\cbRGB\expandafter{\detokenize{211,255,211}}{this\strut} \setlength{\fboxsep}{0pt}\def\cbRGB{\colorbox[RGB]}\expandafter\cbRGB\expandafter{\detokenize{233,255,233}}{bottle\strut} \setlength{\fboxsep}{0pt}\def\cbRGB{\colorbox[RGB]}\expandafter\cbRGB\expandafter{\detokenize{249,255,249}}{could\strut} \setlength{\fboxsep}{0pt}\def\cbRGB{\colorbox[RGB]}\expandafter\cbRGB\expandafter{\detokenize{253,255,253}}{be\strut} \setlength{\fboxsep}{0pt}\def\cbRGB{\colorbox[RGB]}\expandafter\cbRGB\expandafter{\detokenize{245,255,245}}{old\strut} \setlength{\fboxsep}{0pt}\def\cbRGB{\colorbox[RGB]}\expandafter\cbRGB\expandafter{\detokenize{243,255,243}}{,\strut} \setlength{\fboxsep}{0pt}\def\cbRGB{\colorbox[RGB]}\expandafter\cbRGB\expandafter{\detokenize{227,255,227}}{or\strut} \setlength{\fboxsep}{0pt}\def\cbRGB{\colorbox[RGB]}\expandafter\cbRGB\expandafter{\detokenize{225,255,225}}{the\strut} \setlength{\fboxsep}{0pt}\def\cbRGB{\colorbox[RGB]}\expandafter\cbRGB\expandafter{\detokenize{249,255,249}}{flavors\strut} \setlength{\fboxsep}{0pt}\def\cbRGB{\colorbox[RGB]}\expandafter\cbRGB\expandafter{\detokenize{255,255,255}}{could\strut} \setlength{\fboxsep}{0pt}\def\cbRGB{\colorbox[RGB]}\expandafter\cbRGB\expandafter{\detokenize{255,255,255}}{be\strut} \setlength{\fboxsep}{0pt}\def\cbRGB{\colorbox[RGB]}\expandafter\cbRGB\expandafter{\detokenize{255,255,255}}{``\strut} \setlength{\fboxsep}{0pt}\def\cbRGB{\colorbox[RGB]}\expandafter\cbRGB\expandafter{\detokenize{255,255,255}}{off\strut} \setlength{\fboxsep}{0pt}\def\cbRGB{\colorbox[RGB]}\expandafter\cbRGB\expandafter{\detokenize{255,255,255}}{''\strut} \setlength{\fboxsep}{0pt}\def\cbRGB{\colorbox[RGB]}\expandafter\cbRGB\expandafter{\detokenize{255,255,255}}{from\strut} \setlength{\fboxsep}{0pt}\def\cbRGB{\colorbox[RGB]}\expandafter\cbRGB\expandafter{\detokenize{241,255,241}}{the\strut} \setlength{\fboxsep}{0pt}\def\cbRGB{\colorbox[RGB]}\expandafter\cbRGB\expandafter{\detokenize{211,255,211}}{room\strut} \setlength{\fboxsep}{0pt}\def\cbRGB{\colorbox[RGB]}\expandafter\cbRGB\expandafter{\detokenize{209,255,209}}{temp\strut} \setlength{\fboxsep}{0pt}\def\cbRGB{\colorbox[RGB]}\expandafter\cbRGB\expandafter{\detokenize{223,255,223}}{unk\strut} \setlength{\fboxsep}{0pt}\def\cbRGB{\colorbox[RGB]}\expandafter\cbRGB\expandafter{\detokenize{251,255,251}}{.\strut} 

\setlength{\fboxsep}{0pt}\def\cbRGB{\colorbox[RGB]}\expandafter\cbRGB\expandafter{\detokenize{156,156,255}}{poured\strut} \setlength{\fboxsep}{0pt}\def\cbRGB{\colorbox[RGB]}\expandafter\cbRGB\expandafter{\detokenize{200,200,255}}{into\strut} \setlength{\fboxsep}{0pt}\def\cbRGB{\colorbox[RGB]}\expandafter\cbRGB\expandafter{\detokenize{229,229,255}}{a\strut} \setlength{\fboxsep}{0pt}\def\cbRGB{\colorbox[RGB]}\expandafter\cbRGB\expandafter{\detokenize{229,229,255}}{nonic\strut} \setlength{\fboxsep}{0pt}\def\cbRGB{\colorbox[RGB]}\expandafter\cbRGB\expandafter{\detokenize{239,239,255}}{pint\strut} \setlength{\fboxsep}{0pt}\def\cbRGB{\colorbox[RGB]}\expandafter\cbRGB\expandafter{\detokenize{234,234,255}}{glass\strut} \setlength{\fboxsep}{0pt}\def\cbRGB{\colorbox[RGB]}\expandafter\cbRGB\expandafter{\detokenize{215,215,255}}{{\ldots}\strut} \setlength{\fboxsep}{0pt}\def\cbRGB{\colorbox[RGB]}\expandafter\cbRGB\expandafter{\detokenize{182,182,255}}{jet\strut} \setlength{\fboxsep}{0pt}\def\cbRGB{\colorbox[RGB]}\expandafter\cbRGB\expandafter{\detokenize{165,165,255}}{black\strut} \setlength{\fboxsep}{0pt}\def\cbRGB{\colorbox[RGB]}\expandafter\cbRGB\expandafter{\detokenize{129,129,255}}{,\strut} \setlength{\fboxsep}{0pt}\def\cbRGB{\colorbox[RGB]}\expandafter\cbRGB\expandafter{\detokenize{87,87,255}}{unk\strut} \setlength{\fboxsep}{0pt}\def\cbRGB{\colorbox[RGB]}\expandafter\cbRGB\expandafter{\detokenize{62,62,255}}{no\strut} \setlength{\fboxsep}{0pt}\def\cbRGB{\colorbox[RGB]}\expandafter\cbRGB\expandafter{\detokenize{65,65,255}}{highlights\strut} \setlength{\fboxsep}{0pt}\def\cbRGB{\colorbox[RGB]}\expandafter\cbRGB\expandafter{\detokenize{81,81,255}}{around\strut} \setlength{\fboxsep}{0pt}\def\cbRGB{\colorbox[RGB]}\expandafter\cbRGB\expandafter{\detokenize{117,117,255}}{the\strut} \setlength{\fboxsep}{0pt}\def\cbRGB{\colorbox[RGB]}\expandafter\cbRGB\expandafter{\detokenize{169,169,255}}{edges\strut} \setlength{\fboxsep}{0pt}\def\cbRGB{\colorbox[RGB]}\expandafter\cbRGB\expandafter{\detokenize{216,216,255}}{at\strut} \setlength{\fboxsep}{0pt}\def\cbRGB{\colorbox[RGB]}\expandafter\cbRGB\expandafter{\detokenize{241,241,255}}{all\strut} \setlength{\fboxsep}{0pt}\def\cbRGB{\colorbox[RGB]}\expandafter\cbRGB\expandafter{\detokenize{252,252,255}}{,\strut} \setlength{\fboxsep}{0pt}\def\cbRGB{\colorbox[RGB]}\expandafter\cbRGB\expandafter{\detokenize{236,236,255}}{dense\strut} \setlength{\fboxsep}{0pt}\def\cbRGB{\colorbox[RGB]}\expandafter\cbRGB\expandafter{\detokenize{197,197,255}}{tan\strut} \setlength{\fboxsep}{0pt}\def\cbRGB{\colorbox[RGB]}\expandafter\cbRGB\expandafter{\detokenize{155,155,255}}{unk\strut} \setlength{\fboxsep}{0pt}\def\cbRGB{\colorbox[RGB]}\expandafter\cbRGB\expandafter{\detokenize{131,131,255}}{head\strut} \setlength{\fboxsep}{0pt}\def\cbRGB{\colorbox[RGB]}\expandafter\cbRGB\expandafter{\detokenize{106,106,255}}{{\ldots}\strut} \setlength{\fboxsep}{0pt}\def\cbRGB{\colorbox[RGB]}\expandafter\cbRGB\expandafter{\detokenize{111,111,255}}{looks\strut} \setlength{\fboxsep}{0pt}\def\cbRGB{\colorbox[RGB]}\expandafter\cbRGB\expandafter{\detokenize{145,145,255}}{great\strut} \setlength{\fboxsep}{0pt}\def\cbRGB{\colorbox[RGB]}\expandafter\cbRGB\expandafter{\detokenize{198,198,255}}{!\strut} \setlength{\fboxsep}{0pt}\def\cbRGB{\colorbox[RGB]}\expandafter\cbRGB\expandafter{\detokenize{243,243,255}}{a\strut} \setlength{\fboxsep}{0pt}\def\cbRGB{\colorbox[RGB]}\expandafter\cbRGB\expandafter{\detokenize{255,255,255}}{bit\strut} \setlength{\fboxsep}{0pt}\def\cbRGB{\colorbox[RGB]}\expandafter\cbRGB\expandafter{\detokenize{255,255,255}}{of\strut} \setlength{\fboxsep}{0pt}\def\cbRGB{\colorbox[RGB]}\expandafter\cbRGB\expandafter{\detokenize{255,255,255}}{sediment\strut} \setlength{\fboxsep}{0pt}\def\cbRGB{\colorbox[RGB]}\expandafter\cbRGB\expandafter{\detokenize{255,255,255}}{in\strut} \setlength{\fboxsep}{0pt}\def\cbRGB{\colorbox[RGB]}\expandafter\cbRGB\expandafter{\detokenize{255,255,255}}{the\strut} \setlength{\fboxsep}{0pt}\def\cbRGB{\colorbox[RGB]}\expandafter\cbRGB\expandafter{\detokenize{255,255,255}}{bottom\strut} \setlength{\fboxsep}{0pt}\def\cbRGB{\colorbox[RGB]}\expandafter\cbRGB\expandafter{\detokenize{255,255,255}}{of\strut} \setlength{\fboxsep}{0pt}\def\cbRGB{\colorbox[RGB]}\expandafter\cbRGB\expandafter{\detokenize{255,255,255}}{the\strut} \setlength{\fboxsep}{0pt}\def\cbRGB{\colorbox[RGB]}\expandafter\cbRGB\expandafter{\detokenize{255,255,255}}{bottle\strut} \setlength{\fboxsep}{0pt}\def\cbRGB{\colorbox[RGB]}\expandafter\cbRGB\expandafter{\detokenize{255,255,255}}{.\strut} \setlength{\fboxsep}{0pt}\def\cbRGB{\colorbox[RGB]}\expandafter\cbRGB\expandafter{\detokenize{255,255,255}}{bottle\strut} \setlength{\fboxsep}{0pt}\def\cbRGB{\colorbox[RGB]}\expandafter\cbRGB\expandafter{\detokenize{255,255,255}}{read\strut} \setlength{\fboxsep}{0pt}\def\cbRGB{\colorbox[RGB]}\expandafter\cbRGB\expandafter{\detokenize{253,253,255}}{``\strut} \setlength{\fboxsep}{0pt}\def\cbRGB{\colorbox[RGB]}\expandafter\cbRGB\expandafter{\detokenize{228,228,255}}{live\strut} \setlength{\fboxsep}{0pt}\def\cbRGB{\colorbox[RGB]}\expandafter\cbRGB\expandafter{\detokenize{231,231,255}}{ale\strut} \setlength{\fboxsep}{0pt}\def\cbRGB{\colorbox[RGB]}\expandafter\cbRGB\expandafter{\detokenize{235,235,255}}{,\strut} \setlength{\fboxsep}{0pt}\def\cbRGB{\colorbox[RGB]}\expandafter\cbRGB\expandafter{\detokenize{255,255,255}}{keep\strut} \setlength{\fboxsep}{0pt}\def\cbRGB{\colorbox[RGB]}\expandafter\cbRGB\expandafter{\detokenize{255,255,255}}{unk\strut} \setlength{\fboxsep}{0pt}\def\cbRGB{\colorbox[RGB]}\expandafter\cbRGB\expandafter{\detokenize{255,255,255}}{.\strut} \setlength{\fboxsep}{0pt}\def\cbRGB{\colorbox[RGB]}\expandafter\cbRGB\expandafter{\detokenize{255,255,255}}{''\strut} \setlength{\fboxsep}{0pt}\def\cbRGB{\colorbox[RGB]}\expandafter\cbRGB\expandafter{\detokenize{255,255,255}}{the\strut} \setlength{\fboxsep}{0pt}\def\cbRGB{\colorbox[RGB]}\expandafter\cbRGB\expandafter{\detokenize{255,255,255}}{store\strut} \setlength{\fboxsep}{0pt}\def\cbRGB{\colorbox[RGB]}\expandafter\cbRGB\expandafter{\detokenize{255,255,255}}{where\strut} \setlength{\fboxsep}{0pt}\def\cbRGB{\colorbox[RGB]}\expandafter\cbRGB\expandafter{\detokenize{255,255,255}}{i\strut} \setlength{\fboxsep}{0pt}\def\cbRGB{\colorbox[RGB]}\expandafter\cbRGB\expandafter{\detokenize{255,255,255}}{bought\strut} \setlength{\fboxsep}{0pt}\def\cbRGB{\colorbox[RGB]}\expandafter\cbRGB\expandafter{\detokenize{255,255,255}}{it\strut} \setlength{\fboxsep}{0pt}\def\cbRGB{\colorbox[RGB]}\expandafter\cbRGB\expandafter{\detokenize{255,255,255}}{from\strut} \setlength{\fboxsep}{0pt}\def\cbRGB{\colorbox[RGB]}\expandafter\cbRGB\expandafter{\detokenize{255,255,255}}{had\strut} \setlength{\fboxsep}{0pt}\def\cbRGB{\colorbox[RGB]}\expandafter\cbRGB\expandafter{\detokenize{255,255,255}}{it\strut} \setlength{\fboxsep}{0pt}\def\cbRGB{\colorbox[RGB]}\expandafter\cbRGB\expandafter{\detokenize{255,255,255}}{on\strut} \setlength{\fboxsep}{0pt}\def\cbRGB{\colorbox[RGB]}\expandafter\cbRGB\expandafter{\detokenize{255,255,255}}{the\strut} \setlength{\fboxsep}{0pt}\def\cbRGB{\colorbox[RGB]}\expandafter\cbRGB\expandafter{\detokenize{255,255,255}}{shelf\strut} \setlength{\fboxsep}{0pt}\def\cbRGB{\colorbox[RGB]}\expandafter\cbRGB\expandafter{\detokenize{255,255,255}}{at\strut} \setlength{\fboxsep}{0pt}\def\cbRGB{\colorbox[RGB]}\expandafter\cbRGB\expandafter{\detokenize{251,251,255}}{room\strut} \setlength{\fboxsep}{0pt}\def\cbRGB{\colorbox[RGB]}\expandafter\cbRGB\expandafter{\detokenize{243,243,255}}{temp\strut} \setlength{\fboxsep}{0pt}\def\cbRGB{\colorbox[RGB]}\expandafter\cbRGB\expandafter{\detokenize{255,255,255}}{:\strut} \setlength{\fboxsep}{0pt}\def\cbRGB{\colorbox[RGB]}\expandafter\cbRGB\expandafter{\detokenize{255,255,255}}{(\strut} \setlength{\fboxsep}{0pt}\def\cbRGB{\colorbox[RGB]}\expandafter\cbRGB\expandafter{\detokenize{255,255,255}}{the\strut} \setlength{\fboxsep}{0pt}\def\cbRGB{\colorbox[RGB]}\expandafter\cbRGB\expandafter{\detokenize{255,255,255}}{smell\strut} \setlength{\fboxsep}{0pt}\def\cbRGB{\colorbox[RGB]}\expandafter\cbRGB\expandafter{\detokenize{255,255,255}}{was\strut} \setlength{\fboxsep}{0pt}\def\cbRGB{\colorbox[RGB]}\expandafter\cbRGB\expandafter{\detokenize{255,255,255}}{surprising\strut} \setlength{\fboxsep}{0pt}\def\cbRGB{\colorbox[RGB]}\expandafter\cbRGB\expandafter{\detokenize{255,255,255}}{{\ldots}\strut} \setlength{\fboxsep}{0pt}\def\cbRGB{\colorbox[RGB]}\expandafter\cbRGB\expandafter{\detokenize{255,255,255}}{an\strut} \setlength{\fboxsep}{0pt}\def\cbRGB{\colorbox[RGB]}\expandafter\cbRGB\expandafter{\detokenize{255,255,255}}{earthy\strut} \setlength{\fboxsep}{0pt}\def\cbRGB{\colorbox[RGB]}\expandafter\cbRGB\expandafter{\detokenize{255,255,255}}{roasted\strut} \setlength{\fboxsep}{0pt}\def\cbRGB{\colorbox[RGB]}\expandafter\cbRGB\expandafter{\detokenize{255,255,255}}{smell\strut} \setlength{\fboxsep}{0pt}\def\cbRGB{\colorbox[RGB]}\expandafter\cbRGB\expandafter{\detokenize{255,255,255}}{,\strut} \setlength{\fboxsep}{0pt}\def\cbRGB{\colorbox[RGB]}\expandafter\cbRGB\expandafter{\detokenize{255,255,255}}{mixed\strut} \setlength{\fboxsep}{0pt}\def\cbRGB{\colorbox[RGB]}\expandafter\cbRGB\expandafter{\detokenize{255,255,255}}{with\strut} \setlength{\fboxsep}{0pt}\def\cbRGB{\colorbox[RGB]}\expandafter\cbRGB\expandafter{\detokenize{255,255,255}}{day\strut} \setlength{\fboxsep}{0pt}\def\cbRGB{\colorbox[RGB]}\expandafter\cbRGB\expandafter{\detokenize{255,255,255}}{old\strut} \setlength{\fboxsep}{0pt}\def\cbRGB{\colorbox[RGB]}\expandafter\cbRGB\expandafter{\detokenize{255,255,255}}{coffee\strut} \setlength{\fboxsep}{0pt}\def\cbRGB{\colorbox[RGB]}\expandafter\cbRGB\expandafter{\detokenize{255,255,255}}{.\strut} \setlength{\fboxsep}{0pt}\def\cbRGB{\colorbox[RGB]}\expandafter\cbRGB\expandafter{\detokenize{255,255,255}}{also\strut} \setlength{\fboxsep}{0pt}\def\cbRGB{\colorbox[RGB]}\expandafter\cbRGB\expandafter{\detokenize{255,255,255}}{some\strut} \setlength{\fboxsep}{0pt}\def\cbRGB{\colorbox[RGB]}\expandafter\cbRGB\expandafter{\detokenize{255,255,255}}{weird\strut} \setlength{\fboxsep}{0pt}\def\cbRGB{\colorbox[RGB]}\expandafter\cbRGB\expandafter{\detokenize{255,255,255}}{``\strut} \setlength{\fboxsep}{0pt}\def\cbRGB{\colorbox[RGB]}\expandafter\cbRGB\expandafter{\detokenize{255,255,255}}{off\strut} \setlength{\fboxsep}{0pt}\def\cbRGB{\colorbox[RGB]}\expandafter\cbRGB\expandafter{\detokenize{255,255,255}}{''\strut} \setlength{\fboxsep}{0pt}\def\cbRGB{\colorbox[RGB]}\expandafter\cbRGB\expandafter{\detokenize{255,255,255}}{licorice\strut} \setlength{\fboxsep}{0pt}\def\cbRGB{\colorbox[RGB]}\expandafter\cbRGB\expandafter{\detokenize{255,255,255}}{notes\strut} \setlength{\fboxsep}{0pt}\def\cbRGB{\colorbox[RGB]}\expandafter\cbRGB\expandafter{\detokenize{255,255,255}}{.\strut} \setlength{\fboxsep}{0pt}\def\cbRGB{\colorbox[RGB]}\expandafter\cbRGB\expandafter{\detokenize{255,255,255}}{taste\strut} \setlength{\fboxsep}{0pt}\def\cbRGB{\colorbox[RGB]}\expandafter\cbRGB\expandafter{\detokenize{255,255,255}}{was\strut} \setlength{\fboxsep}{0pt}\def\cbRGB{\colorbox[RGB]}\expandafter\cbRGB\expandafter{\detokenize{255,255,255}}{dry\strut} \setlength{\fboxsep}{0pt}\def\cbRGB{\colorbox[RGB]}\expandafter\cbRGB\expandafter{\detokenize{255,255,255}}{at\strut} \setlength{\fboxsep}{0pt}\def\cbRGB{\colorbox[RGB]}\expandafter\cbRGB\expandafter{\detokenize{255,255,255}}{the\strut} \setlength{\fboxsep}{0pt}\def\cbRGB{\colorbox[RGB]}\expandafter\cbRGB\expandafter{\detokenize{255,255,255}}{start\strut} \setlength{\fboxsep}{0pt}\def\cbRGB{\colorbox[RGB]}\expandafter\cbRGB\expandafter{\detokenize{255,255,255}}{,\strut} \setlength{\fboxsep}{0pt}\def\cbRGB{\colorbox[RGB]}\expandafter\cbRGB\expandafter{\detokenize{255,255,255}}{and\strut} \setlength{\fboxsep}{0pt}\def\cbRGB{\colorbox[RGB]}\expandafter\cbRGB\expandafter{\detokenize{255,255,255}}{very\strut} \setlength{\fboxsep}{0pt}\def\cbRGB{\colorbox[RGB]}\expandafter\cbRGB\expandafter{\detokenize{255,255,255}}{dry\strut} \setlength{\fboxsep}{0pt}\def\cbRGB{\colorbox[RGB]}\expandafter\cbRGB\expandafter{\detokenize{255,255,255}}{on\strut} \setlength{\fboxsep}{0pt}\def\cbRGB{\colorbox[RGB]}\expandafter\cbRGB\expandafter{\detokenize{255,255,255}}{the\strut} \setlength{\fboxsep}{0pt}\def\cbRGB{\colorbox[RGB]}\expandafter\cbRGB\expandafter{\detokenize{255,255,255}}{finish\strut} \setlength{\fboxsep}{0pt}\def\cbRGB{\colorbox[RGB]}\expandafter\cbRGB\expandafter{\detokenize{255,255,255}}{.\strut} \setlength{\fboxsep}{0pt}\def\cbRGB{\colorbox[RGB]}\expandafter\cbRGB\expandafter{\detokenize{255,255,255}}{bitter\strut} \setlength{\fboxsep}{0pt}\def\cbRGB{\colorbox[RGB]}\expandafter\cbRGB\expandafter{\detokenize{255,255,255}}{roast\strut} \setlength{\fboxsep}{0pt}\def\cbRGB{\colorbox[RGB]}\expandafter\cbRGB\expandafter{\detokenize{255,255,255}}{notes\strut} \setlength{\fboxsep}{0pt}\def\cbRGB{\colorbox[RGB]}\expandafter\cbRGB\expandafter{\detokenize{255,255,255}}{with\strut} \setlength{\fboxsep}{0pt}\def\cbRGB{\colorbox[RGB]}\expandafter\cbRGB\expandafter{\detokenize{255,255,255}}{a\strut} \setlength{\fboxsep}{0pt}\def\cbRGB{\colorbox[RGB]}\expandafter\cbRGB\expandafter{\detokenize{255,255,255}}{sort\strut} \setlength{\fboxsep}{0pt}\def\cbRGB{\colorbox[RGB]}\expandafter\cbRGB\expandafter{\detokenize{237,237,255}}{of\strut} \setlength{\fboxsep}{0pt}\def\cbRGB{\colorbox[RGB]}\expandafter\cbRGB\expandafter{\detokenize{176,176,255}}{unk\strut} \setlength{\fboxsep}{0pt}\def\cbRGB{\colorbox[RGB]}\expandafter\cbRGB\expandafter{\detokenize{113,113,255}}{.\strut} \setlength{\fboxsep}{0pt}\def\cbRGB{\colorbox[RGB]}\expandafter\cbRGB\expandafter{\detokenize{54,54,255}}{full\strut} \setlength{\fboxsep}{0pt}\def\cbRGB{\colorbox[RGB]}\expandafter\cbRGB\expandafter{\detokenize{12,12,255}}{bodied\strut} \setlength{\fboxsep}{0pt}\def\cbRGB{\colorbox[RGB]}\expandafter\cbRGB\expandafter{\detokenize{0,0,255}}{with\strut} \setlength{\fboxsep}{0pt}\def\cbRGB{\colorbox[RGB]}\expandafter\cbRGB\expandafter{\detokenize{3,3,255}}{appropriate\strut} \setlength{\fboxsep}{0pt}\def\cbRGB{\colorbox[RGB]}\expandafter\cbRGB\expandafter{\detokenize{16,16,255}}{carbonation\strut} \setlength{\fboxsep}{0pt}\def\cbRGB{\colorbox[RGB]}\expandafter\cbRGB\expandafter{\detokenize{37,37,255}}{.\strut} \setlength{\fboxsep}{0pt}\def\cbRGB{\colorbox[RGB]}\expandafter\cbRGB\expandafter{\detokenize{70,70,255}}{i\strut} \setlength{\fboxsep}{0pt}\def\cbRGB{\colorbox[RGB]}\expandafter\cbRGB\expandafter{\detokenize{74,74,255}}{was\strut} \setlength{\fboxsep}{0pt}\def\cbRGB{\colorbox[RGB]}\expandafter\cbRGB\expandafter{\detokenize{89,89,255}}{very\strut} \setlength{\fboxsep}{0pt}\def\cbRGB{\colorbox[RGB]}\expandafter\cbRGB\expandafter{\detokenize{119,119,255}}{excited\strut} \setlength{\fboxsep}{0pt}\def\cbRGB{\colorbox[RGB]}\expandafter\cbRGB\expandafter{\detokenize{157,157,255}}{to\strut} \setlength{\fboxsep}{0pt}\def\cbRGB{\colorbox[RGB]}\expandafter\cbRGB\expandafter{\detokenize{181,181,255}}{try\strut} \setlength{\fboxsep}{0pt}\def\cbRGB{\colorbox[RGB]}\expandafter\cbRGB\expandafter{\detokenize{236,236,255}}{this\strut} \setlength{\fboxsep}{0pt}\def\cbRGB{\colorbox[RGB]}\expandafter\cbRGB\expandafter{\detokenize{255,255,255}}{beer\strut} \setlength{\fboxsep}{0pt}\def\cbRGB{\colorbox[RGB]}\expandafter\cbRGB\expandafter{\detokenize{255,255,255}}{,\strut} \setlength{\fboxsep}{0pt}\def\cbRGB{\colorbox[RGB]}\expandafter\cbRGB\expandafter{\detokenize{255,255,255}}{and\strut} \setlength{\fboxsep}{0pt}\def\cbRGB{\colorbox[RGB]}\expandafter\cbRGB\expandafter{\detokenize{255,255,255}}{i\strut} \setlength{\fboxsep}{0pt}\def\cbRGB{\colorbox[RGB]}\expandafter\cbRGB\expandafter{\detokenize{224,224,255}}{was\strut} \setlength{\fboxsep}{0pt}\def\cbRGB{\colorbox[RGB]}\expandafter\cbRGB\expandafter{\detokenize{172,172,255}}{pretty\strut} \setlength{\fboxsep}{0pt}\def\cbRGB{\colorbox[RGB]}\expandafter\cbRGB\expandafter{\detokenize{119,119,255}}{disappointed\strut} \setlength{\fboxsep}{0pt}\def\cbRGB{\colorbox[RGB]}\expandafter\cbRGB\expandafter{\detokenize{91,91,255}}{.\strut} \setlength{\fboxsep}{0pt}\def\cbRGB{\colorbox[RGB]}\expandafter\cbRGB\expandafter{\detokenize{93,93,255}}{this\strut} \setlength{\fboxsep}{0pt}\def\cbRGB{\colorbox[RGB]}\expandafter\cbRGB\expandafter{\detokenize{124,124,255}}{bottle\strut} \setlength{\fboxsep}{0pt}\def\cbRGB{\colorbox[RGB]}\expandafter\cbRGB\expandafter{\detokenize{161,161,255}}{could\strut} \setlength{\fboxsep}{0pt}\def\cbRGB{\colorbox[RGB]}\expandafter\cbRGB\expandafter{\detokenize{220,220,255}}{be\strut} \setlength{\fboxsep}{0pt}\def\cbRGB{\colorbox[RGB]}\expandafter\cbRGB\expandafter{\detokenize{255,255,255}}{old\strut} \setlength{\fboxsep}{0pt}\def\cbRGB{\colorbox[RGB]}\expandafter\cbRGB\expandafter{\detokenize{255,255,255}}{,\strut} \setlength{\fboxsep}{0pt}\def\cbRGB{\colorbox[RGB]}\expandafter\cbRGB\expandafter{\detokenize{255,255,255}}{or\strut} \setlength{\fboxsep}{0pt}\def\cbRGB{\colorbox[RGB]}\expandafter\cbRGB\expandafter{\detokenize{255,255,255}}{the\strut} \setlength{\fboxsep}{0pt}\def\cbRGB{\colorbox[RGB]}\expandafter\cbRGB\expandafter{\detokenize{255,255,255}}{flavors\strut} \setlength{\fboxsep}{0pt}\def\cbRGB{\colorbox[RGB]}\expandafter\cbRGB\expandafter{\detokenize{255,255,255}}{could\strut} \setlength{\fboxsep}{0pt}\def\cbRGB{\colorbox[RGB]}\expandafter\cbRGB\expandafter{\detokenize{255,255,255}}{be\strut} \setlength{\fboxsep}{0pt}\def\cbRGB{\colorbox[RGB]}\expandafter\cbRGB\expandafter{\detokenize{255,255,255}}{``\strut} \setlength{\fboxsep}{0pt}\def\cbRGB{\colorbox[RGB]}\expandafter\cbRGB\expandafter{\detokenize{255,255,255}}{off\strut} \setlength{\fboxsep}{0pt}\def\cbRGB{\colorbox[RGB]}\expandafter\cbRGB\expandafter{\detokenize{255,255,255}}{''\strut} \setlength{\fboxsep}{0pt}\def\cbRGB{\colorbox[RGB]}\expandafter\cbRGB\expandafter{\detokenize{255,255,255}}{from\strut} \setlength{\fboxsep}{0pt}\def\cbRGB{\colorbox[RGB]}\expandafter\cbRGB\expandafter{\detokenize{255,255,255}}{the\strut} \setlength{\fboxsep}{0pt}\def\cbRGB{\colorbox[RGB]}\expandafter\cbRGB\expandafter{\detokenize{255,255,255}}{room\strut} \setlength{\fboxsep}{0pt}\def\cbRGB{\colorbox[RGB]}\expandafter\cbRGB\expandafter{\detokenize{255,255,255}}{temp\strut} \setlength{\fboxsep}{0pt}\def\cbRGB{\colorbox[RGB]}\expandafter\cbRGB\expandafter{\detokenize{255,255,255}}{unk\strut} \setlength{\fboxsep}{0pt}\def\cbRGB{\colorbox[RGB]}\expandafter\cbRGB\expandafter{\detokenize{255,255,255}}{.\strut} 

\setlength{\fboxsep}{0pt}\def\cbRGB{\colorbox[RGB]}\expandafter\cbRGB\expandafter{\detokenize{255,255,77}}{poured\strut} \setlength{\fboxsep}{0pt}\def\cbRGB{\colorbox[RGB]}\expandafter\cbRGB\expandafter{\detokenize{255,255,168}}{into\strut} \setlength{\fboxsep}{0pt}\def\cbRGB{\colorbox[RGB]}\expandafter\cbRGB\expandafter{\detokenize{255,255,229}}{a\strut} \setlength{\fboxsep}{0pt}\def\cbRGB{\colorbox[RGB]}\expandafter\cbRGB\expandafter{\detokenize{255,255,233}}{nonic\strut} \setlength{\fboxsep}{0pt}\def\cbRGB{\colorbox[RGB]}\expandafter\cbRGB\expandafter{\detokenize{255,255,208}}{pint\strut} \setlength{\fboxsep}{0pt}\def\cbRGB{\colorbox[RGB]}\expandafter\cbRGB\expandafter{\detokenize{255,255,208}}{glass\strut} \setlength{\fboxsep}{0pt}\def\cbRGB{\colorbox[RGB]}\expandafter\cbRGB\expandafter{\detokenize{255,255,208}}{{\ldots}\strut} \setlength{\fboxsep}{0pt}\def\cbRGB{\colorbox[RGB]}\expandafter\cbRGB\expandafter{\detokenize{255,255,207}}{jet\strut} \setlength{\fboxsep}{0pt}\def\cbRGB{\colorbox[RGB]}\expandafter\cbRGB\expandafter{\detokenize{255,255,255}}{black\strut} \setlength{\fboxsep}{0pt}\def\cbRGB{\colorbox[RGB]}\expandafter\cbRGB\expandafter{\detokenize{255,255,255}}{,\strut} \setlength{\fboxsep}{0pt}\def\cbRGB{\colorbox[RGB]}\expandafter\cbRGB\expandafter{\detokenize{255,255,255}}{unk\strut} \setlength{\fboxsep}{0pt}\def\cbRGB{\colorbox[RGB]}\expandafter\cbRGB\expandafter{\detokenize{255,255,255}}{no\strut} \setlength{\fboxsep}{0pt}\def\cbRGB{\colorbox[RGB]}\expandafter\cbRGB\expandafter{\detokenize{255,255,255}}{highlights\strut} \setlength{\fboxsep}{0pt}\def\cbRGB{\colorbox[RGB]}\expandafter\cbRGB\expandafter{\detokenize{255,255,255}}{around\strut} \setlength{\fboxsep}{0pt}\def\cbRGB{\colorbox[RGB]}\expandafter\cbRGB\expandafter{\detokenize{255,255,255}}{the\strut} \setlength{\fboxsep}{0pt}\def\cbRGB{\colorbox[RGB]}\expandafter\cbRGB\expandafter{\detokenize{255,255,255}}{edges\strut} \setlength{\fboxsep}{0pt}\def\cbRGB{\colorbox[RGB]}\expandafter\cbRGB\expandafter{\detokenize{255,255,255}}{at\strut} \setlength{\fboxsep}{0pt}\def\cbRGB{\colorbox[RGB]}\expandafter\cbRGB\expandafter{\detokenize{255,255,255}}{all\strut} \setlength{\fboxsep}{0pt}\def\cbRGB{\colorbox[RGB]}\expandafter\cbRGB\expandafter{\detokenize{255,255,255}}{,\strut} \setlength{\fboxsep}{0pt}\def\cbRGB{\colorbox[RGB]}\expandafter\cbRGB\expandafter{\detokenize{255,255,255}}{dense\strut} \setlength{\fboxsep}{0pt}\def\cbRGB{\colorbox[RGB]}\expandafter\cbRGB\expandafter{\detokenize{255,255,255}}{tan\strut} \setlength{\fboxsep}{0pt}\def\cbRGB{\colorbox[RGB]}\expandafter\cbRGB\expandafter{\detokenize{255,255,255}}{unk\strut} \setlength{\fboxsep}{0pt}\def\cbRGB{\colorbox[RGB]}\expandafter\cbRGB\expandafter{\detokenize{255,255,255}}{head\strut} \setlength{\fboxsep}{0pt}\def\cbRGB{\colorbox[RGB]}\expandafter\cbRGB\expandafter{\detokenize{255,255,241}}{{\ldots}\strut} \setlength{\fboxsep}{0pt}\def\cbRGB{\colorbox[RGB]}\expandafter\cbRGB\expandafter{\detokenize{255,255,205}}{looks\strut} \setlength{\fboxsep}{0pt}\def\cbRGB{\colorbox[RGB]}\expandafter\cbRGB\expandafter{\detokenize{255,255,222}}{great\strut} \setlength{\fboxsep}{0pt}\def\cbRGB{\colorbox[RGB]}\expandafter\cbRGB\expandafter{\detokenize{255,255,255}}{!\strut} \setlength{\fboxsep}{0pt}\def\cbRGB{\colorbox[RGB]}\expandafter\cbRGB\expandafter{\detokenize{255,255,255}}{a\strut} \setlength{\fboxsep}{0pt}\def\cbRGB{\colorbox[RGB]}\expandafter\cbRGB\expandafter{\detokenize{255,255,255}}{bit\strut} \setlength{\fboxsep}{0pt}\def\cbRGB{\colorbox[RGB]}\expandafter\cbRGB\expandafter{\detokenize{255,255,255}}{of\strut} \setlength{\fboxsep}{0pt}\def\cbRGB{\colorbox[RGB]}\expandafter\cbRGB\expandafter{\detokenize{255,255,255}}{sediment\strut} \setlength{\fboxsep}{0pt}\def\cbRGB{\colorbox[RGB]}\expandafter\cbRGB\expandafter{\detokenize{255,255,255}}{in\strut} \setlength{\fboxsep}{0pt}\def\cbRGB{\colorbox[RGB]}\expandafter\cbRGB\expandafter{\detokenize{255,255,217}}{the\strut} \setlength{\fboxsep}{0pt}\def\cbRGB{\colorbox[RGB]}\expandafter\cbRGB\expandafter{\detokenize{255,255,220}}{bottom\strut} \setlength{\fboxsep}{0pt}\def\cbRGB{\colorbox[RGB]}\expandafter\cbRGB\expandafter{\detokenize{255,255,235}}{of\strut} \setlength{\fboxsep}{0pt}\def\cbRGB{\colorbox[RGB]}\expandafter\cbRGB\expandafter{\detokenize{255,255,251}}{the\strut} \setlength{\fboxsep}{0pt}\def\cbRGB{\colorbox[RGB]}\expandafter\cbRGB\expandafter{\detokenize{255,255,255}}{bottle\strut} \setlength{\fboxsep}{0pt}\def\cbRGB{\colorbox[RGB]}\expandafter\cbRGB\expandafter{\detokenize{255,255,255}}{.\strut} \setlength{\fboxsep}{0pt}\def\cbRGB{\colorbox[RGB]}\expandafter\cbRGB\expandafter{\detokenize{255,255,255}}{bottle\strut} \setlength{\fboxsep}{0pt}\def\cbRGB{\colorbox[RGB]}\expandafter\cbRGB\expandafter{\detokenize{255,255,255}}{read\strut} \setlength{\fboxsep}{0pt}\def\cbRGB{\colorbox[RGB]}\expandafter\cbRGB\expandafter{\detokenize{255,255,255}}{``\strut} \setlength{\fboxsep}{0pt}\def\cbRGB{\colorbox[RGB]}\expandafter\cbRGB\expandafter{\detokenize{255,255,255}}{live\strut} \setlength{\fboxsep}{0pt}\def\cbRGB{\colorbox[RGB]}\expandafter\cbRGB\expandafter{\detokenize{255,255,255}}{ale\strut} \setlength{\fboxsep}{0pt}\def\cbRGB{\colorbox[RGB]}\expandafter\cbRGB\expandafter{\detokenize{255,255,255}}{,\strut} \setlength{\fboxsep}{0pt}\def\cbRGB{\colorbox[RGB]}\expandafter\cbRGB\expandafter{\detokenize{255,255,255}}{keep\strut} \setlength{\fboxsep}{0pt}\def\cbRGB{\colorbox[RGB]}\expandafter\cbRGB\expandafter{\detokenize{255,255,255}}{unk\strut} \setlength{\fboxsep}{0pt}\def\cbRGB{\colorbox[RGB]}\expandafter\cbRGB\expandafter{\detokenize{255,255,255}}{.\strut} \setlength{\fboxsep}{0pt}\def\cbRGB{\colorbox[RGB]}\expandafter\cbRGB\expandafter{\detokenize{255,255,193}}{''\strut} \setlength{\fboxsep}{0pt}\def\cbRGB{\colorbox[RGB]}\expandafter\cbRGB\expandafter{\detokenize{255,255,133}}{the\strut} \setlength{\fboxsep}{0pt}\def\cbRGB{\colorbox[RGB]}\expandafter\cbRGB\expandafter{\detokenize{255,255,113}}{store\strut} \setlength{\fboxsep}{0pt}\def\cbRGB{\colorbox[RGB]}\expandafter\cbRGB\expandafter{\detokenize{255,255,151}}{where\strut} \setlength{\fboxsep}{0pt}\def\cbRGB{\colorbox[RGB]}\expandafter\cbRGB\expandafter{\detokenize{255,255,215}}{i\strut} \setlength{\fboxsep}{0pt}\def\cbRGB{\colorbox[RGB]}\expandafter\cbRGB\expandafter{\detokenize{255,255,255}}{bought\strut} \setlength{\fboxsep}{0pt}\def\cbRGB{\colorbox[RGB]}\expandafter\cbRGB\expandafter{\detokenize{255,255,255}}{it\strut} \setlength{\fboxsep}{0pt}\def\cbRGB{\colorbox[RGB]}\expandafter\cbRGB\expandafter{\detokenize{255,255,255}}{from\strut} \setlength{\fboxsep}{0pt}\def\cbRGB{\colorbox[RGB]}\expandafter\cbRGB\expandafter{\detokenize{255,255,255}}{had\strut} \setlength{\fboxsep}{0pt}\def\cbRGB{\colorbox[RGB]}\expandafter\cbRGB\expandafter{\detokenize{255,255,255}}{it\strut} \setlength{\fboxsep}{0pt}\def\cbRGB{\colorbox[RGB]}\expandafter\cbRGB\expandafter{\detokenize{255,255,255}}{on\strut} \setlength{\fboxsep}{0pt}\def\cbRGB{\colorbox[RGB]}\expandafter\cbRGB\expandafter{\detokenize{255,255,255}}{the\strut} \setlength{\fboxsep}{0pt}\def\cbRGB{\colorbox[RGB]}\expandafter\cbRGB\expandafter{\detokenize{255,255,255}}{shelf\strut} \setlength{\fboxsep}{0pt}\def\cbRGB{\colorbox[RGB]}\expandafter\cbRGB\expandafter{\detokenize{255,255,255}}{at\strut} \setlength{\fboxsep}{0pt}\def\cbRGB{\colorbox[RGB]}\expandafter\cbRGB\expandafter{\detokenize{255,255,255}}{room\strut} \setlength{\fboxsep}{0pt}\def\cbRGB{\colorbox[RGB]}\expandafter\cbRGB\expandafter{\detokenize{255,255,255}}{temp\strut} \setlength{\fboxsep}{0pt}\def\cbRGB{\colorbox[RGB]}\expandafter\cbRGB\expandafter{\detokenize{255,255,255}}{:\strut} \setlength{\fboxsep}{0pt}\def\cbRGB{\colorbox[RGB]}\expandafter\cbRGB\expandafter{\detokenize{255,255,255}}{(\strut} \setlength{\fboxsep}{0pt}\def\cbRGB{\colorbox[RGB]}\expandafter\cbRGB\expandafter{\detokenize{255,255,255}}{the\strut} \setlength{\fboxsep}{0pt}\def\cbRGB{\colorbox[RGB]}\expandafter\cbRGB\expandafter{\detokenize{255,255,255}}{smell\strut} \setlength{\fboxsep}{0pt}\def\cbRGB{\colorbox[RGB]}\expandafter\cbRGB\expandafter{\detokenize{255,255,255}}{was\strut} \setlength{\fboxsep}{0pt}\def\cbRGB{\colorbox[RGB]}\expandafter\cbRGB\expandafter{\detokenize{255,255,255}}{surprising\strut} \setlength{\fboxsep}{0pt}\def\cbRGB{\colorbox[RGB]}\expandafter\cbRGB\expandafter{\detokenize{255,255,255}}{{\ldots}\strut} \setlength{\fboxsep}{0pt}\def\cbRGB{\colorbox[RGB]}\expandafter\cbRGB\expandafter{\detokenize{255,255,197}}{an\strut} \setlength{\fboxsep}{0pt}\def\cbRGB{\colorbox[RGB]}\expandafter\cbRGB\expandafter{\detokenize{255,255,118}}{earthy\strut} \setlength{\fboxsep}{0pt}\def\cbRGB{\colorbox[RGB]}\expandafter\cbRGB\expandafter{\detokenize{255,255,38}}{roasted\strut} \setlength{\fboxsep}{0pt}\def\cbRGB{\colorbox[RGB]}\expandafter\cbRGB\expandafter{\detokenize{255,255,15}}{smell\strut} \setlength{\fboxsep}{0pt}\def\cbRGB{\colorbox[RGB]}\expandafter\cbRGB\expandafter{\detokenize{255,255,39}}{,\strut} \setlength{\fboxsep}{0pt}\def\cbRGB{\colorbox[RGB]}\expandafter\cbRGB\expandafter{\detokenize{255,255,66}}{mixed\strut} \setlength{\fboxsep}{0pt}\def\cbRGB{\colorbox[RGB]}\expandafter\cbRGB\expandafter{\detokenize{255,255,125}}{with\strut} \setlength{\fboxsep}{0pt}\def\cbRGB{\colorbox[RGB]}\expandafter\cbRGB\expandafter{\detokenize{255,255,186}}{day\strut} \setlength{\fboxsep}{0pt}\def\cbRGB{\colorbox[RGB]}\expandafter\cbRGB\expandafter{\detokenize{255,255,219}}{old\strut} \setlength{\fboxsep}{0pt}\def\cbRGB{\colorbox[RGB]}\expandafter\cbRGB\expandafter{\detokenize{255,255,239}}{coffee\strut} \setlength{\fboxsep}{0pt}\def\cbRGB{\colorbox[RGB]}\expandafter\cbRGB\expandafter{\detokenize{255,255,255}}{.\strut} \setlength{\fboxsep}{0pt}\def\cbRGB{\colorbox[RGB]}\expandafter\cbRGB\expandafter{\detokenize{255,255,255}}{also\strut} \setlength{\fboxsep}{0pt}\def\cbRGB{\colorbox[RGB]}\expandafter\cbRGB\expandafter{\detokenize{255,255,255}}{some\strut} \setlength{\fboxsep}{0pt}\def\cbRGB{\colorbox[RGB]}\expandafter\cbRGB\expandafter{\detokenize{255,255,255}}{weird\strut} \setlength{\fboxsep}{0pt}\def\cbRGB{\colorbox[RGB]}\expandafter\cbRGB\expandafter{\detokenize{255,255,255}}{``\strut} \setlength{\fboxsep}{0pt}\def\cbRGB{\colorbox[RGB]}\expandafter\cbRGB\expandafter{\detokenize{255,255,255}}{off\strut} \setlength{\fboxsep}{0pt}\def\cbRGB{\colorbox[RGB]}\expandafter\cbRGB\expandafter{\detokenize{255,255,255}}{''\strut} \setlength{\fboxsep}{0pt}\def\cbRGB{\colorbox[RGB]}\expandafter\cbRGB\expandafter{\detokenize{255,255,255}}{licorice\strut} \setlength{\fboxsep}{0pt}\def\cbRGB{\colorbox[RGB]}\expandafter\cbRGB\expandafter{\detokenize{255,255,255}}{notes\strut} \setlength{\fboxsep}{0pt}\def\cbRGB{\colorbox[RGB]}\expandafter\cbRGB\expandafter{\detokenize{255,255,236}}{.\strut} \setlength{\fboxsep}{0pt}\def\cbRGB{\colorbox[RGB]}\expandafter\cbRGB\expandafter{\detokenize{255,255,203}}{taste\strut} \setlength{\fboxsep}{0pt}\def\cbRGB{\colorbox[RGB]}\expandafter\cbRGB\expandafter{\detokenize{255,255,197}}{was\strut} \setlength{\fboxsep}{0pt}\def\cbRGB{\colorbox[RGB]}\expandafter\cbRGB\expandafter{\detokenize{255,255,232}}{dry\strut} \setlength{\fboxsep}{0pt}\def\cbRGB{\colorbox[RGB]}\expandafter\cbRGB\expandafter{\detokenize{255,255,255}}{at\strut} \setlength{\fboxsep}{0pt}\def\cbRGB{\colorbox[RGB]}\expandafter\cbRGB\expandafter{\detokenize{255,255,255}}{the\strut} \setlength{\fboxsep}{0pt}\def\cbRGB{\colorbox[RGB]}\expandafter\cbRGB\expandafter{\detokenize{255,255,255}}{start\strut} \setlength{\fboxsep}{0pt}\def\cbRGB{\colorbox[RGB]}\expandafter\cbRGB\expandafter{\detokenize{255,255,255}}{,\strut} \setlength{\fboxsep}{0pt}\def\cbRGB{\colorbox[RGB]}\expandafter\cbRGB\expandafter{\detokenize{255,255,255}}{and\strut} \setlength{\fboxsep}{0pt}\def\cbRGB{\colorbox[RGB]}\expandafter\cbRGB\expandafter{\detokenize{255,255,255}}{very\strut} \setlength{\fboxsep}{0pt}\def\cbRGB{\colorbox[RGB]}\expandafter\cbRGB\expandafter{\detokenize{255,255,255}}{dry\strut} \setlength{\fboxsep}{0pt}\def\cbRGB{\colorbox[RGB]}\expandafter\cbRGB\expandafter{\detokenize{255,255,255}}{on\strut} \setlength{\fboxsep}{0pt}\def\cbRGB{\colorbox[RGB]}\expandafter\cbRGB\expandafter{\detokenize{255,255,234}}{the\strut} \setlength{\fboxsep}{0pt}\def\cbRGB{\colorbox[RGB]}\expandafter\cbRGB\expandafter{\detokenize{255,255,196}}{finish\strut} \setlength{\fboxsep}{0pt}\def\cbRGB{\colorbox[RGB]}\expandafter\cbRGB\expandafter{\detokenize{255,255,160}}{.\strut} \setlength{\fboxsep}{0pt}\def\cbRGB{\colorbox[RGB]}\expandafter\cbRGB\expandafter{\detokenize{255,255,119}}{bitter\strut} \setlength{\fboxsep}{0pt}\def\cbRGB{\colorbox[RGB]}\expandafter\cbRGB\expandafter{\detokenize{255,255,120}}{roast\strut} \setlength{\fboxsep}{0pt}\def\cbRGB{\colorbox[RGB]}\expandafter\cbRGB\expandafter{\detokenize{255,255,173}}{notes\strut} \setlength{\fboxsep}{0pt}\def\cbRGB{\colorbox[RGB]}\expandafter\cbRGB\expandafter{\detokenize{255,255,222}}{with\strut} \setlength{\fboxsep}{0pt}\def\cbRGB{\colorbox[RGB]}\expandafter\cbRGB\expandafter{\detokenize{255,255,245}}{a\strut} \setlength{\fboxsep}{0pt}\def\cbRGB{\colorbox[RGB]}\expandafter\cbRGB\expandafter{\detokenize{255,255,237}}{sort\strut} \setlength{\fboxsep}{0pt}\def\cbRGB{\colorbox[RGB]}\expandafter\cbRGB\expandafter{\detokenize{255,255,215}}{of\strut} \setlength{\fboxsep}{0pt}\def\cbRGB{\colorbox[RGB]}\expandafter\cbRGB\expandafter{\detokenize{255,255,163}}{unk\strut} \setlength{\fboxsep}{0pt}\def\cbRGB{\colorbox[RGB]}\expandafter\cbRGB\expandafter{\detokenize{255,255,91}}{.\strut} \setlength{\fboxsep}{0pt}\def\cbRGB{\colorbox[RGB]}\expandafter\cbRGB\expandafter{\detokenize{255,255,62}}{full\strut} \setlength{\fboxsep}{0pt}\def\cbRGB{\colorbox[RGB]}\expandafter\cbRGB\expandafter{\detokenize{255,255,92}}{bodied\strut} \setlength{\fboxsep}{0pt}\def\cbRGB{\colorbox[RGB]}\expandafter\cbRGB\expandafter{\detokenize{255,255,116}}{with\strut} \setlength{\fboxsep}{0pt}\def\cbRGB{\colorbox[RGB]}\expandafter\cbRGB\expandafter{\detokenize{255,255,145}}{appropriate\strut} \setlength{\fboxsep}{0pt}\def\cbRGB{\colorbox[RGB]}\expandafter\cbRGB\expandafter{\detokenize{255,255,188}}{carbonation\strut} \setlength{\fboxsep}{0pt}\def\cbRGB{\colorbox[RGB]}\expandafter\cbRGB\expandafter{\detokenize{255,255,202}}{.\strut} \setlength{\fboxsep}{0pt}\def\cbRGB{\colorbox[RGB]}\expandafter\cbRGB\expandafter{\detokenize{255,255,168}}{i\strut} \setlength{\fboxsep}{0pt}\def\cbRGB{\colorbox[RGB]}\expandafter\cbRGB\expandafter{\detokenize{255,255,129}}{was\strut} \setlength{\fboxsep}{0pt}\def\cbRGB{\colorbox[RGB]}\expandafter\cbRGB\expandafter{\detokenize{255,255,88}}{very\strut} \setlength{\fboxsep}{0pt}\def\cbRGB{\colorbox[RGB]}\expandafter\cbRGB\expandafter{\detokenize{255,255,72}}{excited\strut} \setlength{\fboxsep}{0pt}\def\cbRGB{\colorbox[RGB]}\expandafter\cbRGB\expandafter{\detokenize{255,255,87}}{to\strut} \setlength{\fboxsep}{0pt}\def\cbRGB{\colorbox[RGB]}\expandafter\cbRGB\expandafter{\detokenize{255,255,140}}{try\strut} \setlength{\fboxsep}{0pt}\def\cbRGB{\colorbox[RGB]}\expandafter\cbRGB\expandafter{\detokenize{255,255,202}}{this\strut} \setlength{\fboxsep}{0pt}\def\cbRGB{\colorbox[RGB]}\expandafter\cbRGB\expandafter{\detokenize{255,255,255}}{beer\strut} \setlength{\fboxsep}{0pt}\def\cbRGB{\colorbox[RGB]}\expandafter\cbRGB\expandafter{\detokenize{255,255,255}}{,\strut} \setlength{\fboxsep}{0pt}\def\cbRGB{\colorbox[RGB]}\expandafter\cbRGB\expandafter{\detokenize{255,255,255}}{and\strut} \setlength{\fboxsep}{0pt}\def\cbRGB{\colorbox[RGB]}\expandafter\cbRGB\expandafter{\detokenize{255,255,255}}{i\strut} \setlength{\fboxsep}{0pt}\def\cbRGB{\colorbox[RGB]}\expandafter\cbRGB\expandafter{\detokenize{255,255,228}}{was\strut} \setlength{\fboxsep}{0pt}\def\cbRGB{\colorbox[RGB]}\expandafter\cbRGB\expandafter{\detokenize{255,255,159}}{pretty\strut} \setlength{\fboxsep}{0pt}\def\cbRGB{\colorbox[RGB]}\expandafter\cbRGB\expandafter{\detokenize{255,255,53}}{disappointed\strut} \setlength{\fboxsep}{0pt}\def\cbRGB{\colorbox[RGB]}\expandafter\cbRGB\expandafter{\detokenize{255,255,0}}{.\strut} \setlength{\fboxsep}{0pt}\def\cbRGB{\colorbox[RGB]}\expandafter\cbRGB\expandafter{\detokenize{255,255,2}}{this\strut} \setlength{\fboxsep}{0pt}\def\cbRGB{\colorbox[RGB]}\expandafter\cbRGB\expandafter{\detokenize{255,255,23}}{bottle\strut} \setlength{\fboxsep}{0pt}\def\cbRGB{\colorbox[RGB]}\expandafter\cbRGB\expandafter{\detokenize{255,255,44}}{could\strut} \setlength{\fboxsep}{0pt}\def\cbRGB{\colorbox[RGB]}\expandafter\cbRGB\expandafter{\detokenize{255,255,104}}{be\strut} \setlength{\fboxsep}{0pt}\def\cbRGB{\colorbox[RGB]}\expandafter\cbRGB\expandafter{\detokenize{255,255,141}}{old\strut} \setlength{\fboxsep}{0pt}\def\cbRGB{\colorbox[RGB]}\expandafter\cbRGB\expandafter{\detokenize{255,255,156}}{,\strut} \setlength{\fboxsep}{0pt}\def\cbRGB{\colorbox[RGB]}\expandafter\cbRGB\expandafter{\detokenize{255,255,161}}{or\strut} \setlength{\fboxsep}{0pt}\def\cbRGB{\colorbox[RGB]}\expandafter\cbRGB\expandafter{\detokenize{255,255,203}}{the\strut} \setlength{\fboxsep}{0pt}\def\cbRGB{\colorbox[RGB]}\expandafter\cbRGB\expandafter{\detokenize{255,255,252}}{flavors\strut} \setlength{\fboxsep}{0pt}\def\cbRGB{\colorbox[RGB]}\expandafter\cbRGB\expandafter{\detokenize{255,255,255}}{could\strut} \setlength{\fboxsep}{0pt}\def\cbRGB{\colorbox[RGB]}\expandafter\cbRGB\expandafter{\detokenize{255,255,255}}{be\strut} \setlength{\fboxsep}{0pt}\def\cbRGB{\colorbox[RGB]}\expandafter\cbRGB\expandafter{\detokenize{255,255,255}}{``\strut} \setlength{\fboxsep}{0pt}\def\cbRGB{\colorbox[RGB]}\expandafter\cbRGB\expandafter{\detokenize{255,255,255}}{off\strut} \setlength{\fboxsep}{0pt}\def\cbRGB{\colorbox[RGB]}\expandafter\cbRGB\expandafter{\detokenize{255,255,255}}{''\strut} \setlength{\fboxsep}{0pt}\def\cbRGB{\colorbox[RGB]}\expandafter\cbRGB\expandafter{\detokenize{255,255,255}}{from\strut} \setlength{\fboxsep}{0pt}\def\cbRGB{\colorbox[RGB]}\expandafter\cbRGB\expandafter{\detokenize{255,255,255}}{the\strut} \setlength{\fboxsep}{0pt}\def\cbRGB{\colorbox[RGB]}\expandafter\cbRGB\expandafter{\detokenize{255,255,255}}{room\strut} \setlength{\fboxsep}{0pt}\def\cbRGB{\colorbox[RGB]}\expandafter\cbRGB\expandafter{\detokenize{255,255,255}}{temp\strut} \setlength{\fboxsep}{0pt}\def\cbRGB{\colorbox[RGB]}\expandafter\cbRGB\expandafter{\detokenize{255,255,255}}{unk\strut} \setlength{\fboxsep}{0pt}\def\cbRGB{\colorbox[RGB]}\expandafter\cbRGB\expandafter{\detokenize{255,255,255}}{.\strut} 

\par
\textbf{Example 3}

\setlength{\fboxsep}{0pt}\def\cbRGB{\colorbox[RGB]}\expandafter\cbRGB\expandafter{\detokenize{255,237,237}}{this\strut} \setlength{\fboxsep}{0pt}\def\cbRGB{\colorbox[RGB]}\expandafter\cbRGB\expandafter{\detokenize{255,222,222}}{is\strut} \setlength{\fboxsep}{0pt}\def\cbRGB{\colorbox[RGB]}\expandafter\cbRGB\expandafter{\detokenize{255,208,208}}{a\strut} \setlength{\fboxsep}{0pt}\def\cbRGB{\colorbox[RGB]}\expandafter\cbRGB\expandafter{\detokenize{255,242,242}}{real\strut} \setlength{\fboxsep}{0pt}\def\cbRGB{\colorbox[RGB]}\expandafter\cbRGB\expandafter{\detokenize{255,252,252}}{``\strut} \setlength{\fboxsep}{0pt}\def\cbRGB{\colorbox[RGB]}\expandafter\cbRGB\expandafter{\detokenize{255,252,252}}{nothing\strut} \setlength{\fboxsep}{0pt}\def\cbRGB{\colorbox[RGB]}\expandafter\cbRGB\expandafter{\detokenize{255,255,255}}{''\strut} \setlength{\fboxsep}{0pt}\def\cbRGB{\colorbox[RGB]}\expandafter\cbRGB\expandafter{\detokenize{255,255,255}}{beer\strut} \setlength{\fboxsep}{0pt}\def\cbRGB{\colorbox[RGB]}\expandafter\cbRGB\expandafter{\detokenize{255,225,225}}{.\strut} \setlength{\fboxsep}{0pt}\def\cbRGB{\colorbox[RGB]}\expandafter\cbRGB\expandafter{\detokenize{255,170,170}}{pours\strut} \setlength{\fboxsep}{0pt}\def\cbRGB{\colorbox[RGB]}\expandafter\cbRGB\expandafter{\detokenize{255,108,108}}{a\strut} \setlength{\fboxsep}{0pt}\def\cbRGB{\colorbox[RGB]}\expandafter\cbRGB\expandafter{\detokenize{255,31,31}}{unk\strut} \setlength{\fboxsep}{0pt}\def\cbRGB{\colorbox[RGB]}\expandafter\cbRGB\expandafter{\detokenize{255,0,0}}{yellow\strut} \setlength{\fboxsep}{0pt}\def\cbRGB{\colorbox[RGB]}\expandafter\cbRGB\expandafter{\detokenize{255,15,15}}{color\strut} \setlength{\fboxsep}{0pt}\def\cbRGB{\colorbox[RGB]}\expandafter\cbRGB\expandafter{\detokenize{255,77,77}}{,\strut} \setlength{\fboxsep}{0pt}\def\cbRGB{\colorbox[RGB]}\expandafter\cbRGB\expandafter{\detokenize{255,115,115}}{looking\strut} \setlength{\fboxsep}{0pt}\def\cbRGB{\colorbox[RGB]}\expandafter\cbRGB\expandafter{\detokenize{255,170,170}}{more\strut} \setlength{\fboxsep}{0pt}\def\cbRGB{\colorbox[RGB]}\expandafter\cbRGB\expandafter{\detokenize{255,201,201}}{like\strut} \setlength{\fboxsep}{0pt}\def\cbRGB{\colorbox[RGB]}\expandafter\cbRGB\expandafter{\detokenize{255,208,208}}{unk\strut} \setlength{\fboxsep}{0pt}\def\cbRGB{\colorbox[RGB]}\expandafter\cbRGB\expandafter{\detokenize{255,193,193}}{water\strut} \setlength{\fboxsep}{0pt}\def\cbRGB{\colorbox[RGB]}\expandafter\cbRGB\expandafter{\detokenize{255,224,224}}{than\strut} \setlength{\fboxsep}{0pt}\def\cbRGB{\colorbox[RGB]}\expandafter\cbRGB\expandafter{\detokenize{255,248,248}}{beer\strut} \setlength{\fboxsep}{0pt}\def\cbRGB{\colorbox[RGB]}\expandafter\cbRGB\expandafter{\detokenize{255,255,255}}{.\strut} \setlength{\fboxsep}{0pt}\def\cbRGB{\colorbox[RGB]}\expandafter\cbRGB\expandafter{\detokenize{255,255,255}}{strange\strut} \setlength{\fboxsep}{0pt}\def\cbRGB{\colorbox[RGB]}\expandafter\cbRGB\expandafter{\detokenize{255,255,255}}{unk\strut} \setlength{\fboxsep}{0pt}\def\cbRGB{\colorbox[RGB]}\expandafter\cbRGB\expandafter{\detokenize{255,255,255}}{smell\strut} \setlength{\fboxsep}{0pt}\def\cbRGB{\colorbox[RGB]}\expandafter\cbRGB\expandafter{\detokenize{255,255,255}}{,\strut} \setlength{\fboxsep}{0pt}\def\cbRGB{\colorbox[RGB]}\expandafter\cbRGB\expandafter{\detokenize{255,255,255}}{with\strut} \setlength{\fboxsep}{0pt}\def\cbRGB{\colorbox[RGB]}\expandafter\cbRGB\expandafter{\detokenize{255,255,255}}{minimal\strut} \setlength{\fboxsep}{0pt}\def\cbRGB{\colorbox[RGB]}\expandafter\cbRGB\expandafter{\detokenize{255,255,255}}{unk\strut} \setlength{\fboxsep}{0pt}\def\cbRGB{\colorbox[RGB]}\expandafter\cbRGB\expandafter{\detokenize{255,255,255}}{hop\strut} \setlength{\fboxsep}{0pt}\def\cbRGB{\colorbox[RGB]}\expandafter\cbRGB\expandafter{\detokenize{255,255,255}}{aroma\strut} \setlength{\fboxsep}{0pt}\def\cbRGB{\colorbox[RGB]}\expandafter\cbRGB\expandafter{\detokenize{255,255,255}}{unk\strut} \setlength{\fboxsep}{0pt}\def\cbRGB{\colorbox[RGB]}\expandafter\cbRGB\expandafter{\detokenize{255,255,255}}{away\strut} \setlength{\fboxsep}{0pt}\def\cbRGB{\colorbox[RGB]}\expandafter\cbRGB\expandafter{\detokenize{255,255,255}}{behind\strut} \setlength{\fboxsep}{0pt}\def\cbRGB{\colorbox[RGB]}\expandafter\cbRGB\expandafter{\detokenize{255,255,255}}{whatever\strut} \setlength{\fboxsep}{0pt}\def\cbRGB{\colorbox[RGB]}\expandafter\cbRGB\expandafter{\detokenize{255,255,255}}{it\strut} \setlength{\fboxsep}{0pt}\def\cbRGB{\colorbox[RGB]}\expandafter\cbRGB\expandafter{\detokenize{255,255,255}}{is\strut} \setlength{\fboxsep}{0pt}\def\cbRGB{\colorbox[RGB]}\expandafter\cbRGB\expandafter{\detokenize{255,255,255}}{that\strut} \setlength{\fboxsep}{0pt}\def\cbRGB{\colorbox[RGB]}\expandafter\cbRGB\expandafter{\detokenize{255,255,255}}{unk\strut} \setlength{\fboxsep}{0pt}\def\cbRGB{\colorbox[RGB]}\expandafter\cbRGB\expandafter{\detokenize{255,255,255}}{in\strut} \setlength{\fboxsep}{0pt}\def\cbRGB{\colorbox[RGB]}\expandafter\cbRGB\expandafter{\detokenize{255,255,255}}{this\strut} \setlength{\fboxsep}{0pt}\def\cbRGB{\colorbox[RGB]}\expandafter\cbRGB\expandafter{\detokenize{255,255,255}}{brew\strut} \setlength{\fboxsep}{0pt}\def\cbRGB{\colorbox[RGB]}\expandafter\cbRGB\expandafter{\detokenize{255,255,255}}{.\strut} \setlength{\fboxsep}{0pt}\def\cbRGB{\colorbox[RGB]}\expandafter\cbRGB\expandafter{\detokenize{255,255,255}}{taste\strut} \setlength{\fboxsep}{0pt}\def\cbRGB{\colorbox[RGB]}\expandafter\cbRGB\expandafter{\detokenize{255,255,255}}{is\strut} \setlength{\fboxsep}{0pt}\def\cbRGB{\colorbox[RGB]}\expandafter\cbRGB\expandafter{\detokenize{255,255,255}}{equally\strut} \setlength{\fboxsep}{0pt}\def\cbRGB{\colorbox[RGB]}\expandafter\cbRGB\expandafter{\detokenize{255,255,255}}{as\strut} \setlength{\fboxsep}{0pt}\def\cbRGB{\colorbox[RGB]}\expandafter\cbRGB\expandafter{\detokenize{255,255,255}}{unk\strut} \setlength{\fboxsep}{0pt}\def\cbRGB{\colorbox[RGB]}\expandafter\cbRGB\expandafter{\detokenize{255,255,255}}{.\strut} \setlength{\fboxsep}{0pt}\def\cbRGB{\colorbox[RGB]}\expandafter\cbRGB\expandafter{\detokenize{255,255,255}}{nearly\strut} \setlength{\fboxsep}{0pt}\def\cbRGB{\colorbox[RGB]}\expandafter\cbRGB\expandafter{\detokenize{255,255,255}}{unk\strut} \setlength{\fboxsep}{0pt}\def\cbRGB{\colorbox[RGB]}\expandafter\cbRGB\expandafter{\detokenize{255,255,255}}{,\strut} \setlength{\fboxsep}{0pt}\def\cbRGB{\colorbox[RGB]}\expandafter\cbRGB\expandafter{\detokenize{255,244,244}}{the\strut} \setlength{\fboxsep}{0pt}\def\cbRGB{\colorbox[RGB]}\expandafter\cbRGB\expandafter{\detokenize{255,204,204}}{most\strut} \setlength{\fboxsep}{0pt}\def\cbRGB{\colorbox[RGB]}\expandafter\cbRGB\expandafter{\detokenize{255,158,158}}{you\strut} \setlength{\fboxsep}{0pt}\def\cbRGB{\colorbox[RGB]}\expandafter\cbRGB\expandafter{\detokenize{255,113,113}}{get\strut} \setlength{\fboxsep}{0pt}\def\cbRGB{\colorbox[RGB]}\expandafter\cbRGB\expandafter{\detokenize{255,114,114}}{from\strut} \setlength{\fboxsep}{0pt}\def\cbRGB{\colorbox[RGB]}\expandafter\cbRGB\expandafter{\detokenize{255,159,159}}{this\strut} \setlength{\fboxsep}{0pt}\def\cbRGB{\colorbox[RGB]}\expandafter\cbRGB\expandafter{\detokenize{255,204,204}}{beer\strut} \setlength{\fboxsep}{0pt}\def\cbRGB{\colorbox[RGB]}\expandafter\cbRGB\expandafter{\detokenize{255,250,250}}{is\strut} \setlength{\fboxsep}{0pt}\def\cbRGB{\colorbox[RGB]}\expandafter\cbRGB\expandafter{\detokenize{255,255,255}}{a\strut} \setlength{\fboxsep}{0pt}\def\cbRGB{\colorbox[RGB]}\expandafter\cbRGB\expandafter{\detokenize{255,255,255}}{slightly\strut} \setlength{\fboxsep}{0pt}\def\cbRGB{\colorbox[RGB]}\expandafter\cbRGB\expandafter{\detokenize{255,255,255}}{sweet\strut} \setlength{\fboxsep}{0pt}\def\cbRGB{\colorbox[RGB]}\expandafter\cbRGB\expandafter{\detokenize{255,255,255}}{unk\strut} \setlength{\fboxsep}{0pt}\def\cbRGB{\colorbox[RGB]}\expandafter\cbRGB\expandafter{\detokenize{255,255,255}}{flavor\strut} \setlength{\fboxsep}{0pt}\def\cbRGB{\colorbox[RGB]}\expandafter\cbRGB\expandafter{\detokenize{255,255,255}}{and\strut} \setlength{\fboxsep}{0pt}\def\cbRGB{\colorbox[RGB]}\expandafter\cbRGB\expandafter{\detokenize{255,255,255}}{a\strut} \setlength{\fboxsep}{0pt}\def\cbRGB{\colorbox[RGB]}\expandafter\cbRGB\expandafter{\detokenize{255,248,248}}{lot\strut} \setlength{\fboxsep}{0pt}\def\cbRGB{\colorbox[RGB]}\expandafter\cbRGB\expandafter{\detokenize{255,225,225}}{of\strut} \setlength{\fboxsep}{0pt}\def\cbRGB{\colorbox[RGB]}\expandafter\cbRGB\expandafter{\detokenize{255,202,202}}{carbonation\strut} \setlength{\fboxsep}{0pt}\def\cbRGB{\colorbox[RGB]}\expandafter\cbRGB\expandafter{\detokenize{255,202,202}}{in\strut} \setlength{\fboxsep}{0pt}\def\cbRGB{\colorbox[RGB]}\expandafter\cbRGB\expandafter{\detokenize{255,202,202}}{the\strut} \setlength{\fboxsep}{0pt}\def\cbRGB{\colorbox[RGB]}\expandafter\cbRGB\expandafter{\detokenize{255,201,201}}{beginning\strut} \setlength{\fboxsep}{0pt}\def\cbRGB{\colorbox[RGB]}\expandafter\cbRGB\expandafter{\detokenize{255,201,201}}{.\strut} \setlength{\fboxsep}{0pt}\def\cbRGB{\colorbox[RGB]}\expandafter\cbRGB\expandafter{\detokenize{255,201,201}}{watery\strut} \setlength{\fboxsep}{0pt}\def\cbRGB{\colorbox[RGB]}\expandafter\cbRGB\expandafter{\detokenize{255,224,224}}{and\strut} \setlength{\fboxsep}{0pt}\def\cbRGB{\colorbox[RGB]}\expandafter\cbRGB\expandafter{\detokenize{255,248,248}}{thin\strut} \setlength{\fboxsep}{0pt}\def\cbRGB{\colorbox[RGB]}\expandafter\cbRGB\expandafter{\detokenize{255,255,255}}{,\strut} \setlength{\fboxsep}{0pt}\def\cbRGB{\colorbox[RGB]}\expandafter\cbRGB\expandafter{\detokenize{255,255,255}}{but\strut} \setlength{\fboxsep}{0pt}\def\cbRGB{\colorbox[RGB]}\expandafter\cbRGB\expandafter{\detokenize{255,255,255}}{something\strut} \setlength{\fboxsep}{0pt}\def\cbRGB{\colorbox[RGB]}\expandafter\cbRGB\expandafter{\detokenize{255,255,255}}{that\strut} \setlength{\fboxsep}{0pt}\def\cbRGB{\colorbox[RGB]}\expandafter\cbRGB\expandafter{\detokenize{255,255,255}}{goes\strut} \setlength{\fboxsep}{0pt}\def\cbRGB{\colorbox[RGB]}\expandafter\cbRGB\expandafter{\detokenize{255,255,255}}{down\strut} \setlength{\fboxsep}{0pt}\def\cbRGB{\colorbox[RGB]}\expandafter\cbRGB\expandafter{\detokenize{255,255,255}}{easy\strut} \setlength{\fboxsep}{0pt}\def\cbRGB{\colorbox[RGB]}\expandafter\cbRGB\expandafter{\detokenize{255,255,255}}{so\strut} \setlength{\fboxsep}{0pt}\def\cbRGB{\colorbox[RGB]}\expandafter\cbRGB\expandafter{\detokenize{255,255,255}}{the\strut} \setlength{\fboxsep}{0pt}\def\cbRGB{\colorbox[RGB]}\expandafter\cbRGB\expandafter{\detokenize{255,255,255}}{drinkability\strut} \setlength{\fboxsep}{0pt}\def\cbRGB{\colorbox[RGB]}\expandafter\cbRGB\expandafter{\detokenize{255,255,255}}{is\strut} \setlength{\fboxsep}{0pt}\def\cbRGB{\colorbox[RGB]}\expandafter\cbRGB\expandafter{\detokenize{255,255,255}}{unk\strut} \setlength{\fboxsep}{0pt}\def\cbRGB{\colorbox[RGB]}\expandafter\cbRGB\expandafter{\detokenize{255,255,255}}{up\strut} \setlength{\fboxsep}{0pt}\def\cbRGB{\colorbox[RGB]}\expandafter\cbRGB\expandafter{\detokenize{255,255,255}}{slightly\strut} \setlength{\fboxsep}{0pt}\def\cbRGB{\colorbox[RGB]}\expandafter\cbRGB\expandafter{\detokenize{255,255,255}}{.\strut} \setlength{\fboxsep}{0pt}\def\cbRGB{\colorbox[RGB]}\expandafter\cbRGB\expandafter{\detokenize{255,255,255}}{unk\strut} \setlength{\fboxsep}{0pt}\def\cbRGB{\colorbox[RGB]}\expandafter\cbRGB\expandafter{\detokenize{255,255,255}}{.\strut} 

\setlength{\fboxsep}{0pt}\def\cbRGB{\colorbox[RGB]}\expandafter\cbRGB\expandafter{\detokenize{81,255,81}}{this\strut} \setlength{\fboxsep}{0pt}\def\cbRGB{\colorbox[RGB]}\expandafter\cbRGB\expandafter{\detokenize{146,255,146}}{is\strut} \setlength{\fboxsep}{0pt}\def\cbRGB{\colorbox[RGB]}\expandafter\cbRGB\expandafter{\detokenize{206,255,206}}{a\strut} \setlength{\fboxsep}{0pt}\def\cbRGB{\colorbox[RGB]}\expandafter\cbRGB\expandafter{\detokenize{244,255,244}}{real\strut} \setlength{\fboxsep}{0pt}\def\cbRGB{\colorbox[RGB]}\expandafter\cbRGB\expandafter{\detokenize{255,255,255}}{``\strut} \setlength{\fboxsep}{0pt}\def\cbRGB{\colorbox[RGB]}\expandafter\cbRGB\expandafter{\detokenize{255,255,255}}{nothing\strut} \setlength{\fboxsep}{0pt}\def\cbRGB{\colorbox[RGB]}\expandafter\cbRGB\expandafter{\detokenize{255,255,255}}{''\strut} \setlength{\fboxsep}{0pt}\def\cbRGB{\colorbox[RGB]}\expandafter\cbRGB\expandafter{\detokenize{255,255,255}}{beer\strut} \setlength{\fboxsep}{0pt}\def\cbRGB{\colorbox[RGB]}\expandafter\cbRGB\expandafter{\detokenize{255,255,255}}{.\strut} \setlength{\fboxsep}{0pt}\def\cbRGB{\colorbox[RGB]}\expandafter\cbRGB\expandafter{\detokenize{255,255,255}}{pours\strut} \setlength{\fboxsep}{0pt}\def\cbRGB{\colorbox[RGB]}\expandafter\cbRGB\expandafter{\detokenize{255,255,255}}{a\strut} \setlength{\fboxsep}{0pt}\def\cbRGB{\colorbox[RGB]}\expandafter\cbRGB\expandafter{\detokenize{255,255,255}}{unk\strut} \setlength{\fboxsep}{0pt}\def\cbRGB{\colorbox[RGB]}\expandafter\cbRGB\expandafter{\detokenize{255,255,255}}{yellow\strut} \setlength{\fboxsep}{0pt}\def\cbRGB{\colorbox[RGB]}\expandafter\cbRGB\expandafter{\detokenize{255,255,255}}{color\strut} \setlength{\fboxsep}{0pt}\def\cbRGB{\colorbox[RGB]}\expandafter\cbRGB\expandafter{\detokenize{255,255,255}}{,\strut} \setlength{\fboxsep}{0pt}\def\cbRGB{\colorbox[RGB]}\expandafter\cbRGB\expandafter{\detokenize{255,255,255}}{looking\strut} \setlength{\fboxsep}{0pt}\def\cbRGB{\colorbox[RGB]}\expandafter\cbRGB\expandafter{\detokenize{255,255,255}}{more\strut} \setlength{\fboxsep}{0pt}\def\cbRGB{\colorbox[RGB]}\expandafter\cbRGB\expandafter{\detokenize{255,255,255}}{like\strut} \setlength{\fboxsep}{0pt}\def\cbRGB{\colorbox[RGB]}\expandafter\cbRGB\expandafter{\detokenize{255,255,255}}{unk\strut} \setlength{\fboxsep}{0pt}\def\cbRGB{\colorbox[RGB]}\expandafter\cbRGB\expandafter{\detokenize{255,255,255}}{water\strut} \setlength{\fboxsep}{0pt}\def\cbRGB{\colorbox[RGB]}\expandafter\cbRGB\expandafter{\detokenize{255,255,255}}{than\strut} \setlength{\fboxsep}{0pt}\def\cbRGB{\colorbox[RGB]}\expandafter\cbRGB\expandafter{\detokenize{223,255,223}}{beer\strut} \setlength{\fboxsep}{0pt}\def\cbRGB{\colorbox[RGB]}\expandafter\cbRGB\expandafter{\detokenize{161,255,161}}{.\strut} \setlength{\fboxsep}{0pt}\def\cbRGB{\colorbox[RGB]}\expandafter\cbRGB\expandafter{\detokenize{100,255,100}}{strange\strut} \setlength{\fboxsep}{0pt}\def\cbRGB{\colorbox[RGB]}\expandafter\cbRGB\expandafter{\detokenize{59,255,59}}{unk\strut} \setlength{\fboxsep}{0pt}\def\cbRGB{\colorbox[RGB]}\expandafter\cbRGB\expandafter{\detokenize{36,255,36}}{smell\strut} \setlength{\fboxsep}{0pt}\def\cbRGB{\colorbox[RGB]}\expandafter\cbRGB\expandafter{\detokenize{49,255,49}}{,\strut} \setlength{\fboxsep}{0pt}\def\cbRGB{\colorbox[RGB]}\expandafter\cbRGB\expandafter{\detokenize{58,255,58}}{with\strut} \setlength{\fboxsep}{0pt}\def\cbRGB{\colorbox[RGB]}\expandafter\cbRGB\expandafter{\detokenize{47,255,47}}{minimal\strut} \setlength{\fboxsep}{0pt}\def\cbRGB{\colorbox[RGB]}\expandafter\cbRGB\expandafter{\detokenize{36,255,36}}{unk\strut} \setlength{\fboxsep}{0pt}\def\cbRGB{\colorbox[RGB]}\expandafter\cbRGB\expandafter{\detokenize{27,255,27}}{hop\strut} \setlength{\fboxsep}{0pt}\def\cbRGB{\colorbox[RGB]}\expandafter\cbRGB\expandafter{\detokenize{10,255,10}}{aroma\strut} \setlength{\fboxsep}{0pt}\def\cbRGB{\colorbox[RGB]}\expandafter\cbRGB\expandafter{\detokenize{0,255,0}}{unk\strut} \setlength{\fboxsep}{0pt}\def\cbRGB{\colorbox[RGB]}\expandafter\cbRGB\expandafter{\detokenize{30,255,30}}{away\strut} \setlength{\fboxsep}{0pt}\def\cbRGB{\colorbox[RGB]}\expandafter\cbRGB\expandafter{\detokenize{81,255,81}}{behind\strut} \setlength{\fboxsep}{0pt}\def\cbRGB{\colorbox[RGB]}\expandafter\cbRGB\expandafter{\detokenize{152,255,152}}{whatever\strut} \setlength{\fboxsep}{0pt}\def\cbRGB{\colorbox[RGB]}\expandafter\cbRGB\expandafter{\detokenize{217,255,217}}{it\strut} \setlength{\fboxsep}{0pt}\def\cbRGB{\colorbox[RGB]}\expandafter\cbRGB\expandafter{\detokenize{255,255,255}}{is\strut} \setlength{\fboxsep}{0pt}\def\cbRGB{\colorbox[RGB]}\expandafter\cbRGB\expandafter{\detokenize{255,255,255}}{that\strut} \setlength{\fboxsep}{0pt}\def\cbRGB{\colorbox[RGB]}\expandafter\cbRGB\expandafter{\detokenize{255,255,255}}{unk\strut} \setlength{\fboxsep}{0pt}\def\cbRGB{\colorbox[RGB]}\expandafter\cbRGB\expandafter{\detokenize{255,255,255}}{in\strut} \setlength{\fboxsep}{0pt}\def\cbRGB{\colorbox[RGB]}\expandafter\cbRGB\expandafter{\detokenize{255,255,255}}{this\strut} \setlength{\fboxsep}{0pt}\def\cbRGB{\colorbox[RGB]}\expandafter\cbRGB\expandafter{\detokenize{255,255,255}}{brew\strut} \setlength{\fboxsep}{0pt}\def\cbRGB{\colorbox[RGB]}\expandafter\cbRGB\expandafter{\detokenize{255,255,255}}{.\strut} \setlength{\fboxsep}{0pt}\def\cbRGB{\colorbox[RGB]}\expandafter\cbRGB\expandafter{\detokenize{255,255,255}}{taste\strut} \setlength{\fboxsep}{0pt}\def\cbRGB{\colorbox[RGB]}\expandafter\cbRGB\expandafter{\detokenize{255,255,255}}{is\strut} \setlength{\fboxsep}{0pt}\def\cbRGB{\colorbox[RGB]}\expandafter\cbRGB\expandafter{\detokenize{255,255,255}}{equally\strut} \setlength{\fboxsep}{0pt}\def\cbRGB{\colorbox[RGB]}\expandafter\cbRGB\expandafter{\detokenize{255,255,255}}{as\strut} \setlength{\fboxsep}{0pt}\def\cbRGB{\colorbox[RGB]}\expandafter\cbRGB\expandafter{\detokenize{239,255,239}}{unk\strut} \setlength{\fboxsep}{0pt}\def\cbRGB{\colorbox[RGB]}\expandafter\cbRGB\expandafter{\detokenize{255,255,255}}{.\strut} \setlength{\fboxsep}{0pt}\def\cbRGB{\colorbox[RGB]}\expandafter\cbRGB\expandafter{\detokenize{255,255,255}}{nearly\strut} \setlength{\fboxsep}{0pt}\def\cbRGB{\colorbox[RGB]}\expandafter\cbRGB\expandafter{\detokenize{255,255,255}}{unk\strut} \setlength{\fboxsep}{0pt}\def\cbRGB{\colorbox[RGB]}\expandafter\cbRGB\expandafter{\detokenize{255,255,255}}{,\strut} \setlength{\fboxsep}{0pt}\def\cbRGB{\colorbox[RGB]}\expandafter\cbRGB\expandafter{\detokenize{255,255,255}}{the\strut} \setlength{\fboxsep}{0pt}\def\cbRGB{\colorbox[RGB]}\expandafter\cbRGB\expandafter{\detokenize{255,255,255}}{most\strut} \setlength{\fboxsep}{0pt}\def\cbRGB{\colorbox[RGB]}\expandafter\cbRGB\expandafter{\detokenize{255,255,255}}{you\strut} \setlength{\fboxsep}{0pt}\def\cbRGB{\colorbox[RGB]}\expandafter\cbRGB\expandafter{\detokenize{255,255,255}}{get\strut} \setlength{\fboxsep}{0pt}\def\cbRGB{\colorbox[RGB]}\expandafter\cbRGB\expandafter{\detokenize{255,255,255}}{from\strut} \setlength{\fboxsep}{0pt}\def\cbRGB{\colorbox[RGB]}\expandafter\cbRGB\expandafter{\detokenize{255,255,255}}{this\strut} \setlength{\fboxsep}{0pt}\def\cbRGB{\colorbox[RGB]}\expandafter\cbRGB\expandafter{\detokenize{255,255,255}}{beer\strut} \setlength{\fboxsep}{0pt}\def\cbRGB{\colorbox[RGB]}\expandafter\cbRGB\expandafter{\detokenize{255,255,255}}{is\strut} \setlength{\fboxsep}{0pt}\def\cbRGB{\colorbox[RGB]}\expandafter\cbRGB\expandafter{\detokenize{255,255,255}}{a\strut} \setlength{\fboxsep}{0pt}\def\cbRGB{\colorbox[RGB]}\expandafter\cbRGB\expandafter{\detokenize{255,255,255}}{slightly\strut} \setlength{\fboxsep}{0pt}\def\cbRGB{\colorbox[RGB]}\expandafter\cbRGB\expandafter{\detokenize{255,255,255}}{sweet\strut} \setlength{\fboxsep}{0pt}\def\cbRGB{\colorbox[RGB]}\expandafter\cbRGB\expandafter{\detokenize{255,255,255}}{unk\strut} \setlength{\fboxsep}{0pt}\def\cbRGB{\colorbox[RGB]}\expandafter\cbRGB\expandafter{\detokenize{248,255,248}}{flavor\strut} \setlength{\fboxsep}{0pt}\def\cbRGB{\colorbox[RGB]}\expandafter\cbRGB\expandafter{\detokenize{255,255,255}}{and\strut} \setlength{\fboxsep}{0pt}\def\cbRGB{\colorbox[RGB]}\expandafter\cbRGB\expandafter{\detokenize{255,255,255}}{a\strut} \setlength{\fboxsep}{0pt}\def\cbRGB{\colorbox[RGB]}\expandafter\cbRGB\expandafter{\detokenize{255,255,255}}{lot\strut} \setlength{\fboxsep}{0pt}\def\cbRGB{\colorbox[RGB]}\expandafter\cbRGB\expandafter{\detokenize{255,255,255}}{of\strut} \setlength{\fboxsep}{0pt}\def\cbRGB{\colorbox[RGB]}\expandafter\cbRGB\expandafter{\detokenize{255,255,255}}{carbonation\strut} \setlength{\fboxsep}{0pt}\def\cbRGB{\colorbox[RGB]}\expandafter\cbRGB\expandafter{\detokenize{255,255,255}}{in\strut} \setlength{\fboxsep}{0pt}\def\cbRGB{\colorbox[RGB]}\expandafter\cbRGB\expandafter{\detokenize{255,255,255}}{the\strut} \setlength{\fboxsep}{0pt}\def\cbRGB{\colorbox[RGB]}\expandafter\cbRGB\expandafter{\detokenize{255,255,255}}{beginning\strut} \setlength{\fboxsep}{0pt}\def\cbRGB{\colorbox[RGB]}\expandafter\cbRGB\expandafter{\detokenize{255,255,255}}{.\strut} \setlength{\fboxsep}{0pt}\def\cbRGB{\colorbox[RGB]}\expandafter\cbRGB\expandafter{\detokenize{242,255,242}}{watery\strut} \setlength{\fboxsep}{0pt}\def\cbRGB{\colorbox[RGB]}\expandafter\cbRGB\expandafter{\detokenize{255,255,255}}{and\strut} \setlength{\fboxsep}{0pt}\def\cbRGB{\colorbox[RGB]}\expandafter\cbRGB\expandafter{\detokenize{255,255,255}}{thin\strut} \setlength{\fboxsep}{0pt}\def\cbRGB{\colorbox[RGB]}\expandafter\cbRGB\expandafter{\detokenize{255,255,255}}{,\strut} \setlength{\fboxsep}{0pt}\def\cbRGB{\colorbox[RGB]}\expandafter\cbRGB\expandafter{\detokenize{255,255,255}}{but\strut} \setlength{\fboxsep}{0pt}\def\cbRGB{\colorbox[RGB]}\expandafter\cbRGB\expandafter{\detokenize{255,255,255}}{something\strut} \setlength{\fboxsep}{0pt}\def\cbRGB{\colorbox[RGB]}\expandafter\cbRGB\expandafter{\detokenize{255,255,255}}{that\strut} \setlength{\fboxsep}{0pt}\def\cbRGB{\colorbox[RGB]}\expandafter\cbRGB\expandafter{\detokenize{227,255,227}}{goes\strut} \setlength{\fboxsep}{0pt}\def\cbRGB{\colorbox[RGB]}\expandafter\cbRGB\expandafter{\detokenize{178,255,178}}{down\strut} \setlength{\fboxsep}{0pt}\def\cbRGB{\colorbox[RGB]}\expandafter\cbRGB\expandafter{\detokenize{170,255,170}}{easy\strut} \setlength{\fboxsep}{0pt}\def\cbRGB{\colorbox[RGB]}\expandafter\cbRGB\expandafter{\detokenize{163,255,163}}{so\strut} \setlength{\fboxsep}{0pt}\def\cbRGB{\colorbox[RGB]}\expandafter\cbRGB\expandafter{\detokenize{170,255,170}}{the\strut} \setlength{\fboxsep}{0pt}\def\cbRGB{\colorbox[RGB]}\expandafter\cbRGB\expandafter{\detokenize{219,255,219}}{drinkability\strut} \setlength{\fboxsep}{0pt}\def\cbRGB{\colorbox[RGB]}\expandafter\cbRGB\expandafter{\detokenize{255,255,255}}{is\strut} \setlength{\fboxsep}{0pt}\def\cbRGB{\colorbox[RGB]}\expandafter\cbRGB\expandafter{\detokenize{255,255,255}}{unk\strut} \setlength{\fboxsep}{0pt}\def\cbRGB{\colorbox[RGB]}\expandafter\cbRGB\expandafter{\detokenize{255,255,255}}{up\strut} \setlength{\fboxsep}{0pt}\def\cbRGB{\colorbox[RGB]}\expandafter\cbRGB\expandafter{\detokenize{255,255,255}}{slightly\strut} \setlength{\fboxsep}{0pt}\def\cbRGB{\colorbox[RGB]}\expandafter\cbRGB\expandafter{\detokenize{255,255,255}}{.\strut} \setlength{\fboxsep}{0pt}\def\cbRGB{\colorbox[RGB]}\expandafter\cbRGB\expandafter{\detokenize{255,255,255}}{unk\strut} \setlength{\fboxsep}{0pt}\def\cbRGB{\colorbox[RGB]}\expandafter\cbRGB\expandafter{\detokenize{255,255,255}}{.\strut} 

\setlength{\fboxsep}{0pt}\def\cbRGB{\colorbox[RGB]}\expandafter\cbRGB\expandafter{\detokenize{20,20,255}}{this\strut} \setlength{\fboxsep}{0pt}\def\cbRGB{\colorbox[RGB]}\expandafter\cbRGB\expandafter{\detokenize{69,69,255}}{is\strut} \setlength{\fboxsep}{0pt}\def\cbRGB{\colorbox[RGB]}\expandafter\cbRGB\expandafter{\detokenize{136,136,255}}{a\strut} \setlength{\fboxsep}{0pt}\def\cbRGB{\colorbox[RGB]}\expandafter\cbRGB\expandafter{\detokenize{185,185,255}}{real\strut} \setlength{\fboxsep}{0pt}\def\cbRGB{\colorbox[RGB]}\expandafter\cbRGB\expandafter{\detokenize{217,217,255}}{``\strut} \setlength{\fboxsep}{0pt}\def\cbRGB{\colorbox[RGB]}\expandafter\cbRGB\expandafter{\detokenize{230,230,255}}{nothing\strut} \setlength{\fboxsep}{0pt}\def\cbRGB{\colorbox[RGB]}\expandafter\cbRGB\expandafter{\detokenize{255,255,255}}{''\strut} \setlength{\fboxsep}{0pt}\def\cbRGB{\colorbox[RGB]}\expandafter\cbRGB\expandafter{\detokenize{255,255,255}}{beer\strut} \setlength{\fboxsep}{0pt}\def\cbRGB{\colorbox[RGB]}\expandafter\cbRGB\expandafter{\detokenize{244,244,255}}{.\strut} \setlength{\fboxsep}{0pt}\def\cbRGB{\colorbox[RGB]}\expandafter\cbRGB\expandafter{\detokenize{225,225,255}}{pours\strut} \setlength{\fboxsep}{0pt}\def\cbRGB{\colorbox[RGB]}\expandafter\cbRGB\expandafter{\detokenize{209,209,255}}{a\strut} \setlength{\fboxsep}{0pt}\def\cbRGB{\colorbox[RGB]}\expandafter\cbRGB\expandafter{\detokenize{175,175,255}}{unk\strut} \setlength{\fboxsep}{0pt}\def\cbRGB{\colorbox[RGB]}\expandafter\cbRGB\expandafter{\detokenize{175,175,255}}{yellow\strut} \setlength{\fboxsep}{0pt}\def\cbRGB{\colorbox[RGB]}\expandafter\cbRGB\expandafter{\detokenize{211,211,255}}{color\strut} \setlength{\fboxsep}{0pt}\def\cbRGB{\colorbox[RGB]}\expandafter\cbRGB\expandafter{\detokenize{247,247,255}}{,\strut} \setlength{\fboxsep}{0pt}\def\cbRGB{\colorbox[RGB]}\expandafter\cbRGB\expandafter{\detokenize{255,255,255}}{looking\strut} \setlength{\fboxsep}{0pt}\def\cbRGB{\colorbox[RGB]}\expandafter\cbRGB\expandafter{\detokenize{255,255,255}}{more\strut} \setlength{\fboxsep}{0pt}\def\cbRGB{\colorbox[RGB]}\expandafter\cbRGB\expandafter{\detokenize{255,255,255}}{like\strut} \setlength{\fboxsep}{0pt}\def\cbRGB{\colorbox[RGB]}\expandafter\cbRGB\expandafter{\detokenize{255,255,255}}{unk\strut} \setlength{\fboxsep}{0pt}\def\cbRGB{\colorbox[RGB]}\expandafter\cbRGB\expandafter{\detokenize{249,249,255}}{water\strut} \setlength{\fboxsep}{0pt}\def\cbRGB{\colorbox[RGB]}\expandafter\cbRGB\expandafter{\detokenize{255,255,255}}{than\strut} \setlength{\fboxsep}{0pt}\def\cbRGB{\colorbox[RGB]}\expandafter\cbRGB\expandafter{\detokenize{255,255,255}}{beer\strut} \setlength{\fboxsep}{0pt}\def\cbRGB{\colorbox[RGB]}\expandafter\cbRGB\expandafter{\detokenize{255,255,255}}{.\strut} \setlength{\fboxsep}{0pt}\def\cbRGB{\colorbox[RGB]}\expandafter\cbRGB\expandafter{\detokenize{255,255,255}}{strange\strut} \setlength{\fboxsep}{0pt}\def\cbRGB{\colorbox[RGB]}\expandafter\cbRGB\expandafter{\detokenize{255,255,255}}{unk\strut} \setlength{\fboxsep}{0pt}\def\cbRGB{\colorbox[RGB]}\expandafter\cbRGB\expandafter{\detokenize{255,255,255}}{smell\strut} \setlength{\fboxsep}{0pt}\def\cbRGB{\colorbox[RGB]}\expandafter\cbRGB\expandafter{\detokenize{255,255,255}}{,\strut} \setlength{\fboxsep}{0pt}\def\cbRGB{\colorbox[RGB]}\expandafter\cbRGB\expandafter{\detokenize{255,255,255}}{with\strut} \setlength{\fboxsep}{0pt}\def\cbRGB{\colorbox[RGB]}\expandafter\cbRGB\expandafter{\detokenize{255,255,255}}{minimal\strut} \setlength{\fboxsep}{0pt}\def\cbRGB{\colorbox[RGB]}\expandafter\cbRGB\expandafter{\detokenize{255,255,255}}{unk\strut} \setlength{\fboxsep}{0pt}\def\cbRGB{\colorbox[RGB]}\expandafter\cbRGB\expandafter{\detokenize{255,255,255}}{hop\strut} \setlength{\fboxsep}{0pt}\def\cbRGB{\colorbox[RGB]}\expandafter\cbRGB\expandafter{\detokenize{255,255,255}}{aroma\strut} \setlength{\fboxsep}{0pt}\def\cbRGB{\colorbox[RGB]}\expandafter\cbRGB\expandafter{\detokenize{255,255,255}}{unk\strut} \setlength{\fboxsep}{0pt}\def\cbRGB{\colorbox[RGB]}\expandafter\cbRGB\expandafter{\detokenize{255,255,255}}{away\strut} \setlength{\fboxsep}{0pt}\def\cbRGB{\colorbox[RGB]}\expandafter\cbRGB\expandafter{\detokenize{255,255,255}}{behind\strut} \setlength{\fboxsep}{0pt}\def\cbRGB{\colorbox[RGB]}\expandafter\cbRGB\expandafter{\detokenize{255,255,255}}{whatever\strut} \setlength{\fboxsep}{0pt}\def\cbRGB{\colorbox[RGB]}\expandafter\cbRGB\expandafter{\detokenize{255,255,255}}{it\strut} \setlength{\fboxsep}{0pt}\def\cbRGB{\colorbox[RGB]}\expandafter\cbRGB\expandafter{\detokenize{255,255,255}}{is\strut} \setlength{\fboxsep}{0pt}\def\cbRGB{\colorbox[RGB]}\expandafter\cbRGB\expandafter{\detokenize{255,255,255}}{that\strut} \setlength{\fboxsep}{0pt}\def\cbRGB{\colorbox[RGB]}\expandafter\cbRGB\expandafter{\detokenize{255,255,255}}{unk\strut} \setlength{\fboxsep}{0pt}\def\cbRGB{\colorbox[RGB]}\expandafter\cbRGB\expandafter{\detokenize{255,255,255}}{in\strut} \setlength{\fboxsep}{0pt}\def\cbRGB{\colorbox[RGB]}\expandafter\cbRGB\expandafter{\detokenize{255,255,255}}{this\strut} \setlength{\fboxsep}{0pt}\def\cbRGB{\colorbox[RGB]}\expandafter\cbRGB\expandafter{\detokenize{255,255,255}}{brew\strut} \setlength{\fboxsep}{0pt}\def\cbRGB{\colorbox[RGB]}\expandafter\cbRGB\expandafter{\detokenize{255,255,255}}{.\strut} \setlength{\fboxsep}{0pt}\def\cbRGB{\colorbox[RGB]}\expandafter\cbRGB\expandafter{\detokenize{255,255,255}}{taste\strut} \setlength{\fboxsep}{0pt}\def\cbRGB{\colorbox[RGB]}\expandafter\cbRGB\expandafter{\detokenize{255,255,255}}{is\strut} \setlength{\fboxsep}{0pt}\def\cbRGB{\colorbox[RGB]}\expandafter\cbRGB\expandafter{\detokenize{255,255,255}}{equally\strut} \setlength{\fboxsep}{0pt}\def\cbRGB{\colorbox[RGB]}\expandafter\cbRGB\expandafter{\detokenize{255,255,255}}{as\strut} \setlength{\fboxsep}{0pt}\def\cbRGB{\colorbox[RGB]}\expandafter\cbRGB\expandafter{\detokenize{255,255,255}}{unk\strut} \setlength{\fboxsep}{0pt}\def\cbRGB{\colorbox[RGB]}\expandafter\cbRGB\expandafter{\detokenize{255,255,255}}{.\strut} \setlength{\fboxsep}{0pt}\def\cbRGB{\colorbox[RGB]}\expandafter\cbRGB\expandafter{\detokenize{255,255,255}}{nearly\strut} \setlength{\fboxsep}{0pt}\def\cbRGB{\colorbox[RGB]}\expandafter\cbRGB\expandafter{\detokenize{255,255,255}}{unk\strut} \setlength{\fboxsep}{0pt}\def\cbRGB{\colorbox[RGB]}\expandafter\cbRGB\expandafter{\detokenize{255,255,255}}{,\strut} \setlength{\fboxsep}{0pt}\def\cbRGB{\colorbox[RGB]}\expandafter\cbRGB\expandafter{\detokenize{255,255,255}}{the\strut} \setlength{\fboxsep}{0pt}\def\cbRGB{\colorbox[RGB]}\expandafter\cbRGB\expandafter{\detokenize{255,255,255}}{most\strut} \setlength{\fboxsep}{0pt}\def\cbRGB{\colorbox[RGB]}\expandafter\cbRGB\expandafter{\detokenize{255,255,255}}{you\strut} \setlength{\fboxsep}{0pt}\def\cbRGB{\colorbox[RGB]}\expandafter\cbRGB\expandafter{\detokenize{255,255,255}}{get\strut} \setlength{\fboxsep}{0pt}\def\cbRGB{\colorbox[RGB]}\expandafter\cbRGB\expandafter{\detokenize{255,255,255}}{from\strut} \setlength{\fboxsep}{0pt}\def\cbRGB{\colorbox[RGB]}\expandafter\cbRGB\expandafter{\detokenize{255,255,255}}{this\strut} \setlength{\fboxsep}{0pt}\def\cbRGB{\colorbox[RGB]}\expandafter\cbRGB\expandafter{\detokenize{255,255,255}}{beer\strut} \setlength{\fboxsep}{0pt}\def\cbRGB{\colorbox[RGB]}\expandafter\cbRGB\expandafter{\detokenize{255,255,255}}{is\strut} \setlength{\fboxsep}{0pt}\def\cbRGB{\colorbox[RGB]}\expandafter\cbRGB\expandafter{\detokenize{255,255,255}}{a\strut} \setlength{\fboxsep}{0pt}\def\cbRGB{\colorbox[RGB]}\expandafter\cbRGB\expandafter{\detokenize{255,255,255}}{slightly\strut} \setlength{\fboxsep}{0pt}\def\cbRGB{\colorbox[RGB]}\expandafter\cbRGB\expandafter{\detokenize{255,255,255}}{sweet\strut} \setlength{\fboxsep}{0pt}\def\cbRGB{\colorbox[RGB]}\expandafter\cbRGB\expandafter{\detokenize{255,255,255}}{unk\strut} \setlength{\fboxsep}{0pt}\def\cbRGB{\colorbox[RGB]}\expandafter\cbRGB\expandafter{\detokenize{255,255,255}}{flavor\strut} \setlength{\fboxsep}{0pt}\def\cbRGB{\colorbox[RGB]}\expandafter\cbRGB\expandafter{\detokenize{255,255,255}}{and\strut} \setlength{\fboxsep}{0pt}\def\cbRGB{\colorbox[RGB]}\expandafter\cbRGB\expandafter{\detokenize{254,254,255}}{a\strut} \setlength{\fboxsep}{0pt}\def\cbRGB{\colorbox[RGB]}\expandafter\cbRGB\expandafter{\detokenize{214,214,255}}{lot\strut} \setlength{\fboxsep}{0pt}\def\cbRGB{\colorbox[RGB]}\expandafter\cbRGB\expandafter{\detokenize{173,173,255}}{of\strut} \setlength{\fboxsep}{0pt}\def\cbRGB{\colorbox[RGB]}\expandafter\cbRGB\expandafter{\detokenize{122,122,255}}{carbonation\strut} \setlength{\fboxsep}{0pt}\def\cbRGB{\colorbox[RGB]}\expandafter\cbRGB\expandafter{\detokenize{80,80,255}}{in\strut} \setlength{\fboxsep}{0pt}\def\cbRGB{\colorbox[RGB]}\expandafter\cbRGB\expandafter{\detokenize{56,56,255}}{the\strut} \setlength{\fboxsep}{0pt}\def\cbRGB{\colorbox[RGB]}\expandafter\cbRGB\expandafter{\detokenize{33,33,255}}{beginning\strut} \setlength{\fboxsep}{0pt}\def\cbRGB{\colorbox[RGB]}\expandafter\cbRGB\expandafter{\detokenize{9,9,255}}{.\strut} \setlength{\fboxsep}{0pt}\def\cbRGB{\colorbox[RGB]}\expandafter\cbRGB\expandafter{\detokenize{0,0,255}}{watery\strut} \setlength{\fboxsep}{0pt}\def\cbRGB{\colorbox[RGB]}\expandafter\cbRGB\expandafter{\detokenize{3,3,255}}{and\strut} \setlength{\fboxsep}{0pt}\def\cbRGB{\colorbox[RGB]}\expandafter\cbRGB\expandafter{\detokenize{24,24,255}}{thin\strut} \setlength{\fboxsep}{0pt}\def\cbRGB{\colorbox[RGB]}\expandafter\cbRGB\expandafter{\detokenize{71,71,255}}{,\strut} \setlength{\fboxsep}{0pt}\def\cbRGB{\colorbox[RGB]}\expandafter\cbRGB\expandafter{\detokenize{120,120,255}}{but\strut} \setlength{\fboxsep}{0pt}\def\cbRGB{\colorbox[RGB]}\expandafter\cbRGB\expandafter{\detokenize{186,186,255}}{something\strut} \setlength{\fboxsep}{0pt}\def\cbRGB{\colorbox[RGB]}\expandafter\cbRGB\expandafter{\detokenize{244,244,255}}{that\strut} \setlength{\fboxsep}{0pt}\def\cbRGB{\colorbox[RGB]}\expandafter\cbRGB\expandafter{\detokenize{255,255,255}}{goes\strut} \setlength{\fboxsep}{0pt}\def\cbRGB{\colorbox[RGB]}\expandafter\cbRGB\expandafter{\detokenize{255,255,255}}{down\strut} \setlength{\fboxsep}{0pt}\def\cbRGB{\colorbox[RGB]}\expandafter\cbRGB\expandafter{\detokenize{255,255,255}}{easy\strut} \setlength{\fboxsep}{0pt}\def\cbRGB{\colorbox[RGB]}\expandafter\cbRGB\expandafter{\detokenize{255,255,255}}{so\strut} \setlength{\fboxsep}{0pt}\def\cbRGB{\colorbox[RGB]}\expandafter\cbRGB\expandafter{\detokenize{255,255,255}}{the\strut} \setlength{\fboxsep}{0pt}\def\cbRGB{\colorbox[RGB]}\expandafter\cbRGB\expandafter{\detokenize{245,245,255}}{drinkability\strut} \setlength{\fboxsep}{0pt}\def\cbRGB{\colorbox[RGB]}\expandafter\cbRGB\expandafter{\detokenize{219,219,255}}{is\strut} \setlength{\fboxsep}{0pt}\def\cbRGB{\colorbox[RGB]}\expandafter\cbRGB\expandafter{\detokenize{214,214,255}}{unk\strut} \setlength{\fboxsep}{0pt}\def\cbRGB{\colorbox[RGB]}\expandafter\cbRGB\expandafter{\detokenize{239,239,255}}{up\strut} \setlength{\fboxsep}{0pt}\def\cbRGB{\colorbox[RGB]}\expandafter\cbRGB\expandafter{\detokenize{255,255,255}}{slightly\strut} \setlength{\fboxsep}{0pt}\def\cbRGB{\colorbox[RGB]}\expandafter\cbRGB\expandafter{\detokenize{255,255,255}}{.\strut} \setlength{\fboxsep}{0pt}\def\cbRGB{\colorbox[RGB]}\expandafter\cbRGB\expandafter{\detokenize{255,255,255}}{unk\strut} \setlength{\fboxsep}{0pt}\def\cbRGB{\colorbox[RGB]}\expandafter\cbRGB\expandafter{\detokenize{255,255,255}}{.\strut} 

\setlength{\fboxsep}{0pt}\def\cbRGB{\colorbox[RGB]}\expandafter\cbRGB\expandafter{\detokenize{255,255,121}}{this\strut} \setlength{\fboxsep}{0pt}\def\cbRGB{\colorbox[RGB]}\expandafter\cbRGB\expandafter{\detokenize{255,255,186}}{is\strut} \setlength{\fboxsep}{0pt}\def\cbRGB{\colorbox[RGB]}\expandafter\cbRGB\expandafter{\detokenize{255,255,246}}{a\strut} \setlength{\fboxsep}{0pt}\def\cbRGB{\colorbox[RGB]}\expandafter\cbRGB\expandafter{\detokenize{255,255,255}}{real\strut} \setlength{\fboxsep}{0pt}\def\cbRGB{\colorbox[RGB]}\expandafter\cbRGB\expandafter{\detokenize{255,255,255}}{``\strut} \setlength{\fboxsep}{0pt}\def\cbRGB{\colorbox[RGB]}\expandafter\cbRGB\expandafter{\detokenize{255,255,255}}{nothing\strut} \setlength{\fboxsep}{0pt}\def\cbRGB{\colorbox[RGB]}\expandafter\cbRGB\expandafter{\detokenize{255,255,255}}{''\strut} \setlength{\fboxsep}{0pt}\def\cbRGB{\colorbox[RGB]}\expandafter\cbRGB\expandafter{\detokenize{255,255,255}}{beer\strut} \setlength{\fboxsep}{0pt}\def\cbRGB{\colorbox[RGB]}\expandafter\cbRGB\expandafter{\detokenize{255,255,255}}{.\strut} \setlength{\fboxsep}{0pt}\def\cbRGB{\colorbox[RGB]}\expandafter\cbRGB\expandafter{\detokenize{255,255,255}}{pours\strut} \setlength{\fboxsep}{0pt}\def\cbRGB{\colorbox[RGB]}\expandafter\cbRGB\expandafter{\detokenize{255,255,255}}{a\strut} \setlength{\fboxsep}{0pt}\def\cbRGB{\colorbox[RGB]}\expandafter\cbRGB\expandafter{\detokenize{255,255,255}}{unk\strut} \setlength{\fboxsep}{0pt}\def\cbRGB{\colorbox[RGB]}\expandafter\cbRGB\expandafter{\detokenize{255,255,255}}{yellow\strut} \setlength{\fboxsep}{0pt}\def\cbRGB{\colorbox[RGB]}\expandafter\cbRGB\expandafter{\detokenize{255,255,255}}{color\strut} \setlength{\fboxsep}{0pt}\def\cbRGB{\colorbox[RGB]}\expandafter\cbRGB\expandafter{\detokenize{255,255,255}}{,\strut} \setlength{\fboxsep}{0pt}\def\cbRGB{\colorbox[RGB]}\expandafter\cbRGB\expandafter{\detokenize{255,255,255}}{looking\strut} \setlength{\fboxsep}{0pt}\def\cbRGB{\colorbox[RGB]}\expandafter\cbRGB\expandafter{\detokenize{255,255,255}}{more\strut} \setlength{\fboxsep}{0pt}\def\cbRGB{\colorbox[RGB]}\expandafter\cbRGB\expandafter{\detokenize{255,255,255}}{like\strut} \setlength{\fboxsep}{0pt}\def\cbRGB{\colorbox[RGB]}\expandafter\cbRGB\expandafter{\detokenize{255,255,255}}{unk\strut} \setlength{\fboxsep}{0pt}\def\cbRGB{\colorbox[RGB]}\expandafter\cbRGB\expandafter{\detokenize{255,255,255}}{water\strut} \setlength{\fboxsep}{0pt}\def\cbRGB{\colorbox[RGB]}\expandafter\cbRGB\expandafter{\detokenize{255,255,255}}{than\strut} \setlength{\fboxsep}{0pt}\def\cbRGB{\colorbox[RGB]}\expandafter\cbRGB\expandafter{\detokenize{255,255,255}}{beer\strut} \setlength{\fboxsep}{0pt}\def\cbRGB{\colorbox[RGB]}\expandafter\cbRGB\expandafter{\detokenize{255,255,255}}{.\strut} \setlength{\fboxsep}{0pt}\def\cbRGB{\colorbox[RGB]}\expandafter\cbRGB\expandafter{\detokenize{255,255,255}}{strange\strut} \setlength{\fboxsep}{0pt}\def\cbRGB{\colorbox[RGB]}\expandafter\cbRGB\expandafter{\detokenize{255,255,255}}{unk\strut} \setlength{\fboxsep}{0pt}\def\cbRGB{\colorbox[RGB]}\expandafter\cbRGB\expandafter{\detokenize{255,255,255}}{smell\strut} \setlength{\fboxsep}{0pt}\def\cbRGB{\colorbox[RGB]}\expandafter\cbRGB\expandafter{\detokenize{255,255,255}}{,\strut} \setlength{\fboxsep}{0pt}\def\cbRGB{\colorbox[RGB]}\expandafter\cbRGB\expandafter{\detokenize{255,255,255}}{with\strut} \setlength{\fboxsep}{0pt}\def\cbRGB{\colorbox[RGB]}\expandafter\cbRGB\expandafter{\detokenize{255,255,255}}{minimal\strut} \setlength{\fboxsep}{0pt}\def\cbRGB{\colorbox[RGB]}\expandafter\cbRGB\expandafter{\detokenize{255,255,255}}{unk\strut} \setlength{\fboxsep}{0pt}\def\cbRGB{\colorbox[RGB]}\expandafter\cbRGB\expandafter{\detokenize{255,255,255}}{hop\strut} \setlength{\fboxsep}{0pt}\def\cbRGB{\colorbox[RGB]}\expandafter\cbRGB\expandafter{\detokenize{255,255,255}}{aroma\strut} \setlength{\fboxsep}{0pt}\def\cbRGB{\colorbox[RGB]}\expandafter\cbRGB\expandafter{\detokenize{255,255,255}}{unk\strut} \setlength{\fboxsep}{0pt}\def\cbRGB{\colorbox[RGB]}\expandafter\cbRGB\expandafter{\detokenize{255,255,255}}{away\strut} \setlength{\fboxsep}{0pt}\def\cbRGB{\colorbox[RGB]}\expandafter\cbRGB\expandafter{\detokenize{255,255,255}}{behind\strut} \setlength{\fboxsep}{0pt}\def\cbRGB{\colorbox[RGB]}\expandafter\cbRGB\expandafter{\detokenize{255,255,232}}{whatever\strut} \setlength{\fboxsep}{0pt}\def\cbRGB{\colorbox[RGB]}\expandafter\cbRGB\expandafter{\detokenize{255,255,218}}{it\strut} \setlength{\fboxsep}{0pt}\def\cbRGB{\colorbox[RGB]}\expandafter\cbRGB\expandafter{\detokenize{255,255,227}}{is\strut} \setlength{\fboxsep}{0pt}\def\cbRGB{\colorbox[RGB]}\expandafter\cbRGB\expandafter{\detokenize{255,255,233}}{that\strut} \setlength{\fboxsep}{0pt}\def\cbRGB{\colorbox[RGB]}\expandafter\cbRGB\expandafter{\detokenize{255,255,240}}{unk\strut} \setlength{\fboxsep}{0pt}\def\cbRGB{\colorbox[RGB]}\expandafter\cbRGB\expandafter{\detokenize{255,255,250}}{in\strut} \setlength{\fboxsep}{0pt}\def\cbRGB{\colorbox[RGB]}\expandafter\cbRGB\expandafter{\detokenize{255,255,253}}{this\strut} \setlength{\fboxsep}{0pt}\def\cbRGB{\colorbox[RGB]}\expandafter\cbRGB\expandafter{\detokenize{255,255,239}}{brew\strut} \setlength{\fboxsep}{0pt}\def\cbRGB{\colorbox[RGB]}\expandafter\cbRGB\expandafter{\detokenize{255,255,218}}{.\strut} \setlength{\fboxsep}{0pt}\def\cbRGB{\colorbox[RGB]}\expandafter\cbRGB\expandafter{\detokenize{255,255,188}}{taste\strut} \setlength{\fboxsep}{0pt}\def\cbRGB{\colorbox[RGB]}\expandafter\cbRGB\expandafter{\detokenize{255,255,154}}{is\strut} \setlength{\fboxsep}{0pt}\def\cbRGB{\colorbox[RGB]}\expandafter\cbRGB\expandafter{\detokenize{255,255,134}}{equally\strut} \setlength{\fboxsep}{0pt}\def\cbRGB{\colorbox[RGB]}\expandafter\cbRGB\expandafter{\detokenize{255,255,139}}{as\strut} \setlength{\fboxsep}{0pt}\def\cbRGB{\colorbox[RGB]}\expandafter\cbRGB\expandafter{\detokenize{255,255,169}}{unk\strut} \setlength{\fboxsep}{0pt}\def\cbRGB{\colorbox[RGB]}\expandafter\cbRGB\expandafter{\detokenize{255,255,216}}{.\strut} \setlength{\fboxsep}{0pt}\def\cbRGB{\colorbox[RGB]}\expandafter\cbRGB\expandafter{\detokenize{255,255,255}}{nearly\strut} \setlength{\fboxsep}{0pt}\def\cbRGB{\colorbox[RGB]}\expandafter\cbRGB\expandafter{\detokenize{255,255,255}}{unk\strut} \setlength{\fboxsep}{0pt}\def\cbRGB{\colorbox[RGB]}\expandafter\cbRGB\expandafter{\detokenize{255,255,255}}{,\strut} \setlength{\fboxsep}{0pt}\def\cbRGB{\colorbox[RGB]}\expandafter\cbRGB\expandafter{\detokenize{255,255,250}}{the\strut} \setlength{\fboxsep}{0pt}\def\cbRGB{\colorbox[RGB]}\expandafter\cbRGB\expandafter{\detokenize{255,255,219}}{most\strut} \setlength{\fboxsep}{0pt}\def\cbRGB{\colorbox[RGB]}\expandafter\cbRGB\expandafter{\detokenize{255,255,197}}{you\strut} \setlength{\fboxsep}{0pt}\def\cbRGB{\colorbox[RGB]}\expandafter\cbRGB\expandafter{\detokenize{255,255,202}}{get\strut} \setlength{\fboxsep}{0pt}\def\cbRGB{\colorbox[RGB]}\expandafter\cbRGB\expandafter{\detokenize{255,255,237}}{from\strut} \setlength{\fboxsep}{0pt}\def\cbRGB{\colorbox[RGB]}\expandafter\cbRGB\expandafter{\detokenize{255,255,255}}{this\strut} \setlength{\fboxsep}{0pt}\def\cbRGB{\colorbox[RGB]}\expandafter\cbRGB\expandafter{\detokenize{255,255,255}}{beer\strut} \setlength{\fboxsep}{0pt}\def\cbRGB{\colorbox[RGB]}\expandafter\cbRGB\expandafter{\detokenize{255,255,255}}{is\strut} \setlength{\fboxsep}{0pt}\def\cbRGB{\colorbox[RGB]}\expandafter\cbRGB\expandafter{\detokenize{255,255,255}}{a\strut} \setlength{\fboxsep}{0pt}\def\cbRGB{\colorbox[RGB]}\expandafter\cbRGB\expandafter{\detokenize{255,255,255}}{slightly\strut} \setlength{\fboxsep}{0pt}\def\cbRGB{\colorbox[RGB]}\expandafter\cbRGB\expandafter{\detokenize{255,255,255}}{sweet\strut} \setlength{\fboxsep}{0pt}\def\cbRGB{\colorbox[RGB]}\expandafter\cbRGB\expandafter{\detokenize{255,255,255}}{unk\strut} \setlength{\fboxsep}{0pt}\def\cbRGB{\colorbox[RGB]}\expandafter\cbRGB\expandafter{\detokenize{255,255,255}}{flavor\strut} \setlength{\fboxsep}{0pt}\def\cbRGB{\colorbox[RGB]}\expandafter\cbRGB\expandafter{\detokenize{255,255,255}}{and\strut} \setlength{\fboxsep}{0pt}\def\cbRGB{\colorbox[RGB]}\expandafter\cbRGB\expandafter{\detokenize{255,255,255}}{a\strut} \setlength{\fboxsep}{0pt}\def\cbRGB{\colorbox[RGB]}\expandafter\cbRGB\expandafter{\detokenize{255,255,255}}{lot\strut} \setlength{\fboxsep}{0pt}\def\cbRGB{\colorbox[RGB]}\expandafter\cbRGB\expandafter{\detokenize{255,255,255}}{of\strut} \setlength{\fboxsep}{0pt}\def\cbRGB{\colorbox[RGB]}\expandafter\cbRGB\expandafter{\detokenize{255,255,255}}{carbonation\strut} \setlength{\fboxsep}{0pt}\def\cbRGB{\colorbox[RGB]}\expandafter\cbRGB\expandafter{\detokenize{255,255,255}}{in\strut} \setlength{\fboxsep}{0pt}\def\cbRGB{\colorbox[RGB]}\expandafter\cbRGB\expandafter{\detokenize{255,255,255}}{the\strut} \setlength{\fboxsep}{0pt}\def\cbRGB{\colorbox[RGB]}\expandafter\cbRGB\expandafter{\detokenize{255,255,243}}{beginning\strut} \setlength{\fboxsep}{0pt}\def\cbRGB{\colorbox[RGB]}\expandafter\cbRGB\expandafter{\detokenize{255,255,182}}{.\strut} \setlength{\fboxsep}{0pt}\def\cbRGB{\colorbox[RGB]}\expandafter\cbRGB\expandafter{\detokenize{255,255,107}}{watery\strut} \setlength{\fboxsep}{0pt}\def\cbRGB{\colorbox[RGB]}\expandafter\cbRGB\expandafter{\detokenize{255,255,49}}{and\strut} \setlength{\fboxsep}{0pt}\def\cbRGB{\colorbox[RGB]}\expandafter\cbRGB\expandafter{\detokenize{255,255,8}}{thin\strut} \setlength{\fboxsep}{0pt}\def\cbRGB{\colorbox[RGB]}\expandafter\cbRGB\expandafter{\detokenize{255,255,0}}{,\strut} \setlength{\fboxsep}{0pt}\def\cbRGB{\colorbox[RGB]}\expandafter\cbRGB\expandafter{\detokenize{255,255,14}}{but\strut} \setlength{\fboxsep}{0pt}\def\cbRGB{\colorbox[RGB]}\expandafter\cbRGB\expandafter{\detokenize{255,255,59}}{something\strut} \setlength{\fboxsep}{0pt}\def\cbRGB{\colorbox[RGB]}\expandafter\cbRGB\expandafter{\detokenize{255,255,119}}{that\strut} \setlength{\fboxsep}{0pt}\def\cbRGB{\colorbox[RGB]}\expandafter\cbRGB\expandafter{\detokenize{255,255,190}}{goes\strut} \setlength{\fboxsep}{0pt}\def\cbRGB{\colorbox[RGB]}\expandafter\cbRGB\expandafter{\detokenize{255,255,241}}{down\strut} \setlength{\fboxsep}{0pt}\def\cbRGB{\colorbox[RGB]}\expandafter\cbRGB\expandafter{\detokenize{255,255,255}}{easy\strut} \setlength{\fboxsep}{0pt}\def\cbRGB{\colorbox[RGB]}\expandafter\cbRGB\expandafter{\detokenize{255,255,255}}{so\strut} \setlength{\fboxsep}{0pt}\def\cbRGB{\colorbox[RGB]}\expandafter\cbRGB\expandafter{\detokenize{255,255,255}}{the\strut} \setlength{\fboxsep}{0pt}\def\cbRGB{\colorbox[RGB]}\expandafter\cbRGB\expandafter{\detokenize{255,255,244}}{drinkability\strut} \setlength{\fboxsep}{0pt}\def\cbRGB{\colorbox[RGB]}\expandafter\cbRGB\expandafter{\detokenize{255,255,221}}{is\strut} \setlength{\fboxsep}{0pt}\def\cbRGB{\colorbox[RGB]}\expandafter\cbRGB\expandafter{\detokenize{255,255,207}}{unk\strut} \setlength{\fboxsep}{0pt}\def\cbRGB{\colorbox[RGB]}\expandafter\cbRGB\expandafter{\detokenize{255,255,210}}{up\strut} \setlength{\fboxsep}{0pt}\def\cbRGB{\colorbox[RGB]}\expandafter\cbRGB\expandafter{\detokenize{255,255,234}}{slightly\strut} \setlength{\fboxsep}{0pt}\def\cbRGB{\colorbox[RGB]}\expandafter\cbRGB\expandafter{\detokenize{255,255,255}}{.\strut} \setlength{\fboxsep}{0pt}\def\cbRGB{\colorbox[RGB]}\expandafter\cbRGB\expandafter{\detokenize{255,255,255}}{unk\strut} \setlength{\fboxsep}{0pt}\def\cbRGB{\colorbox[RGB]}\expandafter\cbRGB\expandafter{\detokenize{255,255,255}}{.\strut} 

\subsection{Yelp!/TripAdvisor Dataset}
Color Legend : 
\setlength{\fboxsep}{0pt}\def\cbRGB{\colorbox[RGB]}\expandafter\cbRGB\expandafter{\detokenize{255,0,0}}{Sentiment\strut} 
\setlength{\fboxsep}{0pt}\def\cbRGB{\colorbox[RGB]}\expandafter\cbRGB\expandafter{\detokenize{0,0,255}}{Domain\strut} 

\textbf{Example 1}
\setlength{\fboxsep}{0pt}\def\cbRGB{\colorbox[RGB]}\expandafter\cbRGB\expandafter{\detokenize{255,92,92}}{i\strut} \setlength{\fboxsep}{0pt}\def\cbRGB{\colorbox[RGB]}\expandafter\cbRGB\expandafter{\detokenize{255,21,21}}{love\strut} \setlength{\fboxsep}{0pt}\def\cbRGB{\colorbox[RGB]}\expandafter\cbRGB\expandafter{\detokenize{255,0,0}}{everything\strut} \setlength{\fboxsep}{0pt}\def\cbRGB{\colorbox[RGB]}\expandafter\cbRGB\expandafter{\detokenize{255,102,102}}{about\strut} \setlength{\fboxsep}{0pt}\def\cbRGB{\colorbox[RGB]}\expandafter\cbRGB\expandafter{\detokenize{255,181,181}}{this\strut} \setlength{\fboxsep}{0pt}\def\cbRGB{\colorbox[RGB]}\expandafter\cbRGB\expandafter{\detokenize{255,205,205}}{place\strut} \setlength{\fboxsep}{0pt}\def\cbRGB{\colorbox[RGB]}\expandafter\cbRGB\expandafter{\detokenize{255,221,221}}{,\strut} \setlength{\fboxsep}{0pt}\def\cbRGB{\colorbox[RGB]}\expandafter\cbRGB\expandafter{\detokenize{255,220,220}}{i\strut} \setlength{\fboxsep}{0pt}\def\cbRGB{\colorbox[RGB]}\expandafter\cbRGB\expandafter{\detokenize{255,212,212}}{find\strut} \setlength{\fboxsep}{0pt}\def\cbRGB{\colorbox[RGB]}\expandafter\cbRGB\expandafter{\detokenize{255,213,213}}{my\strut} \setlength{\fboxsep}{0pt}\def\cbRGB{\colorbox[RGB]}\expandafter\cbRGB\expandafter{\detokenize{255,222,222}}{self\strut} \setlength{\fboxsep}{0pt}\def\cbRGB{\colorbox[RGB]}\expandafter\cbRGB\expandafter{\detokenize{255,231,231}}{there\strut} \setlength{\fboxsep}{0pt}\def\cbRGB{\colorbox[RGB]}\expandafter\cbRGB\expandafter{\detokenize{255,151,151}}{more\strut} \setlength{\fboxsep}{0pt}\def\cbRGB{\colorbox[RGB]}\expandafter\cbRGB\expandafter{\detokenize{255,115,115}}{often\strut} \setlength{\fboxsep}{0pt}\def\cbRGB{\colorbox[RGB]}\expandafter\cbRGB\expandafter{\detokenize{255,116,116}}{then\strut} \setlength{\fboxsep}{0pt}\def\cbRGB{\colorbox[RGB]}\expandafter\cbRGB\expandafter{\detokenize{255,200,200}}{i\strut} \setlength{\fboxsep}{0pt}\def\cbRGB{\colorbox[RGB]}\expandafter\cbRGB\expandafter{\detokenize{255,237,237}}{want\strut} \setlength{\fboxsep}{0pt}\def\cbRGB{\colorbox[RGB]}\expandafter\cbRGB\expandafter{\detokenize{255,226,226}}{,\strut} \setlength{\fboxsep}{0pt}\def\cbRGB{\colorbox[RGB]}\expandafter\cbRGB\expandafter{\detokenize{255,231,231}}{the\strut} \setlength{\fboxsep}{0pt}\def\cbRGB{\colorbox[RGB]}\expandafter\cbRGB\expandafter{\detokenize{255,216,216}}{deserts\strut} \setlength{\fboxsep}{0pt}\def\cbRGB{\colorbox[RGB]}\expandafter\cbRGB\expandafter{\detokenize{255,28,28}}{are\strut} \setlength{\fboxsep}{0pt}\def\cbRGB{\colorbox[RGB]}\expandafter\cbRGB\expandafter{\detokenize{255,24,24}}{amazing\strut} \setlength{\fboxsep}{0pt}\def\cbRGB{\colorbox[RGB]}\expandafter\cbRGB\expandafter{\detokenize{255,30,30}}{,\strut} \setlength{\fboxsep}{0pt}\def\cbRGB{\colorbox[RGB]}\expandafter\cbRGB\expandafter{\detokenize{255,216,216}}{so\strut} \setlength{\fboxsep}{0pt}\def\cbRGB{\colorbox[RGB]}\expandafter\cbRGB\expandafter{\detokenize{255,209,209}}{many\strut} \setlength{\fboxsep}{0pt}\def\cbRGB{\colorbox[RGB]}\expandafter\cbRGB\expandafter{\detokenize{255,216,216}}{choices\strut} \setlength{\fboxsep}{0pt}\def\cbRGB{\colorbox[RGB]}\expandafter\cbRGB\expandafter{\detokenize{255,217,217}}{,\strut} \setlength{\fboxsep}{0pt}\def\cbRGB{\colorbox[RGB]}\expandafter\cbRGB\expandafter{\detokenize{255,218,218}}{the\strut} \setlength{\fboxsep}{0pt}\def\cbRGB{\colorbox[RGB]}\expandafter\cbRGB\expandafter{\detokenize{255,223,223}}{deli\strut} \setlength{\fboxsep}{0pt}\def\cbRGB{\colorbox[RGB]}\expandafter\cbRGB\expandafter{\detokenize{255,217,217}}{is\strut} \setlength{\fboxsep}{0pt}\def\cbRGB{\colorbox[RGB]}\expandafter\cbRGB\expandafter{\detokenize{255,35,35}}{ \strut} \setlength{\fboxsep}{0pt}\def\cbRGB{\colorbox[RGB]}\expandafter\cbRGB\expandafter{\detokenize{255,23,23}}{awesome\strut} \setlength{\fboxsep}{0pt}\def\cbRGB{\colorbox[RGB]}\expandafter\cbRGB\expandafter{\detokenize{255,22,22}}{,\strut} \setlength{\fboxsep}{0pt}\def\cbRGB{\colorbox[RGB]}\expandafter\cbRGB\expandafter{\detokenize{255,205,205}}{the\strut} \setlength{\fboxsep}{0pt}\def\cbRGB{\colorbox[RGB]}\expandafter\cbRGB\expandafter{\detokenize{255,210,210}}{sushi\strut} \setlength{\fboxsep}{0pt}\def\cbRGB{\colorbox[RGB]}\expandafter\cbRGB\expandafter{\detokenize{255,219,219}}{bar\strut} \setlength{\fboxsep}{0pt}\def\cbRGB{\colorbox[RGB]}\expandafter\cbRGB\expandafter{\detokenize{255,116,116}}{is\strut} \setlength{\fboxsep}{0pt}\def\cbRGB{\colorbox[RGB]}\expandafter\cbRGB\expandafter{\detokenize{255,122,122}}{great\strut} \setlength{\fboxsep}{0pt}\def\cbRGB{\colorbox[RGB]}\expandafter\cbRGB\expandafter{\detokenize{255,154,154}}{,\strut} \setlength{\fboxsep}{0pt}\def\cbRGB{\colorbox[RGB]}\expandafter\cbRGB\expandafter{\detokenize{255,255,255}}{you\strut} \setlength{\fboxsep}{0pt}\def\cbRGB{\colorbox[RGB]}\expandafter\cbRGB\expandafter{\detokenize{255,206,206}}{can\strut} \setlength{\fboxsep}{0pt}\def\cbRGB{\colorbox[RGB]}\expandafter\cbRGB\expandafter{\detokenize{255,197,197}}{not\strut} \setlength{\fboxsep}{0pt}\def\cbRGB{\colorbox[RGB]}\expandafter\cbRGB\expandafter{\detokenize{255,178,178}}{go\strut} \setlength{\fboxsep}{0pt}\def\cbRGB{\colorbox[RGB]}\expandafter\cbRGB\expandafter{\detokenize{255,221,221}}{wrong\strut} \setlength{\fboxsep}{0pt}\def\cbRGB{\colorbox[RGB]}\expandafter\cbRGB\expandafter{\detokenize{255,202,202}}{with\strut} \setlength{\fboxsep}{0pt}\def\cbRGB{\colorbox[RGB]}\expandafter\cbRGB\expandafter{\detokenize{255,223,223}}{this\strut} \setlength{\fboxsep}{0pt}\def\cbRGB{\colorbox[RGB]}\expandafter\cbRGB\expandafter{\detokenize{255,225,225}}{place\strut} \setlength{\fboxsep}{0pt}\def\cbRGB{\colorbox[RGB]}\expandafter\cbRGB\expandafter{\detokenize{255,219,219}}{,\strut} \setlength{\fboxsep}{0pt}\def\cbRGB{\colorbox[RGB]}\expandafter\cbRGB\expandafter{\detokenize{255,220,220}}{if\strut} \setlength{\fboxsep}{0pt}\def\cbRGB{\colorbox[RGB]}\expandafter\cbRGB\expandafter{\detokenize{255,211,211}}{you\strut} \setlength{\fboxsep}{0pt}\def\cbRGB{\colorbox[RGB]}\expandafter\cbRGB\expandafter{\detokenize{255,219,219}}{want\strut} \setlength{\fboxsep}{0pt}\def\cbRGB{\colorbox[RGB]}\expandafter\cbRGB\expandafter{\detokenize{255,220,220}}{a\strut} \setlength{\fboxsep}{0pt}\def\cbRGB{\colorbox[RGB]}\expandafter\cbRGB\expandafter{\detokenize{255,198,198}}{formal\strut} \setlength{\fboxsep}{0pt}\def\cbRGB{\colorbox[RGB]}\expandafter\cbRGB\expandafter{\detokenize{255,194,194}}{dining\strut} \setlength{\fboxsep}{0pt}\def\cbRGB{\colorbox[RGB]}\expandafter\cbRGB\expandafter{\detokenize{255,161,161}}{you\strut} \setlength{\fboxsep}{0pt}\def\cbRGB{\colorbox[RGB]}\expandafter\cbRGB\expandafter{\detokenize{255,185,185}}{can\strut} \setlength{\fboxsep}{0pt}\def\cbRGB{\colorbox[RGB]}\expandafter\cbRGB\expandafter{\detokenize{255,187,187}}{sit\strut} \setlength{\fboxsep}{0pt}\def\cbRGB{\colorbox[RGB]}\expandafter\cbRGB\expandafter{\detokenize{255,220,220}}{at\strut} \setlength{\fboxsep}{0pt}\def\cbRGB{\colorbox[RGB]}\expandafter\cbRGB\expandafter{\detokenize{255,213,213}}{the\strut} \setlength{\fboxsep}{0pt}\def\cbRGB{\colorbox[RGB]}\expandafter\cbRGB\expandafter{\detokenize{255,211,211}}{full\strut} \setlength{\fboxsep}{0pt}\def\cbRGB{\colorbox[RGB]}\expandafter\cbRGB\expandafter{\detokenize{255,212,212}}{service\strut} \setlength{\fboxsep}{0pt}\def\cbRGB{\colorbox[RGB]}\expandafter\cbRGB\expandafter{\detokenize{255,225,225}}{restaurant\strut} \setlength{\fboxsep}{0pt}\def\cbRGB{\colorbox[RGB]}\expandafter\cbRGB\expandafter{\detokenize{255,216,216}}{.\strut} 

\setlength{\fboxsep}{0pt}\def\cbRGB{\colorbox[RGB]}\expandafter\cbRGB\expandafter{\detokenize{131,131,255}}{i\strut} \setlength{\fboxsep}{0pt}\def\cbRGB{\colorbox[RGB]}\expandafter\cbRGB\expandafter{\detokenize{157,157,255}}{love\strut} \setlength{\fboxsep}{0pt}\def\cbRGB{\colorbox[RGB]}\expandafter\cbRGB\expandafter{\detokenize{151,151,255}}{everything\strut} \setlength{\fboxsep}{0pt}\def\cbRGB{\colorbox[RGB]}\expandafter\cbRGB\expandafter{\detokenize{200,200,255}}{about\strut} \setlength{\fboxsep}{0pt}\def\cbRGB{\colorbox[RGB]}\expandafter\cbRGB\expandafter{\detokenize{201,201,255}}{this\strut} \setlength{\fboxsep}{0pt}\def\cbRGB{\colorbox[RGB]}\expandafter\cbRGB\expandafter{\detokenize{199,199,255}}{place\strut} \setlength{\fboxsep}{0pt}\def\cbRGB{\colorbox[RGB]}\expandafter\cbRGB\expandafter{\detokenize{190,190,255}}{,\strut} \setlength{\fboxsep}{0pt}\def\cbRGB{\colorbox[RGB]}\expandafter\cbRGB\expandafter{\detokenize{191,191,255}}{i\strut} \setlength{\fboxsep}{0pt}\def\cbRGB{\colorbox[RGB]}\expandafter\cbRGB\expandafter{\detokenize{190,190,255}}{find\strut} \setlength{\fboxsep}{0pt}\def\cbRGB{\colorbox[RGB]}\expandafter\cbRGB\expandafter{\detokenize{196,196,255}}{my\strut} \setlength{\fboxsep}{0pt}\def\cbRGB{\colorbox[RGB]}\expandafter\cbRGB\expandafter{\detokenize{189,189,255}}{self\strut} \setlength{\fboxsep}{0pt}\def\cbRGB{\colorbox[RGB]}\expandafter\cbRGB\expandafter{\detokenize{186,186,255}}{there\strut} \setlength{\fboxsep}{0pt}\def\cbRGB{\colorbox[RGB]}\expandafter\cbRGB\expandafter{\detokenize{182,182,255}}{more\strut} \setlength{\fboxsep}{0pt}\def\cbRGB{\colorbox[RGB]}\expandafter\cbRGB\expandafter{\detokenize{184,184,255}}{often\strut} \setlength{\fboxsep}{0pt}\def\cbRGB{\colorbox[RGB]}\expandafter\cbRGB\expandafter{\detokenize{187,187,255}}{then\strut} \setlength{\fboxsep}{0pt}\def\cbRGB{\colorbox[RGB]}\expandafter\cbRGB\expandafter{\detokenize{186,186,255}}{i\strut} \setlength{\fboxsep}{0pt}\def\cbRGB{\colorbox[RGB]}\expandafter\cbRGB\expandafter{\detokenize{192,192,255}}{want\strut} \setlength{\fboxsep}{0pt}\def\cbRGB{\colorbox[RGB]}\expandafter\cbRGB\expandafter{\detokenize{189,189,255}}{,\strut} \setlength{\fboxsep}{0pt}\def\cbRGB{\colorbox[RGB]}\expandafter\cbRGB\expandafter{\detokenize{189,189,255}}{the\strut} \setlength{\fboxsep}{0pt}\def\cbRGB{\colorbox[RGB]}\expandafter\cbRGB\expandafter{\detokenize{183,183,255}}{deserts\strut} \setlength{\fboxsep}{0pt}\def\cbRGB{\colorbox[RGB]}\expandafter\cbRGB\expandafter{\detokenize{187,187,255}}{are\strut} \setlength{\fboxsep}{0pt}\def\cbRGB{\colorbox[RGB]}\expandafter\cbRGB\expandafter{\detokenize{193,193,255}}{amazing\strut} \setlength{\fboxsep}{0pt}\def\cbRGB{\colorbox[RGB]}\expandafter\cbRGB\expandafter{\detokenize{193,193,255}}{,\strut} \setlength{\fboxsep}{0pt}\def\cbRGB{\colorbox[RGB]}\expandafter\cbRGB\expandafter{\detokenize{190,190,255}}{so\strut} \setlength{\fboxsep}{0pt}\def\cbRGB{\colorbox[RGB]}\expandafter\cbRGB\expandafter{\detokenize{184,184,255}}{many\strut} \setlength{\fboxsep}{0pt}\def\cbRGB{\colorbox[RGB]}\expandafter\cbRGB\expandafter{\detokenize{190,190,255}}{choices\strut} \setlength{\fboxsep}{0pt}\def\cbRGB{\colorbox[RGB]}\expandafter\cbRGB\expandafter{\detokenize{189,189,255}}{,\strut} \setlength{\fboxsep}{0pt}\def\cbRGB{\colorbox[RGB]}\expandafter\cbRGB\expandafter{\detokenize{169,169,255}}{the\strut} \setlength{\fboxsep}{0pt}\def\cbRGB{\colorbox[RGB]}\expandafter\cbRGB\expandafter{\detokenize{171,171,255}}{deli\strut} \setlength{\fboxsep}{0pt}\def\cbRGB{\colorbox[RGB]}\expandafter\cbRGB\expandafter{\detokenize{255,255,255}}{is\strut} \setlength{\fboxsep}{0pt}\def\cbRGB{\colorbox[RGB]}\expandafter\cbRGB\expandafter{\detokenize{125,125,255}}{ \strut} \setlength{\fboxsep}{0pt}\def\cbRGB{\colorbox[RGB]}\expandafter\cbRGB\expandafter{\detokenize{143,143,255}}{awesome\strut} \setlength{\fboxsep}{0pt}\def\cbRGB{\colorbox[RGB]}\expandafter\cbRGB\expandafter{\detokenize{103,103,255}}{,\strut} \setlength{\fboxsep}{0pt}\def\cbRGB{\colorbox[RGB]}\expandafter\cbRGB\expandafter{\detokenize{124,124,255}}{the\strut} \setlength{\fboxsep}{0pt}\def\cbRGB{\colorbox[RGB]}\expandafter\cbRGB\expandafter{\detokenize{110,110,255}}{sushi\strut} \setlength{\fboxsep}{0pt}\def\cbRGB{\colorbox[RGB]}\expandafter\cbRGB\expandafter{\detokenize{82,82,255}}{bar\strut} \setlength{\fboxsep}{0pt}\def\cbRGB{\colorbox[RGB]}\expandafter\cbRGB\expandafter{\detokenize{209,209,255}}{is\strut} \setlength{\fboxsep}{0pt}\def\cbRGB{\colorbox[RGB]}\expandafter\cbRGB\expandafter{\detokenize{204,204,255}}{great\strut} \setlength{\fboxsep}{0pt}\def\cbRGB{\colorbox[RGB]}\expandafter\cbRGB\expandafter{\detokenize{190,190,255}}{,\strut} \setlength{\fboxsep}{0pt}\def\cbRGB{\colorbox[RGB]}\expandafter\cbRGB\expandafter{\detokenize{190,190,255}}{you\strut} \setlength{\fboxsep}{0pt}\def\cbRGB{\colorbox[RGB]}\expandafter\cbRGB\expandafter{\detokenize{182,182,255}}{can\strut} \setlength{\fboxsep}{0pt}\def\cbRGB{\colorbox[RGB]}\expandafter\cbRGB\expandafter{\detokenize{186,186,255}}{not\strut} \setlength{\fboxsep}{0pt}\def\cbRGB{\colorbox[RGB]}\expandafter\cbRGB\expandafter{\detokenize{189,189,255}}{go\strut} \setlength{\fboxsep}{0pt}\def\cbRGB{\colorbox[RGB]}\expandafter\cbRGB\expandafter{\detokenize{189,189,255}}{wrong\strut} \setlength{\fboxsep}{0pt}\def\cbRGB{\colorbox[RGB]}\expandafter\cbRGB\expandafter{\detokenize{193,193,255}}{with\strut} \setlength{\fboxsep}{0pt}\def\cbRGB{\colorbox[RGB]}\expandafter\cbRGB\expandafter{\detokenize{195,195,255}}{this\strut} \setlength{\fboxsep}{0pt}\def\cbRGB{\colorbox[RGB]}\expandafter\cbRGB\expandafter{\detokenize{201,201,255}}{place\strut} \setlength{\fboxsep}{0pt}\def\cbRGB{\colorbox[RGB]}\expandafter\cbRGB\expandafter{\detokenize{192,192,255}}{,\strut} \setlength{\fboxsep}{0pt}\def\cbRGB{\colorbox[RGB]}\expandafter\cbRGB\expandafter{\detokenize{188,188,255}}{if\strut} \setlength{\fboxsep}{0pt}\def\cbRGB{\colorbox[RGB]}\expandafter\cbRGB\expandafter{\detokenize{184,184,255}}{you\strut} \setlength{\fboxsep}{0pt}\def\cbRGB{\colorbox[RGB]}\expandafter\cbRGB\expandafter{\detokenize{189,189,255}}{want\strut} \setlength{\fboxsep}{0pt}\def\cbRGB{\colorbox[RGB]}\expandafter\cbRGB\expandafter{\detokenize{189,189,255}}{a\strut} \setlength{\fboxsep}{0pt}\def\cbRGB{\colorbox[RGB]}\expandafter\cbRGB\expandafter{\detokenize{191,191,255}}{formal\strut} \setlength{\fboxsep}{0pt}\def\cbRGB{\colorbox[RGB]}\expandafter\cbRGB\expandafter{\detokenize{186,186,255}}{dining\strut} \setlength{\fboxsep}{0pt}\def\cbRGB{\colorbox[RGB]}\expandafter\cbRGB\expandafter{\detokenize{186,186,255}}{you\strut} \setlength{\fboxsep}{0pt}\def\cbRGB{\colorbox[RGB]}\expandafter\cbRGB\expandafter{\detokenize{191,191,255}}{can\strut} \setlength{\fboxsep}{0pt}\def\cbRGB{\colorbox[RGB]}\expandafter\cbRGB\expandafter{\detokenize{197,197,255}}{sit\strut} \setlength{\fboxsep}{0pt}\def\cbRGB{\colorbox[RGB]}\expandafter\cbRGB\expandafter{\detokenize{200,200,255}}{at\strut} \setlength{\fboxsep}{0pt}\def\cbRGB{\colorbox[RGB]}\expandafter\cbRGB\expandafter{\detokenize{25,25,255}}{the\strut} \setlength{\fboxsep}{0pt}\def\cbRGB{\colorbox[RGB]}\expandafter\cbRGB\expandafter{\detokenize{21,21,255}}{full\strut} \setlength{\fboxsep}{0pt}\def\cbRGB{\colorbox[RGB]}\expandafter\cbRGB\expandafter{\detokenize{0,0,255}}{service\strut} \setlength{\fboxsep}{0pt}\def\cbRGB{\colorbox[RGB]}\expandafter\cbRGB\expandafter{\detokenize{175,175,255}}{restaurant\strut} \setlength{\fboxsep}{0pt}\def\cbRGB{\colorbox[RGB]}\expandafter\cbRGB\expandafter{\detokenize{150,150,255}}{.\strut} 

\par
\textbf{Example 2}
\setlength{\fboxsep}{0pt}\def\cbRGB{\colorbox[RGB]}\expandafter\cbRGB\expandafter{\detokenize{255,201,201}}{we\strut} \setlength{\fboxsep}{0pt}\def\cbRGB{\colorbox[RGB]}\expandafter\cbRGB\expandafter{\detokenize{255,207,207}}{have\strut} \setlength{\fboxsep}{0pt}\def\cbRGB{\colorbox[RGB]}\expandafter\cbRGB\expandafter{\detokenize{255,206,206}}{just\strut} \setlength{\fboxsep}{0pt}\def\cbRGB{\colorbox[RGB]}\expandafter\cbRGB\expandafter{\detokenize{255,207,207}}{returned\strut} \setlength{\fboxsep}{0pt}\def\cbRGB{\colorbox[RGB]}\expandafter\cbRGB\expandafter{\detokenize{255,207,207}}{from\strut} \setlength{\fboxsep}{0pt}\def\cbRGB{\colorbox[RGB]}\expandafter\cbRGB\expandafter{\detokenize{255,207,207}}{our\strut} \setlength{\fboxsep}{0pt}\def\cbRGB{\colorbox[RGB]}\expandafter\cbRGB\expandafter{\detokenize{255,208,208}}{first\strut} \setlength{\fboxsep}{0pt}\def\cbRGB{\colorbox[RGB]}\expandafter\cbRGB\expandafter{\detokenize{255,215,215}}{trip\strut} \setlength{\fboxsep}{0pt}\def\cbRGB{\colorbox[RGB]}\expandafter\cbRGB\expandafter{\detokenize{255,188,188}}{to\strut} \setlength{\fboxsep}{0pt}\def\cbRGB{\colorbox[RGB]}\expandafter\cbRGB\expandafter{\detokenize{255,191,191}}{rome\strut} \setlength{\fboxsep}{0pt}\def\cbRGB{\colorbox[RGB]}\expandafter\cbRGB\expandafter{\detokenize{255,188,188}}{and\strut} \setlength{\fboxsep}{0pt}\def\cbRGB{\colorbox[RGB]}\expandafter\cbRGB\expandafter{\detokenize{255,217,217}}{what\strut} \setlength{\fboxsep}{0pt}\def\cbRGB{\colorbox[RGB]}\expandafter\cbRGB\expandafter{\detokenize{255,111,111}}{a\strut} \setlength{\fboxsep}{0pt}\def\cbRGB{\colorbox[RGB]}\expandafter\cbRGB\expandafter{\detokenize{255,111,111}}{wonderful\strut} \setlength{\fboxsep}{0pt}\def\cbRGB{\colorbox[RGB]}\expandafter\cbRGB\expandafter{\detokenize{255,110,110}}{time\strut} \setlength{\fboxsep}{0pt}\def\cbRGB{\colorbox[RGB]}\expandafter\cbRGB\expandafter{\detokenize{255,209,209}}{we\strut} \setlength{\fboxsep}{0pt}\def\cbRGB{\colorbox[RGB]}\expandafter\cbRGB\expandafter{\detokenize{255,212,212}}{had\strut} \setlength{\fboxsep}{0pt}\def\cbRGB{\colorbox[RGB]}\expandafter\cbRGB\expandafter{\detokenize{255,208,208}}{!\strut} \setlength{\fboxsep}{0pt}\def\cbRGB{\colorbox[RGB]}\expandafter\cbRGB\expandafter{\detokenize{255,208,208}}{our\strut} \setlength{\fboxsep}{0pt}\def\cbRGB{\colorbox[RGB]}\expandafter\cbRGB\expandafter{\detokenize{255,202,202}}{stay\strut} \setlength{\fboxsep}{0pt}\def\cbRGB{\colorbox[RGB]}\expandafter\cbRGB\expandafter{\detokenize{255,205,205}}{at\strut} \setlength{\fboxsep}{0pt}\def\cbRGB{\colorbox[RGB]}\expandafter\cbRGB\expandafter{\detokenize{255,213,213}}{unk\strut} \setlength{\fboxsep}{0pt}\def\cbRGB{\colorbox[RGB]}\expandafter\cbRGB\expandafter{\detokenize{255,209,209}}{unk\strut} \setlength{\fboxsep}{0pt}\def\cbRGB{\colorbox[RGB]}\expandafter\cbRGB\expandafter{\detokenize{255,140,140}}{was\strut} \setlength{\fboxsep}{0pt}\def\cbRGB{\colorbox[RGB]}\expandafter\cbRGB\expandafter{\detokenize{255,126,126}}{everything\strut} \setlength{\fboxsep}{0pt}\def\cbRGB{\colorbox[RGB]}\expandafter\cbRGB\expandafter{\detokenize{255,119,119}}{we\strut} \setlength{\fboxsep}{0pt}\def\cbRGB{\colorbox[RGB]}\expandafter\cbRGB\expandafter{\detokenize{255,185,185}}{had\strut} \setlength{\fboxsep}{0pt}\def\cbRGB{\colorbox[RGB]}\expandafter\cbRGB\expandafter{\detokenize{255,193,193}}{hoped\strut} \setlength{\fboxsep}{0pt}\def\cbRGB{\colorbox[RGB]}\expandafter\cbRGB\expandafter{\detokenize{255,209,209}}{for\strut} \setlength{\fboxsep}{0pt}\def\cbRGB{\colorbox[RGB]}\expandafter\cbRGB\expandafter{\detokenize{255,211,211}}{and\strut} \setlength{\fboxsep}{0pt}\def\cbRGB{\colorbox[RGB]}\expandafter\cbRGB\expandafter{\detokenize{255,209,209}}{all\strut} \setlength{\fboxsep}{0pt}\def\cbRGB{\colorbox[RGB]}\expandafter\cbRGB\expandafter{\detokenize{255,203,203}}{we\strut} \setlength{\fboxsep}{0pt}\def\cbRGB{\colorbox[RGB]}\expandafter\cbRGB\expandafter{\detokenize{255,204,204}}{had\strut} \setlength{\fboxsep}{0pt}\def\cbRGB{\colorbox[RGB]}\expandafter\cbRGB\expandafter{\detokenize{255,206,206}}{read\strut} \setlength{\fboxsep}{0pt}\def\cbRGB{\colorbox[RGB]}\expandafter\cbRGB\expandafter{\detokenize{255,206,206}}{about\strut} \setlength{\fboxsep}{0pt}\def\cbRGB{\colorbox[RGB]}\expandafter\cbRGB\expandafter{\detokenize{255,202,202}}{it\strut} \setlength{\fboxsep}{0pt}\def\cbRGB{\colorbox[RGB]}\expandafter\cbRGB\expandafter{\detokenize{255,201,201}}{was\strut} \setlength{\fboxsep}{0pt}\def\cbRGB{\colorbox[RGB]}\expandafter\cbRGB\expandafter{\detokenize{255,209,209}}{true\strut} \setlength{\fboxsep}{0pt}\def\cbRGB{\colorbox[RGB]}\expandafter\cbRGB\expandafter{\detokenize{255,207,207}}{!\strut} \setlength{\fboxsep}{0pt}\def\cbRGB{\colorbox[RGB]}\expandafter\cbRGB\expandafter{\detokenize{255,207,207}}{the\strut} \setlength{\fboxsep}{0pt}\def\cbRGB{\colorbox[RGB]}\expandafter\cbRGB\expandafter{\detokenize{255,200,200}}{room\strut} \setlength{\fboxsep}{0pt}\def\cbRGB{\colorbox[RGB]}\expandafter\cbRGB\expandafter{\detokenize{255,204,204}}{was\strut} \setlength{\fboxsep}{0pt}\def\cbRGB{\colorbox[RGB]}\expandafter\cbRGB\expandafter{\detokenize{255,213,213}}{small\strut} \setlength{\fboxsep}{0pt}\def\cbRGB{\colorbox[RGB]}\expandafter\cbRGB\expandafter{\detokenize{255,212,212}}{,\strut} \setlength{\fboxsep}{0pt}\def\cbRGB{\colorbox[RGB]}\expandafter\cbRGB\expandafter{\detokenize{255,204,204}}{but\strut} \setlength{\fboxsep}{0pt}\def\cbRGB{\colorbox[RGB]}\expandafter\cbRGB\expandafter{\detokenize{255,99,99}}{very\strut} \setlength{\fboxsep}{0pt}\def\cbRGB{\colorbox[RGB]}\expandafter\cbRGB\expandafter{\detokenize{255,103,103}}{clean\strut} \setlength{\fboxsep}{0pt}\def\cbRGB{\colorbox[RGB]}\expandafter\cbRGB\expandafter{\detokenize{255,0,0}}{and\strut} \setlength{\fboxsep}{0pt}\def\cbRGB{\colorbox[RGB]}\expandafter\cbRGB\expandafter{\detokenize{255,107,107}}{comfortable\strut} \setlength{\fboxsep}{0pt}\def\cbRGB{\colorbox[RGB]}\expandafter\cbRGB\expandafter{\detokenize{255,112,112}}{.\strut} \setlength{\fboxsep}{0pt}\def\cbRGB{\colorbox[RGB]}\expandafter\cbRGB\expandafter{\detokenize{255,234,234}}{the\strut} \setlength{\fboxsep}{0pt}\def\cbRGB{\colorbox[RGB]}\expandafter\cbRGB\expandafter{\detokenize{255,220,220}}{staff\strut} \setlength{\fboxsep}{0pt}\def\cbRGB{\colorbox[RGB]}\expandafter\cbRGB\expandafter{\detokenize{255,77,77}}{was\strut} \setlength{\fboxsep}{0pt}\def\cbRGB{\colorbox[RGB]}\expandafter\cbRGB\expandafter{\detokenize{255,66,66}}{fantastic\strut} \setlength{\fboxsep}{0pt}\def\cbRGB{\colorbox[RGB]}\expandafter\cbRGB\expandafter{\detokenize{255,16,16}}{and\strut} \setlength{\fboxsep}{0pt}\def\cbRGB{\colorbox[RGB]}\expandafter\cbRGB\expandafter{\detokenize{255,160,160}}{friendly\strut} \setlength{\fboxsep}{0pt}\def\cbRGB{\colorbox[RGB]}\expandafter\cbRGB\expandafter{\detokenize{255,161,161}}{,\strut} \setlength{\fboxsep}{0pt}\def\cbRGB{\colorbox[RGB]}\expandafter\cbRGB\expandafter{\detokenize{255,159,159}}{very\strut} \setlength{\fboxsep}{0pt}\def\cbRGB{\colorbox[RGB]}\expandafter\cbRGB\expandafter{\detokenize{255,153,153}}{helpful\strut} \setlength{\fboxsep}{0pt}\def\cbRGB{\colorbox[RGB]}\expandafter\cbRGB\expandafter{\detokenize{255,154,154}}{and\strut} \setlength{\fboxsep}{0pt}\def\cbRGB{\colorbox[RGB]}\expandafter\cbRGB\expandafter{\detokenize{255,209,209}}{the\strut} \setlength{\fboxsep}{0pt}\def\cbRGB{\colorbox[RGB]}\expandafter\cbRGB\expandafter{\detokenize{255,206,206}}{breakfast\strut} \setlength{\fboxsep}{0pt}\def\cbRGB{\colorbox[RGB]}\expandafter\cbRGB\expandafter{\detokenize{255,211,211}}{every\strut} \setlength{\fboxsep}{0pt}\def\cbRGB{\colorbox[RGB]}\expandafter\cbRGB\expandafter{\detokenize{255,206,206}}{day\strut} \setlength{\fboxsep}{0pt}\def\cbRGB{\colorbox[RGB]}\expandafter\cbRGB\expandafter{\detokenize{255,98,98}}{was\strut} \setlength{\fboxsep}{0pt}\def\cbRGB{\colorbox[RGB]}\expandafter\cbRGB\expandafter{\detokenize{255,109,109}}{wonderful\strut} \setlength{\fboxsep}{0pt}\def\cbRGB{\colorbox[RGB]}\expandafter\cbRGB\expandafter{\detokenize{255,110,110}}{!\strut} \setlength{\fboxsep}{0pt}\def\cbRGB{\colorbox[RGB]}\expandafter\cbRGB\expandafter{\detokenize{255,131,131}}{we\strut} \setlength{\fboxsep}{0pt}\def\cbRGB{\colorbox[RGB]}\expandafter\cbRGB\expandafter{\detokenize{255,140,140}}{loved\strut} \setlength{\fboxsep}{0pt}\def\cbRGB{\colorbox[RGB]}\expandafter\cbRGB\expandafter{\detokenize{255,126,126}}{the\strut} \setlength{\fboxsep}{0pt}\def\cbRGB{\colorbox[RGB]}\expandafter\cbRGB\expandafter{\detokenize{255,228,228}}{lift\strut} \setlength{\fboxsep}{0pt}\def\cbRGB{\colorbox[RGB]}\expandafter\cbRGB\expandafter{\detokenize{255,222,222}}{up\strut} \setlength{\fboxsep}{0pt}\def\cbRGB{\colorbox[RGB]}\expandafter\cbRGB\expandafter{\detokenize{255,239,239}}{to\strut} \setlength{\fboxsep}{0pt}\def\cbRGB{\colorbox[RGB]}\expandafter\cbRGB\expandafter{\detokenize{255,222,222}}{the\strut} \setlength{\fboxsep}{0pt}\def\cbRGB{\colorbox[RGB]}\expandafter\cbRGB\expandafter{\detokenize{255,211,211}}{rooms\strut} \setlength{\fboxsep}{0pt}\def\cbRGB{\colorbox[RGB]}\expandafter\cbRGB\expandafter{\detokenize{255,210,210}}{!\strut} \setlength{\fboxsep}{0pt}\def\cbRGB{\colorbox[RGB]}\expandafter\cbRGB\expandafter{\detokenize{255,213,213}}{the\strut} \setlength{\fboxsep}{0pt}\def\cbRGB{\colorbox[RGB]}\expandafter\cbRGB\expandafter{\detokenize{255,206,206}}{location\strut} \setlength{\fboxsep}{0pt}\def\cbRGB{\colorbox[RGB]}\expandafter\cbRGB\expandafter{\detokenize{255,125,125}}{was\strut} \setlength{\fboxsep}{0pt}\def\cbRGB{\colorbox[RGB]}\expandafter\cbRGB\expandafter{\detokenize{255,130,130}}{great\strut} \setlength{\fboxsep}{0pt}\def\cbRGB{\colorbox[RGB]}\expandafter\cbRGB\expandafter{\detokenize{255,131,131}}{and\strut} \setlength{\fboxsep}{0pt}\def\cbRGB{\colorbox[RGB]}\expandafter\cbRGB\expandafter{\detokenize{255,214,214}}{at\strut} \setlength{\fboxsep}{0pt}\def\cbRGB{\colorbox[RGB]}\expandafter\cbRGB\expandafter{\detokenize{255,208,208}}{night\strut} \setlength{\fboxsep}{0pt}\def\cbRGB{\colorbox[RGB]}\expandafter\cbRGB\expandafter{\detokenize{255,206,206}}{we\strut} \setlength{\fboxsep}{0pt}\def\cbRGB{\colorbox[RGB]}\expandafter\cbRGB\expandafter{\detokenize{255,207,207}}{left\strut} \setlength{\fboxsep}{0pt}\def\cbRGB{\colorbox[RGB]}\expandafter\cbRGB\expandafter{\detokenize{255,206,206}}{the\strut} \setlength{\fboxsep}{0pt}\def\cbRGB{\colorbox[RGB]}\expandafter\cbRGB\expandafter{\detokenize{255,206,206}}{windows\strut} \setlength{\fboxsep}{0pt}\def\cbRGB{\colorbox[RGB]}\expandafter\cbRGB\expandafter{\detokenize{255,210,210}}{open\strut} \setlength{\fboxsep}{0pt}\def\cbRGB{\colorbox[RGB]}\expandafter\cbRGB\expandafter{\detokenize{255,210,210}}{and\strut} \setlength{\fboxsep}{0pt}\def\cbRGB{\colorbox[RGB]}\expandafter\cbRGB\expandafter{\detokenize{255,219,219}}{listened\strut} \setlength{\fboxsep}{0pt}\def\cbRGB{\colorbox[RGB]}\expandafter\cbRGB\expandafter{\detokenize{255,214,214}}{to\strut} \setlength{\fboxsep}{0pt}\def\cbRGB{\colorbox[RGB]}\expandafter\cbRGB\expandafter{\detokenize{255,170,170}}{the\strut} \setlength{\fboxsep}{0pt}\def\cbRGB{\colorbox[RGB]}\expandafter\cbRGB\expandafter{\detokenize{255,169,169}}{bustle\strut} \setlength{\fboxsep}{0pt}\def\cbRGB{\colorbox[RGB]}\expandafter\cbRGB\expandafter{\detokenize{255,184,184}}{of\strut} \setlength{\fboxsep}{0pt}\def\cbRGB{\colorbox[RGB]}\expandafter\cbRGB\expandafter{\detokenize{255,229,229}}{the\strut} \setlength{\fboxsep}{0pt}\def\cbRGB{\colorbox[RGB]}\expandafter\cbRGB\expandafter{\detokenize{255,145,145}}{street\strut} \setlength{\fboxsep}{0pt}\def\cbRGB{\colorbox[RGB]}\expandafter\cbRGB\expandafter{\detokenize{255,147,147}}{below\strut} \setlength{\fboxsep}{0pt}\def\cbRGB{\colorbox[RGB]}\expandafter\cbRGB\expandafter{\detokenize{255,159,159}}{and\strut} \setlength{\fboxsep}{0pt}\def\cbRGB{\colorbox[RGB]}\expandafter\cbRGB\expandafter{\detokenize{255,255,255}}{a\strut} \setlength{\fboxsep}{0pt}\def\cbRGB{\colorbox[RGB]}\expandafter\cbRGB\expandafter{\detokenize{255,253,253}}{unk\strut} \setlength{\fboxsep}{0pt}\def\cbRGB{\colorbox[RGB]}\expandafter\cbRGB\expandafter{\detokenize{255,175,175}}{off\strut} \setlength{\fboxsep}{0pt}\def\cbRGB{\colorbox[RGB]}\expandafter\cbRGB\expandafter{\detokenize{255,176,176}}{somewhere\strut} \setlength{\fboxsep}{0pt}\def\cbRGB{\colorbox[RGB]}\expandafter\cbRGB\expandafter{\detokenize{255,180,180}}{{\ldots}\strut} \setlength{\fboxsep}{0pt}\def\cbRGB{\colorbox[RGB]}\expandafter\cbRGB\expandafter{\detokenize{255,246,246}}{we\strut} \setlength{\fboxsep}{0pt}\def\cbRGB{\colorbox[RGB]}\expandafter\cbRGB\expandafter{\detokenize{255,228,228}}{felt\strut} \setlength{\fboxsep}{0pt}\def\cbRGB{\colorbox[RGB]}\expandafter\cbRGB\expandafter{\detokenize{255,207,207}}{so\strut} \setlength{\fboxsep}{0pt}\def\cbRGB{\colorbox[RGB]}\expandafter\cbRGB\expandafter{\detokenize{255,176,176}}{'\strut} \setlength{\fboxsep}{0pt}\def\cbRGB{\colorbox[RGB]}\expandafter\cbRGB\expandafter{\detokenize{255,171,171}}{roman\strut} \setlength{\fboxsep}{0pt}\def\cbRGB{\colorbox[RGB]}\expandafter\cbRGB\expandafter{\detokenize{255,176,176}}{"\strut} \setlength{\fboxsep}{0pt}\def\cbRGB{\colorbox[RGB]}\expandafter\cbRGB\expandafter{\detokenize{255,202,202}}{!\strut} \setlength{\fboxsep}{0pt}\def\cbRGB{\colorbox[RGB]}\expandafter\cbRGB\expandafter{\detokenize{255,205,205}}{they\strut} \setlength{\fboxsep}{0pt}\def\cbRGB{\colorbox[RGB]}\expandafter\cbRGB\expandafter{\detokenize{255,203,203}}{gave\strut} \setlength{\fboxsep}{0pt}\def\cbRGB{\colorbox[RGB]}\expandafter\cbRGB\expandafter{\detokenize{255,200,200}}{us\strut} \setlength{\fboxsep}{0pt}\def\cbRGB{\colorbox[RGB]}\expandafter\cbRGB\expandafter{\detokenize{255,202,202}}{champagne\strut} \setlength{\fboxsep}{0pt}\def\cbRGB{\colorbox[RGB]}\expandafter\cbRGB\expandafter{\detokenize{255,202,202}}{on\strut} \setlength{\fboxsep}{0pt}\def\cbRGB{\colorbox[RGB]}\expandafter\cbRGB\expandafter{\detokenize{255,208,208}}{our\strut} \setlength{\fboxsep}{0pt}\def\cbRGB{\colorbox[RGB]}\expandafter\cbRGB\expandafter{\detokenize{255,205,205}}{last\strut} \setlength{\fboxsep}{0pt}\def\cbRGB{\colorbox[RGB]}\expandafter\cbRGB\expandafter{\detokenize{255,225,225}}{night\strut} \setlength{\fboxsep}{0pt}\def\cbRGB{\colorbox[RGB]}\expandafter\cbRGB\expandafter{\detokenize{255,239,239}}{!\strut} \setlength{\fboxsep}{0pt}\def\cbRGB{\colorbox[RGB]}\expandafter\cbRGB\expandafter{\detokenize{255,169,169}}{the\strut} \setlength{\fboxsep}{0pt}\def\cbRGB{\colorbox[RGB]}\expandafter\cbRGB\expandafter{\detokenize{255,164,164}}{gentleman\strut} \setlength{\fboxsep}{0pt}\def\cbRGB{\colorbox[RGB]}\expandafter\cbRGB\expandafter{\detokenize{255,168,168}}{who\strut} \setlength{\fboxsep}{0pt}\def\cbRGB{\colorbox[RGB]}\expandafter\cbRGB\expandafter{\detokenize{255,240,240}}{works\strut} \setlength{\fboxsep}{0pt}\def\cbRGB{\colorbox[RGB]}\expandafter\cbRGB\expandafter{\detokenize{255,223,223}}{the\strut} \setlength{\fboxsep}{0pt}\def\cbRGB{\colorbox[RGB]}\expandafter\cbRGB\expandafter{\detokenize{255,208,208}}{night\strut} \setlength{\fboxsep}{0pt}\def\cbRGB{\colorbox[RGB]}\expandafter\cbRGB\expandafter{\detokenize{255,207,207}}{shift\strut} \setlength{\fboxsep}{0pt}\def\cbRGB{\colorbox[RGB]}\expandafter\cbRGB\expandafter{\detokenize{255,207,207}}{even\strut} \setlength{\fboxsep}{0pt}\def\cbRGB{\colorbox[RGB]}\expandafter\cbRGB\expandafter{\detokenize{255,206,206}}{gave\strut} \setlength{\fboxsep}{0pt}\def\cbRGB{\colorbox[RGB]}\expandafter\cbRGB\expandafter{\detokenize{255,213,213}}{us\strut} \setlength{\fboxsep}{0pt}\def\cbRGB{\colorbox[RGB]}\expandafter\cbRGB\expandafter{\detokenize{255,213,213}}{a\strut} \setlength{\fboxsep}{0pt}\def\cbRGB{\colorbox[RGB]}\expandafter\cbRGB\expandafter{\detokenize{255,202,202}}{personal\strut} \setlength{\fboxsep}{0pt}\def\cbRGB{\colorbox[RGB]}\expandafter\cbRGB\expandafter{\detokenize{255,205,205}}{escort\strut} \setlength{\fboxsep}{0pt}\def\cbRGB{\colorbox[RGB]}\expandafter\cbRGB\expandafter{\detokenize{255,206,206}}{to\strut} \setlength{\fboxsep}{0pt}\def\cbRGB{\colorbox[RGB]}\expandafter\cbRGB\expandafter{\detokenize{255,217,217}}{unk\strut} \setlength{\fboxsep}{0pt}\def\cbRGB{\colorbox[RGB]}\expandafter\cbRGB\expandafter{\detokenize{255,216,216}}{unk\strut} \setlength{\fboxsep}{0pt}\def\cbRGB{\colorbox[RGB]}\expandafter\cbRGB\expandafter{\detokenize{255,218,218}}{,\strut} \setlength{\fboxsep}{0pt}\def\cbRGB{\colorbox[RGB]}\expandafter\cbRGB\expandafter{\detokenize{255,193,193}}{a\strut} \setlength{\fboxsep}{0pt}\def\cbRGB{\colorbox[RGB]}\expandafter\cbRGB\expandafter{\detokenize{255,193,193}}{maze\strut} \setlength{\fboxsep}{0pt}\def\cbRGB{\colorbox[RGB]}\expandafter\cbRGB\expandafter{\detokenize{255,189,189}}{of\strut} \setlength{\fboxsep}{0pt}\def\cbRGB{\colorbox[RGB]}\expandafter\cbRGB\expandafter{\detokenize{255,179,179}}{metro\strut} \setlength{\fboxsep}{0pt}\def\cbRGB{\colorbox[RGB]}\expandafter\cbRGB\expandafter{\detokenize{255,178,178}}{stops\strut} \setlength{\fboxsep}{0pt}\def\cbRGB{\colorbox[RGB]}\expandafter\cbRGB\expandafter{\detokenize{255,183,183}}{,\strut} \setlength{\fboxsep}{0pt}\def\cbRGB{\colorbox[RGB]}\expandafter\cbRGB\expandafter{\detokenize{255,216,216}}{and\strut} \setlength{\fboxsep}{0pt}\def\cbRGB{\colorbox[RGB]}\expandafter\cbRGB\expandafter{\detokenize{255,209,209}}{delivered\strut} \setlength{\fboxsep}{0pt}\def\cbRGB{\colorbox[RGB]}\expandafter\cbRGB\expandafter{\detokenize{255,203,203}}{us\strut} \setlength{\fboxsep}{0pt}\def\cbRGB{\colorbox[RGB]}\expandafter\cbRGB\expandafter{\detokenize{255,208,208}}{right\strut} \setlength{\fboxsep}{0pt}\def\cbRGB{\colorbox[RGB]}\expandafter\cbRGB\expandafter{\detokenize{255,207,207}}{at\strut} \setlength{\fboxsep}{0pt}\def\cbRGB{\colorbox[RGB]}\expandafter\cbRGB\expandafter{\detokenize{255,170,170}}{the\strut} \setlength{\fboxsep}{0pt}\def\cbRGB{\colorbox[RGB]}\expandafter\cbRGB\expandafter{\detokenize{255,175,175}}{gate\strut} \setlength{\fboxsep}{0pt}\def\cbRGB{\colorbox[RGB]}\expandafter\cbRGB\expandafter{\detokenize{255,176,176}}{!\strut} \setlength{\fboxsep}{0pt}\def\cbRGB{\colorbox[RGB]}\expandafter\cbRGB\expandafter{\detokenize{255,208,208}}{unk\strut} \setlength{\fboxsep}{0pt}\def\cbRGB{\colorbox[RGB]}\expandafter\cbRGB\expandafter{\detokenize{255,204,204}}{was\strut} \setlength{\fboxsep}{0pt}\def\cbRGB{\colorbox[RGB]}\expandafter\cbRGB\expandafter{\detokenize{255,205,205}}{just\strut} \setlength{\fboxsep}{0pt}\def\cbRGB{\colorbox[RGB]}\expandafter\cbRGB\expandafter{\detokenize{255,161,161}}{so\strut} \setlength{\fboxsep}{0pt}\def\cbRGB{\colorbox[RGB]}\expandafter\cbRGB\expandafter{\detokenize{255,170,170}}{nice\strut} \setlength{\fboxsep}{0pt}\def\cbRGB{\colorbox[RGB]}\expandafter\cbRGB\expandafter{\detokenize{255,159,159}}{!\strut} \setlength{\fboxsep}{0pt}\def\cbRGB{\colorbox[RGB]}\expandafter\cbRGB\expandafter{\detokenize{255,59,59}}{i\strut} \setlength{\fboxsep}{0pt}\def\cbRGB{\colorbox[RGB]}\expandafter\cbRGB\expandafter{\detokenize{255,53,53}}{highly\strut} \setlength{\fboxsep}{0pt}\def\cbRGB{\colorbox[RGB]}\expandafter\cbRGB\expandafter{\detokenize{255,74,74}}{recommend\strut} \setlength{\fboxsep}{0pt}\def\cbRGB{\colorbox[RGB]}\expandafter\cbRGB\expandafter{\detokenize{255,220,220}}{this\strut} \setlength{\fboxsep}{0pt}\def\cbRGB{\colorbox[RGB]}\expandafter\cbRGB\expandafter{\detokenize{255,213,213}}{small\strut} \setlength{\fboxsep}{0pt}\def\cbRGB{\colorbox[RGB]}\expandafter\cbRGB\expandafter{\detokenize{255,209,209}}{hotel\strut} \setlength{\fboxsep}{0pt}\def\cbRGB{\colorbox[RGB]}\expandafter\cbRGB\expandafter{\detokenize{255,208,208}}{and\strut} \setlength{\fboxsep}{0pt}\def\cbRGB{\colorbox[RGB]}\expandafter\cbRGB\expandafter{\detokenize{255,209,209}}{would\strut} \setlength{\fboxsep}{0pt}\def\cbRGB{\colorbox[RGB]}\expandafter\cbRGB\expandafter{\detokenize{255,204,204}}{absolutely\strut} \setlength{\fboxsep}{0pt}\def\cbRGB{\colorbox[RGB]}\expandafter\cbRGB\expandafter{\detokenize{255,205,205}}{stay\strut} \setlength{\fboxsep}{0pt}\def\cbRGB{\colorbox[RGB]}\expandafter\cbRGB\expandafter{\detokenize{255,205,205}}{there\strut} \setlength{\fboxsep}{0pt}\def\cbRGB{\colorbox[RGB]}\expandafter\cbRGB\expandafter{\detokenize{255,205,205}}{again\strut} \setlength{\fboxsep}{0pt}\def\cbRGB{\colorbox[RGB]}\expandafter\cbRGB\expandafter{\detokenize{255,207,207}}{if\strut} \setlength{\fboxsep}{0pt}\def\cbRGB{\colorbox[RGB]}\expandafter\cbRGB\expandafter{\detokenize{255,206,206}}{we\strut} \setlength{\fboxsep}{0pt}\def\cbRGB{\colorbox[RGB]}\expandafter\cbRGB\expandafter{\detokenize{255,207,207}}{are\strut} \setlength{\fboxsep}{0pt}\def\cbRGB{\colorbox[RGB]}\expandafter\cbRGB\expandafter{\detokenize{255,207,207}}{lucky\strut} \setlength{\fboxsep}{0pt}\def\cbRGB{\colorbox[RGB]}\expandafter\cbRGB\expandafter{\detokenize{255,211,211}}{enough\strut} \setlength{\fboxsep}{0pt}\def\cbRGB{\colorbox[RGB]}\expandafter\cbRGB\expandafter{\detokenize{255,209,209}}{to\strut} \setlength{\fboxsep}{0pt}\def\cbRGB{\colorbox[RGB]}\expandafter\cbRGB\expandafter{\detokenize{255,217,217}}{return\strut} \setlength{\fboxsep}{0pt}\def\cbRGB{\colorbox[RGB]}\expandafter\cbRGB\expandafter{\detokenize{255,208,208}}{!\strut} \setlength{\fboxsep}{0pt}\def\cbRGB{\colorbox[RGB]}\expandafter\cbRGB\expandafter{\detokenize{255,182,182}}{they\strut} \setlength{\fboxsep}{0pt}\def\cbRGB{\colorbox[RGB]}\expandafter\cbRGB\expandafter{\detokenize{255,176,176}}{deserve\strut} \setlength{\fboxsep}{0pt}\def\cbRGB{\colorbox[RGB]}\expandafter\cbRGB\expandafter{\detokenize{255,157,157}}{their\strut} \setlength{\fboxsep}{0pt}\def\cbRGB{\colorbox[RGB]}\expandafter\cbRGB\expandafter{\detokenize{255,198,198}}{\#\strut} \setlength{\fboxsep}{0pt}\def\cbRGB{\colorbox[RGB]}\expandafter\cbRGB\expandafter{\detokenize{255,196,196}}{qqq\strut} \setlength{\fboxsep}{0pt}\def\cbRGB{\colorbox[RGB]}\expandafter\cbRGB\expandafter{\detokenize{255,219,219}}{tripadvisor\strut} \setlength{\fboxsep}{0pt}\def\cbRGB{\colorbox[RGB]}\expandafter\cbRGB\expandafter{\detokenize{255,213,213}}{rating\strut} \setlength{\fboxsep}{0pt}\def\cbRGB{\colorbox[RGB]}\expandafter\cbRGB\expandafter{\detokenize{255,181,181}}{!\strut} \setlength{\fboxsep}{0pt}\def\cbRGB{\colorbox[RGB]}\expandafter\cbRGB\expandafter{\detokenize{255,177,177}}{thanks\strut} \setlength{\fboxsep}{0pt}\def\cbRGB{\colorbox[RGB]}\expandafter\cbRGB\expandafter{\detokenize{255,178,178}}{you\strut} \setlength{\fboxsep}{0pt}\def\cbRGB{\colorbox[RGB]}\expandafter\cbRGB\expandafter{\detokenize{255,208,208}}{,\strut} \setlength{\fboxsep}{0pt}\def\cbRGB{\colorbox[RGB]}\expandafter\cbRGB\expandafter{\detokenize{255,211,211}}{unk\strut} \setlength{\fboxsep}{0pt}\def\cbRGB{\colorbox[RGB]}\expandafter\cbRGB\expandafter{\detokenize{255,212,212}}{unk\strut} \setlength{\fboxsep}{0pt}\def\cbRGB{\colorbox[RGB]}\expandafter\cbRGB\expandafter{\detokenize{255,214,214}}{,\strut} \setlength{\fboxsep}{0pt}\def\cbRGB{\colorbox[RGB]}\expandafter\cbRGB\expandafter{\detokenize{255,207,207}}{for\strut} \setlength{\fboxsep}{0pt}\def\cbRGB{\colorbox[RGB]}\expandafter\cbRGB\expandafter{\detokenize{255,200,200}}{your\strut} \setlength{\fboxsep}{0pt}\def\cbRGB{\colorbox[RGB]}\expandafter\cbRGB\expandafter{\detokenize{255,201,201}}{part\strut} \setlength{\fboxsep}{0pt}\def\cbRGB{\colorbox[RGB]}\expandafter\cbRGB\expandafter{\detokenize{255,210,210}}{in\strut} \setlength{\fboxsep}{0pt}\def\cbRGB{\colorbox[RGB]}\expandafter\cbRGB\expandafter{\detokenize{255,208,208}}{making\strut} \setlength{\fboxsep}{0pt}\def\cbRGB{\colorbox[RGB]}\expandafter\cbRGB\expandafter{\detokenize{255,207,207}}{this\strut} \setlength{\fboxsep}{0pt}\def\cbRGB{\colorbox[RGB]}\expandafter\cbRGB\expandafter{\detokenize{255,213,213}}{trip\strut} \setlength{\fboxsep}{0pt}\def\cbRGB{\colorbox[RGB]}\expandafter\cbRGB\expandafter{\detokenize{255,213,213}}{such\strut} \setlength{\fboxsep}{0pt}\def\cbRGB{\colorbox[RGB]}\expandafter\cbRGB\expandafter{\detokenize{255,128,128}}{a\strut} \setlength{\fboxsep}{0pt}\def\cbRGB{\colorbox[RGB]}\expandafter\cbRGB\expandafter{\detokenize{255,123,123}}{pleasure\strut} \setlength{\fboxsep}{0pt}\def\cbRGB{\colorbox[RGB]}\expandafter\cbRGB\expandafter{\detokenize{255,116,116}}{!\strut} 

\setlength{\fboxsep}{0pt}\def\cbRGB{\colorbox[RGB]}\expandafter\cbRGB\expandafter{\detokenize{229,229,255}}{we\strut} \setlength{\fboxsep}{0pt}\def\cbRGB{\colorbox[RGB]}\expandafter\cbRGB\expandafter{\detokenize{238,238,255}}{have\strut} \setlength{\fboxsep}{0pt}\def\cbRGB{\colorbox[RGB]}\expandafter\cbRGB\expandafter{\detokenize{238,238,255}}{just\strut} \setlength{\fboxsep}{0pt}\def\cbRGB{\colorbox[RGB]}\expandafter\cbRGB\expandafter{\detokenize{238,238,255}}{returned\strut} \setlength{\fboxsep}{0pt}\def\cbRGB{\colorbox[RGB]}\expandafter\cbRGB\expandafter{\detokenize{237,237,255}}{from\strut} \setlength{\fboxsep}{0pt}\def\cbRGB{\colorbox[RGB]}\expandafter\cbRGB\expandafter{\detokenize{237,237,255}}{our\strut} \setlength{\fboxsep}{0pt}\def\cbRGB{\colorbox[RGB]}\expandafter\cbRGB\expandafter{\detokenize{238,238,255}}{first\strut} \setlength{\fboxsep}{0pt}\def\cbRGB{\colorbox[RGB]}\expandafter\cbRGB\expandafter{\detokenize{241,241,255}}{trip\strut} \setlength{\fboxsep}{0pt}\def\cbRGB{\colorbox[RGB]}\expandafter\cbRGB\expandafter{\detokenize{144,144,255}}{to\strut} \setlength{\fboxsep}{0pt}\def\cbRGB{\colorbox[RGB]}\expandafter\cbRGB\expandafter{\detokenize{142,142,255}}{rome\strut} \setlength{\fboxsep}{0pt}\def\cbRGB{\colorbox[RGB]}\expandafter\cbRGB\expandafter{\detokenize{124,124,255}}{and\strut} \setlength{\fboxsep}{0pt}\def\cbRGB{\colorbox[RGB]}\expandafter\cbRGB\expandafter{\detokenize{223,223,255}}{what\strut} \setlength{\fboxsep}{0pt}\def\cbRGB{\colorbox[RGB]}\expandafter\cbRGB\expandafter{\detokenize{224,224,255}}{a\strut} \setlength{\fboxsep}{0pt}\def\cbRGB{\colorbox[RGB]}\expandafter\cbRGB\expandafter{\detokenize{240,240,255}}{wonderful\strut} \setlength{\fboxsep}{0pt}\def\cbRGB{\colorbox[RGB]}\expandafter\cbRGB\expandafter{\detokenize{237,237,255}}{time\strut} \setlength{\fboxsep}{0pt}\def\cbRGB{\colorbox[RGB]}\expandafter\cbRGB\expandafter{\detokenize{237,237,255}}{we\strut} \setlength{\fboxsep}{0pt}\def\cbRGB{\colorbox[RGB]}\expandafter\cbRGB\expandafter{\detokenize{228,228,255}}{had\strut} \setlength{\fboxsep}{0pt}\def\cbRGB{\colorbox[RGB]}\expandafter\cbRGB\expandafter{\detokenize{201,201,255}}{!\strut} \setlength{\fboxsep}{0pt}\def\cbRGB{\colorbox[RGB]}\expandafter\cbRGB\expandafter{\detokenize{53,53,255}}{our\strut} \setlength{\fboxsep}{0pt}\def\cbRGB{\colorbox[RGB]}\expandafter\cbRGB\expandafter{\detokenize{38,38,255}}{stay\strut} \setlength{\fboxsep}{0pt}\def\cbRGB{\colorbox[RGB]}\expandafter\cbRGB\expandafter{\detokenize{41,41,255}}{at\strut} \setlength{\fboxsep}{0pt}\def\cbRGB{\colorbox[RGB]}\expandafter\cbRGB\expandafter{\detokenize{190,190,255}}{unk\strut} \setlength{\fboxsep}{0pt}\def\cbRGB{\colorbox[RGB]}\expandafter\cbRGB\expandafter{\detokenize{212,212,255}}{unk\strut} \setlength{\fboxsep}{0pt}\def\cbRGB{\colorbox[RGB]}\expandafter\cbRGB\expandafter{\detokenize{237,237,255}}{was\strut} \setlength{\fboxsep}{0pt}\def\cbRGB{\colorbox[RGB]}\expandafter\cbRGB\expandafter{\detokenize{236,236,255}}{everything\strut} \setlength{\fboxsep}{0pt}\def\cbRGB{\colorbox[RGB]}\expandafter\cbRGB\expandafter{\detokenize{236,236,255}}{we\strut} \setlength{\fboxsep}{0pt}\def\cbRGB{\colorbox[RGB]}\expandafter\cbRGB\expandafter{\detokenize{238,238,255}}{had\strut} \setlength{\fboxsep}{0pt}\def\cbRGB{\colorbox[RGB]}\expandafter\cbRGB\expandafter{\detokenize{238,238,255}}{hoped\strut} \setlength{\fboxsep}{0pt}\def\cbRGB{\colorbox[RGB]}\expandafter\cbRGB\expandafter{\detokenize{239,239,255}}{for\strut} \setlength{\fboxsep}{0pt}\def\cbRGB{\colorbox[RGB]}\expandafter\cbRGB\expandafter{\detokenize{238,238,255}}{and\strut} \setlength{\fboxsep}{0pt}\def\cbRGB{\colorbox[RGB]}\expandafter\cbRGB\expandafter{\detokenize{238,238,255}}{all\strut} \setlength{\fboxsep}{0pt}\def\cbRGB{\colorbox[RGB]}\expandafter\cbRGB\expandafter{\detokenize{237,237,255}}{we\strut} \setlength{\fboxsep}{0pt}\def\cbRGB{\colorbox[RGB]}\expandafter\cbRGB\expandafter{\detokenize{237,237,255}}{had\strut} \setlength{\fboxsep}{0pt}\def\cbRGB{\colorbox[RGB]}\expandafter\cbRGB\expandafter{\detokenize{238,238,255}}{read\strut} \setlength{\fboxsep}{0pt}\def\cbRGB{\colorbox[RGB]}\expandafter\cbRGB\expandafter{\detokenize{238,238,255}}{about\strut} \setlength{\fboxsep}{0pt}\def\cbRGB{\colorbox[RGB]}\expandafter\cbRGB\expandafter{\detokenize{237,237,255}}{it\strut} \setlength{\fboxsep}{0pt}\def\cbRGB{\colorbox[RGB]}\expandafter\cbRGB\expandafter{\detokenize{236,236,255}}{was\strut} \setlength{\fboxsep}{0pt}\def\cbRGB{\colorbox[RGB]}\expandafter\cbRGB\expandafter{\detokenize{242,242,255}}{true\strut} \setlength{\fboxsep}{0pt}\def\cbRGB{\colorbox[RGB]}\expandafter\cbRGB\expandafter{\detokenize{242,242,255}}{!\strut} \setlength{\fboxsep}{0pt}\def\cbRGB{\colorbox[RGB]}\expandafter\cbRGB\expandafter{\detokenize{150,150,255}}{the\strut} \setlength{\fboxsep}{0pt}\def\cbRGB{\colorbox[RGB]}\expandafter\cbRGB\expandafter{\detokenize{141,141,255}}{room\strut} \setlength{\fboxsep}{0pt}\def\cbRGB{\colorbox[RGB]}\expandafter\cbRGB\expandafter{\detokenize{127,127,255}}{was\strut} \setlength{\fboxsep}{0pt}\def\cbRGB{\colorbox[RGB]}\expandafter\cbRGB\expandafter{\detokenize{223,223,255}}{small\strut} \setlength{\fboxsep}{0pt}\def\cbRGB{\colorbox[RGB]}\expandafter\cbRGB\expandafter{\detokenize{225,225,255}}{,\strut} \setlength{\fboxsep}{0pt}\def\cbRGB{\colorbox[RGB]}\expandafter\cbRGB\expandafter{\detokenize{239,239,255}}{but\strut} \setlength{\fboxsep}{0pt}\def\cbRGB{\colorbox[RGB]}\expandafter\cbRGB\expandafter{\detokenize{236,236,255}}{very\strut} \setlength{\fboxsep}{0pt}\def\cbRGB{\colorbox[RGB]}\expandafter\cbRGB\expandafter{\detokenize{237,237,255}}{clean\strut} \setlength{\fboxsep}{0pt}\def\cbRGB{\colorbox[RGB]}\expandafter\cbRGB\expandafter{\detokenize{237,237,255}}{and\strut} \setlength{\fboxsep}{0pt}\def\cbRGB{\colorbox[RGB]}\expandafter\cbRGB\expandafter{\detokenize{240,240,255}}{comfortable\strut} \setlength{\fboxsep}{0pt}\def\cbRGB{\colorbox[RGB]}\expandafter\cbRGB\expandafter{\detokenize{239,239,255}}{.\strut} \setlength{\fboxsep}{0pt}\def\cbRGB{\colorbox[RGB]}\expandafter\cbRGB\expandafter{\detokenize{239,239,255}}{the\strut} \setlength{\fboxsep}{0pt}\def\cbRGB{\colorbox[RGB]}\expandafter\cbRGB\expandafter{\detokenize{236,236,255}}{staff\strut} \setlength{\fboxsep}{0pt}\def\cbRGB{\colorbox[RGB]}\expandafter\cbRGB\expandafter{\detokenize{237,237,255}}{was\strut} \setlength{\fboxsep}{0pt}\def\cbRGB{\colorbox[RGB]}\expandafter\cbRGB\expandafter{\detokenize{237,237,255}}{fantastic\strut} \setlength{\fboxsep}{0pt}\def\cbRGB{\colorbox[RGB]}\expandafter\cbRGB\expandafter{\detokenize{238,238,255}}{and\strut} \setlength{\fboxsep}{0pt}\def\cbRGB{\colorbox[RGB]}\expandafter\cbRGB\expandafter{\detokenize{240,240,255}}{friendly\strut} \setlength{\fboxsep}{0pt}\def\cbRGB{\colorbox[RGB]}\expandafter\cbRGB\expandafter{\detokenize{237,237,255}}{,\strut} \setlength{\fboxsep}{0pt}\def\cbRGB{\colorbox[RGB]}\expandafter\cbRGB\expandafter{\detokenize{193,193,255}}{very\strut} \setlength{\fboxsep}{0pt}\def\cbRGB{\colorbox[RGB]}\expandafter\cbRGB\expandafter{\detokenize{191,191,255}}{helpful\strut} \setlength{\fboxsep}{0pt}\def\cbRGB{\colorbox[RGB]}\expandafter\cbRGB\expandafter{\detokenize{194,194,255}}{and\strut} \setlength{\fboxsep}{0pt}\def\cbRGB{\colorbox[RGB]}\expandafter\cbRGB\expandafter{\detokenize{237,237,255}}{the\strut} \setlength{\fboxsep}{0pt}\def\cbRGB{\colorbox[RGB]}\expandafter\cbRGB\expandafter{\detokenize{237,237,255}}{breakfast\strut} \setlength{\fboxsep}{0pt}\def\cbRGB{\colorbox[RGB]}\expandafter\cbRGB\expandafter{\detokenize{238,238,255}}{every\strut} \setlength{\fboxsep}{0pt}\def\cbRGB{\colorbox[RGB]}\expandafter\cbRGB\expandafter{\detokenize{237,237,255}}{day\strut} \setlength{\fboxsep}{0pt}\def\cbRGB{\colorbox[RGB]}\expandafter\cbRGB\expandafter{\detokenize{236,236,255}}{was\strut} \setlength{\fboxsep}{0pt}\def\cbRGB{\colorbox[RGB]}\expandafter\cbRGB\expandafter{\detokenize{240,240,255}}{wonderful\strut} \setlength{\fboxsep}{0pt}\def\cbRGB{\colorbox[RGB]}\expandafter\cbRGB\expandafter{\detokenize{240,240,255}}{!\strut} \setlength{\fboxsep}{0pt}\def\cbRGB{\colorbox[RGB]}\expandafter\cbRGB\expandafter{\detokenize{240,240,255}}{we\strut} \setlength{\fboxsep}{0pt}\def\cbRGB{\colorbox[RGB]}\expandafter\cbRGB\expandafter{\detokenize{236,236,255}}{loved\strut} \setlength{\fboxsep}{0pt}\def\cbRGB{\colorbox[RGB]}\expandafter\cbRGB\expandafter{\detokenize{234,234,255}}{the\strut} \setlength{\fboxsep}{0pt}\def\cbRGB{\colorbox[RGB]}\expandafter\cbRGB\expandafter{\detokenize{235,235,255}}{lift\strut} \setlength{\fboxsep}{0pt}\def\cbRGB{\colorbox[RGB]}\expandafter\cbRGB\expandafter{\detokenize{235,235,255}}{up\strut} \setlength{\fboxsep}{0pt}\def\cbRGB{\colorbox[RGB]}\expandafter\cbRGB\expandafter{\detokenize{232,232,255}}{to\strut} \setlength{\fboxsep}{0pt}\def\cbRGB{\colorbox[RGB]}\expandafter\cbRGB\expandafter{\detokenize{39,39,255}}{the\strut} \setlength{\fboxsep}{0pt}\def\cbRGB{\colorbox[RGB]}\expandafter\cbRGB\expandafter{\detokenize{35,35,255}}{rooms\strut} \setlength{\fboxsep}{0pt}\def\cbRGB{\colorbox[RGB]}\expandafter\cbRGB\expandafter{\detokenize{0,0,255}}{!\strut} \setlength{\fboxsep}{0pt}\def\cbRGB{\colorbox[RGB]}\expandafter\cbRGB\expandafter{\detokenize{191,191,255}}{the\strut} \setlength{\fboxsep}{0pt}\def\cbRGB{\colorbox[RGB]}\expandafter\cbRGB\expandafter{\detokenize{195,195,255}}{location\strut} \setlength{\fboxsep}{0pt}\def\cbRGB{\colorbox[RGB]}\expandafter\cbRGB\expandafter{\detokenize{237,237,255}}{was\strut} \setlength{\fboxsep}{0pt}\def\cbRGB{\colorbox[RGB]}\expandafter\cbRGB\expandafter{\detokenize{238,238,255}}{great\strut} \setlength{\fboxsep}{0pt}\def\cbRGB{\colorbox[RGB]}\expandafter\cbRGB\expandafter{\detokenize{239,239,255}}{and\strut} \setlength{\fboxsep}{0pt}\def\cbRGB{\colorbox[RGB]}\expandafter\cbRGB\expandafter{\detokenize{238,238,255}}{at\strut} \setlength{\fboxsep}{0pt}\def\cbRGB{\colorbox[RGB]}\expandafter\cbRGB\expandafter{\detokenize{237,237,255}}{night\strut} \setlength{\fboxsep}{0pt}\def\cbRGB{\colorbox[RGB]}\expandafter\cbRGB\expandafter{\detokenize{237,237,255}}{we\strut} \setlength{\fboxsep}{0pt}\def\cbRGB{\colorbox[RGB]}\expandafter\cbRGB\expandafter{\detokenize{244,244,255}}{left\strut} \setlength{\fboxsep}{0pt}\def\cbRGB{\colorbox[RGB]}\expandafter\cbRGB\expandafter{\detokenize{251,251,255}}{the\strut} \setlength{\fboxsep}{0pt}\def\cbRGB{\colorbox[RGB]}\expandafter\cbRGB\expandafter{\detokenize{221,221,255}}{windows\strut} \setlength{\fboxsep}{0pt}\def\cbRGB{\colorbox[RGB]}\expandafter\cbRGB\expandafter{\detokenize{221,221,255}}{open\strut} \setlength{\fboxsep}{0pt}\def\cbRGB{\colorbox[RGB]}\expandafter\cbRGB\expandafter{\detokenize{221,221,255}}{and\strut} \setlength{\fboxsep}{0pt}\def\cbRGB{\colorbox[RGB]}\expandafter\cbRGB\expandafter{\detokenize{255,255,255}}{listened\strut} \setlength{\fboxsep}{0pt}\def\cbRGB{\colorbox[RGB]}\expandafter\cbRGB\expandafter{\detokenize{247,247,255}}{to\strut} \setlength{\fboxsep}{0pt}\def\cbRGB{\colorbox[RGB]}\expandafter\cbRGB\expandafter{\detokenize{239,239,255}}{the\strut} \setlength{\fboxsep}{0pt}\def\cbRGB{\colorbox[RGB]}\expandafter\cbRGB\expandafter{\detokenize{237,237,255}}{bustle\strut} \setlength{\fboxsep}{0pt}\def\cbRGB{\colorbox[RGB]}\expandafter\cbRGB\expandafter{\detokenize{237,237,255}}{of\strut} \setlength{\fboxsep}{0pt}\def\cbRGB{\colorbox[RGB]}\expandafter\cbRGB\expandafter{\detokenize{237,237,255}}{the\strut} \setlength{\fboxsep}{0pt}\def\cbRGB{\colorbox[RGB]}\expandafter\cbRGB\expandafter{\detokenize{237,237,255}}{street\strut} \setlength{\fboxsep}{0pt}\def\cbRGB{\colorbox[RGB]}\expandafter\cbRGB\expandafter{\detokenize{238,238,255}}{below\strut} \setlength{\fboxsep}{0pt}\def\cbRGB{\colorbox[RGB]}\expandafter\cbRGB\expandafter{\detokenize{240,240,255}}{and\strut} \setlength{\fboxsep}{0pt}\def\cbRGB{\colorbox[RGB]}\expandafter\cbRGB\expandafter{\detokenize{240,240,255}}{a\strut} \setlength{\fboxsep}{0pt}\def\cbRGB{\colorbox[RGB]}\expandafter\cbRGB\expandafter{\detokenize{239,239,255}}{unk\strut} \setlength{\fboxsep}{0pt}\def\cbRGB{\colorbox[RGB]}\expandafter\cbRGB\expandafter{\detokenize{237,237,255}}{off\strut} \setlength{\fboxsep}{0pt}\def\cbRGB{\colorbox[RGB]}\expandafter\cbRGB\expandafter{\detokenize{240,240,255}}{somewhere\strut} \setlength{\fboxsep}{0pt}\def\cbRGB{\colorbox[RGB]}\expandafter\cbRGB\expandafter{\detokenize{239,239,255}}{{\ldots}\strut} \setlength{\fboxsep}{0pt}\def\cbRGB{\colorbox[RGB]}\expandafter\cbRGB\expandafter{\detokenize{239,239,255}}{we\strut} \setlength{\fboxsep}{0pt}\def\cbRGB{\colorbox[RGB]}\expandafter\cbRGB\expandafter{\detokenize{238,238,255}}{felt\strut} \setlength{\fboxsep}{0pt}\def\cbRGB{\colorbox[RGB]}\expandafter\cbRGB\expandafter{\detokenize{241,241,255}}{so\strut} \setlength{\fboxsep}{0pt}\def\cbRGB{\colorbox[RGB]}\expandafter\cbRGB\expandafter{\detokenize{150,150,255}}{'\strut} \setlength{\fboxsep}{0pt}\def\cbRGB{\colorbox[RGB]}\expandafter\cbRGB\expandafter{\detokenize{147,147,255}}{roman\strut} \setlength{\fboxsep}{0pt}\def\cbRGB{\colorbox[RGB]}\expandafter\cbRGB\expandafter{\detokenize{135,135,255}}{"\strut} \setlength{\fboxsep}{0pt}\def\cbRGB{\colorbox[RGB]}\expandafter\cbRGB\expandafter{\detokenize{227,227,255}}{!\strut} \setlength{\fboxsep}{0pt}\def\cbRGB{\colorbox[RGB]}\expandafter\cbRGB\expandafter{\detokenize{233,233,255}}{they\strut} \setlength{\fboxsep}{0pt}\def\cbRGB{\colorbox[RGB]}\expandafter\cbRGB\expandafter{\detokenize{244,244,255}}{gave\strut} \setlength{\fboxsep}{0pt}\def\cbRGB{\colorbox[RGB]}\expandafter\cbRGB\expandafter{\detokenize{212,212,255}}{us\strut} \setlength{\fboxsep}{0pt}\def\cbRGB{\colorbox[RGB]}\expandafter\cbRGB\expandafter{\detokenize{208,208,255}}{champagne\strut} \setlength{\fboxsep}{0pt}\def\cbRGB{\colorbox[RGB]}\expandafter\cbRGB\expandafter{\detokenize{203,203,255}}{on\strut} \setlength{\fboxsep}{0pt}\def\cbRGB{\colorbox[RGB]}\expandafter\cbRGB\expandafter{\detokenize{237,237,255}}{our\strut} \setlength{\fboxsep}{0pt}\def\cbRGB{\colorbox[RGB]}\expandafter\cbRGB\expandafter{\detokenize{235,235,255}}{last\strut} \setlength{\fboxsep}{0pt}\def\cbRGB{\colorbox[RGB]}\expandafter\cbRGB\expandafter{\detokenize{241,241,255}}{night\strut} \setlength{\fboxsep}{0pt}\def\cbRGB{\colorbox[RGB]}\expandafter\cbRGB\expandafter{\detokenize{240,240,255}}{!\strut} \setlength{\fboxsep}{0pt}\def\cbRGB{\colorbox[RGB]}\expandafter\cbRGB\expandafter{\detokenize{240,240,255}}{the\strut} \setlength{\fboxsep}{0pt}\def\cbRGB{\colorbox[RGB]}\expandafter\cbRGB\expandafter{\detokenize{236,236,255}}{gentleman\strut} \setlength{\fboxsep}{0pt}\def\cbRGB{\colorbox[RGB]}\expandafter\cbRGB\expandafter{\detokenize{237,237,255}}{who\strut} \setlength{\fboxsep}{0pt}\def\cbRGB{\colorbox[RGB]}\expandafter\cbRGB\expandafter{\detokenize{237,237,255}}{works\strut} \setlength{\fboxsep}{0pt}\def\cbRGB{\colorbox[RGB]}\expandafter\cbRGB\expandafter{\detokenize{236,236,255}}{the\strut} \setlength{\fboxsep}{0pt}\def\cbRGB{\colorbox[RGB]}\expandafter\cbRGB\expandafter{\detokenize{237,237,255}}{night\strut} \setlength{\fboxsep}{0pt}\def\cbRGB{\colorbox[RGB]}\expandafter\cbRGB\expandafter{\detokenize{237,237,255}}{shift\strut} \setlength{\fboxsep}{0pt}\def\cbRGB{\colorbox[RGB]}\expandafter\cbRGB\expandafter{\detokenize{240,240,255}}{even\strut} \setlength{\fboxsep}{0pt}\def\cbRGB{\colorbox[RGB]}\expandafter\cbRGB\expandafter{\detokenize{239,239,255}}{gave\strut} \setlength{\fboxsep}{0pt}\def\cbRGB{\colorbox[RGB]}\expandafter\cbRGB\expandafter{\detokenize{241,241,255}}{us\strut} \setlength{\fboxsep}{0pt}\def\cbRGB{\colorbox[RGB]}\expandafter\cbRGB\expandafter{\detokenize{239,239,255}}{a\strut} \setlength{\fboxsep}{0pt}\def\cbRGB{\colorbox[RGB]}\expandafter\cbRGB\expandafter{\detokenize{239,239,255}}{personal\strut} \setlength{\fboxsep}{0pt}\def\cbRGB{\colorbox[RGB]}\expandafter\cbRGB\expandafter{\detokenize{240,240,255}}{escort\strut} \setlength{\fboxsep}{0pt}\def\cbRGB{\colorbox[RGB]}\expandafter\cbRGB\expandafter{\detokenize{239,239,255}}{to\strut} \setlength{\fboxsep}{0pt}\def\cbRGB{\colorbox[RGB]}\expandafter\cbRGB\expandafter{\detokenize{240,240,255}}{unk\strut} \setlength{\fboxsep}{0pt}\def\cbRGB{\colorbox[RGB]}\expandafter\cbRGB\expandafter{\detokenize{242,242,255}}{unk\strut} \setlength{\fboxsep}{0pt}\def\cbRGB{\colorbox[RGB]}\expandafter\cbRGB\expandafter{\detokenize{246,246,255}}{,\strut} \setlength{\fboxsep}{0pt}\def\cbRGB{\colorbox[RGB]}\expandafter\cbRGB\expandafter{\detokenize{200,200,255}}{a\strut} \setlength{\fboxsep}{0pt}\def\cbRGB{\colorbox[RGB]}\expandafter\cbRGB\expandafter{\detokenize{183,183,255}}{maze\strut} \setlength{\fboxsep}{0pt}\def\cbRGB{\colorbox[RGB]}\expandafter\cbRGB\expandafter{\detokenize{102,102,255}}{of\strut} \setlength{\fboxsep}{0pt}\def\cbRGB{\colorbox[RGB]}\expandafter\cbRGB\expandafter{\detokenize{102,102,255}}{metro\strut} \setlength{\fboxsep}{0pt}\def\cbRGB{\colorbox[RGB]}\expandafter\cbRGB\expandafter{\detokenize{104,104,255}}{stops\strut} \setlength{\fboxsep}{0pt}\def\cbRGB{\colorbox[RGB]}\expandafter\cbRGB\expandafter{\detokenize{179,179,255}}{,\strut} \setlength{\fboxsep}{0pt}\def\cbRGB{\colorbox[RGB]}\expandafter\cbRGB\expandafter{\detokenize{225,225,255}}{and\strut} \setlength{\fboxsep}{0pt}\def\cbRGB{\colorbox[RGB]}\expandafter\cbRGB\expandafter{\detokenize{236,236,255}}{delivered\strut} \setlength{\fboxsep}{0pt}\def\cbRGB{\colorbox[RGB]}\expandafter\cbRGB\expandafter{\detokenize{239,239,255}}{us\strut} \setlength{\fboxsep}{0pt}\def\cbRGB{\colorbox[RGB]}\expandafter\cbRGB\expandafter{\detokenize{238,238,255}}{right\strut} \setlength{\fboxsep}{0pt}\def\cbRGB{\colorbox[RGB]}\expandafter\cbRGB\expandafter{\detokenize{238,238,255}}{at\strut} \setlength{\fboxsep}{0pt}\def\cbRGB{\colorbox[RGB]}\expandafter\cbRGB\expandafter{\detokenize{236,236,255}}{the\strut} \setlength{\fboxsep}{0pt}\def\cbRGB{\colorbox[RGB]}\expandafter\cbRGB\expandafter{\detokenize{240,240,255}}{gate\strut} \setlength{\fboxsep}{0pt}\def\cbRGB{\colorbox[RGB]}\expandafter\cbRGB\expandafter{\detokenize{241,241,255}}{!\strut} \setlength{\fboxsep}{0pt}\def\cbRGB{\colorbox[RGB]}\expandafter\cbRGB\expandafter{\detokenize{240,240,255}}{unk\strut} \setlength{\fboxsep}{0pt}\def\cbRGB{\colorbox[RGB]}\expandafter\cbRGB\expandafter{\detokenize{237,237,255}}{was\strut} \setlength{\fboxsep}{0pt}\def\cbRGB{\colorbox[RGB]}\expandafter\cbRGB\expandafter{\detokenize{237,237,255}}{just\strut} \setlength{\fboxsep}{0pt}\def\cbRGB{\colorbox[RGB]}\expandafter\cbRGB\expandafter{\detokenize{238,238,255}}{so\strut} \setlength{\fboxsep}{0pt}\def\cbRGB{\colorbox[RGB]}\expandafter\cbRGB\expandafter{\detokenize{241,241,255}}{nice\strut} \setlength{\fboxsep}{0pt}\def\cbRGB{\colorbox[RGB]}\expandafter\cbRGB\expandafter{\detokenize{241,241,255}}{!\strut} \setlength{\fboxsep}{0pt}\def\cbRGB{\colorbox[RGB]}\expandafter\cbRGB\expandafter{\detokenize{241,241,255}}{i\strut} \setlength{\fboxsep}{0pt}\def\cbRGB{\colorbox[RGB]}\expandafter\cbRGB\expandafter{\detokenize{239,239,255}}{highly\strut} \setlength{\fboxsep}{0pt}\def\cbRGB{\colorbox[RGB]}\expandafter\cbRGB\expandafter{\detokenize{243,243,255}}{recommend\strut} \setlength{\fboxsep}{0pt}\def\cbRGB{\colorbox[RGB]}\expandafter\cbRGB\expandafter{\detokenize{242,242,255}}{this\strut} \setlength{\fboxsep}{0pt}\def\cbRGB{\colorbox[RGB]}\expandafter\cbRGB\expandafter{\detokenize{107,107,255}}{small\strut} \setlength{\fboxsep}{0pt}\def\cbRGB{\colorbox[RGB]}\expandafter\cbRGB\expandafter{\detokenize{98,98,255}}{hotel\strut} \setlength{\fboxsep}{0pt}\def\cbRGB{\colorbox[RGB]}\expandafter\cbRGB\expandafter{\detokenize{57,57,255}}{and\strut} \setlength{\fboxsep}{0pt}\def\cbRGB{\colorbox[RGB]}\expandafter\cbRGB\expandafter{\detokenize{166,166,255}}{would\strut} \setlength{\fboxsep}{0pt}\def\cbRGB{\colorbox[RGB]}\expandafter\cbRGB\expandafter{\detokenize{19,19,255}}{absolutely\strut} \setlength{\fboxsep}{0pt}\def\cbRGB{\colorbox[RGB]}\expandafter\cbRGB\expandafter{\detokenize{38,38,255}}{stay\strut} \setlength{\fboxsep}{0pt}\def\cbRGB{\colorbox[RGB]}\expandafter\cbRGB\expandafter{\detokenize{39,39,255}}{there\strut} \setlength{\fboxsep}{0pt}\def\cbRGB{\colorbox[RGB]}\expandafter\cbRGB\expandafter{\detokenize{190,190,255}}{again\strut} \setlength{\fboxsep}{0pt}\def\cbRGB{\colorbox[RGB]}\expandafter\cbRGB\expandafter{\detokenize{212,212,255}}{if\strut} \setlength{\fboxsep}{0pt}\def\cbRGB{\colorbox[RGB]}\expandafter\cbRGB\expandafter{\detokenize{238,238,255}}{we\strut} \setlength{\fboxsep}{0pt}\def\cbRGB{\colorbox[RGB]}\expandafter\cbRGB\expandafter{\detokenize{238,238,255}}{are\strut} \setlength{\fboxsep}{0pt}\def\cbRGB{\colorbox[RGB]}\expandafter\cbRGB\expandafter{\detokenize{238,238,255}}{lucky\strut} \setlength{\fboxsep}{0pt}\def\cbRGB{\colorbox[RGB]}\expandafter\cbRGB\expandafter{\detokenize{240,240,255}}{enough\strut} \setlength{\fboxsep}{0pt}\def\cbRGB{\colorbox[RGB]}\expandafter\cbRGB\expandafter{\detokenize{239,239,255}}{to\strut} \setlength{\fboxsep}{0pt}\def\cbRGB{\colorbox[RGB]}\expandafter\cbRGB\expandafter{\detokenize{242,242,255}}{return\strut} \setlength{\fboxsep}{0pt}\def\cbRGB{\colorbox[RGB]}\expandafter\cbRGB\expandafter{\detokenize{240,240,255}}{!\strut} \setlength{\fboxsep}{0pt}\def\cbRGB{\colorbox[RGB]}\expandafter\cbRGB\expandafter{\detokenize{243,243,255}}{they\strut} \setlength{\fboxsep}{0pt}\def\cbRGB{\colorbox[RGB]}\expandafter\cbRGB\expandafter{\detokenize{239,239,255}}{deserve\strut} \setlength{\fboxsep}{0pt}\def\cbRGB{\colorbox[RGB]}\expandafter\cbRGB\expandafter{\detokenize{237,237,255}}{their\strut} \setlength{\fboxsep}{0pt}\def\cbRGB{\colorbox[RGB]}\expandafter\cbRGB\expandafter{\detokenize{242,242,255}}{\#\strut} \setlength{\fboxsep}{0pt}\def\cbRGB{\colorbox[RGB]}\expandafter\cbRGB\expandafter{\detokenize{243,243,255}}{qqq\strut} \setlength{\fboxsep}{0pt}\def\cbRGB{\colorbox[RGB]}\expandafter\cbRGB\expandafter{\detokenize{243,243,255}}{tripadvisor\strut} \setlength{\fboxsep}{0pt}\def\cbRGB{\colorbox[RGB]}\expandafter\cbRGB\expandafter{\detokenize{240,240,255}}{rating\strut} \setlength{\fboxsep}{0pt}\def\cbRGB{\colorbox[RGB]}\expandafter\cbRGB\expandafter{\detokenize{239,239,255}}{!\strut} \setlength{\fboxsep}{0pt}\def\cbRGB{\colorbox[RGB]}\expandafter\cbRGB\expandafter{\detokenize{240,240,255}}{thanks\strut} \setlength{\fboxsep}{0pt}\def\cbRGB{\colorbox[RGB]}\expandafter\cbRGB\expandafter{\detokenize{238,238,255}}{you\strut} \setlength{\fboxsep}{0pt}\def\cbRGB{\colorbox[RGB]}\expandafter\cbRGB\expandafter{\detokenize{239,239,255}}{,\strut} \setlength{\fboxsep}{0pt}\def\cbRGB{\colorbox[RGB]}\expandafter\cbRGB\expandafter{\detokenize{239,239,255}}{unk\strut} \setlength{\fboxsep}{0pt}\def\cbRGB{\colorbox[RGB]}\expandafter\cbRGB\expandafter{\detokenize{239,239,255}}{unk\strut} \setlength{\fboxsep}{0pt}\def\cbRGB{\colorbox[RGB]}\expandafter\cbRGB\expandafter{\detokenize{239,239,255}}{,\strut} \setlength{\fboxsep}{0pt}\def\cbRGB{\colorbox[RGB]}\expandafter\cbRGB\expandafter{\detokenize{240,240,255}}{for\strut} \setlength{\fboxsep}{0pt}\def\cbRGB{\colorbox[RGB]}\expandafter\cbRGB\expandafter{\detokenize{237,237,255}}{your\strut} \setlength{\fboxsep}{0pt}\def\cbRGB{\colorbox[RGB]}\expandafter\cbRGB\expandafter{\detokenize{236,236,255}}{part\strut} \setlength{\fboxsep}{0pt}\def\cbRGB{\colorbox[RGB]}\expandafter\cbRGB\expandafter{\detokenize{239,239,255}}{in\strut} \setlength{\fboxsep}{0pt}\def\cbRGB{\colorbox[RGB]}\expandafter\cbRGB\expandafter{\detokenize{243,243,255}}{making\strut} \setlength{\fboxsep}{0pt}\def\cbRGB{\colorbox[RGB]}\expandafter\cbRGB\expandafter{\detokenize{243,243,255}}{this\strut} \setlength{\fboxsep}{0pt}\def\cbRGB{\colorbox[RGB]}\expandafter\cbRGB\expandafter{\detokenize{240,240,255}}{trip\strut} \setlength{\fboxsep}{0pt}\def\cbRGB{\colorbox[RGB]}\expandafter\cbRGB\expandafter{\detokenize{238,238,255}}{such\strut} \setlength{\fboxsep}{0pt}\def\cbRGB{\colorbox[RGB]}\expandafter\cbRGB\expandafter{\detokenize{238,238,255}}{a\strut} \setlength{\fboxsep}{0pt}\def\cbRGB{\colorbox[RGB]}\expandafter\cbRGB\expandafter{\detokenize{242,242,255}}{pleasure\strut} \setlength{\fboxsep}{0pt}\def\cbRGB{\colorbox[RGB]}\expandafter\cbRGB\expandafter{\detokenize{233,233,255}}{!\strut} 

\par
\textbf{Example 3}

\setlength{\fboxsep}{0pt}\def\cbRGB{\colorbox[RGB]}\expandafter\cbRGB\expandafter{\detokenize{255,222,222}}{oh\strut} \setlength{\fboxsep}{0pt}\def\cbRGB{\colorbox[RGB]}\expandafter\cbRGB\expandafter{\detokenize{255,226,226}}{,\strut} \setlength{\fboxsep}{0pt}\def\cbRGB{\colorbox[RGB]}\expandafter\cbRGB\expandafter{\detokenize{255,229,229}}{unk\strut} \setlength{\fboxsep}{0pt}\def\cbRGB{\colorbox[RGB]}\expandafter\cbRGB\expandafter{\detokenize{255,226,226}}{.\strut} \setlength{\fboxsep}{0pt}\def\cbRGB{\colorbox[RGB]}\expandafter\cbRGB\expandafter{\detokenize{255,229,229}}{true\strut} \setlength{\fboxsep}{0pt}\def\cbRGB{\colorbox[RGB]}\expandafter\cbRGB\expandafter{\detokenize{255,225,225}}{to\strut} \setlength{\fboxsep}{0pt}\def\cbRGB{\colorbox[RGB]}\expandafter\cbRGB\expandafter{\detokenize{255,228,228}}{name\strut} \setlength{\fboxsep}{0pt}\def\cbRGB{\colorbox[RGB]}\expandafter\cbRGB\expandafter{\detokenize{255,225,225}}{,\strut} \setlength{\fboxsep}{0pt}\def\cbRGB{\colorbox[RGB]}\expandafter\cbRGB\expandafter{\detokenize{255,225,225}}{you\strut} \setlength{\fboxsep}{0pt}\def\cbRGB{\colorbox[RGB]}\expandafter\cbRGB\expandafter{\detokenize{255,225,225}}{'re\strut} \setlength{\fboxsep}{0pt}\def\cbRGB{\colorbox[RGB]}\expandafter\cbRGB\expandafter{\detokenize{255,231,231}}{like\strut} \setlength{\fboxsep}{0pt}\def\cbRGB{\colorbox[RGB]}\expandafter\cbRGB\expandafter{\detokenize{255,135,135}}{a\strut} \setlength{\fboxsep}{0pt}\def\cbRGB{\colorbox[RGB]}\expandafter\cbRGB\expandafter{\detokenize{255,66,66}}{guilty\strut} \setlength{\fboxsep}{0pt}\def\cbRGB{\colorbox[RGB]}\expandafter\cbRGB\expandafter{\detokenize{255,61,61}}{indulgence\strut} \setlength{\fboxsep}{0pt}\def\cbRGB{\colorbox[RGB]}\expandafter\cbRGB\expandafter{\detokenize{255,152,152}}{that\strut} \setlength{\fboxsep}{0pt}\def\cbRGB{\colorbox[RGB]}\expandafter\cbRGB\expandafter{\detokenize{255,217,217}}{i\strut} \setlength{\fboxsep}{0pt}\def\cbRGB{\colorbox[RGB]}\expandafter\cbRGB\expandafter{\detokenize{255,216,216}}{want\strut} \setlength{\fboxsep}{0pt}\def\cbRGB{\colorbox[RGB]}\expandafter\cbRGB\expandafter{\detokenize{255,223,223}}{more\strut} \setlength{\fboxsep}{0pt}\def\cbRGB{\colorbox[RGB]}\expandafter\cbRGB\expandafter{\detokenize{255,227,227}}{of\strut} \setlength{\fboxsep}{0pt}\def\cbRGB{\colorbox[RGB]}\expandafter\cbRGB\expandafter{\detokenize{255,227,227}}{.\strut} \setlength{\fboxsep}{0pt}\def\cbRGB{\colorbox[RGB]}\expandafter\cbRGB\expandafter{\detokenize{255,228,228}}{

\strut} \setlength{\fboxsep}{0pt}\def\cbRGB{\colorbox[RGB]}\expandafter\cbRGB\expandafter{\detokenize{255,224,224}}{while\strut} \setlength{\fboxsep}{0pt}\def\cbRGB{\colorbox[RGB]}\expandafter\cbRGB\expandafter{\detokenize{255,169,169}}{visiting\strut} \setlength{\fboxsep}{0pt}\def\cbRGB{\colorbox[RGB]}\expandafter\cbRGB\expandafter{\detokenize{255,170,170}}{cleveland\strut} \setlength{\fboxsep}{0pt}\def\cbRGB{\colorbox[RGB]}\expandafter\cbRGB\expandafter{\detokenize{255,171,171}}{,\strut} \setlength{\fboxsep}{0pt}\def\cbRGB{\colorbox[RGB]}\expandafter\cbRGB\expandafter{\detokenize{255,225,225}}{i\strut} \setlength{\fboxsep}{0pt}\def\cbRGB{\colorbox[RGB]}\expandafter\cbRGB\expandafter{\detokenize{255,226,226}}{went\strut} \setlength{\fboxsep}{0pt}\def\cbRGB{\colorbox[RGB]}\expandafter\cbRGB\expandafter{\detokenize{255,226,226}}{to\strut} \setlength{\fboxsep}{0pt}\def\cbRGB{\colorbox[RGB]}\expandafter\cbRGB\expandafter{\detokenize{255,226,226}}{unk\strut} \setlength{\fboxsep}{0pt}\def\cbRGB{\colorbox[RGB]}\expandafter\cbRGB\expandafter{\detokenize{255,225,225}}{for\strut} \setlength{\fboxsep}{0pt}\def\cbRGB{\colorbox[RGB]}\expandafter\cbRGB\expandafter{\detokenize{255,224,224}}{a\strut} \setlength{\fboxsep}{0pt}\def\cbRGB{\colorbox[RGB]}\expandafter\cbRGB\expandafter{\detokenize{255,225,225}}{late\strut} \setlength{\fboxsep}{0pt}\def\cbRGB{\colorbox[RGB]}\expandafter\cbRGB\expandafter{\detokenize{255,223,223}}{dinner\strut} \setlength{\fboxsep}{0pt}\def\cbRGB{\colorbox[RGB]}\expandafter\cbRGB\expandafter{\detokenize{255,228,228}}{around\strut} \setlength{\fboxsep}{0pt}\def\cbRGB{\colorbox[RGB]}\expandafter\cbRGB\expandafter{\detokenize{255,227,227}}{qqq\strut} \setlength{\fboxsep}{0pt}\def\cbRGB{\colorbox[RGB]}\expandafter\cbRGB\expandafter{\detokenize{255,230,230}}{on\strut} \setlength{\fboxsep}{0pt}\def\cbRGB{\colorbox[RGB]}\expandafter\cbRGB\expandafter{\detokenize{255,225,225}}{a\strut} \setlength{\fboxsep}{0pt}\def\cbRGB{\colorbox[RGB]}\expandafter\cbRGB\expandafter{\detokenize{255,229,229}}{thursday\strut} \setlength{\fboxsep}{0pt}\def\cbRGB{\colorbox[RGB]}\expandafter\cbRGB\expandafter{\detokenize{255,227,227}}{.\strut} \setlength{\fboxsep}{0pt}\def\cbRGB{\colorbox[RGB]}\expandafter\cbRGB\expandafter{\detokenize{255,227,227}}{i\strut} \setlength{\fboxsep}{0pt}\def\cbRGB{\colorbox[RGB]}\expandafter\cbRGB\expandafter{\detokenize{255,223,223}}{sat\strut} \setlength{\fboxsep}{0pt}\def\cbRGB{\colorbox[RGB]}\expandafter\cbRGB\expandafter{\detokenize{255,223,223}}{at\strut} \setlength{\fboxsep}{0pt}\def\cbRGB{\colorbox[RGB]}\expandafter\cbRGB\expandafter{\detokenize{255,222,222}}{the\strut} \setlength{\fboxsep}{0pt}\def\cbRGB{\colorbox[RGB]}\expandafter\cbRGB\expandafter{\detokenize{255,225,225}}{bar\strut} \setlength{\fboxsep}{0pt}\def\cbRGB{\colorbox[RGB]}\expandafter\cbRGB\expandafter{\detokenize{255,227,227}}{,\strut} \setlength{\fboxsep}{0pt}\def\cbRGB{\colorbox[RGB]}\expandafter\cbRGB\expandafter{\detokenize{255,227,227}}{and\strut} \setlength{\fboxsep}{0pt}\def\cbRGB{\colorbox[RGB]}\expandafter\cbRGB\expandafter{\detokenize{255,223,223}}{there\strut} \setlength{\fboxsep}{0pt}\def\cbRGB{\colorbox[RGB]}\expandafter\cbRGB\expandafter{\detokenize{255,221,221}}{was\strut} \setlength{\fboxsep}{0pt}\def\cbRGB{\colorbox[RGB]}\expandafter\cbRGB\expandafter{\detokenize{255,225,225}}{quite\strut} \setlength{\fboxsep}{0pt}\def\cbRGB{\colorbox[RGB]}\expandafter\cbRGB\expandafter{\detokenize{255,223,223}}{a\strut} \setlength{\fboxsep}{0pt}\def\cbRGB{\colorbox[RGB]}\expandafter\cbRGB\expandafter{\detokenize{255,224,224}}{full\strut} \setlength{\fboxsep}{0pt}\def\cbRGB{\colorbox[RGB]}\expandafter\cbRGB\expandafter{\detokenize{255,221,221}}{crowd\strut} \setlength{\fboxsep}{0pt}\def\cbRGB{\colorbox[RGB]}\expandafter\cbRGB\expandafter{\detokenize{255,223,223}}{because\strut} \setlength{\fboxsep}{0pt}\def\cbRGB{\colorbox[RGB]}\expandafter\cbRGB\expandafter{\detokenize{255,222,222}}{they\strut} \setlength{\fboxsep}{0pt}\def\cbRGB{\colorbox[RGB]}\expandafter\cbRGB\expandafter{\detokenize{255,224,224}}{have\strut} \setlength{\fboxsep}{0pt}\def\cbRGB{\colorbox[RGB]}\expandafter\cbRGB\expandafter{\detokenize{255,224,224}}{a\strut} \setlength{\fboxsep}{0pt}\def\cbRGB{\colorbox[RGB]}\expandafter\cbRGB\expandafter{\detokenize{255,230,230}}{second\strut} \setlength{\fboxsep}{0pt}\def\cbRGB{\colorbox[RGB]}\expandafter\cbRGB\expandafter{\detokenize{255,229,229}}{,\strut} \setlength{\fboxsep}{0pt}\def\cbRGB{\colorbox[RGB]}\expandafter\cbRGB\expandafter{\detokenize{255,191,191}}{late\strut} \setlength{\fboxsep}{0pt}\def\cbRGB{\colorbox[RGB]}\expandafter\cbRGB\expandafter{\detokenize{255,187,187}}{happy\strut} \setlength{\fboxsep}{0pt}\def\cbRGB{\colorbox[RGB]}\expandafter\cbRGB\expandafter{\detokenize{255,188,188}}{hour\strut} \setlength{\fboxsep}{0pt}\def\cbRGB{\colorbox[RGB]}\expandafter\cbRGB\expandafter{\detokenize{255,234,234}}{from\strut} \setlength{\fboxsep}{0pt}\def\cbRGB{\colorbox[RGB]}\expandafter\cbRGB\expandafter{\detokenize{255,233,233}}{qqq\strut} \setlength{\fboxsep}{0pt}\def\cbRGB{\colorbox[RGB]}\expandafter\cbRGB\expandafter{\detokenize{255,184,184}}{with\strut} \setlength{\fboxsep}{0pt}\def\cbRGB{\colorbox[RGB]}\expandafter\cbRGB\expandafter{\detokenize{255,177,177}}{great\strut} \setlength{\fboxsep}{0pt}\def\cbRGB{\colorbox[RGB]}\expandafter\cbRGB\expandafter{\detokenize{255,177,177}}{deals\strut} \setlength{\fboxsep}{0pt}\def\cbRGB{\colorbox[RGB]}\expandafter\cbRGB\expandafter{\detokenize{255,225,225}}{on\strut} \setlength{\fboxsep}{0pt}\def\cbRGB{\colorbox[RGB]}\expandafter\cbRGB\expandafter{\detokenize{255,226,226}}{drinks\strut} \setlength{\fboxsep}{0pt}\def\cbRGB{\colorbox[RGB]}\expandafter\cbRGB\expandafter{\detokenize{255,224,224}}{and\strut} \setlength{\fboxsep}{0pt}\def\cbRGB{\colorbox[RGB]}\expandafter\cbRGB\expandafter{\detokenize{255,224,224}}{small\strut} \setlength{\fboxsep}{0pt}\def\cbRGB{\colorbox[RGB]}\expandafter\cbRGB\expandafter{\detokenize{255,225,225}}{plates\strut} \setlength{\fboxsep}{0pt}\def\cbRGB{\colorbox[RGB]}\expandafter\cbRGB\expandafter{\detokenize{255,229,229}}{.\strut} \setlength{\fboxsep}{0pt}\def\cbRGB{\colorbox[RGB]}\expandafter\cbRGB\expandafter{\detokenize{255,231,231}}{

\strut} \setlength{\fboxsep}{0pt}\def\cbRGB{\colorbox[RGB]}\expandafter\cbRGB\expandafter{\detokenize{255,212,212}}{i\strut} \setlength{\fboxsep}{0pt}\def\cbRGB{\colorbox[RGB]}\expandafter\cbRGB\expandafter{\detokenize{255,211,211}}{ordered\strut} \setlength{\fboxsep}{0pt}\def\cbRGB{\colorbox[RGB]}\expandafter\cbRGB\expandafter{\detokenize{255,212,212}}{the\strut} \setlength{\fboxsep}{0pt}\def\cbRGB{\colorbox[RGB]}\expandafter\cbRGB\expandafter{\detokenize{255,226,226}}{house\strut} \setlength{\fboxsep}{0pt}\def\cbRGB{\colorbox[RGB]}\expandafter\cbRGB\expandafter{\detokenize{255,229,229}}{red\strut} \setlength{\fboxsep}{0pt}\def\cbRGB{\colorbox[RGB]}\expandafter\cbRGB\expandafter{\detokenize{255,233,233}}{wine\strut} \setlength{\fboxsep}{0pt}\def\cbRGB{\colorbox[RGB]}\expandafter\cbRGB\expandafter{\detokenize{255,221,221}}{(\strut} \setlength{\fboxsep}{0pt}\def\cbRGB{\colorbox[RGB]}\expandafter\cbRGB\expandafter{\detokenize{255,231,231}}{\$\strut} \setlength{\fboxsep}{0pt}\def\cbRGB{\colorbox[RGB]}\expandafter\cbRGB\expandafter{\detokenize{255,234,234}}{qqq\strut} \setlength{\fboxsep}{0pt}\def\cbRGB{\colorbox[RGB]}\expandafter\cbRGB\expandafter{\detokenize{255,208,208}}{at\strut} \setlength{\fboxsep}{0pt}\def\cbRGB{\colorbox[RGB]}\expandafter\cbRGB\expandafter{\detokenize{255,193,193}}{happy\strut} \setlength{\fboxsep}{0pt}\def\cbRGB{\colorbox[RGB]}\expandafter\cbRGB\expandafter{\detokenize{255,188,188}}{hour\strut} \setlength{\fboxsep}{0pt}\def\cbRGB{\colorbox[RGB]}\expandafter\cbRGB\expandafter{\detokenize{255,225,225}}{!\strut} \setlength{\fboxsep}{0pt}\def\cbRGB{\colorbox[RGB]}\expandafter\cbRGB\expandafter{\detokenize{255,228,228}}{)\strut} \setlength{\fboxsep}{0pt}\def\cbRGB{\colorbox[RGB]}\expandafter\cbRGB\expandafter{\detokenize{255,230,230}}{,\strut} \setlength{\fboxsep}{0pt}\def\cbRGB{\colorbox[RGB]}\expandafter\cbRGB\expandafter{\detokenize{255,223,223}}{and\strut} \setlength{\fboxsep}{0pt}\def\cbRGB{\colorbox[RGB]}\expandafter\cbRGB\expandafter{\detokenize{255,187,187}}{two\strut} \setlength{\fboxsep}{0pt}\def\cbRGB{\colorbox[RGB]}\expandafter\cbRGB\expandafter{\detokenize{255,195,195}}{suggestions\strut} \setlength{\fboxsep}{0pt}\def\cbRGB{\colorbox[RGB]}\expandafter\cbRGB\expandafter{\detokenize{255,205,205}}{of\strut} \setlength{\fboxsep}{0pt}\def\cbRGB{\colorbox[RGB]}\expandafter\cbRGB\expandafter{\detokenize{255,213,213}}{the\strut} \setlength{\fboxsep}{0pt}\def\cbRGB{\colorbox[RGB]}\expandafter\cbRGB\expandafter{\detokenize{255,105,105}}{bartender\strut} \setlength{\fboxsep}{0pt}\def\cbRGB{\colorbox[RGB]}\expandafter\cbRGB\expandafter{\detokenize{255,105,105}}{:\strut} \setlength{\fboxsep}{0pt}\def\cbRGB{\colorbox[RGB]}\expandafter\cbRGB\expandafter{\detokenize{255,107,107}}{the\strut} \setlength{\fboxsep}{0pt}\def\cbRGB{\colorbox[RGB]}\expandafter\cbRGB\expandafter{\detokenize{255,203,203}}{roasted\strut} \setlength{\fboxsep}{0pt}\def\cbRGB{\colorbox[RGB]}\expandafter\cbRGB\expandafter{\detokenize{255,204,204}}{dates\strut} \setlength{\fboxsep}{0pt}\def\cbRGB{\colorbox[RGB]}\expandafter\cbRGB\expandafter{\detokenize{255,225,225}}{to\strut} \setlength{\fboxsep}{0pt}\def\cbRGB{\colorbox[RGB]}\expandafter\cbRGB\expandafter{\detokenize{255,227,227}}{start\strut} \setlength{\fboxsep}{0pt}\def\cbRGB{\colorbox[RGB]}\expandafter\cbRGB\expandafter{\detokenize{255,223,223}}{and\strut} \setlength{\fboxsep}{0pt}\def\cbRGB{\colorbox[RGB]}\expandafter\cbRGB\expandafter{\detokenize{255,223,223}}{the\strut} \setlength{\fboxsep}{0pt}\def\cbRGB{\colorbox[RGB]}\expandafter\cbRGB\expandafter{\detokenize{255,221,221}}{steak\strut} \setlength{\fboxsep}{0pt}\def\cbRGB{\colorbox[RGB]}\expandafter\cbRGB\expandafter{\detokenize{255,222,222}}{entree\strut} \setlength{\fboxsep}{0pt}\def\cbRGB{\colorbox[RGB]}\expandafter\cbRGB\expandafter{\detokenize{255,224,224}}{with\strut} \setlength{\fboxsep}{0pt}\def\cbRGB{\colorbox[RGB]}\expandafter\cbRGB\expandafter{\detokenize{255,223,223}}{a\strut} \setlength{\fboxsep}{0pt}\def\cbRGB{\colorbox[RGB]}\expandafter\cbRGB\expandafter{\detokenize{255,225,225}}{side\strut} \setlength{\fboxsep}{0pt}\def\cbRGB{\colorbox[RGB]}\expandafter\cbRGB\expandafter{\detokenize{255,223,223}}{of\strut} \setlength{\fboxsep}{0pt}\def\cbRGB{\colorbox[RGB]}\expandafter\cbRGB\expandafter{\detokenize{255,224,224}}{brussels\strut} \setlength{\fboxsep}{0pt}\def\cbRGB{\colorbox[RGB]}\expandafter\cbRGB\expandafter{\detokenize{255,226,226}}{sprouts\strut} \setlength{\fboxsep}{0pt}\def\cbRGB{\colorbox[RGB]}\expandafter\cbRGB\expandafter{\detokenize{255,225,225}}{.\strut} \setlength{\fboxsep}{0pt}\def\cbRGB{\colorbox[RGB]}\expandafter\cbRGB\expandafter{\detokenize{255,230,230}}{the\strut} \setlength{\fboxsep}{0pt}\def\cbRGB{\colorbox[RGB]}\expandafter\cbRGB\expandafter{\detokenize{255,232,232}}{dates\strut} \setlength{\fboxsep}{0pt}\def\cbRGB{\colorbox[RGB]}\expandafter\cbRGB\expandafter{\detokenize{255,221,221}}{(\strut} \setlength{\fboxsep}{0pt}\def\cbRGB{\colorbox[RGB]}\expandafter\cbRGB\expandafter{\detokenize{255,226,226}}{\$\strut} \setlength{\fboxsep}{0pt}\def\cbRGB{\colorbox[RGB]}\expandafter\cbRGB\expandafter{\detokenize{255,227,227}}{qqq\strut} \setlength{\fboxsep}{0pt}\def\cbRGB{\colorbox[RGB]}\expandafter\cbRGB\expandafter{\detokenize{255,247,247}}{)\strut} \setlength{\fboxsep}{0pt}\def\cbRGB{\colorbox[RGB]}\expandafter\cbRGB\expandafter{\detokenize{255,228,228}}{were\strut} \setlength{\fboxsep}{0pt}\def\cbRGB{\colorbox[RGB]}\expandafter\cbRGB\expandafter{\detokenize{255,129,129}}{incredibly\strut} \setlength{\fboxsep}{0pt}\def\cbRGB{\colorbox[RGB]}\expandafter\cbRGB\expandafter{\detokenize{255,0,0}}{decadent\strut} \setlength{\fboxsep}{0pt}\def\cbRGB{\colorbox[RGB]}\expandafter\cbRGB\expandafter{\detokenize{255,7,7}}{:\strut} \setlength{\fboxsep}{0pt}\def\cbRGB{\colorbox[RGB]}\expandafter\cbRGB\expandafter{\detokenize{255,78,78}}{roasted\strut} \setlength{\fboxsep}{0pt}\def\cbRGB{\colorbox[RGB]}\expandafter\cbRGB\expandafter{\detokenize{255,199,199}}{and\strut} \setlength{\fboxsep}{0pt}\def\cbRGB{\colorbox[RGB]}\expandafter\cbRGB\expandafter{\detokenize{255,200,200}}{topped\strut} \setlength{\fboxsep}{0pt}\def\cbRGB{\colorbox[RGB]}\expandafter\cbRGB\expandafter{\detokenize{255,220,220}}{with\strut} \setlength{\fboxsep}{0pt}\def\cbRGB{\colorbox[RGB]}\expandafter\cbRGB\expandafter{\detokenize{255,223,223}}{almonds\strut} \setlength{\fboxsep}{0pt}\def\cbRGB{\colorbox[RGB]}\expandafter\cbRGB\expandafter{\detokenize{255,224,224}}{,\strut} \setlength{\fboxsep}{0pt}\def\cbRGB{\colorbox[RGB]}\expandafter\cbRGB\expandafter{\detokenize{255,227,227}}{bacon\strut} \setlength{\fboxsep}{0pt}\def\cbRGB{\colorbox[RGB]}\expandafter\cbRGB\expandafter{\detokenize{255,223,223}}{,\strut} \setlength{\fboxsep}{0pt}\def\cbRGB{\colorbox[RGB]}\expandafter\cbRGB\expandafter{\detokenize{255,224,224}}{unk\strut} \setlength{\fboxsep}{0pt}\def\cbRGB{\colorbox[RGB]}\expandafter\cbRGB\expandafter{\detokenize{255,221,221}}{and\strut} \setlength{\fboxsep}{0pt}\def\cbRGB{\colorbox[RGB]}\expandafter\cbRGB\expandafter{\detokenize{255,226,226}}{parsley\strut} \setlength{\fboxsep}{0pt}\def\cbRGB{\colorbox[RGB]}\expandafter\cbRGB\expandafter{\detokenize{255,224,224}}{,\strut} \setlength{\fboxsep}{0pt}\def\cbRGB{\colorbox[RGB]}\expandafter\cbRGB\expandafter{\detokenize{255,225,225}}{they\strut} \setlength{\fboxsep}{0pt}\def\cbRGB{\colorbox[RGB]}\expandafter\cbRGB\expandafter{\detokenize{255,224,224}}{are\strut} \setlength{\fboxsep}{0pt}\def\cbRGB{\colorbox[RGB]}\expandafter\cbRGB\expandafter{\detokenize{255,225,225}}{a\strut} \setlength{\fboxsep}{0pt}\def\cbRGB{\colorbox[RGB]}\expandafter\cbRGB\expandafter{\detokenize{255,226,226}}{party\strut} \setlength{\fboxsep}{0pt}\def\cbRGB{\colorbox[RGB]}\expandafter\cbRGB\expandafter{\detokenize{255,218,218}}{in\strut} \setlength{\fboxsep}{0pt}\def\cbRGB{\colorbox[RGB]}\expandafter\cbRGB\expandafter{\detokenize{255,218,218}}{your\strut} \setlength{\fboxsep}{0pt}\def\cbRGB{\colorbox[RGB]}\expandafter\cbRGB\expandafter{\detokenize{255,218,218}}{both\strut} \setlength{\fboxsep}{0pt}\def\cbRGB{\colorbox[RGB]}\expandafter\cbRGB\expandafter{\detokenize{255,233,233}}{of\strut} \setlength{\fboxsep}{0pt}\def\cbRGB{\colorbox[RGB]}\expandafter\cbRGB\expandafter{\detokenize{255,243,243}}{sweet\strut} \setlength{\fboxsep}{0pt}\def\cbRGB{\colorbox[RGB]}\expandafter\cbRGB\expandafter{\detokenize{255,181,181}}{and\strut} \setlength{\fboxsep}{0pt}\def\cbRGB{\colorbox[RGB]}\expandafter\cbRGB\expandafter{\detokenize{255,185,185}}{salty\strut} \setlength{\fboxsep}{0pt}\def\cbRGB{\colorbox[RGB]}\expandafter\cbRGB\expandafter{\detokenize{255,187,187}}{and\strut} \setlength{\fboxsep}{0pt}\def\cbRGB{\colorbox[RGB]}\expandafter\cbRGB\expandafter{\detokenize{255,255,255}}{spicy\strut} \setlength{\fboxsep}{0pt}\def\cbRGB{\colorbox[RGB]}\expandafter\cbRGB\expandafter{\detokenize{255,242,242}}{.\strut} \setlength{\fboxsep}{0pt}\def\cbRGB{\colorbox[RGB]}\expandafter\cbRGB\expandafter{\detokenize{255,215,215}}{the\strut} \setlength{\fboxsep}{0pt}\def\cbRGB{\colorbox[RGB]}\expandafter\cbRGB\expandafter{\detokenize{255,218,218}}{sirloin\strut} \setlength{\fboxsep}{0pt}\def\cbRGB{\colorbox[RGB]}\expandafter\cbRGB\expandafter{\detokenize{255,206,206}}{(\strut} \setlength{\fboxsep}{0pt}\def\cbRGB{\colorbox[RGB]}\expandafter\cbRGB\expandafter{\detokenize{255,231,231}}{\$\strut} \setlength{\fboxsep}{0pt}\def\cbRGB{\colorbox[RGB]}\expandafter\cbRGB\expandafter{\detokenize{255,229,229}}{qqq\strut} \setlength{\fboxsep}{0pt}\def\cbRGB{\colorbox[RGB]}\expandafter\cbRGB\expandafter{\detokenize{255,243,243}}{)\strut} \setlength{\fboxsep}{0pt}\def\cbRGB{\colorbox[RGB]}\expandafter\cbRGB\expandafter{\detokenize{255,235,235}}{,\strut} \setlength{\fboxsep}{0pt}\def\cbRGB{\colorbox[RGB]}\expandafter\cbRGB\expandafter{\detokenize{255,239,239}}{which\strut} \setlength{\fboxsep}{0pt}\def\cbRGB{\colorbox[RGB]}\expandafter\cbRGB\expandafter{\detokenize{255,236,236}}{the\strut} \setlength{\fboxsep}{0pt}\def\cbRGB{\colorbox[RGB]}\expandafter\cbRGB\expandafter{\detokenize{255,211,211}}{bartender\strut} \setlength{\fboxsep}{0pt}\def\cbRGB{\colorbox[RGB]}\expandafter\cbRGB\expandafter{\detokenize{255,211,211}}{told\strut} \setlength{\fboxsep}{0pt}\def\cbRGB{\colorbox[RGB]}\expandafter\cbRGB\expandafter{\detokenize{255,220,220}}{me\strut} \setlength{\fboxsep}{0pt}\def\cbRGB{\colorbox[RGB]}\expandafter\cbRGB\expandafter{\detokenize{255,245,245}}{is\strut} \setlength{\fboxsep}{0pt}\def\cbRGB{\colorbox[RGB]}\expandafter\cbRGB\expandafter{\detokenize{255,235,235}}{a\strut} \setlength{\fboxsep}{0pt}\def\cbRGB{\colorbox[RGB]}\expandafter\cbRGB\expandafter{\detokenize{255,225,225}}{new\strut} \setlength{\fboxsep}{0pt}\def\cbRGB{\colorbox[RGB]}\expandafter\cbRGB\expandafter{\detokenize{255,222,222}}{menu\strut} \setlength{\fboxsep}{0pt}\def\cbRGB{\colorbox[RGB]}\expandafter\cbRGB\expandafter{\detokenize{255,229,229}}{item\strut} \setlength{\fboxsep}{0pt}\def\cbRGB{\colorbox[RGB]}\expandafter\cbRGB\expandafter{\detokenize{255,233,233}}{,\strut} \setlength{\fboxsep}{0pt}\def\cbRGB{\colorbox[RGB]}\expandafter\cbRGB\expandafter{\detokenize{255,195,195}}{was\strut} \setlength{\fboxsep}{0pt}\def\cbRGB{\colorbox[RGB]}\expandafter\cbRGB\expandafter{\detokenize{255,175,175}}{beautifully\strut} \setlength{\fboxsep}{0pt}\def\cbRGB{\colorbox[RGB]}\expandafter\cbRGB\expandafter{\detokenize{255,179,179}}{cooked\strut} \setlength{\fboxsep}{0pt}\def\cbRGB{\colorbox[RGB]}\expandafter\cbRGB\expandafter{\detokenize{255,220,220}}{and\strut} \setlength{\fboxsep}{0pt}\def\cbRGB{\colorbox[RGB]}\expandafter\cbRGB\expandafter{\detokenize{255,231,231}}{topped\strut} \setlength{\fboxsep}{0pt}\def\cbRGB{\colorbox[RGB]}\expandafter\cbRGB\expandafter{\detokenize{255,223,223}}{with\strut} \setlength{\fboxsep}{0pt}\def\cbRGB{\colorbox[RGB]}\expandafter\cbRGB\expandafter{\detokenize{255,224,224}}{parmesan\strut} \setlength{\fboxsep}{0pt}\def\cbRGB{\colorbox[RGB]}\expandafter\cbRGB\expandafter{\detokenize{255,224,224}}{,\strut} \setlength{\fboxsep}{0pt}\def\cbRGB{\colorbox[RGB]}\expandafter\cbRGB\expandafter{\detokenize{255,224,224}}{truffle\strut} \setlength{\fboxsep}{0pt}\def\cbRGB{\colorbox[RGB]}\expandafter\cbRGB\expandafter{\detokenize{255,224,224}}{butter\strut} \setlength{\fboxsep}{0pt}\def\cbRGB{\colorbox[RGB]}\expandafter\cbRGB\expandafter{\detokenize{255,224,224}}{,\strut} \setlength{\fboxsep}{0pt}\def\cbRGB{\colorbox[RGB]}\expandafter\cbRGB\expandafter{\detokenize{255,226,226}}{arugula\strut} \setlength{\fboxsep}{0pt}\def\cbRGB{\colorbox[RGB]}\expandafter\cbRGB\expandafter{\detokenize{255,223,223}}{and\strut} \setlength{\fboxsep}{0pt}\def\cbRGB{\colorbox[RGB]}\expandafter\cbRGB\expandafter{\detokenize{255,228,228}}{mushrooms\strut} \setlength{\fboxsep}{0pt}\def\cbRGB{\colorbox[RGB]}\expandafter\cbRGB\expandafter{\detokenize{255,228,228}}{.\strut} \setlength{\fboxsep}{0pt}\def\cbRGB{\colorbox[RGB]}\expandafter\cbRGB\expandafter{\detokenize{255,229,229}}{it\strut} \setlength{\fboxsep}{0pt}\def\cbRGB{\colorbox[RGB]}\expandafter\cbRGB\expandafter{\detokenize{255,201,201}}{was\strut} \setlength{\fboxsep}{0pt}\def\cbRGB{\colorbox[RGB]}\expandafter\cbRGB\expandafter{\detokenize{255,204,204}}{good\strut} \setlength{\fboxsep}{0pt}\def\cbRGB{\colorbox[RGB]}\expandafter\cbRGB\expandafter{\detokenize{255,205,205}}{,\strut} \setlength{\fboxsep}{0pt}\def\cbRGB{\colorbox[RGB]}\expandafter\cbRGB\expandafter{\detokenize{255,228,228}}{although\strut} \setlength{\fboxsep}{0pt}\def\cbRGB{\colorbox[RGB]}\expandafter\cbRGB\expandafter{\detokenize{255,223,223}}{there\strut} \setlength{\fboxsep}{0pt}\def\cbRGB{\colorbox[RGB]}\expandafter\cbRGB\expandafter{\detokenize{255,222,222}}{are\strut} \setlength{\fboxsep}{0pt}\def\cbRGB{\colorbox[RGB]}\expandafter\cbRGB\expandafter{\detokenize{255,223,223}}{other\strut} \setlength{\fboxsep}{0pt}\def\cbRGB{\colorbox[RGB]}\expandafter\cbRGB\expandafter{\detokenize{255,222,222}}{things\strut} \setlength{\fboxsep}{0pt}\def\cbRGB{\colorbox[RGB]}\expandafter\cbRGB\expandafter{\detokenize{255,222,222}}{on\strut} \setlength{\fboxsep}{0pt}\def\cbRGB{\colorbox[RGB]}\expandafter\cbRGB\expandafter{\detokenize{255,223,223}}{the\strut} \setlength{\fboxsep}{0pt}\def\cbRGB{\colorbox[RGB]}\expandafter\cbRGB\expandafter{\detokenize{255,222,222}}{menu\strut} \setlength{\fboxsep}{0pt}\def\cbRGB{\colorbox[RGB]}\expandafter\cbRGB\expandafter{\detokenize{255,223,223}}{that\strut} \setlength{\fboxsep}{0pt}\def\cbRGB{\colorbox[RGB]}\expandafter\cbRGB\expandafter{\detokenize{255,224,224}}{might\strut} \setlength{\fboxsep}{0pt}\def\cbRGB{\colorbox[RGB]}\expandafter\cbRGB\expandafter{\detokenize{255,224,224}}{be\strut} \setlength{\fboxsep}{0pt}\def\cbRGB{\colorbox[RGB]}\expandafter\cbRGB\expandafter{\detokenize{255,205,205}}{more\strut} \setlength{\fboxsep}{0pt}\def\cbRGB{\colorbox[RGB]}\expandafter\cbRGB\expandafter{\detokenize{255,205,205}}{exciting\strut} \setlength{\fboxsep}{0pt}\def\cbRGB{\colorbox[RGB]}\expandafter\cbRGB\expandafter{\detokenize{255,203,203}}{.\strut} \setlength{\fboxsep}{0pt}\def\cbRGB{\colorbox[RGB]}\expandafter\cbRGB\expandafter{\detokenize{255,222,222}}{the\strut} \setlength{\fboxsep}{0pt}\def\cbRGB{\colorbox[RGB]}\expandafter\cbRGB\expandafter{\detokenize{255,219,219}}{stand\strut} \setlength{\fboxsep}{0pt}\def\cbRGB{\colorbox[RGB]}\expandafter\cbRGB\expandafter{\detokenize{255,222,222}}{out\strut} \setlength{\fboxsep}{0pt}\def\cbRGB{\colorbox[RGB]}\expandafter\cbRGB\expandafter{\detokenize{255,224,224}}{dish\strut} \setlength{\fboxsep}{0pt}\def\cbRGB{\colorbox[RGB]}\expandafter\cbRGB\expandafter{\detokenize{255,224,224}}{of\strut} \setlength{\fboxsep}{0pt}\def\cbRGB{\colorbox[RGB]}\expandafter\cbRGB\expandafter{\detokenize{255,223,223}}{the\strut} \setlength{\fboxsep}{0pt}\def\cbRGB{\colorbox[RGB]}\expandafter\cbRGB\expandafter{\detokenize{255,221,221}}{evening\strut} \setlength{\fboxsep}{0pt}\def\cbRGB{\colorbox[RGB]}\expandafter\cbRGB\expandafter{\detokenize{255,222,222}}{was\strut} \setlength{\fboxsep}{0pt}\def\cbRGB{\colorbox[RGB]}\expandafter\cbRGB\expandafter{\detokenize{255,223,223}}{the\strut} \setlength{\fboxsep}{0pt}\def\cbRGB{\colorbox[RGB]}\expandafter\cbRGB\expandafter{\detokenize{255,223,223}}{fried\strut} \setlength{\fboxsep}{0pt}\def\cbRGB{\colorbox[RGB]}\expandafter\cbRGB\expandafter{\detokenize{255,223,223}}{brussels\strut} \setlength{\fboxsep}{0pt}\def\cbRGB{\colorbox[RGB]}\expandafter\cbRGB\expandafter{\detokenize{255,226,226}}{sprouts\strut} \setlength{\fboxsep}{0pt}\def\cbRGB{\colorbox[RGB]}\expandafter\cbRGB\expandafter{\detokenize{255,226,226}}{,\strut} \setlength{\fboxsep}{0pt}\def\cbRGB{\colorbox[RGB]}\expandafter\cbRGB\expandafter{\detokenize{255,225,225}}{which\strut} \setlength{\fboxsep}{0pt}\def\cbRGB{\colorbox[RGB]}\expandafter\cbRGB\expandafter{\detokenize{255,221,221}}{i\strut} \setlength{\fboxsep}{0pt}\def\cbRGB{\colorbox[RGB]}\expandafter\cbRGB\expandafter{\detokenize{255,222,222}}{would\strut} \setlength{\fboxsep}{0pt}\def\cbRGB{\colorbox[RGB]}\expandafter\cbRGB\expandafter{\detokenize{255,223,223}}{very\strut} \setlength{\fboxsep}{0pt}\def\cbRGB{\colorbox[RGB]}\expandafter\cbRGB\expandafter{\detokenize{255,223,223}}{much\strut} \setlength{\fboxsep}{0pt}\def\cbRGB{\colorbox[RGB]}\expandafter\cbRGB\expandafter{\detokenize{255,220,220}}{like\strut} \setlength{\fboxsep}{0pt}\def\cbRGB{\colorbox[RGB]}\expandafter\cbRGB\expandafter{\detokenize{255,198,198}}{the\strut} \setlength{\fboxsep}{0pt}\def\cbRGB{\colorbox[RGB]}\expandafter\cbRGB\expandafter{\detokenize{255,199,199}}{recipe\strut} \setlength{\fboxsep}{0pt}\def\cbRGB{\colorbox[RGB]}\expandafter\cbRGB\expandafter{\detokenize{255,205,205}}{for\strut} \setlength{\fboxsep}{0pt}\def\cbRGB{\colorbox[RGB]}\expandafter\cbRGB\expandafter{\detokenize{255,228,228}}{so\strut} \setlength{\fboxsep}{0pt}\def\cbRGB{\colorbox[RGB]}\expandafter\cbRGB\expandafter{\detokenize{255,212,212}}{i\strut} \setlength{\fboxsep}{0pt}\def\cbRGB{\colorbox[RGB]}\expandafter\cbRGB\expandafter{\detokenize{255,209,209}}{can\strut} \setlength{\fboxsep}{0pt}\def\cbRGB{\colorbox[RGB]}\expandafter\cbRGB\expandafter{\detokenize{255,213,213}}{unk\strut} \setlength{\fboxsep}{0pt}\def\cbRGB{\colorbox[RGB]}\expandafter\cbRGB\expandafter{\detokenize{255,231,231}}{myself\strut} \setlength{\fboxsep}{0pt}\def\cbRGB{\colorbox[RGB]}\expandafter\cbRGB\expandafter{\detokenize{255,224,224}}{on\strut} \setlength{\fboxsep}{0pt}\def\cbRGB{\colorbox[RGB]}\expandafter\cbRGB\expandafter{\detokenize{255,223,223}}{them\strut} \setlength{\fboxsep}{0pt}\def\cbRGB{\colorbox[RGB]}\expandafter\cbRGB\expandafter{\detokenize{255,226,226}}{every\strut} \setlength{\fboxsep}{0pt}\def\cbRGB{\colorbox[RGB]}\expandafter\cbRGB\expandafter{\detokenize{255,237,237}}{night\strut} \setlength{\fboxsep}{0pt}\def\cbRGB{\colorbox[RGB]}\expandafter\cbRGB\expandafter{\detokenize{255,194,194}}{.\strut} \setlength{\fboxsep}{0pt}\def\cbRGB{\colorbox[RGB]}\expandafter\cbRGB\expandafter{\detokenize{255,174,174}}{those\strut} \setlength{\fboxsep}{0pt}\def\cbRGB{\colorbox[RGB]}\expandafter\cbRGB\expandafter{\detokenize{255,179,179}}{babies\strut} \setlength{\fboxsep}{0pt}\def\cbRGB{\colorbox[RGB]}\expandafter\cbRGB\expandafter{\detokenize{255,221,221}}{,\strut} \setlength{\fboxsep}{0pt}\def\cbRGB{\colorbox[RGB]}\expandafter\cbRGB\expandafter{\detokenize{255,244,244}}{topped\strut} \setlength{\fboxsep}{0pt}\def\cbRGB{\colorbox[RGB]}\expandafter\cbRGB\expandafter{\detokenize{255,197,197}}{with\strut} \setlength{\fboxsep}{0pt}\def\cbRGB{\colorbox[RGB]}\expandafter\cbRGB\expandafter{\detokenize{255,160,160}}{crispy\strut} \setlength{\fboxsep}{0pt}\def\cbRGB{\colorbox[RGB]}\expandafter\cbRGB\expandafter{\detokenize{255,164,164}}{crumbled\strut} \setlength{\fboxsep}{0pt}\def\cbRGB{\colorbox[RGB]}\expandafter\cbRGB\expandafter{\detokenize{255,210,210}}{bits\strut} \setlength{\fboxsep}{0pt}\def\cbRGB{\colorbox[RGB]}\expandafter\cbRGB\expandafter{\detokenize{255,242,242}}{of\strut} \setlength{\fboxsep}{0pt}\def\cbRGB{\colorbox[RGB]}\expandafter\cbRGB\expandafter{\detokenize{255,232,232}}{unk\strut} \setlength{\fboxsep}{0pt}\def\cbRGB{\colorbox[RGB]}\expandafter\cbRGB\expandafter{\detokenize{255,223,223}}{,\strut} \setlength{\fboxsep}{0pt}\def\cbRGB{\colorbox[RGB]}\expandafter\cbRGB\expandafter{\detokenize{255,225,225}}{capers\strut} \setlength{\fboxsep}{0pt}\def\cbRGB{\colorbox[RGB]}\expandafter\cbRGB\expandafter{\detokenize{255,222,222}}{and\strut} \setlength{\fboxsep}{0pt}\def\cbRGB{\colorbox[RGB]}\expandafter\cbRGB\expandafter{\detokenize{255,229,229}}{walnuts\strut} \setlength{\fboxsep}{0pt}\def\cbRGB{\colorbox[RGB]}\expandafter\cbRGB\expandafter{\detokenize{255,228,228}}{,\strut} \setlength{\fboxsep}{0pt}\def\cbRGB{\colorbox[RGB]}\expandafter\cbRGB\expandafter{\detokenize{255,169,169}}{were\strut} \setlength{\fboxsep}{0pt}\def\cbRGB{\colorbox[RGB]}\expandafter\cbRGB\expandafter{\detokenize{255,136,136}}{pure\strut} \setlength{\fboxsep}{0pt}\def\cbRGB{\colorbox[RGB]}\expandafter\cbRGB\expandafter{\detokenize{255,142,142}}{bliss\strut} \setlength{\fboxsep}{0pt}\def\cbRGB{\colorbox[RGB]}\expandafter\cbRGB\expandafter{\detokenize{255,204,204}}{.\strut} \setlength{\fboxsep}{0pt}\def\cbRGB{\colorbox[RGB]}\expandafter\cbRGB\expandafter{\detokenize{255,231,231}}{and\strut} \setlength{\fboxsep}{0pt}\def\cbRGB{\colorbox[RGB]}\expandafter\cbRGB\expandafter{\detokenize{255,226,226}}{let\strut} \setlength{\fboxsep}{0pt}\def\cbRGB{\colorbox[RGB]}\expandafter\cbRGB\expandafter{\detokenize{255,221,221}}{'s\strut} \setlength{\fboxsep}{0pt}\def\cbRGB{\colorbox[RGB]}\expandafter\cbRGB\expandafter{\detokenize{255,222,222}}{be\strut} \setlength{\fboxsep}{0pt}\def\cbRGB{\colorbox[RGB]}\expandafter\cbRGB\expandafter{\detokenize{255,229,229}}{honest\strut} \setlength{\fboxsep}{0pt}\def\cbRGB{\colorbox[RGB]}\expandafter\cbRGB\expandafter{\detokenize{255,228,228}}{,\strut} \setlength{\fboxsep}{0pt}\def\cbRGB{\colorbox[RGB]}\expandafter\cbRGB\expandafter{\detokenize{255,183,183}}{how\strut} \setlength{\fboxsep}{0pt}\def\cbRGB{\colorbox[RGB]}\expandafter\cbRGB\expandafter{\detokenize{255,174,174}}{often\strut} \setlength{\fboxsep}{0pt}\def\cbRGB{\colorbox[RGB]}\expandafter\cbRGB\expandafter{\detokenize{255,173,173}}{do\strut} \setlength{\fboxsep}{0pt}\def\cbRGB{\colorbox[RGB]}\expandafter\cbRGB\expandafter{\detokenize{255,218,218}}{you\strut} \setlength{\fboxsep}{0pt}\def\cbRGB{\colorbox[RGB]}\expandafter\cbRGB\expandafter{\detokenize{255,220,220}}{really\strut} \setlength{\fboxsep}{0pt}\def\cbRGB{\colorbox[RGB]}\expandafter\cbRGB\expandafter{\detokenize{255,222,222}}{have\strut} \setlength{\fboxsep}{0pt}\def\cbRGB{\colorbox[RGB]}\expandafter\cbRGB\expandafter{\detokenize{255,226,226}}{the\strut} \setlength{\fboxsep}{0pt}\def\cbRGB{\colorbox[RGB]}\expandafter\cbRGB\expandafter{\detokenize{255,234,234}}{opportunity\strut} \setlength{\fboxsep}{0pt}\def\cbRGB{\colorbox[RGB]}\expandafter\cbRGB\expandafter{\detokenize{255,224,224}}{to\strut} \setlength{\fboxsep}{0pt}\def\cbRGB{\colorbox[RGB]}\expandafter\cbRGB\expandafter{\detokenize{255,225,225}}{call\strut} \setlength{\fboxsep}{0pt}\def\cbRGB{\colorbox[RGB]}\expandafter\cbRGB\expandafter{\detokenize{255,223,223}}{unk\strut} \setlength{\fboxsep}{0pt}\def\cbRGB{\colorbox[RGB]}\expandafter\cbRGB\expandafter{\detokenize{255,233,233}}{sprouts\strut} \setlength{\fboxsep}{0pt}\def\cbRGB{\colorbox[RGB]}\expandafter\cbRGB\expandafter{\detokenize{255,237,237}}{unk\strut} \setlength{\fboxsep}{0pt}\def\cbRGB{\colorbox[RGB]}\expandafter\cbRGB\expandafter{\detokenize{255,239,239}}{?\strut} \setlength{\fboxsep}{0pt}\def\cbRGB{\colorbox[RGB]}\expandafter\cbRGB\expandafter{\detokenize{255,221,221}}{probably\strut} \setlength{\fboxsep}{0pt}\def\cbRGB{\colorbox[RGB]}\expandafter\cbRGB\expandafter{\detokenize{255,224,224}}{not\strut} \setlength{\fboxsep}{0pt}\def\cbRGB{\colorbox[RGB]}\expandafter\cbRGB\expandafter{\detokenize{255,199,199}}{very\strut} \setlength{\fboxsep}{0pt}\def\cbRGB{\colorbox[RGB]}\expandafter\cbRGB\expandafter{\detokenize{255,218,218}}{often\strut} \setlength{\fboxsep}{0pt}\def\cbRGB{\colorbox[RGB]}\expandafter\cbRGB\expandafter{\detokenize{255,205,205}}{.\strut} \setlength{\fboxsep}{0pt}\def\cbRGB{\colorbox[RGB]}\expandafter\cbRGB\expandafter{\detokenize{255,223,223}}{

\strut} \setlength{\fboxsep}{0pt}\def\cbRGB{\colorbox[RGB]}\expandafter\cbRGB\expandafter{\detokenize{255,221,221}}{i\strut} \setlength{\fboxsep}{0pt}\def\cbRGB{\colorbox[RGB]}\expandafter\cbRGB\expandafter{\detokenize{255,222,222}}{unk\strut} \setlength{\fboxsep}{0pt}\def\cbRGB{\colorbox[RGB]}\expandafter\cbRGB\expandafter{\detokenize{255,222,222}}{off\strut} \setlength{\fboxsep}{0pt}\def\cbRGB{\colorbox[RGB]}\expandafter\cbRGB\expandafter{\detokenize{255,222,222}}{the\strut} \setlength{\fboxsep}{0pt}\def\cbRGB{\colorbox[RGB]}\expandafter\cbRGB\expandafter{\detokenize{255,222,222}}{evening\strut} \setlength{\fboxsep}{0pt}\def\cbRGB{\colorbox[RGB]}\expandafter\cbRGB\expandafter{\detokenize{255,224,224}}{with\strut} \setlength{\fboxsep}{0pt}\def\cbRGB{\colorbox[RGB]}\expandafter\cbRGB\expandafter{\detokenize{255,224,224}}{a\strut} \setlength{\fboxsep}{0pt}\def\cbRGB{\colorbox[RGB]}\expandafter\cbRGB\expandafter{\detokenize{255,226,226}}{glass\strut} \setlength{\fboxsep}{0pt}\def\cbRGB{\colorbox[RGB]}\expandafter\cbRGB\expandafter{\detokenize{255,223,223}}{of\strut} \setlength{\fboxsep}{0pt}\def\cbRGB{\colorbox[RGB]}\expandafter\cbRGB\expandafter{\detokenize{255,224,224}}{unk\strut} \setlength{\fboxsep}{0pt}\def\cbRGB{\colorbox[RGB]}\expandafter\cbRGB\expandafter{\detokenize{255,227,227}}{scotch\strut} \setlength{\fboxsep}{0pt}\def\cbRGB{\colorbox[RGB]}\expandafter\cbRGB\expandafter{\detokenize{255,198,198}}{,\strut} \setlength{\fboxsep}{0pt}\def\cbRGB{\colorbox[RGB]}\expandafter\cbRGB\expandafter{\detokenize{255,197,197}}{neat\strut} \setlength{\fboxsep}{0pt}\def\cbRGB{\colorbox[RGB]}\expandafter\cbRGB\expandafter{\detokenize{255,93,93}}{.\strut} \setlength{\fboxsep}{0pt}\def\cbRGB{\colorbox[RGB]}\expandafter\cbRGB\expandafter{\detokenize{255,125,125}}{perfection\strut} \setlength{\fboxsep}{0pt}\def\cbRGB{\colorbox[RGB]}\expandafter\cbRGB\expandafter{\detokenize{255,121,121}}{.\strut} 

\setlength{\fboxsep}{0pt}\def\cbRGB{\colorbox[RGB]}\expandafter\cbRGB\expandafter{\detokenize{132,132,255}}{oh\strut} \setlength{\fboxsep}{0pt}\def\cbRGB{\colorbox[RGB]}\expandafter\cbRGB\expandafter{\detokenize{150,150,255}}{,\strut} \setlength{\fboxsep}{0pt}\def\cbRGB{\colorbox[RGB]}\expandafter\cbRGB\expandafter{\detokenize{157,157,255}}{unk\strut} \setlength{\fboxsep}{0pt}\def\cbRGB{\colorbox[RGB]}\expandafter\cbRGB\expandafter{\detokenize{150,150,255}}{.\strut} \setlength{\fboxsep}{0pt}\def\cbRGB{\colorbox[RGB]}\expandafter\cbRGB\expandafter{\detokenize{156,156,255}}{true\strut} \setlength{\fboxsep}{0pt}\def\cbRGB{\colorbox[RGB]}\expandafter\cbRGB\expandafter{\detokenize{152,152,255}}{to\strut} \setlength{\fboxsep}{0pt}\def\cbRGB{\colorbox[RGB]}\expandafter\cbRGB\expandafter{\detokenize{159,159,255}}{name\strut} \setlength{\fboxsep}{0pt}\def\cbRGB{\colorbox[RGB]}\expandafter\cbRGB\expandafter{\detokenize{153,153,255}}{,\strut} \setlength{\fboxsep}{0pt}\def\cbRGB{\colorbox[RGB]}\expandafter\cbRGB\expandafter{\detokenize{156,156,255}}{you\strut} \setlength{\fboxsep}{0pt}\def\cbRGB{\colorbox[RGB]}\expandafter\cbRGB\expandafter{\detokenize{151,151,255}}{'re\strut} \setlength{\fboxsep}{0pt}\def\cbRGB{\colorbox[RGB]}\expandafter\cbRGB\expandafter{\detokenize{155,155,255}}{like\strut} \setlength{\fboxsep}{0pt}\def\cbRGB{\colorbox[RGB]}\expandafter\cbRGB\expandafter{\detokenize{123,123,255}}{a\strut} \setlength{\fboxsep}{0pt}\def\cbRGB{\colorbox[RGB]}\expandafter\cbRGB\expandafter{\detokenize{120,120,255}}{guilty\strut} \setlength{\fboxsep}{0pt}\def\cbRGB{\colorbox[RGB]}\expandafter\cbRGB\expandafter{\detokenize{117,117,255}}{indulgence\strut} \setlength{\fboxsep}{0pt}\def\cbRGB{\colorbox[RGB]}\expandafter\cbRGB\expandafter{\detokenize{147,147,255}}{that\strut} \setlength{\fboxsep}{0pt}\def\cbRGB{\colorbox[RGB]}\expandafter\cbRGB\expandafter{\detokenize{149,149,255}}{i\strut} \setlength{\fboxsep}{0pt}\def\cbRGB{\colorbox[RGB]}\expandafter\cbRGB\expandafter{\detokenize{148,148,255}}{want\strut} \setlength{\fboxsep}{0pt}\def\cbRGB{\colorbox[RGB]}\expandafter\cbRGB\expandafter{\detokenize{144,144,255}}{more\strut} \setlength{\fboxsep}{0pt}\def\cbRGB{\colorbox[RGB]}\expandafter\cbRGB\expandafter{\detokenize{147,147,255}}{of\strut} \setlength{\fboxsep}{0pt}\def\cbRGB{\colorbox[RGB]}\expandafter\cbRGB\expandafter{\detokenize{158,158,255}}{.\strut} \setlength{\fboxsep}{0pt}\def\cbRGB{\colorbox[RGB]}\expandafter\cbRGB\expandafter{\detokenize{189,189,255}}{

\strut} \setlength{\fboxsep}{0pt}\def\cbRGB{\colorbox[RGB]}\expandafter\cbRGB\expandafter{\detokenize{210,210,255}}{while\strut} \setlength{\fboxsep}{0pt}\def\cbRGB{\colorbox[RGB]}\expandafter\cbRGB\expandafter{\detokenize{35,35,255}}{visiting\strut} \setlength{\fboxsep}{0pt}\def\cbRGB{\colorbox[RGB]}\expandafter\cbRGB\expandafter{\detokenize{38,38,255}}{cleveland\strut} \setlength{\fboxsep}{0pt}\def\cbRGB{\colorbox[RGB]}\expandafter\cbRGB\expandafter{\detokenize{42,42,255}}{,\strut} \setlength{\fboxsep}{0pt}\def\cbRGB{\colorbox[RGB]}\expandafter\cbRGB\expandafter{\detokenize{208,208,255}}{i\strut} \setlength{\fboxsep}{0pt}\def\cbRGB{\colorbox[RGB]}\expandafter\cbRGB\expandafter{\detokenize{181,181,255}}{went\strut} \setlength{\fboxsep}{0pt}\def\cbRGB{\colorbox[RGB]}\expandafter\cbRGB\expandafter{\detokenize{153,153,255}}{to\strut} \setlength{\fboxsep}{0pt}\def\cbRGB{\colorbox[RGB]}\expandafter\cbRGB\expandafter{\detokenize{154,154,255}}{unk\strut} \setlength{\fboxsep}{0pt}\def\cbRGB{\colorbox[RGB]}\expandafter\cbRGB\expandafter{\detokenize{151,151,255}}{for\strut} \setlength{\fboxsep}{0pt}\def\cbRGB{\colorbox[RGB]}\expandafter\cbRGB\expandafter{\detokenize{152,152,255}}{a\strut} \setlength{\fboxsep}{0pt}\def\cbRGB{\colorbox[RGB]}\expandafter\cbRGB\expandafter{\detokenize{151,151,255}}{late\strut} \setlength{\fboxsep}{0pt}\def\cbRGB{\colorbox[RGB]}\expandafter\cbRGB\expandafter{\detokenize{144,144,255}}{dinner\strut} \setlength{\fboxsep}{0pt}\def\cbRGB{\colorbox[RGB]}\expandafter\cbRGB\expandafter{\detokenize{160,160,255}}{around\strut} \setlength{\fboxsep}{0pt}\def\cbRGB{\colorbox[RGB]}\expandafter\cbRGB\expandafter{\detokenize{160,160,255}}{qqq\strut} \setlength{\fboxsep}{0pt}\def\cbRGB{\colorbox[RGB]}\expandafter\cbRGB\expandafter{\detokenize{162,162,255}}{on\strut} \setlength{\fboxsep}{0pt}\def\cbRGB{\colorbox[RGB]}\expandafter\cbRGB\expandafter{\detokenize{143,143,255}}{a\strut} \setlength{\fboxsep}{0pt}\def\cbRGB{\colorbox[RGB]}\expandafter\cbRGB\expandafter{\detokenize{151,151,255}}{thursday\strut} \setlength{\fboxsep}{0pt}\def\cbRGB{\colorbox[RGB]}\expandafter\cbRGB\expandafter{\detokenize{154,154,255}}{.\strut} \setlength{\fboxsep}{0pt}\def\cbRGB{\colorbox[RGB]}\expandafter\cbRGB\expandafter{\detokenize{155,155,255}}{i\strut} \setlength{\fboxsep}{0pt}\def\cbRGB{\colorbox[RGB]}\expandafter\cbRGB\expandafter{\detokenize{150,150,255}}{sat\strut} \setlength{\fboxsep}{0pt}\def\cbRGB{\colorbox[RGB]}\expandafter\cbRGB\expandafter{\detokenize{147,147,255}}{at\strut} \setlength{\fboxsep}{0pt}\def\cbRGB{\colorbox[RGB]}\expandafter\cbRGB\expandafter{\detokenize{146,146,255}}{the\strut} \setlength{\fboxsep}{0pt}\def\cbRGB{\colorbox[RGB]}\expandafter\cbRGB\expandafter{\detokenize{152,152,255}}{bar\strut} \setlength{\fboxsep}{0pt}\def\cbRGB{\colorbox[RGB]}\expandafter\cbRGB\expandafter{\detokenize{154,154,255}}{,\strut} \setlength{\fboxsep}{0pt}\def\cbRGB{\colorbox[RGB]}\expandafter\cbRGB\expandafter{\detokenize{153,153,255}}{and\strut} \setlength{\fboxsep}{0pt}\def\cbRGB{\colorbox[RGB]}\expandafter\cbRGB\expandafter{\detokenize{146,146,255}}{there\strut} \setlength{\fboxsep}{0pt}\def\cbRGB{\colorbox[RGB]}\expandafter\cbRGB\expandafter{\detokenize{148,148,255}}{was\strut} \setlength{\fboxsep}{0pt}\def\cbRGB{\colorbox[RGB]}\expandafter\cbRGB\expandafter{\detokenize{156,156,255}}{quite\strut} \setlength{\fboxsep}{0pt}\def\cbRGB{\colorbox[RGB]}\expandafter\cbRGB\expandafter{\detokenize{19,19,255}}{a\strut} \setlength{\fboxsep}{0pt}\def\cbRGB{\colorbox[RGB]}\expandafter\cbRGB\expandafter{\detokenize{23,23,255}}{full\strut} \setlength{\fboxsep}{0pt}\def\cbRGB{\colorbox[RGB]}\expandafter\cbRGB\expandafter{\detokenize{7,7,255}}{crowd\strut} \setlength{\fboxsep}{0pt}\def\cbRGB{\colorbox[RGB]}\expandafter\cbRGB\expandafter{\detokenize{147,147,255}}{because\strut} \setlength{\fboxsep}{0pt}\def\cbRGB{\colorbox[RGB]}\expandafter\cbRGB\expandafter{\detokenize{139,139,255}}{they\strut} \setlength{\fboxsep}{0pt}\def\cbRGB{\colorbox[RGB]}\expandafter\cbRGB\expandafter{\detokenize{151,151,255}}{have\strut} \setlength{\fboxsep}{0pt}\def\cbRGB{\colorbox[RGB]}\expandafter\cbRGB\expandafter{\detokenize{150,150,255}}{a\strut} \setlength{\fboxsep}{0pt}\def\cbRGB{\colorbox[RGB]}\expandafter\cbRGB\expandafter{\detokenize{156,156,255}}{second\strut} \setlength{\fboxsep}{0pt}\def\cbRGB{\colorbox[RGB]}\expandafter\cbRGB\expandafter{\detokenize{152,152,255}}{,\strut} \setlength{\fboxsep}{0pt}\def\cbRGB{\colorbox[RGB]}\expandafter\cbRGB\expandafter{\detokenize{154,154,255}}{late\strut} \setlength{\fboxsep}{0pt}\def\cbRGB{\colorbox[RGB]}\expandafter\cbRGB\expandafter{\detokenize{148,148,255}}{happy\strut} \setlength{\fboxsep}{0pt}\def\cbRGB{\colorbox[RGB]}\expandafter\cbRGB\expandafter{\detokenize{146,146,255}}{hour\strut} \setlength{\fboxsep}{0pt}\def\cbRGB{\colorbox[RGB]}\expandafter\cbRGB\expandafter{\detokenize{161,161,255}}{from\strut} \setlength{\fboxsep}{0pt}\def\cbRGB{\colorbox[RGB]}\expandafter\cbRGB\expandafter{\detokenize{161,161,255}}{qqq\strut} \setlength{\fboxsep}{0pt}\def\cbRGB{\colorbox[RGB]}\expandafter\cbRGB\expandafter{\detokenize{162,162,255}}{with\strut} \setlength{\fboxsep}{0pt}\def\cbRGB{\colorbox[RGB]}\expandafter\cbRGB\expandafter{\detokenize{145,145,255}}{great\strut} \setlength{\fboxsep}{0pt}\def\cbRGB{\colorbox[RGB]}\expandafter\cbRGB\expandafter{\detokenize{147,147,255}}{deals\strut} \setlength{\fboxsep}{0pt}\def\cbRGB{\colorbox[RGB]}\expandafter\cbRGB\expandafter{\detokenize{158,158,255}}{on\strut} \setlength{\fboxsep}{0pt}\def\cbRGB{\colorbox[RGB]}\expandafter\cbRGB\expandafter{\detokenize{157,157,255}}{drinks\strut} \setlength{\fboxsep}{0pt}\def\cbRGB{\colorbox[RGB]}\expandafter\cbRGB\expandafter{\detokenize{155,155,255}}{and\strut} \setlength{\fboxsep}{0pt}\def\cbRGB{\colorbox[RGB]}\expandafter\cbRGB\expandafter{\detokenize{147,147,255}}{small\strut} \setlength{\fboxsep}{0pt}\def\cbRGB{\colorbox[RGB]}\expandafter\cbRGB\expandafter{\detokenize{150,150,255}}{plates\strut} \setlength{\fboxsep}{0pt}\def\cbRGB{\colorbox[RGB]}\expandafter\cbRGB\expandafter{\detokenize{174,174,255}}{.\strut} \setlength{\fboxsep}{0pt}\def\cbRGB{\colorbox[RGB]}\expandafter\cbRGB\expandafter{\detokenize{184,184,255}}{

\strut} \setlength{\fboxsep}{0pt}\def\cbRGB{\colorbox[RGB]}\expandafter\cbRGB\expandafter{\detokenize{121,121,255}}{i\strut} \setlength{\fboxsep}{0pt}\def\cbRGB{\colorbox[RGB]}\expandafter\cbRGB\expandafter{\detokenize{110,110,255}}{ordered\strut} \setlength{\fboxsep}{0pt}\def\cbRGB{\colorbox[RGB]}\expandafter\cbRGB\expandafter{\detokenize{112,112,255}}{the\strut} \setlength{\fboxsep}{0pt}\def\cbRGB{\colorbox[RGB]}\expandafter\cbRGB\expandafter{\detokenize{175,175,255}}{house\strut} \setlength{\fboxsep}{0pt}\def\cbRGB{\colorbox[RGB]}\expandafter\cbRGB\expandafter{\detokenize{171,171,255}}{red\strut} \setlength{\fboxsep}{0pt}\def\cbRGB{\colorbox[RGB]}\expandafter\cbRGB\expandafter{\detokenize{163,163,255}}{wine\strut} \setlength{\fboxsep}{0pt}\def\cbRGB{\colorbox[RGB]}\expandafter\cbRGB\expandafter{\detokenize{119,119,255}}{(\strut} \setlength{\fboxsep}{0pt}\def\cbRGB{\colorbox[RGB]}\expandafter\cbRGB\expandafter{\detokenize{133,133,255}}{\$\strut} \setlength{\fboxsep}{0pt}\def\cbRGB{\colorbox[RGB]}\expandafter\cbRGB\expandafter{\detokenize{128,128,255}}{qqq\strut} \setlength{\fboxsep}{0pt}\def\cbRGB{\colorbox[RGB]}\expandafter\cbRGB\expandafter{\detokenize{167,167,255}}{at\strut} \setlength{\fboxsep}{0pt}\def\cbRGB{\colorbox[RGB]}\expandafter\cbRGB\expandafter{\detokenize{145,145,255}}{happy\strut} \setlength{\fboxsep}{0pt}\def\cbRGB{\colorbox[RGB]}\expandafter\cbRGB\expandafter{\detokenize{153,153,255}}{hour\strut} \setlength{\fboxsep}{0pt}\def\cbRGB{\colorbox[RGB]}\expandafter\cbRGB\expandafter{\detokenize{157,157,255}}{!\strut} \setlength{\fboxsep}{0pt}\def\cbRGB{\colorbox[RGB]}\expandafter\cbRGB\expandafter{\detokenize{162,162,255}}{)\strut} \setlength{\fboxsep}{0pt}\def\cbRGB{\colorbox[RGB]}\expandafter\cbRGB\expandafter{\detokenize{155,155,255}}{,\strut} \setlength{\fboxsep}{0pt}\def\cbRGB{\colorbox[RGB]}\expandafter\cbRGB\expandafter{\detokenize{153,153,255}}{and\strut} \setlength{\fboxsep}{0pt}\def\cbRGB{\colorbox[RGB]}\expandafter\cbRGB\expandafter{\detokenize{150,150,255}}{two\strut} \setlength{\fboxsep}{0pt}\def\cbRGB{\colorbox[RGB]}\expandafter\cbRGB\expandafter{\detokenize{161,161,255}}{suggestions\strut} \setlength{\fboxsep}{0pt}\def\cbRGB{\colorbox[RGB]}\expandafter\cbRGB\expandafter{\detokenize{171,171,255}}{of\strut} \setlength{\fboxsep}{0pt}\def\cbRGB{\colorbox[RGB]}\expandafter\cbRGB\expandafter{\detokenize{122,122,255}}{the\strut} \setlength{\fboxsep}{0pt}\def\cbRGB{\colorbox[RGB]}\expandafter\cbRGB\expandafter{\detokenize{124,124,255}}{bartender\strut} \setlength{\fboxsep}{0pt}\def\cbRGB{\colorbox[RGB]}\expandafter\cbRGB\expandafter{\detokenize{123,123,255}}{:\strut} \setlength{\fboxsep}{0pt}\def\cbRGB{\colorbox[RGB]}\expandafter\cbRGB\expandafter{\detokenize{175,175,255}}{the\strut} \setlength{\fboxsep}{0pt}\def\cbRGB{\colorbox[RGB]}\expandafter\cbRGB\expandafter{\detokenize{163,163,255}}{roasted\strut} \setlength{\fboxsep}{0pt}\def\cbRGB{\colorbox[RGB]}\expandafter\cbRGB\expandafter{\detokenize{160,160,255}}{dates\strut} \setlength{\fboxsep}{0pt}\def\cbRGB{\colorbox[RGB]}\expandafter\cbRGB\expandafter{\detokenize{156,156,255}}{to\strut} \setlength{\fboxsep}{0pt}\def\cbRGB{\colorbox[RGB]}\expandafter\cbRGB\expandafter{\detokenize{156,156,255}}{start\strut} \setlength{\fboxsep}{0pt}\def\cbRGB{\colorbox[RGB]}\expandafter\cbRGB\expandafter{\detokenize{149,149,255}}{and\strut} \setlength{\fboxsep}{0pt}\def\cbRGB{\colorbox[RGB]}\expandafter\cbRGB\expandafter{\detokenize{150,150,255}}{the\strut} \setlength{\fboxsep}{0pt}\def\cbRGB{\colorbox[RGB]}\expandafter\cbRGB\expandafter{\detokenize{150,150,255}}{steak\strut} \setlength{\fboxsep}{0pt}\def\cbRGB{\colorbox[RGB]}\expandafter\cbRGB\expandafter{\detokenize{151,151,255}}{entree\strut} \setlength{\fboxsep}{0pt}\def\cbRGB{\colorbox[RGB]}\expandafter\cbRGB\expandafter{\detokenize{152,152,255}}{with\strut} \setlength{\fboxsep}{0pt}\def\cbRGB{\colorbox[RGB]}\expandafter\cbRGB\expandafter{\detokenize{154,154,255}}{a\strut} \setlength{\fboxsep}{0pt}\def\cbRGB{\colorbox[RGB]}\expandafter\cbRGB\expandafter{\detokenize{154,154,255}}{side\strut} \setlength{\fboxsep}{0pt}\def\cbRGB{\colorbox[RGB]}\expandafter\cbRGB\expandafter{\detokenize{148,148,255}}{of\strut} \setlength{\fboxsep}{0pt}\def\cbRGB{\colorbox[RGB]}\expandafter\cbRGB\expandafter{\detokenize{148,148,255}}{brussels\strut} \setlength{\fboxsep}{0pt}\def\cbRGB{\colorbox[RGB]}\expandafter\cbRGB\expandafter{\detokenize{154,154,255}}{sprouts\strut} \setlength{\fboxsep}{0pt}\def\cbRGB{\colorbox[RGB]}\expandafter\cbRGB\expandafter{\detokenize{154,154,255}}{.\strut} \setlength{\fboxsep}{0pt}\def\cbRGB{\colorbox[RGB]}\expandafter\cbRGB\expandafter{\detokenize{157,157,255}}{the\strut} \setlength{\fboxsep}{0pt}\def\cbRGB{\colorbox[RGB]}\expandafter\cbRGB\expandafter{\detokenize{156,156,255}}{dates\strut} \setlength{\fboxsep}{0pt}\def\cbRGB{\colorbox[RGB]}\expandafter\cbRGB\expandafter{\detokenize{117,117,255}}{(\strut} \setlength{\fboxsep}{0pt}\def\cbRGB{\colorbox[RGB]}\expandafter\cbRGB\expandafter{\detokenize{133,133,255}}{\$\strut} \setlength{\fboxsep}{0pt}\def\cbRGB{\colorbox[RGB]}\expandafter\cbRGB\expandafter{\detokenize{130,130,255}}{qqq\strut} \setlength{\fboxsep}{0pt}\def\cbRGB{\colorbox[RGB]}\expandafter\cbRGB\expandafter{\detokenize{166,166,255}}{)\strut} \setlength{\fboxsep}{0pt}\def\cbRGB{\colorbox[RGB]}\expandafter\cbRGB\expandafter{\detokenize{154,154,255}}{were\strut} \setlength{\fboxsep}{0pt}\def\cbRGB{\colorbox[RGB]}\expandafter\cbRGB\expandafter{\detokenize{149,149,255}}{incredibly\strut} \setlength{\fboxsep}{0pt}\def\cbRGB{\colorbox[RGB]}\expandafter\cbRGB\expandafter{\detokenize{156,156,255}}{decadent\strut} \setlength{\fboxsep}{0pt}\def\cbRGB{\colorbox[RGB]}\expandafter\cbRGB\expandafter{\detokenize{152,152,255}}{:\strut} \setlength{\fboxsep}{0pt}\def\cbRGB{\colorbox[RGB]}\expandafter\cbRGB\expandafter{\detokenize{155,155,255}}{roasted\strut} \setlength{\fboxsep}{0pt}\def\cbRGB{\colorbox[RGB]}\expandafter\cbRGB\expandafter{\detokenize{164,164,255}}{and\strut} \setlength{\fboxsep}{0pt}\def\cbRGB{\colorbox[RGB]}\expandafter\cbRGB\expandafter{\detokenize{167,167,255}}{topped\strut} \setlength{\fboxsep}{0pt}\def\cbRGB{\colorbox[RGB]}\expandafter\cbRGB\expandafter{\detokenize{146,146,255}}{with\strut} \setlength{\fboxsep}{0pt}\def\cbRGB{\colorbox[RGB]}\expandafter\cbRGB\expandafter{\detokenize{161,161,255}}{almonds\strut} \setlength{\fboxsep}{0pt}\def\cbRGB{\colorbox[RGB]}\expandafter\cbRGB\expandafter{\detokenize{81,81,255}}{,\strut} \setlength{\fboxsep}{0pt}\def\cbRGB{\colorbox[RGB]}\expandafter\cbRGB\expandafter{\detokenize{123,123,255}}{bacon\strut} \setlength{\fboxsep}{0pt}\def\cbRGB{\colorbox[RGB]}\expandafter\cbRGB\expandafter{\detokenize{136,136,255}}{,\strut} \setlength{\fboxsep}{0pt}\def\cbRGB{\colorbox[RGB]}\expandafter\cbRGB\expandafter{\detokenize{231,231,255}}{unk\strut} \setlength{\fboxsep}{0pt}\def\cbRGB{\colorbox[RGB]}\expandafter\cbRGB\expandafter{\detokenize{82,82,255}}{and\strut} \setlength{\fboxsep}{0pt}\def\cbRGB{\colorbox[RGB]}\expandafter\cbRGB\expandafter{\detokenize{75,75,255}}{parsley\strut} \setlength{\fboxsep}{0pt}\def\cbRGB{\colorbox[RGB]}\expandafter\cbRGB\expandafter{\detokenize{75,75,255}}{,\strut} \setlength{\fboxsep}{0pt}\def\cbRGB{\colorbox[RGB]}\expandafter\cbRGB\expandafter{\detokenize{203,203,255}}{they\strut} \setlength{\fboxsep}{0pt}\def\cbRGB{\colorbox[RGB]}\expandafter\cbRGB\expandafter{\detokenize{175,175,255}}{are\strut} \setlength{\fboxsep}{0pt}\def\cbRGB{\colorbox[RGB]}\expandafter\cbRGB\expandafter{\detokenize{154,154,255}}{a\strut} \setlength{\fboxsep}{0pt}\def\cbRGB{\colorbox[RGB]}\expandafter\cbRGB\expandafter{\detokenize{152,152,255}}{party\strut} \setlength{\fboxsep}{0pt}\def\cbRGB{\colorbox[RGB]}\expandafter\cbRGB\expandafter{\detokenize{148,148,255}}{in\strut} \setlength{\fboxsep}{0pt}\def\cbRGB{\colorbox[RGB]}\expandafter\cbRGB\expandafter{\detokenize{161,161,255}}{your\strut} \setlength{\fboxsep}{0pt}\def\cbRGB{\colorbox[RGB]}\expandafter\cbRGB\expandafter{\detokenize{175,175,255}}{both\strut} \setlength{\fboxsep}{0pt}\def\cbRGB{\colorbox[RGB]}\expandafter\cbRGB\expandafter{\detokenize{135,135,255}}{of\strut} \setlength{\fboxsep}{0pt}\def\cbRGB{\colorbox[RGB]}\expandafter\cbRGB\expandafter{\detokenize{153,153,255}}{sweet\strut} \setlength{\fboxsep}{0pt}\def\cbRGB{\colorbox[RGB]}\expandafter\cbRGB\expandafter{\detokenize{77,77,255}}{and\strut} \setlength{\fboxsep}{0pt}\def\cbRGB{\colorbox[RGB]}\expandafter\cbRGB\expandafter{\detokenize{148,148,255}}{salty\strut} \setlength{\fboxsep}{0pt}\def\cbRGB{\colorbox[RGB]}\expandafter\cbRGB\expandafter{\detokenize{71,71,255}}{and\strut} \setlength{\fboxsep}{0pt}\def\cbRGB{\colorbox[RGB]}\expandafter\cbRGB\expandafter{\detokenize{171,171,255}}{spicy\strut} \setlength{\fboxsep}{0pt}\def\cbRGB{\colorbox[RGB]}\expandafter\cbRGB\expandafter{\detokenize{164,164,255}}{.\strut} \setlength{\fboxsep}{0pt}\def\cbRGB{\colorbox[RGB]}\expandafter\cbRGB\expandafter{\detokenize{131,131,255}}{the\strut} \setlength{\fboxsep}{0pt}\def\cbRGB{\colorbox[RGB]}\expandafter\cbRGB\expandafter{\detokenize{120,120,255}}{sirloin\strut} \setlength{\fboxsep}{0pt}\def\cbRGB{\colorbox[RGB]}\expandafter\cbRGB\expandafter{\detokenize{74,74,255}}{(\strut} \setlength{\fboxsep}{0pt}\def\cbRGB{\colorbox[RGB]}\expandafter\cbRGB\expandafter{\detokenize{172,172,255}}{\$\strut} \setlength{\fboxsep}{0pt}\def\cbRGB{\colorbox[RGB]}\expandafter\cbRGB\expandafter{\detokenize{149,149,255}}{qqq\strut} \setlength{\fboxsep}{0pt}\def\cbRGB{\colorbox[RGB]}\expandafter\cbRGB\expandafter{\detokenize{172,172,255}}{)\strut} \setlength{\fboxsep}{0pt}\def\cbRGB{\colorbox[RGB]}\expandafter\cbRGB\expandafter{\detokenize{163,163,255}}{,\strut} \setlength{\fboxsep}{0pt}\def\cbRGB{\colorbox[RGB]}\expandafter\cbRGB\expandafter{\detokenize{169,169,255}}{which\strut} \setlength{\fboxsep}{0pt}\def\cbRGB{\colorbox[RGB]}\expandafter\cbRGB\expandafter{\detokenize{120,120,255}}{the\strut} \setlength{\fboxsep}{0pt}\def\cbRGB{\colorbox[RGB]}\expandafter\cbRGB\expandafter{\detokenize{121,121,255}}{bartender\strut} \setlength{\fboxsep}{0pt}\def\cbRGB{\colorbox[RGB]}\expandafter\cbRGB\expandafter{\detokenize{127,127,255}}{told\strut} \setlength{\fboxsep}{0pt}\def\cbRGB{\colorbox[RGB]}\expandafter\cbRGB\expandafter{\detokenize{174,174,255}}{me\strut} \setlength{\fboxsep}{0pt}\def\cbRGB{\colorbox[RGB]}\expandafter\cbRGB\expandafter{\detokenize{193,193,255}}{is\strut} \setlength{\fboxsep}{0pt}\def\cbRGB{\colorbox[RGB]}\expandafter\cbRGB\expandafter{\detokenize{225,225,255}}{a\strut} \setlength{\fboxsep}{0pt}\def\cbRGB{\colorbox[RGB]}\expandafter\cbRGB\expandafter{\detokenize{118,118,255}}{new\strut} \setlength{\fboxsep}{0pt}\def\cbRGB{\colorbox[RGB]}\expandafter\cbRGB\expandafter{\detokenize{0,0,255}}{menu\strut} \setlength{\fboxsep}{0pt}\def\cbRGB{\colorbox[RGB]}\expandafter\cbRGB\expandafter{\detokenize{2,2,255}}{item\strut} \setlength{\fboxsep}{0pt}\def\cbRGB{\colorbox[RGB]}\expandafter\cbRGB\expandafter{\detokenize{128,128,255}}{,\strut} \setlength{\fboxsep}{0pt}\def\cbRGB{\colorbox[RGB]}\expandafter\cbRGB\expandafter{\detokenize{215,215,255}}{was\strut} \setlength{\fboxsep}{0pt}\def\cbRGB{\colorbox[RGB]}\expandafter\cbRGB\expandafter{\detokenize{167,167,255}}{beautifully\strut} \setlength{\fboxsep}{0pt}\def\cbRGB{\colorbox[RGB]}\expandafter\cbRGB\expandafter{\detokenize{150,150,255}}{cooked\strut} \setlength{\fboxsep}{0pt}\def\cbRGB{\colorbox[RGB]}\expandafter\cbRGB\expandafter{\detokenize{153,153,255}}{and\strut} \setlength{\fboxsep}{0pt}\def\cbRGB{\colorbox[RGB]}\expandafter\cbRGB\expandafter{\detokenize{150,150,255}}{topped\strut} \setlength{\fboxsep}{0pt}\def\cbRGB{\colorbox[RGB]}\expandafter\cbRGB\expandafter{\detokenize{183,183,255}}{with\strut} \setlength{\fboxsep}{0pt}\def\cbRGB{\colorbox[RGB]}\expandafter\cbRGB\expandafter{\detokenize{213,213,255}}{parmesan\strut} \setlength{\fboxsep}{0pt}\def\cbRGB{\colorbox[RGB]}\expandafter\cbRGB\expandafter{\detokenize{63,63,255}}{,\strut} \setlength{\fboxsep}{0pt}\def\cbRGB{\colorbox[RGB]}\expandafter\cbRGB\expandafter{\detokenize{60,60,255}}{truffle\strut} \setlength{\fboxsep}{0pt}\def\cbRGB{\colorbox[RGB]}\expandafter\cbRGB\expandafter{\detokenize{60,60,255}}{butter\strut} \setlength{\fboxsep}{0pt}\def\cbRGB{\colorbox[RGB]}\expandafter\cbRGB\expandafter{\detokenize{251,251,255}}{,\strut} \setlength{\fboxsep}{0pt}\def\cbRGB{\colorbox[RGB]}\expandafter\cbRGB\expandafter{\detokenize{255,255,255}}{arugula\strut} \setlength{\fboxsep}{0pt}\def\cbRGB{\colorbox[RGB]}\expandafter\cbRGB\expandafter{\detokenize{29,29,255}}{and\strut} \setlength{\fboxsep}{0pt}\def\cbRGB{\colorbox[RGB]}\expandafter\cbRGB\expandafter{\detokenize{35,35,255}}{mushrooms\strut} \setlength{\fboxsep}{0pt}\def\cbRGB{\colorbox[RGB]}\expandafter\cbRGB\expandafter{\detokenize{33,33,255}}{.\strut} \setlength{\fboxsep}{0pt}\def\cbRGB{\colorbox[RGB]}\expandafter\cbRGB\expandafter{\detokenize{225,225,255}}{it\strut} \setlength{\fboxsep}{0pt}\def\cbRGB{\colorbox[RGB]}\expandafter\cbRGB\expandafter{\detokenize{179,179,255}}{was\strut} \setlength{\fboxsep}{0pt}\def\cbRGB{\colorbox[RGB]}\expandafter\cbRGB\expandafter{\detokenize{152,152,255}}{good\strut} \setlength{\fboxsep}{0pt}\def\cbRGB{\colorbox[RGB]}\expandafter\cbRGB\expandafter{\detokenize{154,154,255}}{,\strut} \setlength{\fboxsep}{0pt}\def\cbRGB{\colorbox[RGB]}\expandafter\cbRGB\expandafter{\detokenize{149,149,255}}{although\strut} \setlength{\fboxsep}{0pt}\def\cbRGB{\colorbox[RGB]}\expandafter\cbRGB\expandafter{\detokenize{145,145,255}}{there\strut} \setlength{\fboxsep}{0pt}\def\cbRGB{\colorbox[RGB]}\expandafter\cbRGB\expandafter{\detokenize{148,148,255}}{are\strut} \setlength{\fboxsep}{0pt}\def\cbRGB{\colorbox[RGB]}\expandafter\cbRGB\expandafter{\detokenize{150,150,255}}{other\strut} \setlength{\fboxsep}{0pt}\def\cbRGB{\colorbox[RGB]}\expandafter\cbRGB\expandafter{\detokenize{178,178,255}}{things\strut} \setlength{\fboxsep}{0pt}\def\cbRGB{\colorbox[RGB]}\expandafter\cbRGB\expandafter{\detokenize{200,200,255}}{on\strut} \setlength{\fboxsep}{0pt}\def\cbRGB{\colorbox[RGB]}\expandafter\cbRGB\expandafter{\detokenize{76,76,255}}{the\strut} \setlength{\fboxsep}{0pt}\def\cbRGB{\colorbox[RGB]}\expandafter\cbRGB\expandafter{\detokenize{74,74,255}}{menu\strut} \setlength{\fboxsep}{0pt}\def\cbRGB{\colorbox[RGB]}\expandafter\cbRGB\expandafter{\detokenize{76,76,255}}{that\strut} \setlength{\fboxsep}{0pt}\def\cbRGB{\colorbox[RGB]}\expandafter\cbRGB\expandafter{\detokenize{199,199,255}}{might\strut} \setlength{\fboxsep}{0pt}\def\cbRGB{\colorbox[RGB]}\expandafter\cbRGB\expandafter{\detokenize{171,171,255}}{be\strut} \setlength{\fboxsep}{0pt}\def\cbRGB{\colorbox[RGB]}\expandafter\cbRGB\expandafter{\detokenize{145,145,255}}{more\strut} \setlength{\fboxsep}{0pt}\def\cbRGB{\colorbox[RGB]}\expandafter\cbRGB\expandafter{\detokenize{152,152,255}}{exciting\strut} \setlength{\fboxsep}{0pt}\def\cbRGB{\colorbox[RGB]}\expandafter\cbRGB\expandafter{\detokenize{151,151,255}}{.\strut} \setlength{\fboxsep}{0pt}\def\cbRGB{\colorbox[RGB]}\expandafter\cbRGB\expandafter{\detokenize{156,156,255}}{the\strut} \setlength{\fboxsep}{0pt}\def\cbRGB{\colorbox[RGB]}\expandafter\cbRGB\expandafter{\detokenize{149,149,255}}{stand\strut} \setlength{\fboxsep}{0pt}\def\cbRGB{\colorbox[RGB]}\expandafter\cbRGB\expandafter{\detokenize{154,154,255}}{out\strut} \setlength{\fboxsep}{0pt}\def\cbRGB{\colorbox[RGB]}\expandafter\cbRGB\expandafter{\detokenize{149,149,255}}{dish\strut} \setlength{\fboxsep}{0pt}\def\cbRGB{\colorbox[RGB]}\expandafter\cbRGB\expandafter{\detokenize{150,150,255}}{of\strut} \setlength{\fboxsep}{0pt}\def\cbRGB{\colorbox[RGB]}\expandafter\cbRGB\expandafter{\detokenize{146,146,255}}{the\strut} \setlength{\fboxsep}{0pt}\def\cbRGB{\colorbox[RGB]}\expandafter\cbRGB\expandafter{\detokenize{158,158,255}}{evening\strut} \setlength{\fboxsep}{0pt}\def\cbRGB{\colorbox[RGB]}\expandafter\cbRGB\expandafter{\detokenize{173,173,255}}{was\strut} \setlength{\fboxsep}{0pt}\def\cbRGB{\colorbox[RGB]}\expandafter\cbRGB\expandafter{\detokenize{114,114,255}}{the\strut} \setlength{\fboxsep}{0pt}\def\cbRGB{\colorbox[RGB]}\expandafter\cbRGB\expandafter{\detokenize{113,113,255}}{fried\strut} \setlength{\fboxsep}{0pt}\def\cbRGB{\colorbox[RGB]}\expandafter\cbRGB\expandafter{\detokenize{114,114,255}}{brussels\strut} \setlength{\fboxsep}{0pt}\def\cbRGB{\colorbox[RGB]}\expandafter\cbRGB\expandafter{\detokenize{179,179,255}}{sprouts\strut} \setlength{\fboxsep}{0pt}\def\cbRGB{\colorbox[RGB]}\expandafter\cbRGB\expandafter{\detokenize{167,167,255}}{,\strut} \setlength{\fboxsep}{0pt}\def\cbRGB{\colorbox[RGB]}\expandafter\cbRGB\expandafter{\detokenize{154,154,255}}{which\strut} \setlength{\fboxsep}{0pt}\def\cbRGB{\colorbox[RGB]}\expandafter\cbRGB\expandafter{\detokenize{149,149,255}}{i\strut} \setlength{\fboxsep}{0pt}\def\cbRGB{\colorbox[RGB]}\expandafter\cbRGB\expandafter{\detokenize{148,148,255}}{would\strut} \setlength{\fboxsep}{0pt}\def\cbRGB{\colorbox[RGB]}\expandafter\cbRGB\expandafter{\detokenize{144,144,255}}{very\strut} \setlength{\fboxsep}{0pt}\def\cbRGB{\colorbox[RGB]}\expandafter\cbRGB\expandafter{\detokenize{173,173,255}}{much\strut} \setlength{\fboxsep}{0pt}\def\cbRGB{\colorbox[RGB]}\expandafter\cbRGB\expandafter{\detokenize{197,197,255}}{like\strut} \setlength{\fboxsep}{0pt}\def\cbRGB{\colorbox[RGB]}\expandafter\cbRGB\expandafter{\detokenize{84,84,255}}{the\strut} \setlength{\fboxsep}{0pt}\def\cbRGB{\colorbox[RGB]}\expandafter\cbRGB\expandafter{\detokenize{80,80,255}}{recipe\strut} \setlength{\fboxsep}{0pt}\def\cbRGB{\colorbox[RGB]}\expandafter\cbRGB\expandafter{\detokenize{80,80,255}}{for\strut} \setlength{\fboxsep}{0pt}\def\cbRGB{\colorbox[RGB]}\expandafter\cbRGB\expandafter{\detokenize{196,196,255}}{so\strut} \setlength{\fboxsep}{0pt}\def\cbRGB{\colorbox[RGB]}\expandafter\cbRGB\expandafter{\detokenize{171,171,255}}{i\strut} \setlength{\fboxsep}{0pt}\def\cbRGB{\colorbox[RGB]}\expandafter\cbRGB\expandafter{\detokenize{149,149,255}}{can\strut} \setlength{\fboxsep}{0pt}\def\cbRGB{\colorbox[RGB]}\expandafter\cbRGB\expandafter{\detokenize{147,147,255}}{unk\strut} \setlength{\fboxsep}{0pt}\def\cbRGB{\colorbox[RGB]}\expandafter\cbRGB\expandafter{\detokenize{148,148,255}}{myself\strut} \setlength{\fboxsep}{0pt}\def\cbRGB{\colorbox[RGB]}\expandafter\cbRGB\expandafter{\detokenize{153,153,255}}{on\strut} \setlength{\fboxsep}{0pt}\def\cbRGB{\colorbox[RGB]}\expandafter\cbRGB\expandafter{\detokenize{153,153,255}}{them\strut} \setlength{\fboxsep}{0pt}\def\cbRGB{\colorbox[RGB]}\expandafter\cbRGB\expandafter{\detokenize{150,150,255}}{every\strut} \setlength{\fboxsep}{0pt}\def\cbRGB{\colorbox[RGB]}\expandafter\cbRGB\expandafter{\detokenize{151,151,255}}{night\strut} \setlength{\fboxsep}{0pt}\def\cbRGB{\colorbox[RGB]}\expandafter\cbRGB\expandafter{\detokenize{151,151,255}}{.\strut} \setlength{\fboxsep}{0pt}\def\cbRGB{\colorbox[RGB]}\expandafter\cbRGB\expandafter{\detokenize{152,152,255}}{those\strut} \setlength{\fboxsep}{0pt}\def\cbRGB{\colorbox[RGB]}\expandafter\cbRGB\expandafter{\detokenize{151,151,255}}{babies\strut} \setlength{\fboxsep}{0pt}\def\cbRGB{\colorbox[RGB]}\expandafter\cbRGB\expandafter{\detokenize{187,187,255}}{,\strut} \setlength{\fboxsep}{0pt}\def\cbRGB{\colorbox[RGB]}\expandafter\cbRGB\expandafter{\detokenize{216,216,255}}{topped\strut} \setlength{\fboxsep}{0pt}\def\cbRGB{\colorbox[RGB]}\expandafter\cbRGB\expandafter{\detokenize{55,55,255}}{with\strut} \setlength{\fboxsep}{0pt}\def\cbRGB{\colorbox[RGB]}\expandafter\cbRGB\expandafter{\detokenize{40,40,255}}{crispy\strut} \setlength{\fboxsep}{0pt}\def\cbRGB{\colorbox[RGB]}\expandafter\cbRGB\expandafter{\detokenize{40,40,255}}{crumbled\strut} \setlength{\fboxsep}{0pt}\def\cbRGB{\colorbox[RGB]}\expandafter\cbRGB\expandafter{\detokenize{196,196,255}}{bits\strut} \setlength{\fboxsep}{0pt}\def\cbRGB{\colorbox[RGB]}\expandafter\cbRGB\expandafter{\detokenize{175,175,255}}{of\strut} \setlength{\fboxsep}{0pt}\def\cbRGB{\colorbox[RGB]}\expandafter\cbRGB\expandafter{\detokenize{152,152,255}}{unk\strut} \setlength{\fboxsep}{0pt}\def\cbRGB{\colorbox[RGB]}\expandafter\cbRGB\expandafter{\detokenize{152,152,255}}{,\strut} \setlength{\fboxsep}{0pt}\def\cbRGB{\colorbox[RGB]}\expandafter\cbRGB\expandafter{\detokenize{153,153,255}}{capers\strut} \setlength{\fboxsep}{0pt}\def\cbRGB{\colorbox[RGB]}\expandafter\cbRGB\expandafter{\detokenize{150,150,255}}{and\strut} \setlength{\fboxsep}{0pt}\def\cbRGB{\colorbox[RGB]}\expandafter\cbRGB\expandafter{\detokenize{156,156,255}}{walnuts\strut} \setlength{\fboxsep}{0pt}\def\cbRGB{\colorbox[RGB]}\expandafter\cbRGB\expandafter{\detokenize{155,155,255}}{,\strut} \setlength{\fboxsep}{0pt}\def\cbRGB{\colorbox[RGB]}\expandafter\cbRGB\expandafter{\detokenize{147,147,255}}{were\strut} \setlength{\fboxsep}{0pt}\def\cbRGB{\colorbox[RGB]}\expandafter\cbRGB\expandafter{\detokenize{142,142,255}}{pure\strut} \setlength{\fboxsep}{0pt}\def\cbRGB{\colorbox[RGB]}\expandafter\cbRGB\expandafter{\detokenize{149,149,255}}{bliss\strut} \setlength{\fboxsep}{0pt}\def\cbRGB{\colorbox[RGB]}\expandafter\cbRGB\expandafter{\detokenize{153,153,255}}{.\strut} \setlength{\fboxsep}{0pt}\def\cbRGB{\colorbox[RGB]}\expandafter\cbRGB\expandafter{\detokenize{157,157,255}}{and\strut} \setlength{\fboxsep}{0pt}\def\cbRGB{\colorbox[RGB]}\expandafter\cbRGB\expandafter{\detokenize{157,157,255}}{let\strut} \setlength{\fboxsep}{0pt}\def\cbRGB{\colorbox[RGB]}\expandafter\cbRGB\expandafter{\detokenize{156,156,255}}{'s\strut} \setlength{\fboxsep}{0pt}\def\cbRGB{\colorbox[RGB]}\expandafter\cbRGB\expandafter{\detokenize{153,153,255}}{be\strut} \setlength{\fboxsep}{0pt}\def\cbRGB{\colorbox[RGB]}\expandafter\cbRGB\expandafter{\detokenize{152,152,255}}{honest\strut} \setlength{\fboxsep}{0pt}\def\cbRGB{\colorbox[RGB]}\expandafter\cbRGB\expandafter{\detokenize{156,156,255}}{,\strut} \setlength{\fboxsep}{0pt}\def\cbRGB{\colorbox[RGB]}\expandafter\cbRGB\expandafter{\detokenize{155,155,255}}{how\strut} \setlength{\fboxsep}{0pt}\def\cbRGB{\colorbox[RGB]}\expandafter\cbRGB\expandafter{\detokenize{149,149,255}}{often\strut} \setlength{\fboxsep}{0pt}\def\cbRGB{\colorbox[RGB]}\expandafter\cbRGB\expandafter{\detokenize{148,148,255}}{do\strut} \setlength{\fboxsep}{0pt}\def\cbRGB{\colorbox[RGB]}\expandafter\cbRGB\expandafter{\detokenize{148,148,255}}{you\strut} \setlength{\fboxsep}{0pt}\def\cbRGB{\colorbox[RGB]}\expandafter\cbRGB\expandafter{\detokenize{148,148,255}}{really\strut} \setlength{\fboxsep}{0pt}\def\cbRGB{\colorbox[RGB]}\expandafter\cbRGB\expandafter{\detokenize{147,147,255}}{have\strut} \setlength{\fboxsep}{0pt}\def\cbRGB{\colorbox[RGB]}\expandafter\cbRGB\expandafter{\detokenize{148,148,255}}{the\strut} \setlength{\fboxsep}{0pt}\def\cbRGB{\colorbox[RGB]}\expandafter\cbRGB\expandafter{\detokenize{155,155,255}}{opportunity\strut} \setlength{\fboxsep}{0pt}\def\cbRGB{\colorbox[RGB]}\expandafter\cbRGB\expandafter{\detokenize{155,155,255}}{to\strut} \setlength{\fboxsep}{0pt}\def\cbRGB{\colorbox[RGB]}\expandafter\cbRGB\expandafter{\detokenize{152,152,255}}{call\strut} \setlength{\fboxsep}{0pt}\def\cbRGB{\colorbox[RGB]}\expandafter\cbRGB\expandafter{\detokenize{149,149,255}}{unk\strut} \setlength{\fboxsep}{0pt}\def\cbRGB{\colorbox[RGB]}\expandafter\cbRGB\expandafter{\detokenize{150,150,255}}{sprouts\strut} \setlength{\fboxsep}{0pt}\def\cbRGB{\colorbox[RGB]}\expandafter\cbRGB\expandafter{\detokenize{152,152,255}}{unk\strut} \setlength{\fboxsep}{0pt}\def\cbRGB{\colorbox[RGB]}\expandafter\cbRGB\expandafter{\detokenize{155,155,255}}{?\strut} \setlength{\fboxsep}{0pt}\def\cbRGB{\colorbox[RGB]}\expandafter\cbRGB\expandafter{\detokenize{154,154,255}}{probably\strut} \setlength{\fboxsep}{0pt}\def\cbRGB{\colorbox[RGB]}\expandafter\cbRGB\expandafter{\detokenize{150,150,255}}{not\strut} \setlength{\fboxsep}{0pt}\def\cbRGB{\colorbox[RGB]}\expandafter\cbRGB\expandafter{\detokenize{144,144,255}}{very\strut} \setlength{\fboxsep}{0pt}\def\cbRGB{\colorbox[RGB]}\expandafter\cbRGB\expandafter{\detokenize{149,149,255}}{often\strut} \setlength{\fboxsep}{0pt}\def\cbRGB{\colorbox[RGB]}\expandafter\cbRGB\expandafter{\detokenize{161,161,255}}{.\strut} \setlength{\fboxsep}{0pt}\def\cbRGB{\colorbox[RGB]}\expandafter\cbRGB\expandafter{\detokenize{161,161,255}}{\strut} \setlength{\fboxsep}{0pt}\def\cbRGB{\colorbox[RGB]}\expandafter\cbRGB\expandafter{\detokenize{156,156,255}}{i\strut} \setlength{\fboxsep}{0pt}\def\cbRGB{\colorbox[RGB]}\expandafter\cbRGB\expandafter{\detokenize{147,147,255}}{unk\strut} \setlength{\fboxsep}{0pt}\def\cbRGB{\colorbox[RGB]}\expandafter\cbRGB\expandafter{\detokenize{146,146,255}}{off\strut} \setlength{\fboxsep}{0pt}\def\cbRGB{\colorbox[RGB]}\expandafter\cbRGB\expandafter{\detokenize{145,145,255}}{the\strut} \setlength{\fboxsep}{0pt}\def\cbRGB{\colorbox[RGB]}\expandafter\cbRGB\expandafter{\detokenize{146,146,255}}{evening\strut} \setlength{\fboxsep}{0pt}\def\cbRGB{\colorbox[RGB]}\expandafter\cbRGB\expandafter{\detokenize{151,151,255}}{with\strut} \setlength{\fboxsep}{0pt}\def\cbRGB{\colorbox[RGB]}\expandafter\cbRGB\expandafter{\detokenize{152,152,255}}{a\strut} \setlength{\fboxsep}{0pt}\def\cbRGB{\colorbox[RGB]}\expandafter\cbRGB\expandafter{\detokenize{167,167,255}}{glass\strut} \setlength{\fboxsep}{0pt}\def\cbRGB{\colorbox[RGB]}\expandafter\cbRGB\expandafter{\detokenize{176,176,255}}{of\strut} \setlength{\fboxsep}{0pt}\def\cbRGB{\colorbox[RGB]}\expandafter\cbRGB\expandafter{\detokenize{108,108,255}}{unk\strut} \setlength{\fboxsep}{0pt}\def\cbRGB{\colorbox[RGB]}\expandafter\cbRGB\expandafter{\detokenize{110,110,255}}{scotch\strut} \setlength{\fboxsep}{0pt}\def\cbRGB{\colorbox[RGB]}\expandafter\cbRGB\expandafter{\detokenize{111,111,255}}{,\strut} \setlength{\fboxsep}{0pt}\def\cbRGB{\colorbox[RGB]}\expandafter\cbRGB\expandafter{\detokenize{182,182,255}}{neat\strut} \setlength{\fboxsep}{0pt}\def\cbRGB{\colorbox[RGB]}\expandafter\cbRGB\expandafter{\detokenize{160,160,255}}{.\strut} \setlength{\fboxsep}{0pt}\def\cbRGB{\colorbox[RGB]}\expandafter\cbRGB\expandafter{\detokenize{150,150,255}}{perfection\strut} \setlength{\fboxsep}{0pt}\def\cbRGB{\colorbox[RGB]}\expandafter\cbRGB\expandafter{\detokenize{129,129,255}}{.\strut} 

\begin{table*}
\footnotesize
\centering
\begin{tabular}{|c|}
\hline
\textbf{PICO Domain} \\
\hline
 listening, respondent, perceived, motivation, attitude, participant, asking, apprehension, helpfulness, antismoking \\
 virtual, prototype, smart, tracking, internally, locked, haptic, handle, autism, autistic \\
 psychosocial, gluconate, mental, prospective, social, genetic, psychological, environmental, sulphadoxine, reproductive \\
 regain, obviously, drive, relieved, promoted, bid, diego, mdi, bocs, rescue \\
 workplace, caring, carers, helpfulness, occupational, awareness, spiritual, motivating, personal, educational \\
 euro, biochemical, virological, effectivity, yearly, oxidation, dos, reversion, quitter, audiologic \\
 semen, infertile, uncircumcised, sperm, fertility, azoospermia, fe, ejaculation, caput, particulate\\
 obstetrician, radiologist, cardiologist, technician, physician, midwife, nurse, gynecologist, physiotherapist, anaesthetist \\
 nonsignificant, bias, unadjusted, quadratic, constraint, apgar, undue, trend, nonstatistically, ckd \\
 dependency, dependence, comorbid, dependent, type, sensitivity, insensitive, manner, specific, comorbidity \\
 stem, progenitor, resident, concomitant, recruiting, mobilized, promoted, precursor, malondialdehyde, reduces \\
 anemic, cardiorespiratory, normothermic, hemiplegic, thromboelastography, hematological, hemodilution, dyslipidemic \\
 premenopausal, locoregional, postmenopausal, menopausal, cancer, nonmetastatic, nsclc, pelvic, tirilazad, operable \\
 conscious, intubated, anaesthetized, terlipressin, corticotropin, ventilated, resuscitation, anesthetized, resuscitated, vasopressin \\
 quasi, intergroup, midstream, static, voluntary, csa, vienna, proprioceptive, stroop, multinational \\
 trachea, malnourished, intestine, diaphragm, semitendinosus, gastrocnemius, undernourished, pancreas, lung, nonanemic \\
 pound, mile, customized, pack, per, hundred, liter, oz, thousand, litre \\
 lying, correlated, activity, matter, functioning, transported, gut, subscale, phosphodiesterase, impacting \\ 
 count, admission, disparity, spot, parasitaemia, visit, quartile, white, age, wbc \\
\hline
\textbf{Yelp!-TripAdvisor Domain} \\
\hline
 like, hate, grow, love, coming, come, hygienic, desperate, reminds, unsanitary, complainer \\
 food, service, restaurant, resturants, steakhouse, restaraunt, tapa, eats, restraunts, ate, eatery \\
 bunny, chimney, ooze, sliver, masked, adorned, cow, concrete, coating, triangle, beige \\
  cause, get, brain, nt, annoying, guess, time, sometimes, anymore, cuz, hurry \\
 week, bosa, month, today, california, visited, newton, bought, yogurtland, exellent, recent\\
 caring, outgoing, demeanor, gm, personable, engaging, employee, respectful, interaction, associate, incompetent \\
 staff, excellent, personel, good, unseasoned, superb, unremarkable, excellant, tasty, personnel, unfailingly \\
 place, sounding, scenery, cuppa, temptation, kick, indulge, confection, coffeehouse, lavender \\
 sweetened, watery, raisin, flavoring, doughy, mmmm, crumble, hazelnut, consistency, lemonade, powdered \\
 upgrade, premium, room, staff, offer, amenity, perk, suite, whereas, price, pricing \\
 walk, lined, directed, nearest, headed, adjacent, gaze, grab, lovely, surrounded, field \\
 time, afternoon, night, day, moment, evening, went, rushed, place, boyfriend, came \\
 hotel, property, travelodge, radisson, metropolitan, intercontinental, lodging, ibis, sofitel, novotel, doubletree \\
 vegetable, dish, herb, premade, delicacy, chopped, masa, lentil, tamale, canned, omlettes \\
 stay, staying, return, trip, vacation, honeymoon, visit, outing, revisit, celebrate, stayed \\
 great, awesome, amazing, excellent, fantastic, phenomenal, exceptional, nice, outstanding, wonderful, good \\
 room, cupboard, bedcover, drapery, bathroom, dusty, carpeting, laminate, ensuite, washroom, bedspread \\
utilized, day, excercise, recycled, tp, liner, borrowed, depot, vanished, restocked \\
 argued, blamed, demanded, called, registered, referred, transferred, contacted, claiming, denied, questioned \\
 neck, bruise, suspended, souvenir, tragedy, godiva, depot, blazing, peice, investigating \\
 \hline
\textbf{BeerAdvocate Dataset} \\
\hline
 shade, ruddy, transluscent, yellowish, translucent, hue, color, foggy, pours, branded \\
 recommend, like, disturb, conform, recomend, suggest, consider, imagine, liken, resemble\\
 longneck, swingtop, tapered, greeted, squat, ml, stubby, getting, stubbie, oz \\
 earthy, toffee, fruit, fruity, molasses, herbal, woody, floral, pineapple, raisin \\
 scarily, faster, quicker, obliged, inch, trace, warranted, mere, frighteningly, wouldve \\
 dubbel, quad, doppelbock, weizenbock, witbier, barleywine, tripel, dopplebock, brew, dipa \\
 lack, chestnut, purple, mahogany, nt, weakness, burgundy, coppery, fraction, garnet\\
 offering, product, beer, stuff, brew, gueuze, saison, porter, geuze, news \\
 mich, least, bass, pacifico, sumpin, inferior, grolsch, westy, yanjing, everclear \\
 foam, disk, head, heading, buts, edging, remains, lacing, froth, liquid \\
 starbucks, coffe, pepsi, expresso, kahlua, cofee, liqour, esspresso, cuban, grocery \\
 roasty, hop, chocolatey, hoppy, chocolaty, roasted, roast, robust, espresso, chocolately \\
 bodied, body, feel, barley, mouthfeel, tone, enjoyably, texture, moutfeel, mothfeel \\
 lasted, crema, flute, head, retentive, smell, fing, duper, compact, retention \\
 nt, rather, almost, never, kinda, fiddle, manner, easier, memphis, hard\\
 wafted, jumped, onset, proceeds, consisted, rife, consists, lotsa, citrisy, consisting \\
 hop, hoppy, assertive, inclined, tempted, doubtful, manageable, bearable, importantly, prominant\\
 great, good, decent, crazy, fair, killer, ridiculous, serious, insane, tremendous \\
 beer, brew, aperitif, nightcap, simply, intro, truly, ale, iipa, aipa \\
 wayne, harrisburg, massachusetts, arizona, jamaica, lincoln, kc, oklahoma, adult, odd \\
 \hline
\end{tabular}
\caption{Top words generated for each topic using ABAE method}
\end{table*}

\end{document}